\documentclass{article}

\usepackage{times}
\usepackage[utf8]{inputenc} 
\usepackage[T1]{fontenc}    
\usepackage{hyperref, etoolbox}       
\usepackage{url}            
\usepackage{booktabs}       
\usepackage{amsfonts}       
\usepackage{nicefrac}       
\usepackage{microtype}      
\usepackage{xspace}

\usepackage{amsmath, amssymb}
\usepackage{amsthm}
\usepackage{algorithm, algorithmic}
\usepackage{commath}
\usepackage{dsfont}
\usepackage[inline]{enumitem}
\usepackage{xcolor}

\usepackage{graphicx}
\usepackage[accepted]{icml2019}
\usepackage{overpic}
\usepackage{subcaption}
\usepackage{diagbox}
\usepackage[export]{adjustbox}

\newtheorem{definition}{Definition}
\newtheorem{theorem}{Theorem}
\newtheorem{lemma}{Lemma}
\newtheorem{claim}{Claim}
\newtheorem{proposition}{Proposition}
\newtheorem{corollary}{Corollary}
\newcommand{\hide}[1]{}

\def\R{\mathbb{R}}
\def\E{\mathbb{E}}
\def\P{\mathbb{P}}

\DeclareMathOperator{\lcb}{lcb}

\def\one{\mathds{1}}
\def\calX{\mathcal{X}}
\def\calF{\mathcal{F}}
\def\calB{\mathcal{B}}

\renewcommand{\mod}{ \ensuremath{\mathrm{mod}} }
\def\cb{c^b}
\def\hcb{\hat{c}}
\def\cs{c^s}
\def\Db{D^b}
\def\Ds{D^s}

\def\Ss{S}
\def\otau{{\tau'}}
\def\nb{n^b}
\def\ns{n^s}
\def\tns{\tilde{n}^s}
\def\scal{\alpha}
\def\bias{\Delta}
\def\risk{L}
\def\unrisk{\Gamma}
\DeclareMathOperator*{\argmin}{argmin}
\def\Str{S^{\text{tr}}}
\def\Sde{S^{\text{val}}}

\newcommand{\ourname}{\textsc{ARRoW-CB}\xspace}

\newcounter{qcounter}
 {\end{list}}

\newcommand{\regretb}{\mathbf{R}^b(\langle x_t, a_t\rangle_{t=1}^{\nb})}
\newcommand{\regrets}{\mathbf{R}^s(\langle x_t, a_t\rangle_{t=1}^{\nb})}
\newcommand{\nofrac}[2]{#1/#2}

\def\equationautorefname~#1\null{Equation~(#1)\null}
\renewcommand{\sectionautorefname}{\S\kern-0.2em}
\renewcommand{\subsectionautorefname}{\S\kern-0.2em}
\renewcommand{\subsubsectionautorefname}{\S\kern-0.2em}
\makeatletter 

\newcommand{\ALC@lineautorefname}{line} \makeatother
\newcommand{\err}{\texttt{err}}

\icmltitlerunning{Warm-starting Contextual Bandits}
\begin{document}
\twocolumn[
\icmltitle{Warm-starting Contextual Bandits:\\Robustly Combining Supervised and Bandit Feedback}
\begin{icmlauthorlist}
\icmlauthor{Chicheng Zhang}{msr}
\icmlauthor{Alekh Agarwal}{msr}
\icmlauthor{Hal Daum\'{e} III}{msr,umd}
\icmlauthor{John Langford}{msr}
\icmlauthor{Sahand N Negahban}{yale}
\end{icmlauthorlist}
\icmlaffiliation{msr}{Microsoft Research}
\icmlaffiliation{yale}{Yale University}
\icmlaffiliation{umd}{University of Maryland}

\icmlcorrespondingauthor{Chicheng Zhang}{Chicheng.Zhang@microsoft.com}

\vskip 0.3in
]

\printAffiliationsAndNotice{}

\setlength{\abovedisplayskip}{2pt}
\setlength{\belowdisplayskip}{2pt}
\setlength{\textfloatsep}{5pt}

\begin{abstract}
  We investigate the feasibility of learning from a mix of both fully-labeled supervised data and contextual bandit data.
  We specifically consider settings in which the underlying learning signal may be different between these two data sources.
  Theoretically, we state and prove no-regret algorithms for learning that is robust to misaligned cost distributions between the two sources.
  Empirically, we evaluate some of these algorithms on a large selection of datasets, showing that our approach is both feasible, and helpful in practice.
\end{abstract}

\section{Introduction}

In many real-world settings, a system must learn from multiple types of feedback;
we consider the specific setting of learning jointly from fully labeled ``supervised'' examples and from online feedback ``contextual bandit'' (abbrev. CB) examples.
For instance, in a system that chooses personalized content to display on a webpage,
an expert may be able to provide an initial set of fully labeled examples to get a system started.
After deployment, however, the system can only measure its performance (e.g., dwell time) on the content it displays and not other (counterfactual) options. In an automated translation system, professional translators can provide initial translations to seed a system,
but the system may be able to further improve its performance based on, e.g., user satisfaction measures \cite{riezler15downstream,daume17simhuman}.

In both these settings (content display and translation), we desire an approach that is able to use the fully supervised expert data to ``warm-start'' a system, which later learns from CB feedback~\cite{ACFS02, LZ07, LinUCB, DHKKLRZ11, AG13, AHKLLS14}.
Doing so has the added advantage of ensuring that such a system does not need to suffer too much error in an initial exploration phase, which may be necessary in user-facing systems or in error- or safety-critical settings~\cite{tewari2017ads}.
However, it is generally unreasonable to assume that the expert supervision and the CB feedback in such settings are perfectly aligned: the ``best'' decision according to an expert may not necessarily match a user's choice.
We need algorithms that operate well even in the case of unknown degrees of misalignment;
we introduce a hypothesis class-specific notion of cost similarity used in our analysis, but not our algorithms (\autoref{sec:setting}). We also highlight how simple strategies for combining the two sources without robustness to misalignment can perform significantly worse than learning from the ground truth source alone (\autoref{subsec:simple-fail}).

Furthermore, different applications can differ in terms of which source---supervised or CB---is considered ``ground truth''. For example, while the CB feedback from users is the better signal about their preferences in content personalization (\autoref{sec:banditgt}), the expert translations provide the ground truth in the translation setting for which user satisfaction is an imperfect proxy (\autoref{sec:supgt}).
We develop algorithms for \emph{both} settings, which effectively ``search'' for a good balance between fitting the CB feedback and supervised labels. In both cases, we provide regret bounds showing the value of the complementary data sources, dependent on their cost discrepancy and respective sample sizes. Importantly, our theory shows that our methods perform close to an oracle that knows the similarity of the two sources beforehand and uses it to optimally weight their examples, with a small additional penalty from searching for this weighting.

\begin{figure}[t!] 
    \centering
        \includegraphics[width=0.4\textwidth,trim={0 0 0 0},clip]{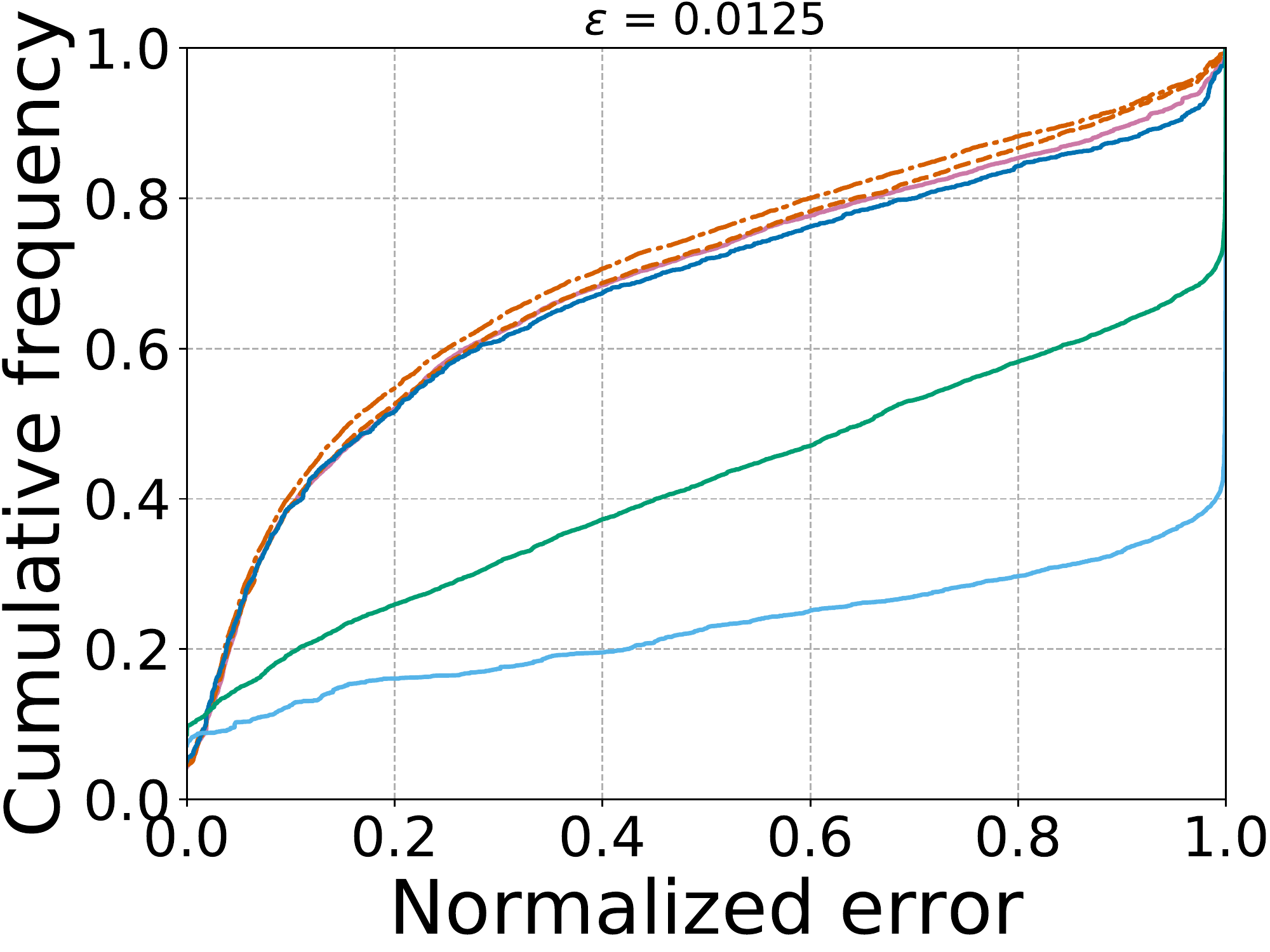}
        \label{fig:ds_6-noiseless}
        \begin{subfigure}[t]{\textwidth}
    \vspace{-1em}
    \hspace{-1.1cm} \includegraphics[width=0.6\textwidth,trim={0 0 0 0},clip]{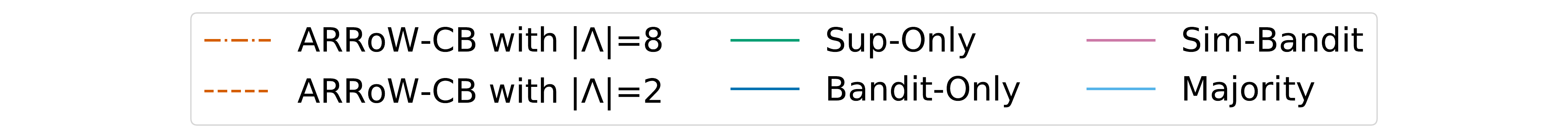}
  \end{subfigure}

    		\caption{Empirical CDFs of the performance of different methods across a number of datasets and experimental conditions (See \autoref{sec:exp} for descriptions of all algorithms, settings, and aggregation method). Our method \ourname has a parameter $|\Lambda|$, and we evaluate it with $|\Lambda|$ set to 8 and 2; \textsc{Sim-Bandit} is a baseline also leveraging the warm-start; \textsc{Sup-Only}, \textsc{Majority}\protect\footnotemark[1] and \textsc{Bandit-Only} learn using only the supervised and CB sources respectively.
        All CB methods use the $\epsilon$-greedy strategy with $\epsilon = 0.0125$.}
		\label{fig:agg-0.0125}
\end{figure}

Empirically, we perform experiments based on fully-labeled examples from which CB feedback is simulated. We focus on the setting when CB data is ground truth and the supervised warm-start might have differing levels of bias.
In an experimental study over hundreds of datasets (\autoref{sec:exp}) we demonstrate the efficacy of our algorithm. As a snapshot, Figure~\ref{fig:agg-0.0125} shows the empirical cumulative distribution functions (CDF) of algorithms across a number of experimental conditions, where each $(x,y)$ value on the curve indicates that there is a $y$ fraction of experimental conditions where the normalized error\footnote{See~\autoref{sec:exp} for a formal definition of normalized error.} of a method is below $x$. The plot aggregates across settings where the CB and supervised signals are perfectly aligned as well as where they are not. Overall, our main algorithm, namely \ourname with $|\Lambda|=8$, outperforms all baselines in this aggregated summary, in particular beating the two algorithms (\ourname with $|\Lambda|=2$ and \textsc{Sim-Bandit})
that leverage both CB and supervised sources. More detailed results are presented in~\autoref{sec:exp}.


\footnotetext[1]{\textsc{Sup-Only} and \textsc{Majority} do not explore or update on CB examples and we plot the average costs of their policies over all CB examples.}
\setcounter{footnote}{2}

\paragraph{Relation to prior work.} A theoretical study of domain adaptation~\cite{BBCKPV10, MMR09} and learning
from multiple sources~\cite{CKW08} are the closest
prior works.  In these works, all data sources provide the same
supervised feedback rather than the supervised/CB modality we investigate here, with the two sources having very different information per sample.
Another related line of work is on ``safe'' CB learning~\cite{KGAR17,SDK17} to maintain performance better than a baseline policy at all times, somewhat related to our supervised ground truth setting. However, they do not study the distributional mismatch concerns central to our work.

Finally, there is a substantial literature on active learning from different
data sources~\cite{DC08, UBS12, YRFD11, MCR14, ZC15, YCJ18}, combining multiple labeling oracles of varying quality. The CB setting studied has important differences from active learning and the techniques do not carry over directly.


\section{Notation and Problem Specification} \label{sec:setting}
We begin with some notation. For an event $A$, $I(A) = 1$ if $A$
is true, and $0$ otherwise. Denote by $[K]$ the set $\cbr{1,2,\ldots,K}$.
We use $\one_K$ to denote the all $1$'s vector in $\R^K$ and $\Delta^{K-1}$ for the $K$ dimensional probability simplex.

In this paper, we study the problem of cost-sensitive interactive learning from multiple data
sources. Specifically, we consider distributions over cost-sensitive examples $(x,c)$,
where $x \in \calX$ is a context and $c$ is a cost vector in $[0,1]^K$; $K$ being the number of actions (or ``classes'').
There are two distributions $\Ds$ (supervised) and $\Db$ (CB), which have identical marginals over the context $x$, but different conditional distributions over cost vectors given $x$.
We use the notation $\cb$ (resp. $\cs$) to denote the cost vector $c$ drawn from $\Db$ (resp. $\Ds$) to avoid writing $\Db$ and $\Ds$ as subscripts in expectations.
The interaction between the learner and the environment is described as follows:
\paragraph{Warm-start:} The learner receives $\Ss$, a dataset of $\ns$ fully supervised examples drawn i.i.d. from $\Ds$.
\vspace{-0.3cm}
\paragraph{Interaction:} For $t = 1,2,\ldots,\nb$, the environment draws $(x_t, \cb_t) \sim \Db$ and reveals $x_t$ to the learner, based on which the learner chooses a (possibly random) action $a_t \in [K]$ and observes $\cb_t(a_t)$, but not the cost of any other action.


In this paper, we focus on two learning settings: {\em CB ground truth setting} and {\em supervised ground truth setting}. In the CB ground truth setting (resp. supervised ground truth setting), the goal of the learner is to optimize the costs drawn from distribution $\Db$ (resp. $\Ds$).

To help make decisions, the learner is given a finite policy class $\Pi$ that contains policies $\pi: \calX \rightarrow [K]$\footnote{More generally, $\pi~:~\calX\rightarrow \Delta^{K-1}$ and $\pi(x)$ is the distribution over actions given $x$.}.
The performance of the algorithm is measured by its \emph{regret} to the retrospective-best policy in $\Pi$.
We consider two notions of regret over the sequence $\langle x_t\rangle_{t=1}^{\nb}$, based on whether we consider the CB costs ($\cb$) or the supervised costs ($\cs$) as the ground truth:
\begin{align}
  \text{CB: } &\regretb =
  \textstyle\sum_{t=1}^{\nb} \E\big[\cb(a_t)~\big|~x_t\big] - \nonumber\\&\qquad \textstyle \min_{\pi \in \Pi} \sum_{t=1}^{\nb} \E\big[\cb(\pi(x_t))~\big|~x_t\big], \label{eq:regretbandit}
\end{align}
\begin{align}
  \text{supervised: } &\regrets =
  \textstyle\sum_{t=1}^{\nb} \E\big[\cs(a_t)~\big|~x_t\big]\,- \nonumber\\&\qquad \textstyle \min_{\pi \in \Pi} \sum_{t=1}^{\nb} \E\big[\cs(\pi(x_t))~\big|~x_t\big].
  \label{eq:regretsup}
\end{align}


In the content recommendation example (CB ground truth), $x_t$ encodes a user profile and the system predicts which articles ($a_t$) to display.
Here, $\cb$ can be the negative dwell-time of users and $\cs$ is the annotation of editors, which can have disagreements with $\cb$. The learner aims to optimize the dwell time over all displayed articles.

In the translation example (supervised ground truth), $x_t$ encodes the text to be translated and $a_t$ encodes its translation.
Here, the learner aims to minimize errors against the expert translation ($\cs$) on $x_t$'s, \emph{despite the fact} that the system never sees these costs in its interaction phase. Note that the learner only observes the user feedback costs ($\cb$) in this interaction phase, which are imperfect proxies for $\cs$, and the only direct observations of $\cs$ are on the warm-start examples. Nevertheless, we seek to optimize the accuracy of our translations given to the users, and hence regret is still measured over the interaction phase.




The utility of non ground truth examples are different in the two learning settings.
In the CB ground truth setting, relying on the CB examples alone is sufficient to ensure vanishing regret asymptotically. The supervised warm-start primarily helps with a smaller regret in the initial phases of learning. On the other hand, in the supervised ground truth setting, the CB examples can have an asymptotically meaningful effect on the regret: for instance, if $\Ds = \Db$, then utilizing CB examples can lead to a vanishing regret, whereas using supervised examples alone cannot.





We can leverage examples from a different source only when the cost structures are at least somewhat related.
Therefore, we introduce a measure of similarity of two distributions over cost-sensitive examples.

\begin{definition}
  $D_2$ is said to be $(\scal, \bias)$-similar to $D_1$ with respect to $\Pi$, if
  for any policy $\pi$,
$\E_{D_2} c(\pi(x)) - \E_{D_2} c(\pi^*(x))  \geq \scal\left(\E_{D_1} c(\pi(x))  - \E_{D_1} c(\pi^*(x))\right) - \bias$, where $\pi^* = \argmin_{\pi \in \Pi} \E_{D_1} c(\pi(x))$.
\label{def:similar}
\end{definition}

If we have a larger $\scal$ and smaller $\bias$, examples from $D_2$ are more useful for learning under $D_1$. Prior similarity notions, such as in~\citet{BBCKPV10}, roughly assume a bound on $\max_{\pi \in \Pi} |\E_{D_1}[c(\pi(x))] - \E_{D_2}[c(\pi(x))]|$. The one-sided bound in our definition (instead of absolute value bound) on regret and an additional scaling factor $\alpha$ yield additional flexibility.
Note that~\autoref{def:similar} is only used in our analysis; our algorithms do not require knowledge of $\scal$ and $\bias$. We give
a more general condition which implies $(\alpha,\Delta)$-similarity, along with several examples in~\autoref{sec:similar}.






Finally, we define some additional notation. 
In the $t$-th interaction round, our algorithms compute $\hcb_t$, an estimate of the unobserved vector $\cb_t$.
We use $\E_S$ to denote sample averages on $S$ and abbreviate $\E_{S_t}$ by $\E_t$ where $S_t = \{(x_\tau,\hcb_\tau)\}_{\tau=1}^t$ is the log of the CB examples up to time $t$.

\subsection{Failure of Simple Strategies}
\label{subsec:simple-fail}
The settings we have described so far might appear deceptively simple. It should be easy to include some additional supervised examples, which contain more feedback, into a CB algorithm. We now illustrate the difficulty of this task when the two distributions $\Ds$ and $\Db$ are misaligned.

Consider the special case of $2$-armed bandits (CB with a dummy context), where the CB source is the ground truth. $\Ds$ and $\Db$ are deterministic with costs $(0.5, 0.5+\tfrac \Delta 2)$ and $(0.5,0.5-\tfrac{\Delta}{2})$ for the two arms respectively, so that they are $(1,\Delta)$-similar. Suppose we see $\ns = \Omega(1/\Delta^3)$ examples in warmstart, and use them to initialize the means and confidence intervals on each arm to run the UCB algorithm~\cite{auer2002finite}.
Proposition~\ref{prop:ucbfail} in Appendix~\ref{app:ucb} show
that the optimal arm according to $\Db$, which is arm 2, is never played for the first $O(\exp(1/\Delta))$ rounds,
incurring regret $\Omega(\Delta\exp(1/\Delta))$. So for any $\Delta < 0.5$, the regret is strictly larger than that of a UCB algorithm which ignores the warm-start and incurs at most $\tilde{O}(1/\Delta)$ regret. On the other hand, if $\Db = \Ds$, then the UCB strategy described above incurs no regret.

What we observe here is a failure in competing simultaneously with two baselines: naively warmstarting by weighting examples from the two sources equally, or just ignoring the supervised source entirely. We will next describe an algorithm to compete not just with these two, but many possible weightings of the two sources. This extreme failure case shows that an arbitrary low-regret CB algorithm cannot handle biased warm-start data without extra care. Using additional randomization can help, but is not adequate by itself as we will see in our theory and experiments.





\section{Contextual Bandit Ground Truth Setting}
\label{sec:banditgt}
In this section, we study the setting where $\Db$, the distribution over CB examples, is considered the ground truth, as in the content recommendation example. Recall that in this setting, one could ignore the supervised warm-start examples and still achieve vanishing regret; the main goal here is to show that using the warm start data can help further reduce the regret, especially in early stages of learning.

\subsection{Algorithm}

\paragraph{Intuition of our approach.} The key challenge in designing an algorithm for the CB ground truth setting is understanding how to effectively combine two data sources which might have unknown differences in their distributions. For the simpler supervised learning setting, Proposition~\ref{prop:minimax} in Appendix~\ref{app:minimax} shows that it suffices to always use one of the two sources depending on the bias and relative number of examples. This has two caveats though: the bias is not known in practice, and completely ignoring one data source is obviously wasteful when the two sources are identical. We choose to instead consider cost minimization on a dataset where the two sources are combined with different weights, and seek to learn these weights adaptively.

With these insights, we return to the actual problem setting of warm-starting a CB learner with supervised examples. Our algorithm for this setting is presented in~\autoref{alg:twosources}. The main idea is to minimize the empirical risk on a weighted dataset containing examples from the two sources. Our algorithm picks the mixture weighting by online model selection over a set of weighting parameters $\Lambda$, where we use the ground truth CB data at each time step to evaluate which $\lambda \in \Lambda$ has the best performance so far.
For each $\lambda \in \Lambda$, we estimate a $\pi^\lambda \in \Pi$ as the empirical risk minimizer (ERM) for the $\lambda$-mixture between CB and supervised examples.
We focus on the simplest $\epsilon$-greedy algorithm for CBs, leaving similar modifications in more advanced CB algorithms for future work.

\begin{algorithm}[!t]
\caption{Adaptive Reweighting for Robustly Warmstarting Contextual Bandits (\ourname)}\label{alg:twosources}
\begin{algorithmic}[1]
  \REQUIRE Supervised dataset $\Ss$ from $\Ds$ of size $\ns$, number of interaction rounds $\nb$,
  exploration probability $\epsilon$, weighted combination parameters $\Lambda$, policy class $\Pi$.
  \FOR{$t=1,2,\ldots,\nb$}

	\STATE Observe instance $x_t$ from $\Db$.



  \STATE Define $p_t := \frac{1 - \epsilon}{t-1} \sum_{\tau=1}^{t-1} \pi_\tau^{\lambda_t}(x_t) + \frac\epsilon K \one_K$ for $t \geq 2$ and $p_t := \frac 1 K \one_K$ for $t=1$. \label{step:uar-history}

	\STATE Predict $a_t \sim p_t$, and receive feedback $\cb_t(a_t)$.

  \STATE Define the inverse propensity score (IPS) cost vector $\hcb_t(a) := \frac{\cb_t(a_t)}{p_{t,a_t}} I(a=a_t)$, for $a \in [K]$.

  \STATE For every $\lambda \in \Lambda$, train $\pi_t^{\lambda}$ by minimizing over $\pi \in \Pi$:
    \begin{equation}
    \lambda \sum_{\tau=1}^{t-1} \hcb_\tau(\pi(x_\tau)) + (1-\lambda) \sum_{(x,\cs) \in \Ss} \cs(\pi(x)).
    \label{eqn:pi-t-lambda}
  \end{equation}
  \STATE Set $\lambda_{t+1} \gets \argmin_{\lambda \in \Lambda} \sum_{\tau=1}^{t} \hcb_\tau(\pi_\tau^{\lambda}(x_\tau)). $
  \label{line:pv}


  \ENDFOR
\end{algorithmic}
\end{algorithm}

So long as $\{0,1\}\subseteq \Lambda$, \autoref{alg:twosources} allows for relying on one source alone, while using a larger set of $\Lambda$ significantly improves its empirical performance (see \autoref{sec:exp}).\footnote{If we approximate the computation of the best policy in Step 6 using an online oracle as in prior works~\cite{AHKLLS14, LZ07},
then the entire algorithm can be implemented in a streaming fashion since~\autoref{line:pv} for selecting the best $\lambda$ also uses an online estimate a la~\citet{blum1999beating} for each $\lambda$ as opposed to a holdout estimate for the current policy $\pi_t^\lambda$.}

We need some additional notation to present our regret bound.
We define $V_t(\lambda)$ that governs the deviation of $\lambda$-weighted empirical costs for all policies in $\Pi$ as
$V_t(\lambda) :=2\sqrt{( \tfrac{\lambda^2K t}{\epsilon} + (1-\lambda)^2 \ns) \ln\tfrac{8\nb|\Pi|}{\delta} } + (\tfrac{\lambda K}{\epsilon} + (1-\lambda)) \ln \tfrac{8\nb|\Pi|}{\delta}$,
and $G_t$ that bounds the excess cost of the ERM solution using weighted combination parameter $\lambda$ as\\  $$G_t(\lambda, \scal, \bias) := \frac{(1-\lambda)\ns \bias + 2V_t(\lambda)}{\lambda t + (1-\lambda) \ns \scal}.$$
%

We prove the following theorem in Appendix~\ref{app:bandit-gt}.
\begin{theorem}
\setlength{\belowdisplayskip}{0pt} \setlength{\belowdisplayshortskip}{0pt}
\setlength{\abovedisplayskip}{0pt} \setlength{\abovedisplayshortskip}{0pt}
\label{thm:bandit-gt}
Suppose $\Ds$ is $(\scal, \bias)$-similar to $\Db$. Then for any $\delta < 1/e$, with probability $1-\delta$, the average CB regret of~\autoref{alg:twosources} can be bounded as:
\begin{align}
  \frac 1 {\nb} &\regretb
\leq\textstyle
\epsilon + 3 \sqrt{\frac{\ln\frac{8 \nb |\Pi|}{\delta}}{\nb}} + 32 \sqrt{\frac{K \ln\frac{8\nb|\Lambda|}\delta}{\nb \epsilon}} +\nonumber\\&\qquad\qquad\qquad
\textstyle\min_{\lambda \in \Lambda} \frac{\ln (e\nb)}{\nb} \sum_{t=1}^{\nb} G_t(\lambda, \scal, \bias) \label{eq:ourbound}
\end{align}
\end{theorem}

%


The bound~\eqref{eq:ourbound} consists of many intuitive terms. The first $\epsilon$ term comes from uniform exploration; the second term is from the deviation of costs under $\Db$.
The next term is the average regret incurred in performing model selection for $\lambda$; in our experiments $|\Lambda|=8$ so that it can be thought of as $\widetilde{O}(\sqrt{K/(\nb\epsilon)})$.
The final term involving a minimum over $\lambda$'s is effectively finding the weighted combination which minimizes a bias-variance tradeoff in combining the two sources. Here the bias is controlled by $\Delta$ and in place of variance we use $V_t(\lambda)$ for high-probability results. Contrasting with learning with CB examples alone, we replace a $\sqrt{\tfrac{K\ln(|\Pi|/\delta)}{\nb \epsilon}}$ term with the middle term independent of $\ln |\Pi|$ and the average of $G_t$'s which can be much smaller in favorable cases as we discuss below.

\paragraph{Identical distributions:} A very friendly setting has $\Ds = \Db$, corresponding to $(1,0)$-similarity. 
Since the theorem holds with a minimum over all $\lambda$'s in the set $\Lambda$, we can pick specific values $\lambda_0$ of our choice. One choice of $\lambda_0$ motivated from prior work~\citep{BBCKPV10} is to pick it such that $\lambda_0/(1-\lambda_0) = \epsilon/K$ to equalize the variance of the two sources, meaning each supervised example is worth $K/\epsilon$ CB examples.
This setting of $\lambda_0 = \frac{\epsilon}{K + \epsilon}$ yields $G_t\left(\lambda_0, 1, 0\right) = O\bigg(\textstyle\sqrt{\frac{K\ln\frac{\nb|\Pi|}{\delta}}{\epsilon t + K\ns}} + \frac{K\ln\frac{\nb|\Pi|}{\delta}}{\epsilon t + K\ns}\bigg)$. That is, after $t$ CB samples, the effective sample size is $\ns + Kt/\epsilon$.


\paragraph{Comparison with no warmstart:} Whenever $1 \in \Lambda$, the minimum over $\lambda \in \Lambda$ in~\autoref{thm:bandit-gt} can be bounded by its value at $\lambda = 1$, which corresponds to ignoring the warmstart examples and using bandit examples alone. For this special case, we have the following corollary.

\begin{corollary}
\label{cor:bandit-gt-01}
Under conditions of~\autoref{thm:bandit-gt}, suppose that $1 \in \Lambda$. Then for any $\delta < 1/e$, with probability $1-\delta$,
\begin{align*}
   \frac 1 {\nb} \regretb
\leq\textstyle
\epsilon + O\bigg(\sqrt{\frac{\ln\frac{\nb |\Pi|\,|\Lambda|}{\delta}}{\nb\epsilon}}\bigg).
\end{align*}
\end{corollary}

The corollary follows from using the value of $G_t(1,\alpha,\Delta)$ along with some algebra, and shows that the regret of \ourname is never worse than using the bandit source alone, up to a term scaling as $\ln|\Lambda|$. In particular, the usual choice of $\epsilon = O((\nb)^{-2/3})$ implies a $O((\nb)^{2/3})$ regret bound. Since a small value of $|\Lambda|$ suffices in our experiments, this is a negligible cost for robustness to arbitrary bias in the warmstart examples. Similarly comparing to $\lambda = 0$ lets us obtain a comparison against using the warmstart alone up to a model selection penalty, when $0 \in \Lambda$. The minimization over a richer set of $\lambda$ leaves room for further improvements as shown in the case of $\Ds = \Db$ above (which used a different setting of $\lambda_0$). Further improvements are also possible in the algorithm by using different $\lambda$ values after reach round, which is not captured in the theory here.

\section{Supervised Ground Truth Setting}
\label{sec:supgt}
In \autoref{sec:banditgt}, we developed an algorithm and proved regret bounds for combining supervised and CB feedback, in the case where the CB cost is considered the ground truth.
In this section, we consider the reverse setting where the supervised source constitutes the ground truth, recalling the motivating example in an automated translation setting from the introduction. Here, we wish to leverage the CB examples for learning the best policy relative to the distribution $\Ds$.

Note that this setting is qualitatively different, since we only have a fixed number $\ns$ of ground-truth examples while the number of CB examples grows over time. If we assign relative weights to individual supervised and CB examples as in~\autoref{alg:twosources}, the CB examples will eventually outweigh the supervised ones for any $\lambda > 0$, which is not desirable when the supervised source is the ground truth. In~\autoref{alg:twosources-sup-gt}, we address this problem by first computing the average costs of every policy on the supervised and CB examples separately, and then choosing a policy that minimizes a weighted combination of these averages.
As a consequence, the relative weight of each CB example diminishes as their number grows, with the overall bias incurred from the CB source staying bounded.

Another difference between~\autoref{alg:twosources-sup-gt} and~\autoref{alg:twosources} is that, as opposed to using the CB examples collected online, we use subsets of warm start examples to guide the selection of weighted combination parameter $\lambda$. To this end, we introduce an epoch structure in the algorithm.
In particular, at each epoch $e$, $\lambda_e$ and $\pi_e^\lambda$'s are updated exactly once, where a separate validation set is used to pick $\lambda$. In addition, we play with uniform randomization around the most recent policy as opposed to a running average of all policies trained so far, an outcome of using a separate validation set (line 12 of~\autoref{alg:twosources-sup-gt}) instead of progressive validation (\autoref{line:pv} of~\autoref{alg:twosources}).
Since the exploration policy at the next epoch depends on the previous validation set, we must use a ``fresh'' validation set at each epoch. 
Avoiding this splitting is an interesting question for future work.


\begin{algorithm}[t]
\caption{Combining contextual bandit and supervised data when supervised source is the ground truth}
\label{alg:twosources-sup-gt}
\begin{algorithmic}[1]
  \REQUIRE Supervised dataset $S$ from $\Ds$ of size $\ns$, number of interaction rounds $\nb$, exploration probability $\epsilon$, weighted combination parameters $\Lambda$, policy set $\Pi$.

  \STATE Let $E = \lceil \log \nb \rceil$ be the number of epochs.
  \STATE Define $t_e = \min(2^e, \nb)$ for $e \geq 1$, and $t_0 = 0$.
	\STATE Partition $S$ to $E+1$ equally sized sets $\Str, \Sde_1, \ldots, \Sde_E$.


	\FOR{$e=1,2,\ldots,E$}
  \FOR{$t=t_{e-1}+1,t_{e-1}+2,\ldots,t_e$}

	\STATE Observe instance $x_t$ from $\Db$.

  \STATE Define $p_t := (1 - \epsilon)\pi_{e-1}^{\lambda_{e-1}}(x_t) + \frac\epsilon K \one_K$ for $e \geq 2$, and $p_t := \frac 1 K \one_K$ for $e = 1$.

  \STATE Predict $a_t \sim p_t$ and receive feedback $\cb_t(a_t)$.

  \STATE Define the IPS cost vector $\hat{c}_t(a) := \frac{\cb_t(a_t)}{p_{t,a_t}} I(a=a_t)$, for $a \in [K]$.

	\ENDFOR

	\STATE For each $\lambda \in \Lambda$, train
	$ \pi_e^{\lambda}$ as:\\\quad  $\arg\min_{\pi \in \Pi}
	\lambda \E_{t_e} \hat{c}(\pi(x)) + (1-\lambda) \E_{\Str} \cs(\pi(x)). $

	\STATE Set $\lambda_e \gets \arg\min_{\lambda \in \Lambda} \E_{\Sde_e} \cs(\pi_e^\lambda(x)) $.
  \label{line:sv}


	\ENDFOR
\end{algorithmic}
\end{algorithm}

For the main result, we need the following notation for the deviation of $\lambda$-weighted empirical costs, where
$E = \lceil \log \nb \rceil$ is the total number of epochs:
\begin{align*}
W_t(\lambda) &:= \textstyle 2 \sqrt{ \left(\frac{\lambda^2 K }{t \epsilon} + \frac{(1-\lambda)^2(E+1)}{\ns}  \right) \ln\frac{8E |\Pi|}{\delta} } +\\&\qquad \qquad\textstyle\left(\frac{\lambda K}{t \epsilon} + \frac{(1-\lambda)(E+1)}{\ns}\right) \ln\frac{8E |\Pi|}{\delta}.
\end{align*}
\begin{theorem}
\label{thm:supervised-gt}
Suppose that $\Db$ is $(\scal, \bias)$-similar to $\Ds$. Then for any $\delta < 1/e$, with probability
$1-\delta$, the average supervised regret of~\autoref{alg:twosources-sup-gt} can be bounded as:
\begin{align}
\textstyle \frac 1 {\nb} \regrets
&\leq
\epsilon + 3\sqrt{\frac{\ln\tfrac{8|\Pi|}{\delta}}{\nb}} + \sqrt{\frac{ 2(E+1) \ln\tfrac{8E|\Lambda|}{\delta}}{\ns}}
\nonumber\\ & + \min_{\lambda \in \Lambda} \frac{2}{\nb} \sum_{t=1}^{\nb} \frac{\lambda \bias + 2W_t(\lambda)}{(1-\lambda) + \lambda \scal}.
\label{eqn:sup-gt-bound}
\end{align}
\end{theorem}

The first term is the cost of exploration, while the second is the gap between the conditional and unconditional expectations over costs in defining the regret. The third term captures the complexity of model selection while the final is the performance upper bound for the best $\lambda$ in our weighted combination set $\Lambda$. As before, this significantly improves upon the $O(\sqrt{\frac{\ln|\Pi|/\delta}\ns})$ bound from using supervised examples alone whenever the two sources have sufficient similarity. The proof can be found in \autoref{app:supervised-gt}.

\paragraph{Identical distributions:} When $\Ds = \Db$, which implies that $\Ds$ is $(1,0)$-similar to $\Db$, a single choice of $\lambda = \frac{\nb \epsilon}{\ns K + \nb \epsilon}$ will ensure that the last term in~\autoref{eqn:sup-gt-bound} is at most
$\tilde{O}\bigg(\textstyle \sqrt{ \frac{ K \ln\frac{8E |\Pi|}{\delta} }{K \ns + \nb \epsilon} } + \frac{ K \ln\frac{8E |\Pi|}{\delta} }{K \ns + \nb \epsilon} \bigg)$
(See~\autoref{prop:one-lambda-sup-gt} in~\autoref{sec:one-lambda-sup-gt}). That is,
after $\nb$ CB samples, the effective sample size is at most $\ns + \nb \epsilon / K$.


\section{Experiments} \label{sec:exp}

Experimentally, we focus on the question of learning with the CB costs as the ground truth (\autoref{sec:banditgt}).
Our experiments seek to address the following questions:
\begin{enumerate*}[label={\alph*)},font={\bfseries}]
\item How much benefit does a small amount of supervised warm-start provide?
\item How much benefit does the bandit feedback provide?
\item How robust is our algorithm under a realistic mismatch in cost structures?
\item How robust is our algorithm under adversarial cost structures (the ``safety'' question)?
\end{enumerate*}

We consider the following set of approaches:\\
\textbf{\textsc{Bandit-Only}}: a baseline that only uses CB examples.\\
\textbf{\textsc{Majority}}: always predicts $a \in \argmin_{a \in [K]} \E_{(x,c)\sim\Db}[c(a)]$ independent of the context, without exploration.\\
\textbf{\textsc{Sup-Only}}: a baseline that uses the best policy on supervised examples, without exploration.\\
\textbf{\textsc{Sim-Bandit}}: a baseline that runs the CB algorithm on warm-start examples as well, providing cost for the chosen action only (from the supervised set) and then continues on the remaining CB examples.\\
\textbf{\textsc{\ourname} with $\Lambda= \{ 0, \frac 1 8 \zeta, \frac 1 4 \zeta, \frac 1 2 \zeta, \zeta, \frac 1 2 + \frac 1 2 \zeta, \frac 3 4 + \frac 1 4 \zeta, 1 \}$}
(abbrev. \textbf{\textsc{\ourname} with $|\Lambda| = 8$}),
where $\zeta = \epsilon / (K+\epsilon)$; this is chosen because $\zeta$ is an approximate minimizer of $G_t(\lambda, 1, 0)$, and the $|\Lambda|$ used aims to ensure that
$\min_{\lambda \in \Lambda} G_t(\lambda, \scal, \bias)$ is close to $\min_{\lambda \in [0,1]} G_t(\lambda, \scal, \bias)$ (see Prop.~\ref{prop:lambda-0-1}).
For computational considerations, we use the last policy $\pi_t^{\lambda_t}$ rather than the averaged policy $\frac1{t-1}\sum_{\tau=1}^{t-1} \pi_\tau^{\lambda_t}$ in line~\ref{step:uar-history} of~\autoref{alg:twosources}.\\
\textbf{\ourname with $\Lambda = \cbr{0,1}$}(abbrev. \textbf{\textsc{\ourname} with $|\Lambda| = 2$}): as argued in Proposition~\ref{prop:lambda-0-1}, choosing $\lambda$ in $\cbr{0,1}$ also approximately minimizes $G_t(\lambda, \scal, \bias)$.


In subsequent discussions, if not explicitly mentioned, \ourname refers to \ourname with $|\Lambda| = 8$.

All the algorithms (other than \textsc{Sup-Only} and \textsc{Majority}, which do not explore) use $\epsilon$-greedy exploration, with most of the results presented using $\epsilon = 0.0125$. We additionally present the results for $\epsilon = 0.1$ and $\epsilon = 0.0625$ in Appendix~\ref{apx:moreresults}. In general, the increased uniform exploration for larger $\epsilon$ leads to some performance penalty in the CB algorithms relative to \textsc{Sup-Only}, when the bias is small. However, the added exploration gives robustness to large bias as it is readily detected in more adversarial noise settings.

\paragraph{Datasets.} We compare these approaches on 524 binary and  multiclass classification datasets from~\citet{BAL18}, which in turn are from \url{openml.org}.
For each dataset, we use the multiclass label in the dataset to generate cost vectors $\cb$ and $\cs$ respectively.
That is, given an example $(x,y) \in \calX \times [K]$, $\cb(a) = I(a \neq y)$.
We vary the number of warm-start examples and CB examples as follows:
for a dataset of size $n$, we vary the number of warm-start examples $\ns$ in $\cbr{0.005n, 0.01n, 0.02n, 0.04n}$, and the number of CB examples $\nb$ in $\cbr{0.92n, 0.46n, 0.23n, 0.115n}$.
Define the \emph{warm-start ratio} as the ratio $\nb/\ns$. We group different settings of (dataset, $\ns$, $\nb$) by $\nb/\ns$, so that a separate plot is generated for each ratio in $R = \cbr{2.875, 5.75, 11.5, 23, 46, 92, 184}$.
We filter out the settings where $\ns$ is below 100.

\begin{figure*}[t!] 
  \centering

  \captionsetup[subfigure]{labelformat=empty}
  \begin{subfigure}[t]{2mm}
    \centering
    \includegraphics[width=3mm,trim={0 43 540 0},clip]{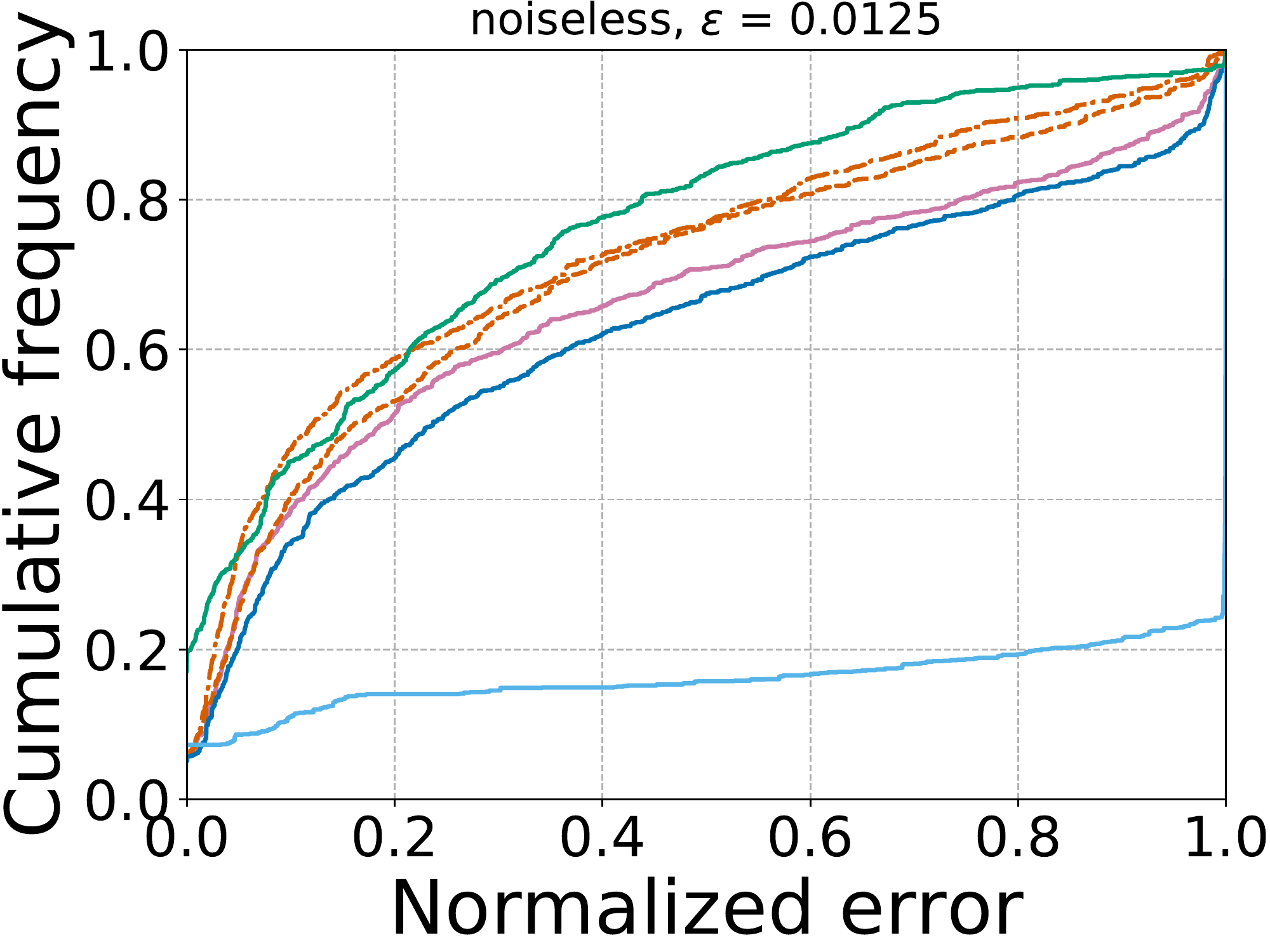}
  \end{subfigure}
  \begin{subfigure}[t]{0.32\textwidth}
    \centering
    \begin{Overpic}{\includegraphics[width=\textwidth,trim={38 43 0 0},clip]{figs/cdfs/agg_ratio_eq/corruption={st,ctws=1,cpws=0.0,cti=1,cpi=0.0},explore_method={expl,eps=0.0125},/cdf_nolegend.pdf}} 
    \end{Overpic}
    \caption{}
    \label{fig:0.0125-noiseless}
  \end{subfigure}%
  \begin{subfigure}[t]{0.32\textwidth}
    \centering
    \begin{Overpic}{\includegraphics[width=\textwidth,trim={38 43 0 0},clip]{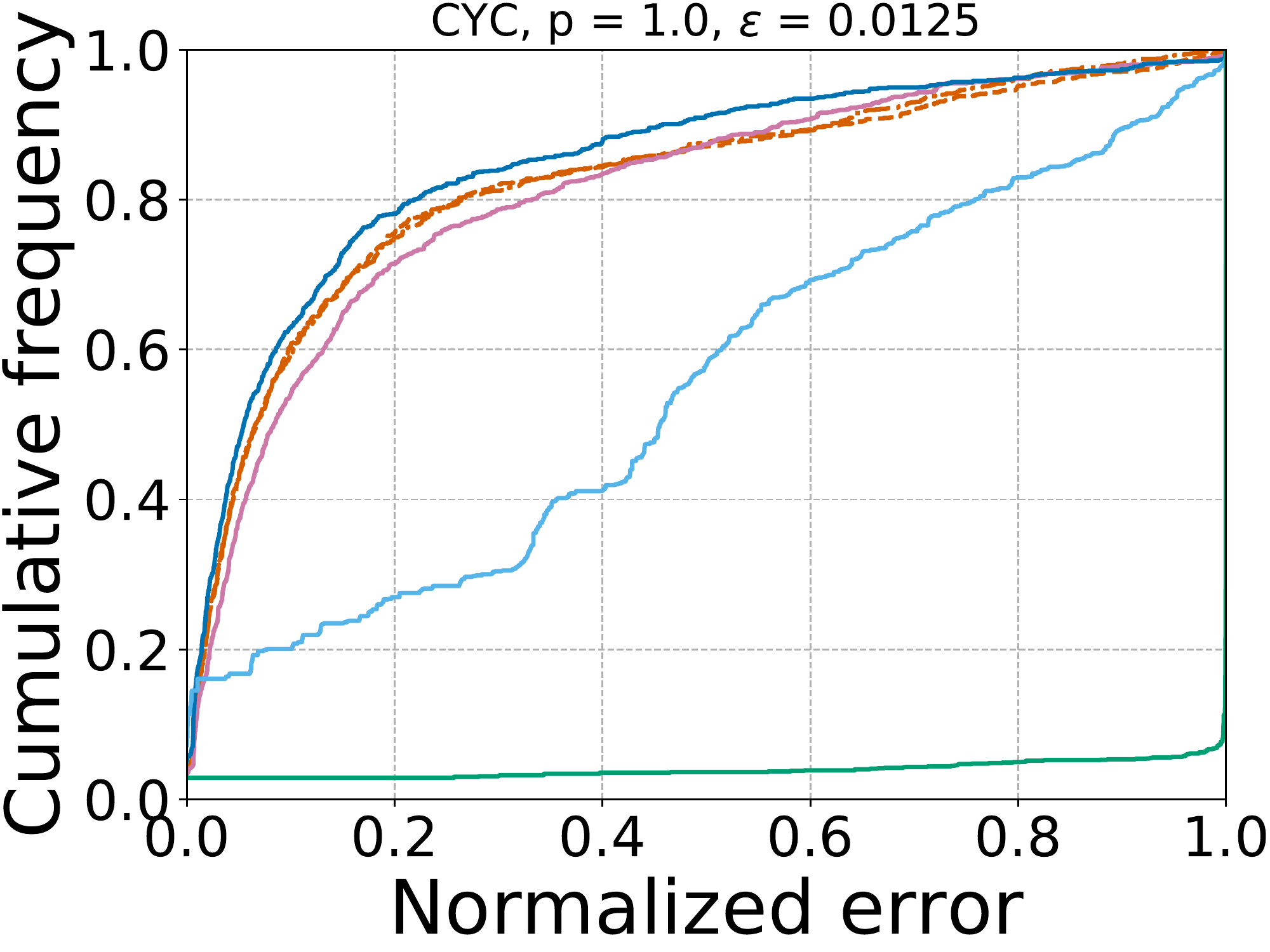}} 
    \end{Overpic}
    \caption{}
    \label{fig:0.0125-2-1.0}
  \end{subfigure}
  \begin{subfigure}[t]{0.32\textwidth}
    \centering
    \begin{Overpic}{\includegraphics[width=\textwidth,trim={38 43 0 0},clip]{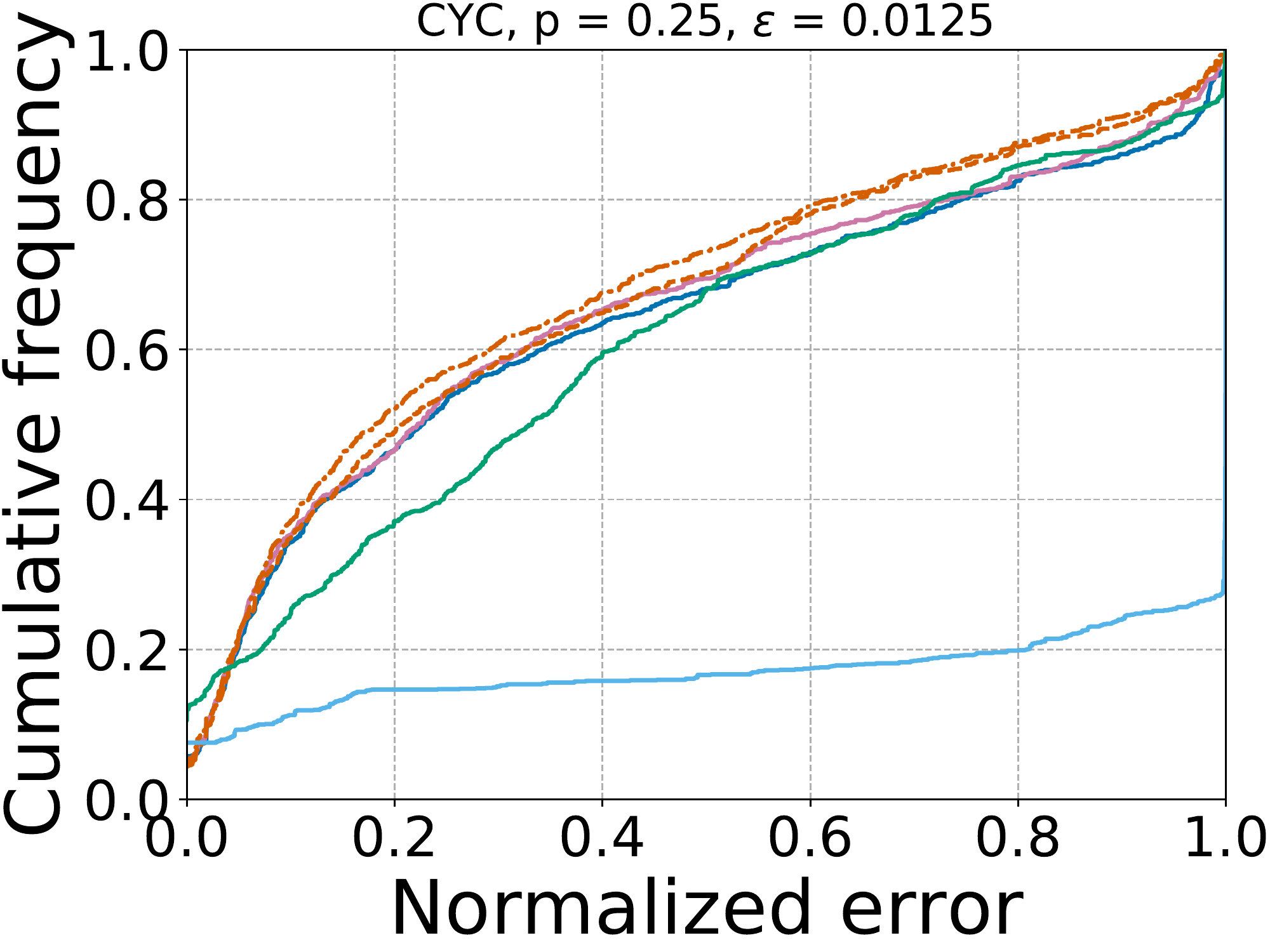}} 
    \end{Overpic}
    \caption{}
    \label{fig:0.0125-2-0.25}
  \end{subfigure}


  %
  \smallskip

  \begin{subfigure}[t]{2mm}
    \centering
    \includegraphics[width=3mm,trim={0 0 540 0},clip]{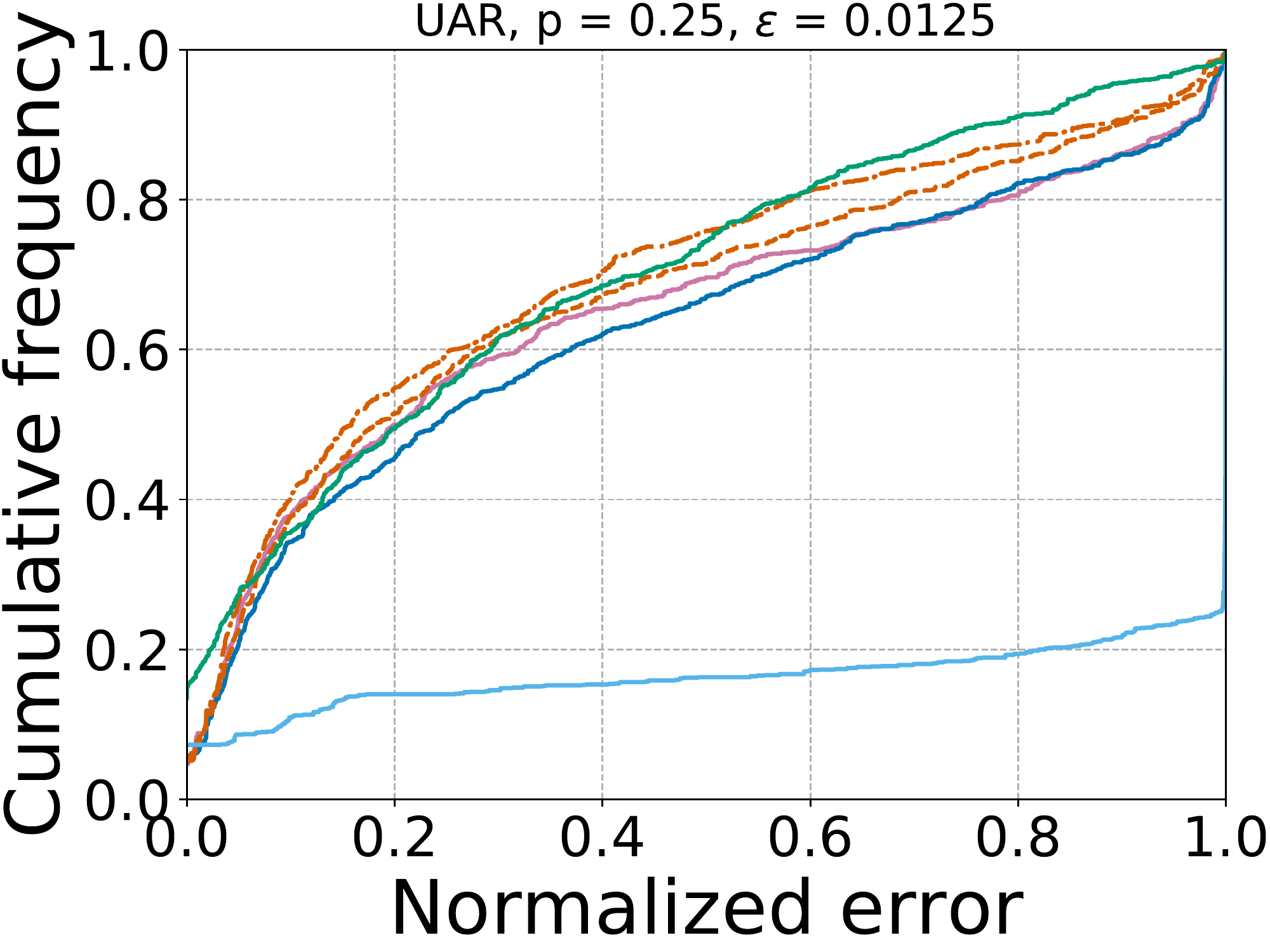}
  \end{subfigure}
  \begin{subfigure}[t]{0.32\textwidth}
    \centering
    \begin{Overpic}{\includegraphics[width=\textwidth,trim={38 0 0 0},clip]{figs/cdfs/agg_ratio_eq/corruption={st,ctws=1,cpws=0.25,cti=1,cpi=0.0},explore_method={expl,eps=0.0125},/cdf_nolegend.pdf}} 
    \end{Overpic}
    \caption{}
    \label{fig:0.0125-1-0.25}
  \end{subfigure}
  \begin{subfigure}[t]{0.32\textwidth}
    \centering
    \begin{Overpic}{\includegraphics[width=\textwidth,trim={38 0 0 0},clip]{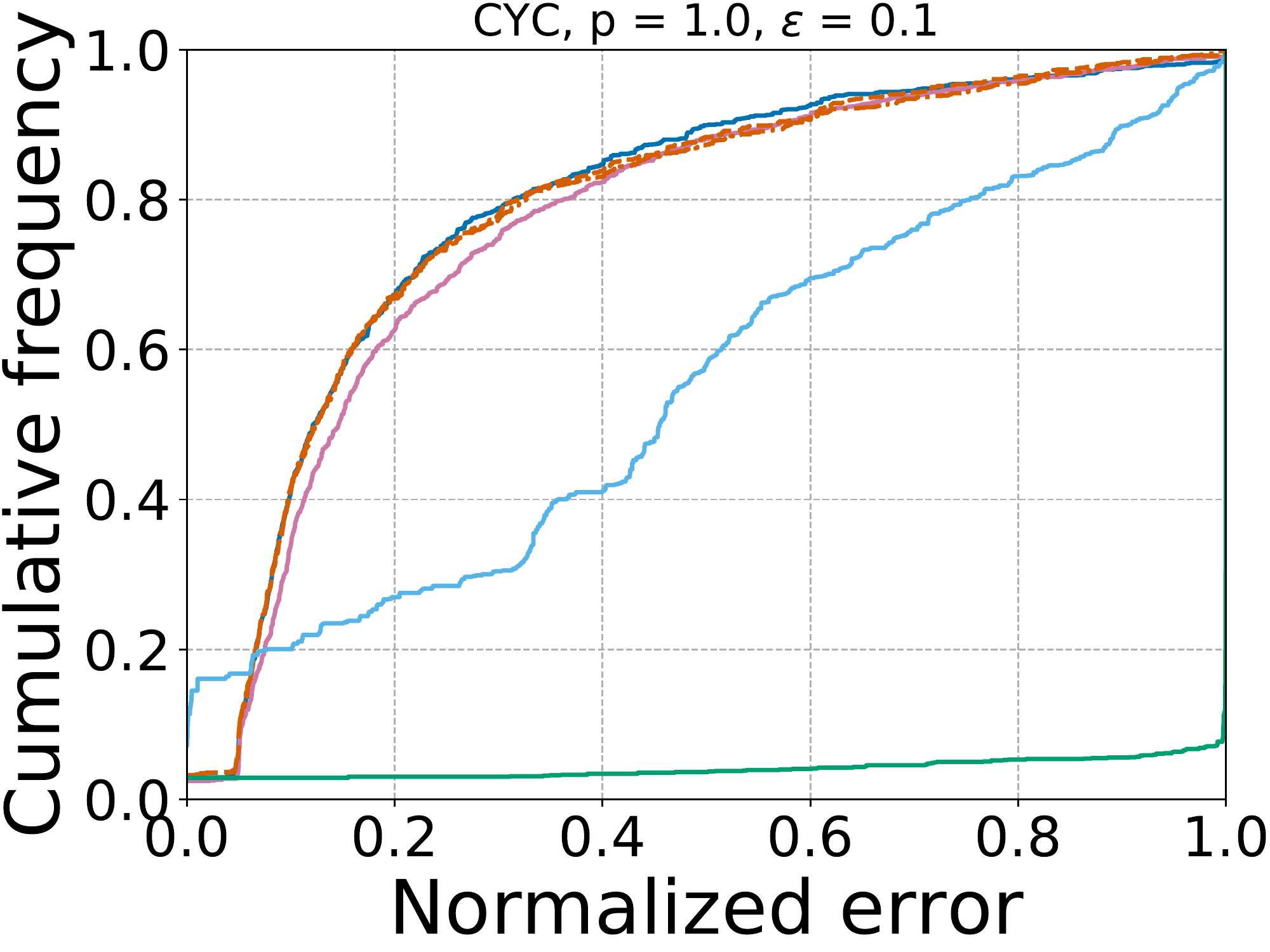}} 
    \end{Overpic}
    \caption{}
    \label{fig:0.1-2-1.0}
    \end{subfigure}
    \begin{subfigure}[t]{0.32\textwidth}
      \centering
      \begin{Overpic}{\includegraphics[width=\textwidth,trim={38 0 0 0},clip]{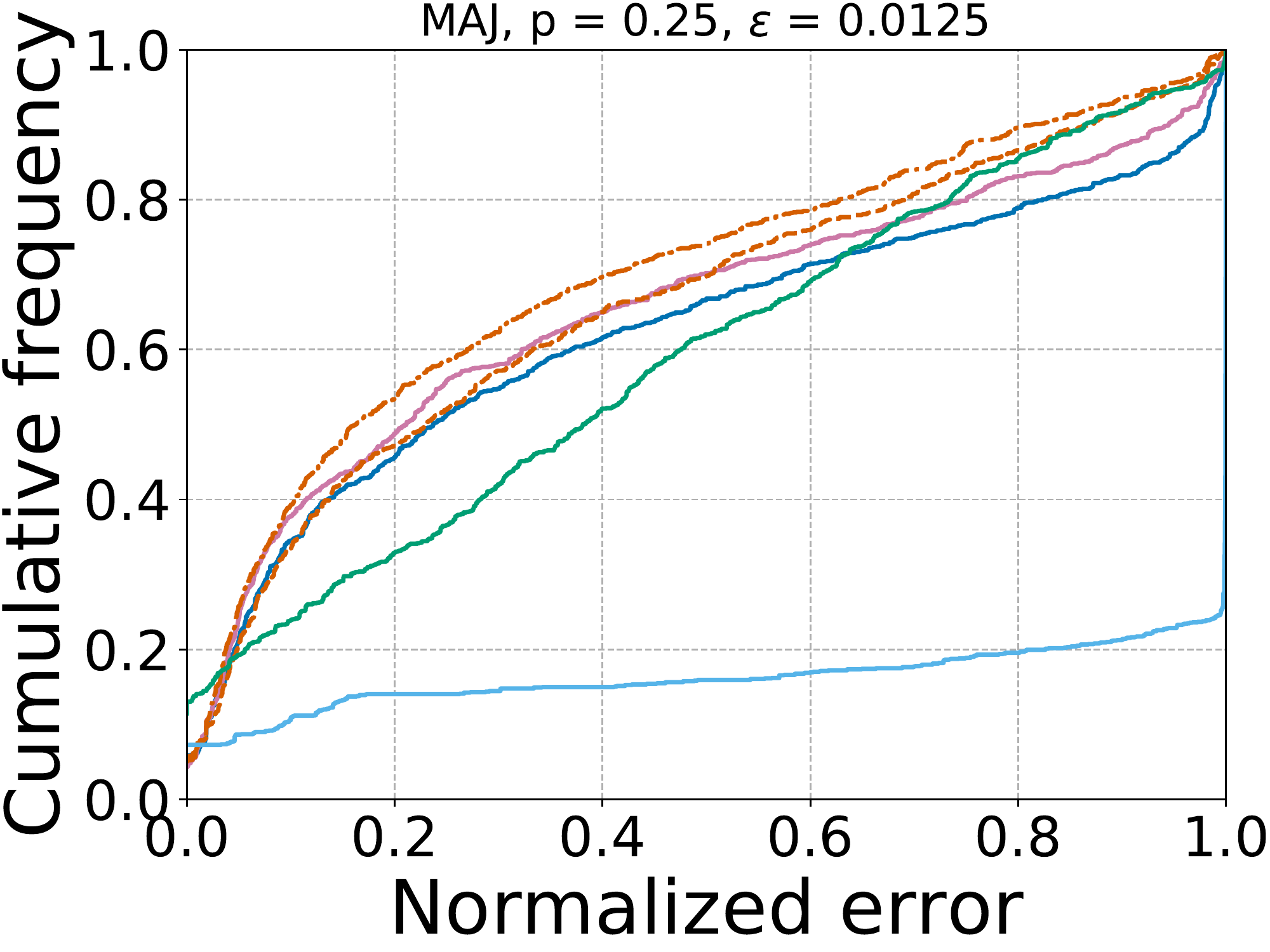}} 
      \end{Overpic}
      \caption{}
      \label{fig:0.0125-3-0.25}
    \end{subfigure}%


  \begin{subfigure}[t]{\textwidth}
    \vspace{-1em}
    \includegraphics[width=\textwidth,trim={0 0 0 0},clip]{figs/cdfs/legends/legend_Lambda=8.pdf}
  \end{subfigure}

  \caption{Comparison of all algorithms in the CB ground truth setting using the empirical CDF of the normalized performance scores.
    Left: unbiased warm-start examples with noiseless (top) and
    UAR with probability 0.5 (down) costs on warm-start examples.
    Middle: extreme noise rate using CYC noise type with probability 1.0. All CB algorithms use $\epsilon = 0.0125$ for exploration (top) and $\epsilon = 0.1$ (bottom).
    Right: moderate and potentially helpful noise rates. The corruption added to the warm-start examples are of
    types CYC (top) and MAJ (down) respectively, with probability 0.25.}
  \label{fig:cdfs}
\end{figure*}

\begin{figure*}[t!] 
  \centering
  \captionsetup[subfigure]{labelformat=empty}

  \begin{subfigure}[t]{2mm}
    \centering
    \includegraphics[width=3mm,trim={0 43 540 0},clip]{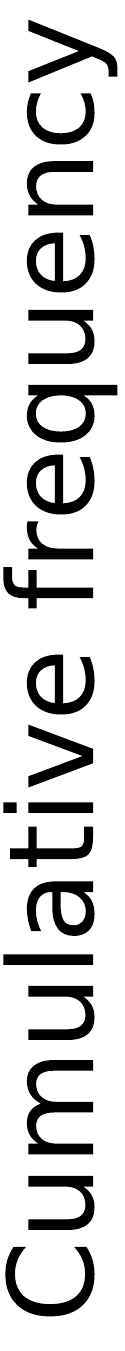}
  \end{subfigure}
  \begin{subfigure}[t]{0.32\textwidth}
    \centering
    \begin{Overpic}{\includegraphics[width=\textwidth,trim={38 43 0 0},clip]{figs/cdfs/no_agg/corruption={st,ctws=3,cpws=0.25,cti=1,cpi=0.0},inter_ws_size_ratio={2.875},explore_method={expl,eps=0.0125},/cdf_nolegend.pdf}} \put(15,55){\textcolor{blue!50!black}{\large (a)}} \end{Overpic}
    \caption{}
    \label{fig:2.875}
  \end{subfigure}%
  \begin{subfigure}[t]{0.32\textwidth}
    \centering
    \begin{Overpic}{\includegraphics[width=\textwidth,trim={38 43 0 0},clip]{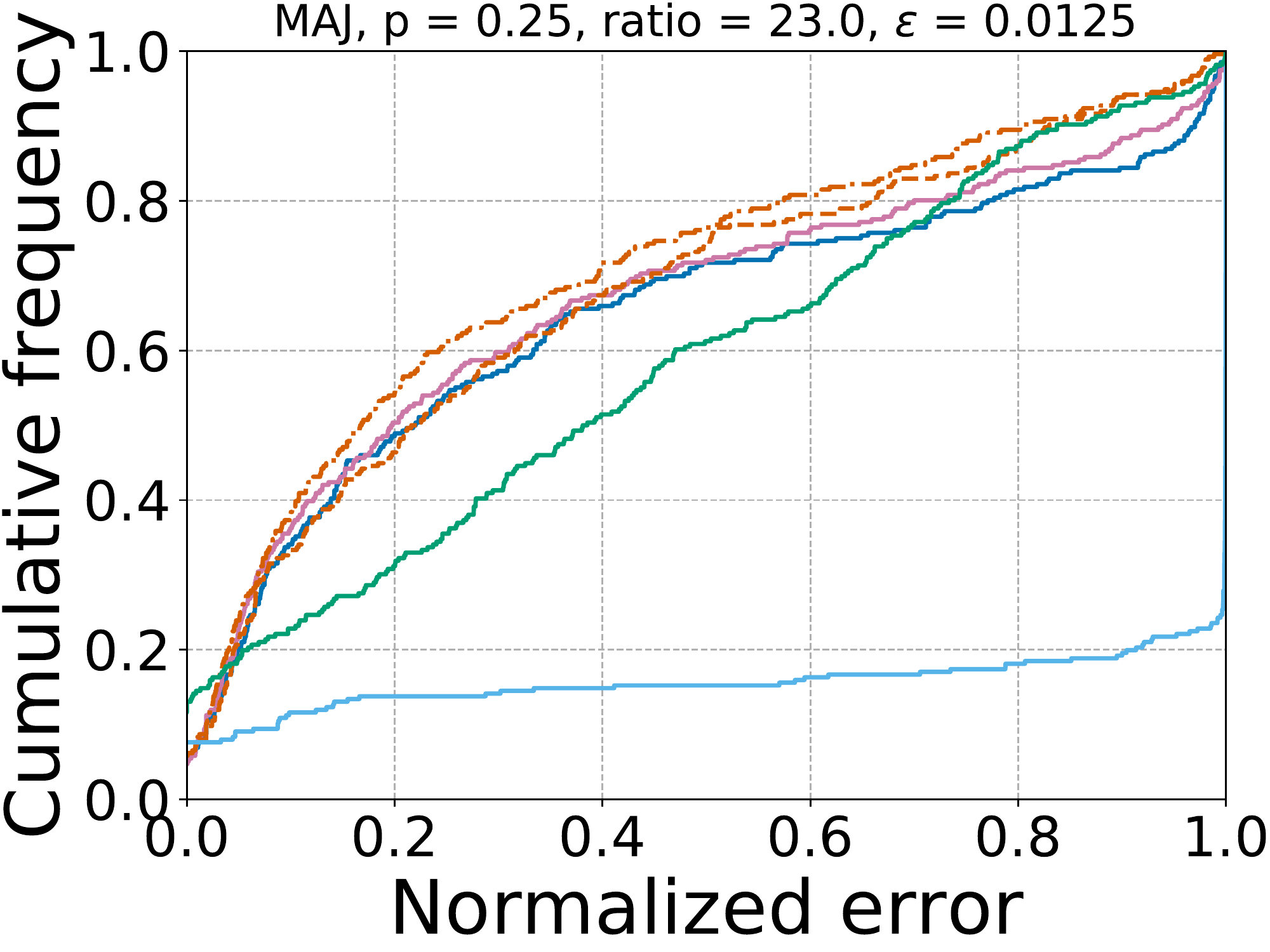}} \put(15,55){\textcolor{blue!50!black}{\large (b)}} \end{Overpic}
    \caption{}
    \label{fig:23}
  \end{subfigure}
  \begin{subfigure}[t]{0.32\textwidth}
    \centering
    \begin{Overpic}{\includegraphics[width=\textwidth,trim={38 43 0 0},clip]{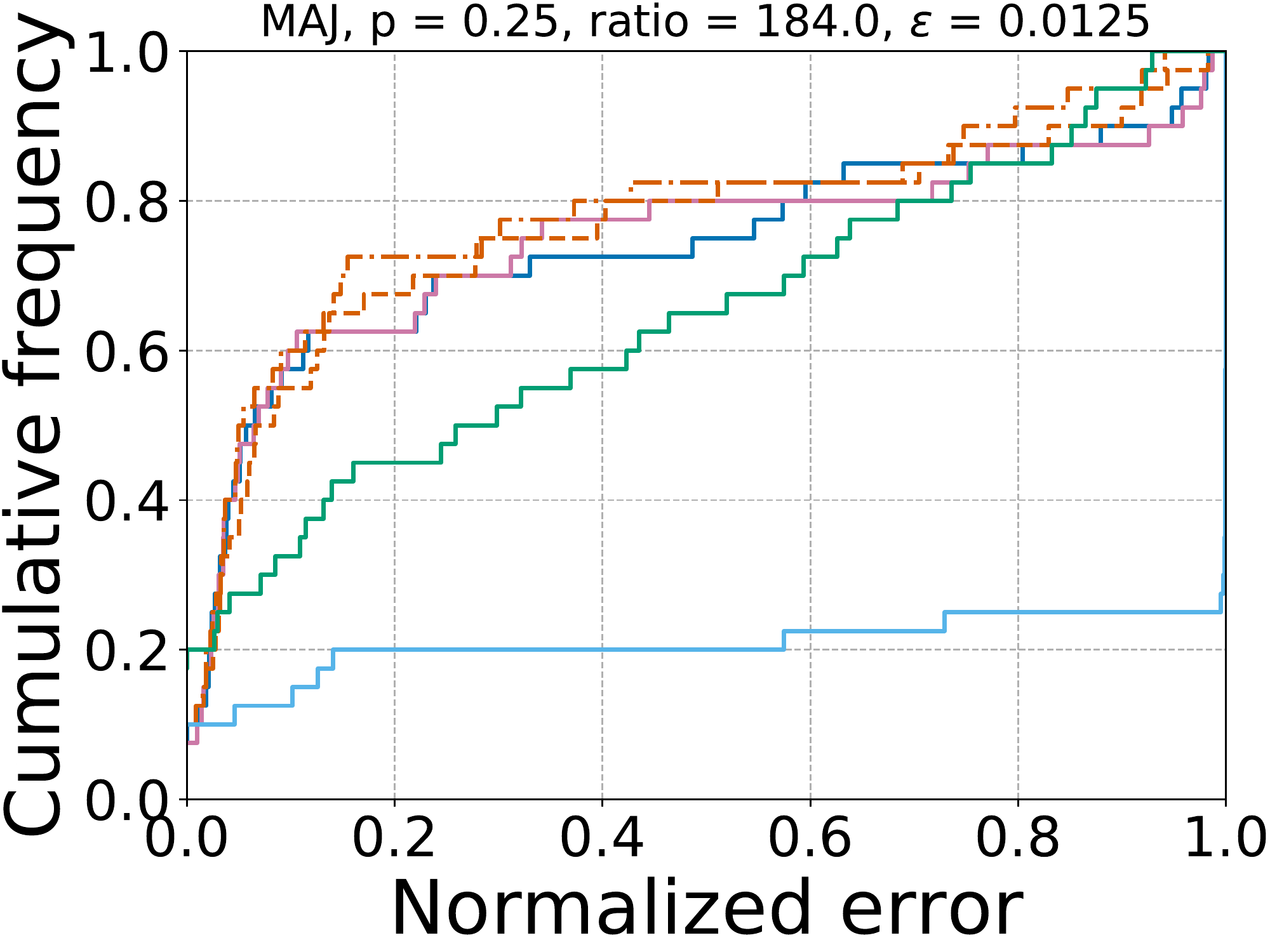}} \put(15,55){\textcolor{blue!50!black}{\large (c)}} \end{Overpic}
    \caption{}
    \label{fig:184}
  \end{subfigure}

  \caption{
  Effect of varying warm-start ratios for MAJ noise with $p = 0.25$. The warm-start ratios vary from 2.875 (left), 23 (middle) to 184 (right). Each CDF aggregates over all conditions of this noise type with the same warm-start ratio. 
  }
  \label{fig:vary-ws-ratio}
\end{figure*}

\paragraph{Evaluation Criteria.}
For each (dataset, $\ns$, $\nb$) combination $c$, we can compute $e_{c,a}$ to be the average cost of algorithm $a$ on the CB examples.
Because the range of $e_{c,a}$ can vary significantly over different settings $c$, we normalize these
to yield the \emph{normalized error} of an algorithm on a dataset: $\err_{c,a} := \frac{ e_{c,a} - e^*_c }{\max_b e_{c,b} - e^*_c}$,
where $e^*_c$ is the error achieved by a fully supervised one-versus-all learning algorithm trained on all the examples with original labels in this dataset.
Lower normalized error indicates better performance.
We plot the cumulative distribution function (CDF) of the normalized errors for each algorithm.
That is, for an algorithm $a$, at each point $x$, the $y$ value is the fraction of $c$'s such that $\err_{c,a} \leq x$.
In general, a high CDF value at a small $x$ indicates that the algorithm is performing well over a large number of (dataset, $\ns$, $\nb$) combinations.

In some of the plots when investigating the effect of a particular type or level of noise, we find it useful to aggregate the plots further over all warm-start ratios in creating the CDF and this aggregation is done by a pointwise averaging of the individual CDFs.

\paragraph{Comparison with baselines using both sources.}
We present the CDFs of all algorithms under various noise models in Figure~\ref{fig:cdfs}, with detailed results for individual noise levels, warm-start ratios and different $\epsilon$ values in Appendix~\ref{apx:moreresults}. In Figure~\ref{fig:cdfs}, we aggregate over warm-start ratios as described earlier. We can see from the figures that \ourname's CDFs (approximately) dominate those of \textsc{Sim-Bandit} and \ourname with $|\Lambda|=2$, which use weightings of $0.5$, and the best of $\{0,1\}$ respectively. These gains highlight the importance of being more careful about selecting a good weighting, despite the earlier intuition from Proposition~\ref{prop:minimax}. We see that there is a potentially added benefit of using different
$\lambda$'s in different phases of learning which might even outperform the best setting in hindsight.
%
%
\vspace{-0.1in}
\paragraph{Results for aligned cost structures.}
In Fig.~\ref{fig:0.0125-noiseless}, we consider the setting $\cs = \cb$. Here, \ourname's CDF dominates all other algorithms other than \textsc{Sup-Only}. For \textsc{Sup-Only}, the warm-start policy is used greedily with no exploration, making it a very strong baseline when there is no bias. We observe that \ourname uses the warm-start much more effectively than both the \textsc{Sim-Bandit} and \ourname with $|\Lambda|=2$ baselines. Our next experiments consider a uniform at random (UAR) noise setting, where the supervised data is unbiased (with respect to $\cb$) but has higher variance. In particular, for every example $(x,\cb)$, with probability $1-p$ we set $\cs=\cb$ and with probability $p$ we set $\cs$ as the classification error against a uniform random label.
From Claim~\ref{claim:uar} in the appendix, $\Ds$ is $(1-p,0)$-similar to $\Db$.
We plot the CDFs of the algorithms in the case where $p=0.25$ in Fig.~\ref{fig:0.0125-1-0.25}. The ordering of the CDFs stays the same, with \textsc{Sup-Only} less dominant (unsurprisingly), and with the gaps between methods reduced with the reduced utility of the warm-start data.
%
\paragraph{Results with adversarial noise.}
We next conduct an experiment where $\cs$ and $\cb$ are highly misaligned in order to understand how robust \ourname is to adversarial conditions.
We consider the cycling noise model (CYC), where we set the supervised costs to be ``off-by-one'' from the CB costs.
Specifically, if $\cb$ declares action $a$ to be the zero-cost action, then, with probability $p$, $\cs$ corrupts the costs so that action $(a+1)~\mod~K$ becomes the zero-cost action.
From Claim~\ref{claim:arbit} in the appendix, $\Ds$ is $(1, 2p)$-similar to $\Db$.
The CDF results for this experiment are in Figs.~\ref{fig:0.0125-2-1.0},~\ref{fig:0.1-2-1.0} ($p = 1$) for $\epsilon = 0.0125$ and $\epsilon = 0.1$, and Fig.~\ref{fig:0.0125-2-0.25} ($p=0.25$) for  $\epsilon = 0.0125$ respectively.
Again, \ourname is dominant amongst methods which use both the sources.
In the case of $p=1.0$, \textsc{Bandit-Only} performs the best as the warm start examples are misleading.  \ourname performs slightly worse (Fig.~\ref{fig:0.0125-2-1.0}) due to the model selection overhead, as discussed following Theorem~\ref{thm:bandit-gt}.
This gap is reduced when we increase the $\epsilon$ value in $\epsilon$-greedy to $0.1$ (Fig.~\ref{fig:0.1-2-1.0}).
In the case of $p=0.25$ (Fig.~\ref{fig:0.0125-2-0.25}), \ourname outperforms all the methods, showing that it can utilize warm start examples even if they are moderately biased.

\paragraph{Results with majority noise.}
Finally, we consider the case of a noise model that replaces the ground truth label with the majority label, roughly modeling a ``lazy annotator'' who occasionally defaults to the most frequent class.
For the majority noise model (MAJ), with probability $1-p$, we set $\cs=\cb$ and with probability $p$ we set $\cs$ to a cost vector that has a zero for the most frequent label in this dataset and one elsewhere.
From Claim~\ref{claim:arbit} in the appendix, $\Ds$ is $(1, 2p)$-similar to $\Db$.
The CDFs for this setting are shown in Figure~\ref{fig:0.0125-3-0.25}, where we again see \ourname dominating all the baselines (similar to Figure~\ref{fig:0.0125-2-0.25}).

In sum, we observe that \ourname is the \emph{only} method which is the best or close across all the noise regimes; no other approach is consistently strong. In practical scenarios, where the extent of bias in the warm-start is difficult or costly to ascertain, this robust performance of \ourname is extremely desirable. If we have some prior information about the noise level, it is prudent to prefer smaller $\epsilon$ when we expect a low noise (to compete well with \textsc{Sup-Only}), while a larger $\epsilon$ is preferred in high noise situations (to quickly detect the extent of bias).

While we present aggregates over warm-start ratios here, plots for each combination of noise type, level and warm-start ratio for three values of $\epsilon$ are shown in Appendix~\ref{apx:moreresults}.

\vspace{-0.1in}
\paragraph{Effect of warm-start ratio.} In Fig.~\ref{fig:vary-ws-ratio}, we pick a moderate noise setting and study the ordering of the different methods as the number of warm-start examples $\ns$ increases relative to $\nb$. We see \ourname outperforming all methods. \textsc{Sup-Only} is strong on the left for a small ratio (2.875), while \textsc{Bandit-Only} does well on the other extreme (184), and \ourname consistently outperform both the baselines combining the two sources.
\vspace{-0.1in}
\paragraph{Overall.}
Overall, we see that effectively using warm-start examples can certainly improve the performance of CB approaches. \ourname provides a way to do this in a robust manner, consistently outperforming most baselines. This is best evidenced in Figure~\ref{fig:agg-0.0125}, which further aggregates performance across the following 10 noise conditions on the warm start examples: noiseless and  \{UAR, CYC, MAJ\} corruptions with probability $p$ in $\{0.25, 0.5, 1.0\}$.




\section{Discussion and Future Work}
In this paper, we study the question of incorporating multiple data sources in CB settings. We see that even in simple cases, obvious techniques do not work robustly, and some care is required to handle biases from the non-ground-truth source.

Building on our results, there are several natural avenues for future work.  Doing a similar modification to more advanced exploration algorithms (e.g.~\citep{AG13,AHKLLS14}) is significantly more challenging. This falls into the general category of selecting the best from an ensemble of CB algorithms (where the ensemble corresponds to different weightings of the supervised and the CB examples). In $\epsilon$-greedy, the policy training corresponds to training the CB algorithm on reweighted data, while the model selection over $\lambda$ induces the action distribution. While the first step is typically straightforward even for other CB algorithms, finding an action distribution which looks good at this round, while allows the CB algorithms for different $\lambda$ values to make subsequent updates is significantly harder (for instance, when using a UCB style strategy, each $\lambda$ value might suggest a completely different action and expect reward feedback about it).
A possible approach is to employ ideas from the \textsc{Corral} algorithm~\citep{agarwal2016corralling}, but the cost of model selection is linear instead of logarithmic in $|\Lambda|$, and the approach is somewhat data inefficient due to restarts.  More ambitiously, it is desirable for the schedule of supervised and CB examples to not be fixed in a warm-start fashion but based on active querying, such as by sending uncertain examples to a labeler for full supervision. Studying this and considering broader sources of feedback are both interesting future research.

\section*{Acknowledgments} We thank Alberto Bietti for kindly sharing the scripts for experiments performed by~\citet{BAL18}, and helping getting the experiments running. We also thank the anonymous reviewers for their helpful feedback.

\newpage
\bibliographystyle{icml2019}
\bibliography{multimodal}
\newpage
\onecolumn
\appendix
\section{Additional Experimental Details}
We run our experiments using Vowpal Wabbit (VW)\footnote{\url{http://hunch.net/~vw/}}. In all our algorithms, we consider a scorer function class $\calF$ that contains functions $f$ that map $(x,a)$ to estimated cost values. The policy class $\Pi$ induced by $\calF$ is defined as:
\[ \Pi := \cbr{ \pi_f: f \in \calF }, \text{ where } \pi_f(x) = \argmin_{a \in [K]} f(x,a). \]
In general, our algorithms do not learn with respect to $\Pi$ directly; instead, at each time $t$, they find some scorer function $f_t$, and use its induced policy $\pi_{f_t}$ in $\Pi$ to perform exploration and exploitation.

In all CB learning algorithms (\textsc{Bandit-Only} and \textsc{Sim-Bandit}), we use the $\epsilon$-greedy exploration strategy with $\epsilon = 0.0125$ in most of the results for the main text, while the results for two other values ($0.1$ and $0.0625$) in Appendix~\ref{apx:moreresults}. We use the importance weighted regression algorithm (IWR)~\cite{BAL18} to compute cost regressors $f_t$ in $\calF$. The function class we use consists of linear functions: $f_w(x,a) = \langle w_a,x\rangle$.
In the supervised learning algorithm (\textsc{Sup-Only}), we use the cost-sensitive one against all algorithm~\cite{BLZ05} to train a cost regressor $f_0$ using all the warm start examples, and make no updates in the interaction stage. In the \textsc{Majority} algorithm, we simply predict using the majority class in the dataset and make no updates in the interaction stage.
\emph{Note that there is no exploration in \textsc{Sup-Only} and \textsc{Majority}}. For \ourname, we also use the same set of $\epsilon$ values as \textsc{Bandit-Only} and \textsc{Sim-Bandit}. At time $t$ and for $\lambda$ in $\Lambda$, we approximate $\pi_t^\lambda$ as follows.
Instead of optimizing
the objective function in Equation~\eqref{eqn:pi-t-lambda},
we start with finding an approximate optimizer of the following objective function:
\[ f_t^\lambda = \argmin_{f \in \calF} \cbr{(1-\lambda) \sum_{(x,c) \in S} \sum_{a=1}^K (f(x, a) - c(a))^2 + \lambda \sum_{\tau=1}^t \frac{1}{p_{\tau,a_\tau}} (f(x_\tau, a_\tau) - c_\tau(a_\tau))^2},  \]
and then take $\pi_t^\lambda = \pi_{f_t^\lambda}$.
For computational efficiency, we choose not to find
the exact empirical cost minimizer $f_t^\lambda$ at each time. Instead, we use a variant of online gradient descent in VW with adaptive~\cite{duchi2011adaptive}, normalized~\cite{ross2013normalized} and importance-weight aware updates~\cite{KL11} on the objective function.


For computational efficiency, in \ourname, at time $t \geq 2$, instead of mixing the average policy of $\cbr{\pi_\tau^\lambda}_{\tau=1}^{t-1}$ with $\epsilon$-uniform exploration, we predict by mixing the most recent policy
$\pi_t^\lambda$ with $\epsilon$-uniform exploration.

We vary the learning rates of all algorithms from $\cbr{0.1, 0.03, 0.3, 0.01, 1.0, 0.003, 3.0, 0.001, 10.0}$. For each algorithm and dataset/setting combination,
we first compute the average cost in the interaction stage. For different algorithm/dataset/setting combinations, we select different learning rates that minimize the corresponding average cost.

The source codes for running the experiments are available at \url{https://github.com/zcc1307/warmcb_scripts}.


\section{Similiarity Conditions between Cost-sensitive Distributions}
\label{sec:similar}

In this section, we introduce an alternative, more intuitive definition of similarity between two
distributions of cost-sensitive examples. Then we show that this
notion is stronger than~\autoref{def:similar}.

\begin{definition}
Distribution $D_2$ is strongly $(\scal, \bias)$-similar to $D_1$ wrt policy class $\Pi$, if there exists a joint distribution $D$ over $(x,c^+,c^-) \in \calX \times [0,1]^{K} \times [-1,1]^{K}$ such that
\[ \E_D[c^+(a) + c^-(a) | x] = \E_{D_2}[c(a) | x], \quad \text{ $\forall$ $a \in [K]$ and $x \in\calX$, } \]
and there exists a policy
$\pi^* \in \argmin_{\pi \in \Pi} \E_{D_1}[c(\pi(x))]$, such that
both of the following hold:
\begin{enumerate}[nolistsep,noitemsep]
  \item For any policy $\pi$ in $\Pi$, $\E_D[c^+(\pi(x))] - \E_D[c^+(\pi^*(x))] \!\geq\! \scal (\E_{D_1}[c(\pi(x))] - \E_{D_1}[c(\pi^*(x))])$.
  \item The expected magnitude of $c^-$ is bounded: for any policy $\pi$, $|\E_D[c^-(\pi(x))]| \leq \bias / 2$.
\end{enumerate}
\label{def:strongly-similar}
\end{definition}

In the definition above, we use $+/-$ superscripts to indicate the utility of $c^+,c^-$, which give a decomposition of the cost structure generated from $D_2$. Intuitively, $c^+$ is the component useful for learning under $D_1$, as low excess cost under $c^+$ implies low excess cost under $D_1$. In contrast, $c^-$ may or may not have such property.

We have the following lemma upper bounding a policy's excess cost on $D_1$ in terms of its excess cost on $D_2$:

\begin{lemma}
  Suppose $D_1$ and $D_2$ are two distributions over cost-sensitive examples. If $D_2$ is strongly $(\scal, \bias)$-similar to $D_1$ with respect to policy class $\Pi$, then $D_2$ is $(\scal, \bias)$-similar to $D_1$ with respect to policy class $\Pi$.
  \label{lem:similar}
\end{lemma}

\begin{proof}
Suppose $(x, c^+, c^-)$ is the decomposition of $(x, c)$ that satisfies
 Definition~\ref{def:strongly-similar}. We have that
 \begin{eqnarray*}
 \E_{D_2}[c(\pi(x))] - \E_{D_2}[c(\pi^*(x))] &=&
 (\E_D[c^+(\pi(x))] - \E_D[c^+(\pi^*(x))]) + (\E_D[c^-(\pi(x))] - \E_D[c^-(\pi^*(x))]) \\
 &\geq&
 \scal (\E_{D_1}[c(\pi(x))] - \E_{D_1}[c(\pi^*(x))]) - \Delta,
\end{eqnarray*}
where the inequality follows from using the respective items in Definition~\ref{def:strongly-similar} to lower bound each term (and observing that $\E_D[c^-(\pi(x))] - \E_D[c^-(\pi^*(x))] \leq \Delta/2 + \Delta/2 = \Delta$). The lemma follows.
\end{proof}

We now present a few examples that satisfy strong similarity.

\paragraph{Example 1: labeling the lowest-cost action.}
Consider a setting where eliciting the costs from an expert for warm-start examples is prohibitively expensive, but they can label the least costly action. Let $D_1$ be the CB source and let $\pi^*$ be the best policy under $D_1$.
Define $D_2$ by first sampling $(x, c_1) \sim D_1$ and returning $(x, \tilde{c})$ where $\tilde{c}(a) = I(a \neq a^*)$, for some $a^* \in \argmin_a c_1(a)$. Define $c^+(a) = I(a \neq \pi^*(x))$ and $c^- = \tilde{c} - c^+$.

\begin{claim}
$D_2$ is $(1,\bias)$-similar to $D_1$, with $\bias = 2\P(\cb(\pi^*(x)) \geq \min_{a \neq \pi^*(x)} \cb(a))$.
\end{claim}
\begin{proof}
On one hand, for any policy $\pi$,
\[ \E_D c^+(\pi(x)) - \E_D c^+(\pi^*(x)) = \P(\pi(x) \neq \pi^*(x)) \geq \E_{D_1} c_1(\pi(x)) - \E_{D_1} c_1(\pi^*(x)). \]

On the other hand,
observe that given a cost sensitive example $(x,c_1)$ and policy $\pi^*$, if $c_1(\pi^*(x)) < \min_{a \neq \pi^*(x)} c_1(a)$, then $\pi^*(x) = a^*$, in which case $\tilde{c} = c^+$ and $c^- = 0$. This implies that
\[ \P(c^- \neq 0) \leq \P(c_1(\pi^*(x)) \geq \min_{a \neq \pi^*(x)} c_1(a)) = \bias/2. \]
Therefore, for all policy $\pi$,
\[ \E|c^-(\pi(x))| \leq \P(c^- \neq 0) = \bias/2. \qedhere \]
\end{proof}


Given a multiclass label $y$, defined its induced zero-one cost vector
$c_y \in [0,1]^K$ as follows: $c_y(a) = 0$ if $a = y$ and $c_y(a) = 1$ otherwise.
From the definition of $c_y$, it can be seen that for any policy $\pi$,
$\E c_y(\pi(x)) = \P(\pi(x) \neq y)$.

\paragraph{Example 2: Uniform-at-random corruption.} Suppose $D_0$ is a joint distribution over multiclass examples $(x, y)$, and the corrupted label $\tilde{y}$ has the following conditional distribution given $(x,y)$: for all $a$ in $[K]$, $\P(\tilde{y} = a| (x,y) ) = (1-p)I(a=y) + \frac p K$. Define $D_1$ and $D_2$ as the joint distribution of $(x,c_y)$ and $(x,c_{\tilde{y}})$, respectively.

\begin{claim}
Suppose $D_1$ and $D_2$ are defined as above. Then $D_2$ is $(1-p, 0)$-strongly similar to $D_1$.
\label{claim:uar}
\end{claim}
\begin{proof}
Suppose $\pi^* \in \argmin_{\pi \in \Pi} \E_{D_1} c(\pi(x))$ is an optimal policy with respect to $D_1$.

For any policy $\pi$, by the definition of $\tilde{y}$ and $c_{\tilde{y}}$, we have
that
\[ \E c_{\tilde{y}}(\pi(x)) = \P(\pi(x) \neq \tilde{y}) = (1-p)\P(\pi(x) \neq y) + p \frac{K-1}{K}. \]
Therefore, for any policy $\pi$, the below identity holds:
\[ \E c_{\tilde{y}}(\pi(x)) - \E c_{\tilde{y}}(\pi^*(x)) = (1-p) (\E c_{y}(\pi(x)) - \E c_{y}(\pi^*(x))). \]
Therefore, taking $c^+ = c_{\tilde{y}}$, $c^- = 0$, $\scal = 1-p$ and $\bias=0$, it can be easily checked that the conditions of Definition~\ref{def:similar} are satisfied.
\end{proof}

\paragraph{Example 3: general corruption.} Suppose $D_0$ is a joint distribution over multiclass examples $(x, y)$, and the corrupted label $\tilde{y}$'s conditional distribution given $(x,y)$ has the following property: $\P(\tilde{y} = y| (x,y) ) \geq 1-p$. Define $D_1$ and $D_2$ as the joint distribution of $(x,c_y)$ and $(x,c_{\tilde{y}})$, respectively.

\begin{claim}
Suppose $D_1$ and $D_2$ are defined as above. Then $D_2$ is $(1, 2p)$-strongly similar to $D_1$.
\label{claim:arbit}
\end{claim}
\begin{proof}
Suppose $\pi^* \in \argmin_{\pi \in \Pi} \E_{D_1} c(\pi(x))$ is an optimal policy with respect to $D_1$.

For every $x$, define deterministically that
$c^+ = \E[c_y | x]$, and $c^- = \E[c_{\tilde{y}}|x] - \E[c_y | x]$.
We have that $\E[c^++c^-|x] = \E[c_{\tilde{y}}|x]$ by the definitions of $c^+$ and $c^-$.
In addition, by the construction of $c^+$, we immediately have that
\[ \E c^+(\pi(x)) - \E c^+(\pi^*(x)) =  \E c_y(\pi(x)) - \E c_y(\pi^*(x)). \]
What remains is to bound $\E c^-(\pi(x))$. By the definitions of $c_y$ and $c_{\tilde{y}}$, we have that
\[ |\E c^-(\pi(x))| = |\P(\pi(x) \neq \tilde{y}) - \P(\pi(x) \neq y)| \leq \P(y \neq \tilde{y}). \]
By the assumption on the conditional distribution of $\tilde{y}$ given $(x,y)$, we have that the right hand side is at most $p$. The claim follows.
\end{proof}

\section{Proof Showing the Failure of Equal Data Weighting}
\label{app:ucb}

In this section we formalize the example presented in~\autoref{subsec:simple-fail}.
To recall, this is a $2$-armed bandit setting (i.e. contextual bandits with a dummy context), where policy class $\Pi := \cbr{\pi_1, \pi_2}$, where $\pi_i$ maps any context to action $i$, $i = 1,2$. In addition, $\Ds$ (resp. $\Db$) is the Dirac measure
on $(x_0, \cs_0)$ (resp. $(x_0, \cb_0)$), and the respective cost vectors are:
\[
  \cs_0 = (0.5, 0.5 + \frac \Delta 2), \quad \mbox{and} \quad \cb_0 = (0.5, 0.5 - \frac \Delta 2).
\]
We consider the following algorithm, which directly extends the UCB1 algorithm~\cite{auer2002finite} by additionally using the warm start examples to estimate the mean costs of the two actions. Note that as it is minimizing its cumulative cost, the algorithm computes lower confidence bounds of the costs and selects the minimum, which is equivalent to computing upper confidence bounds of the rewards and selecting the maximum.

\begin{algorithm}[H]
\caption{A variant of the UCB1 algorithm that accounts for warm start examples}
\begin{algorithmic}[1]\label{alg:ucb}
\REQUIRE Supervised examples $S = \cbr{(x_0, \cs)}$ of size $\ns$, number of interaction rounds $\nb$.
\FOR{$t=1,2,\ldots,\nb$}
  \STATE For $i=1,2$, define $n_{i,t-1} = \sum_{s=1}^{t-1} I(a_t = i)$.

  \STATE For $i = 1,2$, compute empirical mean cost of action $i$: $\hat{\mu}_{i,t} = \frac{\sum_{(x_0,\cs) \in S} \cs(i) + \sum_{s=1}^{t-1} I(a_t = i) \cb_s(i)}{\ns + n_{i,t-1}}$.

  \STATE For $i = 1,2$, compute $\lcb_{i,t} = \hat{\mu}_{i,t} - 2 \sqrt{\frac{\ln t}{\ns + n_{i,t-1}}}$.

  \STATE Take action $a_t = \argmin_{i \in \cbr{1,2}} \lcb_{i,t}$.

  \STATE Observe $\cb_t(a_t)$.

\ENDFOR
\end{algorithmic}
\end{algorithm}

We have the following proposition on a lower bound of the regret of Algorithm~\ref{alg:ucb}, under the above settings of $\Ds$ and $\Db$.

\begin{proposition}
Suppose $\Ds$ and $\Db$ are defined as above. Additionally, Algorithm~\ref{alg:ucb} is run
with input $\ns$ warm start examples drawn from $\Ds$ and number of interaction rounds $\nb \geq \exp(\Delta^2 \ns/16)$. Then, Algorithm~\ref{alg:ucb} incurs a regret of $\Omega( \Delta \exp(\Delta^2 n_s/16))$.
\label{prop:ucbfail}
\end{proposition}

\begin{proof}
Suppose after $t - 1$ rounds of the interaction phase, Algorithm~\ref{alg:ucb} has taken action $i$ $t_i$ times for $i=1,2$. As the cost vectors are deterministic, we can calculate the
 lower confidence bound estimates for the two actions in closed form:
\[
\lcb_{t,1} = 0.5 - 2\sqrt{\frac{\ln (t_1 + t_2 + 1)}{\ns + t_1}} \quad \mbox{and} \quad \lcb_{t,2} = 0.5 + \frac \Delta 2 \,\frac{\ns - t_2}{\ns + t_2} - 2\sqrt{\frac{\ln (t_1 + t_2 + 1)}{\ns + t_2}}
\]


Now, let us consider the first time we play action $2$ in the interaction phase. At this point, $t_2$ is still $0$ and $t_1$ should satisfy
\[
\lcb_{t,2} \leq \lcb_{t,1}, \quad \mbox{that is} \quad \frac \Delta 2 - 2\sqrt{\frac{\ln (t_1 + 1)}{\ns}} \leq -2\sqrt{\frac{\ln (t_1 + 1)}{\ns + t_1}}.
\]

The above condition implies that
\[
2\sqrt{\frac{\ln (t_1 + 1)}{\ns}} \geq \frac\Delta2, \quad \mbox{equivalently} \quad t_1 \geq \exp(\Delta^2 \ns/16) - 1.
\]

Denote by $T_0 := \exp(\Delta^2 \ns/16) - 1$.
Therefore, after $\nb \geq \exp(\Delta^2 \ns/16)$ rounds of interaction, the regret of Algorithm~\ref{alg:ucb} can be lower bounded by:
\[
\sum_{t=1}^{\nb} \cb_0(a_t) - \cb_0(2)  \geq \sum_{t=1}^{T_0 - 1} \cb_0(1) - \cb_0(2) = \frac \Delta 2 \cdot (T_0 - 1) = \Omega( \Delta \exp(\Delta^2 \ns/16) ).
\qedhere \]
\end{proof}

In Algorithm~\ref{alg:ucb}, we used only $\ln(t + 1)$ in the numerator, when an alternative might be to use $\ln(t + \ns + 1)$. However, it is easily checked that after this simple modification, a similar exponential regret lower bound of Algorithm~\ref{alg:ucb} can be proved (with the definition of $T_0$ changed to $T_0 := \exp(\Delta^2 \ns/16) - \ns - 1$).

\section{Concentration Inequalities}
We use a version of Freedman's inequality from~\cite{BLLRS11}.
\begin{lemma}[Freedman's inequality]
Let $X_1, \ldots, X_n$ be a martingale difference sequence adapted to filtration $\cbr{\calB_i}_{i=0}^n$, and $\abs{ X_i } \leq M$ almost surely for all
$i$. Let $V = \sum_{i=1}^n \E[X_i^2 | \calB_{i-1} ]$ be the cumulative conditional variance. Then, with probability $1-\delta$,
\[
\abs{ \sum_{i=1}^n X_i } \leq 2 \sqrt{V \ln \frac{2}{\delta}} + M \ln \frac{2}{\delta}.
\]
\label{lem:freedman}
\end{lemma}

\section{Proof of Theorem~\ref{thm:bandit-gt}}
\label{app:bandit-gt}

We begin with some additional notation used in the analysis.
Throughout this section, we	let $\pi^* = \argmin_{\pi \in \Pi} \E \cb(\pi(x)) $, the optimal policy in $\Pi$ with respect to $\Db$.

Recall that for policy $\pi$ and  $\lambda$ in $[0,1]$,
we define $\E_t[\hcb(\pi(x))] := \frac 1 t \sum_{s=1}^t \hcb_s(\pi(x))$ and
$\E_{\Ss}[\cs(\pi(x))] := \frac 1 {\ns} \sum_{(x,c) \in \Ss} c(\pi(x))$.
We additionally define the $\lambda$-weighted empirical cost
of $\pi$ as
\[
\hat{\risk}_{\lambda,t}(\pi) = \frac{\lambda t \E_t[\hcb(\pi(x))]  + (1-\lambda) \ns \E_{\Ss}[\cs(\pi(x))] }{\lambda t + (1-\lambda)\ns},
\]
and its expectation
\[
\risk_{\lambda,t}(\pi) = \frac{\lambda t \E[\cb(\pi(x))] + (1-\lambda) \ns \E[\cs(\pi(x))] }{\lambda t + (1-\lambda)\ns}.
\]

Observe that
$\hat{\risk}_{1,t}(\pi) = \E_t[\hcb(\pi(x))]$ is $\pi$'s empirical cost on the first $t$ CB examples, $\risk_{1,t}(\pi) = \E \cb(\pi(x))$,
$\hat{\risk}_{0,t}(\pi) = \E_{\Ss}[\cs(\pi(x))]$ is $\pi$'s empirical cost on the $\ns$ supervised examples, and $\risk_{0,t}(\pi) = \E \cs(\pi(x))$.
Denote by $\hat{\unrisk}_{\lambda,t}(\pi) = (\lambda t + (1-\lambda)\ns) \hat{\risk}_{\lambda,t}(\pi)$ the unnormalized $\lambda$-weighted empirical cost
of $\pi$, and $\unrisk_{\lambda,t}(\pi) = (\lambda t + (1-\lambda)\ns) \risk_{\lambda,t}(\pi)$ its expectation. 

Denote by $(x_{\nb+1}, \cs_{\nb+1}), \ldots, (x_{\nb+\ns}, \cs_{\nb+\ns})$ an enumeration of the elements in $\Ss$.
Define filtration $\cbr{\calB_t}_{t=0}^{\nb+\ns}$ as follows: $\calB_0$ is the trivial $\sigma$-algebra, and
\[
\calB_t =
\begin{cases}
	\sigma((x_{\nb+1}, \cs_{\nb+1}), \ldots, (x_{\nb+1}, \cs_{\nb+t})), & t \in \cbr{1,\ldots,\ns}, \\
	\sigma(\Ss, (x_1, \hcb_1), \ldots, (x_{t-\ns}, \hcb_{t-\ns})), & t \in \cbr{\ns+1,\ldots,\ns+\nb}.
  \end{cases}
\]

For reader's convenience, we also recall our earlier notation:
\begin{align*}
\textstyle V_t(\lambda) &= \textstyle2\sqrt{\left(\lambda^2 \frac{K t}{\epsilon} + (1-\lambda)^2 \ns\right) \ln\frac{8\nb|\Pi|}{\delta} } + \textstyle \left(\frac{\lambda K}{\epsilon} + (1-\lambda)\right) \ln \frac{8\nb|\Pi|}{\delta}, \\
\textstyle  G_t(\lambda, \scal, \bias) &= \textstyle\frac{(1-\lambda)\ns \bias + 2V_t(\lambda)}{\lambda t + (1-\lambda) \ns \scal}.
\end{align*}

In addition, denote by
\[ \bar{G}_t(\lambda, \scal, \bias) := \min(1, G_t(\lambda, \scal, \bias)). \]


\begin{proof}[Proof of Theorem~\ref{thm:bandit-gt}]
Define event $I$ as: for all $\pi$ in $\Pi$,
\[  \abs{ \frac{1}{\nb}\sum_{t=1}^{\nb} (\E[\cb_t(\pi(x_t))|x_t] - \E \cb(\pi(x))) } \leq \sqrt{\frac{ \ln\frac{8|\Pi|}{\delta}}{2\nb}}. \]
By Hoeffding's inequality and union bound, $I$ happens with probability $1-\frac\delta{4}$.
Specifically, on event $I$, for every policy $\pi$ in $\Pi$, as $\pi^*$ is the policy in $\Pi$ that minimizes $\E \cb(\pi(x))$, we have
\[
  \E \cb(\pi^*(x)) - \frac{1}{\nb}\sum_{t=1}^{\nb} \E[\cb(\pi(x_t))|x_t]
  \leq \E \cb(\pi(x)) - \frac{1}{\nb}\sum_{t=1}^{\nb} \E[\cb(\pi(x_t))|x_t]
  \leq \sqrt{ \frac{\ln\frac{8|\Pi|}{\delta}}{2\nb}}.
\]
Therefore,
\begin{equation}
	 \E \cb(\pi^*(x)) - \min_{\pi \in \Pi}  \frac{1}{\nb}\sum_{t=1}^{\nb} \E[\cb(\pi(x_t))|x_t] \leq \sqrt{ \frac{\ln\frac{8|\Pi|}{\delta}}{2\nb}}.
	\label{eqn:hoeff-best}
\end{equation}

Recall that randomized policy $\pi_t: \calX \to \Delta^{K-1}$ is defined as $\pi_t(x) = \frac{1 - \epsilon}{t-1} \sum_{\tau=1}^{t-1} \pi_\tau^{\lambda_t}(x) + \frac\epsilon K \one_K$ for $t \geq 2$, and $\pi_1(x) = \frac 1 K \one_K$ for all $x$. With a slight abuse of notation, denote by $\E \cb(\pi_t(x)) := \E_{(x,\cb),a\sim\pi_t(x)} \cb(a)$.
Observe that $\E[ \E[\cb(a_t)|x_t] |\calB_{\ns+t-1}] = \E \cb(\pi_t(x))$. Define event $J$ as:

\begin{equation}
	\abs{ \frac{1}{\nb}\sum_{t=1}^{\nb} \E[\cb(a_t)|x_t] - \frac{1}{\nb} \sum_{t=1}^{\nb} \E \cb(\pi_t(x)) } \leq \sqrt{\frac{\ln\frac{8}{\delta}}{2\nb}}.
	\label{eqn:azuma-pit}
\end{equation}
By Azuma's inequality, $J$ happens with probability $1-\frac \delta 4$.

Denote by $E$ the event that the events $E_t$, $F_t$ defined in Lemmas~\ref{lem:bern-pi} and~\ref{lem:bern-pi-lambda-t} (both given below) and $I$, $J$ hold simultaneously for all $t$. By a union bound over all $E_t,F_t$'s and $I,J$, event $E$ happens with probability $1-\delta$.
We henceforth condition on $E$ happening.

Consider $t$ in $\cbr{2,\ldots,\nb}$. We now give an upper bound on the expected excess cost of using randomized prediction $\pi_t$.
Observe that by the definition of $\pi_t$,
\begin{equation}
	\E \cb(\pi_t(x)) = \frac{1-\epsilon}{t-1} \sum_{\tau=1}^{t-1} \E \cb(\pi_\tau^{\lambda_t}(x)) + \epsilon \frac1K\sum_{a=1}^K\E\cb(a) \leq \frac{1}{t-1} \sum_{\tau=1}^{t-1} \E \cb(\pi_\tau^{\lambda_t}(x)) + \epsilon,
	\label{eqn:epsilon}
\end{equation}
it suffices to upper bound $\frac{1}{t-1} \sum_{\tau=1}^{t-1} \E \cb(\pi_\tau^{\lambda_t}(x))$.


By Lemma~\ref{lem:bern-pi-lambda-t} below, we have that
\begin{equation}
\frac{1}{t-1} \sum_{\tau=1}^{t-1} \E \cb(\pi_\tau^{\lambda_t}(x)) - \min_{\lambda \in \Lambda} \frac{1}{t-1} \sum_{\tau=1}^{t-1} \E \cb(\pi_\tau^{\lambda}(x))
\leq 16 \sqrt{\frac{K\ln\frac{8\nb|\Lambda|}{\delta}}{(t-1)\epsilon}}.
\label{eqn:bern-selection}
\end{equation}

In the above inequality, we can bound the individual summands of the second term by Lemma~\ref{lem:bern-pi}:
\[
\E \cb(\pi_\tau^{\lambda}(x)) - \E \cb(\pi^*(x))
\leq
\bar{G}_{\tau-1}(\lambda, \scal, \bias).
\]

Therefore, rewriting Equation~\eqref{eqn:bern-selection}, we get that
\[ \frac{1}{t-1} \sum_{\tau=1}^{t-1} \E \cb(\pi_\tau^{\lambda_t}(x)) - \min_{\lambda \in \Lambda} \left[\frac{1}{t-1} \sum_{\tau=1}^{t-1} (\E \cb(\pi^*(x)) + \bar{G}_{\tau-1}(\lambda,\alpha,\Delta))\right]
 \leq 16 \sqrt{\frac{K\ln\frac{8\nb|\Lambda|}{\delta}}{(t-1)\epsilon}}. \]
Combining the above with Equation~\eqref{eqn:epsilon}, we get that
\[
\E \cb(\pi_t(x)) - \E \cb(\pi^*(x)) \leq \epsilon + \min_{\lambda \in \Lambda} \left[ \frac{1}{t-1}\sum_{\tau=1}^{t-1} \bar{G}_{\tau-1}(\lambda, \scal, \bias) \right] + 16 \sqrt{\frac{K\ln\frac{8\nb|\Lambda|}{\delta}}{(t-1)\epsilon}}.
\]
Summing the above inequality over $t$ in $\cbr{2,\ldots,\nb}$ and using that $\E \cb(\pi_1(x)) - \E \cb(\pi^*(x)) \leq 1$, we get,
\[
\sum_{t=1}^{\nb} [\E \cb(\pi_t(x)) - \E \cb(\pi^*(x))]  \leq \nb \epsilon +
 \sum_{t=2}^{\nb} \min_{\lambda \in \Lambda} \left[ \frac{1}{t-1}\sum_{\tau=1}^{t-1} \bar{G}_{\tau-1}(\lambda, \scal, \bias) \right]
+ 16 \sum_{t=2}^{\nb} \sqrt{\frac{K\ln\frac{8\nb|\Lambda|}{\delta}}{(t-1)\epsilon}} + 1.
\]
Observe that for any function $\cbr{N_t(\cdot)}_{t=1}^{\nb}$, $\sum_{t=1}^{\nb} \min_{\lambda \in \Lambda} N_t(\lambda) \leq \min_{\lambda \in \Lambda} [ \sum_{t=1}^{\nb} N_t(\lambda) ]$.
We further collect the coefficients on the $\bar{G}_\tau(\lambda,\scal,\bias)$ terms and use the upper bound $\sum_{t = \tau+1}^{\nb} \frac{1}{t-1} \leq \ln (e\nb)$ for all $\tau \geq 1$, getting
\begin{equation}
\sum_{t=1}^{\nb} [\E \cb(\pi_t(x)) - \E \cb(\pi^*(x))]  \leq \nb \epsilon +
 \ln (e\nb) \cdot \min_{\lambda \in \Lambda} \sum_{\tau=0}^{\nb-2} \bar{G}_\tau(\lambda, \scal, \bias)
+ 16 \sum_{t=2}^{\nb} \sqrt{\frac{K\ln\frac{8\nb|\Lambda|}{\delta}}{(t-1)\epsilon}} + 1.
\label{eqn:pregret}
\end{equation}

Therefore, we get:
\begin{eqnarray*}
&& \regretb \\
&\leq& \frac1{\nb}\sum_{t=1}^{\nb} [\E \cb(\pi_t(x)) - \E \cb(\pi^*(x))] + \sqrt{\frac{2 \ln\frac{8|\Pi|}{\delta} }{\nb}} \\
&\leq& \epsilon
+ \frac{\ln (e\nb)}{\nb} \min_{\lambda \in \Lambda} \sum_{t=0}^{\nb-2} \bar{G}_t(\lambda, \scal, \bias)
+ \frac{16}{\nb} \sum_{t=2}^{\nb} \sqrt{\frac{K\ln\frac{8\nb|\Lambda|}{\delta}}{(t-1)\epsilon}}
+ \sqrt{\frac{2 \ln\frac{8|\Pi|}{\delta} }{\nb}}
+ \frac1{\nb} \\
&\leq& \epsilon
+ \frac{\ln (e\nb)}{\nb} \min_{\lambda \in \Lambda} \sum_{t=1}^{\nb} \bar{G}_t(\lambda, \scal, \bias)
+ \frac{16}{\nb} \sum_{t=2}^{\nb} \sqrt{\frac{K\ln\frac{8\nb|\Lambda|}{\delta}}{(t-1)\epsilon}}
+ \sqrt{\frac{2 \ln\frac{8|\Pi|}{\delta} }{\nb}}
+ \frac{\ln(e^2 \nb)}{\nb} \\
&\leq& \epsilon
+ 3 \sqrt{\frac{\ln\frac{8 \nb |\Pi|}{\delta} }{\nb}}
+ 32 \sqrt{\frac{K \ln\frac{8\nb|\Lambda|}\delta}{\nb \epsilon}}
+ \frac{\ln (e\nb)}{\nb} \min_{\lambda \in \Lambda} \sum_{t=1}^{\nb} \bar{G}_t(\lambda, \scal, \bias).
\end{eqnarray*}
where the first inequality is from Equations~\eqref{eqn:hoeff-best} and~\eqref{eqn:azuma-pit} and dividing both sides by $\nb$; the second inequality is from Equation~\eqref{eqn:pregret}; the third inequality is from that $\bar{G}_0(\lambda, \scal, \bias) \leq 1$;
the fourth inequality is from algebra, and our assumption that $\delta < 1/e$. The theorem follows.
\end{proof}

The following lemma upper bounds the excess cost of $\pi_t^\lambda = \argmin_{\pi \in \Pi} \hat{\risk}_{\lambda,t-1}(\pi)$.


\begin{lemma}
\label{lem:bern-pi}
For every $t \in \cbr{1,\ldots,\nb}$, there exists an event $E_t$ with probability $1-\frac\delta{4\nb}$, such that the following holds
for all $\lambda$ in $\Lambda$:
\[
\E \cb(\pi_t^\lambda(x)) - \E \cb(\pi^*(x)) \leq \bar{G}_{t-1}(\lambda, \scal, \bias).
\]
\end{lemma}
\begin{proof}
Define event $E_t$ as: for all $\pi$ in $\Pi$,
\begin{eqnarray}
&& \abs{ [\lambda (t-1) \E \cb(\pi(x)) + (1-\lambda) \ns \E \cs(\pi(x))] - [\lambda (t-1) \E_{t-1} \hcb(\pi(x)) + (1-\lambda) \ns \E_{\Ss} \cs(\pi(x))] } \nonumber \\
&\leq&
2 \sqrt{ (\lambda^2 \frac{K(t-1)}{\epsilon} + (1-\lambda)^2\ns) \ln \frac{8\nb|\Pi|}{\delta} } + ( \frac{\lambda K }{\epsilon} + (1-\lambda)) \ln \frac{8\nb|\Pi|}{\delta}
\label{eqn:freedman-pi}
\end{eqnarray}

In other words,
\begin{eqnarray}
\abs{ \unrisk_{\lambda,t-1}(\pi) - \hat{\unrisk}_{\lambda,t-1}(\pi) }
\leq
V_{t-1}(\lambda).
\label{eqn:lambda-risk-conc}
\end{eqnarray}

For a fixed $\pi$, applying Lemma~\ref{lem:freedman} with $X_i = (1-\lambda)(\cs_{\nb+i}(\pi(x_{\nb+i}) - \E \cs(\pi(x)))$
for $i$ in $\cbr{1,\ldots,\ns}$,
$X_i = \lambda(\hcb_{i-\ns}(\pi(x_{i-\ns})) - \E \cb(\pi(x)))$ for $i$ in $\cbr{\ns+1, \ldots, \ns+t}$, and $M = (1-\lambda) + \frac{\lambda K}{\epsilon}$, and noting that
$|X_i| \leq M$ almost surely for all $i$,
$\E[X_i^2|\calB_{i-1}] \leq (1-\lambda)^2$ for $i$ in $\cbr{1,\ldots,\ns}$, and
$\E[X_i^2|\calB_{i-1}] \leq \lambda^2 \E[\frac{1}{p_{i-\ns,\pi(x_{i-\ns})}}|\calB_{i-1}] \leq
\frac{\lambda^2 K}{\epsilon}$
for $i$ in $\cbr{\ns+1,\ldots,\ns+t-1}$,
we get that Equation~\eqref{eqn:freedman-pi}
holds for $\pi$ with probability $1-\frac{\delta}{4\nb|\Pi|}$. Therefore, by an union bound over all $\pi$ in $\Pi$, $E_t$ happens with probability $1-\frac\delta{4\nb}$. We henceforth condition on $E_t$ happening.

By the optimality of $\pi_t^\lambda$,
\[
\hat{\risk}_{\lambda,t-1}(\pi_t^\lambda) \leq \hat{\risk}_{\lambda,t-1}(\pi^*).
\]
Equivalently,
\[
\hat{\unrisk}_{\lambda,t-1}(\pi_t^\lambda) \leq \hat{\unrisk}_{\lambda,t-1}(\pi^*).
\]

Combining with Equation~\eqref{eqn:lambda-risk-conc} applied to $\pi_t^\lambda$ and $\pi^*$, we get that
\begin{eqnarray*}
\unrisk_{\lambda,t-1}(\pi_t^\lambda) - \unrisk_{\lambda,t-1}(\pi^*)
\leq
2V_{t-1}(\lambda).
\end{eqnarray*}

Using the $(\scal,\bias)$-similarity of $\Ds$ to $\Db$, and Lemma~\ref{lem:similar-weighted} below, we get that
\[
(\E \cb(\pi_t^\lambda(x)) - \E \cb(\pi^*(x)))(\lambda (t-1) + (1-\lambda) \ns \scal) \leq 2 V_{t-1}(\lambda, \scal) + (1-\lambda) \ns \bias.
\]
Therefore, by the definition of $G_t(\lambda, \scal, \bias)$, we have that $\E \cb(\pi_t^\lambda(x)) - \E \cb(\pi^*(x))) \leq G_{t-1}(\lambda, \scal, \bias)$.
Combining the above with the fact that $\E \cb(\pi_t^\lambda(x)) - \E \cb(\pi^*(x)) \leq 1$, the lemma follows.
\end{proof}

\begin{lemma}
\label{lem:similar-weighted}
If $\Ds$ is $(\scal, \bias)$-similar to $\Db$, then for any policy $\pi$,
\[
(\E \cb(\pi(x)) - \E \cb(\pi^*(x)))(\lambda t + (1-\lambda) \ns \scal)
\leq
(\unrisk_{\lambda, t}(\pi) - \unrisk_{\lambda, t}(\pi^*))
+
(1-\lambda) \ns \bias.
\]
\end{lemma}
\begin{proof}
Using the definition of $\unrisk_{\lambda, t}$, we have that
\[ \unrisk_{\lambda, t}(\pi) - \unrisk_{\lambda, t}(\pi^*) = \lambda t (\E \cb(\pi(x)) - \E \cb(\pi^*(x))) + (1-\lambda) \ns (\E \cs(\pi(x)) - \E \cs(\pi^*(x))). \]
Applying Lemma~\ref{lem:similar}, the right hand side is at least
\[
 \lambda t (\E \cb(\pi(x)) - \E \cb(\pi^*(x))) + (1-\lambda) \ns (\scal (\E \cb(\pi(x)) - \E \cb(\pi^*(x))) - \bias ).
\]
The lemma follows immediately by algebra.
\end{proof}



We also bound the cost overhead for selecting $\lambda_t$ from $\Lambda$ compared to using the best
$\lambda$ in hindsight. Define the progressive validation error using $\cbr{\pi_\tau^\lambda}_{\tau \leq t}$ as:
$ \hat{C}_{\lambda,t} = \sum_{\tau=1}^t \hcb_\tau(\pi_\tau^\lambda(x_\tau)) $,
and its expectation as:
$ C_{\lambda,t} = \sum_{\tau=1}^t \E \cb(\pi_\tau^\lambda(x)) $.

\begin{lemma}
\label{lem:bern-pi-lambda-t}
For every $t \in \cbr{2,\ldots,\nb}$, there exists an event $F_t$ with probability $1-\frac\delta{4\nb}$, such that $\lambda_t$ has the following property:
\[
\sum_{\tau=1}^{t-1} \E \cb (\pi_\tau^{\lambda_t}(x)) - \min_{\lambda \in \Lambda} \sum_{\tau=1}^{t-1} \E \cb (\pi_\tau^\lambda(x))
\leq
16 \sqrt{\frac{K(t-1)}{\epsilon} \ln\frac{8\nb|\Lambda|}{\delta}}.
 \]
\end{lemma}

\begin{proof}
	Define event $F_t$ as: for all $\lambda$ in $\Lambda$,
	\[
	\abs{ C_{\lambda,t-1} - \hat{C}_{\lambda,t-1} } \leq 2\sqrt{ \frac{K(t-1)}{\epsilon} \ln \frac{8\nb|\Lambda|}{\delta} } + \frac{K}{\epsilon} \ln \frac{8\nb|\Lambda|}{\delta}.
	\]
	In other words,
	\begin{eqnarray}
	 \abs{ \sum_{\tau=1}^{t-1} \E \cb(\pi_\tau^\lambda(x)) - \sum_{\tau=1}^{t-1} \hcb_\tau(\pi_\tau^\lambda(x_\tau)) }
	\leq
	2\sqrt{\frac{K(t-1)}{\epsilon} \ln \frac{8\nb|\Lambda|}{\delta} } + \frac{K}{\epsilon} \ln \frac{8\nb|\Lambda|}{\delta}.
	\label{eqn:freedman-pi-s-lambda}
	\end{eqnarray}

For a fixed $\lambda$, by Lemma~\ref{lem:freedman}, taking
$X_i = 0$ for $i$ in $[\ns]$,
$X_i = \hcb_{i-\ns}(\pi_{i-\ns}^\lambda(x_{i-\ns})) - \E \cb(\pi_{i-\ns}^\lambda(x))$ for $i$ in
$\cbr{\ns+1,\ldots,\ns+t-1}$, $M = \frac{K}{\epsilon}$,
and noting that
$|X_i| \leq M$ almost surely for all $i$,
$\E[X_i^2|\calB_{i-1}] = 0$ for $i$ in $[\ns]$, and
$\E[X_i^2|\calB_{i-1}] \leq \E[\frac{1}{p_{i-\ns,\pi(x_{i-\ns})}}|\calB_{i-1}] \leq \frac{K}{\epsilon}$ for $i$ in $\cbr{\ns+1,\ldots,\ns+t-1}$,
we get that Equation~\eqref{eqn:freedman-pi-s-lambda} holds with probability $1-\frac{\delta}{4\nb|\Lambda|}$. By an union bound over all $\lambda$ in $\Lambda$, the probability of $F_t$
is at least $1-\frac\delta{4\nb}$. We henceforth condition on $F_t$ happening.

By the optimality of $\lambda_t$, we know that for all $\lambda$ in $\Lambda$,
\[
\hat{C}_{\lambda_t,t-1} \leq \hat{C}_{\lambda,t-1}.
\]
Combining the above inequality with Equation~\eqref{eqn:freedman-pi-s-lambda} applied on $\lambda_t$ and $\lambda$, we get that
\[
C_{\lambda_t,t-1} - C_{\lambda,t-1}
\leq
4\sqrt{\frac{K(t-1)}{\epsilon} \ln\frac{8\nb|\Lambda|}{\delta}}
+ 2\frac{K}{\epsilon} \ln\frac{8\nb|\Lambda|}{\delta}.
\]
In addition, observe that $C_{\lambda_t,t-1} - C_{\lambda,t-1} \leq t-1$ as $\cb \in [0,1]^K$ with probability 1. Combining the above facts with Lemma~\ref{lem:cap} below, we have
that
\[
C_{\lambda_t,t-1} - C_{\lambda,t-1}
\leq
\min\del{4\sqrt{\frac{K(t-1)}{\epsilon} \ln\frac{8\nb|\Lambda|}{\delta}}
+ 2\frac{K}{\epsilon} \ln\frac{8\nb|\Lambda|}{\delta}, t-1}
\leq
16 \sqrt{\frac{K(t-1)}{\epsilon} \ln\frac{8\nb|\Lambda|}{\delta}}.
\]
In other words,
\[
\sum_{\tau=1}^{t-1} \E \cb (\pi_\tau^{\lambda_t}(x)) - \sum_{\tau=1}^{t-1} \E \cb (\pi_\tau^\lambda(x))
\leq
16 \sqrt{\frac{K(t-1)}{\epsilon} \ln\frac{8\nb|\Lambda|}{\delta}}.
 \]
As the above holds for any $\lambda$ in $\Lambda$, the lemma follows.
\end{proof}

\begin{lemma}
\label{lem:cap}
For any positive real numbers $a, b > 0$, we have
\[ \min(\sqrt{ab}+b, a) \leq 2\sqrt{ab}. \]
\end{lemma}
\begin{proof}
The lemma follows from the straightforward calculations below:
\[ \min(\sqrt{ab}+b, a) \leq \min(\sqrt{ab}, a) + \min(b,a) \leq \sqrt{ab} + \sqrt{ab} = 2\sqrt{ab}.
\qedhere
\]
\end{proof}

\section{Proof of Theorem~\ref{thm:supervised-gt}}
\label{app:supervised-gt}
We begin with some notation used in our analysis.
Throughout this section, we	let $\pi^* = \argmin_{\pi \in \Pi} \E \cs(\pi(x)) $, the optimal policy in $\Pi$ with respect to $\Ds$. Define
$$H_t(\lambda, \scal, \bias) = \frac{\lambda \bias + 2W_t(\lambda)}{(1-\lambda) + \lambda \scal}.$$
For policy $\pi$,
$\lambda$ in $[0,1]$ and $t_e$ for $e \in \{1,2,\ldots,E\}$, define the $\lambda$-weighted empirical cost\footnote{Note that previously we used a cost of $\lambda$ per example, whereas now it is $\lambda$ per source, implying that the per example costs are $\lambda/t_e$ and $(1-\lambda)/\ns$ after $t_e$ CB examples.} of $\pi$
as
\[ \hat{M}_{\lambda,t_e}(\pi) = \lambda \E_{t_e} \hat{c}(\pi(x)) + (1-\lambda) \E_{\Str} \cs(\pi(x)) \]
and its expectation
\[ M_{\lambda,t_e}(\pi) = \lambda \E \cb(\pi(x)) + (1-\lambda) \E \cs(\pi(x)) \]

For convenience, we define $\tns = \ns/(E+1)$.

Denote by $(x_{\nb+1}, \cs_{\nb+1}), \ldots, (x_{\nb+\tns}, \cs_{\nb+\tns})$ an enumeration of the elements in $\Str$.
Define filtration $\cbr{\calB_t}_{t=0}^{\nb+\tns}$ as follows: $\calB_0$ is the trivial $\sigma$-algebra, and
\[
\calB_t =
\begin{cases}
	\sigma((x_{\nb+1}, \cs_{\nb+1}), \ldots, (x_{\nb+1}, \cs_{\nb+t})), & t \in \cbr{0,\ldots,\tns}, \\
	\sigma(\Str, (x_1, \hcb_1), (x_{t-\tns}, \hcb_{t-\tns})), & t \in \cbr{\tns+1,\ldots,\tns+\nb}.
  \end{cases}
\]

For reader's convenience, we also recall our earlier notation:
\begin{align*}
  W_t(\lambda) &= \textstyle 2 \sqrt{ \left(\frac{\lambda^2 K }{t \epsilon} + \frac{(1-\lambda)^2(E+1)}{\ns}  \right) \ln\frac{8E |\Pi|}{\delta} } +  \textstyle\left(\frac{\lambda K}{t \epsilon} + \frac{(1-\lambda)(E+1)}{\ns}\right) \ln\frac{8E |\Pi|}{\delta}.
\end{align*}

\begin{proof}[Proof of Theorem~\ref{thm:supervised-gt}]
	Define event $I$ as: for all $\pi$ in $\Pi$,
	\begin{equation}
		  \abs{ \frac{1}{\nb}\sum_{t=1}^{\nb} \E[\cs(\pi(x_t))|x_t] - \E \cs(\pi(x)) } \leq \sqrt{\frac{ \ln\frac{8|\Pi|}{\delta}}{2 \nb}}.
	\end{equation}
    Intuitively, under this event our regret measure for a policy is close to its expected value.
	By Hoeffding's inequality and union bound, $I$ happens with probability $1-\frac \delta 4$.
	Specifically, on event $I$, as $\pi^*$ is the policy in $\Pi$ that minimizes $\E \cs(\pi(x))$, we have that for all $\pi$ in $\Pi$,
  \[
     \E \cs(\pi^*(x)) - \frac{1}{\nb}\sum_{t=1}^{\nb} \E[\cs(\pi(x_t))|x_t] \leq \E \cs(\pi(x)) - \frac{1}{\nb}\sum_{t=1}^{\nb} \E[\cs(\pi(x_t))|x_t] \leq \sqrt{ \frac{\ln\frac{8|\Pi|}{\delta}}{2 \nb}}.
  \]
  Therefore,
	\begin{equation}
		 \E \cs(\pi^*(x)) - \min_{\pi \in \Pi} \frac{1}{\nb}\sum_{t=1}^{\nb} \E[\cs(\pi(x_t))|x_t] \leq \sqrt{ \frac{\ln\frac{8|\Pi|}{\delta}}{2 \nb}}.
		 \label{eqn:i-sup-gt}
	\end{equation}

  Recall that randomized policy $\pi_t: \calX \to \Delta^{K-1}$ is defined as
	\begin{equation}
		\pi_t(x) = (1 - \epsilon) \pi_e^{\lambda_e}(x) + \frac\epsilon K \one_K,
		\label{eqn:pi-t-sup-gt}
	\end{equation}
  for $e$ in $\cbr{1,2,\ldots,E-1}$, $t \in (t_e, t_{e+1}]$,
	and $\pi_t(x) = \frac{1}{K} \one_K$ for all $t$ in $(t_0, t_1]$.
	With a slight abuse of notation, denote by $\E \cs(\pi_t(x)) := \E_{(x,\cs),a\sim\pi_t(x)} \cs(a)$.
	 Observe that for $t$ in $(t_e, t_{e+1}]$, $\E[\E[\cs(a_t)|x_t]|\calB_{\tns+t-1}] = \E \cs(\pi_t(x))$.
	Define event $J$ as:
	\begin{equation}
		\abs{ \frac{1}{\nb}\sum_{t=1}^{\nb} \E[\cs(a_t)|x_t] - \frac{1}{\nb} \sum_{t=1}^{\nb} \E \cs(\pi_t(x)) } \leq \sqrt{\frac{\ln\frac{8}{\delta}}{2\nb}}.
		\label{eqn:j-sup-gt}
	\end{equation}
	By Azuma's inequality, $J$ happens with probability $1-\frac \delta 4$.

	Denote by $E$ the event that the events $E_e$, $F_e$ defined in Lemmas~\ref{lem:bern-pi-sup-gt} and~\ref{lem:bern-pi-lambda-e-sup-gt} (both given below) and $I$, $J$ hold simultaneously for all $t$. By union bound over all $E_e,F_e$'s and $I,J$, event $E$ happens with probability $1-\delta$.
	We henceforth condition on $E$ happening.

	Consider $e$ in $\cbr{1,2,\ldots,E-1}$ and $t$ in $(t_e, t_{e+1}]$. We now upper bound the expected excess cost of using randomized prediction $\pi_t$ as defined in Equation~\eqref{eqn:pi-t-sup-gt}. By the definition of $\pi_t$, we get
	\begin{equation}
		\E \cs(\pi_t(x)) = (1-\epsilon) \E \cs(\pi_e(x)) + \epsilon \frac1K \sum_{a=1}^K \E \cs(a)  \leq \E \cs(\pi_e(x)) + \epsilon.
		\label{eqn:eps-greedy}
	\end{equation}

	Now, combining the above inequality with Lemmas~\ref{lem:bern-pi-sup-gt} and~\ref{lem:bern-pi-lambda-e-sup-gt}, we have that for all $t$ in $(t_e, t_{e+1}]$,
	\begin{eqnarray*}
		 \E \cs(\pi_t(x))
		&\leq& \min_{\lambda \in \Lambda} \E \cs(\pi_e^\lambda(x)) + \epsilon + \sqrt{\frac{2\ln\frac{8E|\Lambda|}{\delta}}{\tns}} \\
		&\leq& \E \cs(\pi^*(x)) + \epsilon + \sqrt{\frac{2\ln\frac{8E|\Lambda|}{\delta}}{\tns}} + \min_{\lambda \in \Lambda} H_{t_e}(\lambda, \scal, \bias).
	\end{eqnarray*}

	Summing the above inequality over all $t=t_1+1,\ldots,\nb$, grouping by epoch $e$,
  and using the fact that $0 \leq \E \cs(\pi_t(x)) \leq 1$ for $t \leq t_1 = 2$, we have
	\[
	\sum_{t=t_1+1}^{\nb} \E \cs(\pi_t(x))
  \leq
	\nb \E \cs(\pi^*(x)) + 2 + \nb \epsilon + (\nb - 2) \sqrt{\frac{2\ln\frac{8E|\Lambda|}{\delta}}{\tns}} + \sum_{e=1}^{E-1} \sum_{t=t_e+1}^{t_{e+1}} \min_{\lambda \in \Lambda} H_{t_e}(\lambda, \scal, \bias).
	\]
	Dividing both sides by $\nb$ and some algebra yields
	\[
	\frac{1}{\nb} \sum_{t=1}^{\nb} \E \cs(\pi_t(x))
  \leq
	\frac2{\nb} + \E \cs(\pi^*(x)) + \epsilon + \sqrt{\frac{2\ln\frac{8E|\Lambda|}{\delta}}{\tns}} + \frac{1}{\nb} \sum_{e=1}^{E-1} \sum_{t=t_e+1}^{t_{e+1}} \min_{\lambda \in \Lambda} H_{t_e}(\lambda, \scal, \bias).
	\]

	Note that for every $t$ in $[t_e+1, t_{e+1}]$, as $t \leq t_{e+1} = 2 t_e$, $W_t(\lambda) \geq \frac12 W_{t_e}(\lambda)$, hence $H_t(\lambda, \scal, \bias) \geq \frac12 H_{t_e}(\lambda, \scal, \bias)$. Therefore, the right hand side can be further upper bounded by
	\begin{eqnarray*}
		&& \frac2{\nb} + \E \cs(\pi^*(x)) + \epsilon + \sqrt{\frac{2\ln\frac{8E|\Lambda|}{\delta}}{\tns}} + \frac2{\nb}\sum_{e=1}^{E-1} \sum_{t=t_e+1}^{t_{e+1}} \min_{\lambda \in \Lambda} H_t(\lambda, \scal, \bias) \\
		&\leq& \frac{2}{\nb} + \E \cs(\pi^*(x)) + \epsilon + \sqrt{\frac{2\ln\frac{8E|\Lambda|}{\delta}}{\tns}} + \min_{\lambda \in \Lambda} \frac2{\nb}\sum_{t=1}^{\nb} H_t(\lambda, \scal, \bias).
	\end{eqnarray*}
	where the inequality is from that for any set of functions $\cbr{N_t(\cdot)}_{t=1}^{\nb}$, $\sum_{t=1}^{\nb} \min_{\lambda \in \Lambda} N_t(\lambda) \leq \min_{\lambda \in \Lambda} [ \sum_{t=1}^{\nb} N_t(\lambda) ]$.

	To summarize, we have that
	\[
	\frac{1}{\nb} \sum_{t=1}^{\nb} \E \cs(\pi_t(x)) - \E \cs(\pi^*(x)) \leq \frac{2}{\nb} + \epsilon + \sqrt{\frac{2\ln\frac{8E|\Lambda|}{\delta}}{\tns}} + \min_{\lambda \in \Lambda} \frac2{\nb}\sum_{t=1}^{\nb} H_t(\lambda, \scal, \bias).
	\]
	Combining with Equations~\eqref{eqn:i-sup-gt} and~\eqref{eqn:j-sup-gt} and using some algebra, we have
	\[
   \frac{1}{\nb}\sum_{t=1}^{\nb} \E[\cs_t(\pi_t(x_t))|x_t] - \min_{\pi \in \Pi} \frac{1}{\nb}\sum_{t=1}^{\nb} \E[\cs(\pi(x_t))|x_t]
	 \leq
	 \epsilon + 3\sqrt{\frac{\ln\frac{8|\Pi|}{\delta}}{\nb}} +
	\sqrt{\frac{2\ln\frac{8E|\Lambda|}{\delta}}{\tns}} + \min_{\lambda \in \Lambda} \frac2{\nb}\sum_{t=1}^{\nb} H_t(\lambda, \scal, \bias).
	\]
	The theorem follows from the definition of $\tns$.
\end{proof}

\begin{lemma}
For every $e$, there exists an event $E_e$ with probability $1-\frac \delta {4 E}$, on which for all $\lambda$ in $\Lambda$, the excess cost of $\pi_e^\lambda$ can be bounded as:
\[ \E \cs(\pi_e^\lambda(x)) - \E \cs(\pi^*(x)) \leq H_{t_e}(\lambda, \scal, \bias). \]
\label{lem:bern-pi-sup-gt}
\end{lemma}

\begin{proof}
We first show the following concentration inequality: with probability $1-\frac\delta{4 E}$, for all $\pi$ in
$\Pi$,
\begin{equation}
	\abs{ \hat{M}_{\lambda,t_e}(\pi) - M_{\lambda,t_e}(\pi) } \leq W_{t_e}(\lambda).
	\label{eqn:conc-dataset-wt-simp}
\end{equation}

To show the above statement, in light of the definitions of $\hat{M}, M$ and $W$, it suffices to show that for every $\pi$ in $\Pi$, with probability
$1-\frac\delta {4E|\Pi|}$, we have
\begin{eqnarray}
	&& \abs{ \sum_{t=1}^{t_e} \frac{\lambda}{t_e} [\hat{c}_t(\pi(x_t)) - \E c(\pi(x))] + \sum_{(x,\cs) \in S} \frac{1-\lambda}{\tns} [\cs(\pi(x)) - \E \cs(\pi(x))] } \nonumber \\
	&\leq& 2\sqrt{ (\frac{\lambda^2 K }{t_e \epsilon} + \frac{(1-\lambda)^2}{\tns}  ) \ln\frac{8E|\Pi|}{\delta} } + (\frac{\lambda K}{t_e \epsilon} + \frac{1-\lambda}{\tns}) \ln\frac{8E|\Pi|}{\delta}.
	\label{eqn:conc-dataset-wt}
\end{eqnarray}

For a fixed $\pi$, applying Lemma~\ref{lem:freedman} with $X_i = \frac{1-\lambda}{\tns}(\cs_{\nb+i}(\pi(x_{\nb+i})) - \E \cs(\pi(x)))$
for $i$ in $\cbr{1,\ldots,\tns}$,
$X_i = \frac{\lambda}{t_e}(\hcb_{i-\tns}(\pi(x_{i-\tns})) - \E \cb(\pi(x)))$ for $i$ in $\cbr{\tns+1, \ldots, \tns+t_e}$, and $M = \frac{1-\lambda}{\tns} + \frac{\lambda K}{t_e \epsilon}$, and note that
$|X_i| \leq M$ almost surely for all $i$,
$\E[X_i^2|\calB_{i-1}] \leq (\frac{1-\lambda}{\tns})^2$ for $i$ in $\cbr{1,\ldots,\tns}$, and
$\E[X_i^2|\calB_{i-1}] \leq (\frac{\lambda}{t_e})^2 \E[\frac{1}{p_{i-\tns,\pi(x_{i-\tns})}}|\calB_{i-1}] \leq
\frac{\lambda^2 K}{t_e^2 \epsilon}$ for $i$ in $\cbr{\tns+1,\ldots,\tns+t_e}$,
we get that Equation~\eqref{eqn:conc-dataset-wt} holds with probability $1-\frac\delta{4E|\Pi|}$.

Therefore, by union bound over all $\pi$ in $\Pi$ Equation~\eqref{eqn:conc-dataset-wt-simp} holds for all $\lambda$ simultaneously with probability $1-\frac\delta{4E}$. As
$\pi_e^\lambda$ minimizes $\hat{M}_{\lambda,t}(\pi)$ over $\Pi$, we have that
$\hat{M}_{\lambda,t}(\pi_e^\lambda) \leq \hat{M}_{\lambda,t}(\pi^*)$. Combining this fact with Equation~\eqref{eqn:conc-dataset-wt-simp} on $\pi_e^\lambda$ and $\pi^*$, we get
\[ M_{\lambda,t_e}(\pi_e^\lambda) - M_{\lambda,t_e}(\pi^*) \leq 2W_{t_e}(\lambda). \]

Observe that Lemma~\ref{lem:similar} and the definition of $M$ implies that the left hand side
of the above equation is at least $(\lambda \scal + (1-\lambda))(\E\cs(\pi_e^\lambda(x)) - \E\cs(\pi^*(x)) ) - \lambda \bias$. The lemma statement follows by straightforward algebra and the definition of $H_t(\cdot,\cdot,\cdot)$.
\end{proof}

\begin{lemma}
For every $e$, there exists an event $F_e$ with probability $1-\frac{\delta}{4E}$, on which the policy $\pi_e^{\lambda_e}$ satisfies that
\[ \E \cs(\pi_e^{\lambda_e}) - \min_{\lambda \in \Lambda} \E \cs(\pi_e^\lambda(x)) \leq \sqrt{\frac{2\ln\frac{8E|\Lambda|}{\delta}}{\tns}}. \]
\label{lem:bern-pi-lambda-e-sup-gt}
\end{lemma}

\begin{proof}
Given any $\lambda$ in $\Lambda$, as $\Sde_e$ is a sample independent of the $\pi_e^\lambda$,
we have by Hoeffding's inequality that with probability $1-\frac{\delta}{4E|\Lambda|}$,
\begin{equation}
	\abs{ \E_{\Sde_e} \cs(\pi_e^\lambda(x)) - \E \cs(\pi_e^\lambda(x)) } \leq \sqrt{\frac{\ln\frac{8E|\Lambda|}{\delta}}{2 \tns}}.
	\label{eqn:sup-validation}
\end{equation}
By union bound, with probability $1-\frac \delta 4$, Equation~\eqref{eqn:sup-validation} holds for all
$\lambda$ in $\Lambda$ simultaneously.

Observe that by the optimality of $\lambda_e$, for all $\lambda$ in $\Lambda$, we have that
\[ \E_{\Sde_e} \cs(\pi_e^{\lambda_e}(x)) \leq \E_{\Sde_e} \cs(\pi_e^{\lambda}(x)). \]

Combining with Equation~\eqref{eqn:sup-validation} on $\lambda$ and $\lambda_e$, we get
\[ \E \cs(\pi_e^{\lambda_e}(x)) \leq \E \cs(\pi_e^{\lambda}(x)) + \sqrt{\frac{2 \ln\frac{8E|\Lambda|}{\delta}}{\tns}}. \]

The lemma follows as the above inequality holds for every $\lambda$ in $\Lambda$.
\end{proof}

\section{Fixed choice of $\lambda$ in~\autoref{alg:twosources-sup-gt}}
\label{sec:one-lambda-sup-gt}

Recall that $$H_t(\lambda, \scal, \bias) = \frac{\lambda \bias + 2W_t(\lambda)}{(1-\lambda) + \lambda \scal},$$
and $$W_t(\lambda) = \textstyle 2 \sqrt{ \left(\frac{\lambda^2 K }{t \epsilon} + \frac{(1-\lambda)^2(E+1)}{\ns}  \right) \ln\frac{8E |\Pi|}{\delta} }
+\textstyle\left(\frac{\lambda K}{t \epsilon} + \frac{(1-\lambda)(E+1)}{\ns}\right) \ln\frac{8E |\Pi|}{\delta}.$$

In addition, recall that $\frac{2}{\nb} \sum_{t=1}^{\nb} H_t(\lambda, \scal, \bias)$ is the last term of the regret bound~\autoref{eqn:sup-gt-bound}. We have the following proposition, showing that a single choice
of $\lambda$ ensures a small upper bound on this term when $\scal=1$ and $\bias=0$.

\begin{proposition}
Suppose $\lambda_0 = \frac{\nb \epsilon}{\ns K + \nb \epsilon}$. Then,
\[ \frac{2}{\nb} \sum_{t=1}^{\nb} H_t(\lambda_0, 1, 0) = \tilde{O}\bigg(\textstyle \sqrt{ \frac{ K \ln\frac{8E |\Pi|}{\delta} }{K \ns + \nb \epsilon} } + \frac{ K \ln\frac{8E |\Pi|}{\delta} }{K \ns + \nb \epsilon} \bigg). \]
\label{prop:one-lambda-sup-gt}
\end{proposition}

\begin{proof}
We have the following:
\begin{eqnarray*}
 \frac{2}{\nb} \sum_{t=1}^{\nb} H_t(\lambda_0, 1, 0)
&=& \frac{4}{\nb} \sum_{t=1}^{\nb} W_t(\lambda_0) \\
&=& 4 \cdot \frac{1}{\nb} \sum_{t=1}^{\nb} \sbr{ \textstyle 2 \sqrt{ \left(\frac{\lambda_0^2 K }{t \epsilon} + \frac{(1-\lambda_0)^2(E+1)}{\ns}  \right) \ln\frac{8E |\Pi|}{\delta} }
+\textstyle\left(\frac{\lambda_0 K}{t \epsilon} + \frac{(1-\lambda_0)(E+1)}{\ns}\right) \ln\frac{8E |\Pi|}{\delta}. } \\
&\leq& 4 \cdot \sbr{ \textstyle 2 \sqrt{ \left(\frac{\lambda_0^2 K (\ln\nb+1) }{\nb \epsilon} + \frac{(1-\lambda_0)^2(E+1)}{\ns}  \right) \ln\frac{8E |\Pi|}{\delta} }
+\textstyle\left(\frac{\lambda_0 K (\ln\nb+1)}{\nb \epsilon} + \frac{(1-\lambda_0)(E+1)}{\ns}\right) \ln\frac{8E |\Pi|}{\delta}. } \\
&=& \tilde{O}\bigg(\textstyle \sqrt{ \left(\frac{\lambda_0^2 K }{\nb \epsilon} + \frac{(1-\lambda_0)^2}{\ns}\right) \ln\frac{8E |\Pi|}{\delta} }
+\textstyle\left(\frac{\lambda_0 K}{\nb \epsilon} + \frac{(1-\lambda_0)}{\ns}\right) \ln\frac{8E |\Pi|}{\delta} \bigg) \\
&=& \tilde{O}\bigg(\textstyle \sqrt{ \frac{ K \ln\frac{8E |\Pi|}{\delta} }{K \ns + \nb \epsilon} } + \frac{ K \ln\frac{8E |\Pi|}{\delta} }{K \ns + \nb \epsilon} \bigg).
\end{eqnarray*}
Where the first inequality is from that $H_t(\lambda_0, 1, 0) = 2 W_t(\lambda_0)$; the second equality is from the definition of $W_t(\lambda)$; the inequality is by Jensen's inequality;
the third equality is by dropping the logarithmic terms; the last inequality is by plugging in the choice
of $\lambda_0$.
\end{proof}

\section{Approximate optimality of $\Lambda = \cbr{0,1}$}
In this section, we show Proposition~\ref{prop:lambda-0-1}, which justifies that
using $\Lambda = \cbr{0,1}$ achieves a near-optimal regret bound in
Theorem~\ref{thm:bandit-gt}. Recall that
\begin{align*}
\textstyle V_t(\lambda) &= \textstyle2\sqrt{\left(\lambda^2 \frac{K t}{\epsilon} + (1-\lambda)^2 \ns\right) \ln\frac{8\nb|\Pi|}{\delta} } + \textstyle \left(\frac{\lambda K}{\epsilon} + (1-\lambda)\right) \ln \frac{8\nb|\Pi|}{\delta}, \\
\textstyle  G_t(\lambda, \scal, \bias) &= \textstyle\frac{(1-\lambda)\ns \bias + 2V_t(\lambda)}{\lambda t + (1-\lambda) \ns \scal}.
\end{align*}

\begin{proposition}
\[
\min_{\lambda \in \{0,1\}} G_t(\lambda, \scal, \bias) \leq \sqrt{2} \min_{\lambda \in [0,1]} G_t(\lambda, \scal, \bias).
\]
\label{prop:lambda-0-1}
\end{proposition}

\begin{proof}
For any $\lambda$ in $[0,1]$, we give a lower bound on $V_t(\lambda)$. Note that by the fact that $\sqrt{a+b} \geq \sqrt{\frac 1 2} (\sqrt{a} + \sqrt{b})$, we have that
\begin{equation}
   V_t(\lambda) \geq \frac 1 {\sqrt{2}} [(1-\lambda) V_t(0) + \lambda V_t(1)].
   \label{eqn:vt}
\end{equation}
Therefore,
\begin{eqnarray*}
&& G_t(\lambda, \scal, \bias) \\
&\geq& \frac{(1-\lambda)\ns \bias + 2 \cdot \sqrt{\frac 1 2} [(1-\lambda) V_t(0) + \lambda V_t(1)]}{\lambda t + (1-\lambda) \ns \scal} \\
&\geq& \frac 1 {\sqrt{2}} \frac{(1-\lambda)\ns \bias + 2 [(1-\lambda) V_t(0) + \lambda V_t(1)]}{\lambda t + (1-\lambda) \ns \scal} \\
&\geq& \frac 1 {\sqrt{2}} \min(\frac{\ns \Delta + 2V_t(0)}{\ns \alpha}, \frac{2V_t(1)}{t}) \\
&=& \frac 1 {\sqrt{2}} \min_{\lambda \in \cbr{0,1}} G_t(\lambda, \scal, \bias),
\end{eqnarray*}
where the first inequality is from Equation~\eqref{eqn:vt}, the second inequality is by algebra, the third inequality is by the quasi-concavity of the LHS with respect to $\lambda$, and the fact that the minimum of a quasi-concave function over a convex set is attained at its boundary. As the above holds for any $\lambda$ in $[0,1]$, the proposition follows.
\end{proof}

\section{Combining Two Sources in Supervised Learning}
\label{app:minimax}

\begin{proposition}
For every policy class $\Pi$ of VC dimension $d$ and $\Delta \in [0, \frac14]$, $m, n \geq 32 d$,  for any algorithm
 that outputs a policy $\hat{\pi}$ based on $n$ examples from $D^1$ and $m$ examples from $D^2$,
 there exists a pair of distributions $(D^1, D^2)$ such that $D^2$ is $(1,\bias)$-similar to $D^1$, and
 \[\textstyle\E [\E_{D^1} c(\hat{\pi}(x)) - \min_{\pi \in \Pi} \E_{D^1} c(\pi(x)) ] \geq \frac1{16} \min\left(\sqrt{\nofrac{d}{m}}+8\Delta, \sqrt{\nofrac{d}{n}}\right)\]
 where the outer expectation is over draws from $D^1$ and $D^2$, and the algorithm's randomness.
 \label{prop:minimax}
\end{proposition}

The lower bound proof follows a similar strategy as in the classical
classification setting with slight modifications. In order to prove
the bound we make use of Assouad's Lemma. The statement below follows
exactly from Yu~\cite{Yu97}.
\begin{theorem}[Assouad's Lemma]
  Let $d \geq 1$ be an integer and let $\mathcal{F}_d = \{P_\tau \mid
  \tau \in \{-1,+1\}^d \}$ be a class of $2^d$ probability measures indexed
  by binary strings of length $d$. Write $\tau \sim \tau'$ if $\tau$ and
  $\tau'$ differ in only one coordinate, and write $\tau \sim_j
  \tau'$ when that coordinate is the $j^{th}$. Suppose that there are
  $d$ psuedo-distances on $\mathcal{D}$ such that for any $x,y \in
  \mathcal{D}$
  \begin{equation*}
    \rho(x,y) = \sum_{j=1}^d \rho_j(x,y),
  \end{equation*}
  and further that, if $\tau \sim_j \tau'$,
  \begin{equation*}
    \rho_j(\theta(P_\tau),\theta(P_{\tau'})) \geq \alpha.
  \end{equation*}
  Then for any estimator $\widehat{\theta}$,
  \begin{equation*}
    \max_\tau \E_\tau \rho(\widehat{\theta},\theta(P_\tau)) \geq d \cdot
    \frac{\alpha}{2} \min(\|P_\tau \wedge P_{\tau'} \| \mid \tau
    \sim \tau' \}.
  \end{equation*}
	\label{thm:assouad}

\end{theorem}

\begin{proof}[Proof of Proposition~\ref{prop:minimax}]
Since the VC dimension of $\Pi$ is $d$, there exists
 a set $A = \{x_1,\hdots,x_d\}$ such that for any binary
sequence $\tau \in \{-1,+1\}^d$, there exists some function $\pi$ in $\Pi$
such that for all $i \in \cbr{1,\ldots,d}$, $\pi(x_i) =
\tau_i$. We now define two distributions for a binary classification problem, but express them as joint distributions over $(x,c)$ in order to be consistent with the rest of the paper with the understanding that $c$ will have a zero-one cost structure (where $c = (c(-1), c(+1))$ represent the costs of predicting labels $-1$ and $+1$). The two distributions $D^1_\tau$ and $D^2_\tau$ are each uniform over $x_i$, and the conditional distributions over costs are given by
\begin{align*}
  D^1_\tau((1,0) | x = x_i) &= \frac{1}{2} + \tau_i \epsilon\\
  D^1_\tau((0,1) | x = x_i ) &= \frac{1}{2} - \tau_i \epsilon, \quad \mbox{and}
\end{align*}
\begin{align*}
  D^2_\tau((1,0) | x = x_i) &= \frac{1}{2} + \tau_i \epsilon - \tau_i \beta\\
  D^2_\tau((0,1) | x = x_i) &=  \frac{1}{2} - \tau_i \epsilon + \tau_i \beta.
\end{align*}

Where $\epsilon \in [0, \frac 1 2]$, $\beta \in [0, \frac \bias 2]$ are free parameters to be determined later. Clearly $D^2_\tau$ is $(1,\bias)$-strongly similar to $D^1_\tau$, therefore $D^2_\tau$ is $(1,\bias)$-similar to $D^1_\tau$. This can be seen by taking
$c^+ = \E_{D^1}[c|x]$ and $c^- = \E_{D^2}[c|x] - \E_{D^1}[c|x]$, and observing that
$|\E c^-(\pi(x))| \leq \beta \leq \bias / 2$.
Define $P_\tau$ as the product distribution of $n$ copies of $D^1_\tau$ and $m$ copies of $D^2_\tau$, which is the joint distribution of the input examples to the algorithm.
In addition, define $\theta(P_\tau) = \tau$, and $\rho_j(\tau, \tau') = I(\tau_j \neq \tau_j')$. Therefore,
$\rho(\tau, \tau') = \sum_{j=1}^d \rho_j(\tau, \tau') = \sum_{j=1}^d I(\tau_j \neq \tau_j')$ is the
Hamming distance between $\tau$ and $\tau'$. Suppose the algorithm returns a policy $\hat{\pi}$,
there exists a binary sequence $\hat{\tau}$ such that $\hat{\pi}(x_i) = \hat{\tau}_i$.
Observe that for any $\tau$, by the definition of $D_\tau^1$,
\[ \E_{D^1_\tau} c(\hat{\pi}(x)) - \min_{\pi \in \Pi} \E_{D^1_\tau} c(\pi(x)) = \frac{2 \epsilon}{d} \rho(\hat{\tau}, \tau). \]
By Assouad's Lemma (Theorem~\ref{thm:assouad}) with the $\rho$ defined above and $\alpha=1$, we have that there exists some $\tau$ in $\cbr{-1,+1}^d$, such that
\[ \E \rho(\hat{\tau}, \tau) \geq \epsilon \min(\|P_\tau \wedge P_{\tau'} \|_1 \mid \tau \sim \tau') \]
This immediately implies that, for any algorithm that returns $\hat{\pi}$, there exists some $\tau$ such that
\[ \E_{P_\tau} [c(\hat{\pi}(x)) - \min_{\pi \in \Pi} \E_{D^1_\tau} c(\pi(x))] \geq \epsilon \min(\|P_\tau \wedge P_{\tau'} \|_1 \mid \tau \sim \tau'). \]

What remains is to bound $\|P_\tau \wedge P_{\otau}\|_1$
for any binary sequences differing in one coordinate. Recall that
$\|P_\tau \wedge P_{\otau} \| = 1 - \frac{1}{2} \|P_\tau -
P_\otau\|_1$. We will bound
$\frac{1}{2} \|P_\tau - P_\otau\|_1 \leq H(P_\tau,P_\otau)$
using the Hellinger distance
$H^2(P_\tau,P_\otau) = \frac{1}{2} \sum_z (\sqrt{P_\tau(z)} -
\sqrt{P_\otau(z)})^2$. We recall that $P_\tau$ is in fact the product
distribution over $n$ copies of $D^1_\tau$ and $m$ copies of $D^2_\tau$. For product measures
\begin{equation*}
 H^2(P_\tau,P_\otau) \leq \sum_{i=1}^n (D^1_\tau,D^1_\otau) +
 \sum_{i=1}^m(D^2_\tau, D^2_\otau)
\end{equation*}
we need to bound the Hellinger distance for the
biased and unbiased distributions.
\paragraph{Bounding the Hellinger Distance}
We have
\begin{align*}
 H^2(D^1_\tau,D^1_\otau)
 & = \frac{1}{2} \sum_{i=1}^d \sum_{c \in \{(0,1),(1,0)\}}
	 (\sqrt{D^1_\tau(x_i,c)} - \sqrt{D^1_\otau(x_i,c)})^2 \\
 & = \frac{1}{2 d} \sum_{c \in \{(0,1),(1,0)\}} (\sqrt{D^1_\tau(c|x_i)} -
	 \sqrt{D^1_\otau(c|x_i)})^2 \\
 & = \frac{1}{d} H^2(B(\frac{1}{2} + \epsilon),B(\frac{1}{2} -
	 \epsilon)) \\
 & \leq \frac{8}{d} \epsilon^2,
\end{align*}
where the first equality is by the definiton of $H^2$; the second inequality is
from that there is exactly one $i$ such that
$\sum_{c \in \{(0,1),(1,0)\}} (\sqrt{D^1_\tau(x_i,c)} - \sqrt{D^1_\otau(x_i,c)})^2$ is nonzero,
as $\tau \sim \tau'$; in the right hand side of third equality, $B(p)$ is the Bernoulli distribution with mean parameter $p$.
Similarly,
\begin{equation*}
 H^2(D^2_\tau,D^2_\otau) \leq \frac{8}{d} (\epsilon-\beta)^2
\end{equation*}
Hence,
\begin{equation*}
 H^2(P_\tau,P_\otau) \leq \frac{8}{d} \left ( n \epsilon^2 + m
	 (\epsilon-\beta)^2 \right )
\end{equation*}

Therefore, we have
\begin{equation}
 \max_\tau \E_{P_\tau} [c(\hat{\pi}(x)) - \min_{\pi \in \Pi} \E_{D^1_\tau} c(\pi(x))]
 \geq \epsilon \left [ 1 - \sqrt{\frac{8}{d} \left ( n \epsilon^2 + m
		 (\epsilon - \beta)^2 \right )} \right ]
		 \label{eqn:minimax-lb}
\end{equation}

We now consider two separate cases regarding the settings of $m$, $n$, $d$ and $\Delta$.
\paragraph{Case 1: $\sqrt{\frac{d}{n}} \leq \sqrt{\frac{d}{m}} + 4\Delta$.} In this case,
we let $\epsilon = \frac18 \sqrt{\frac d n}$ and $\beta = \frac18 \max(0, \sqrt{\frac d n} - \sqrt{\frac d m}) \in [0, \frac \Delta 2]$. This gives that the right hand side of Equation~\eqref{eqn:minimax-lb} is at least
$\frac18 \sqrt{\frac d n} \cdot \frac12 = \frac1{16} \sqrt{\frac d n}$.

\paragraph{Case 2: $\sqrt{\frac{d}{n}} > \sqrt{\frac{d}{m}} + 4\Delta$.} In this case, we let $\epsilon = \frac 1 8 \sqrt{\frac d m} + \frac \Delta 2$ and $\beta = \frac\Delta2$. This gives that the right hand
side of Equation~\eqref{eqn:minimax-lb} is at least $(\frac 1 8 \sqrt{\frac d m} + \frac \Delta 2) \cdot \frac12 = \frac1{16} (\sqrt{\frac d m} + 4 \Delta)$.

In summary, for any choice of $m, n, d, \Delta$ with $m, n \geq 32d$ and $\Delta \in [0, \frac 1 4]$, we can find a pair of distributions $(D^1_\tau, D^2_\tau)$ such that
\[
\E_{P_\tau} [c(\hat{\pi}(x)) - \min_{\pi \in \Pi} \E_{D^1_\tau} c(\pi(x))]
\geq
\frac1{16} \min(\sqrt{\frac d n}, \sqrt{\frac d m} + 4 \Delta).
\]
The proposition follows.
\end{proof}

\section{Additional Experimental Results}
\label{apx:moreresults}
\begin{figure}[H]
\centering
\begin{tabular}{c | @{}c@{ }c@{ }c@{}} 
\toprule
& \multicolumn{3}{c}{ Ratio }
\\
Noise & 2.875 & 5.75 & 11.5
\\\midrule
Noiseless & \includegraphics[width=0.29\textwidth,valign=c,trim={0 0 0 0},clip]{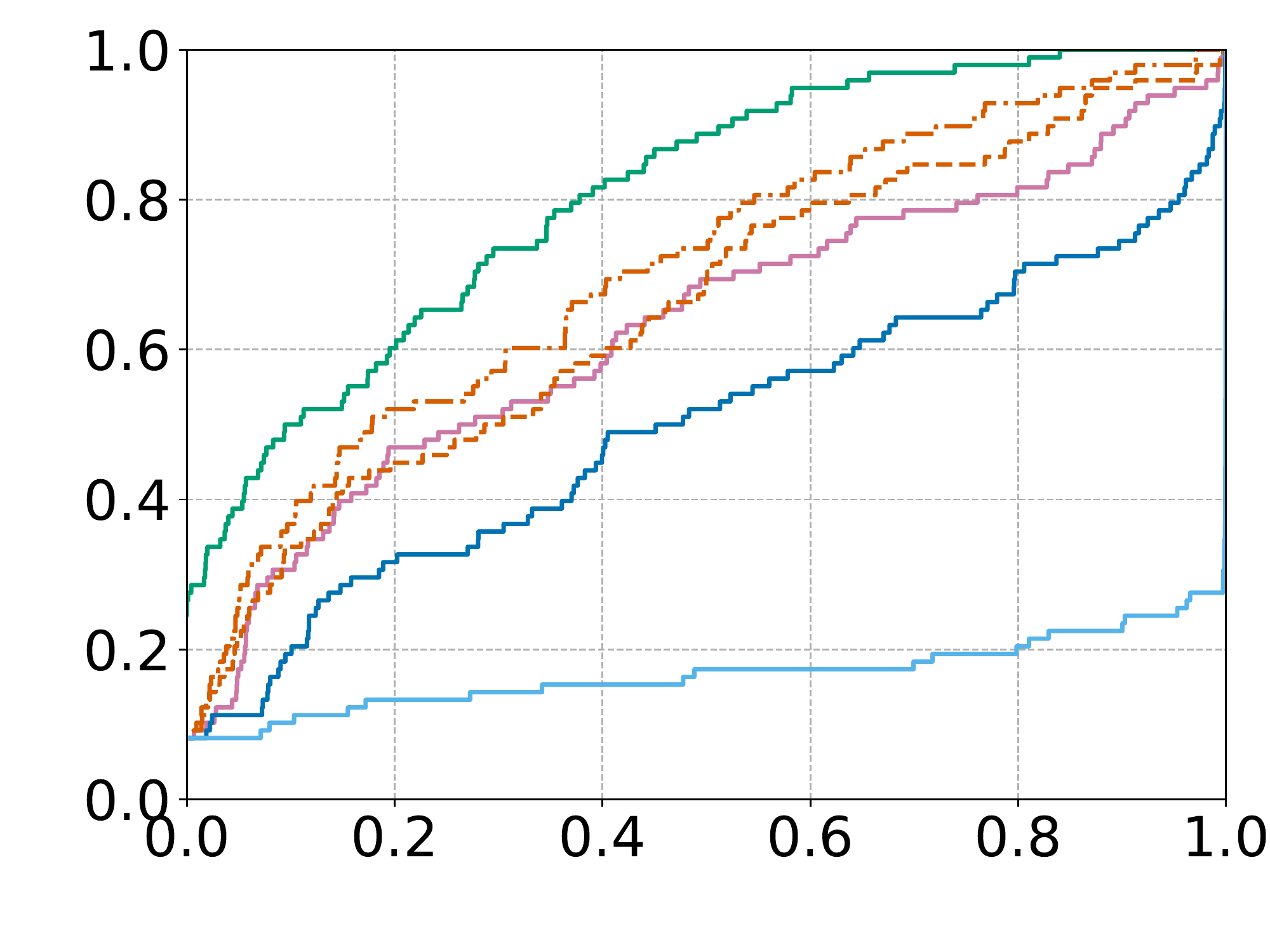}
&
\includegraphics[width=0.29\textwidth,valign=c,trim={0 0 0 0},clip]{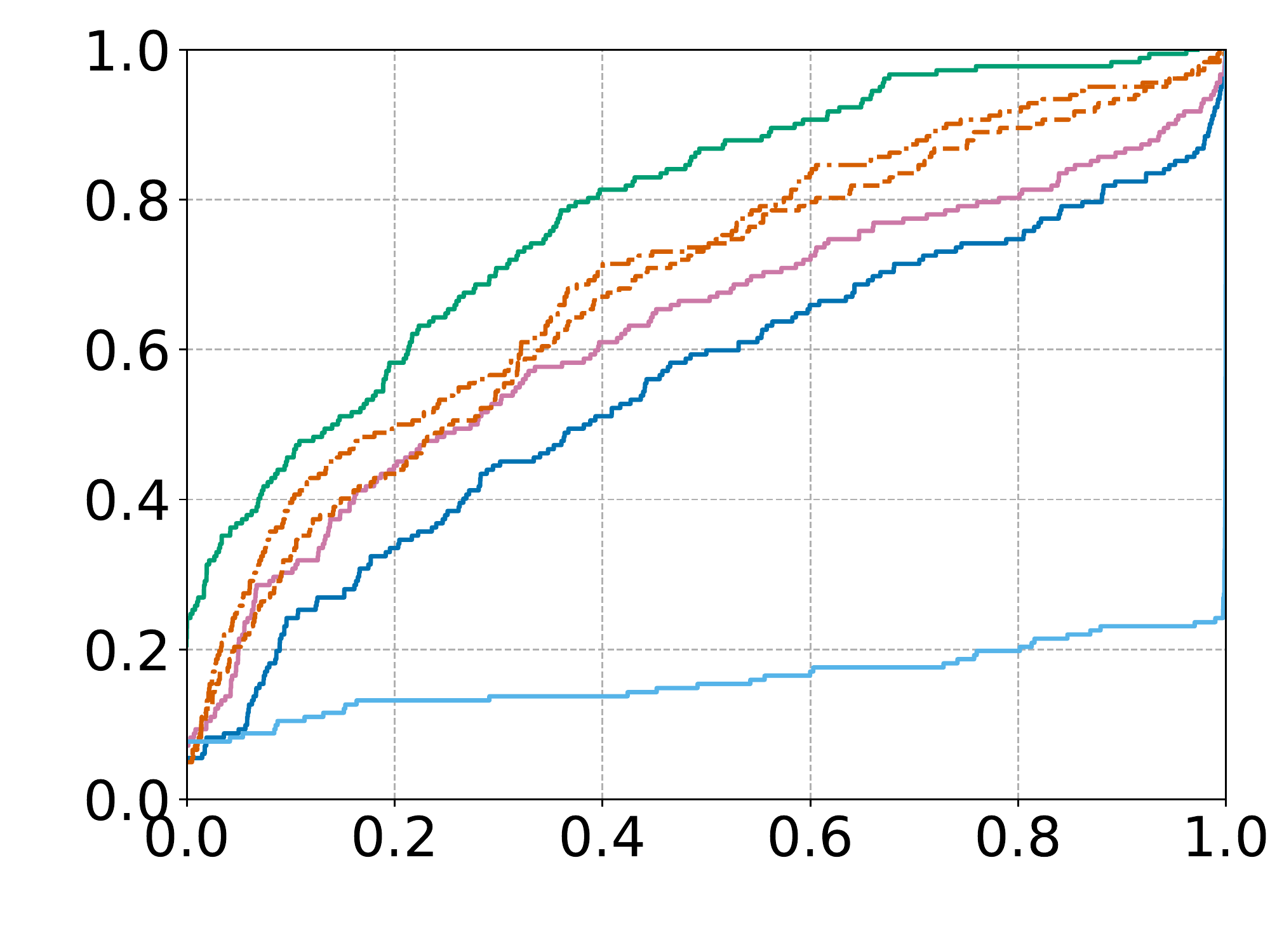}
&
\includegraphics[width=0.29\textwidth,valign=c,trim={0 0 0 0},clip]{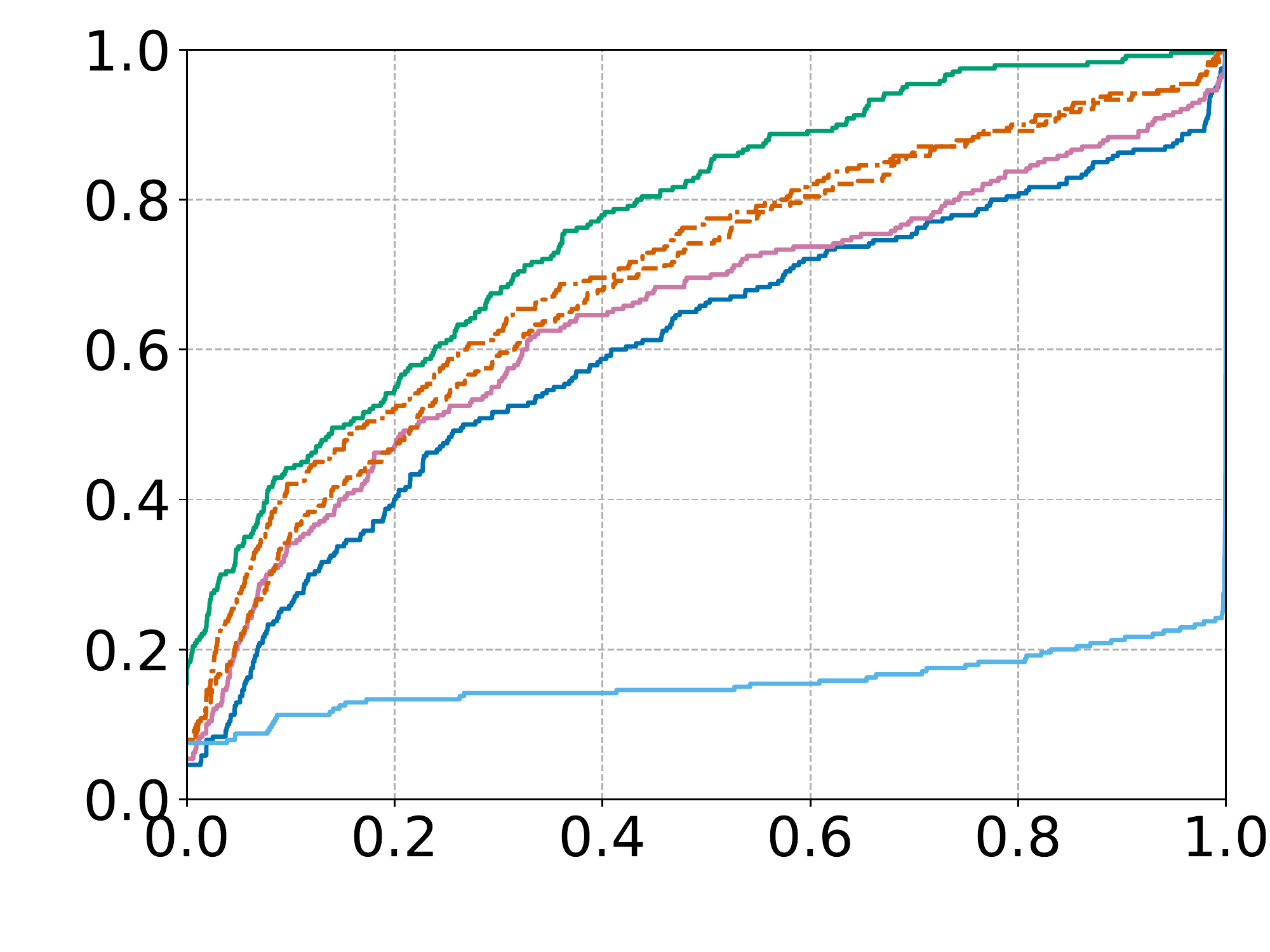}
\\\hline
\begin{tabular}{c}UAR \\ $p=0.25$\end{tabular}
& \includegraphics[width=0.29\textwidth,valign=c,trim={0 0 0 0},clip]{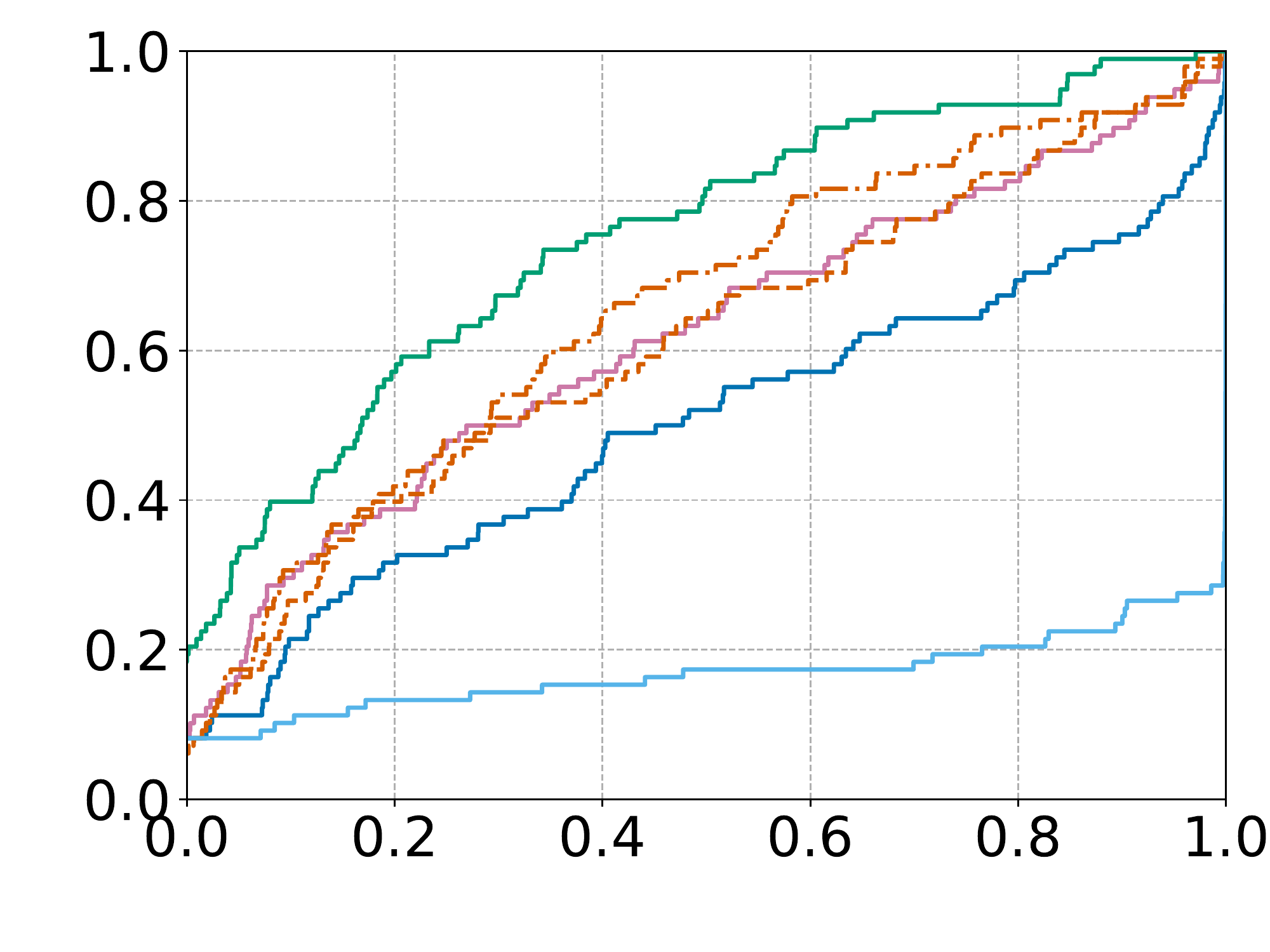}
&
\includegraphics[width=0.29\textwidth,valign=c,trim={0 0 0 0},clip]{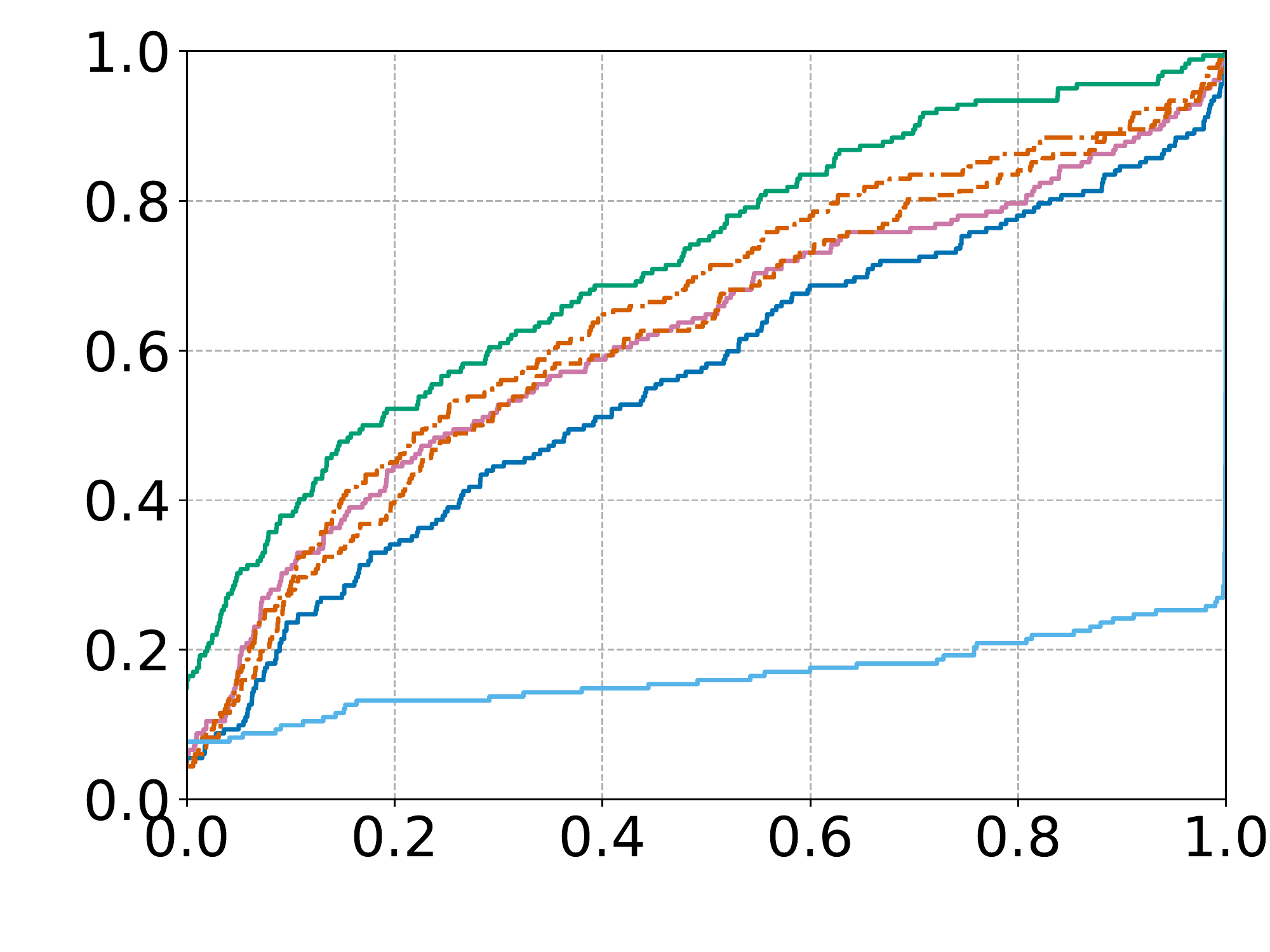}
&
\includegraphics[width=0.29\textwidth,valign=c,trim={0 0 0 0},clip]{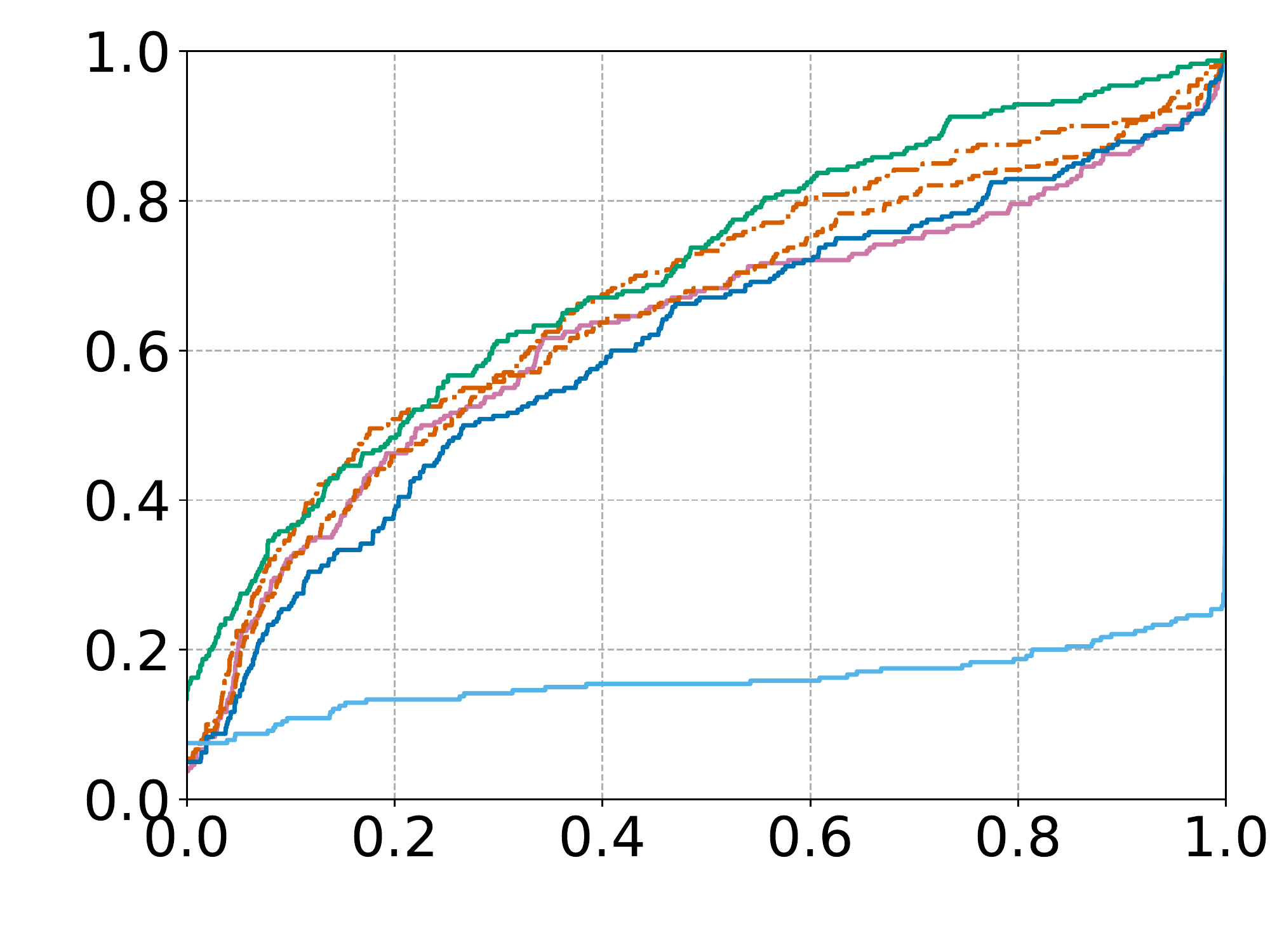}
\\\hline
\begin{tabular}{c}MAJ \\ $p=0.25$\end{tabular}
& \includegraphics[width=0.29\textwidth,valign=c,trim={0 0 0 0},clip]{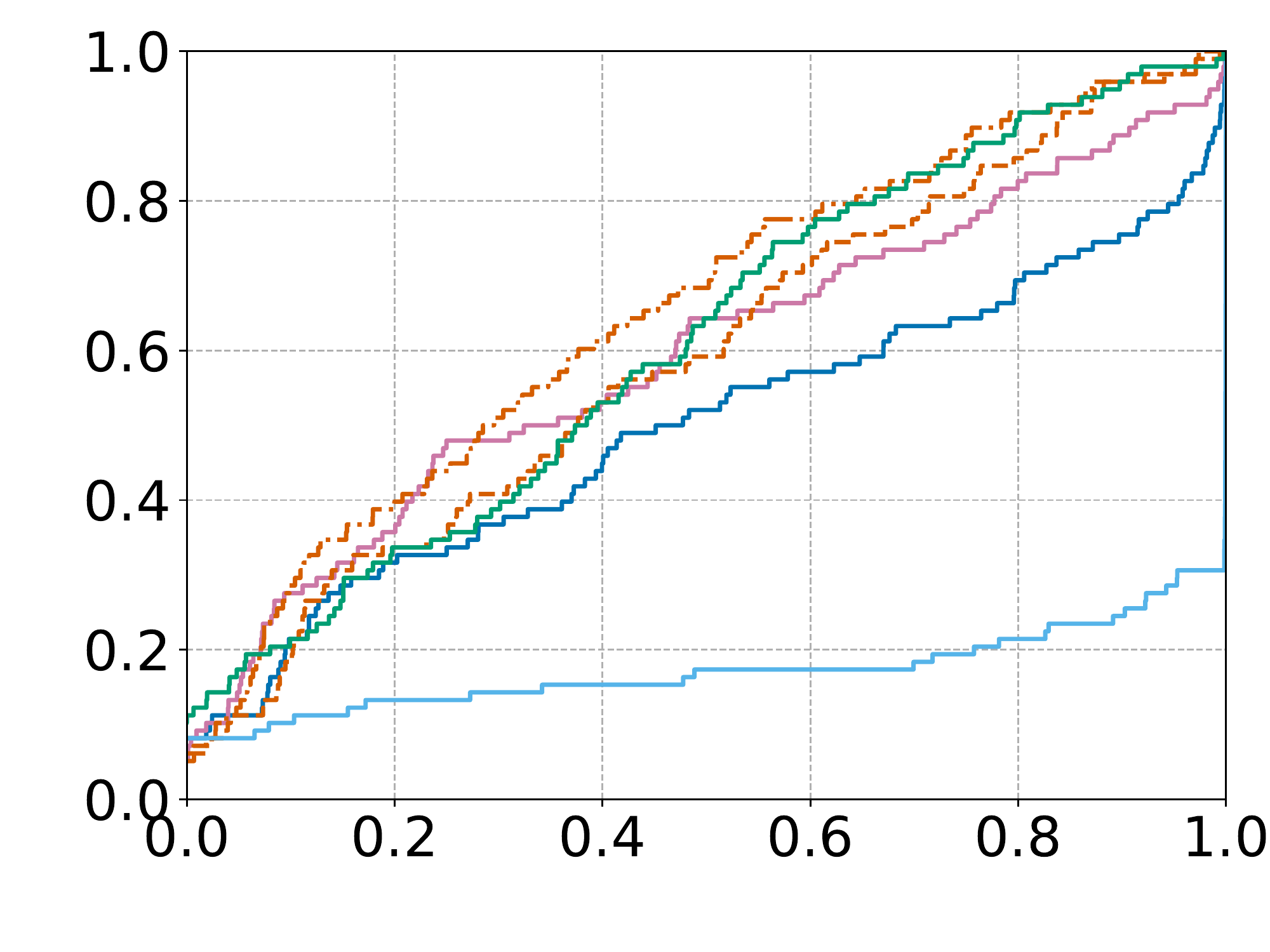}
&
\includegraphics[width=0.29\textwidth,valign=c,trim={0 0 0 0},clip]{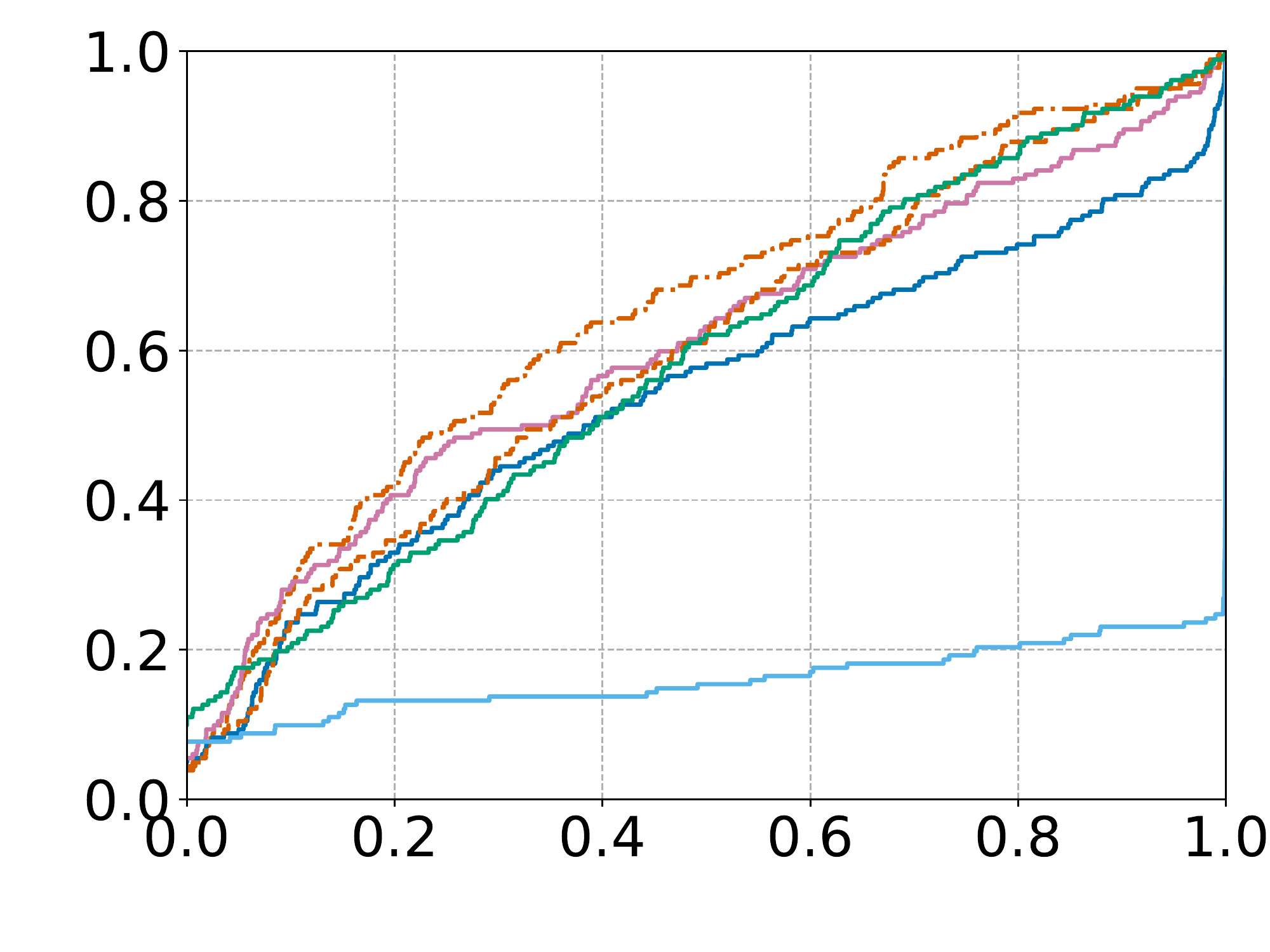}
&
\includegraphics[width=0.29\textwidth,valign=c,trim={0 0 0 0},clip]{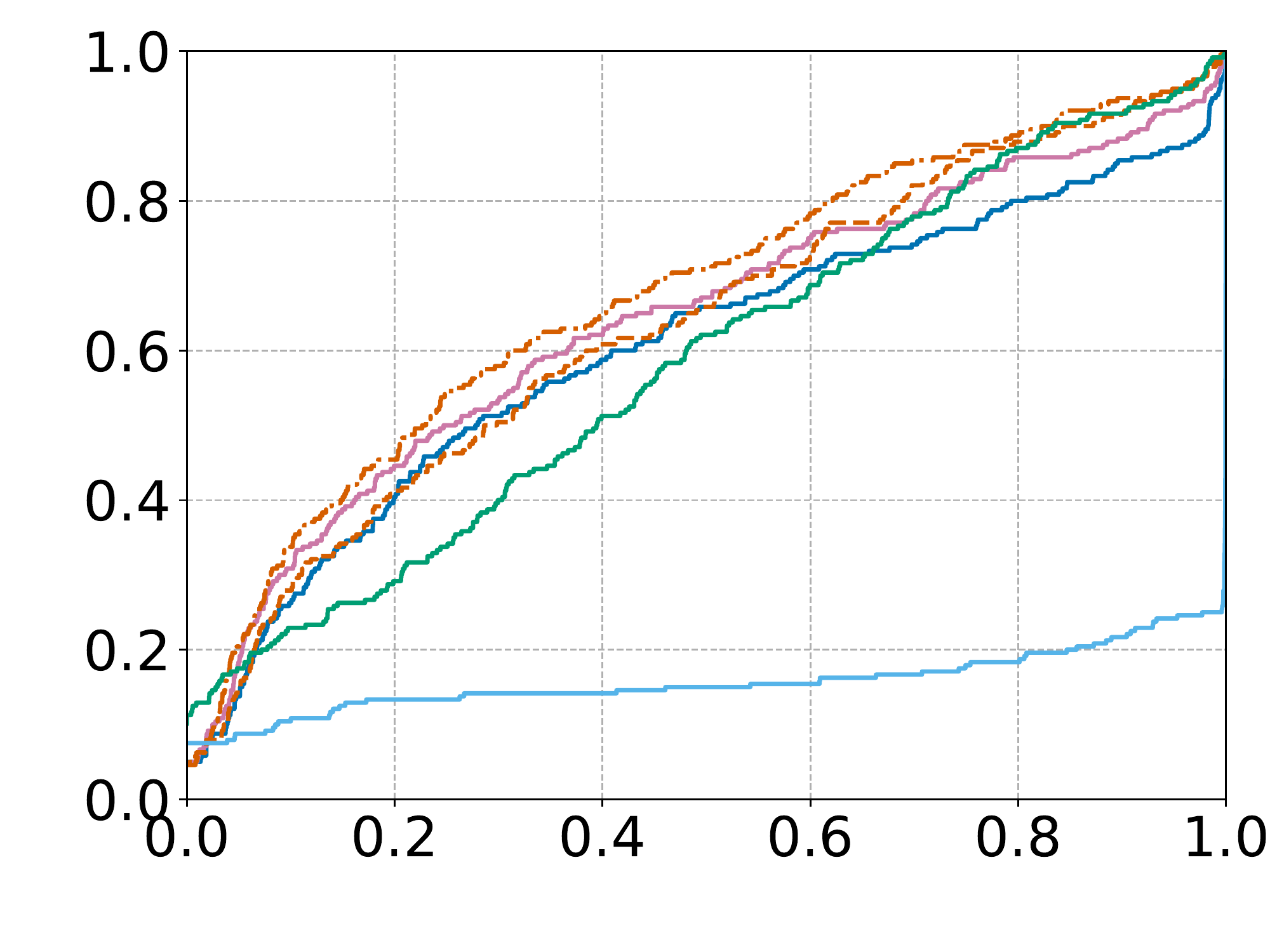}
\\\hline
\begin{tabular}{c}CYC \\ $p=0.25$\end{tabular}
& \includegraphics[width=0.29\textwidth,valign=c,trim={0 0 0 0},clip]{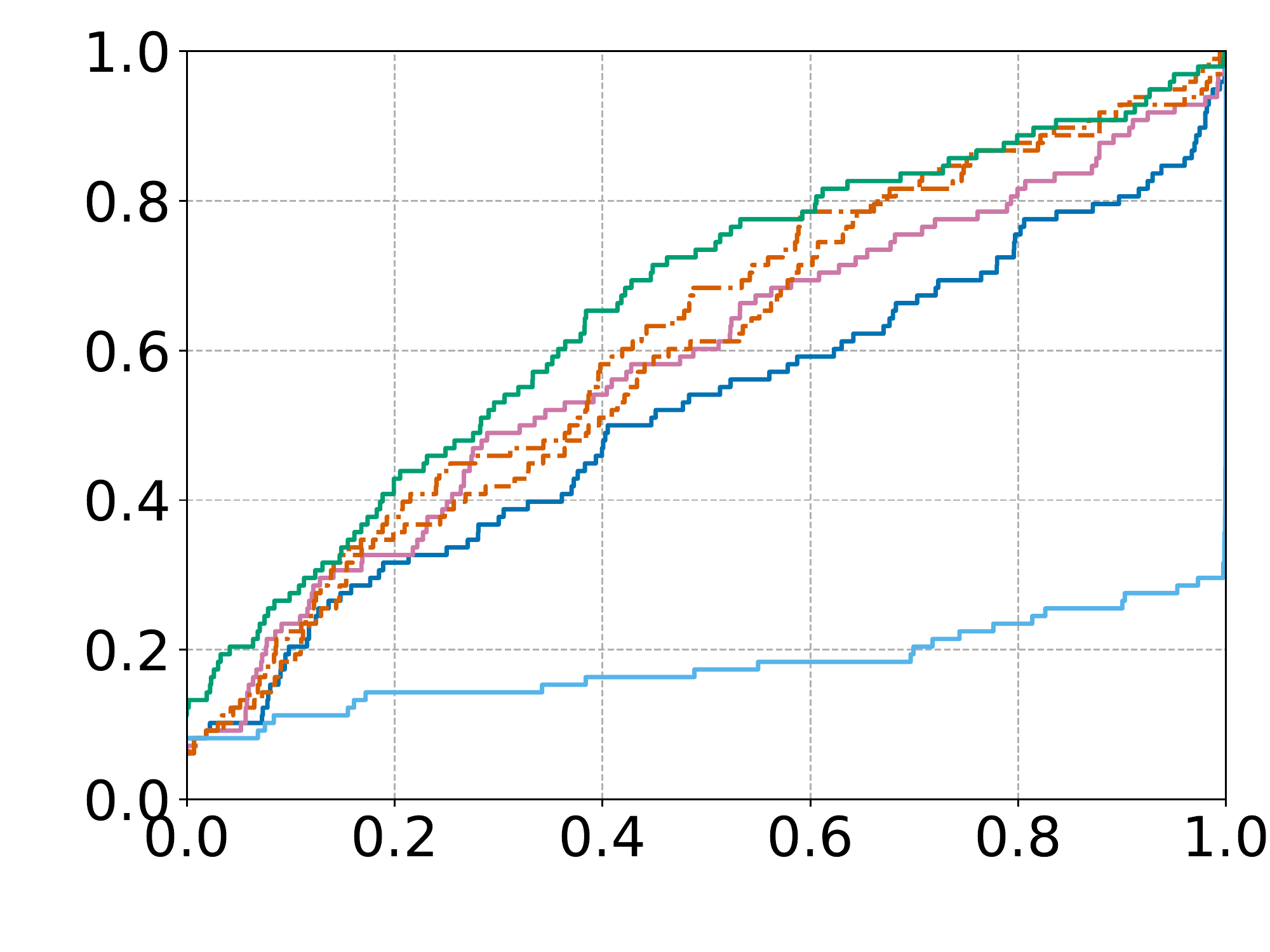}
&
\includegraphics[width=0.29\textwidth,valign=c,trim={0 0 0 0},clip]{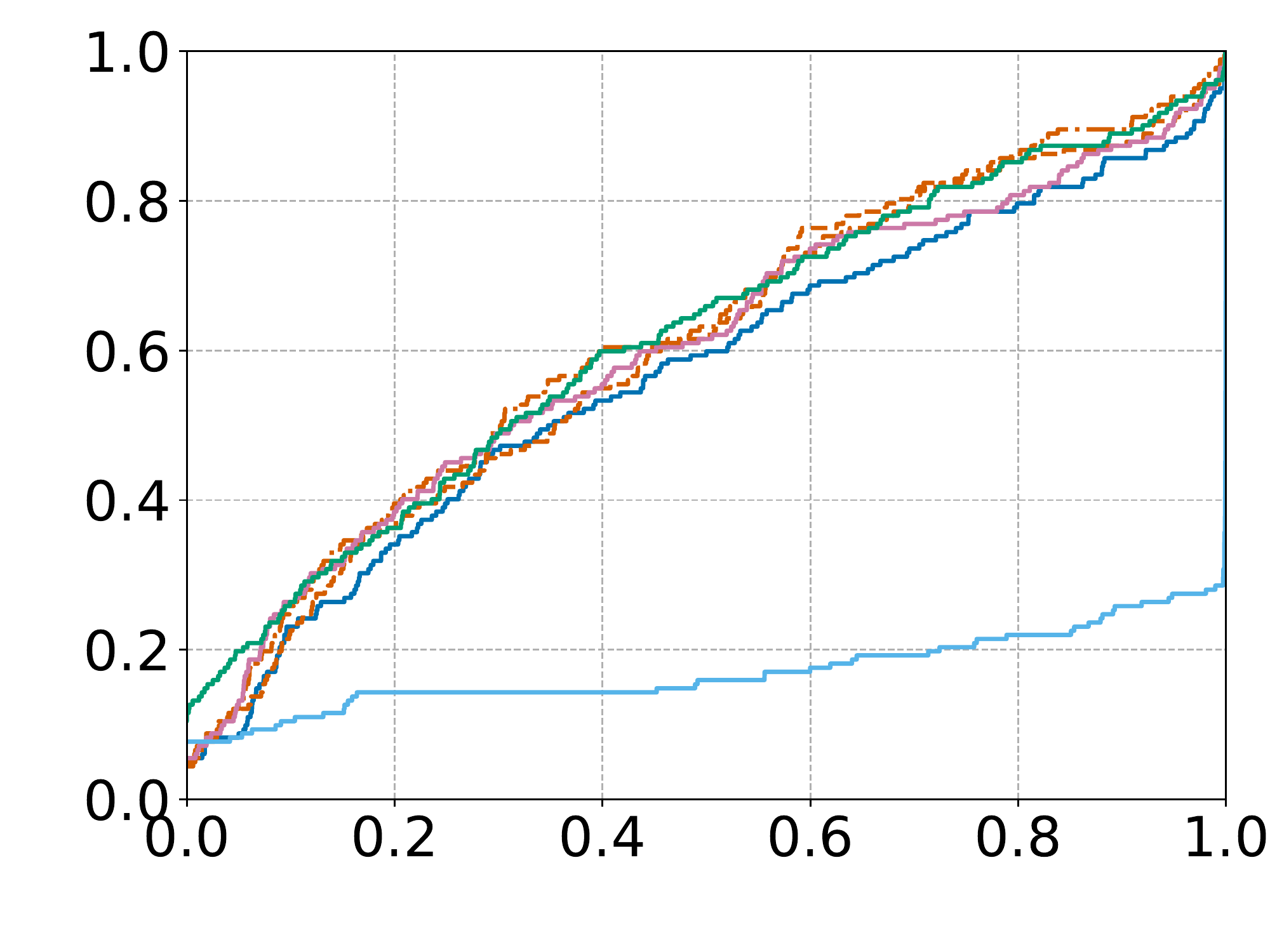}
&
\includegraphics[width=0.29\textwidth,valign=c,trim={0 0 0 0},clip]{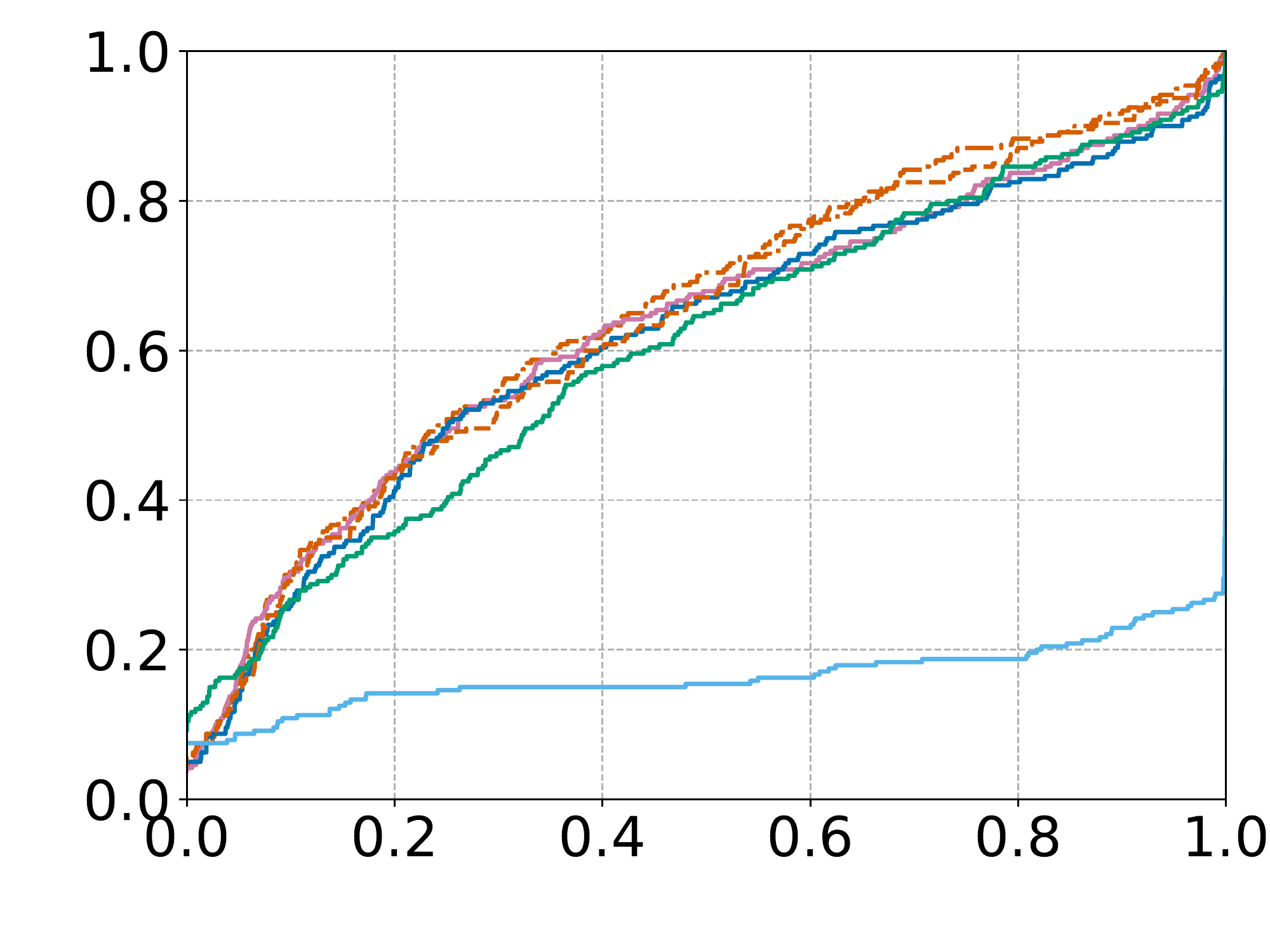}
\\\hline
\begin{tabular}{c}UAR \\ $p=0.5$\end{tabular}
& \includegraphics[width=0.29\textwidth,valign=c,trim={0 0 0 0},clip]{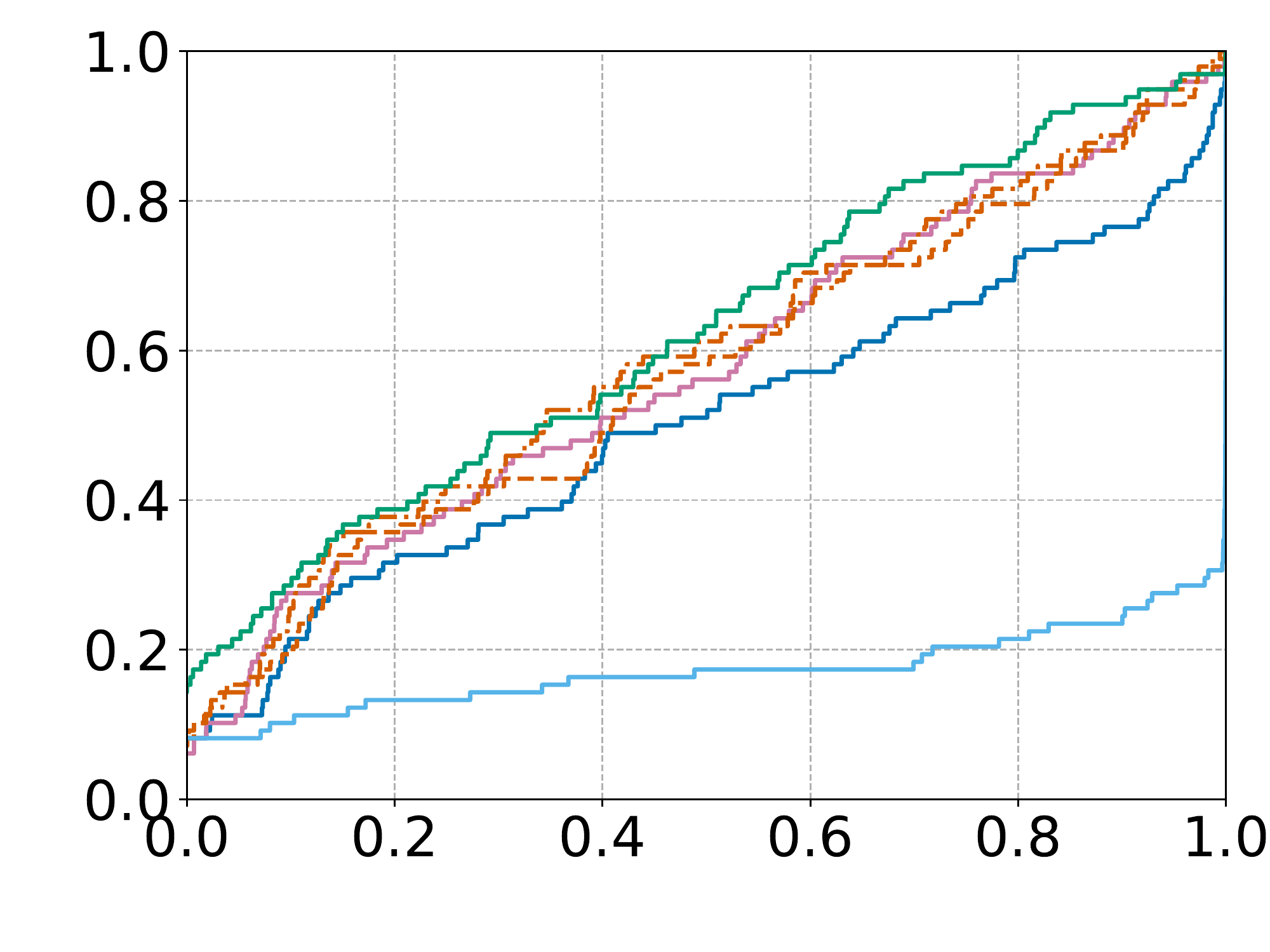}
&
\includegraphics[width=0.29\textwidth,valign=c,trim={0 0 0 0},clip]{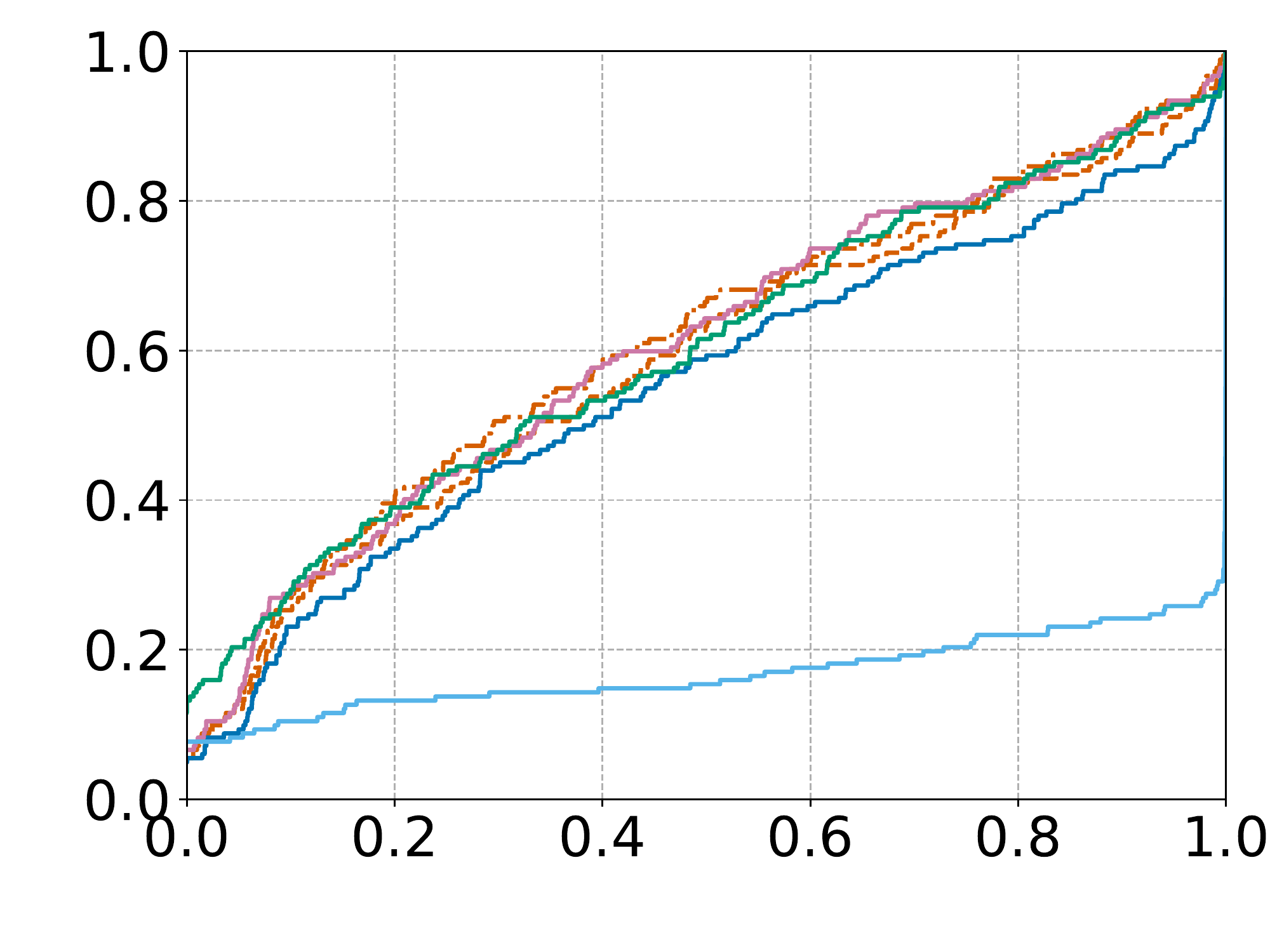}
&
\includegraphics[width=0.29\textwidth,valign=c,trim={0 0 0 0},clip]{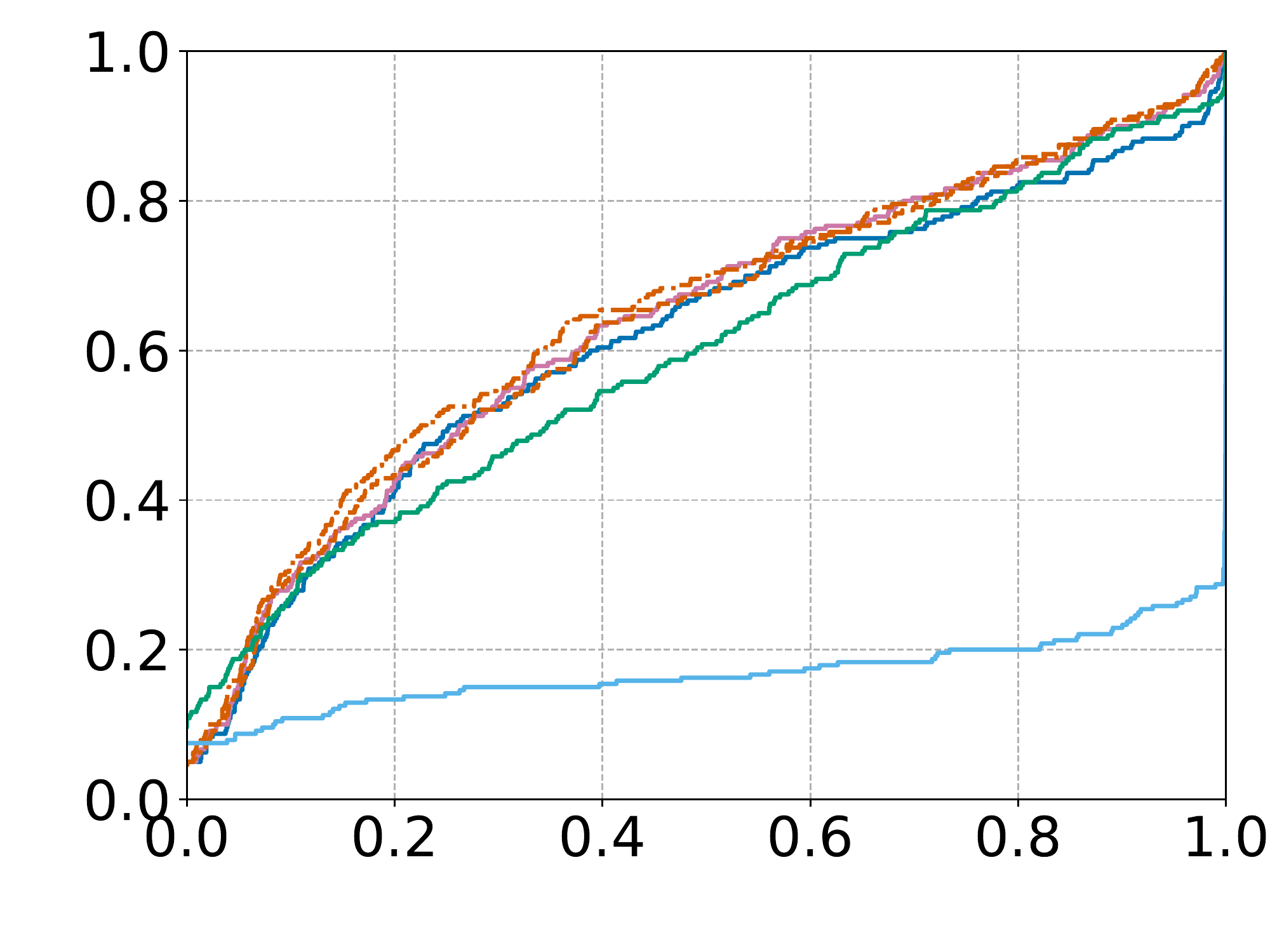}
\\\bottomrule
\multicolumn{4}{c}{
		\begin{subfigure}[t]{\textwidth}
    \includegraphics[width=\textwidth,trim={0 0 0 0},clip]{figs/cdfs/legends/legend_Lambda=8.pdf}
		\end{subfigure} }
\end{tabular}
\caption{Cumulative distribution functions (CDFs) for all evaluated algorithms, against different noise settings and warm start ratios in $\cbr{2.875,5.75,11.5}$. All CB algorithms use $\epsilon$-greedy with $\epsilon=0.0125$. In each of the above plots, the $x$ axis represents scores, while the $y$ axis represents the CDF values.}
\label{fig:cdfs-eps=0.0125-1}
\end{figure}

\begin{figure}[H]
\centering
\begin{tabular}{c | @{}c@{ }c@{ }c@{}} 
\toprule
& \multicolumn{3}{c}{ Ratio }
\\
Noise & 23.0 & 46.0 & 92.0
\\\midrule
Noiseless & \includegraphics[width=0.29\textwidth,valign=c,trim={0 0 0 0},clip]{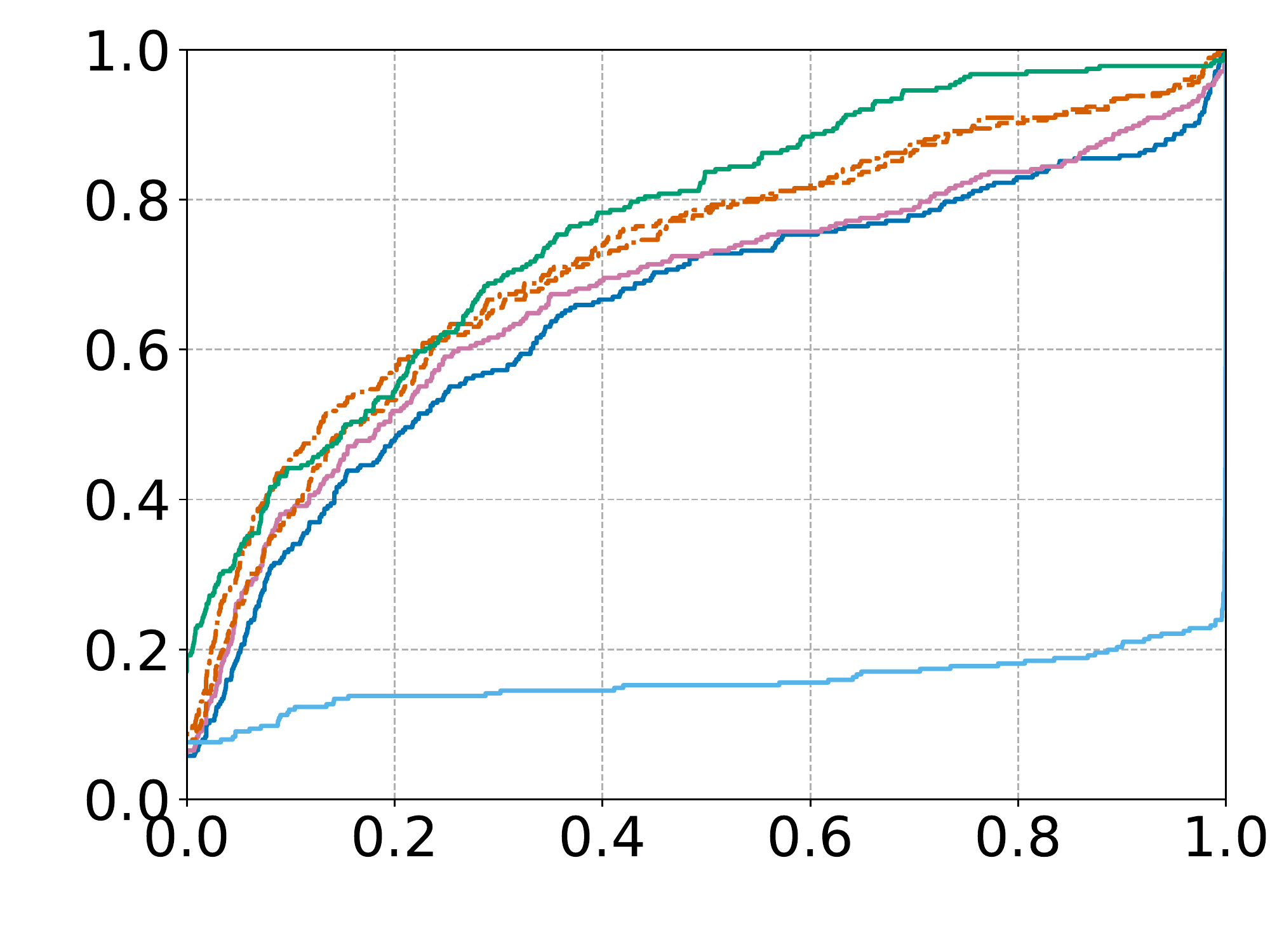}
&
\includegraphics[width=0.29\textwidth,valign=c,trim={0 0 0 0},clip]{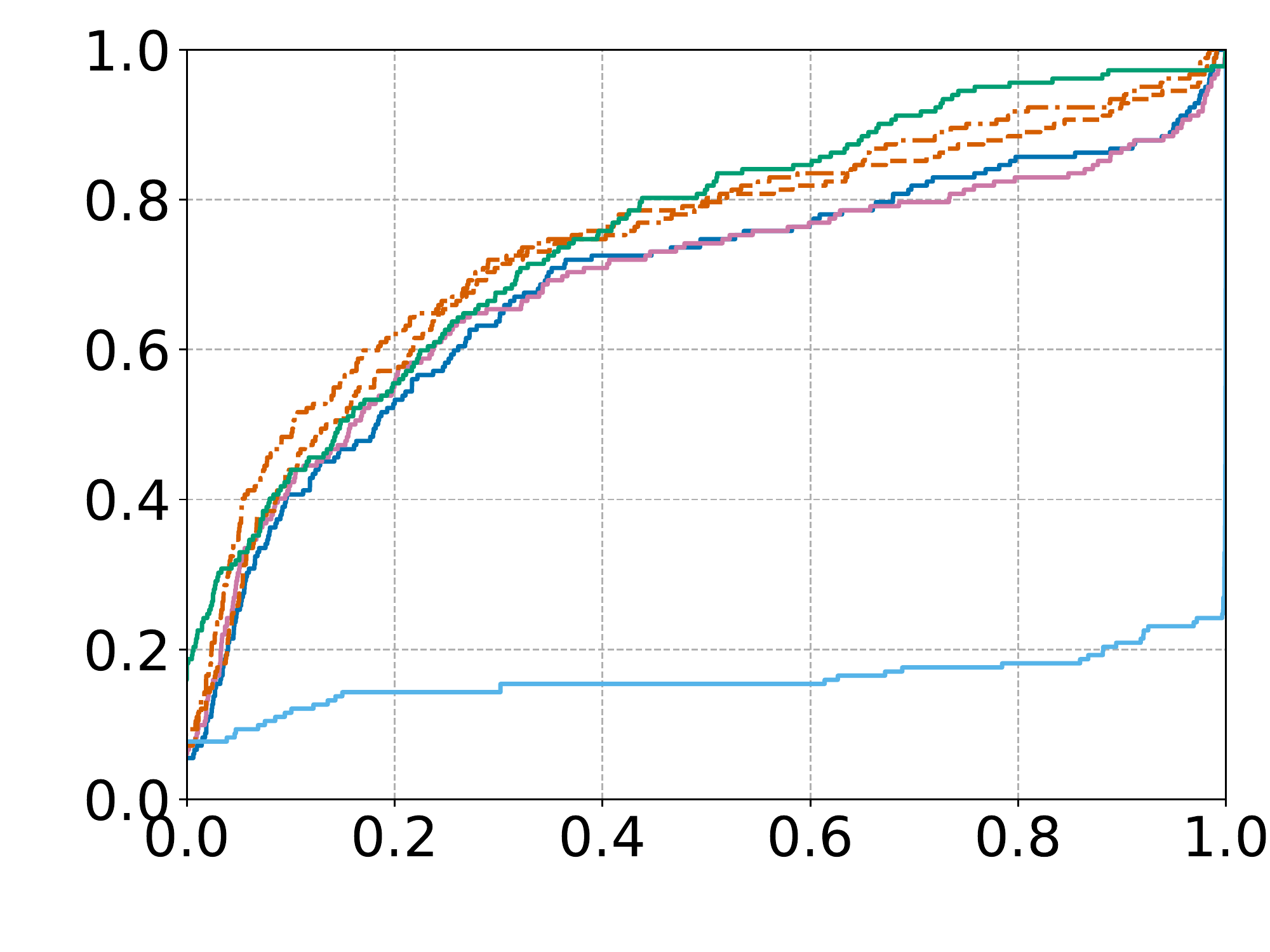}
&
\includegraphics[width=0.29\textwidth,valign=c,trim={0 0 0 0},clip]{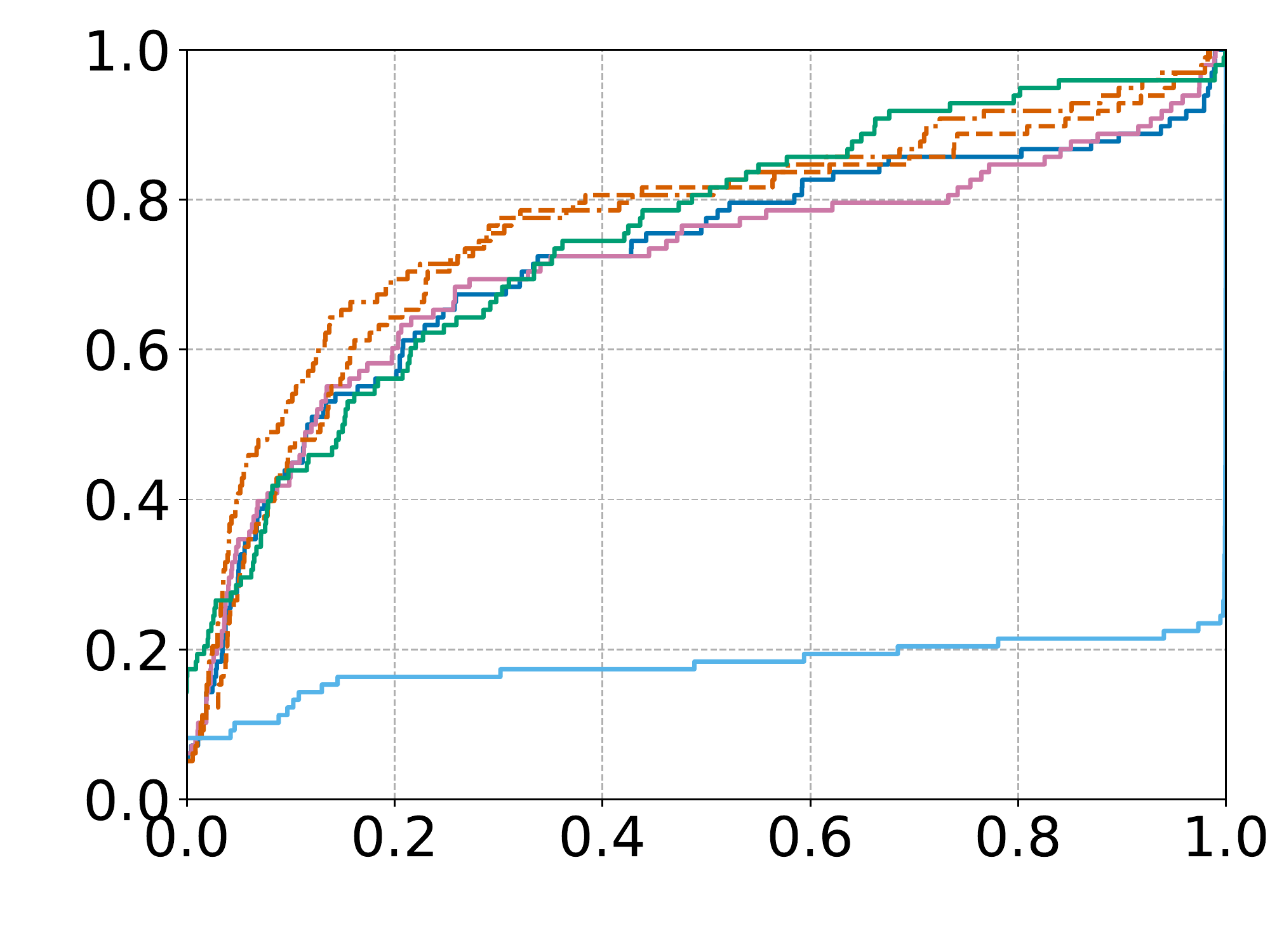}
\\\hline
\begin{tabular}{c}UAR \\ $p=0.25$\end{tabular}
& \includegraphics[width=0.29\textwidth,valign=c,trim={0 0 0 0},clip]{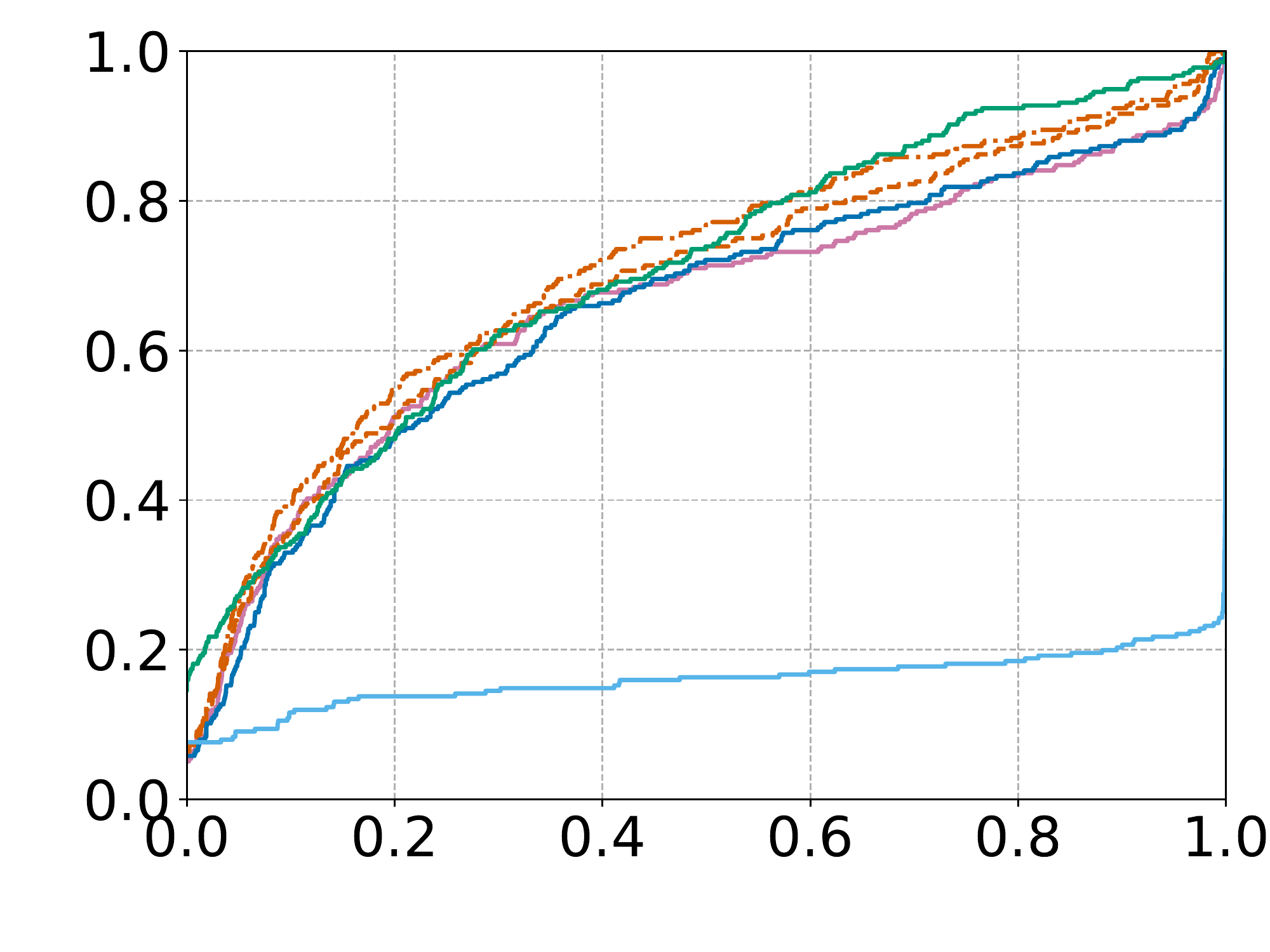}
&
\includegraphics[width=0.29\textwidth,valign=c,trim={0 0 0 0},clip]{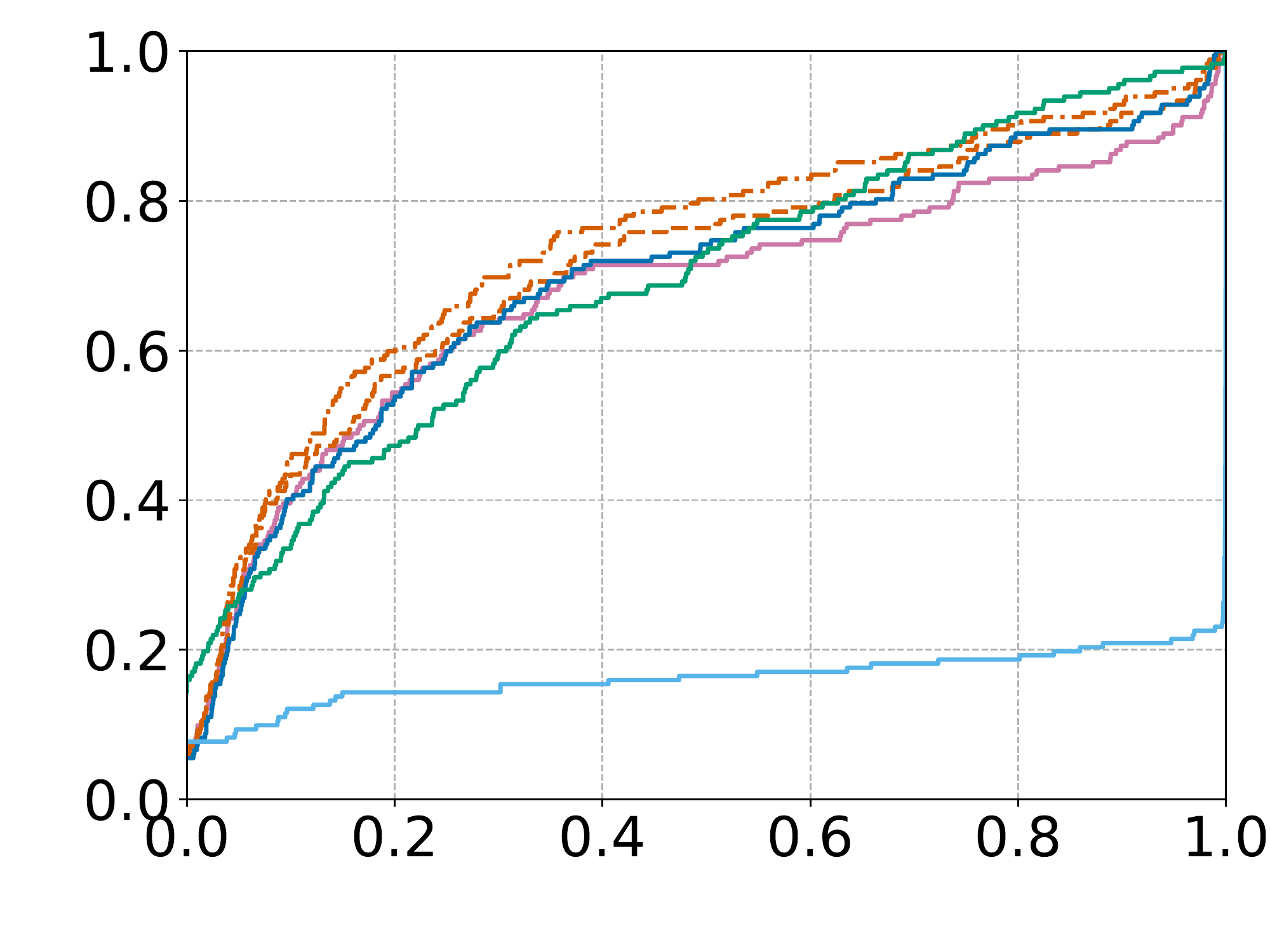}
&
\includegraphics[width=0.29\textwidth,valign=c,trim={0 0 0 0},clip]{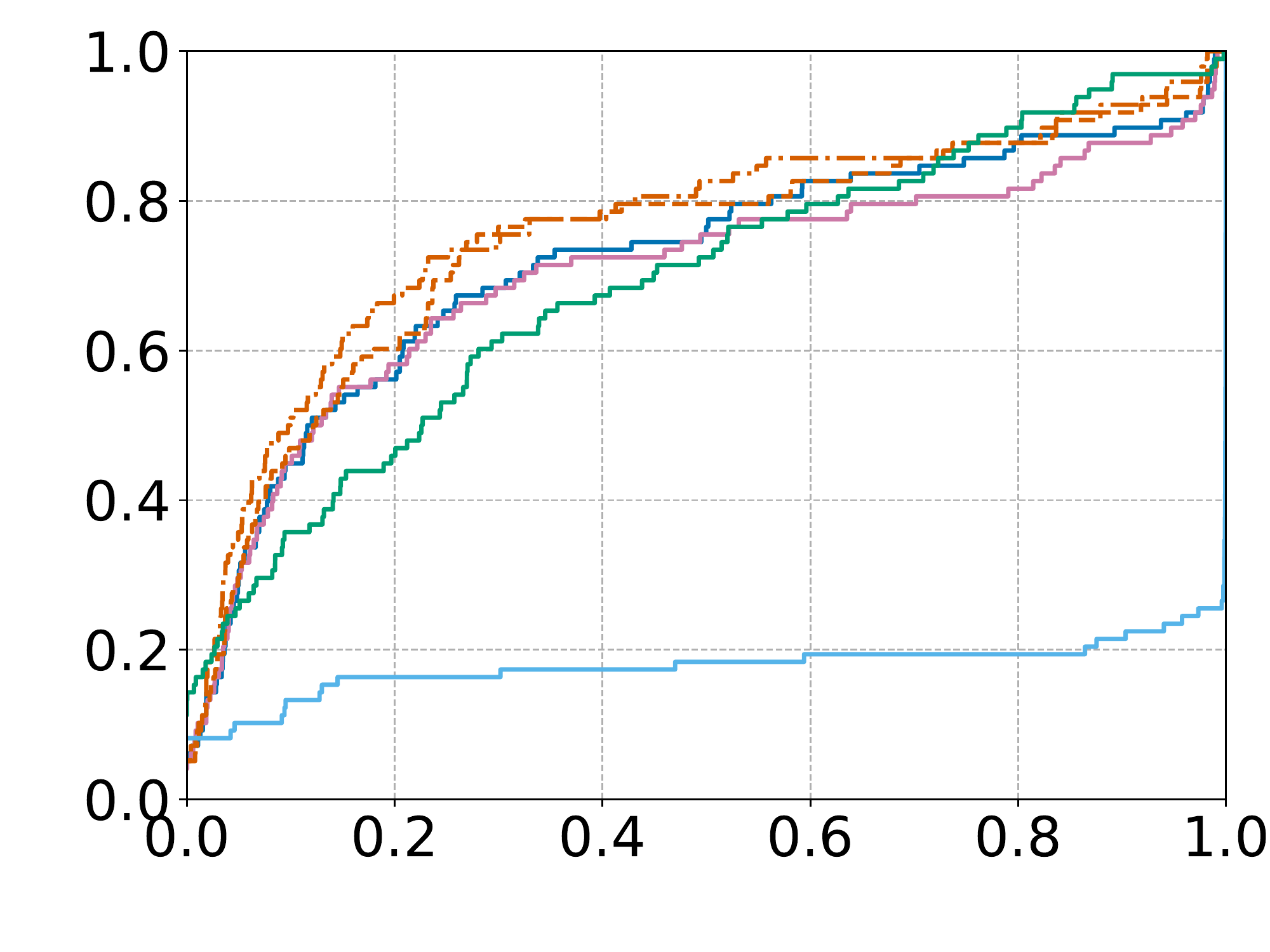}
\\\hline
\begin{tabular}{c}MAJ \\ $p=0.25$\end{tabular}
& \includegraphics[width=0.29\textwidth,valign=c,trim={0 0 0 0},clip]{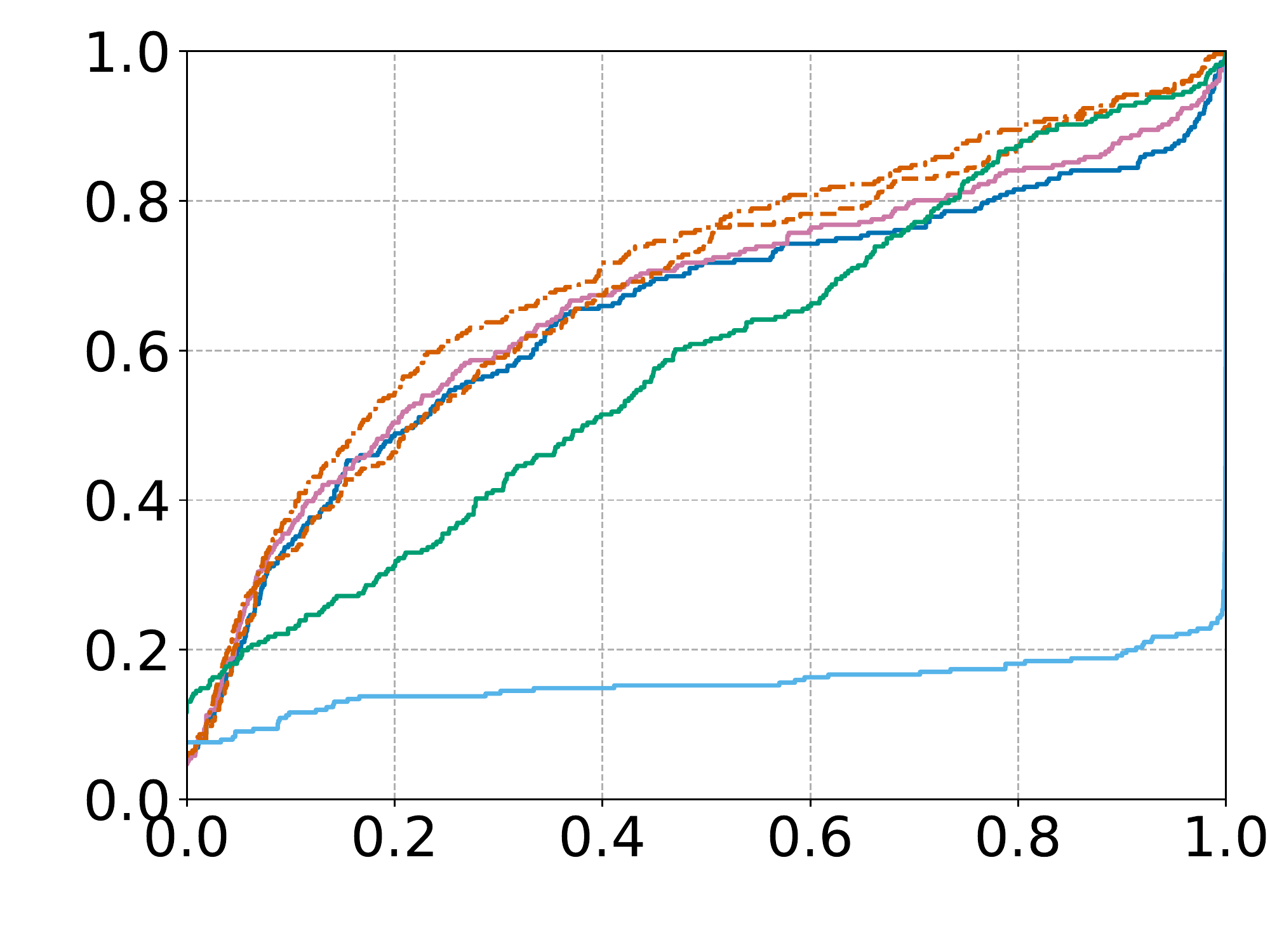}
&
\includegraphics[width=0.29\textwidth,valign=c,trim={0 0 0 0},clip]{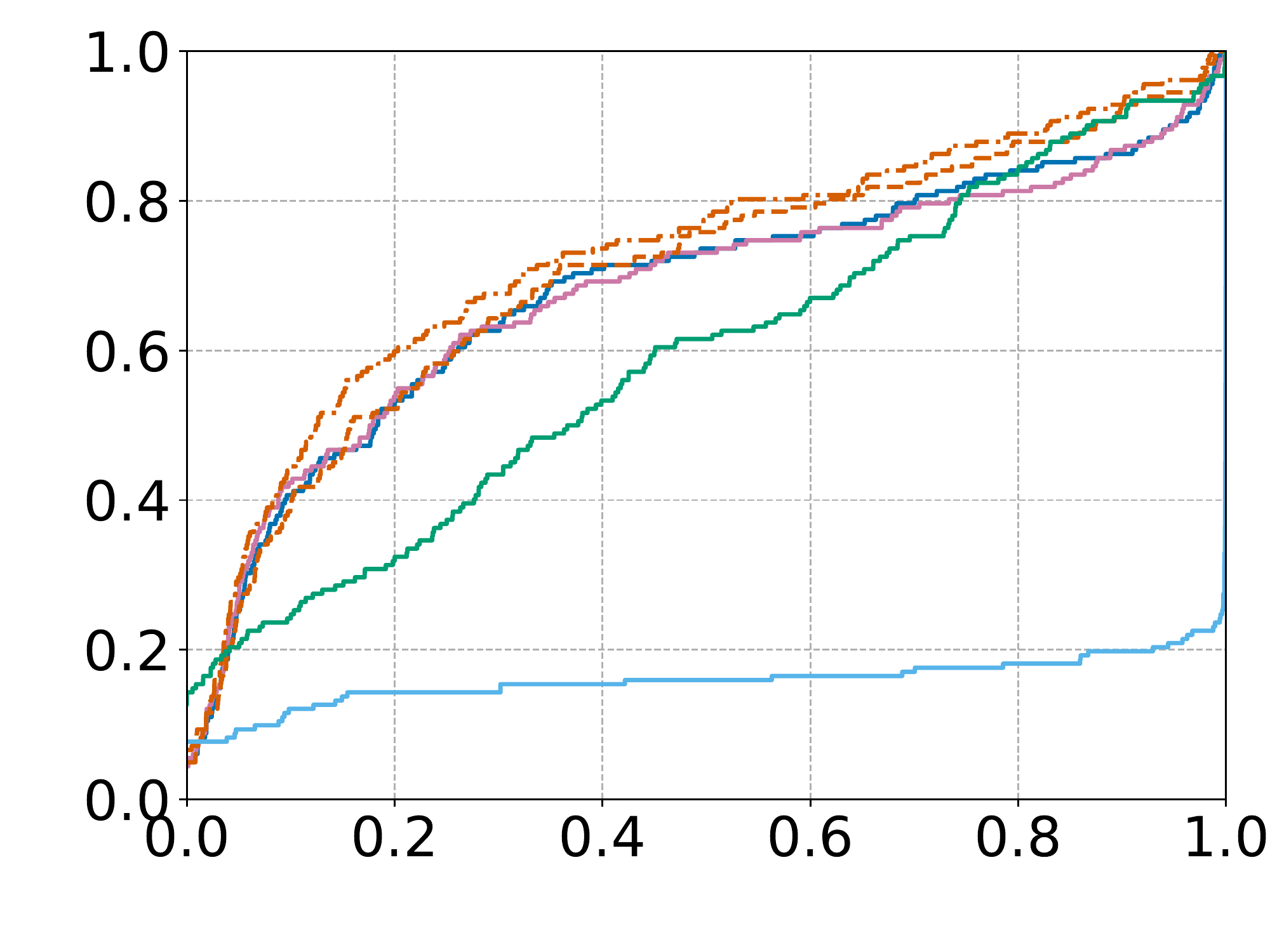}
&
\includegraphics[width=0.29\textwidth,valign=c,trim={0 0 0 0},clip]{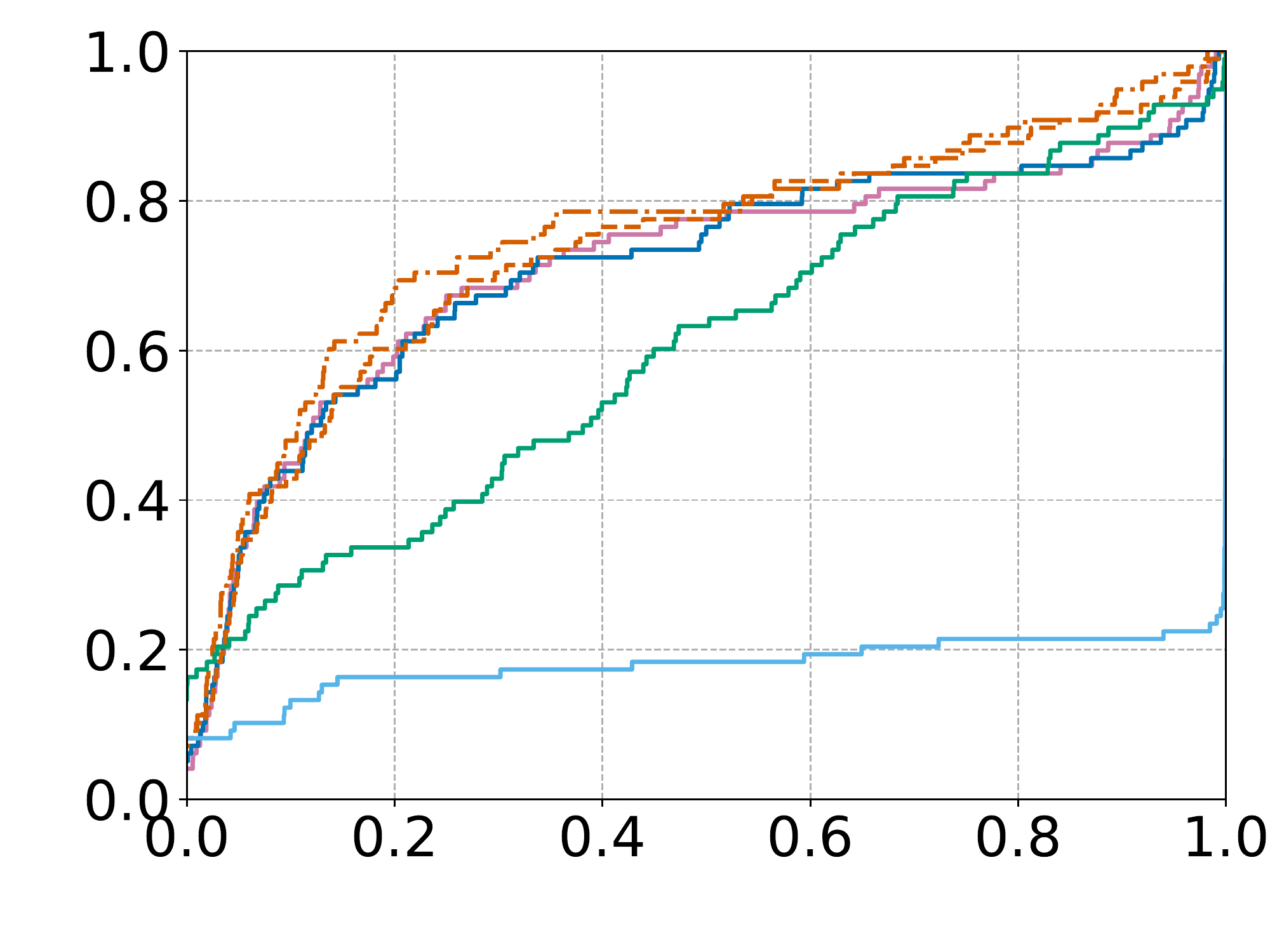}
\\\hline
\begin{tabular}{c}CYC \\ $p=0.25$\end{tabular}
& \includegraphics[width=0.29\textwidth,valign=c,trim={0 0 0 0},clip]{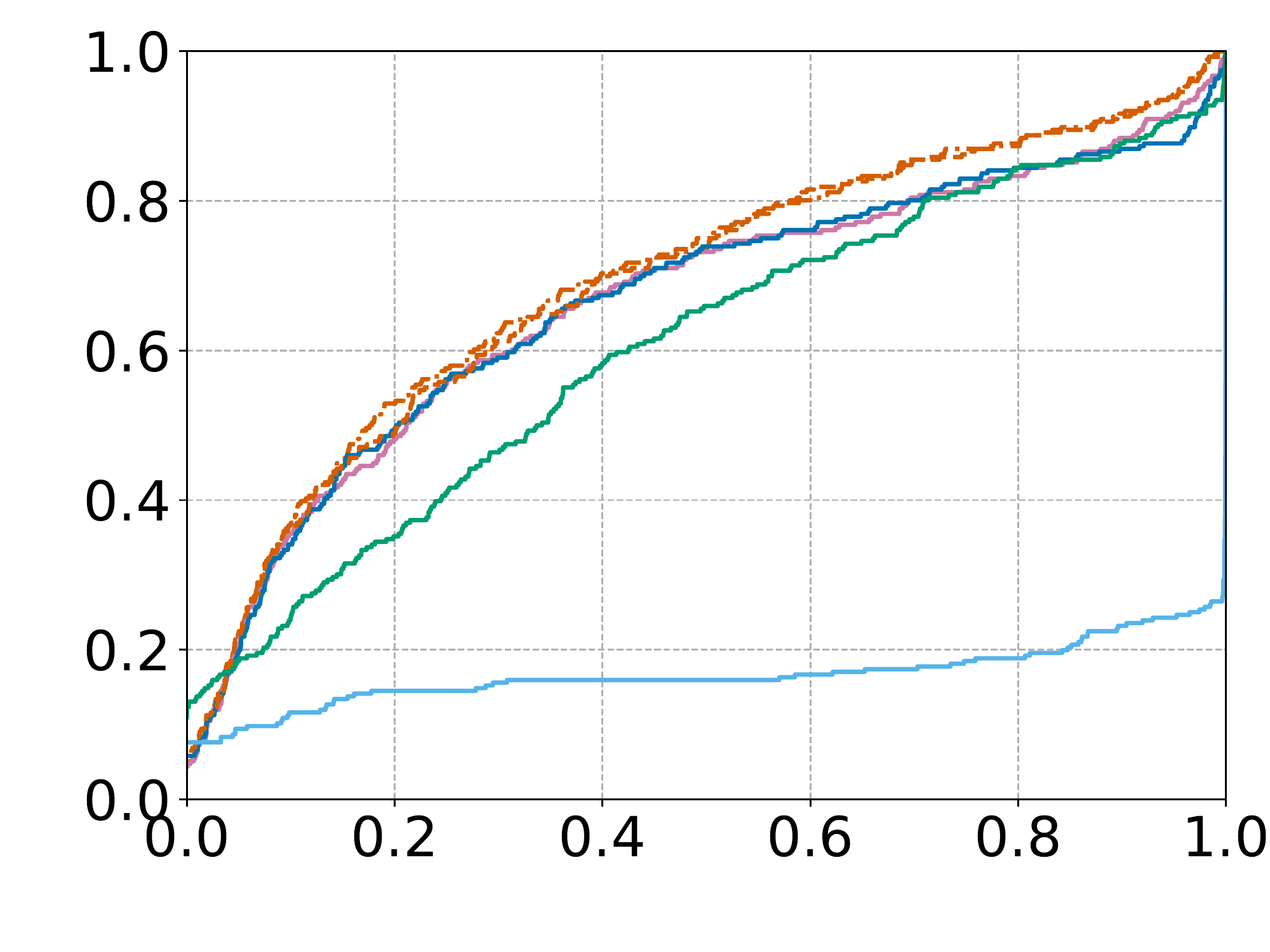}
&
\includegraphics[width=0.29\textwidth,valign=c,trim={0 0 0 0},clip]{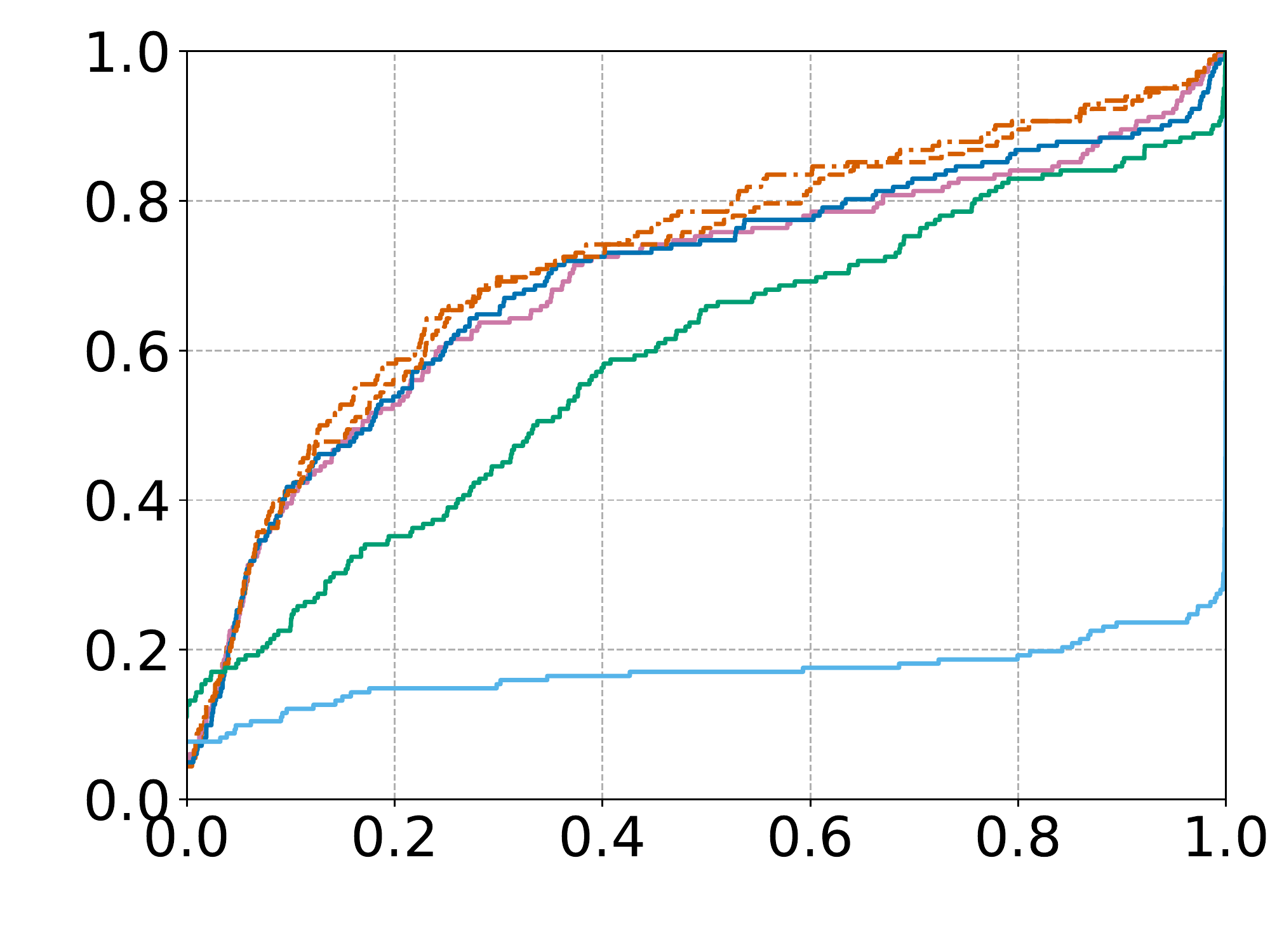}
&
\includegraphics[width=0.29\textwidth,valign=c,trim={0 0 0 0},clip]{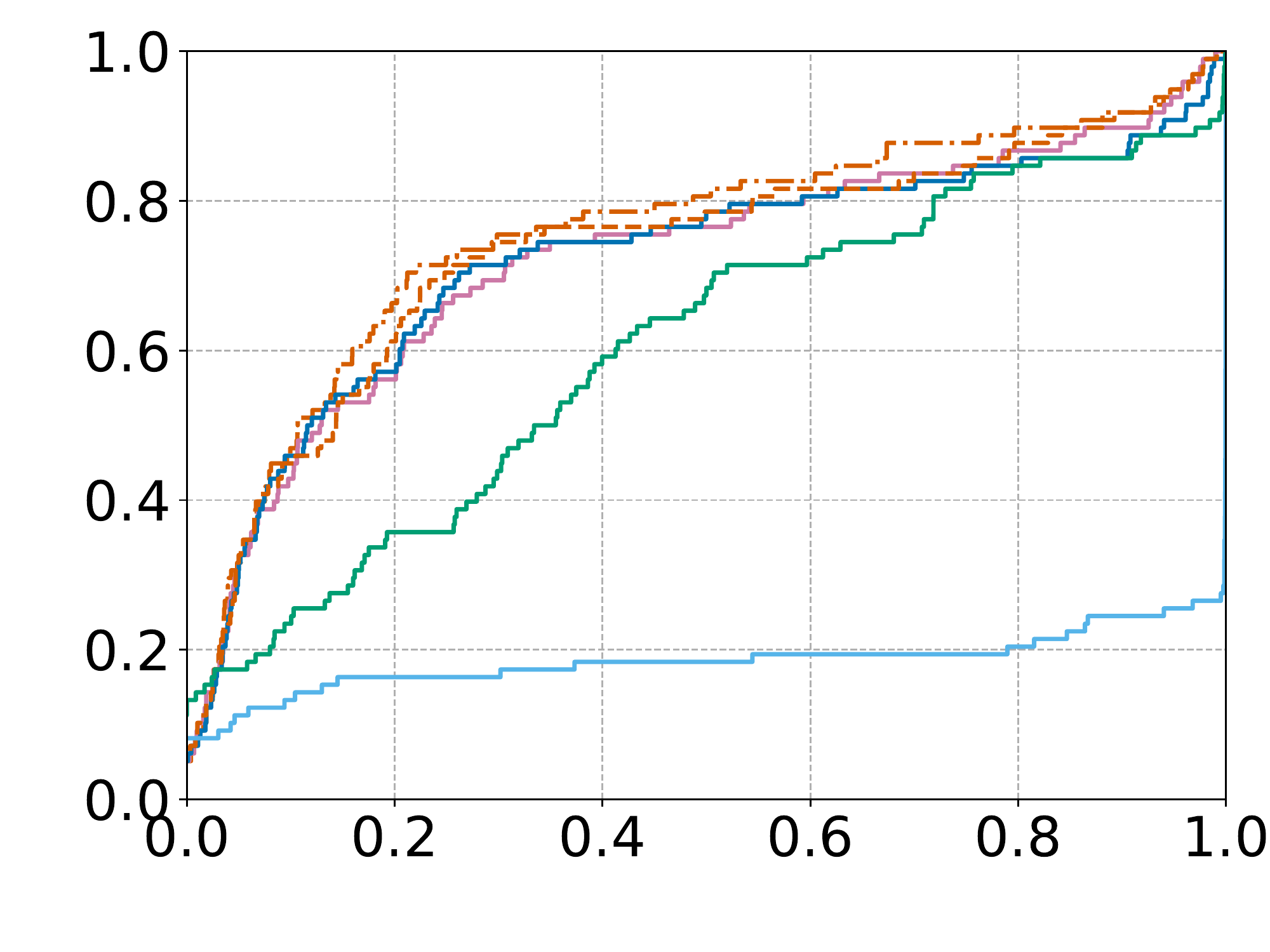}
\\\hline
\begin{tabular}{c}UAR \\ $p=0.5$\end{tabular}
& \includegraphics[width=0.29\textwidth,valign=c,trim={0 0 0 0},clip]{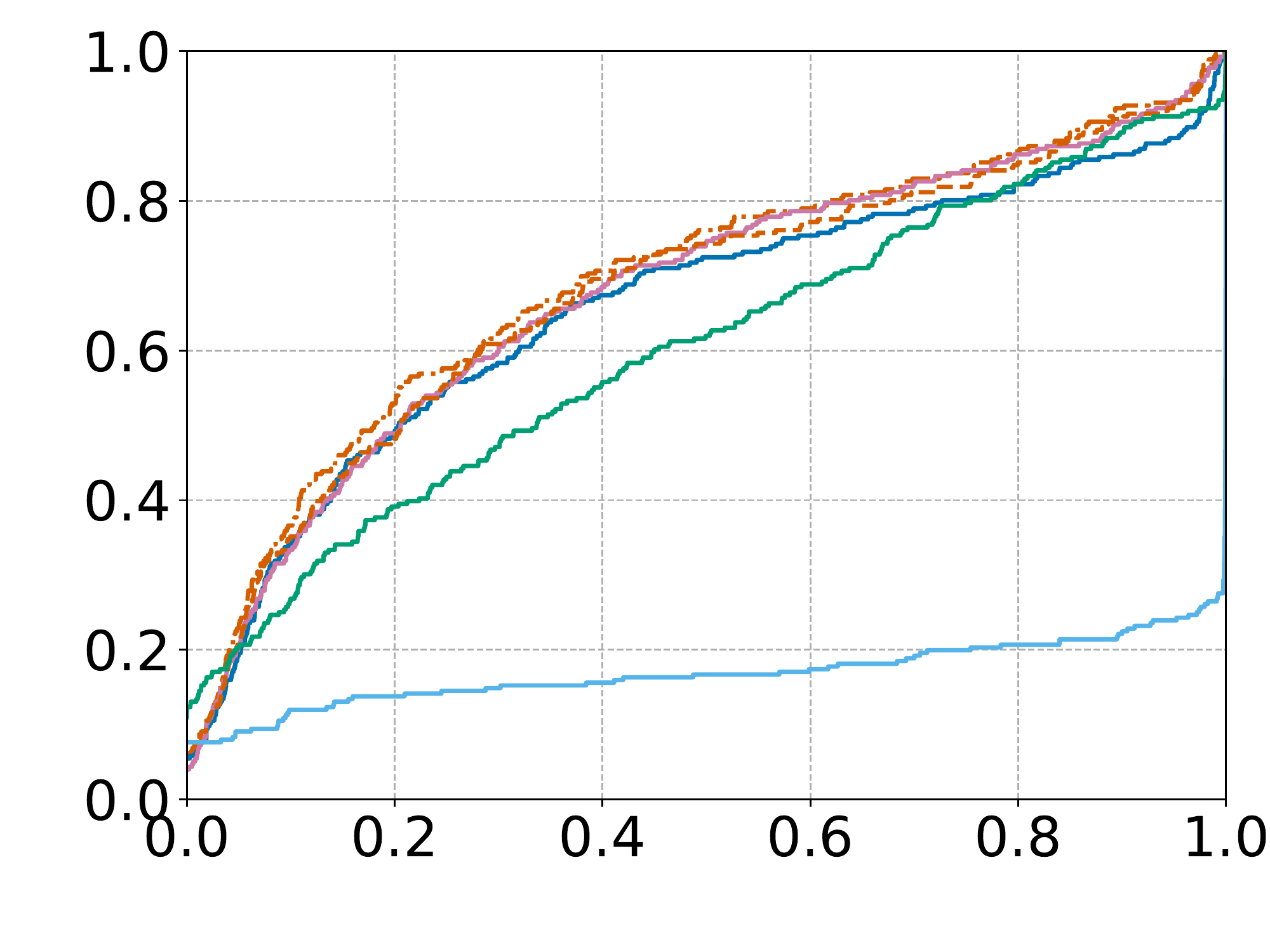}
&
\includegraphics[width=0.29\textwidth,valign=c,trim={0 0 0 0},clip]{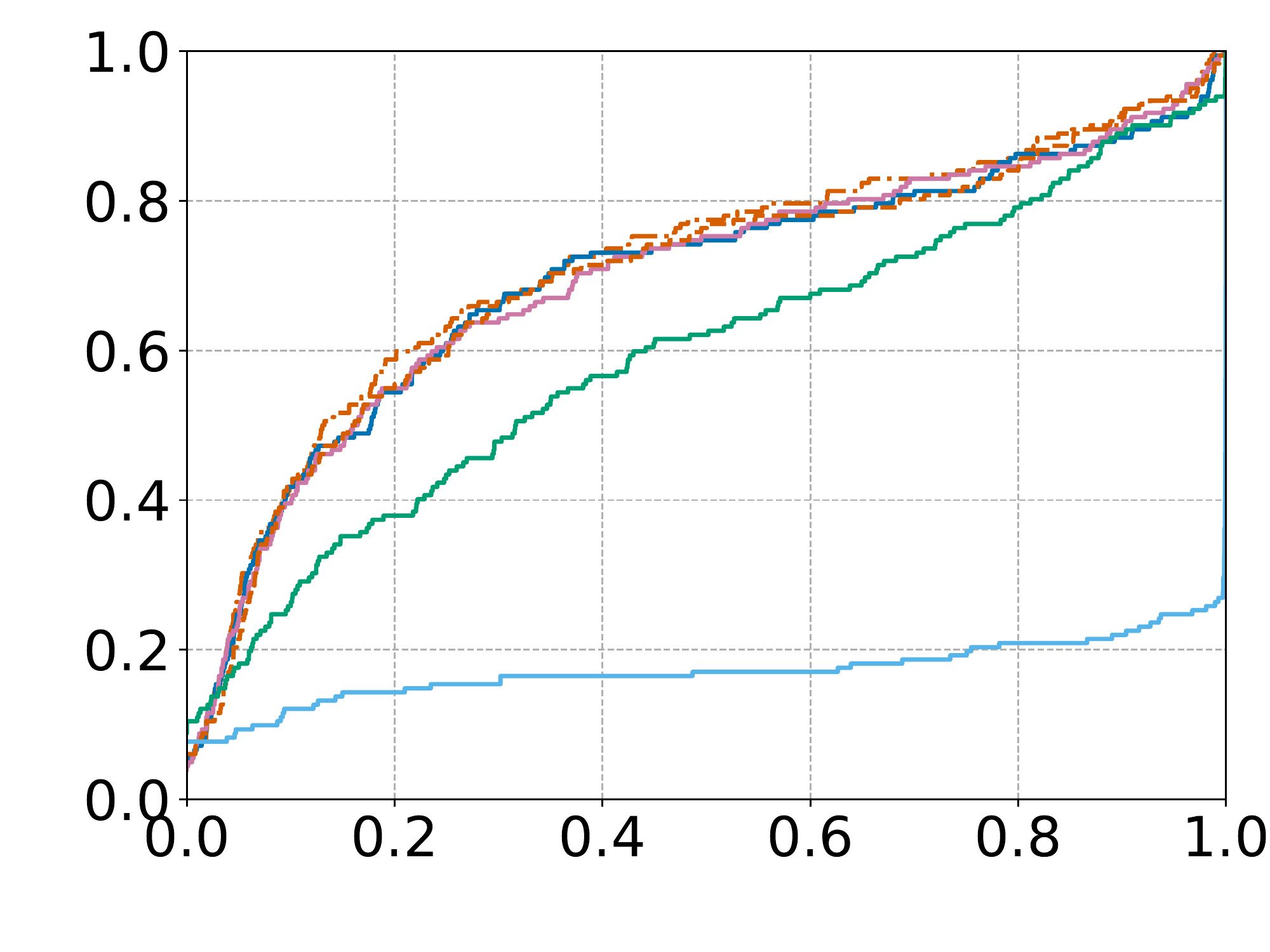}
&
\includegraphics[width=0.29\textwidth,valign=c,trim={0 0 0 0},clip]{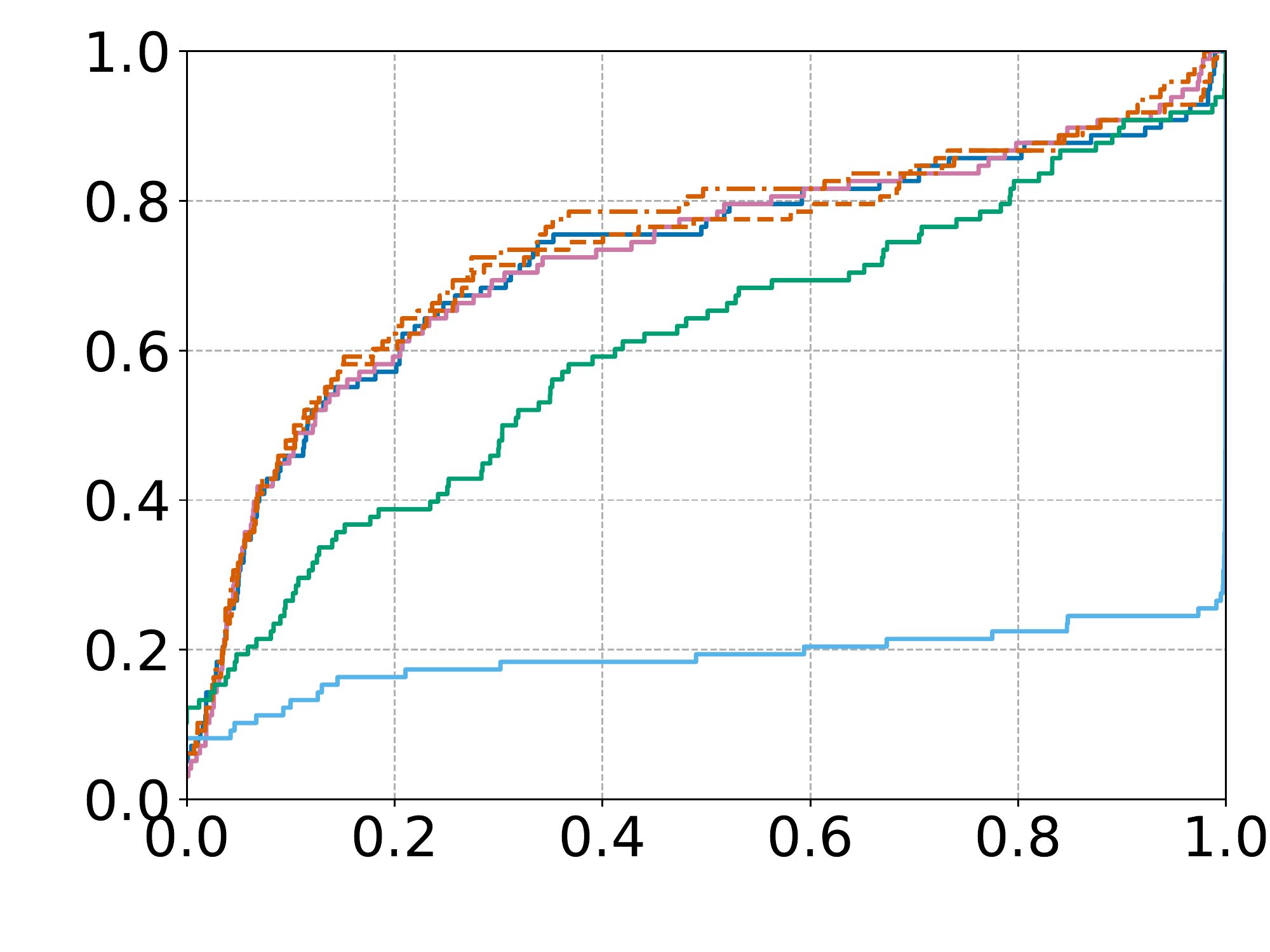}
\\\bottomrule
\multicolumn{4}{c}{
		\begin{subfigure}[t]{\textwidth}
    \includegraphics[width=\textwidth,trim={0 0 0 0},clip]{figs/cdfs/legends/legend_Lambda=8.pdf}
		\end{subfigure} }
\end{tabular}
\caption{Cumulative distribution functions (CDFs) for all evaluated algorithms, against different noise settings and warm start ratios in $\cbr{23.0,46.0,92.0}$. All CB algorithms use $\epsilon$-greedy with $\epsilon=0.0125$. In each of the above plots, the $x$ axis represents scores, while the $y$ axis represents the CDF values.}
\label{fig:cdfs-eps=0.0125-2}
\end{figure}

\begin{figure}[H]
\centering
\begin{tabular}{c |@{}c@{}} 
\toprule
& \multicolumn{1}{c}{ Ratio }
\\
Noise & 184.0
\\\midrule
Noiseless & \includegraphics[width=0.3\textwidth,valign=c,trim={0 0 0 0},clip]{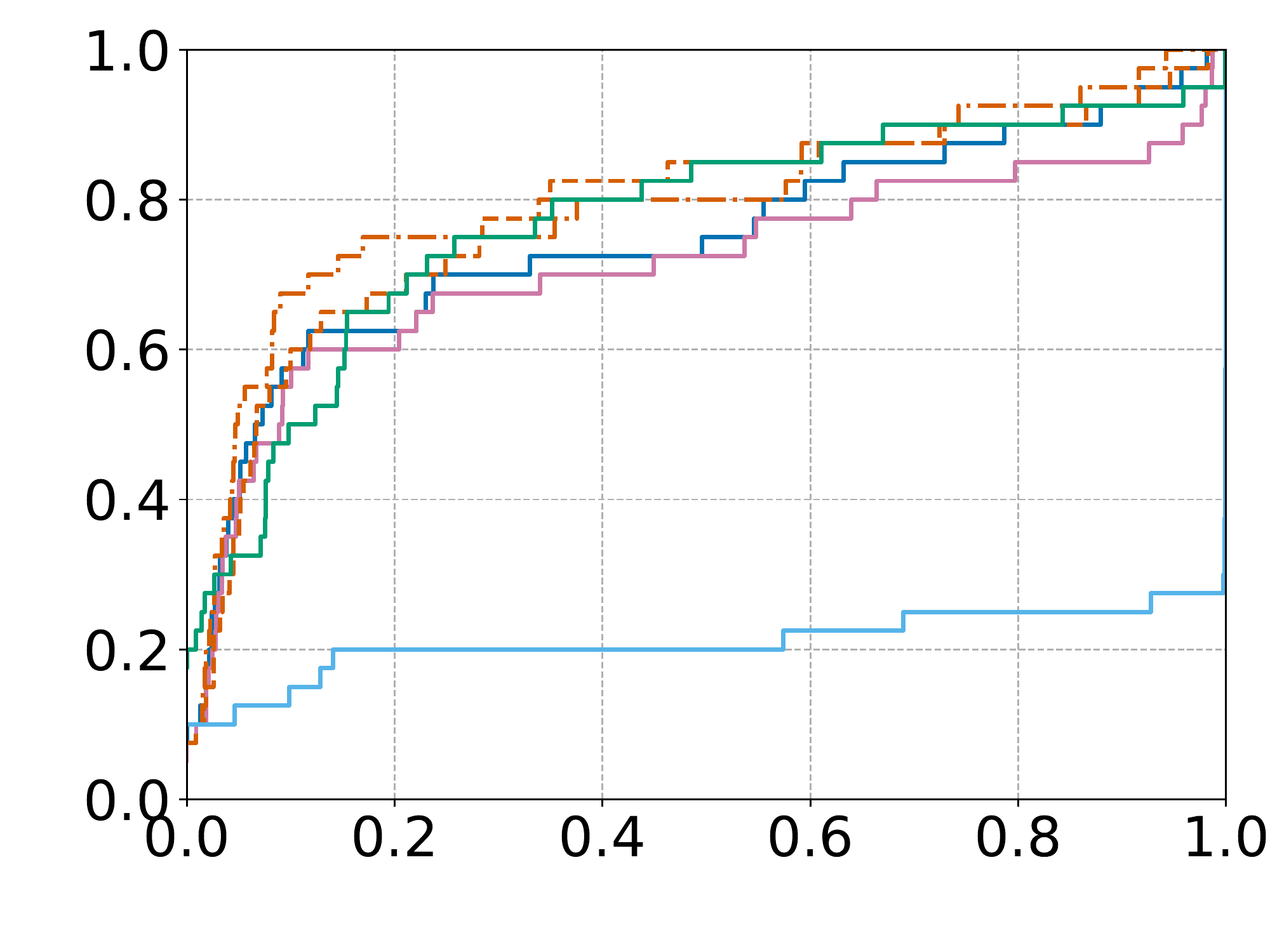}
\\\hline
\begin{tabular}{c}UAR \\ $p=0.25$\end{tabular}
& \includegraphics[width=0.3\textwidth,valign=c,trim={0 0 0 0},clip]{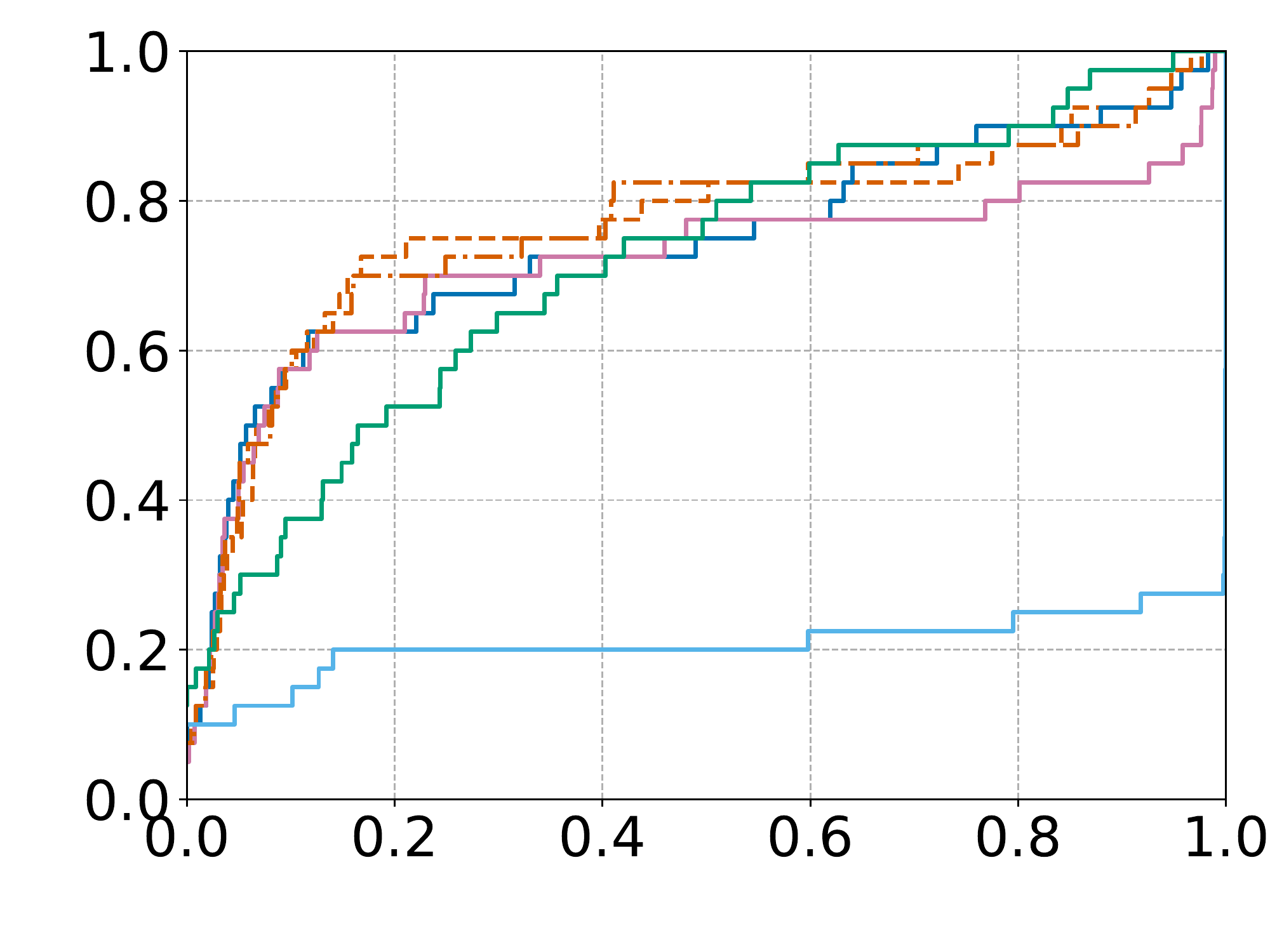}

\\\hline
\begin{tabular}{c}MAJ \\ $p=0.25$\end{tabular}
& \includegraphics[width=0.3\textwidth,valign=c,trim={0 0 0 0},clip]{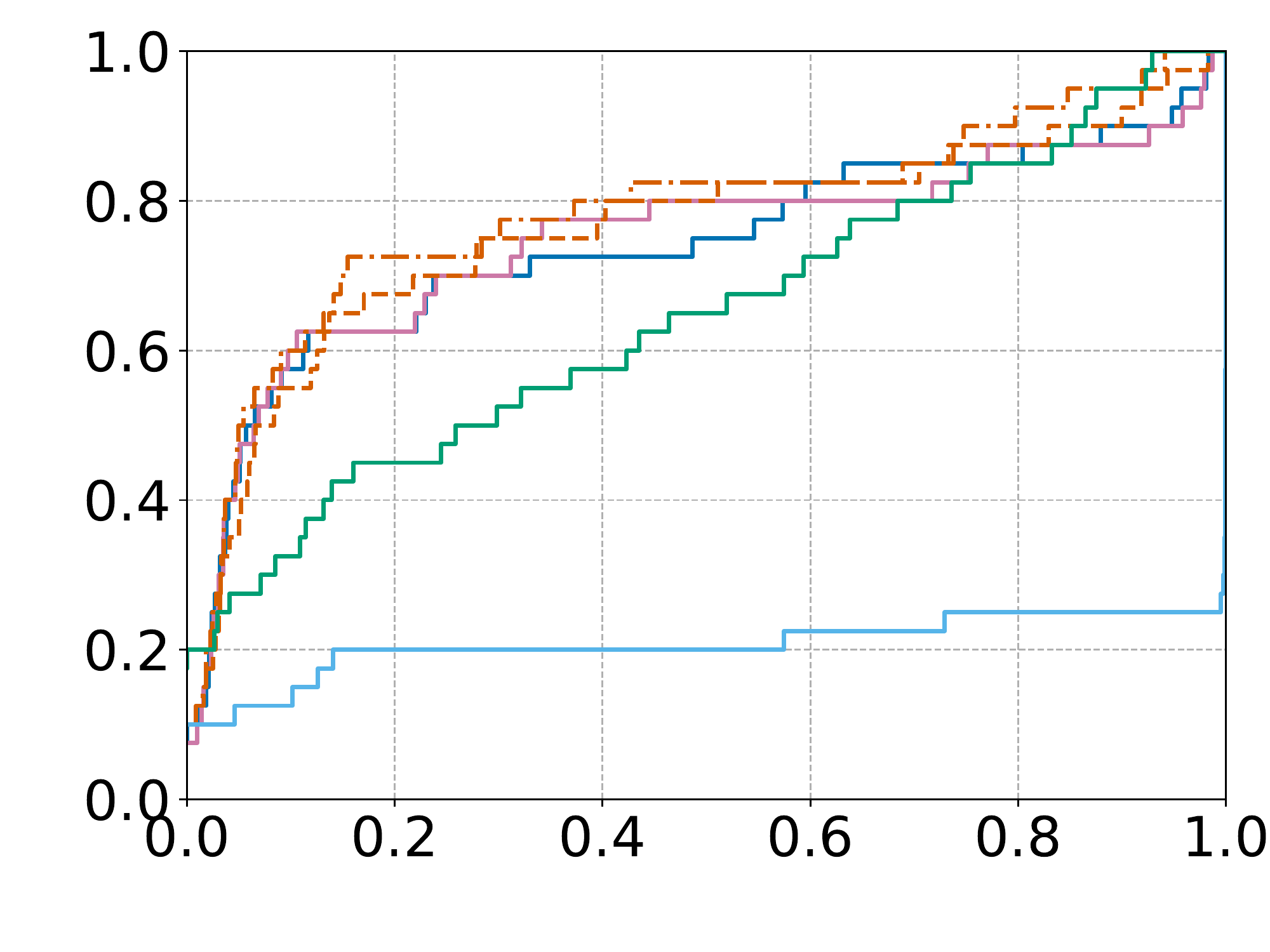}

\\\hline
\begin{tabular}{c}CYC \\ $p=0.25$\end{tabular}
& \includegraphics[width=0.3\textwidth,valign=c,trim={0 0 0 0},clip]{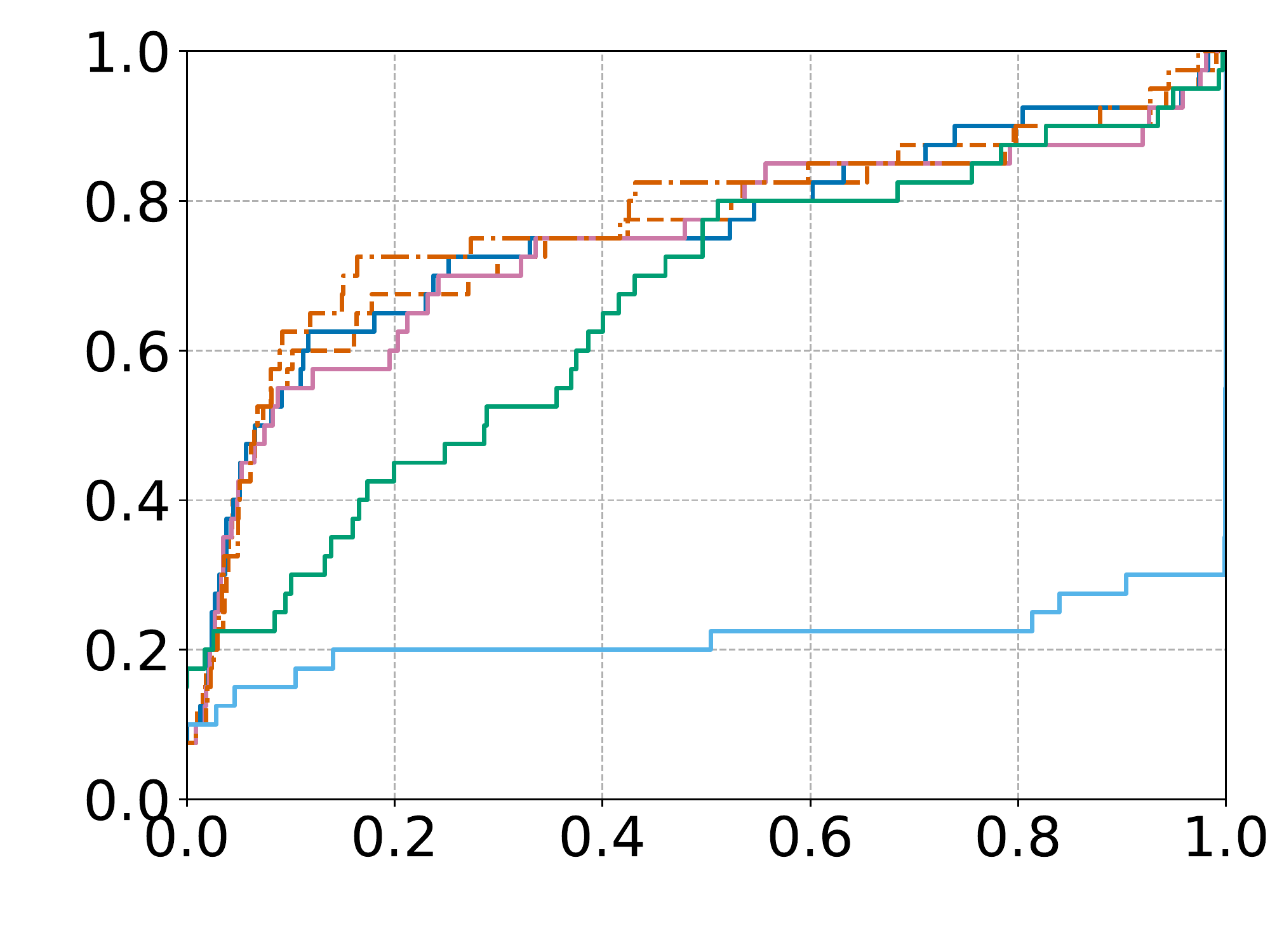}

\\\hline
\begin{tabular}{c}UAR \\ $p=0.5$\end{tabular}
& \includegraphics[width=0.3\textwidth,valign=c,trim={0 0 0 0},clip]{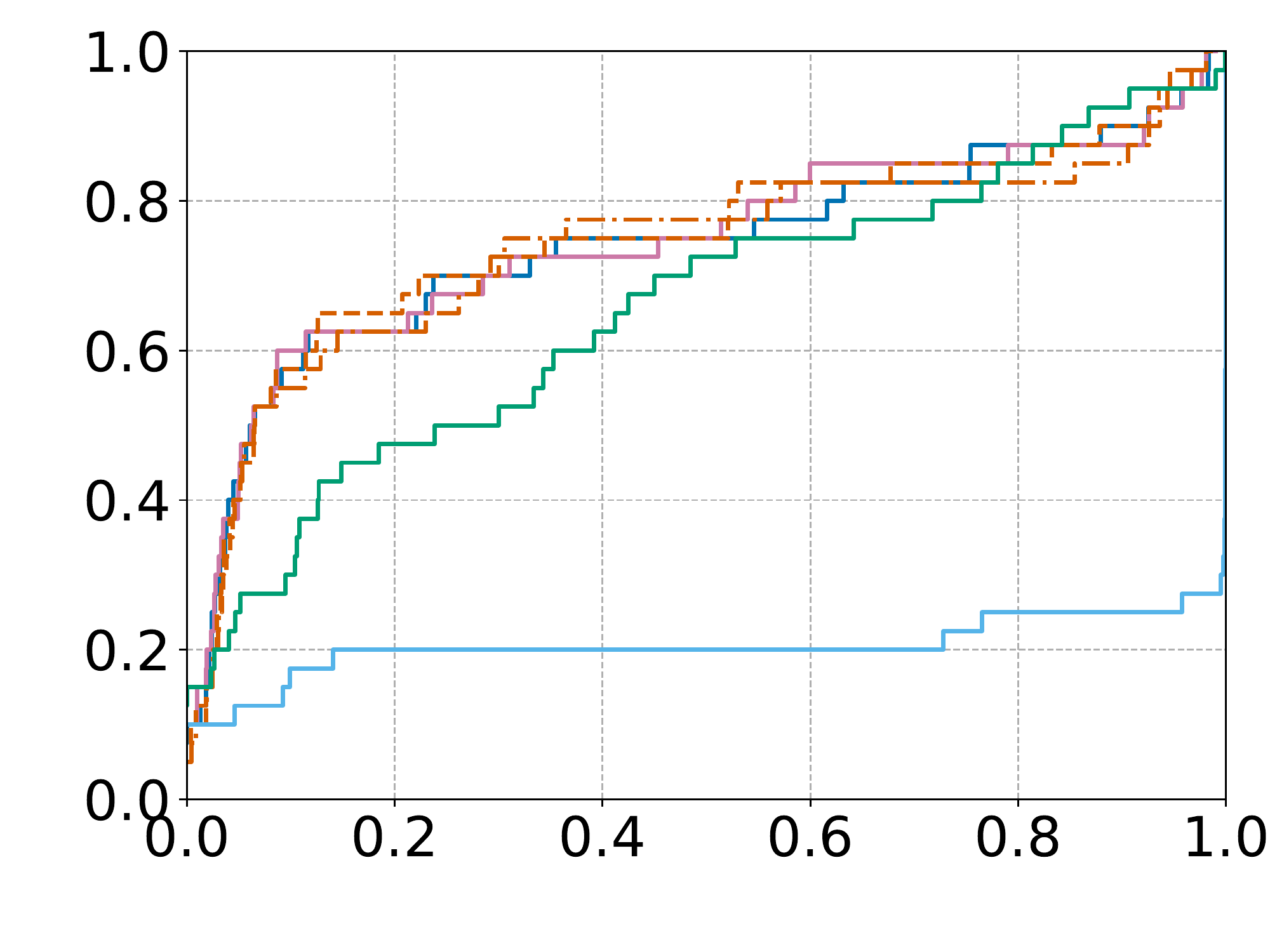}

\\\bottomrule
\multicolumn{2}{c}{
		\begin{subfigure}[t]{\textwidth}
    \includegraphics[width=\textwidth,trim={0 0 0 0},clip]{figs/cdfs/legends/legend_Lambda=8.pdf}
		\end{subfigure} }
\end{tabular}
\caption{Cumulative distribution functions (CDFs) for all evaluated algorithms, against different noise settings and warm start ratios in $\cbr{184.0}$. All CB algorithms use $\epsilon$-greedy with $\epsilon=0.0125$. In each of the above plots, the $x$ axis represents scores, while the $y$ axis represents the CDF values.}
\label{fig:cdfs-eps=0.0125-3}
\end{figure}

\begin{figure}[H]
\centering
\begin{tabular}{c | @{}c@{ }c@{ }c@{}} 
\toprule
& \multicolumn{3}{c}{ Ratio }
\\
Noise & 2.875 & 5.75 & 11.5
\\\midrule
\begin{tabular}{c}MAJ \\ $p=0.5$\end{tabular}
 & \includegraphics[width=0.29\textwidth,valign=c,trim={0 0 0 0},clip]{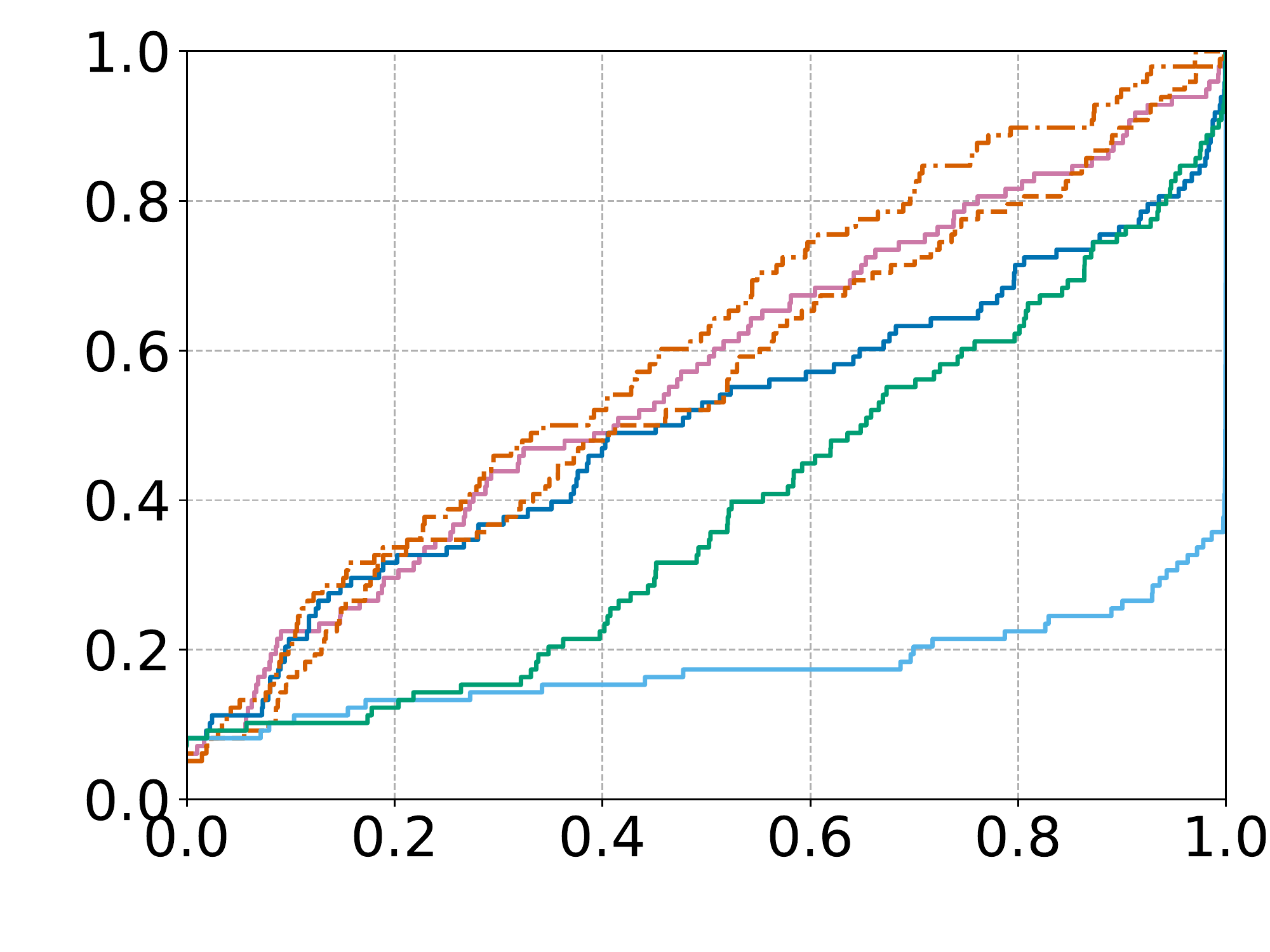}
&
\includegraphics[width=0.29\textwidth,valign=c,trim={0 0 0 0},clip]{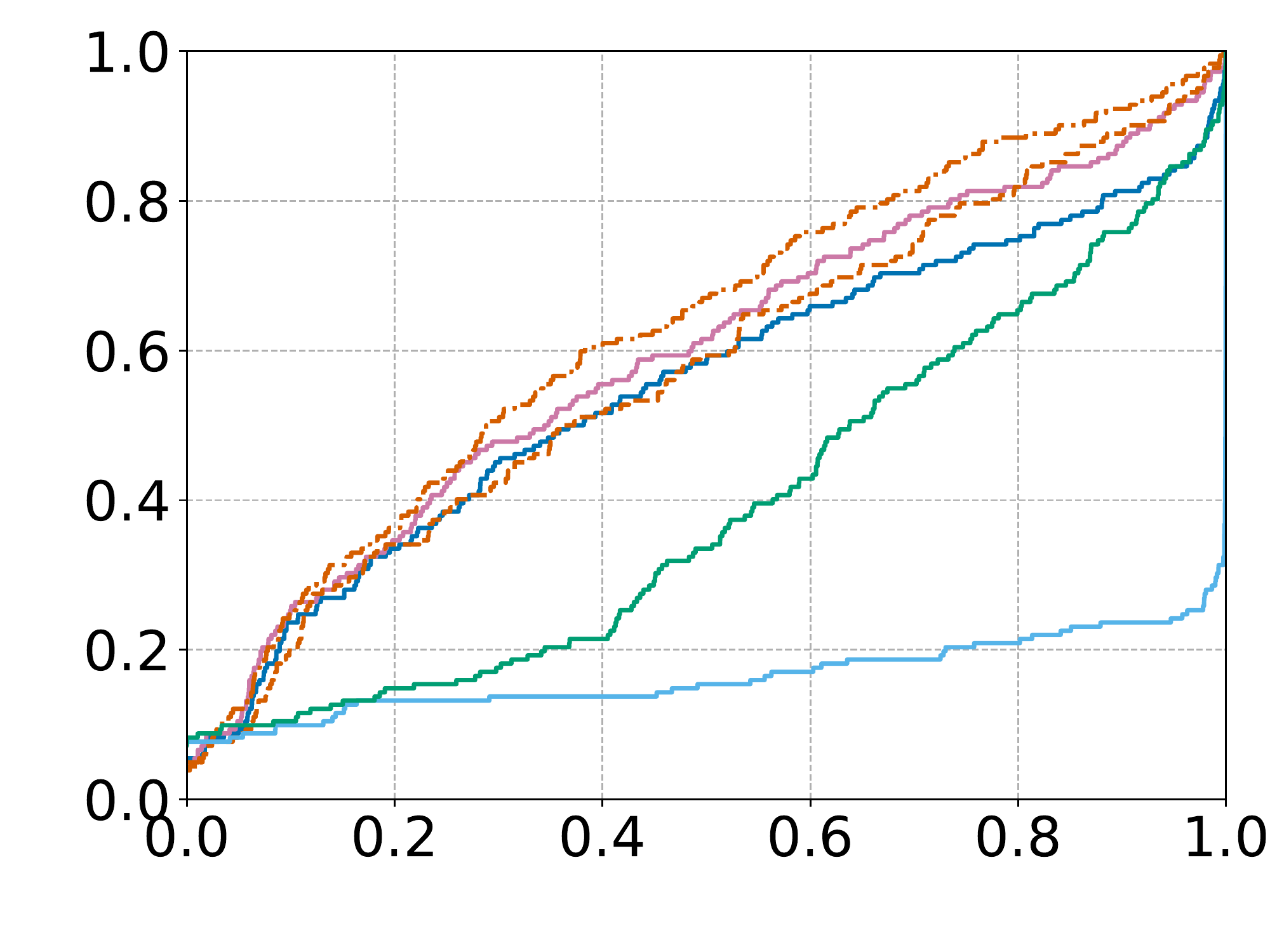}
&
\includegraphics[width=0.29\textwidth,valign=c,trim={0 0 0 0},clip]{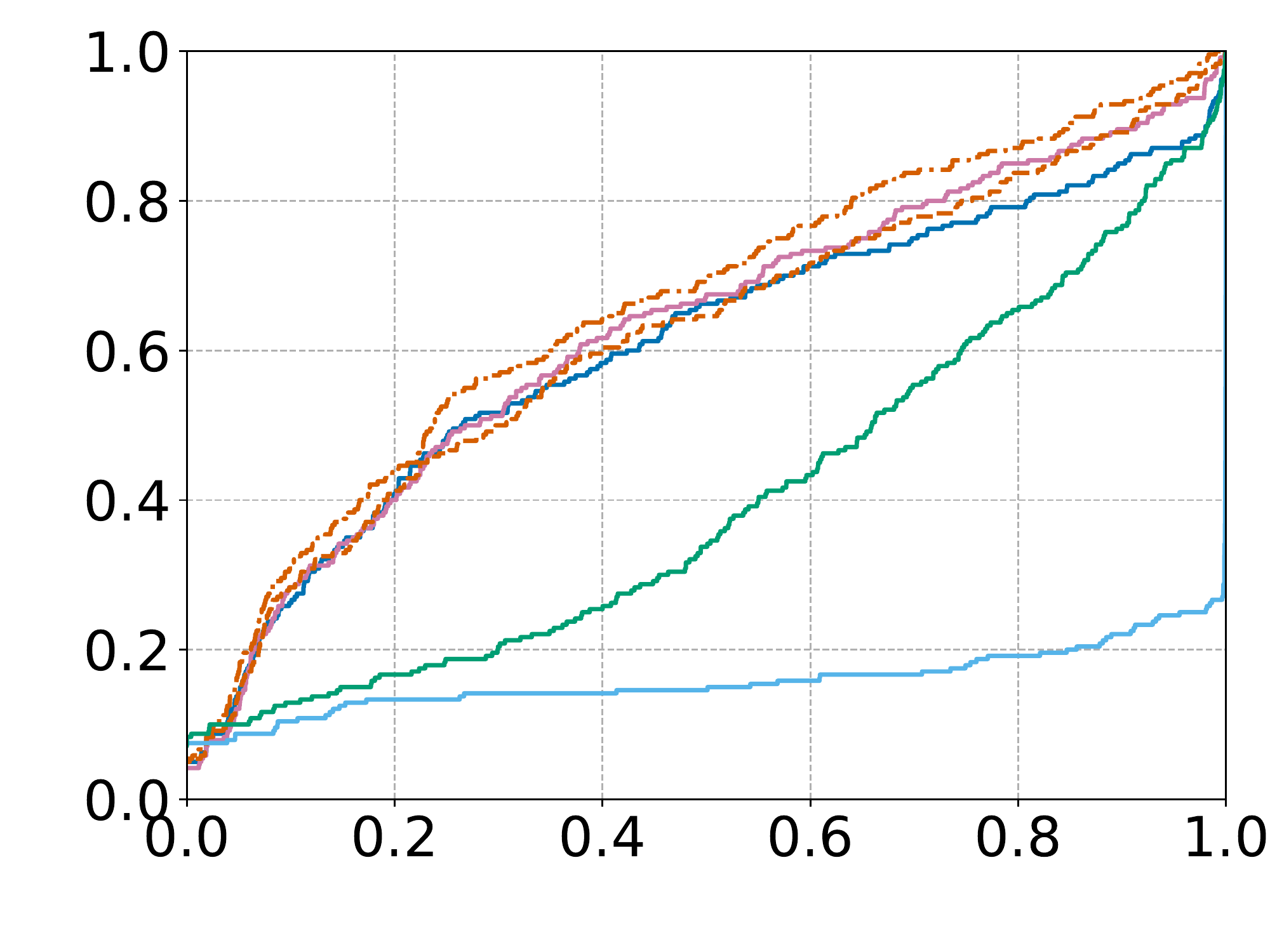}
\\\hline
\begin{tabular}{c}CYC \\ $p=0.5$\end{tabular}
& \includegraphics[width=0.29\textwidth,valign=c,trim={0 0 0 0},clip]{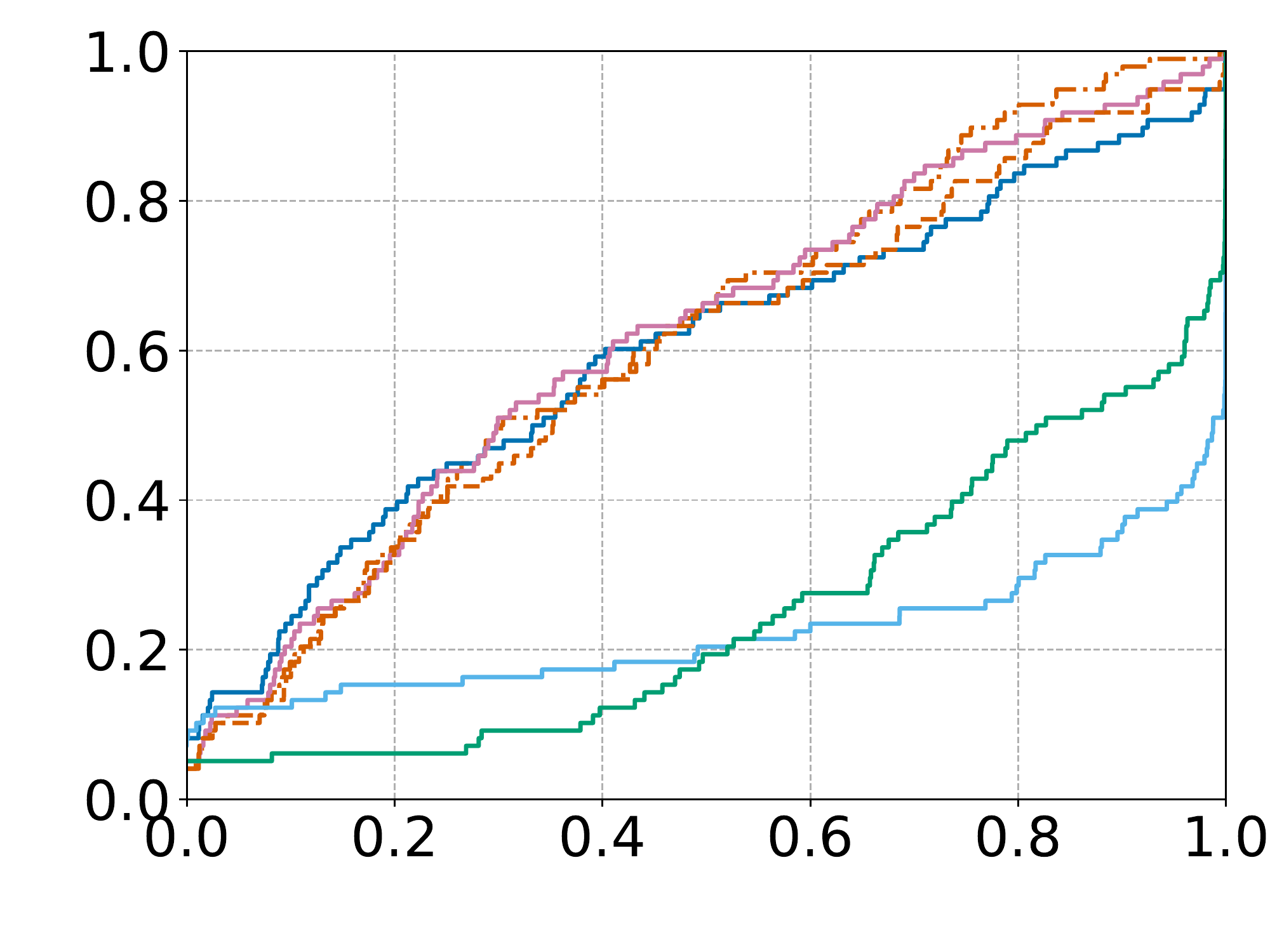}
&
\includegraphics[width=0.29\textwidth,valign=c,trim={0 0 0 0},clip]{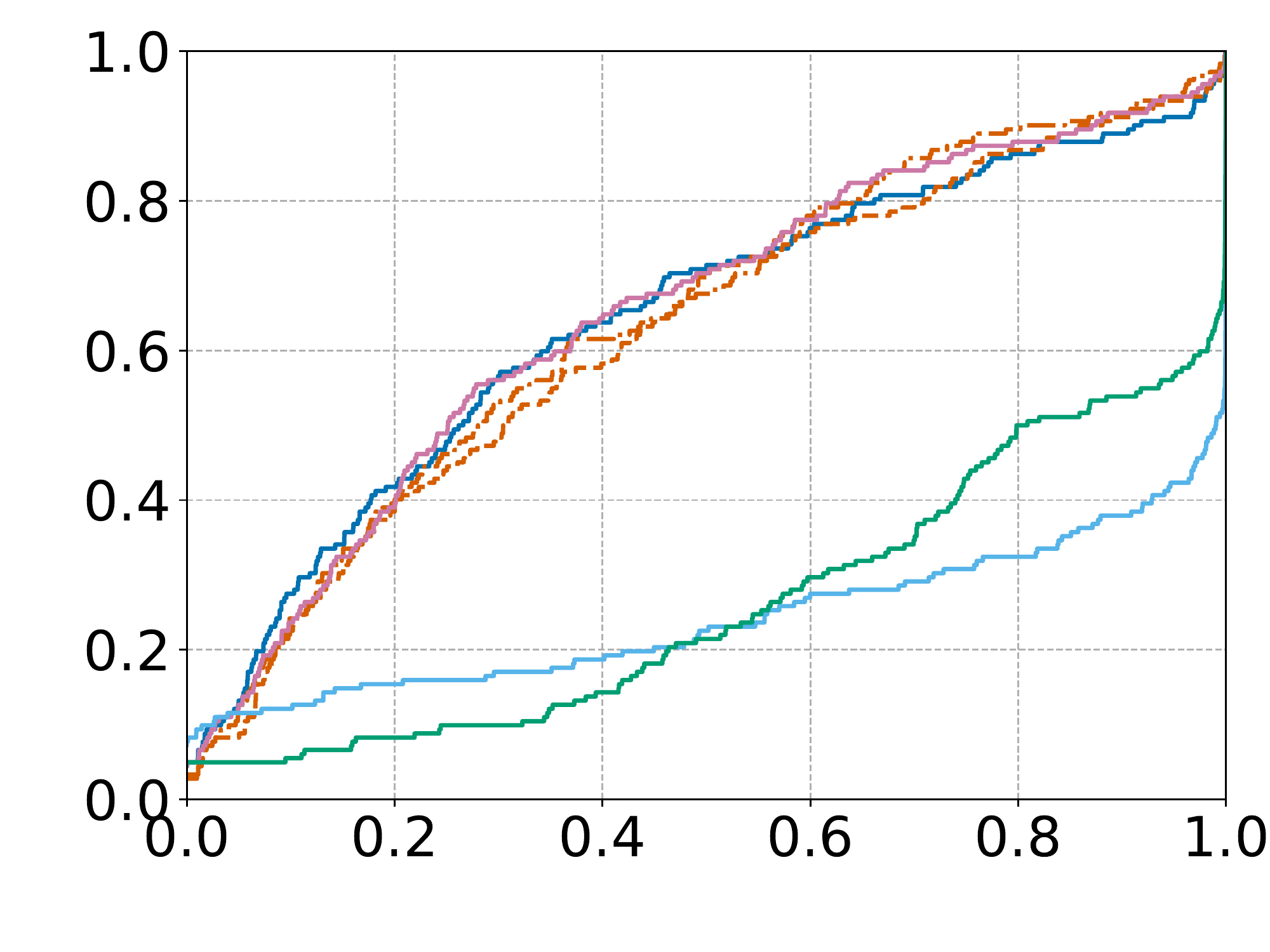}
&
\includegraphics[width=0.29\textwidth,valign=c,trim={0 0 0 0},clip]{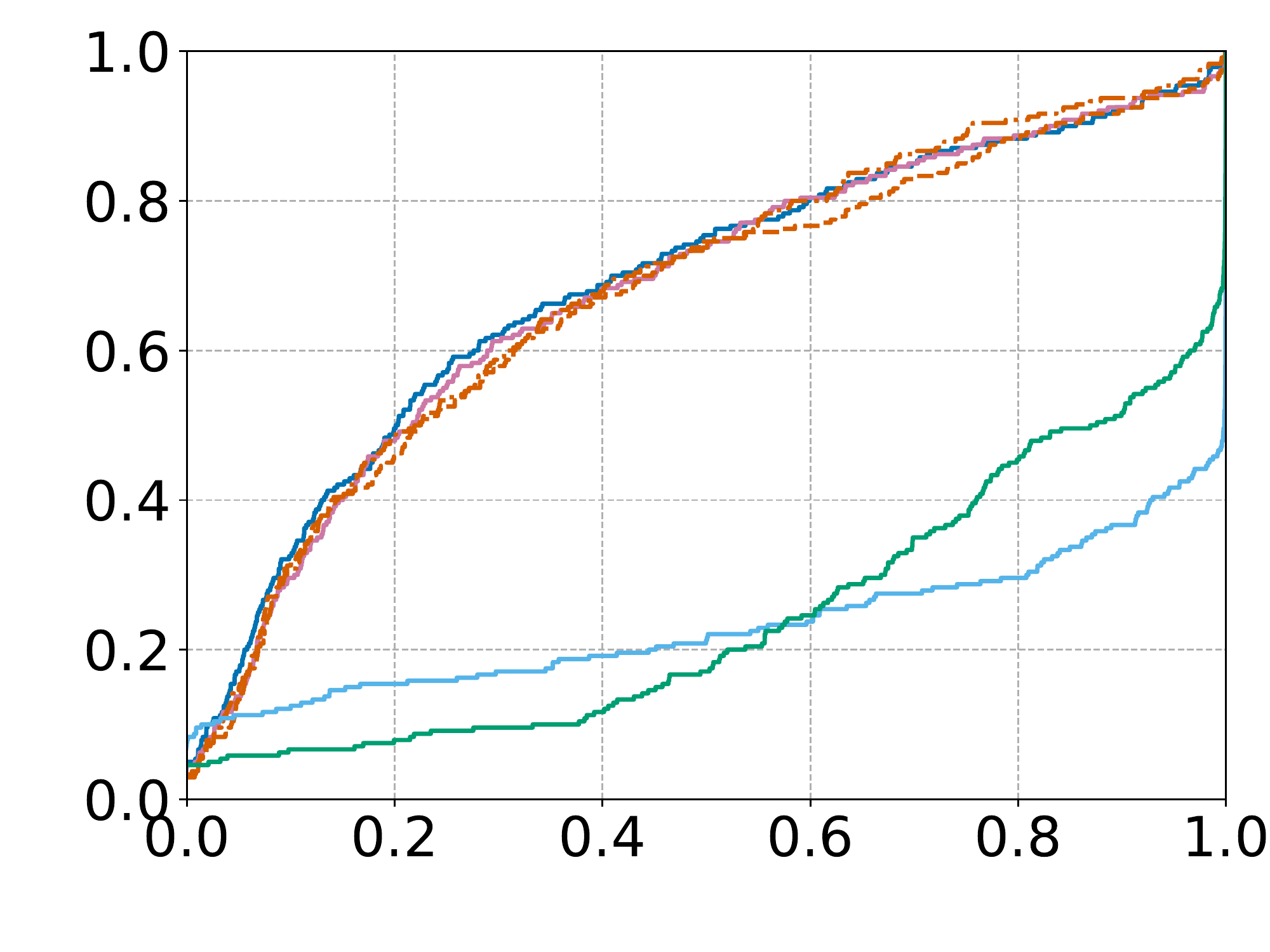}
\\\hline
\begin{tabular}{c}UAR \\ $p=1$\end{tabular}
& \includegraphics[width=0.29\textwidth,valign=c,trim={0 0 0 0},clip]{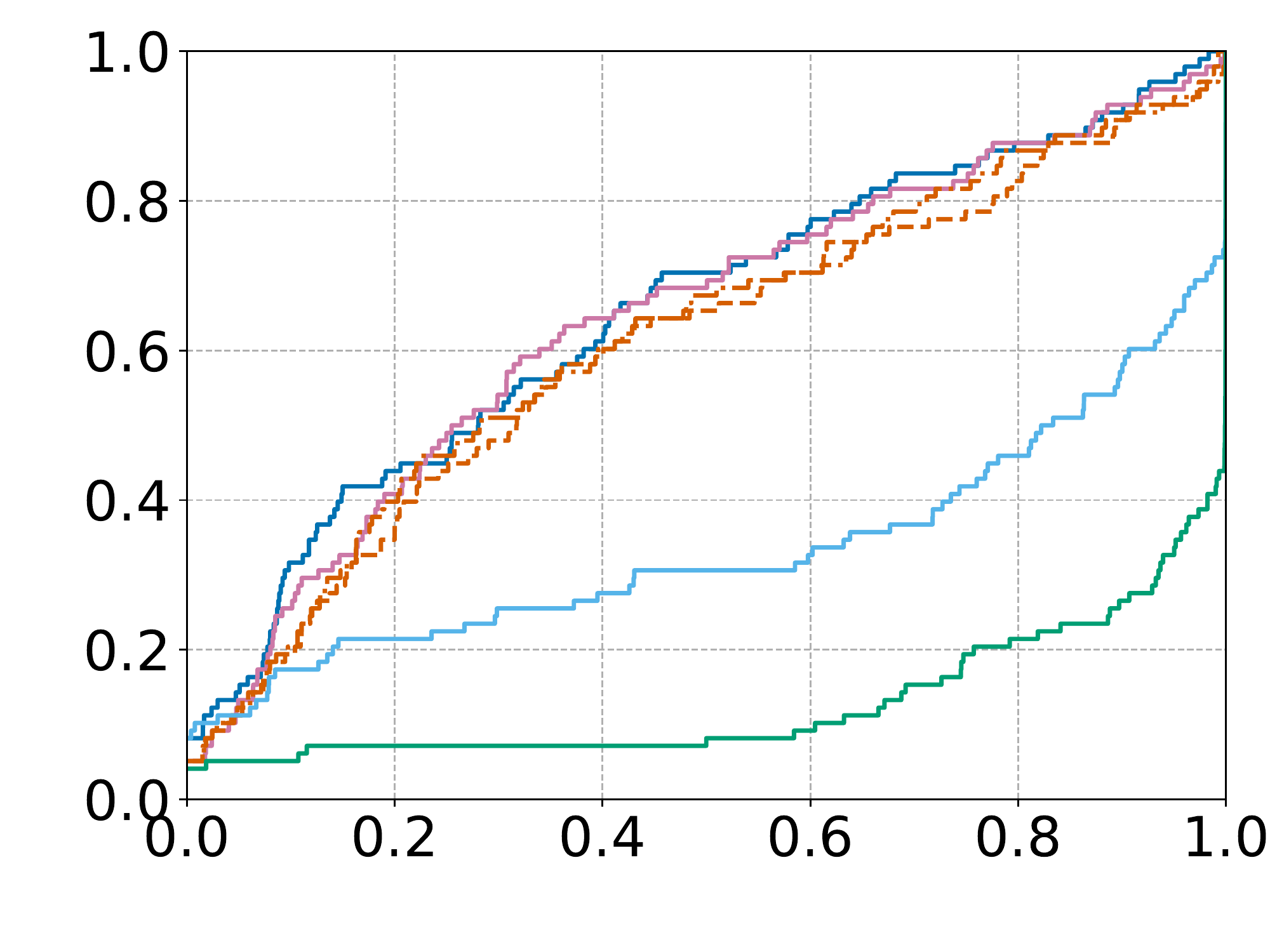}
&
\includegraphics[width=0.29\textwidth,valign=c,trim={0 0 0 0},clip]{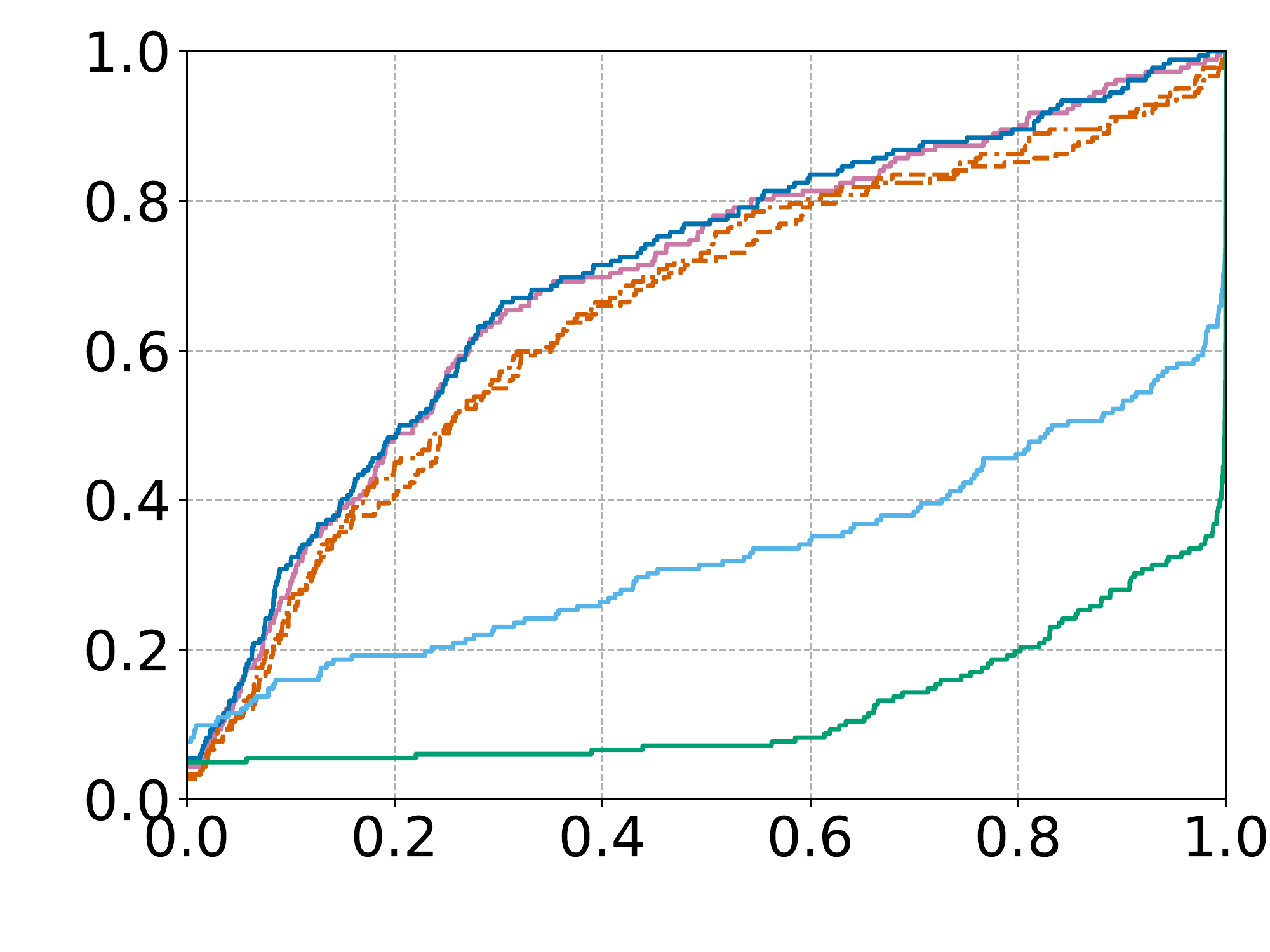}
&
\includegraphics[width=0.29\textwidth,valign=c,trim={0 0 0 0},clip]{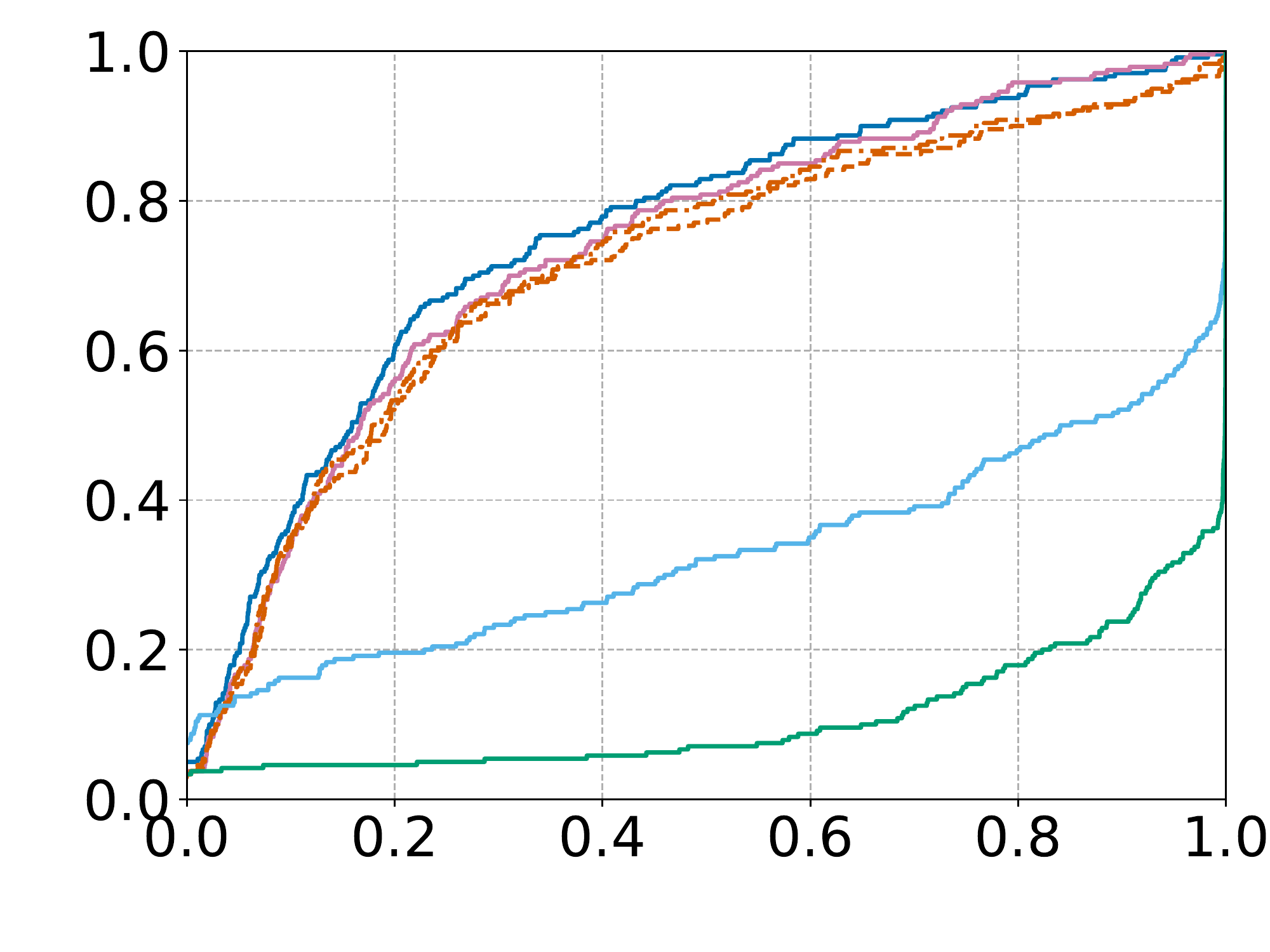}
\\\hline
\begin{tabular}{c}MAJ \\ $p=1$\end{tabular}
& \includegraphics[width=0.29\textwidth,valign=c,trim={0 0 0 0},clip]{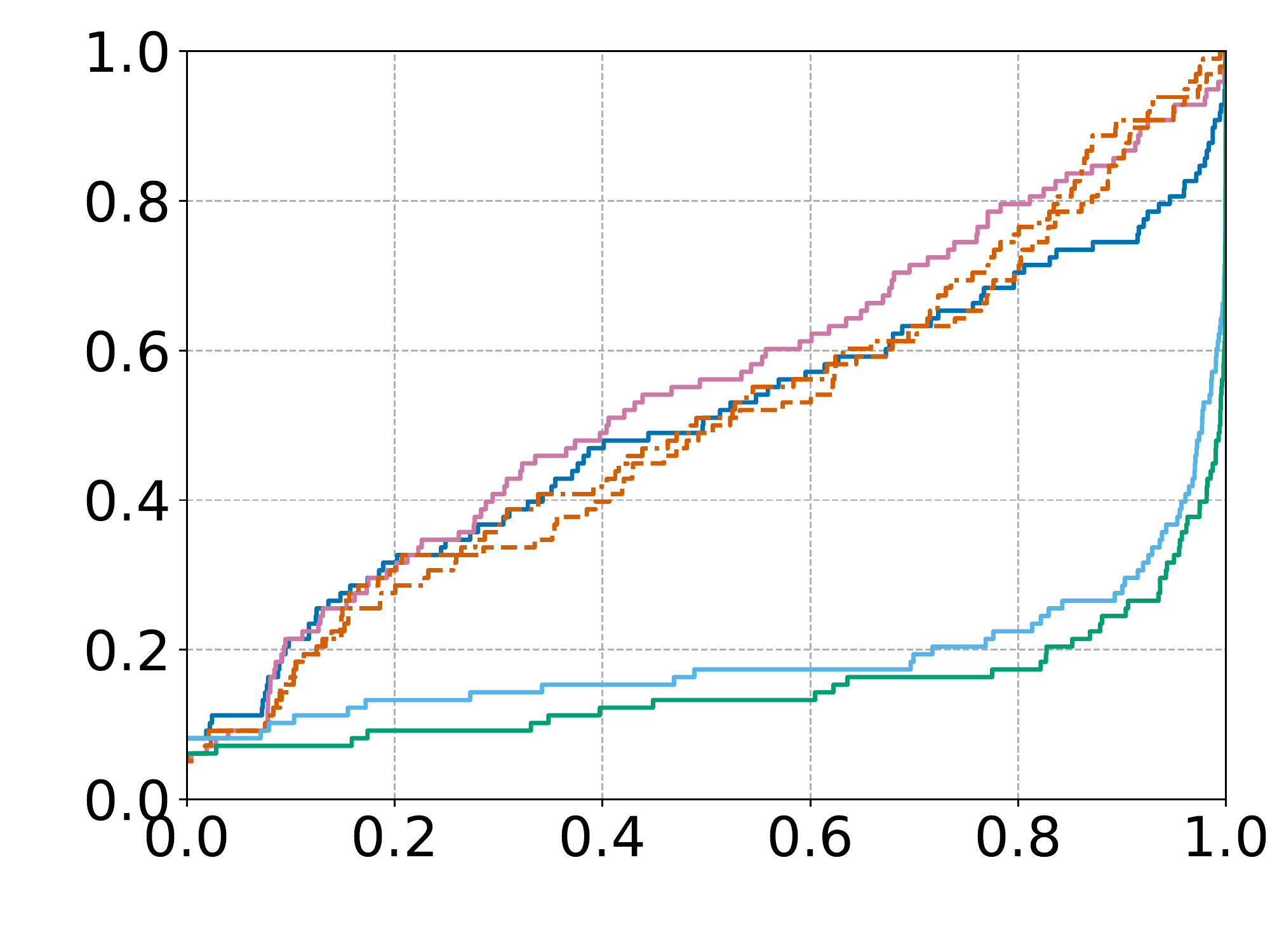}
&
\includegraphics[width=0.29\textwidth,valign=c,trim={0 0 0 0},clip]{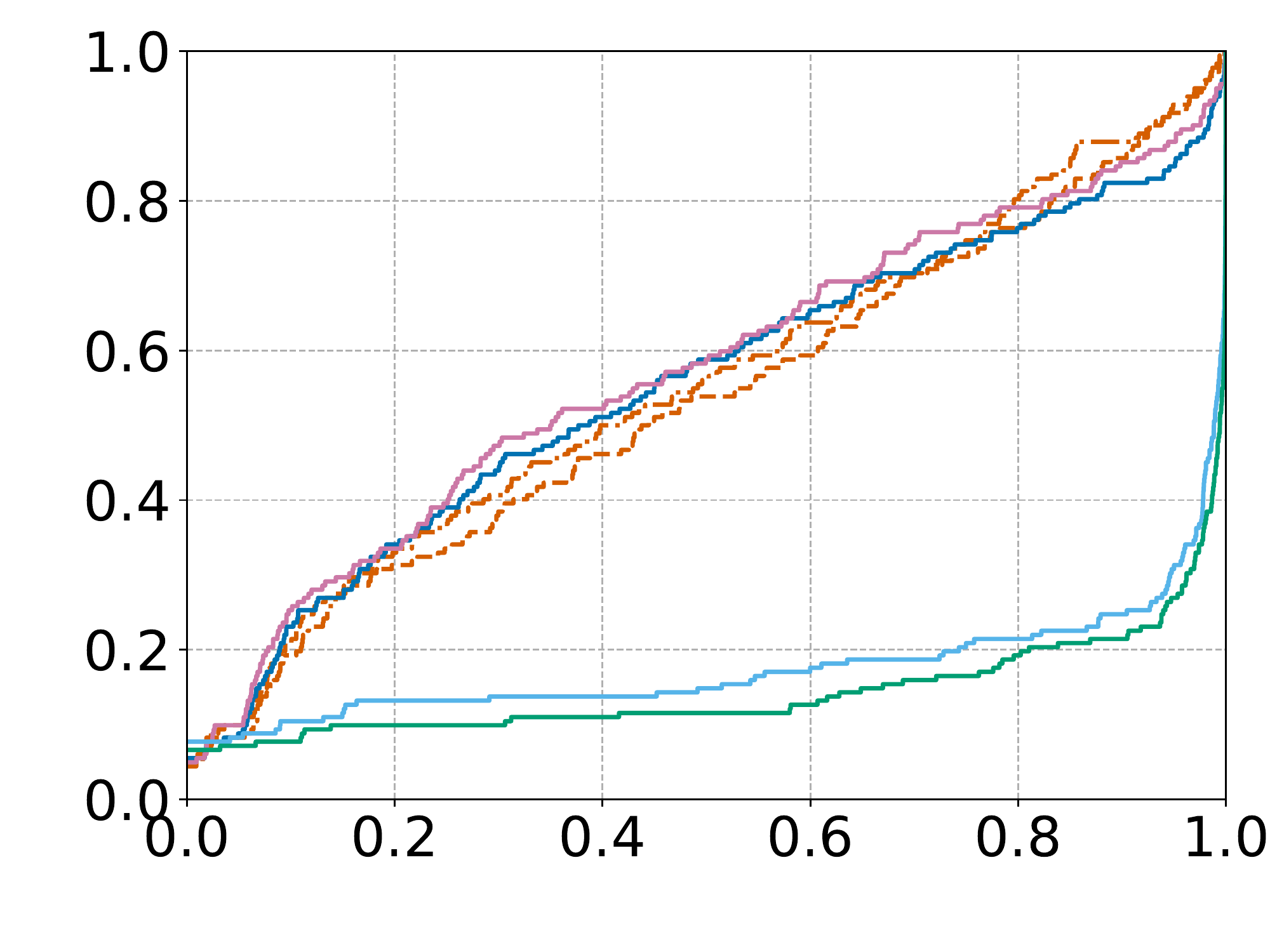}
&
\includegraphics[width=0.29\textwidth,valign=c,trim={0 0 0 0},clip]{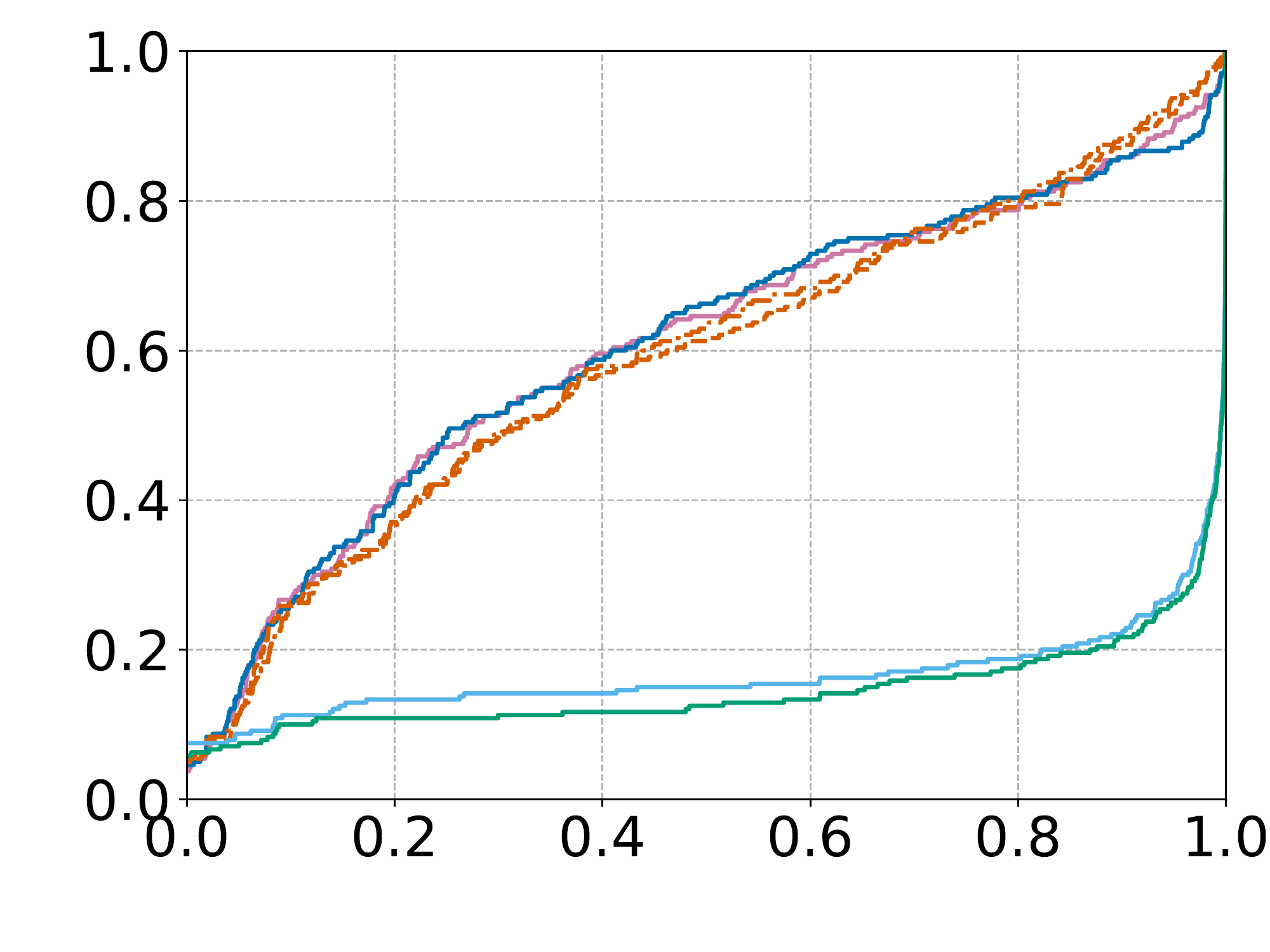}
\\\hline
\begin{tabular}{c}CYC \\ $p=1$\end{tabular}
& \includegraphics[width=0.29\textwidth,valign=c,trim={0 0 0 0},clip]{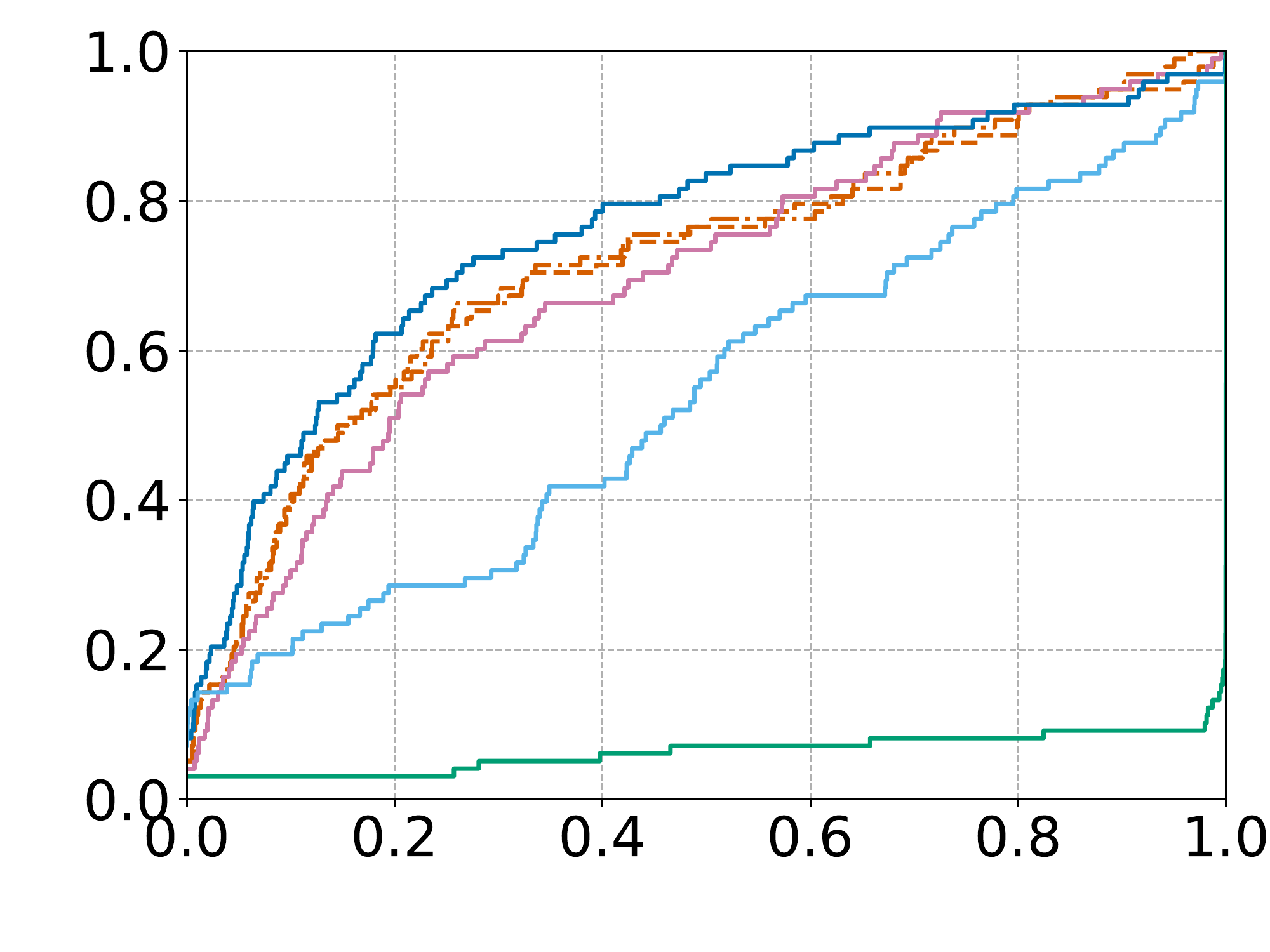}
&
\includegraphics[width=0.29\textwidth,valign=c,trim={0 0 0 0},clip]{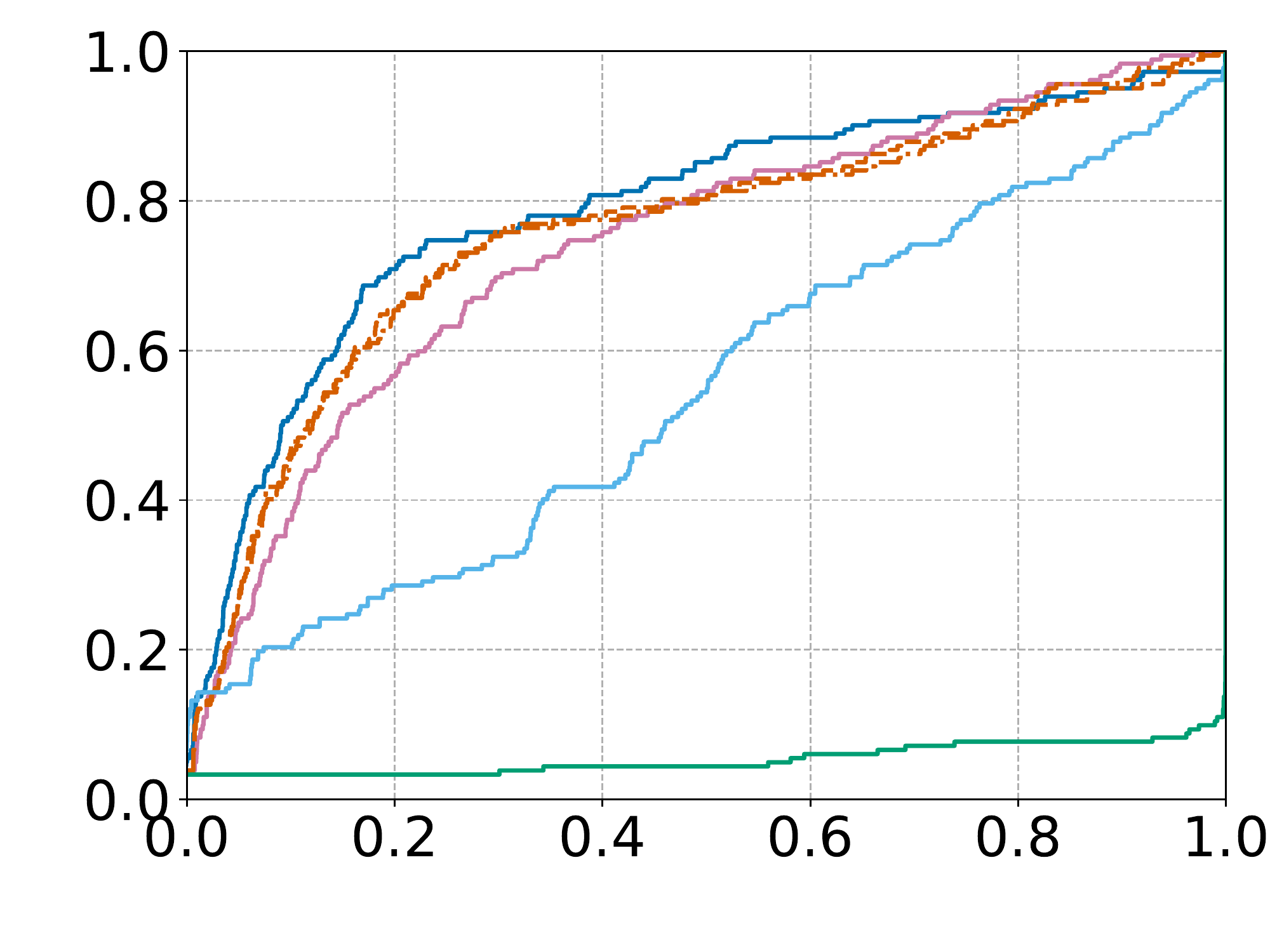}
&
\includegraphics[width=0.29\textwidth,valign=c,trim={0 0 0 0},clip]{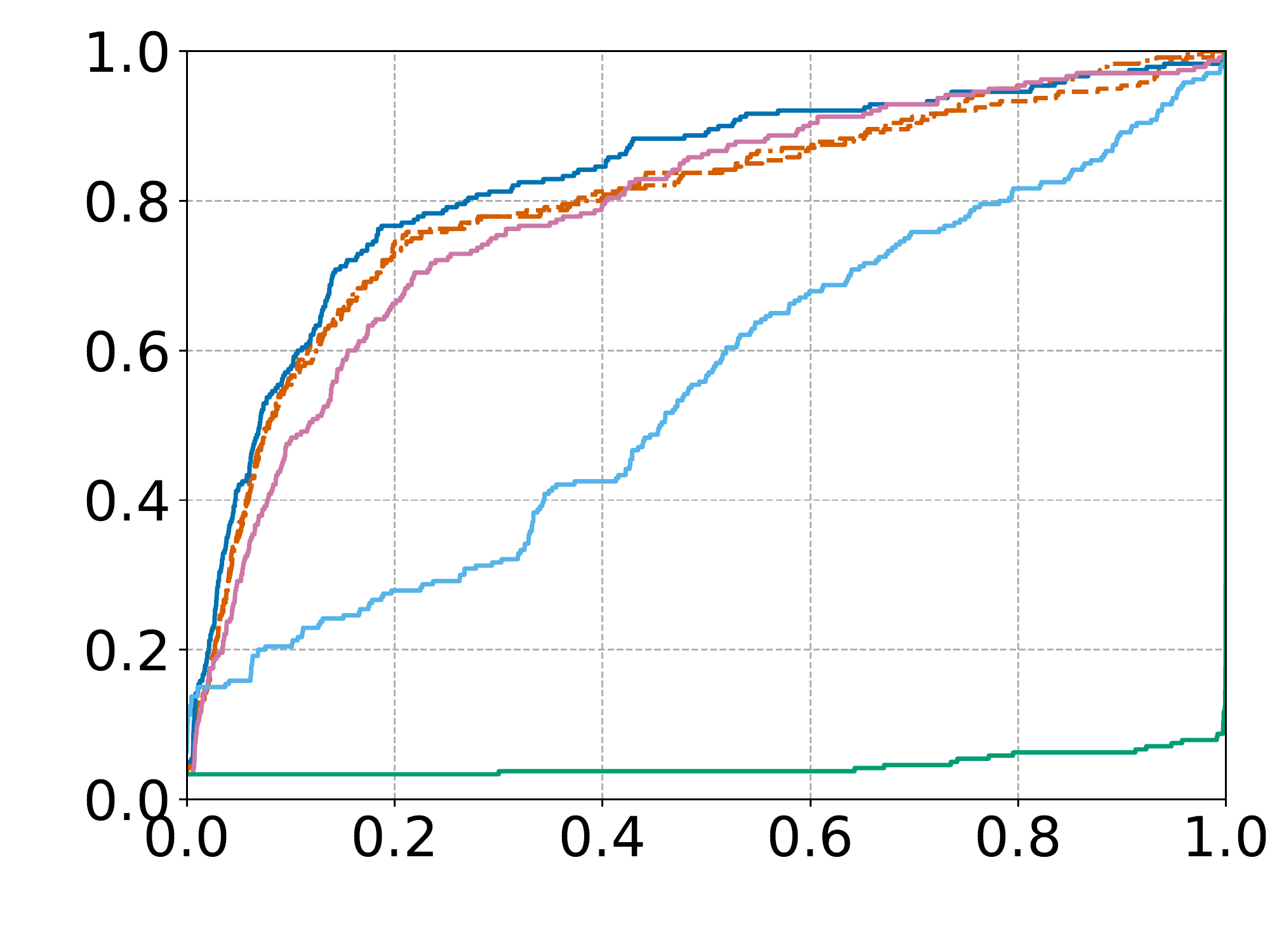}
\\\bottomrule
\multicolumn{4}{c}{
		\begin{subfigure}[t]{\textwidth}
    \includegraphics[width=\textwidth,trim={0 0 0 0},clip]{figs/cdfs/legends/legend_Lambda=8.pdf}
		\end{subfigure} }
\end{tabular}
\caption{Cumulative distribution functions (CDFs) for all evaluated algorithms, against different noise settings and warm start ratios in $\cbr{2.875,5.75,11.5}$. All CB algorithms use $\epsilon$-greedy with $\epsilon=0.0125$. In each of the above plots, the $x$ axis represents scores, while the $y$ axis represents the CDF values.}
\label{fig:cdfs-eps=0.0125-4}
\end{figure}

\begin{figure}[H]
\centering
\begin{tabular}{c | @{}c@{ }c@{ }c@{}} 
\toprule
& \multicolumn{3}{c}{ Ratio }
\\
Noise & 23.0 & 46.0 & 92.0
\\\midrule
\begin{tabular}{c}MAJ \\ $p=0.5$\end{tabular}
 & \includegraphics[width=0.29\textwidth,valign=c,trim={0 0 0 0},clip]{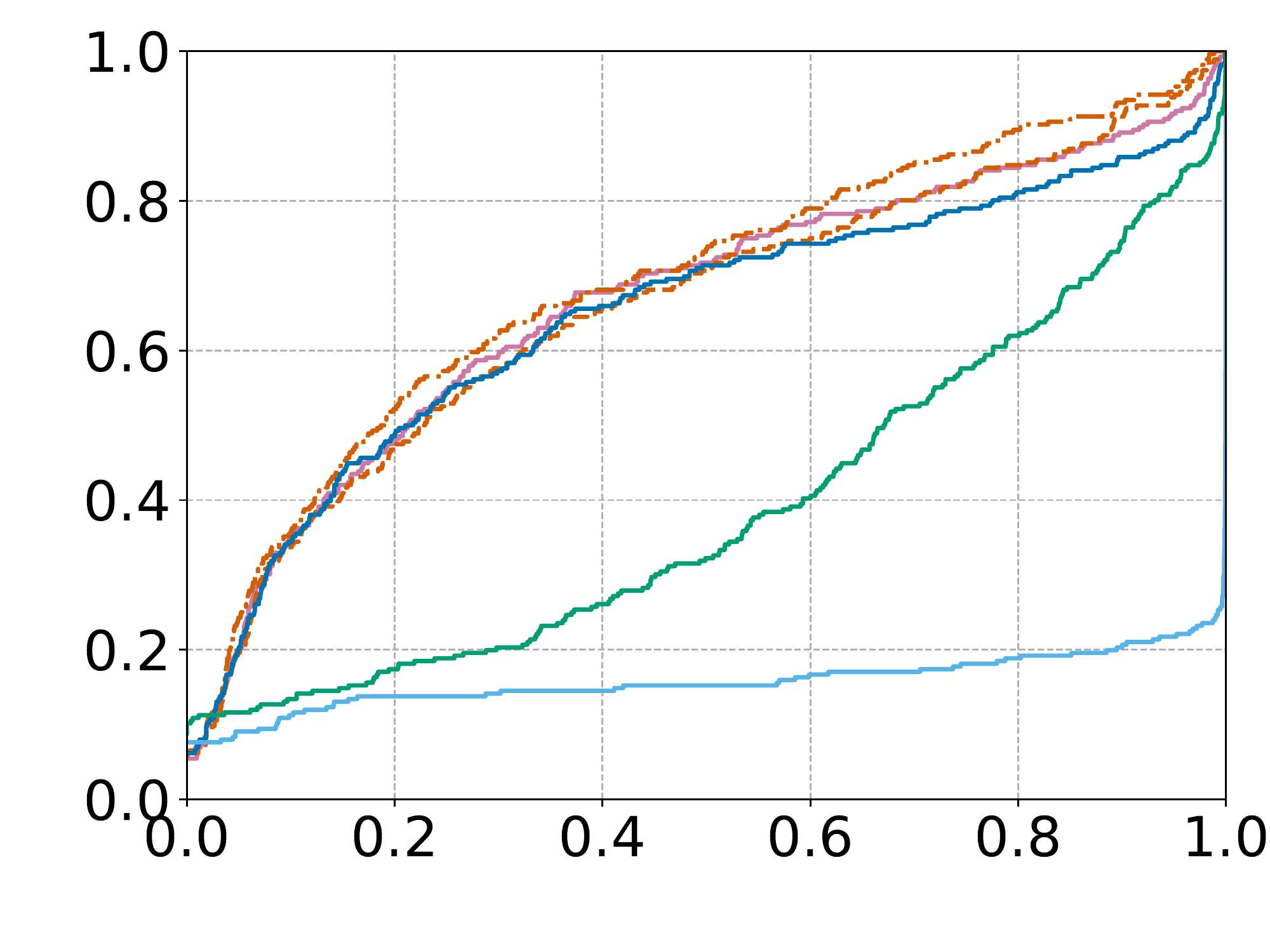}
&
\includegraphics[width=0.29\textwidth,valign=c,trim={0 0 0 0},clip]{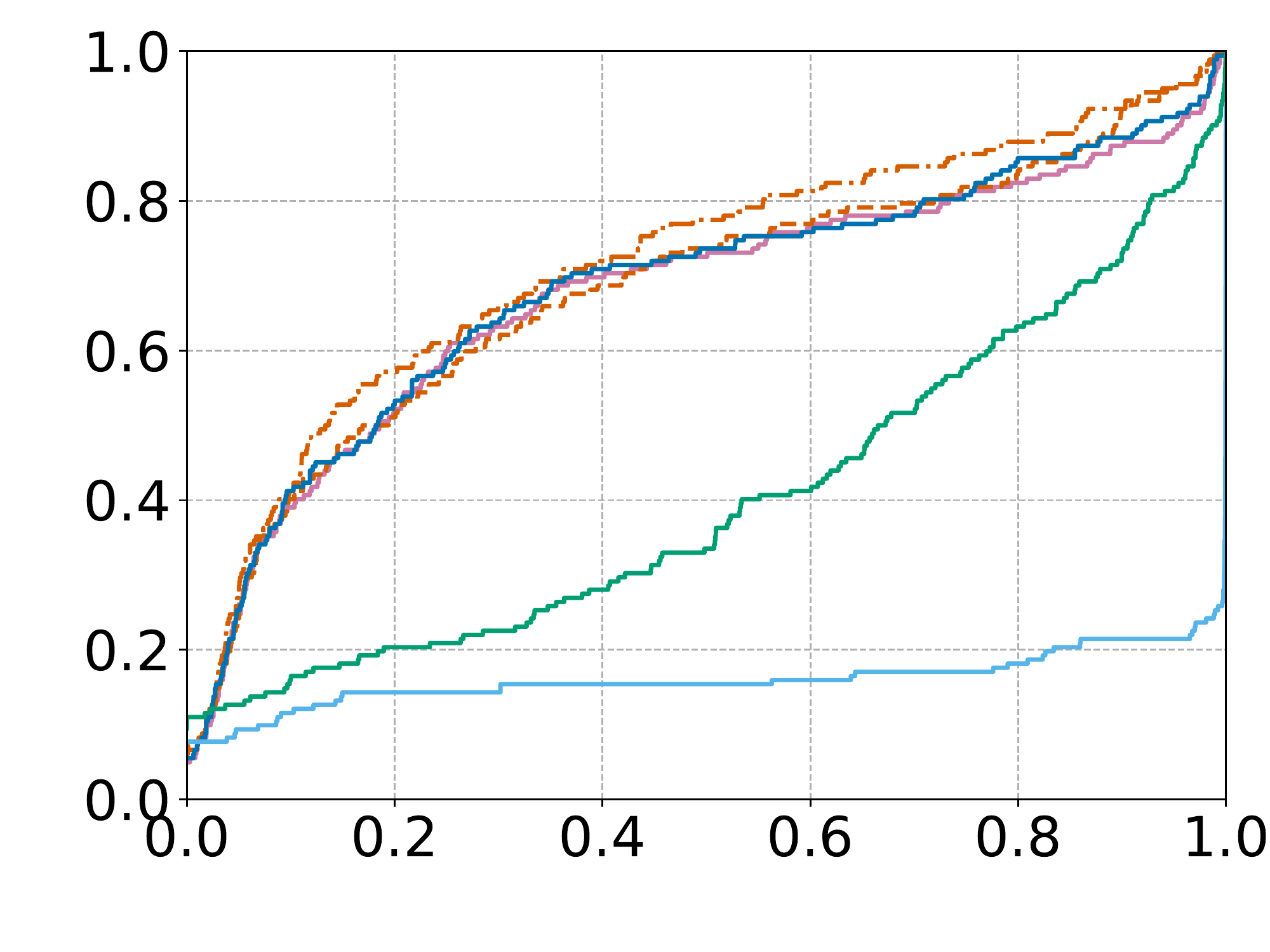}
&
\includegraphics[width=0.29\textwidth,valign=c,trim={0 0 0 0},clip]{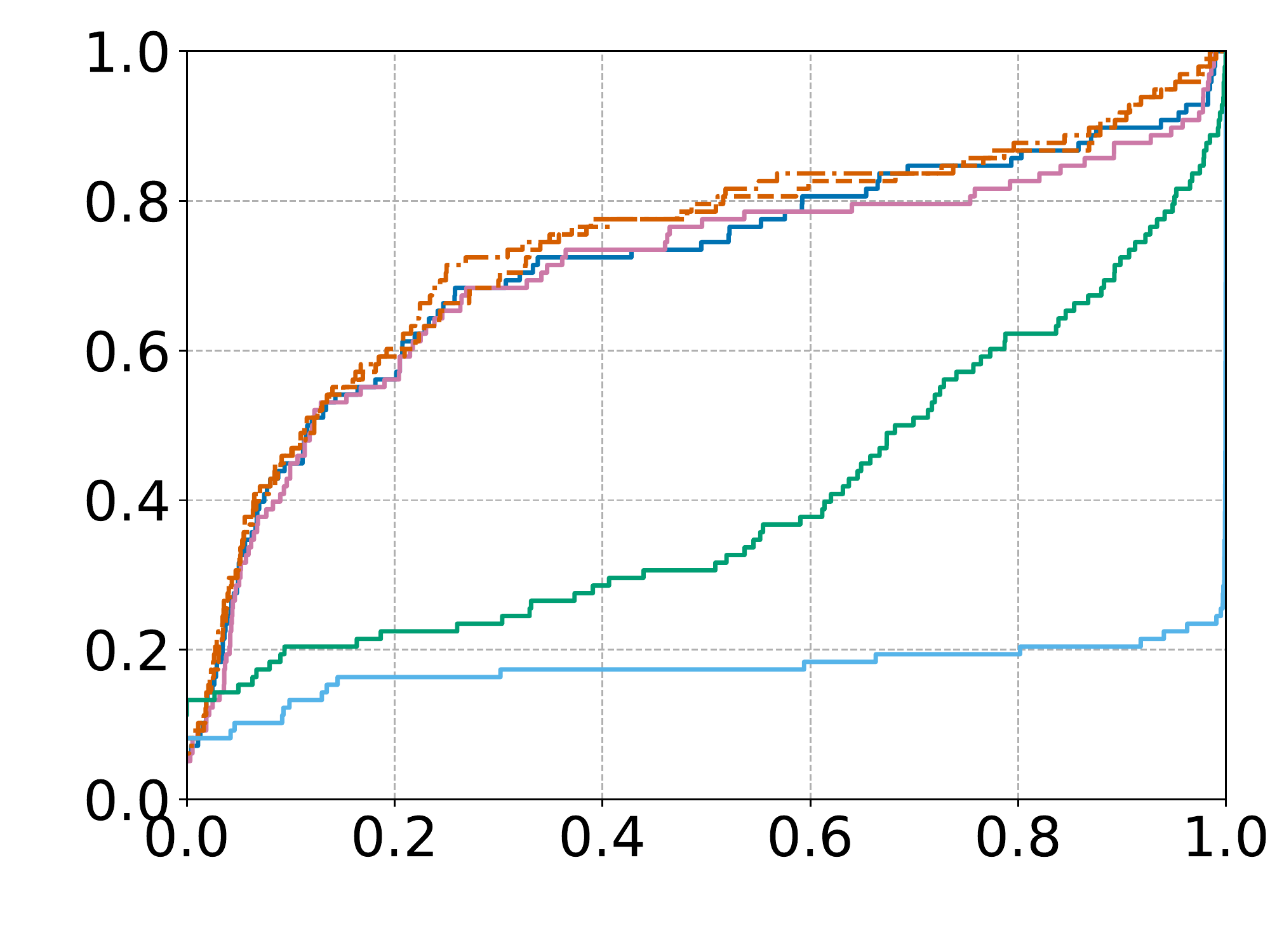}
\\\hline
\begin{tabular}{c}CYC \\ $p=0.5$\end{tabular}
& \includegraphics[width=0.29\textwidth,valign=c,trim={0 0 0 0},clip]{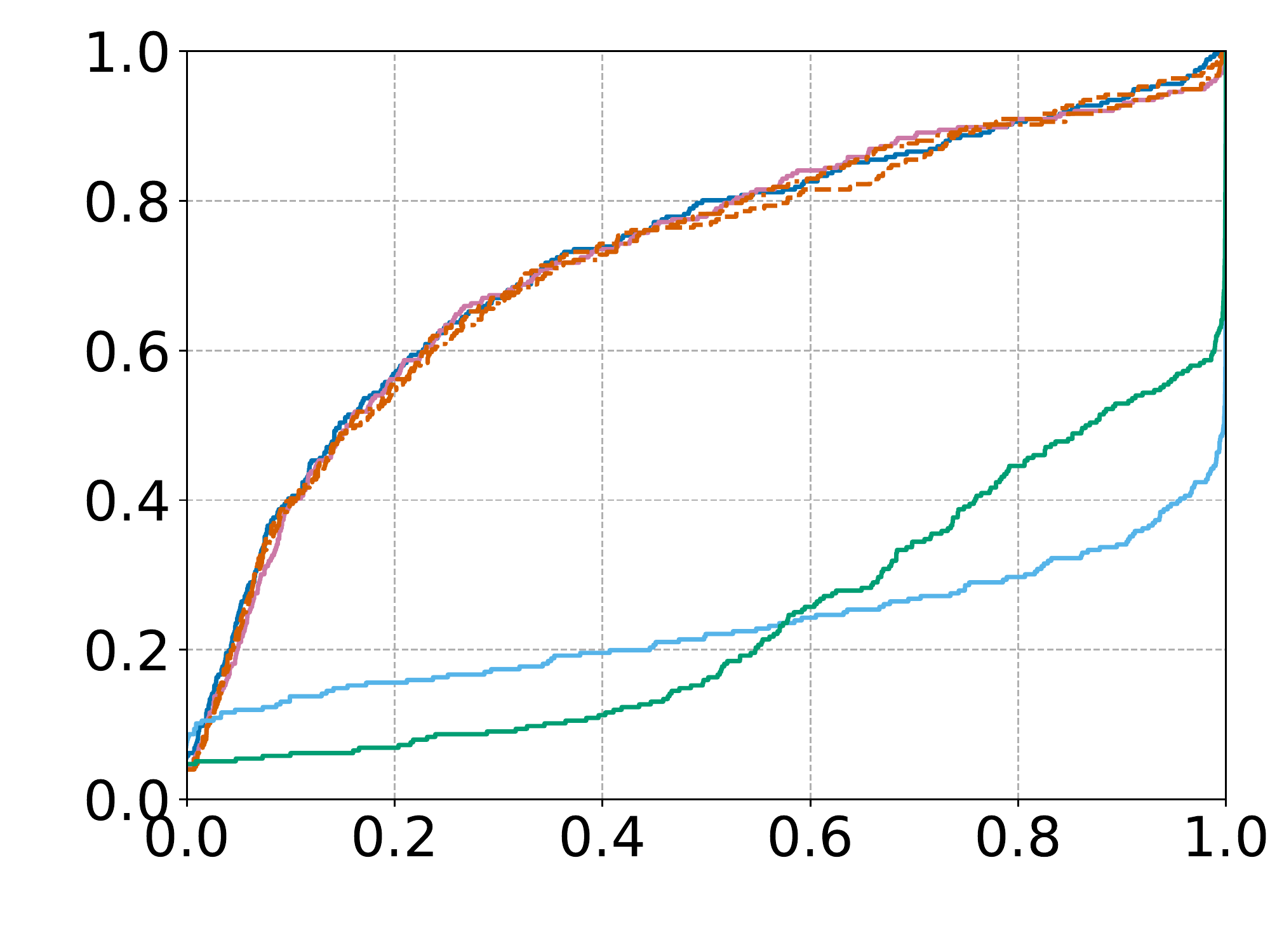}
&
\includegraphics[width=0.29\textwidth,valign=c,trim={0 0 0 0},clip]{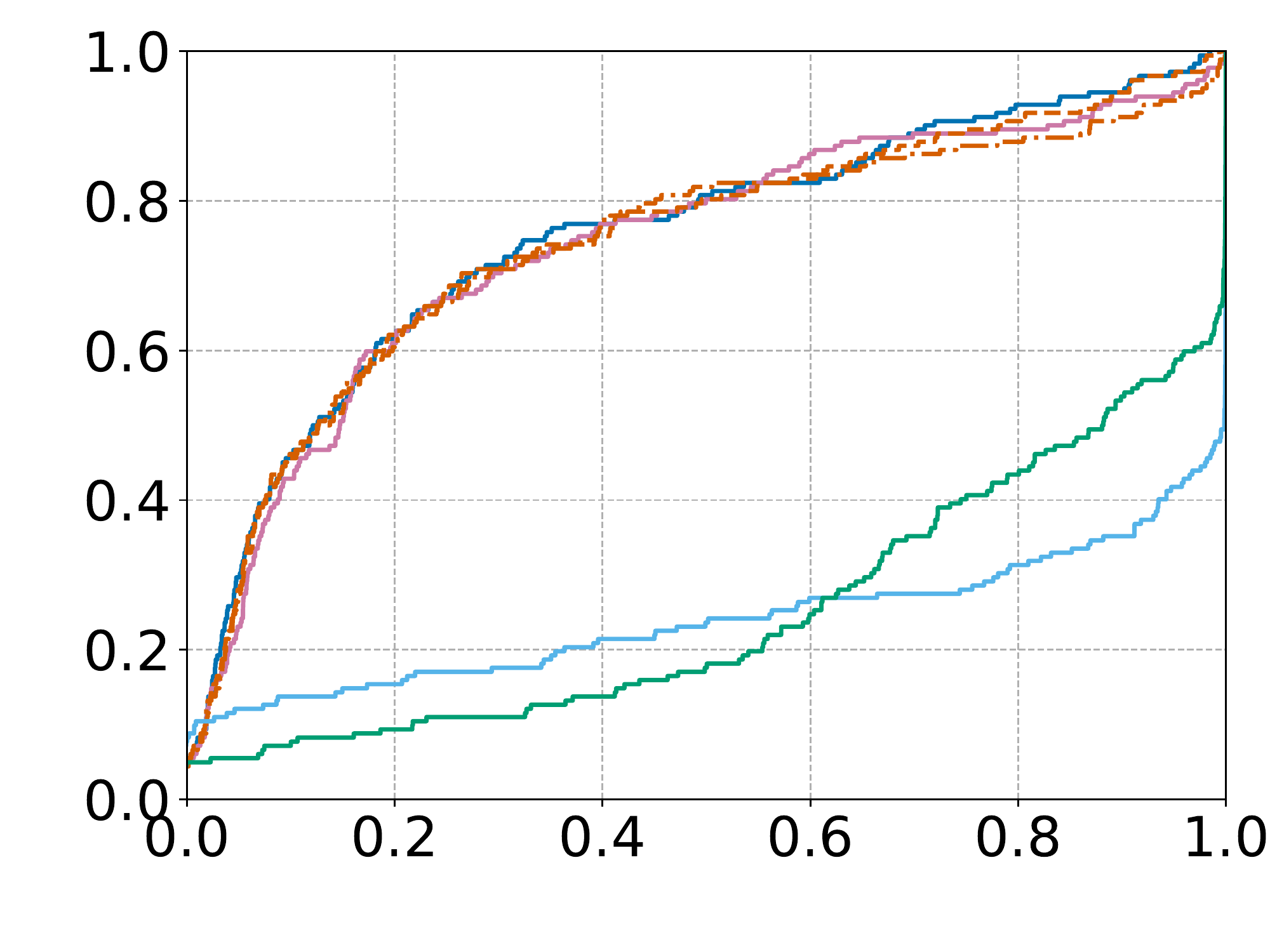}
&
\includegraphics[width=0.29\textwidth,valign=c,trim={0 0 0 0},clip]{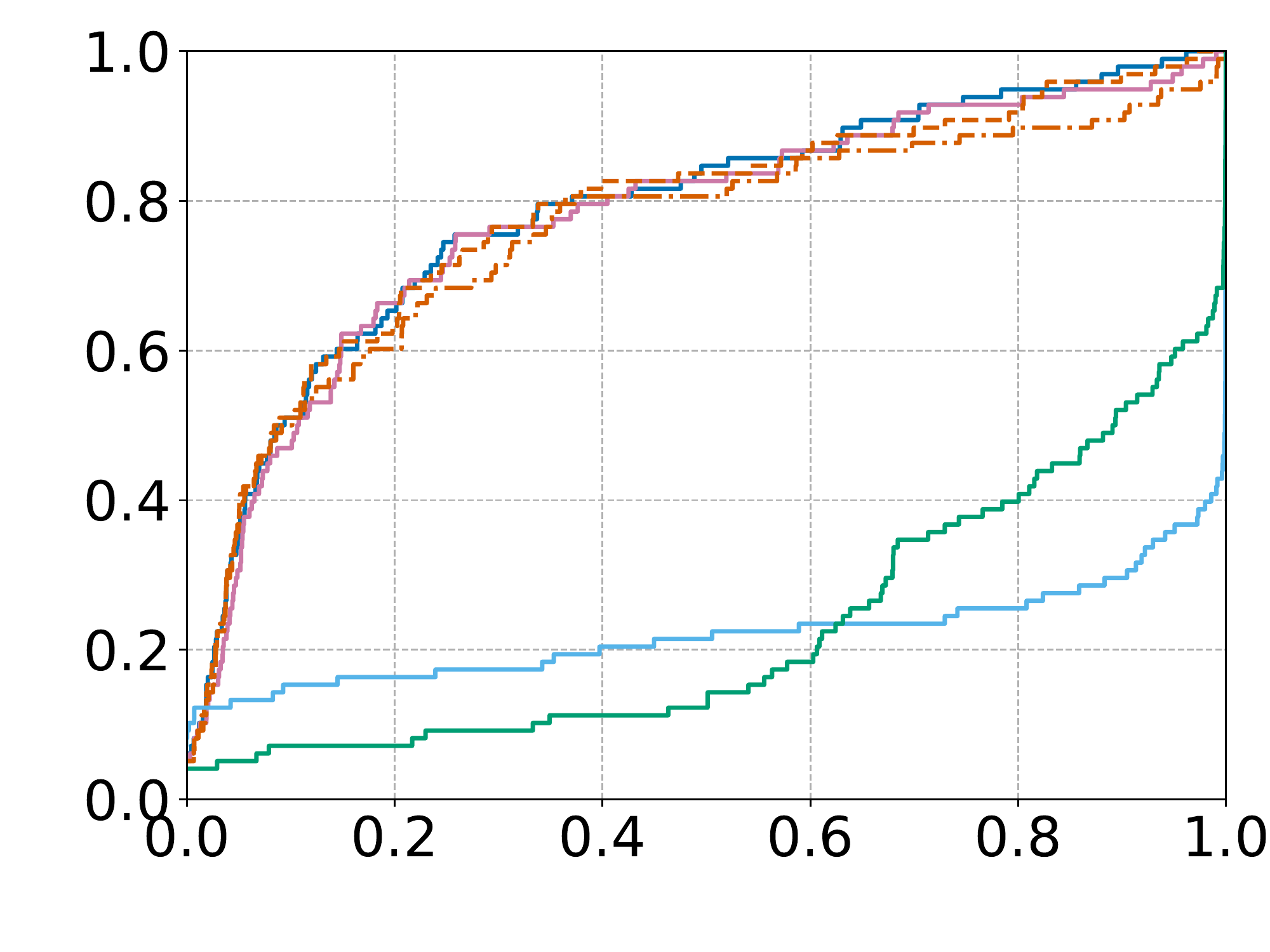}
\\\hline
\begin{tabular}{c}UAR \\ $p=1$\end{tabular}
& \includegraphics[width=0.29\textwidth,valign=c,trim={0 0 0 0},clip]{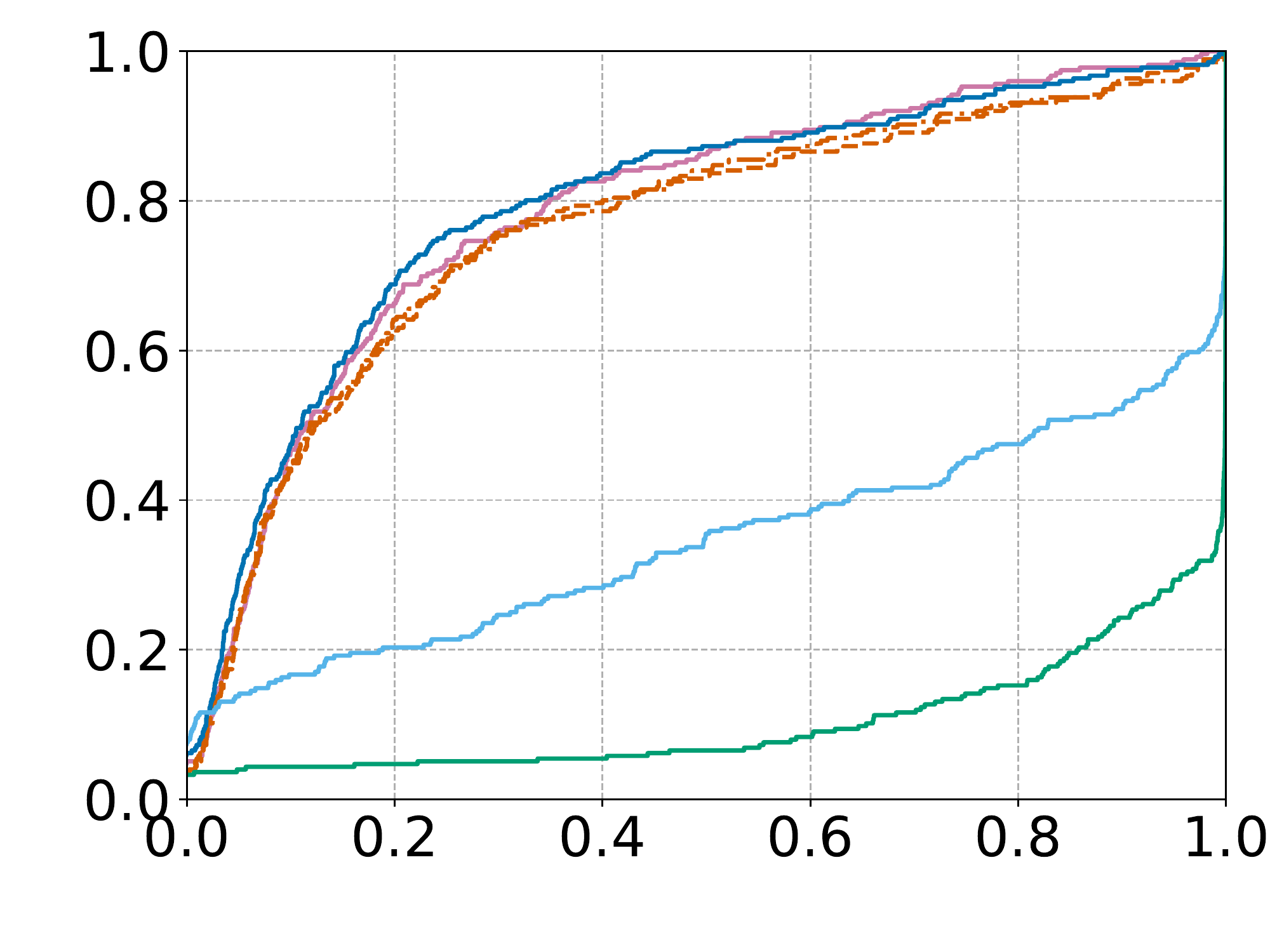}
&
\includegraphics[width=0.29\textwidth,valign=c,trim={0 0 0 0},clip]{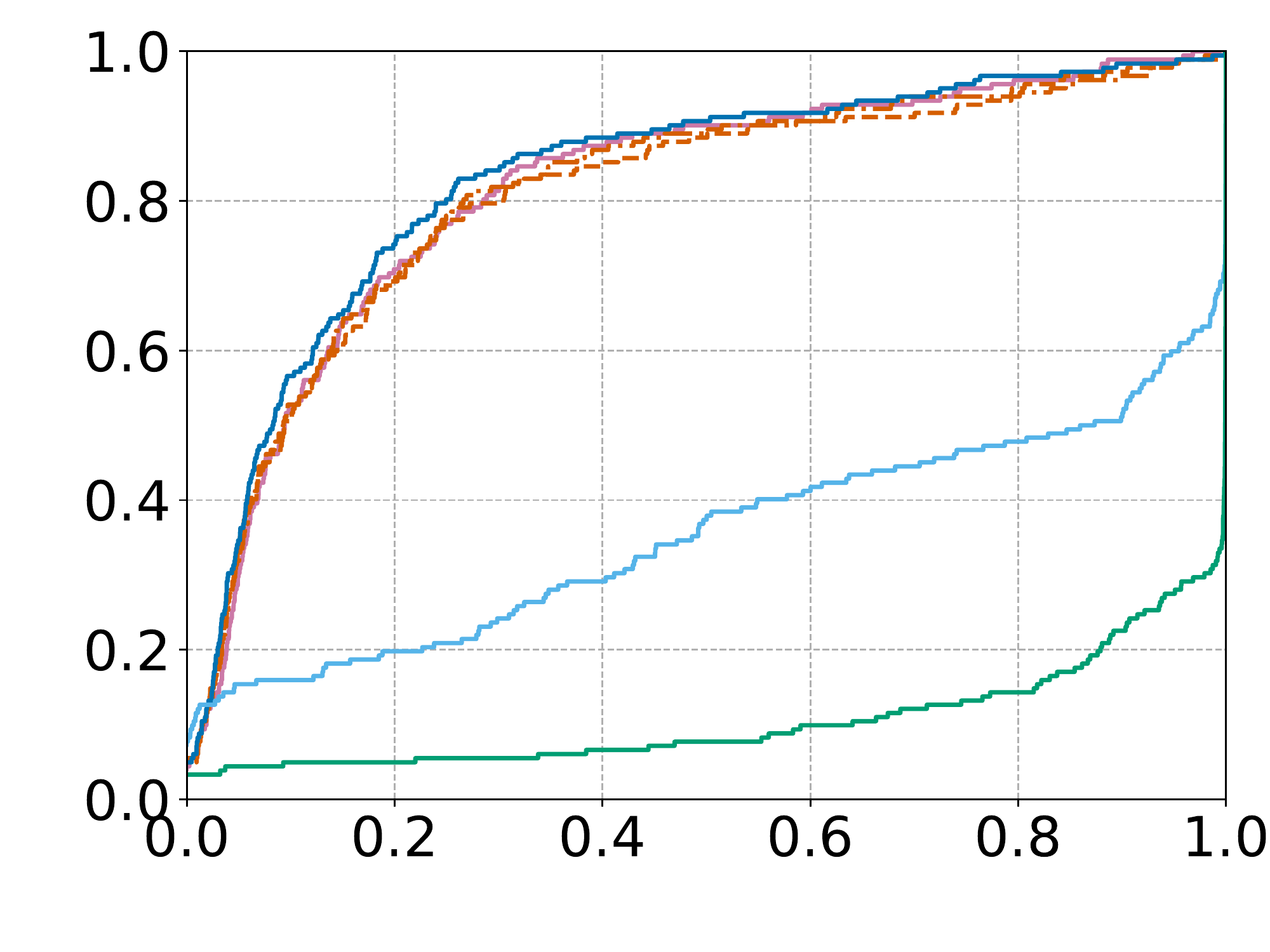}
&
\includegraphics[width=0.29\textwidth,valign=c,trim={0 0 0 0},clip]{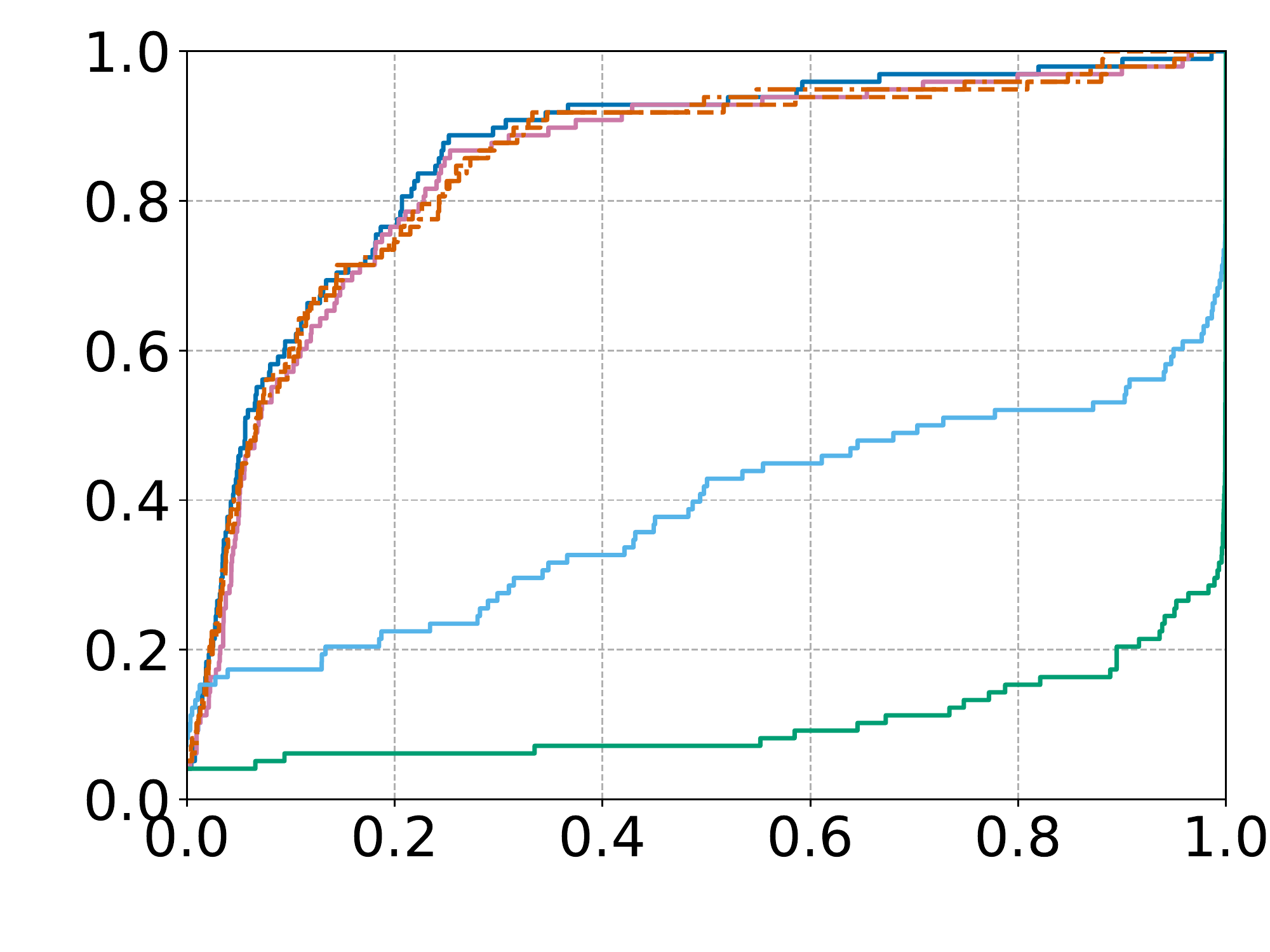}
\\\hline
\begin{tabular}{c}MAJ \\ $p=1$\end{tabular}
& \includegraphics[width=0.29\textwidth,valign=c,trim={0 0 0 0},clip]{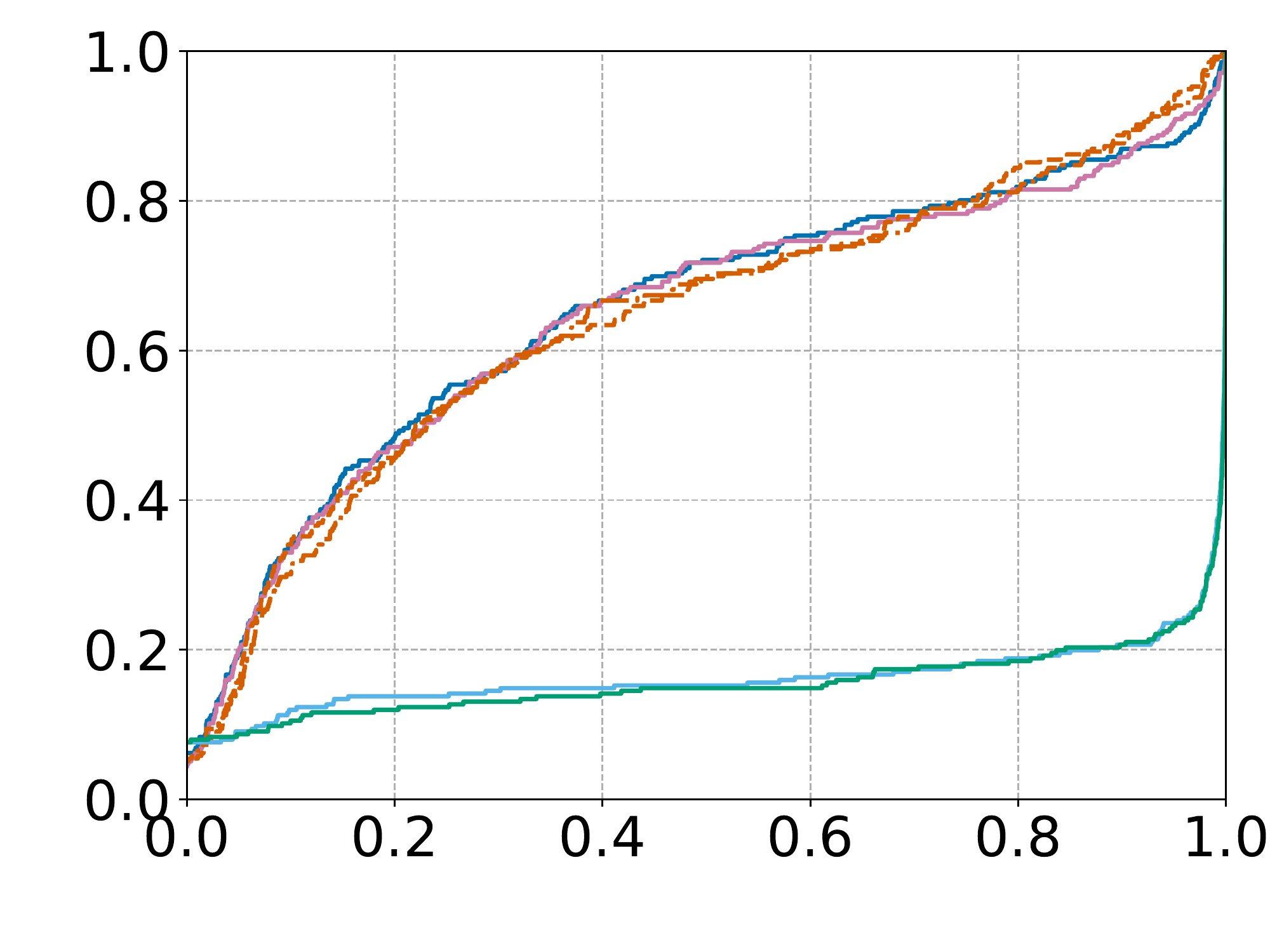}
&
\includegraphics[width=0.29\textwidth,valign=c,trim={0 0 0 0},clip]{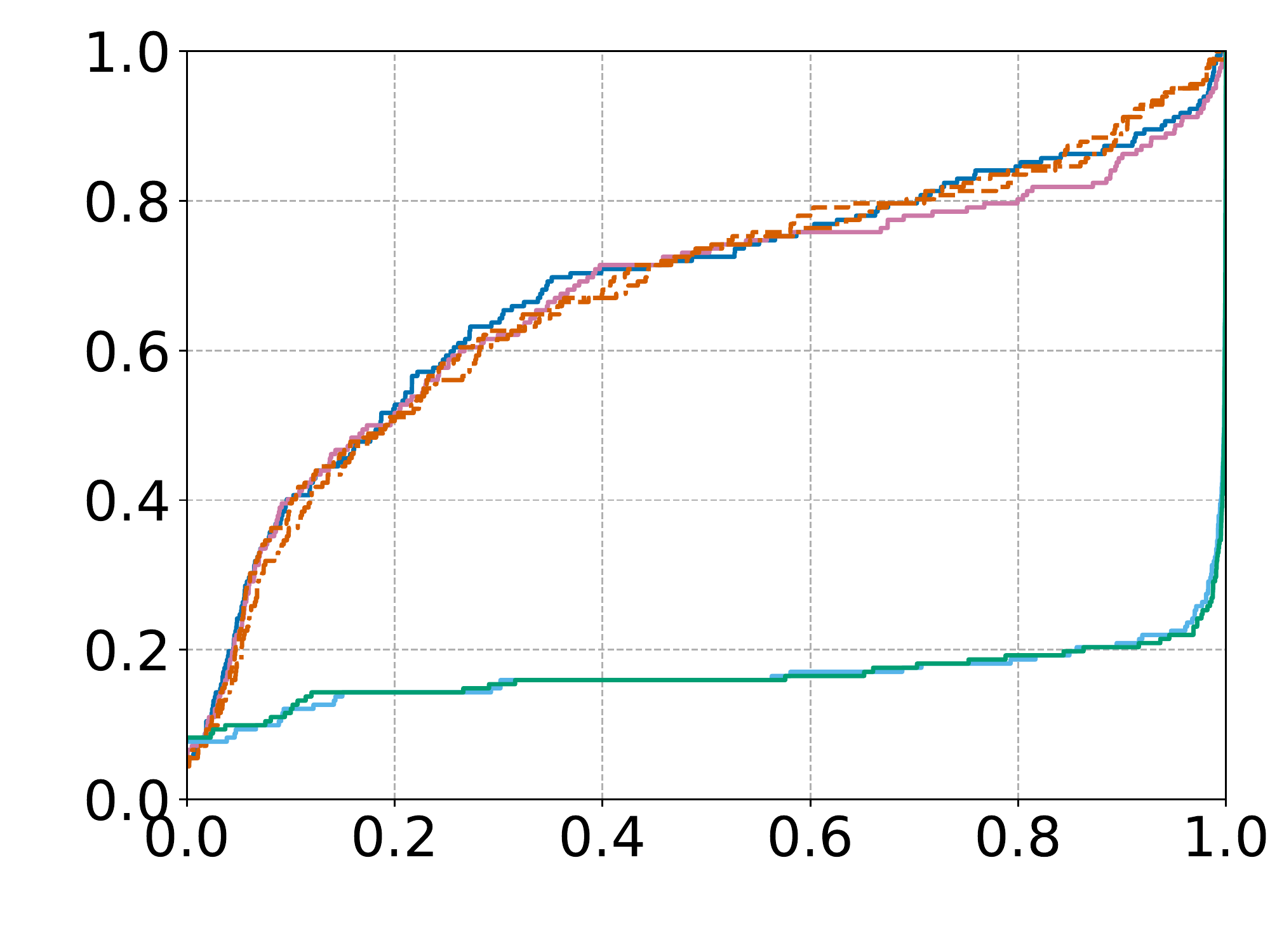}
&
\includegraphics[width=0.29\textwidth,valign=c,trim={0 0 0 0},clip]{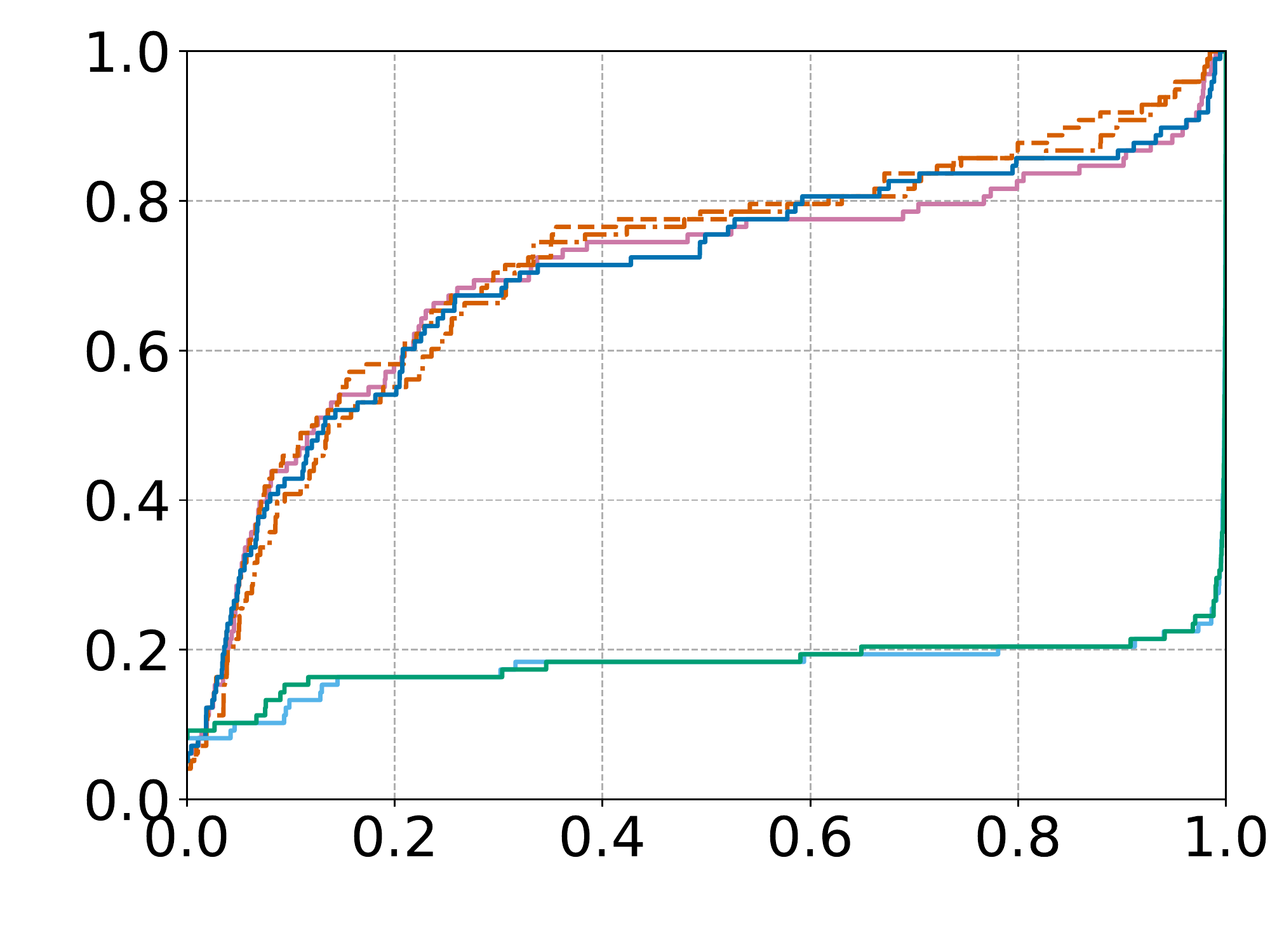}
\\\hline
\begin{tabular}{c}CYC \\ $p=1$\end{tabular}
& \includegraphics[width=0.29\textwidth,valign=c,trim={0 0 0 0},clip]{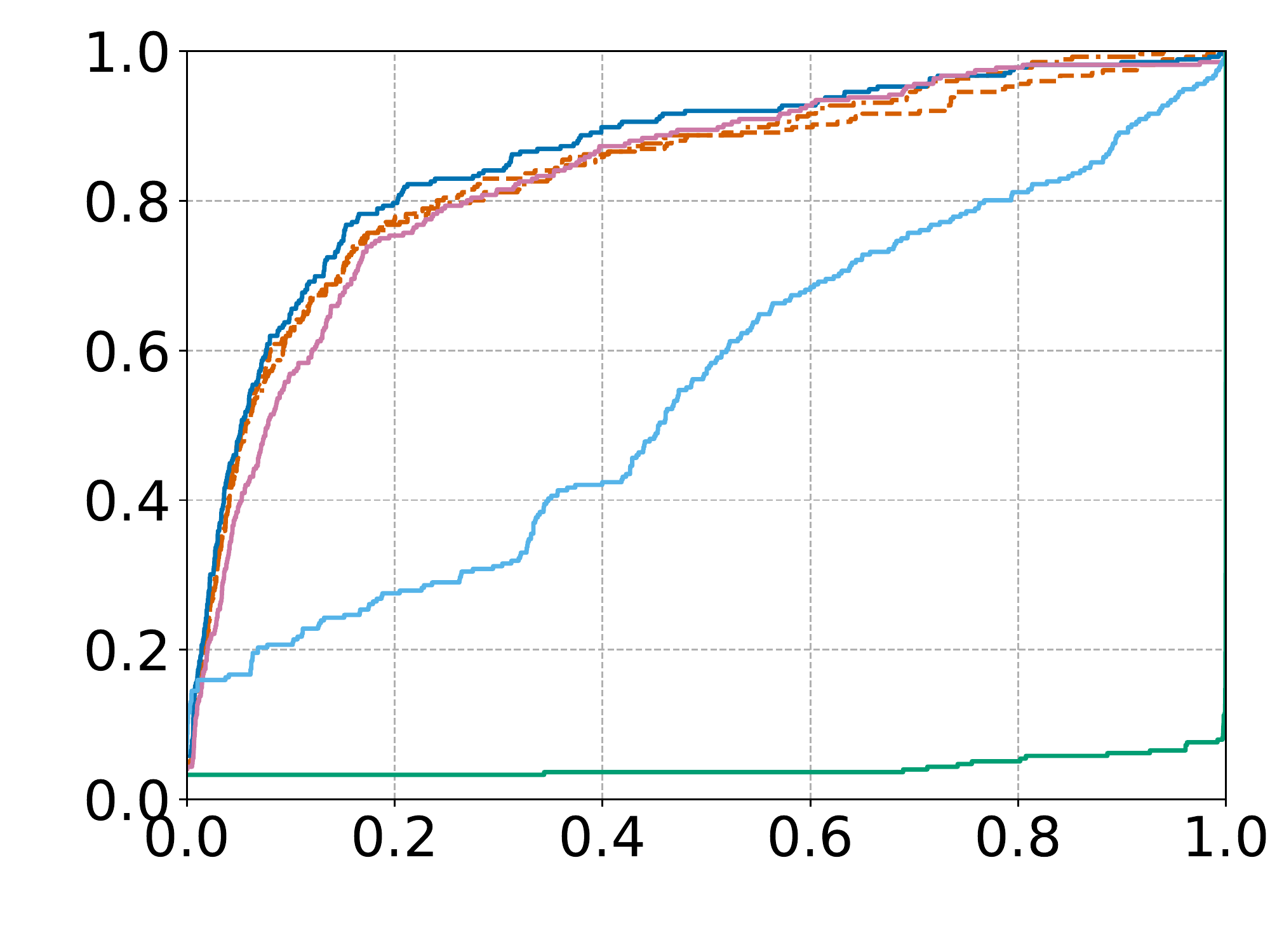}
&
\includegraphics[width=0.29\textwidth,valign=c,trim={0 0 0 0},clip]{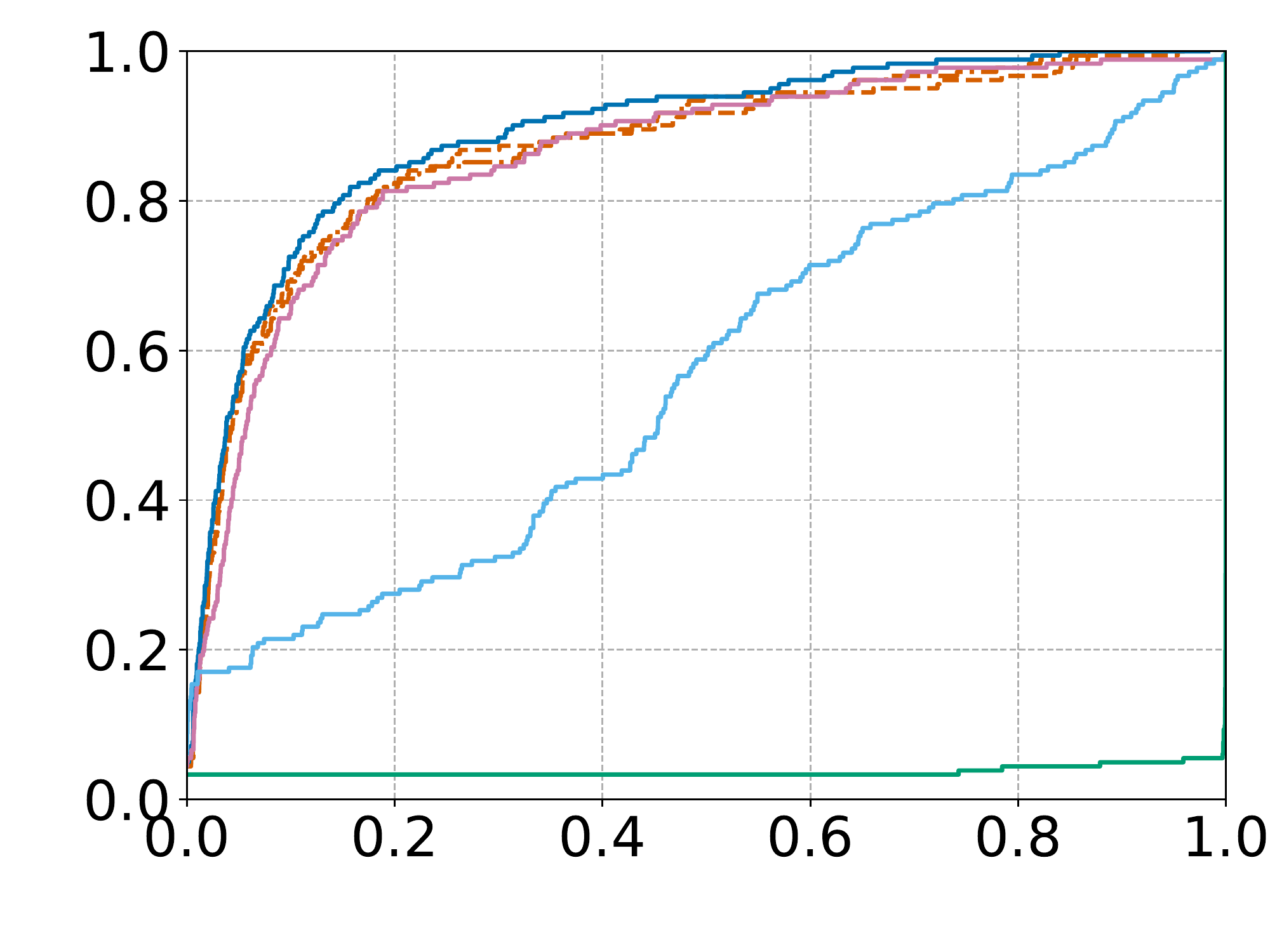}
&
\includegraphics[width=0.29\textwidth,valign=c,trim={0 0 0 0},clip]{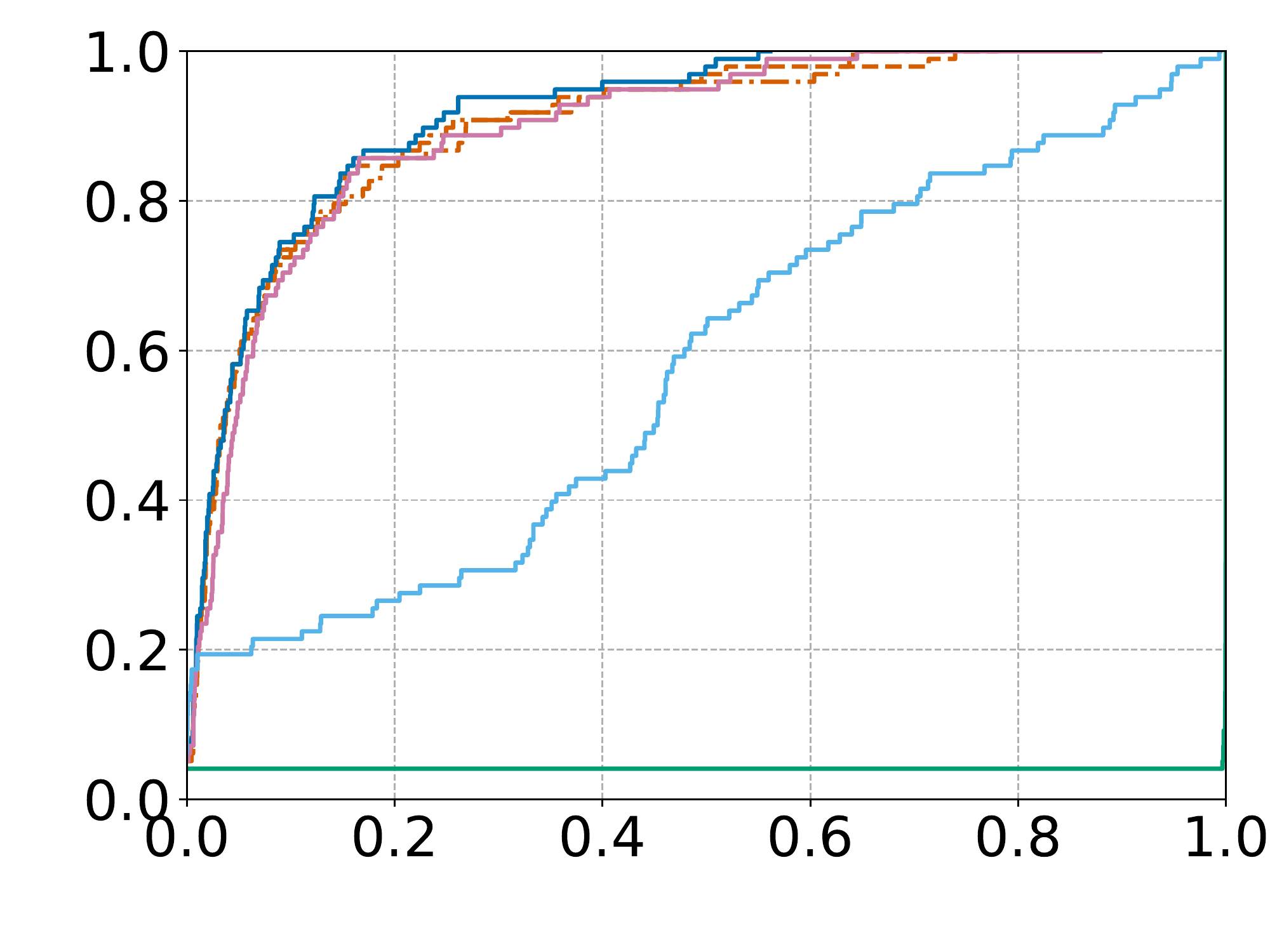}
\\\bottomrule
\multicolumn{4}{c}{
		\begin{subfigure}[t]{\textwidth}
    \includegraphics[width=\textwidth,trim={0 0 0 0},clip]{figs/cdfs/legends/legend_Lambda=8.pdf}
		\end{subfigure} }
\end{tabular}
\caption{Cumulative distribution functions (CDFs) for all evaluated algorithms, against different noise settings and warm start ratios in $\cbr{23.0,46.0,92.0}$. All CB algorithms use $\epsilon$-greedy with $\epsilon=0.0125$. In each of the above plots, the $x$ axis represents scores, while the $y$ axis represents the CDF values.}
\label{fig:cdfs-eps=0.0125-5}
\end{figure}

\begin{figure}[H]
\centering
\begin{tabular}{c | @{}c@{ }c@{ }c@{}} 
\toprule
& \multicolumn{3}{c}{ Ratio }
\\
Noise & 2.875 & 5.75 & 11.5
\\\midrule
Noiseless & \includegraphics[width=0.29\textwidth,valign=c,trim={0 0 0 0},clip]{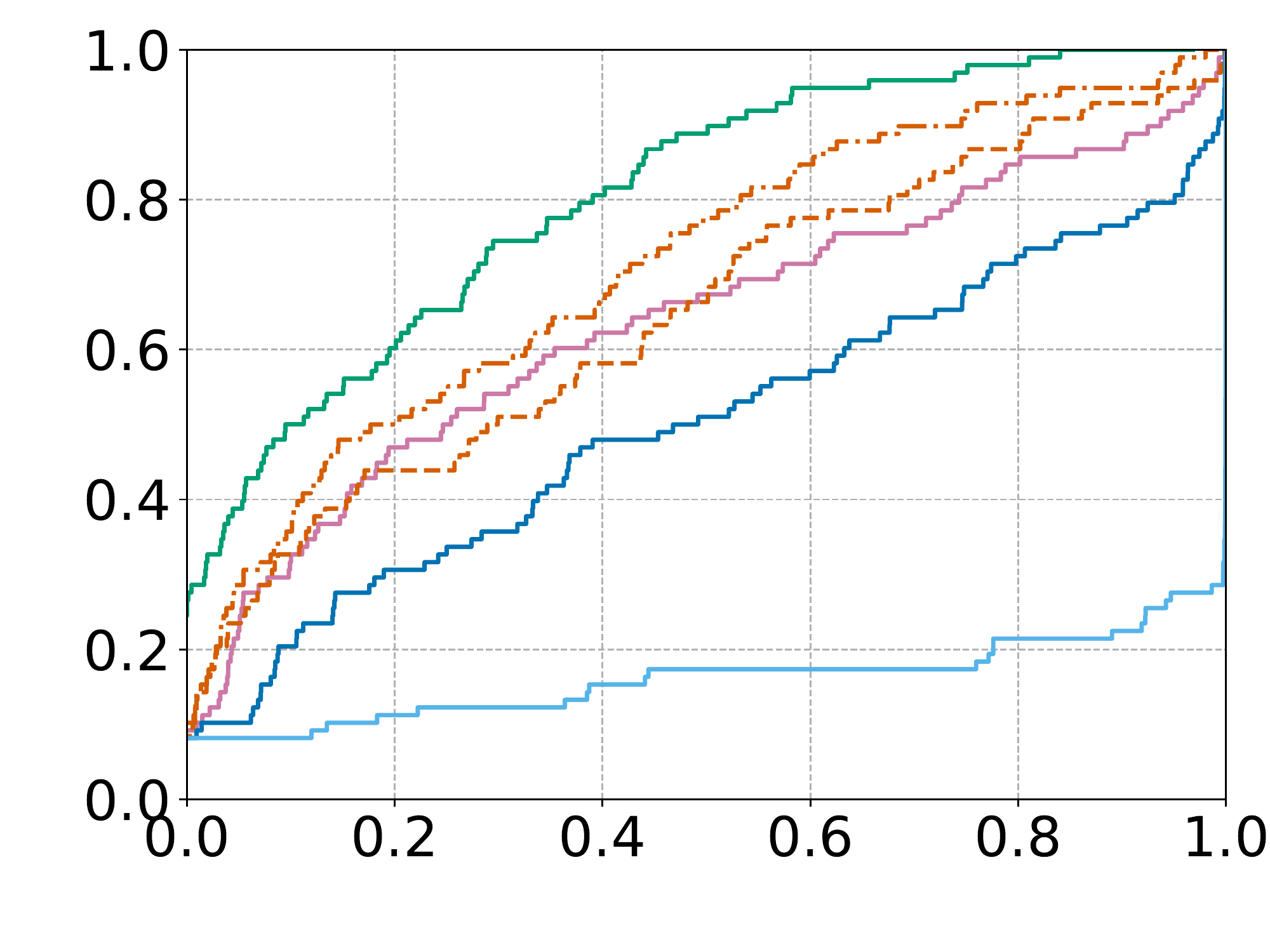}
&
\includegraphics[width=0.29\textwidth,valign=c,trim={0 0 0 0},clip]{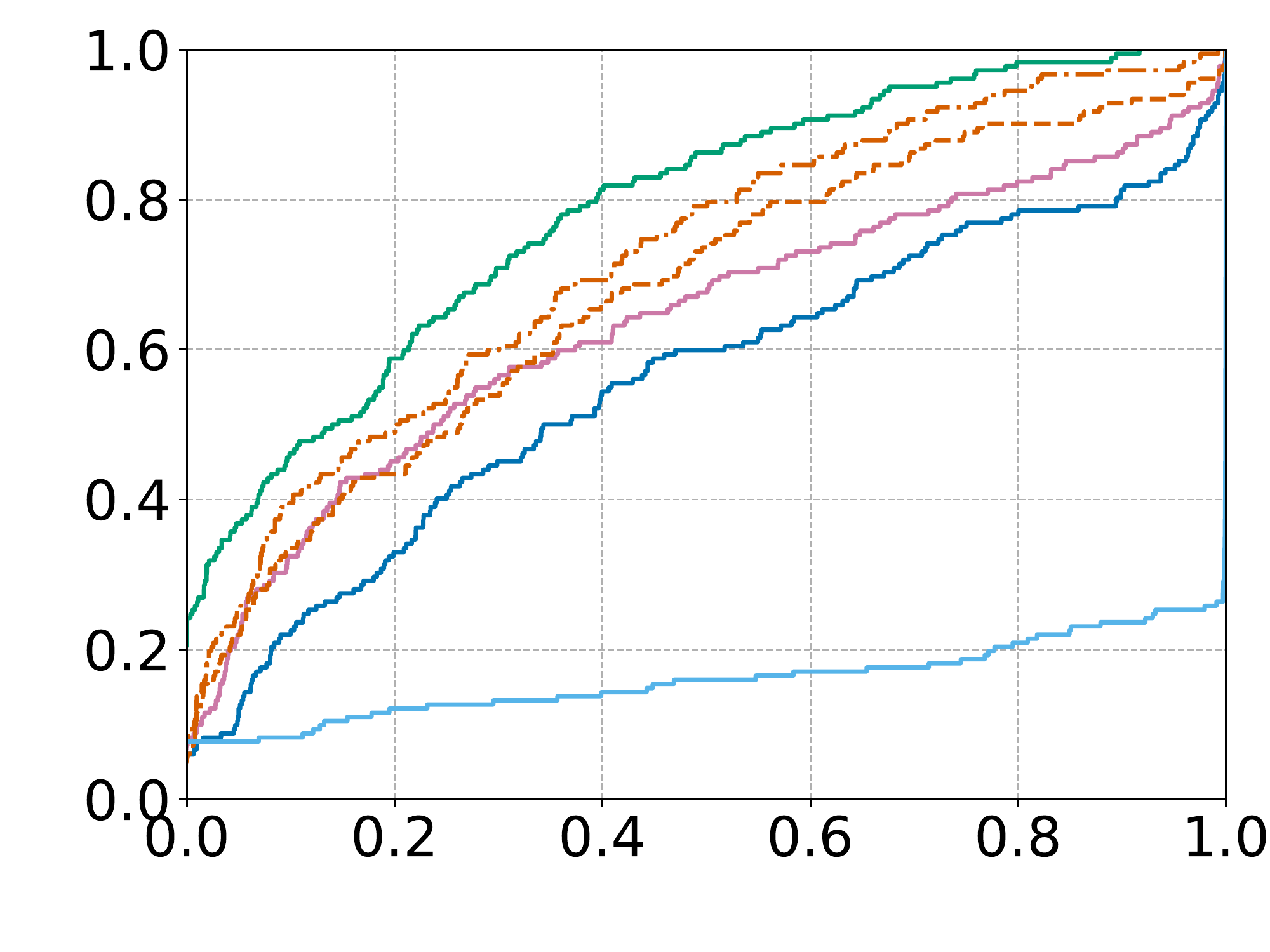}
&
\includegraphics[width=0.29\textwidth,valign=c,trim={0 0 0 0},clip]{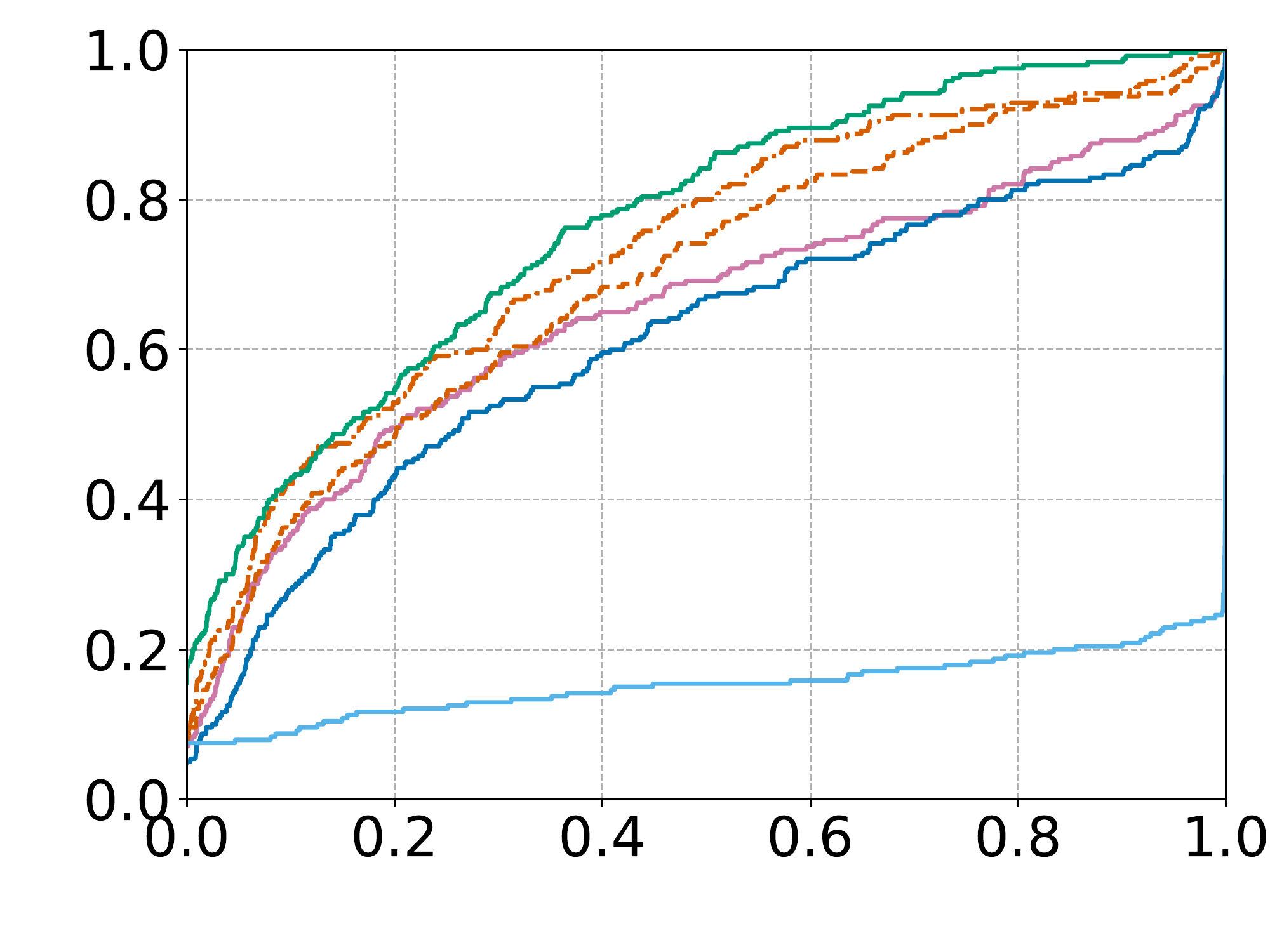}
\\\hline
\begin{tabular}{c}UAR \\ $p=0.25$\end{tabular}
& \includegraphics[width=0.29\textwidth,valign=c,trim={0 0 0 0},clip]{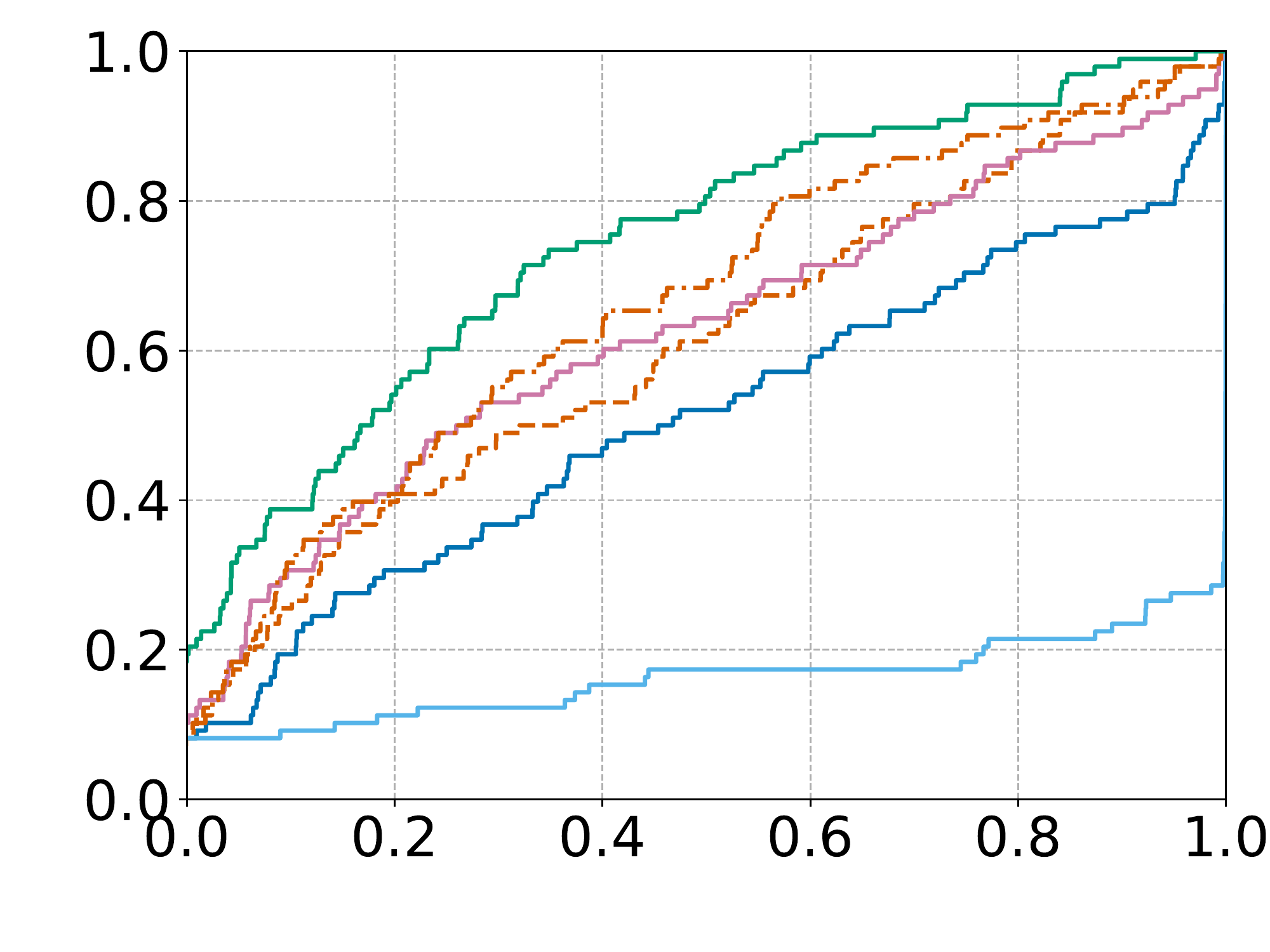}
&
\includegraphics[width=0.29\textwidth,valign=c,trim={0 0 0 0},clip]{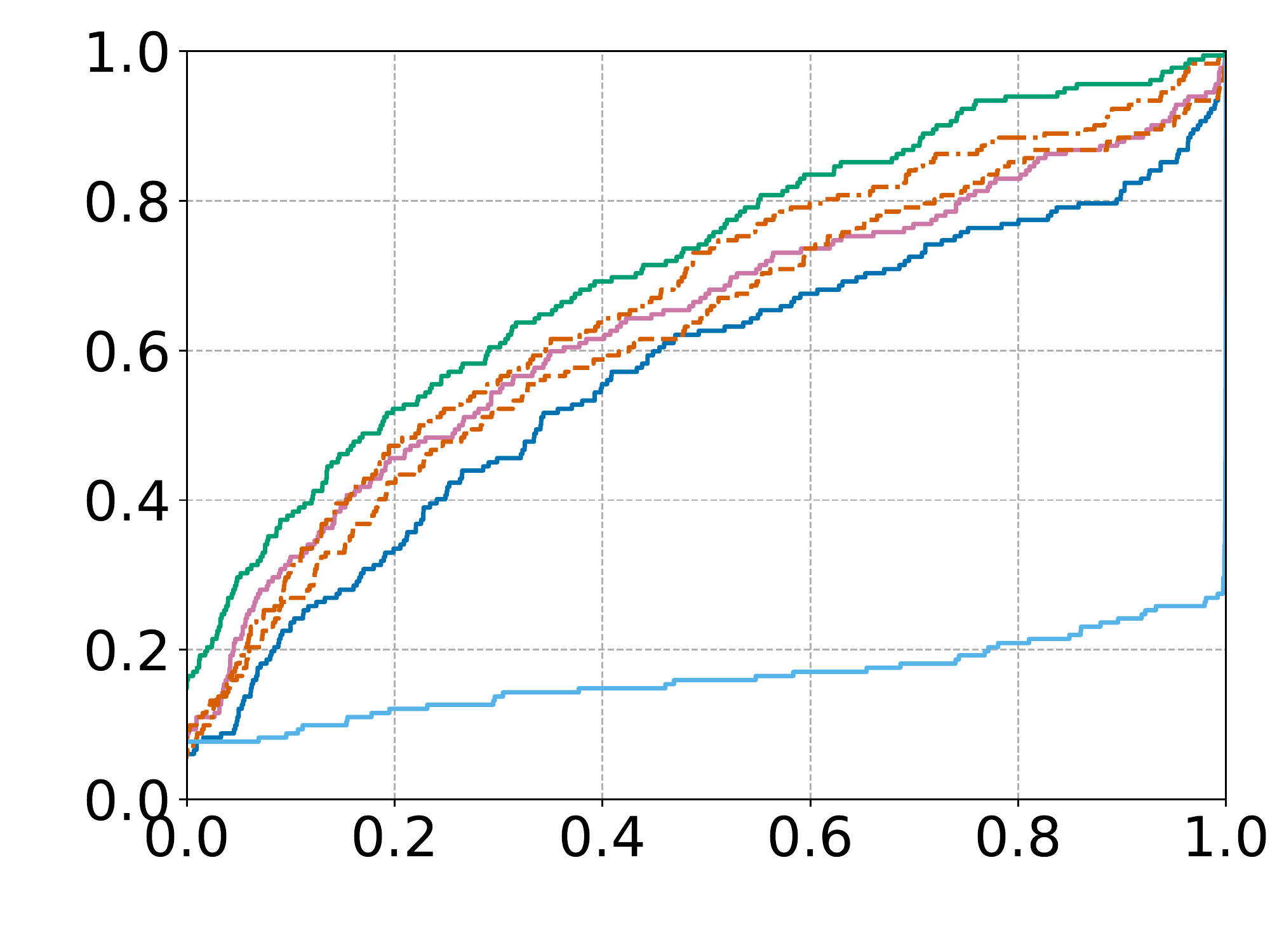}
&
\includegraphics[width=0.29\textwidth,valign=c,trim={0 0 0 0},clip]{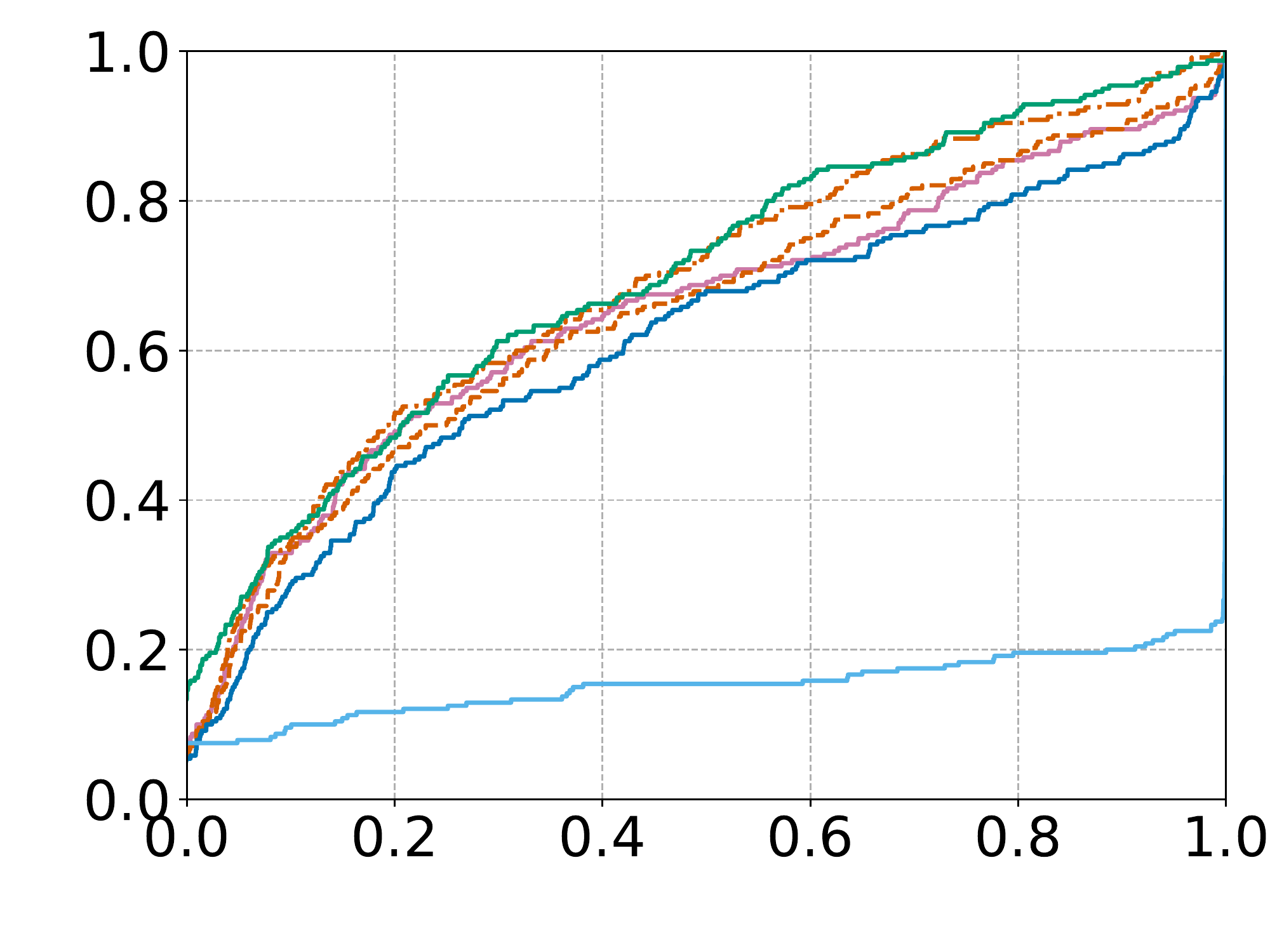}
\\\hline
\begin{tabular}{c}MAJ \\ $p=0.25$\end{tabular}
& \includegraphics[width=0.29\textwidth,valign=c,trim={0 0 0 0},clip]{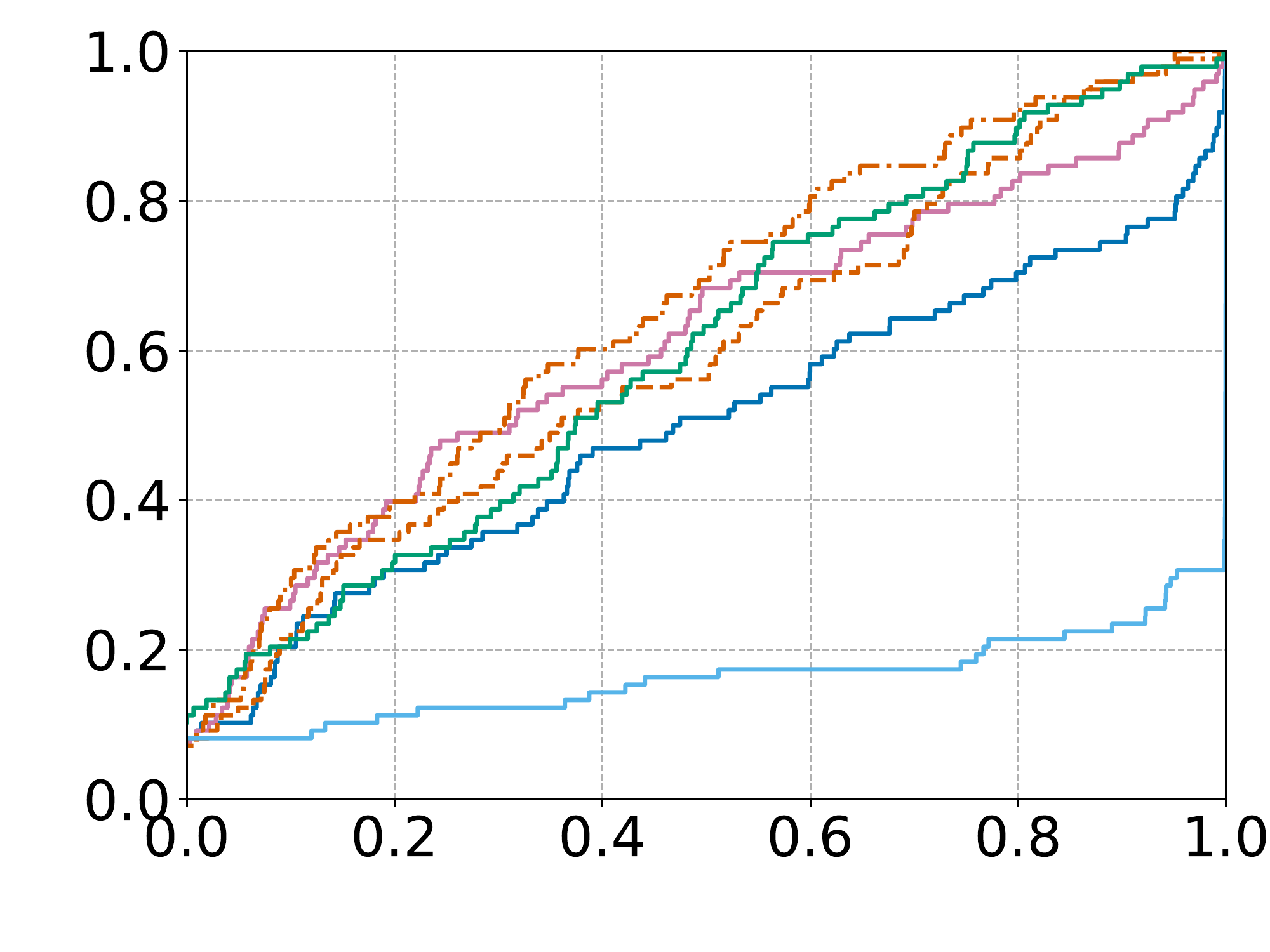}
&
\includegraphics[width=0.29\textwidth,valign=c,trim={0 0 0 0},clip]{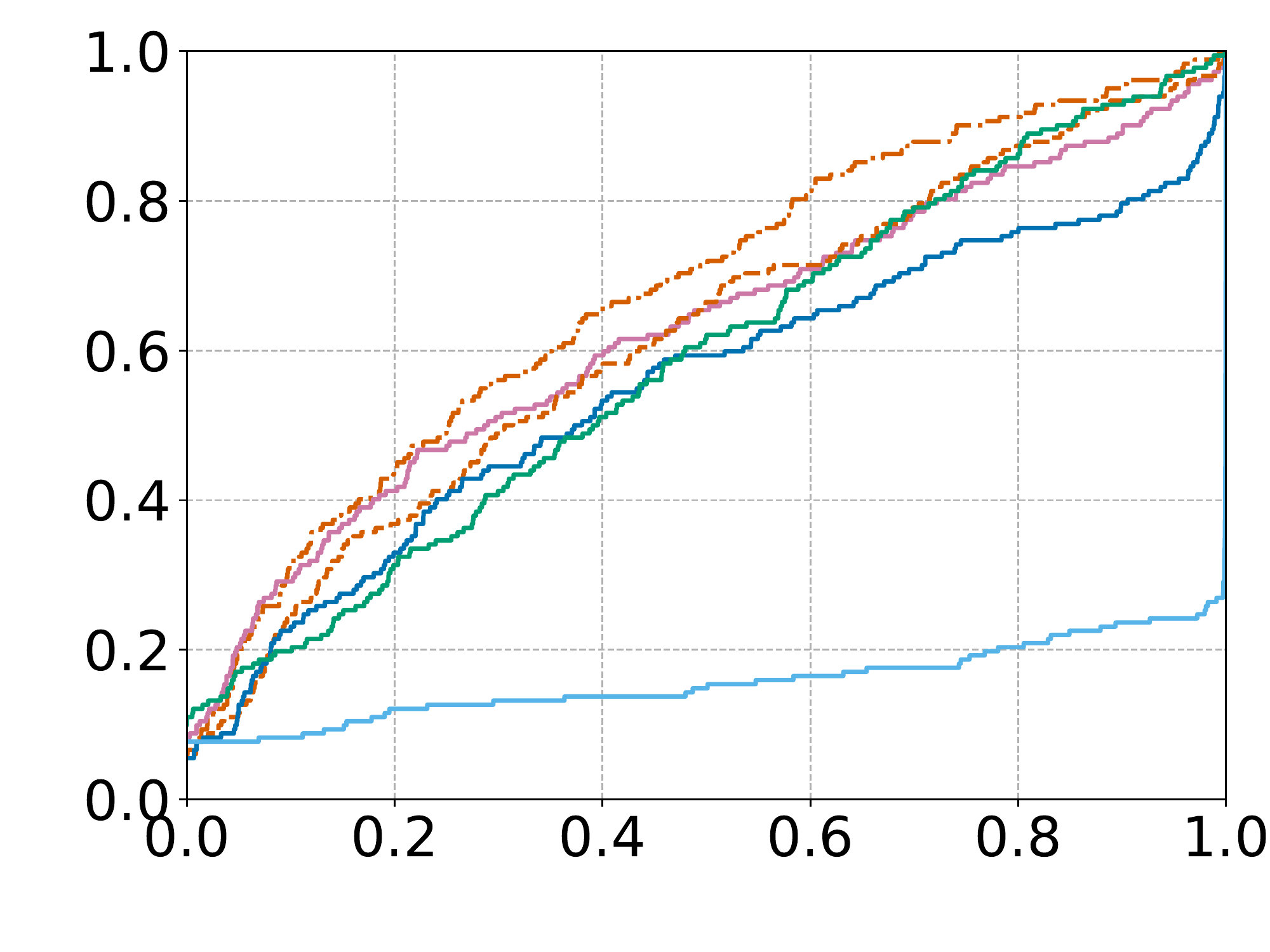}
&
\includegraphics[width=0.29\textwidth,valign=c,trim={0 0 0 0},clip]{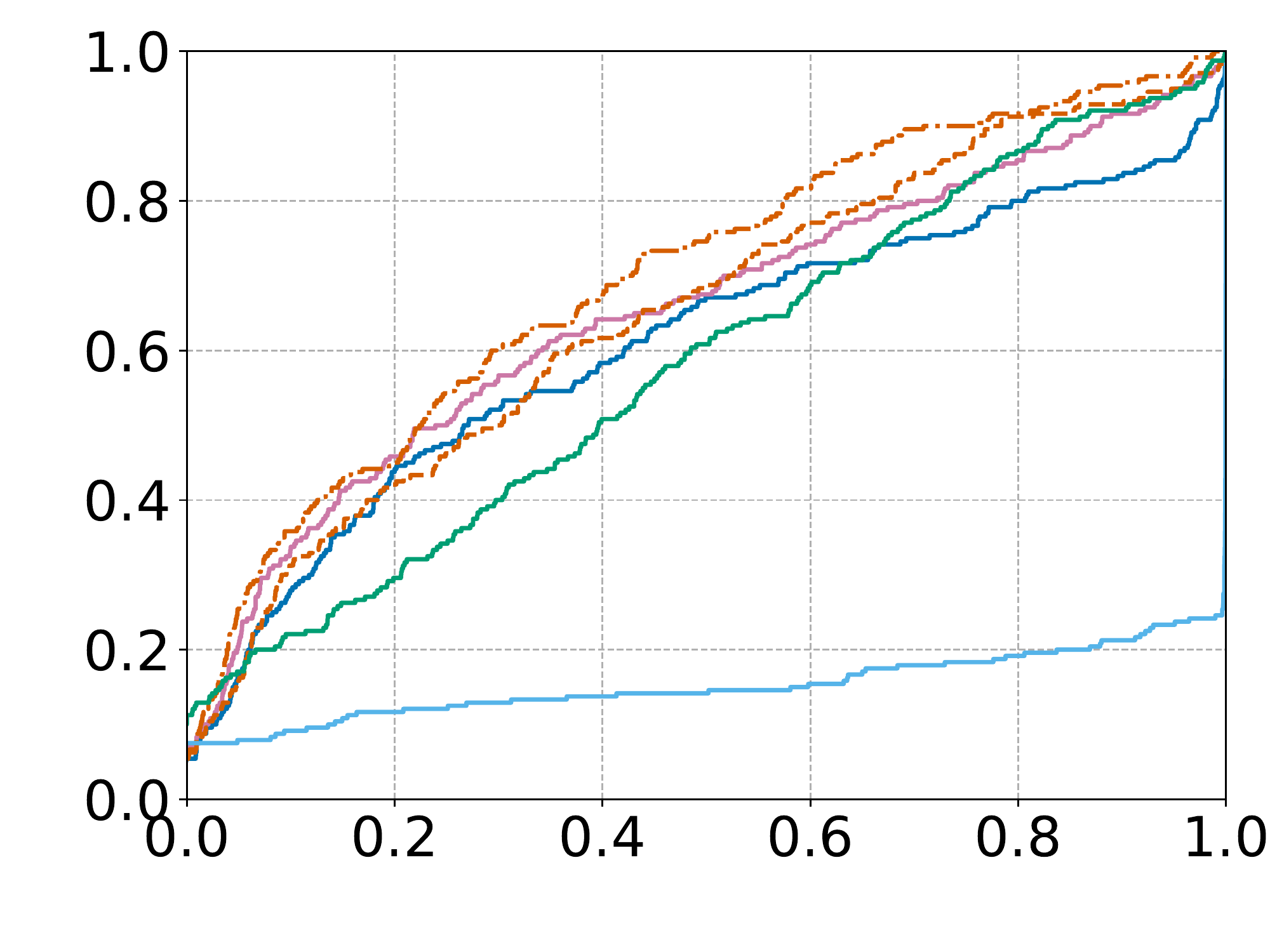}
\\\hline
\begin{tabular}{c}CYC \\ $p=0.25$\end{tabular}
& \includegraphics[width=0.29\textwidth,valign=c,trim={0 0 0 0},clip]{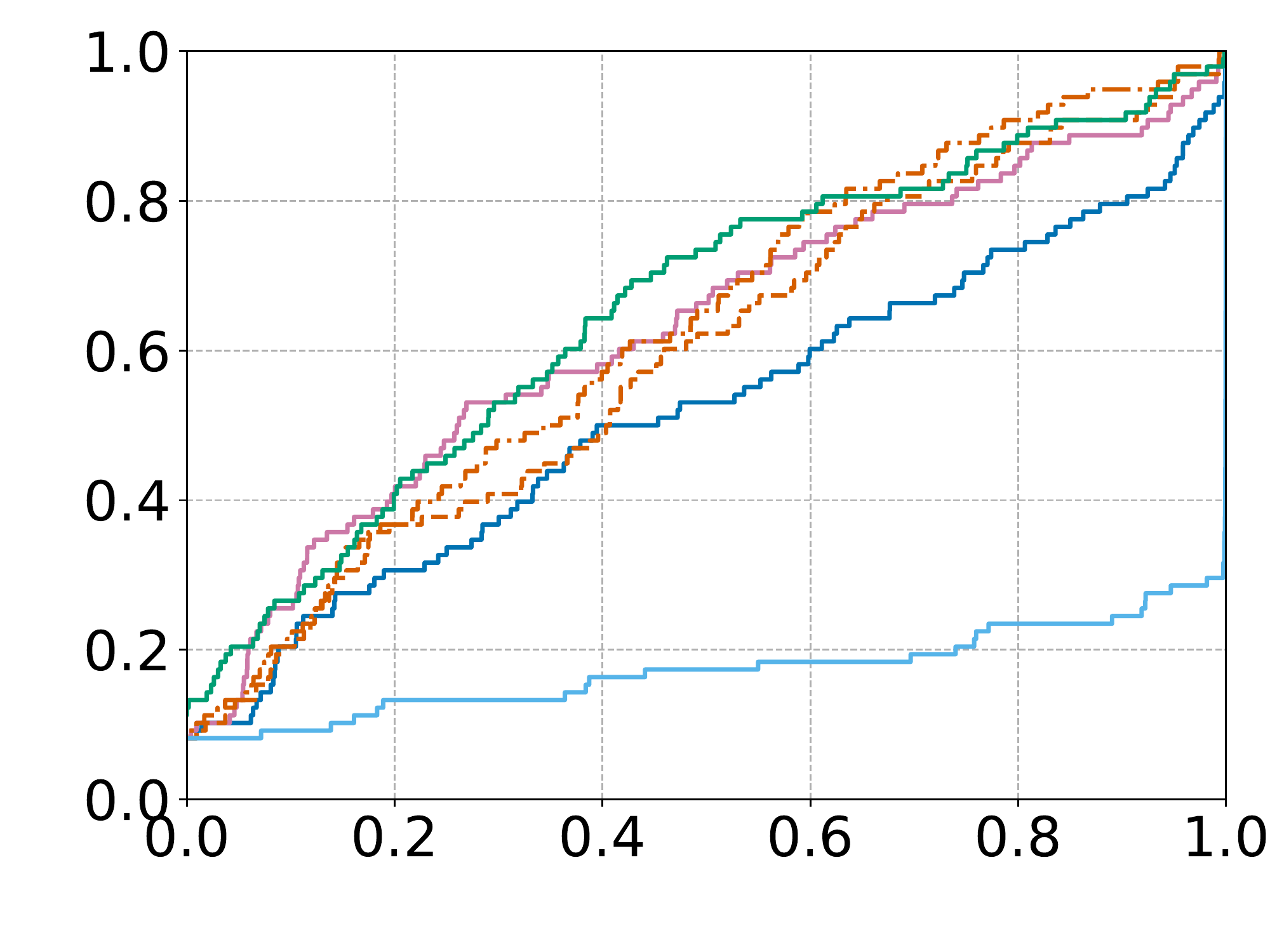}
&
\includegraphics[width=0.29\textwidth,valign=c,trim={0 0 0 0},clip]{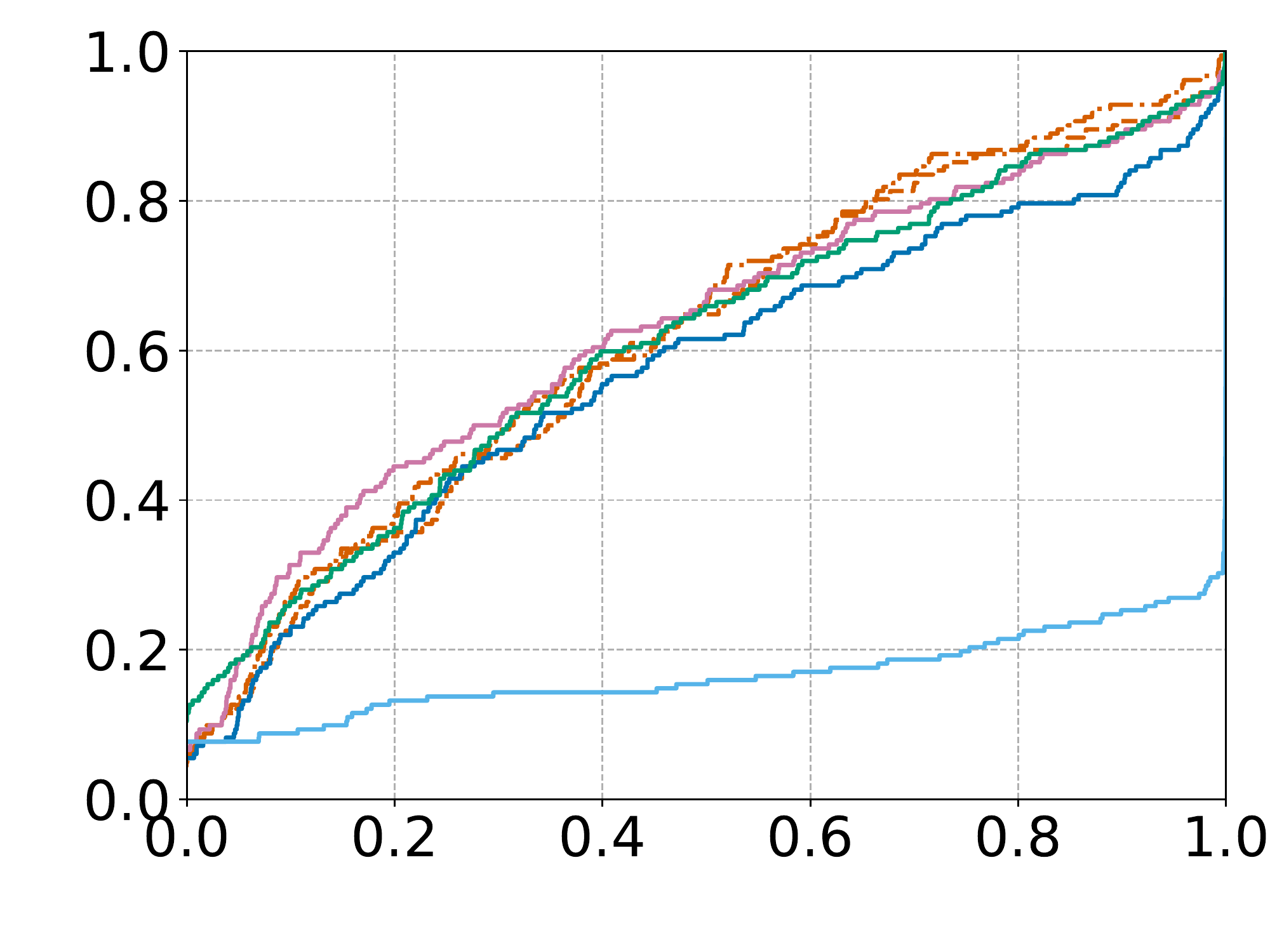}
&
\includegraphics[width=0.29\textwidth,valign=c,trim={0 0 0 0},clip]{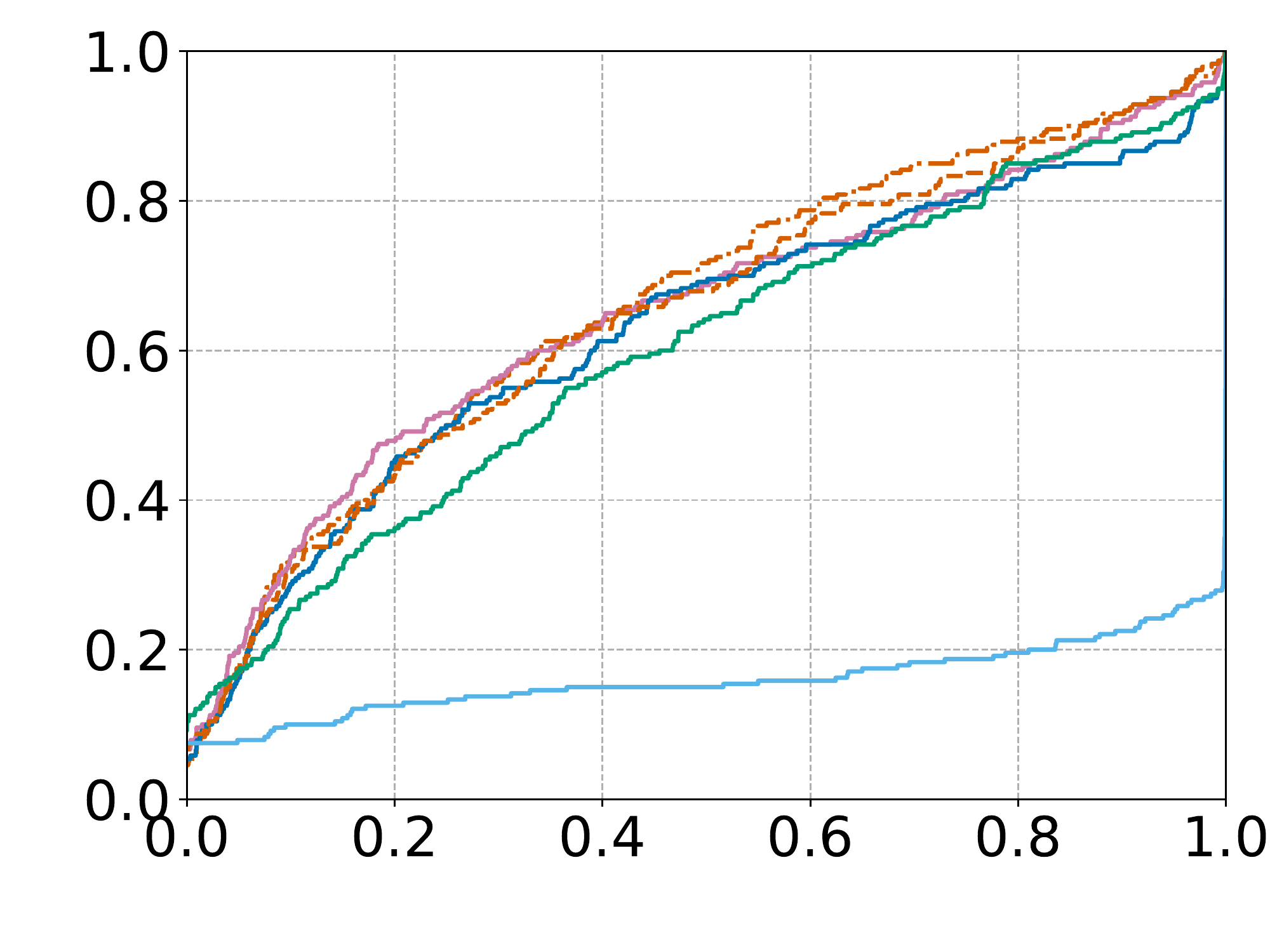}
\\\hline
\begin{tabular}{c}UAR \\ $p=0.5$\end{tabular}
& \includegraphics[width=0.29\textwidth,valign=c,trim={0 0 0 0},clip]{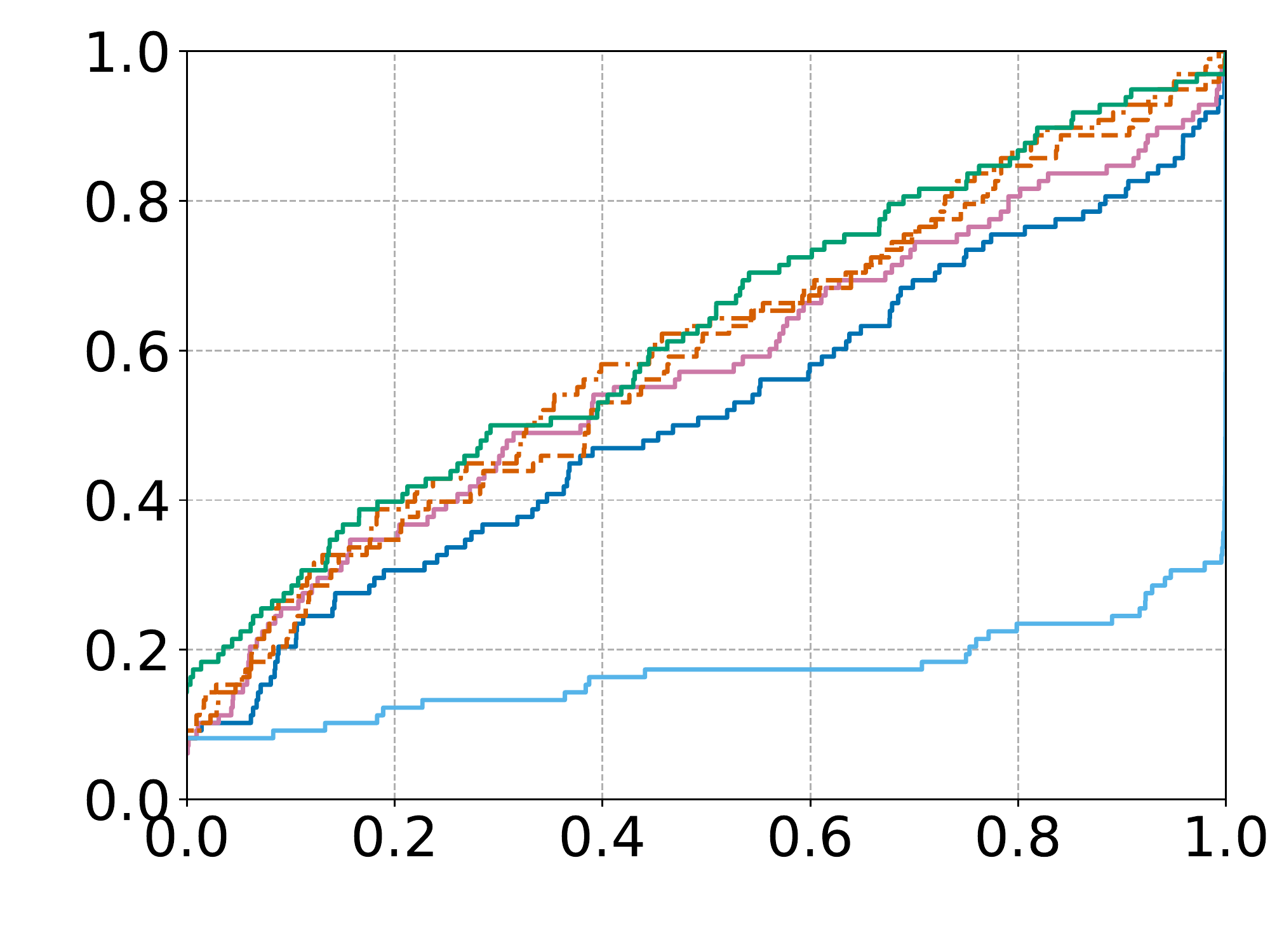}
&
\includegraphics[width=0.29\textwidth,valign=c,trim={0 0 0 0},clip]{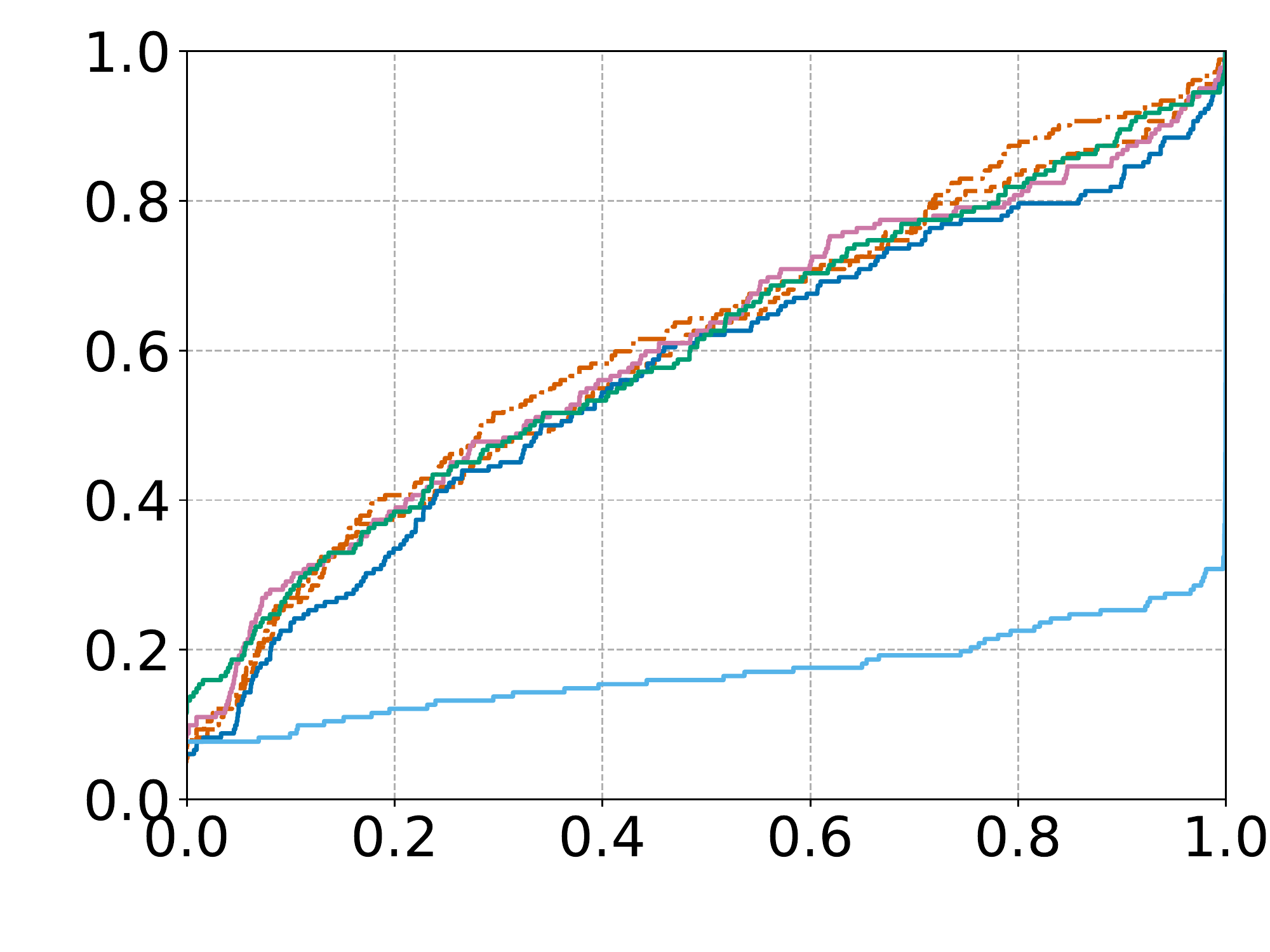}
&
\includegraphics[width=0.29\textwidth,valign=c,trim={0 0 0 0},clip]{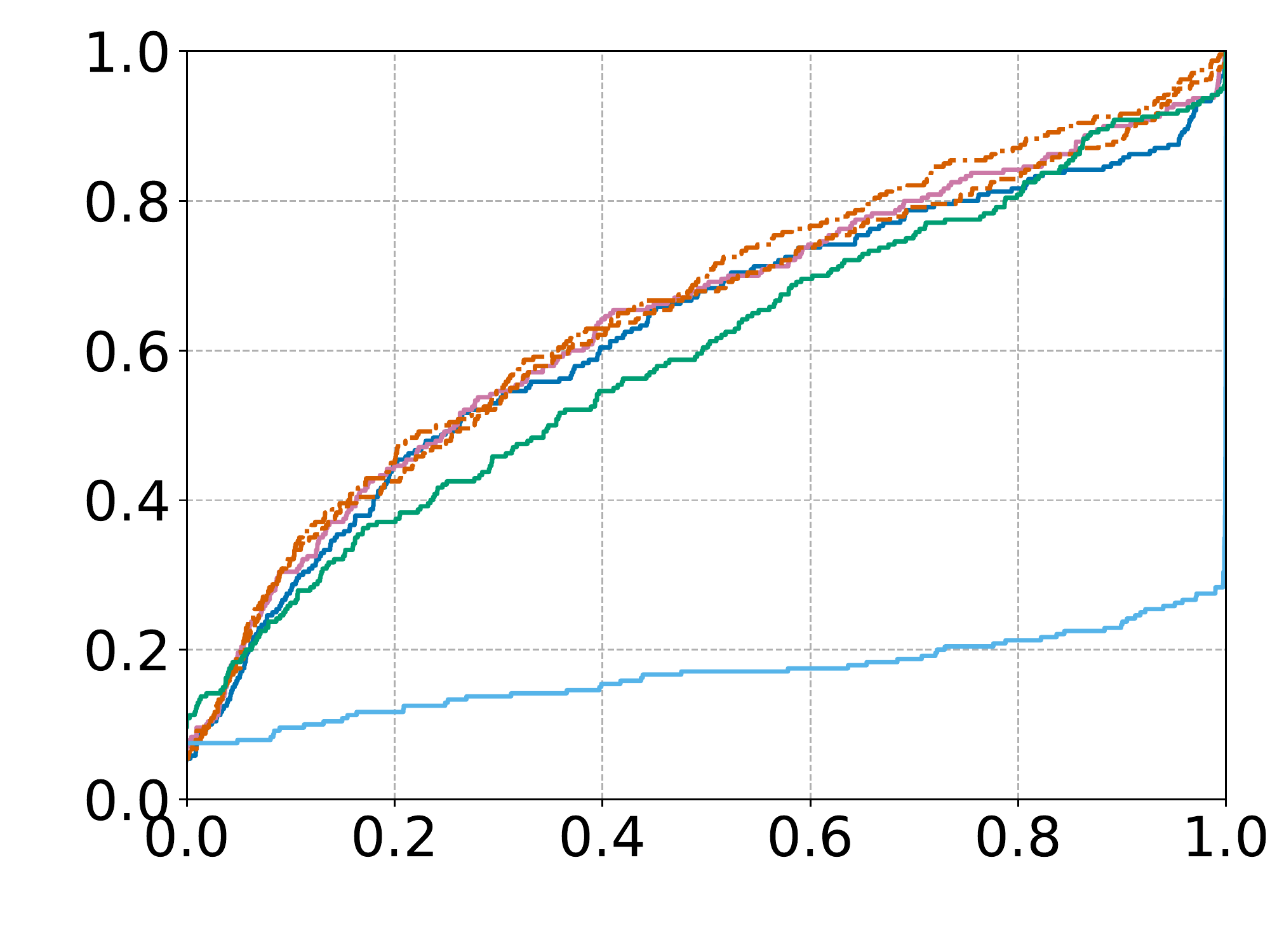}
\\\bottomrule
\multicolumn{4}{c}{
		\begin{subfigure}[t]{\textwidth}
    \includegraphics[width=\textwidth,trim={0 0 0 0},clip]{figs/cdfs/legends/legend_Lambda=8.pdf}
		\end{subfigure} }
\end{tabular}
\caption{Cumulative distribution functions (CDFs) for all evaluated algorithms, against different noise settings and warm start ratios in $\cbr{2.875,5.75,11.5}$. All CB algorithms use $\epsilon$-greedy with $\epsilon=0.00625$. In each of the above plots, the $x$ axis represents scores, while the $y$ axis represents the CDF values.}
\label{fig:cdfs-eps=0.00625-1}
\end{figure}

\begin{figure}[H]
\centering
\begin{tabular}{c | @{}c@{ }c@{ }c@{}} 
\toprule
& \multicolumn{3}{c}{ Ratio }
\\
Noise & 23.0 & 46.0 & 92.0
\\\midrule
Noiseless & \includegraphics[width=0.29\textwidth,valign=c,trim={0 0 0 0},clip]{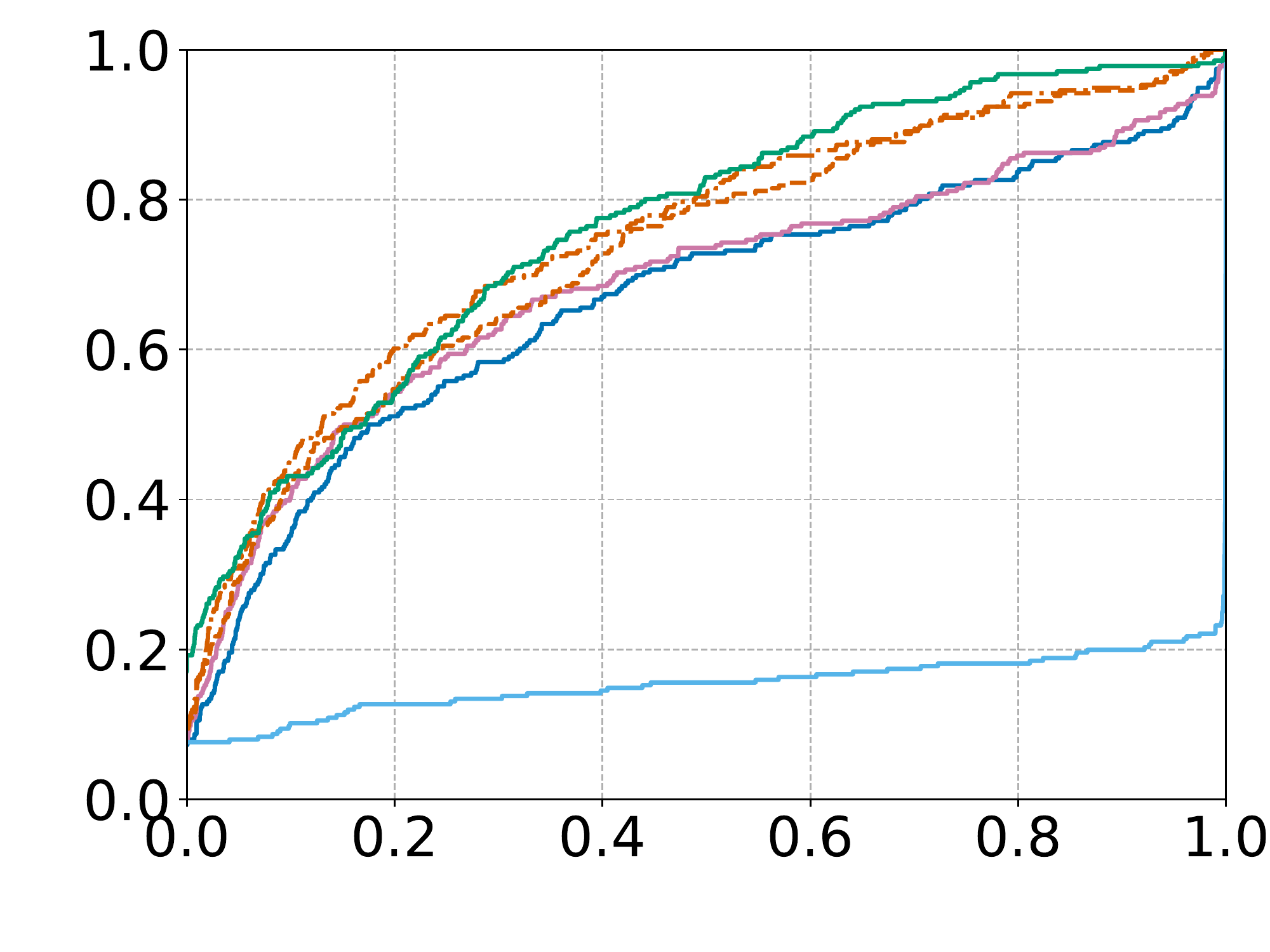}
&
\includegraphics[width=0.29\textwidth,valign=c,trim={0 0 0 0},clip]{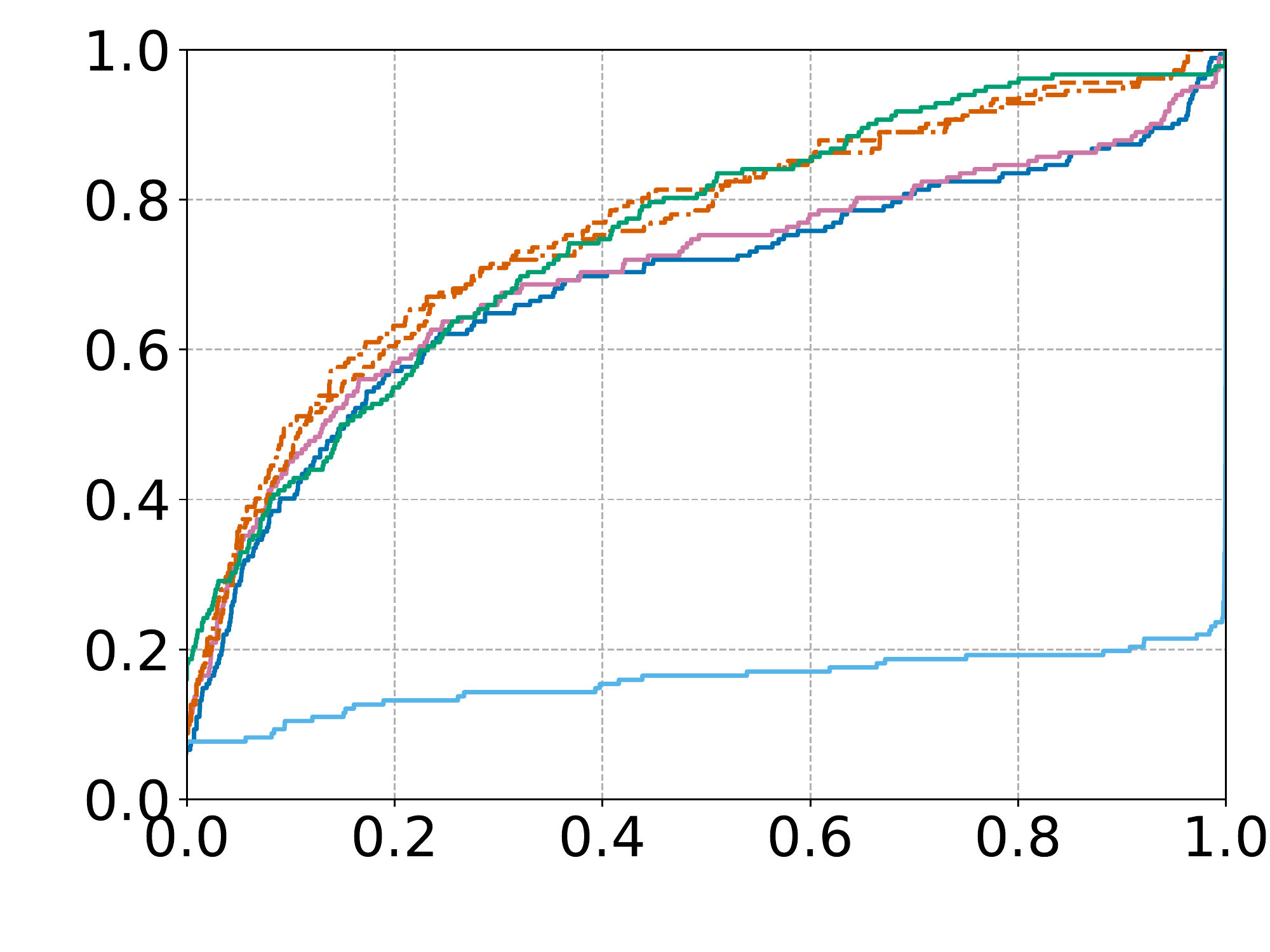}
&
\includegraphics[width=0.29\textwidth,valign=c,trim={0 0 0 0},clip]{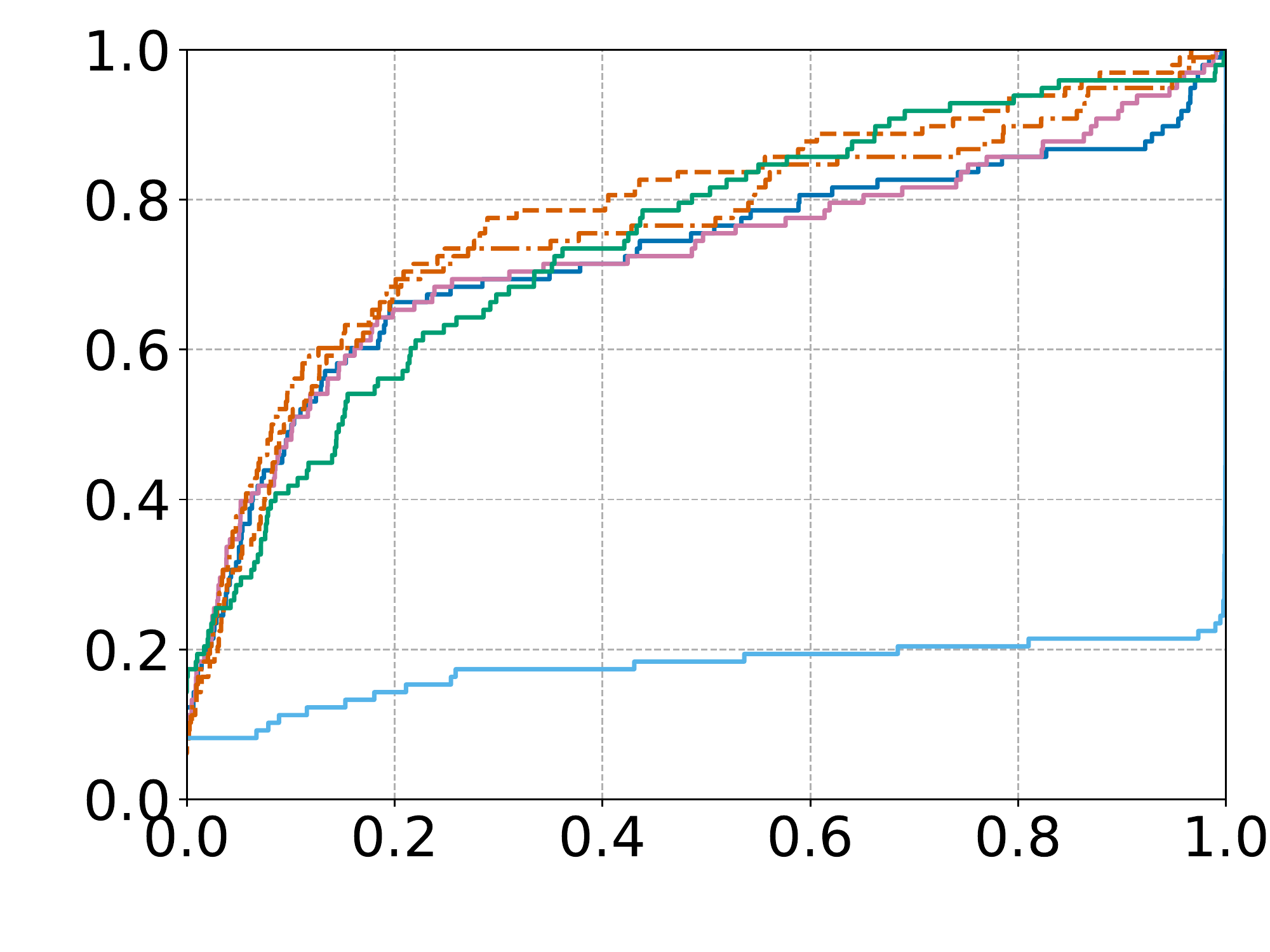}
\\\hline
\begin{tabular}{c}UAR \\ $p=0.25$\end{tabular}
& \includegraphics[width=0.29\textwidth,valign=c,trim={0 0 0 0},clip]{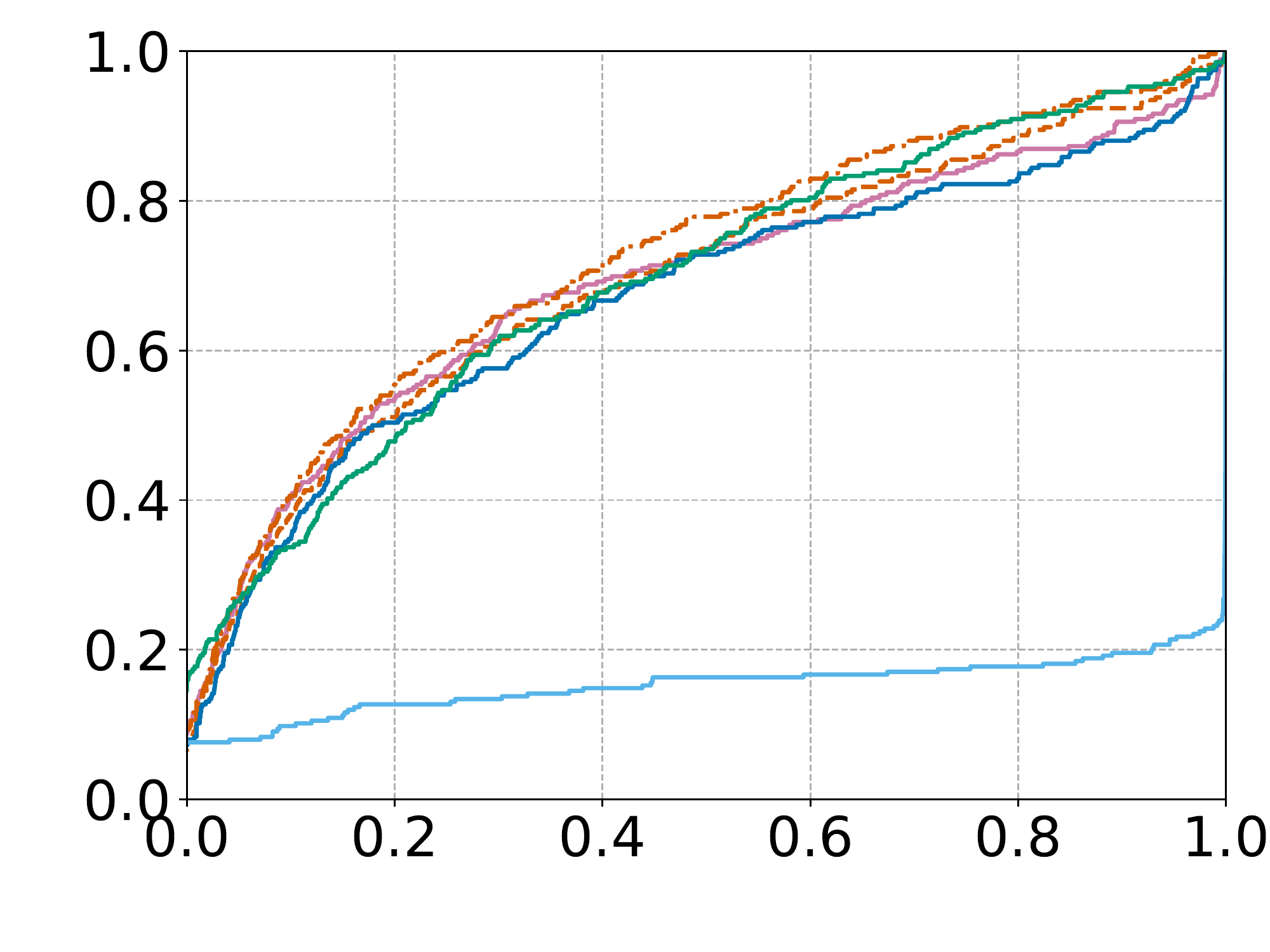}
&
\includegraphics[width=0.29\textwidth,valign=c,trim={0 0 0 0},clip]{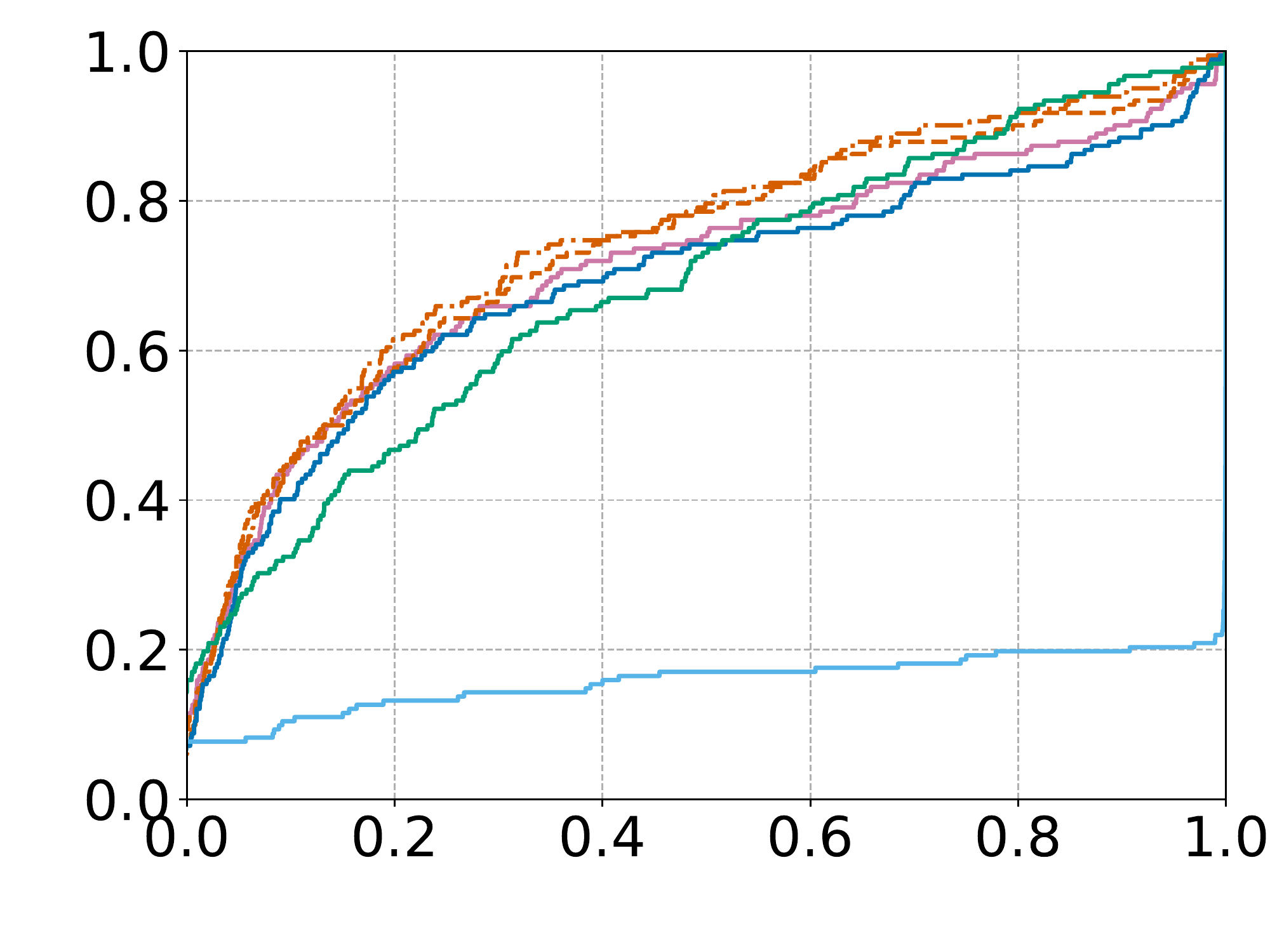}
&
\includegraphics[width=0.29\textwidth,valign=c,trim={0 0 0 0},clip]{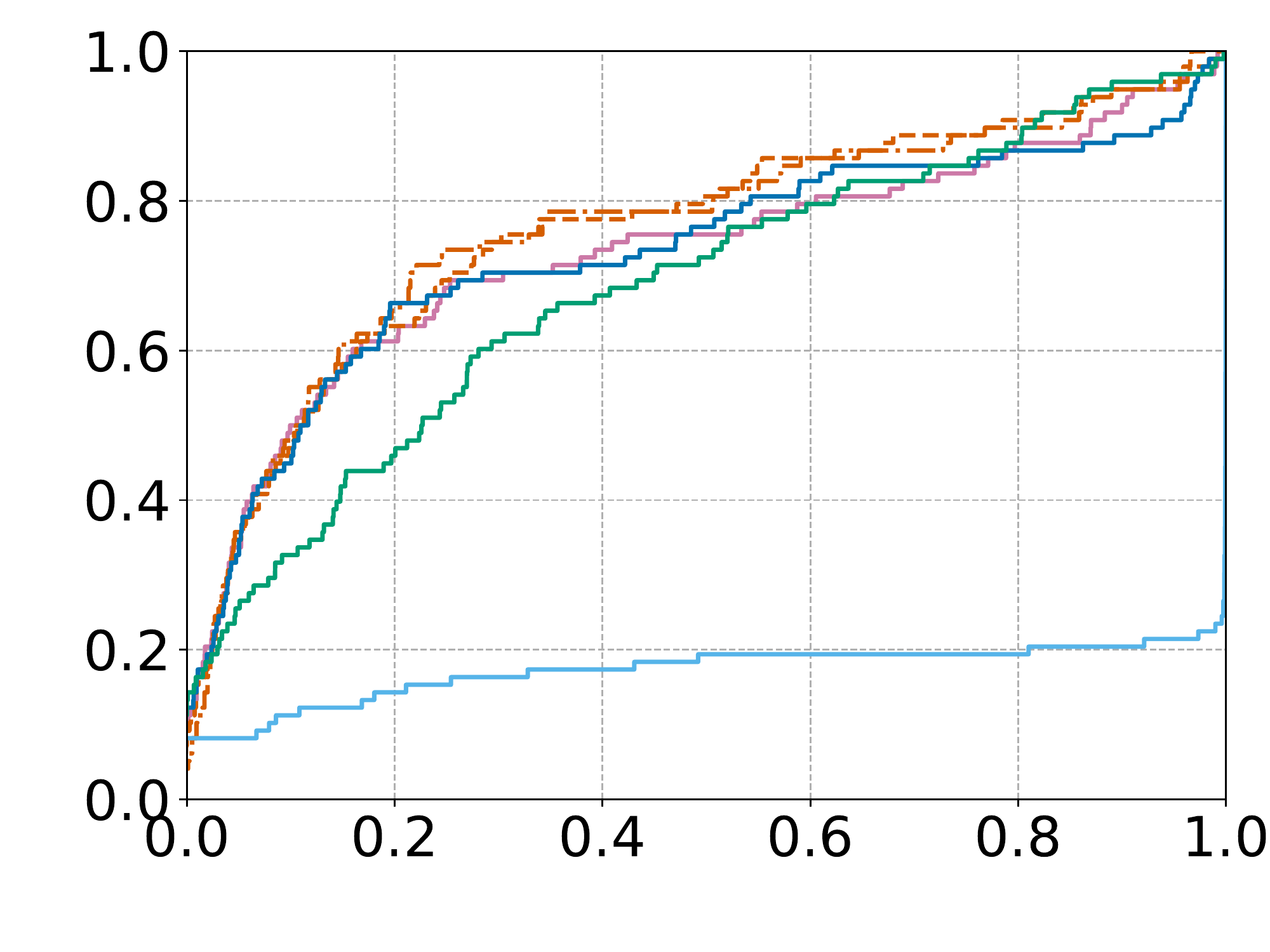}
\\\hline
\begin{tabular}{c}MAJ \\ $p=0.25$\end{tabular}
& \includegraphics[width=0.29\textwidth,valign=c,trim={0 0 0 0},clip]{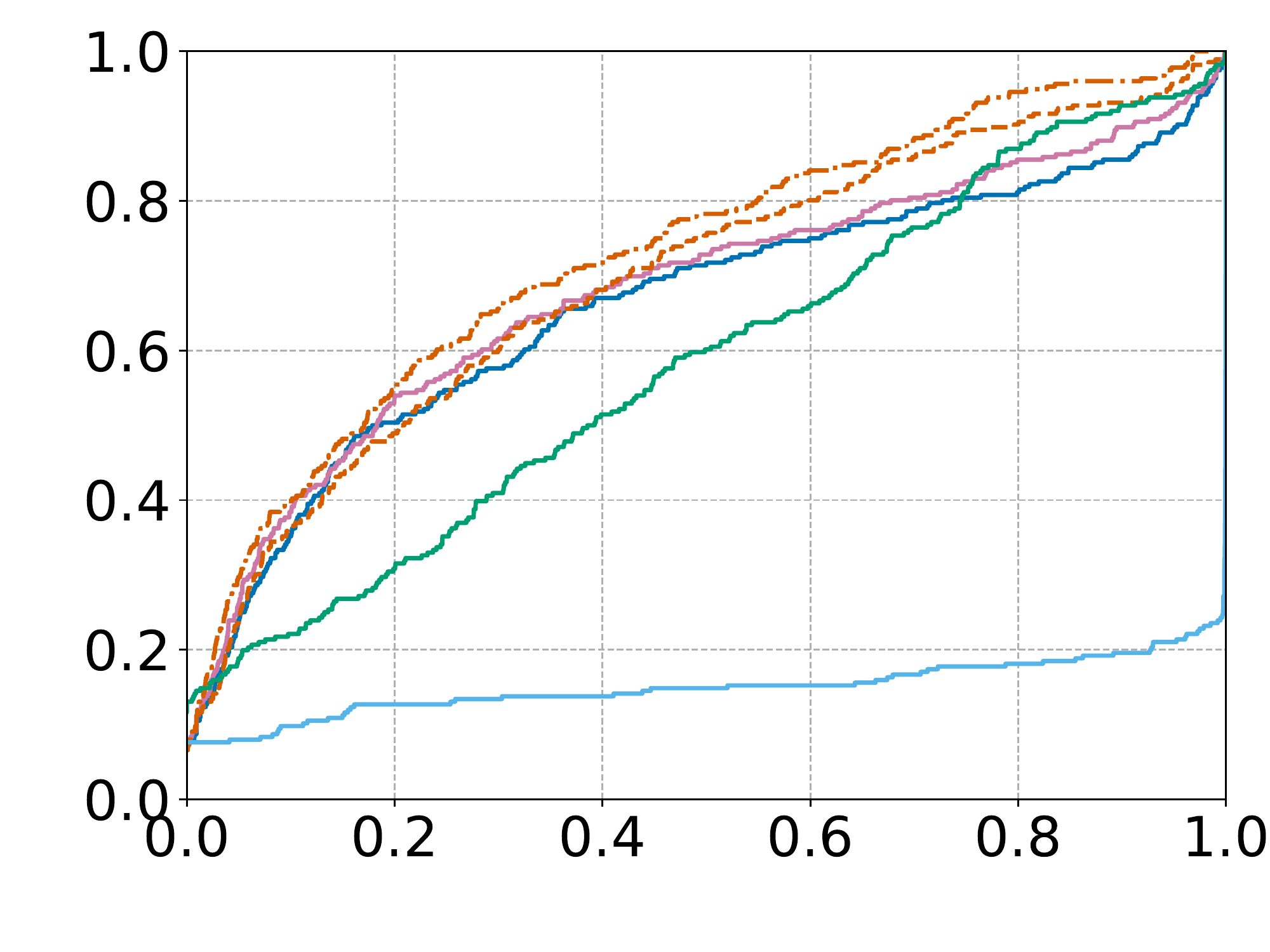}
&
\includegraphics[width=0.29\textwidth,valign=c,trim={0 0 0 0},clip]{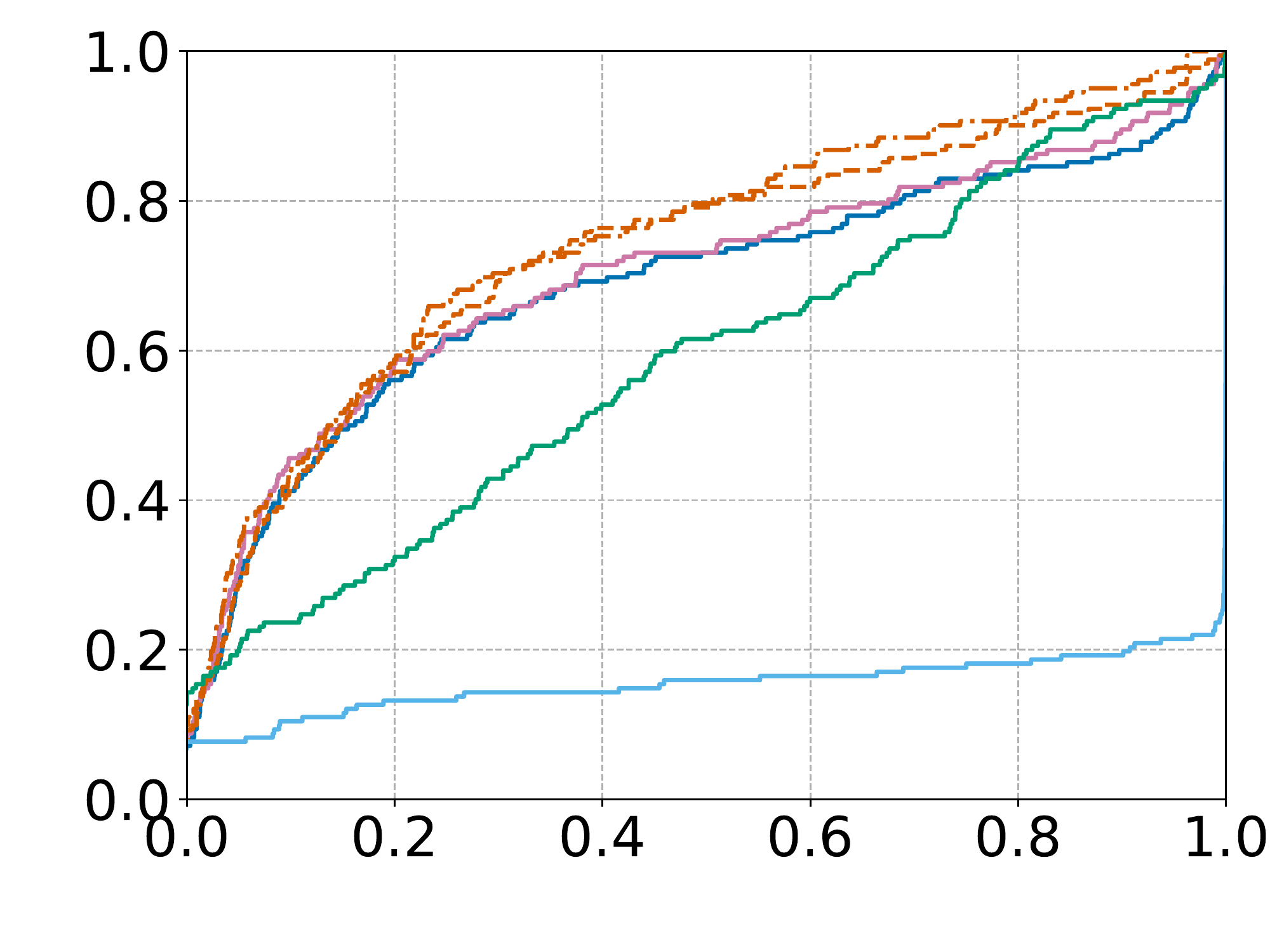}
&
\includegraphics[width=0.29\textwidth,valign=c,trim={0 0 0 0},clip]{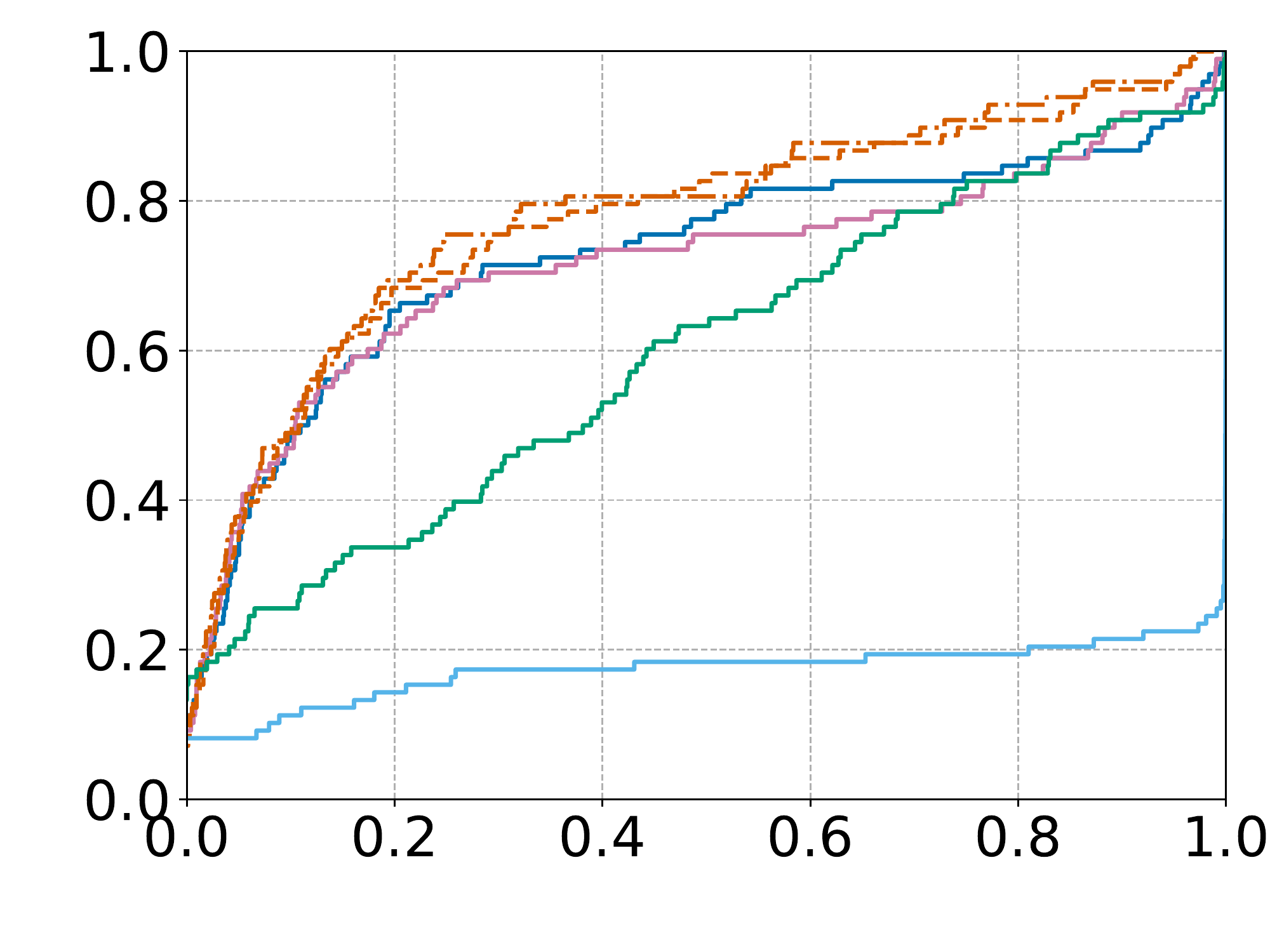}
\\\hline
\begin{tabular}{c}CYC \\ $p=0.25$\end{tabular}
& \includegraphics[width=0.29\textwidth,valign=c,trim={0 0 0 0},clip]{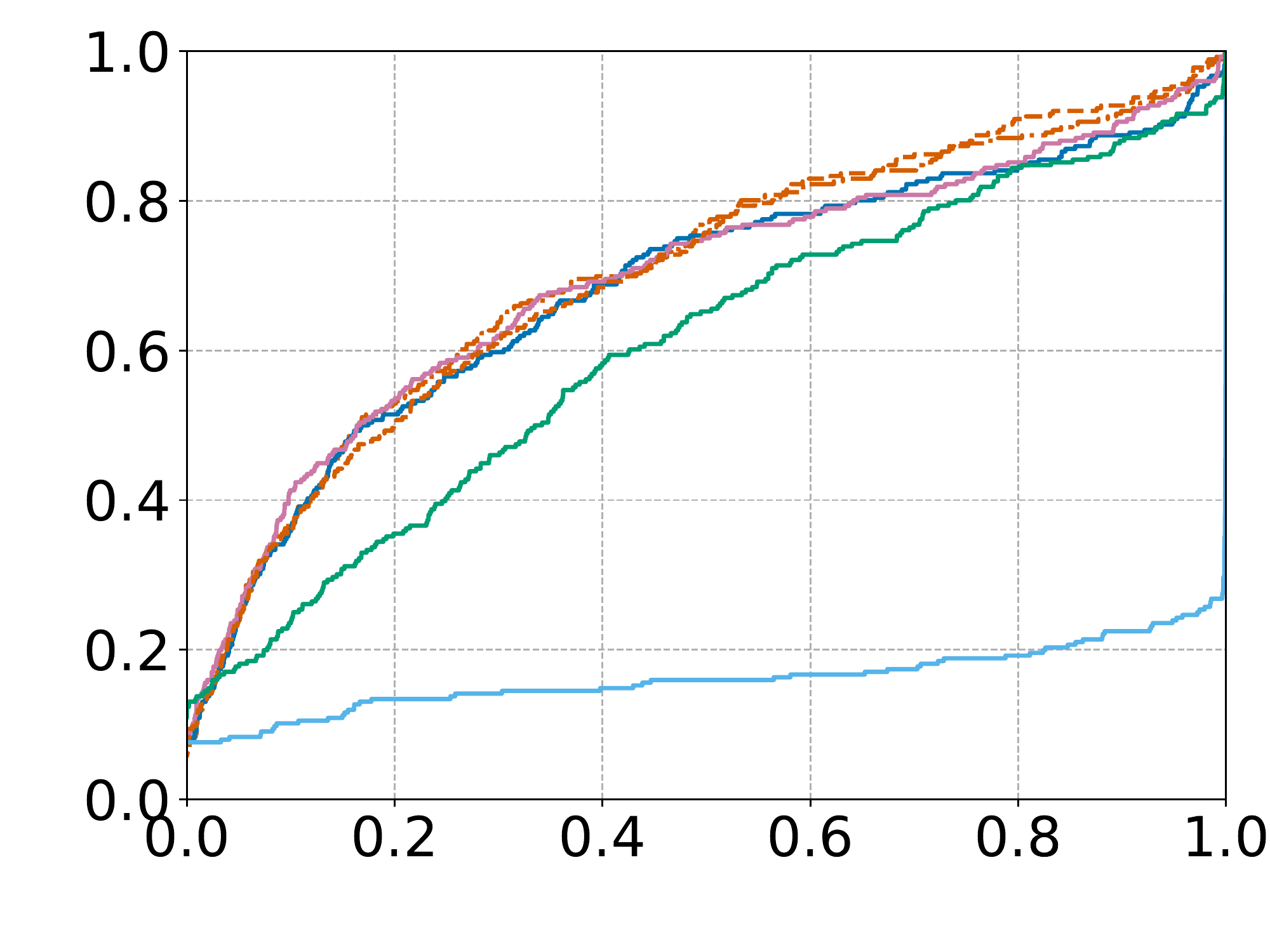}
&
\includegraphics[width=0.29\textwidth,valign=c,trim={0 0 0 0},clip]{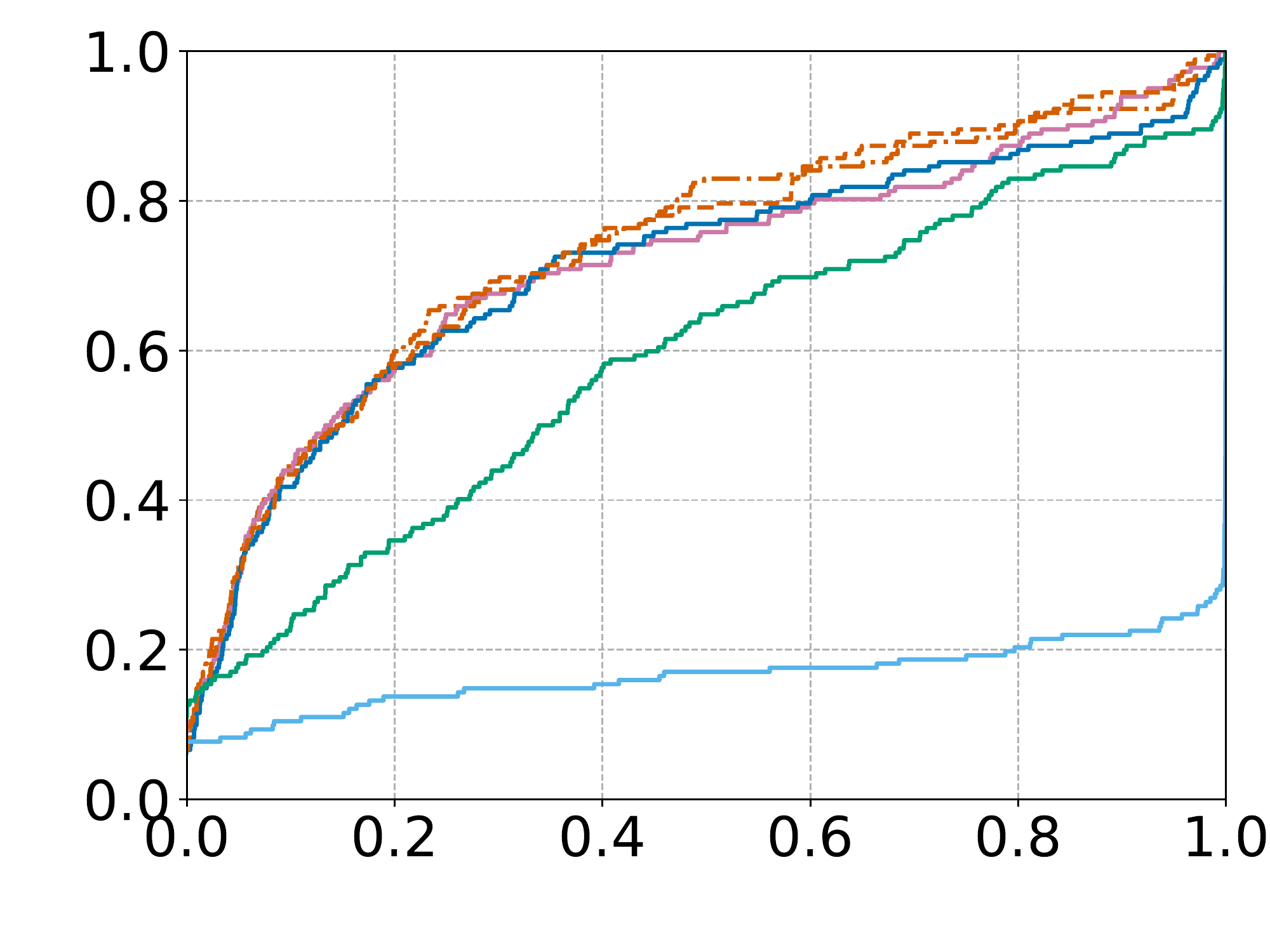}
&
\includegraphics[width=0.29\textwidth,valign=c,trim={0 0 0 0},clip]{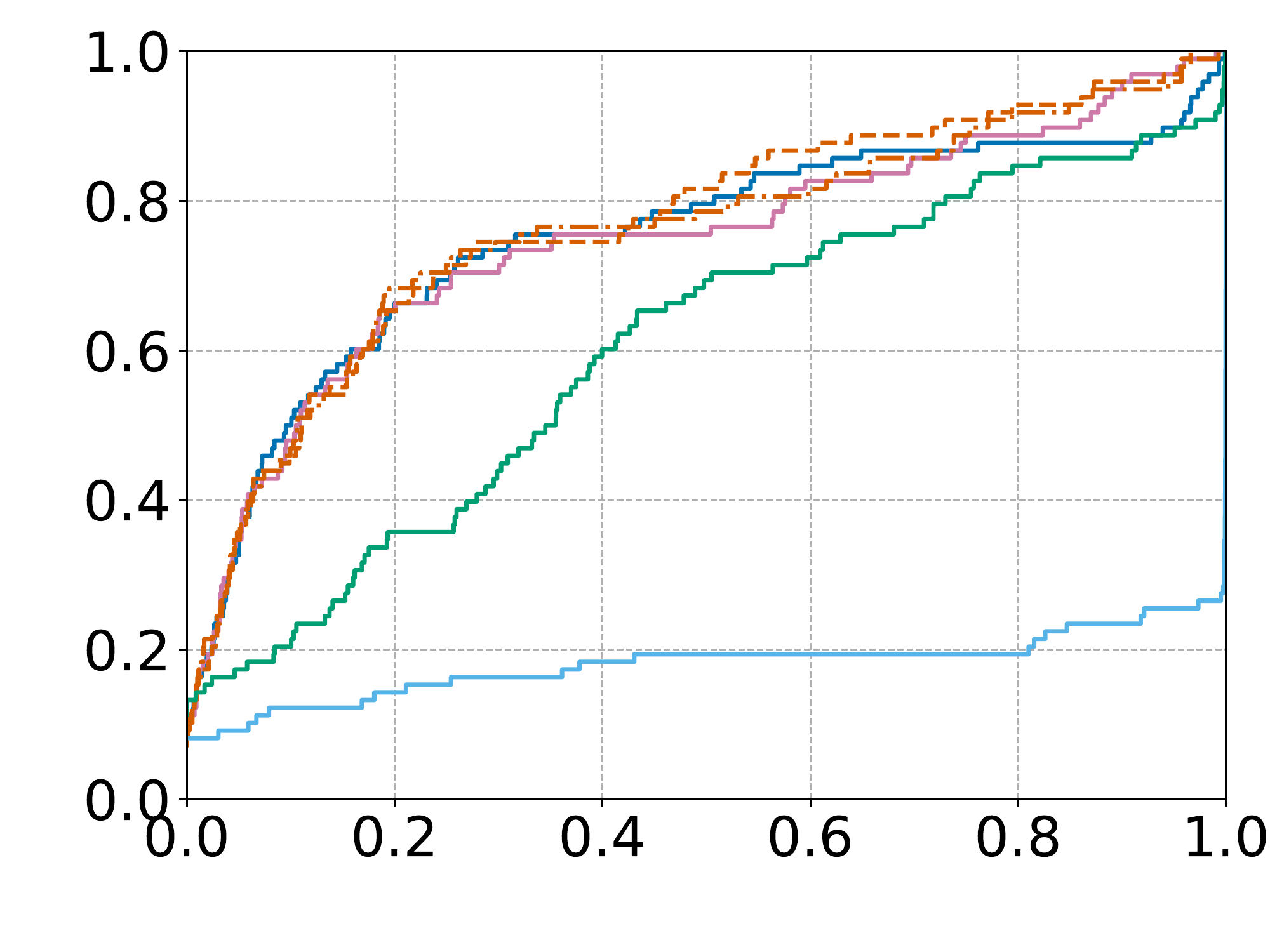}
\\\hline
\begin{tabular}{c}UAR \\ $p=0.5$\end{tabular}
& \includegraphics[width=0.29\textwidth,valign=c,trim={0 0 0 0},clip]{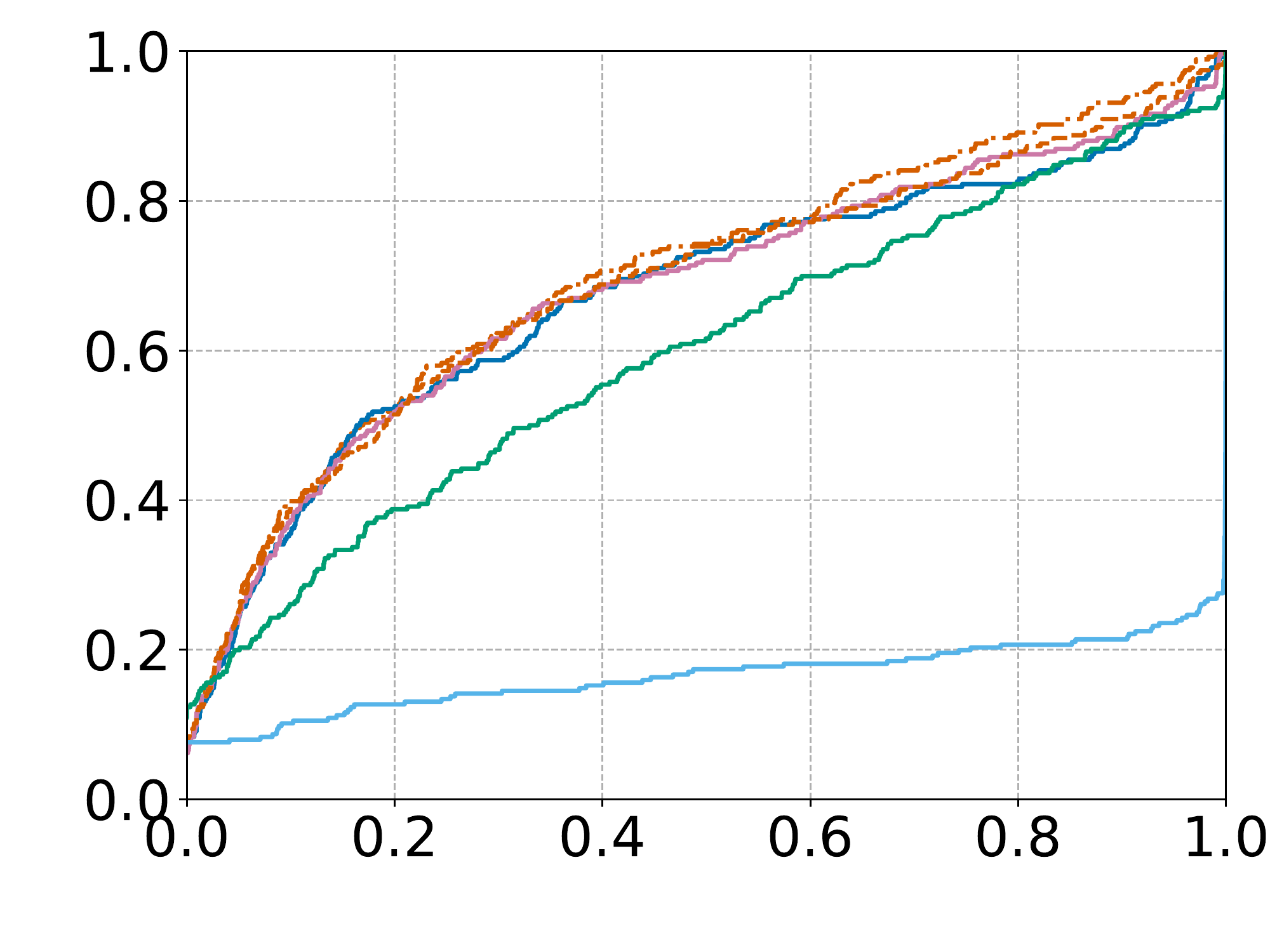}
&
\includegraphics[width=0.29\textwidth,valign=c,trim={0 0 0 0},clip]{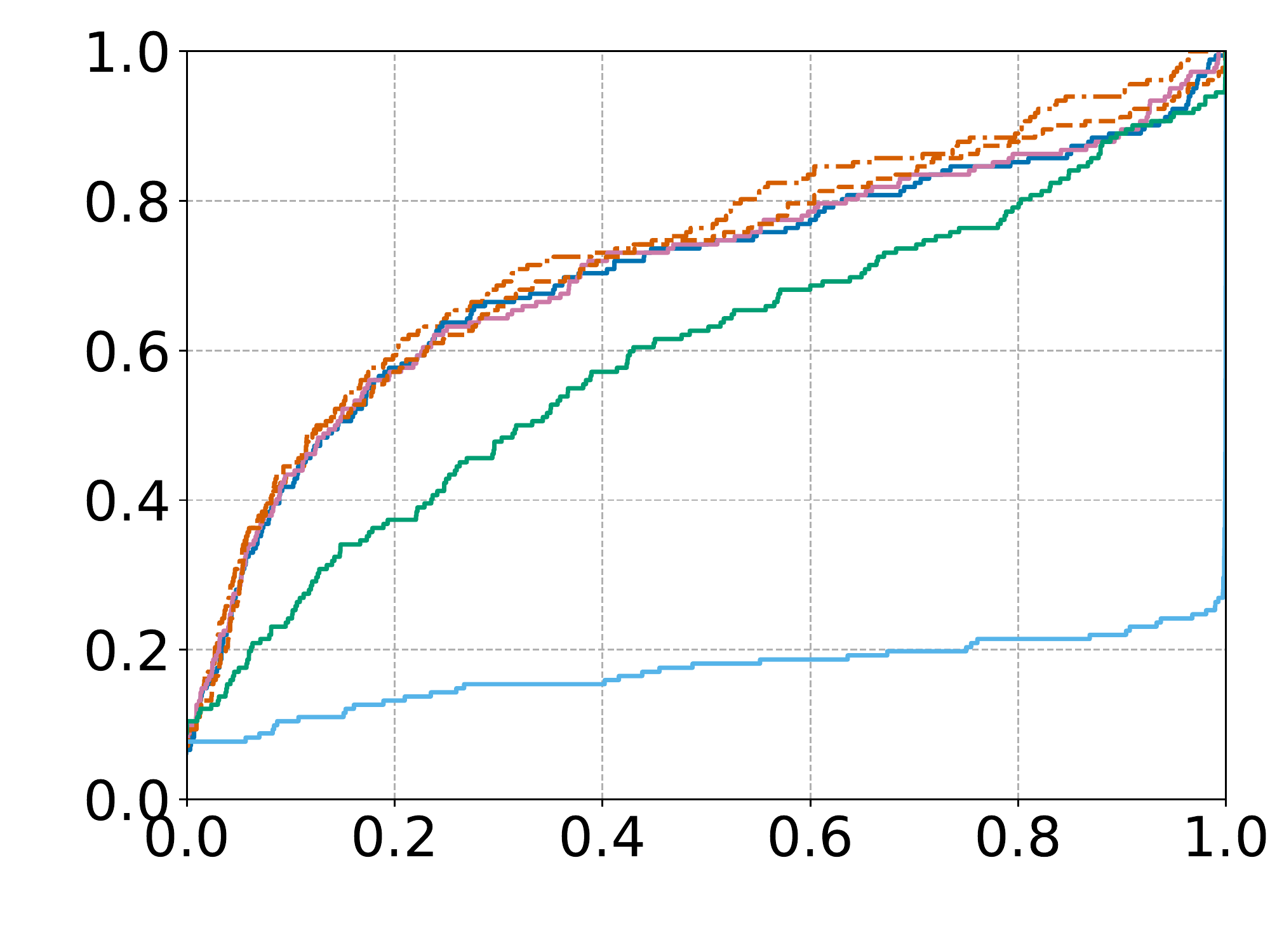}
&
\includegraphics[width=0.29\textwidth,valign=c,trim={0 0 0 0},clip]{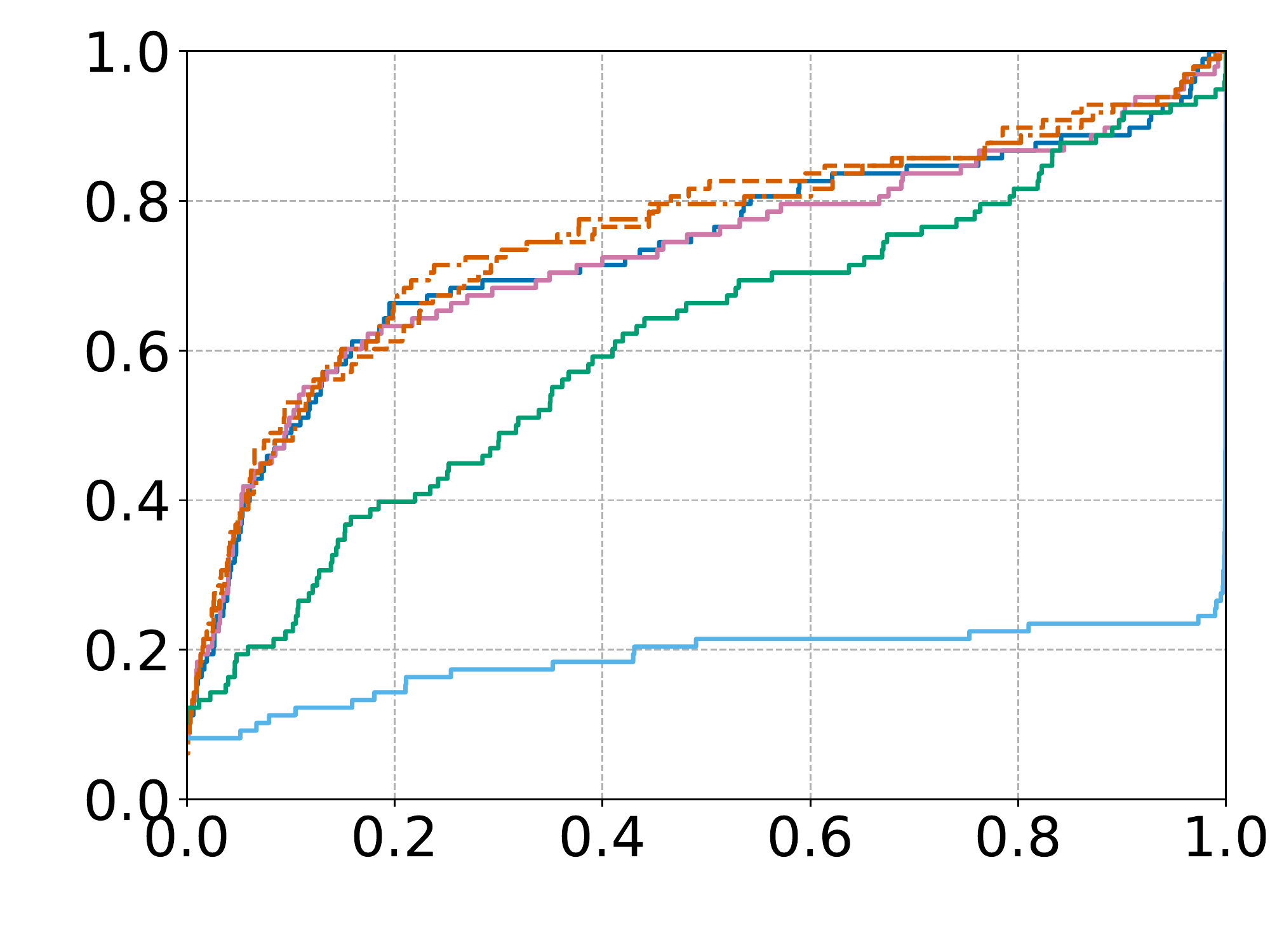}
\\\bottomrule
\multicolumn{4}{c}{
		\begin{subfigure}[t]{\textwidth}
    \includegraphics[width=\textwidth,trim={0 0 0 0},clip]{figs/cdfs/legends/legend_Lambda=8.pdf}
		\end{subfigure} }
\end{tabular}
\caption{Cumulative distribution functions (CDFs) for all evaluated algorithms, against different noise settings and warm start ratios in $\cbr{23.0,46.0,92.0}$. All CB algorithms use $\epsilon$-greedy with $\epsilon=0.00625$. In each of the above plots, the $x$ axis represents scores, while the $y$ axis represents the CDF values.}
\label{fig:cdfs-eps=0.00625-2}
\end{figure}

\begin{figure}[H]
\centering
\begin{tabular}{c |@{}c@{}} 
\toprule
& \multicolumn{1}{c}{ Ratio }
\\
Noise & 184.0
\\\midrule
Noiseless & \includegraphics[width=0.3\textwidth,valign=c,trim={0 0 0 0},clip]{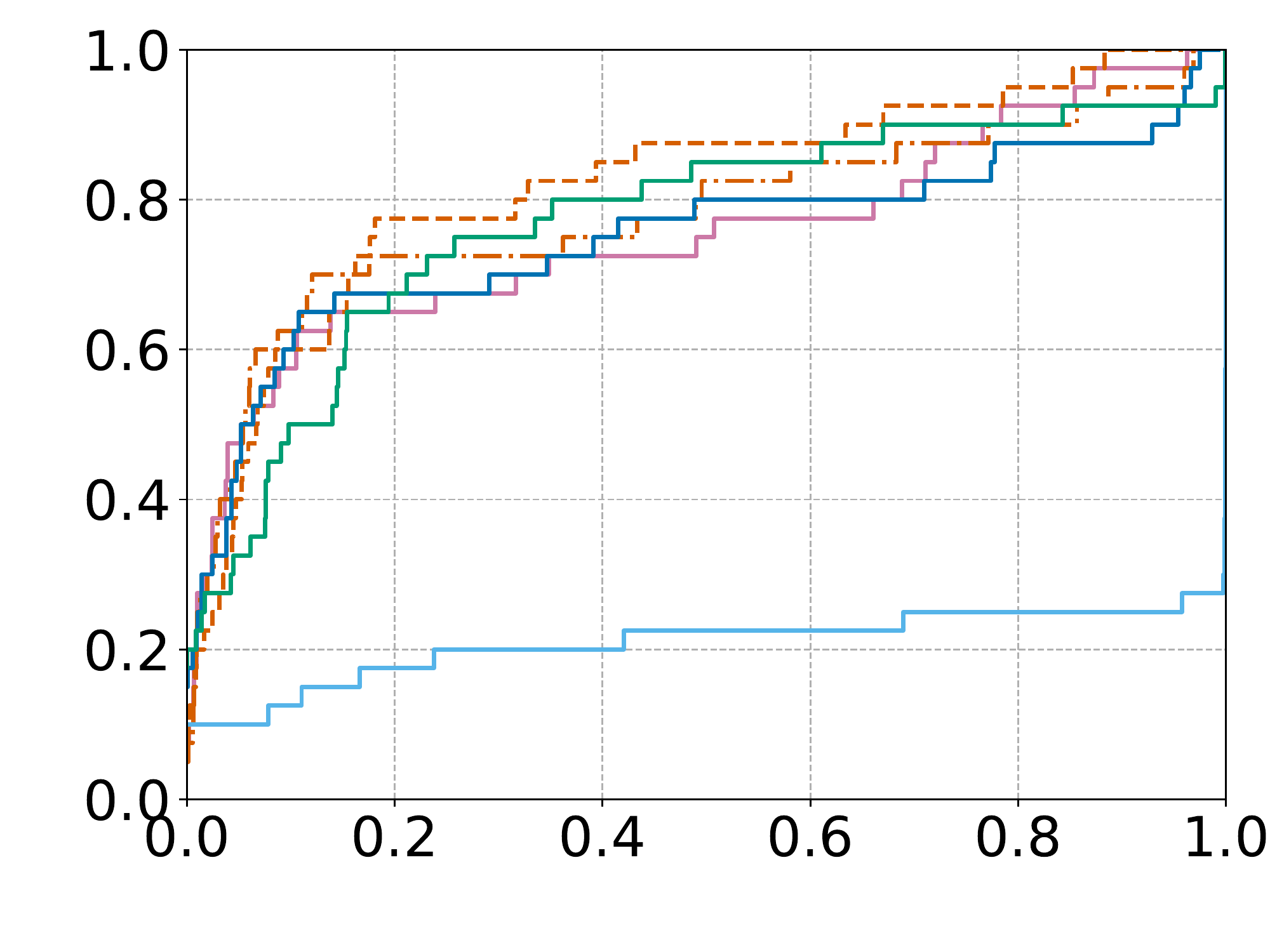}
\\\hline
\begin{tabular}{c}UAR \\ $p=0.25$\end{tabular}
& \includegraphics[width=0.3\textwidth,valign=c,trim={0 0 0 0},clip]{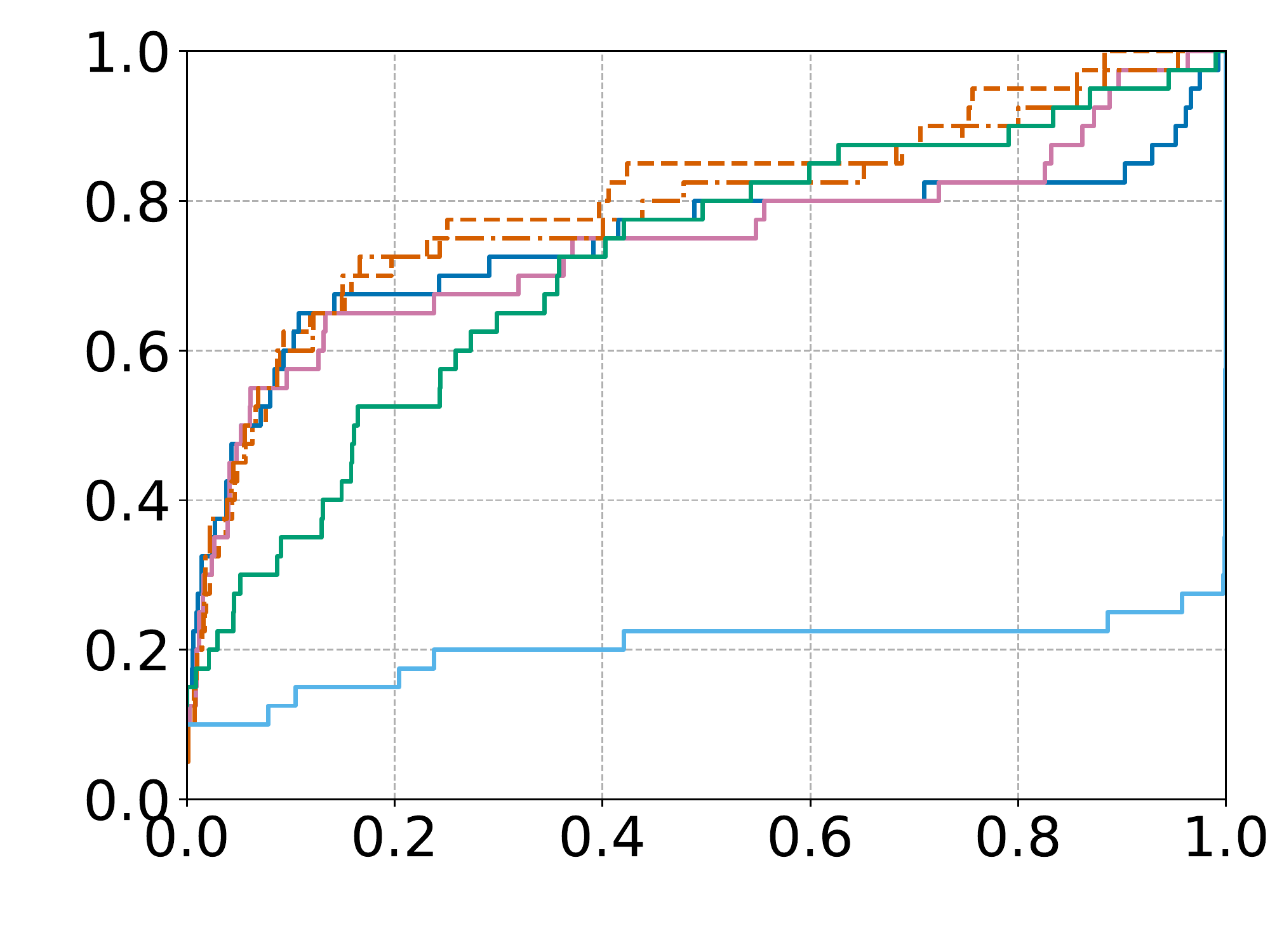}

\\\hline
\begin{tabular}{c}MAJ \\ $p=0.25$\end{tabular}
& \includegraphics[width=0.3\textwidth,valign=c,trim={0 0 0 0},clip]{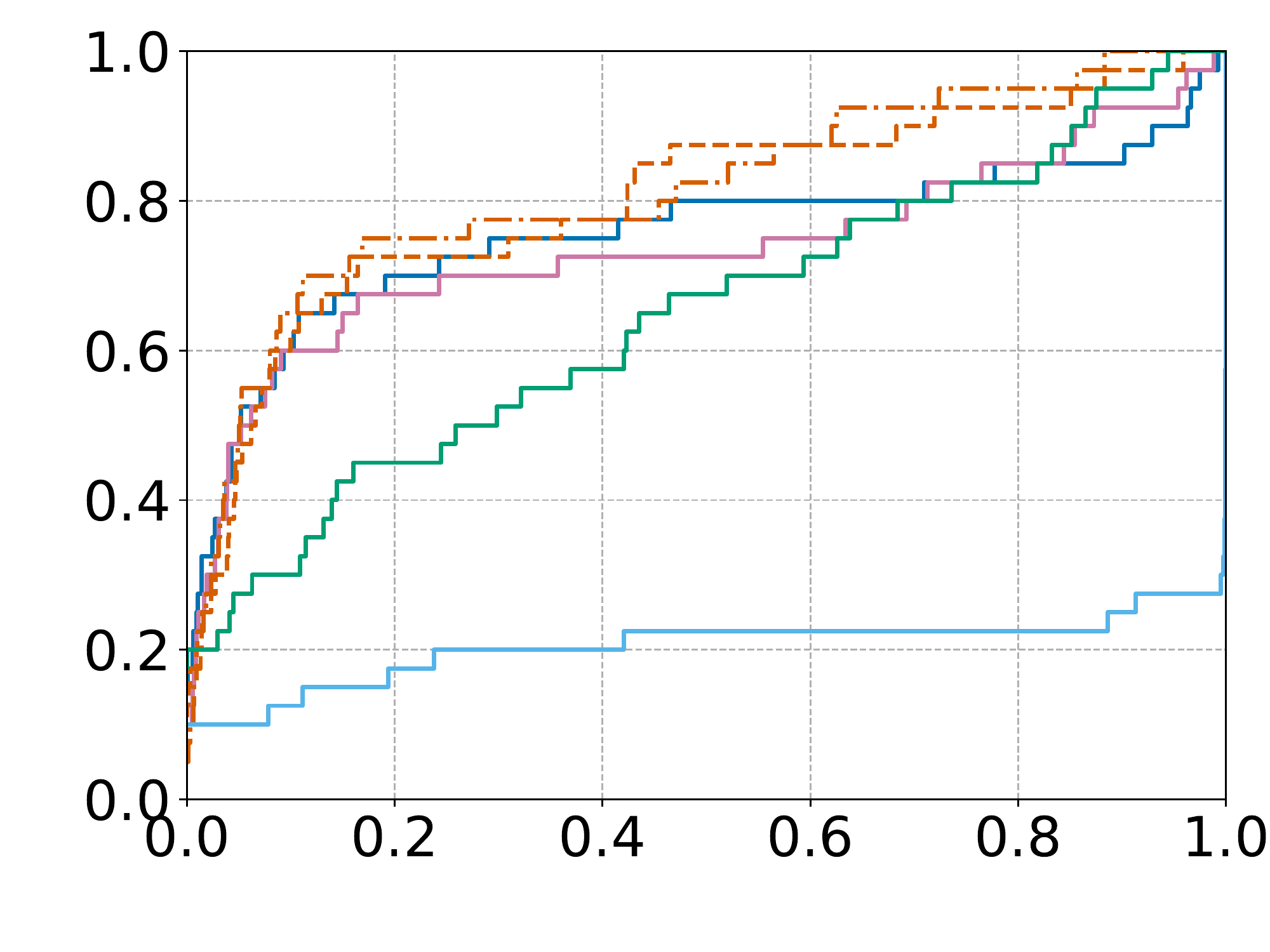}

\\\hline
\begin{tabular}{c}CYC \\ $p=0.25$\end{tabular}
& \includegraphics[width=0.3\textwidth,valign=c,trim={0 0 0 0},clip]{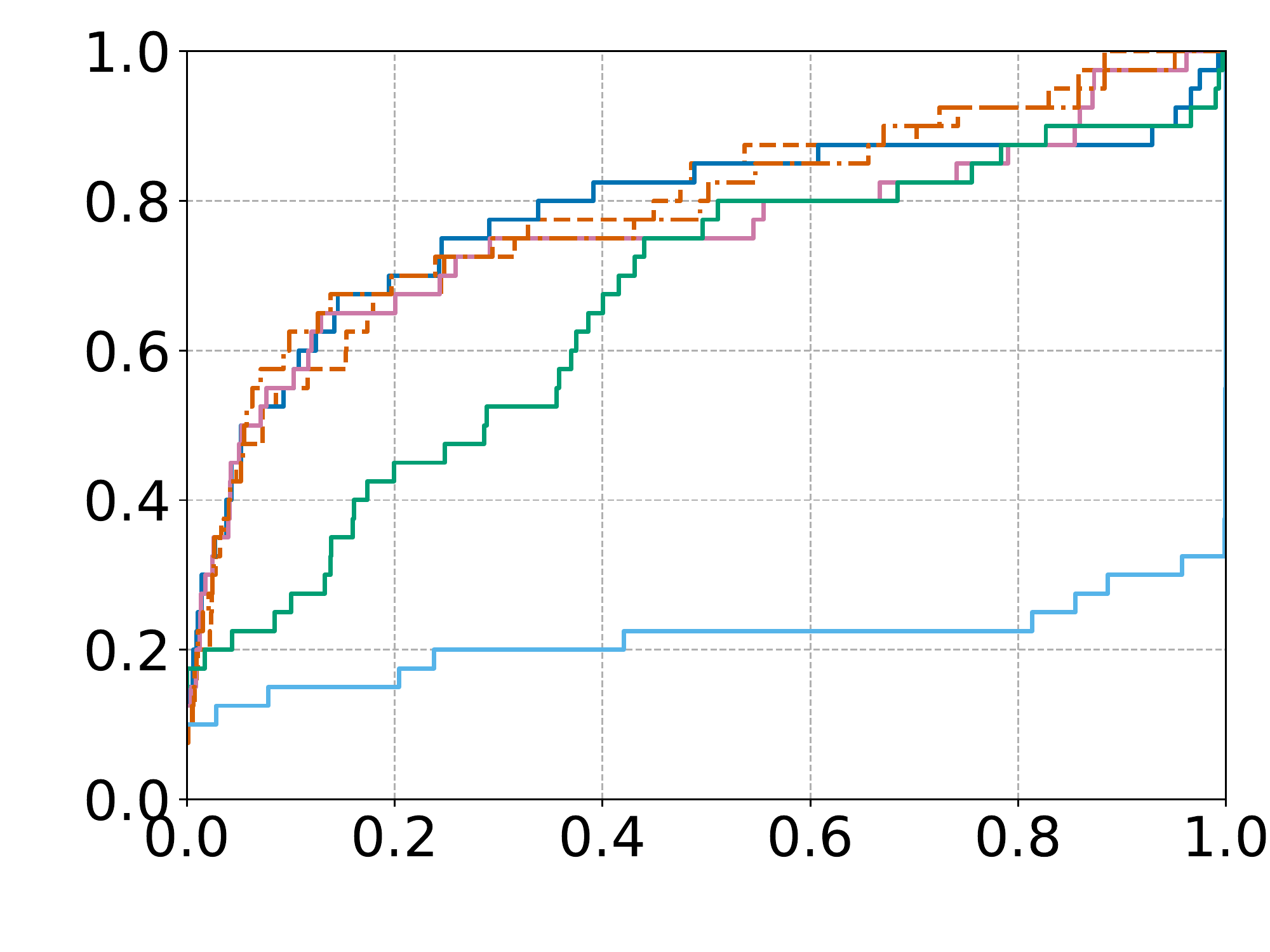}

\\\hline
\begin{tabular}{c}UAR \\ $p=0.5$\end{tabular}
& \includegraphics[width=0.3\textwidth,valign=c,trim={0 0 0 0},clip]{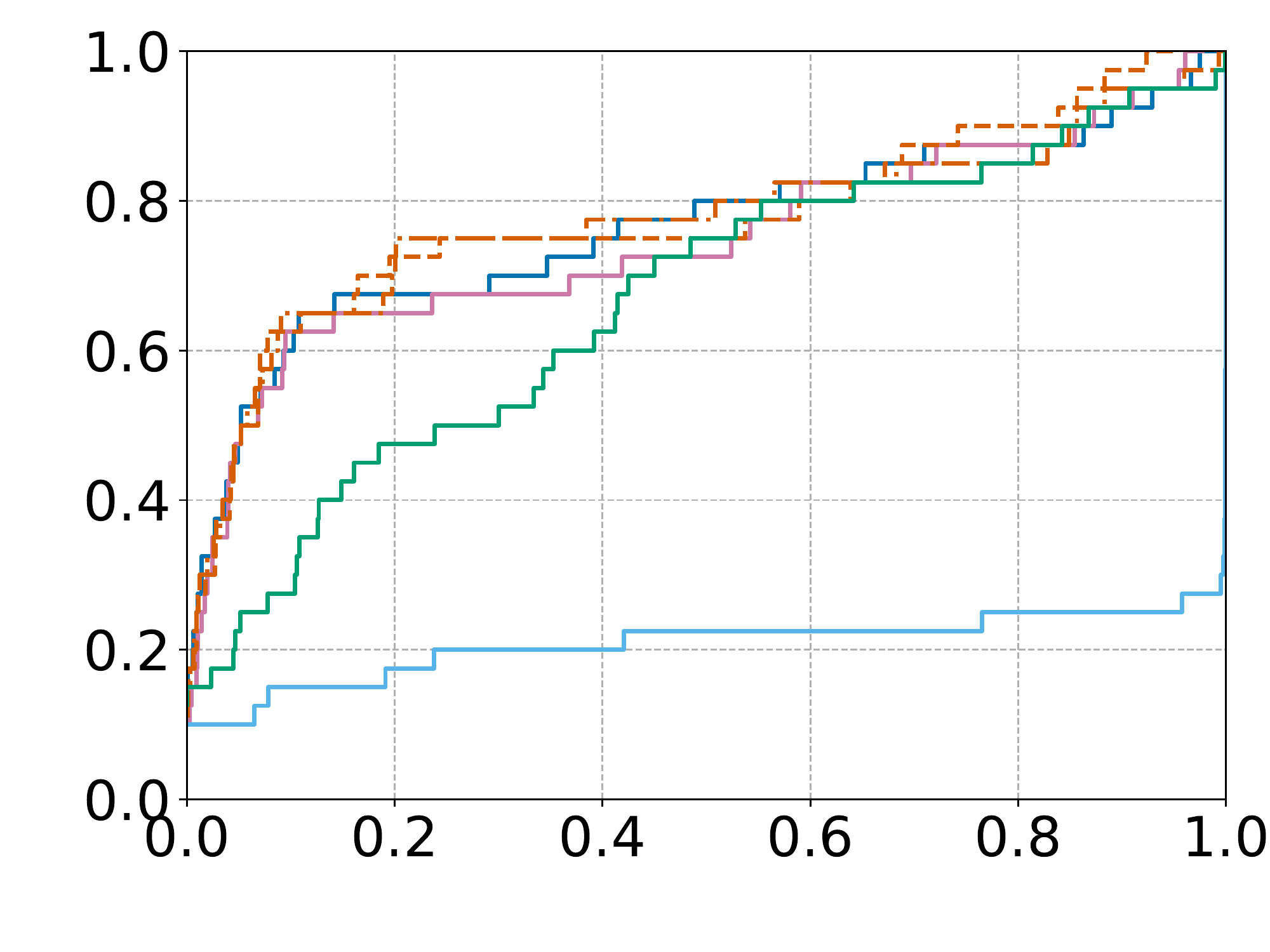}

\\\bottomrule
\multicolumn{2}{c}{
		\begin{subfigure}[t]{\textwidth}
    \includegraphics[width=\textwidth,trim={0 0 0 0},clip]{figs/cdfs/legends/legend_Lambda=8.pdf}
		\end{subfigure} }
\end{tabular}
\caption{Cumulative distribution functions (CDFs) for all evaluated algorithms, against different noise settings and warm start ratios in $\cbr{184.0}$. All CB algorithms use $\epsilon$-greedy with $\epsilon=0.00625$. In each of the above plots, the $x$ axis represents scores, while the $y$ axis represents the CDF values.}
\label{fig:cdfs-eps=0.00625-3}
\end{figure}

\begin{figure}[H]
\centering
\begin{tabular}{c | @{}c@{ }c@{ }c@{}} 
\toprule
& \multicolumn{3}{c}{ Ratio }
\\
Noise & 2.875 & 5.75 & 11.5
\\\midrule
\begin{tabular}{c}MAJ \\ $p=0.5$\end{tabular}
 & \includegraphics[width=0.29\textwidth,valign=c,trim={0 0 0 0},clip]{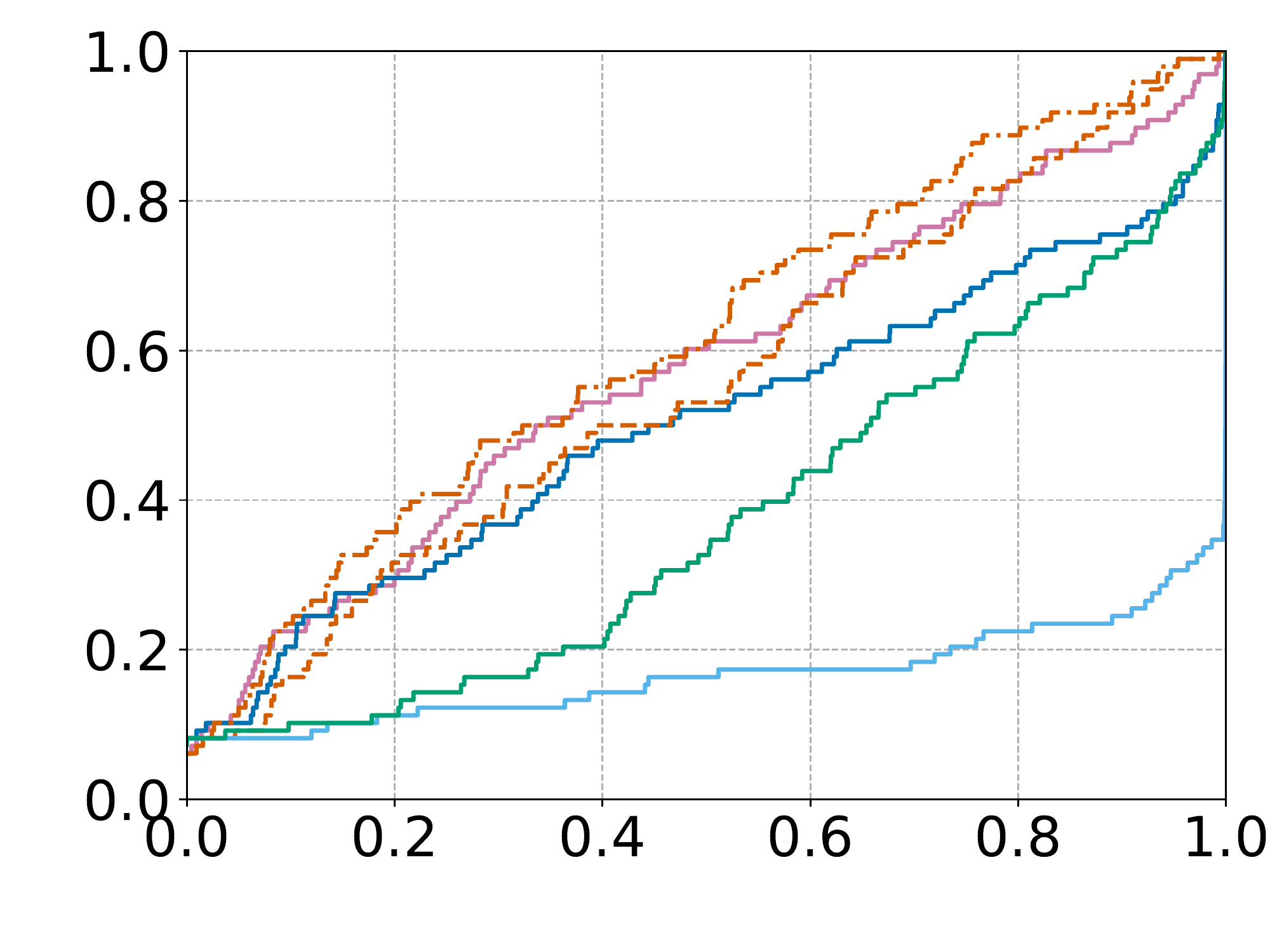}
&
\includegraphics[width=0.29\textwidth,valign=c,trim={0 0 0 0},clip]{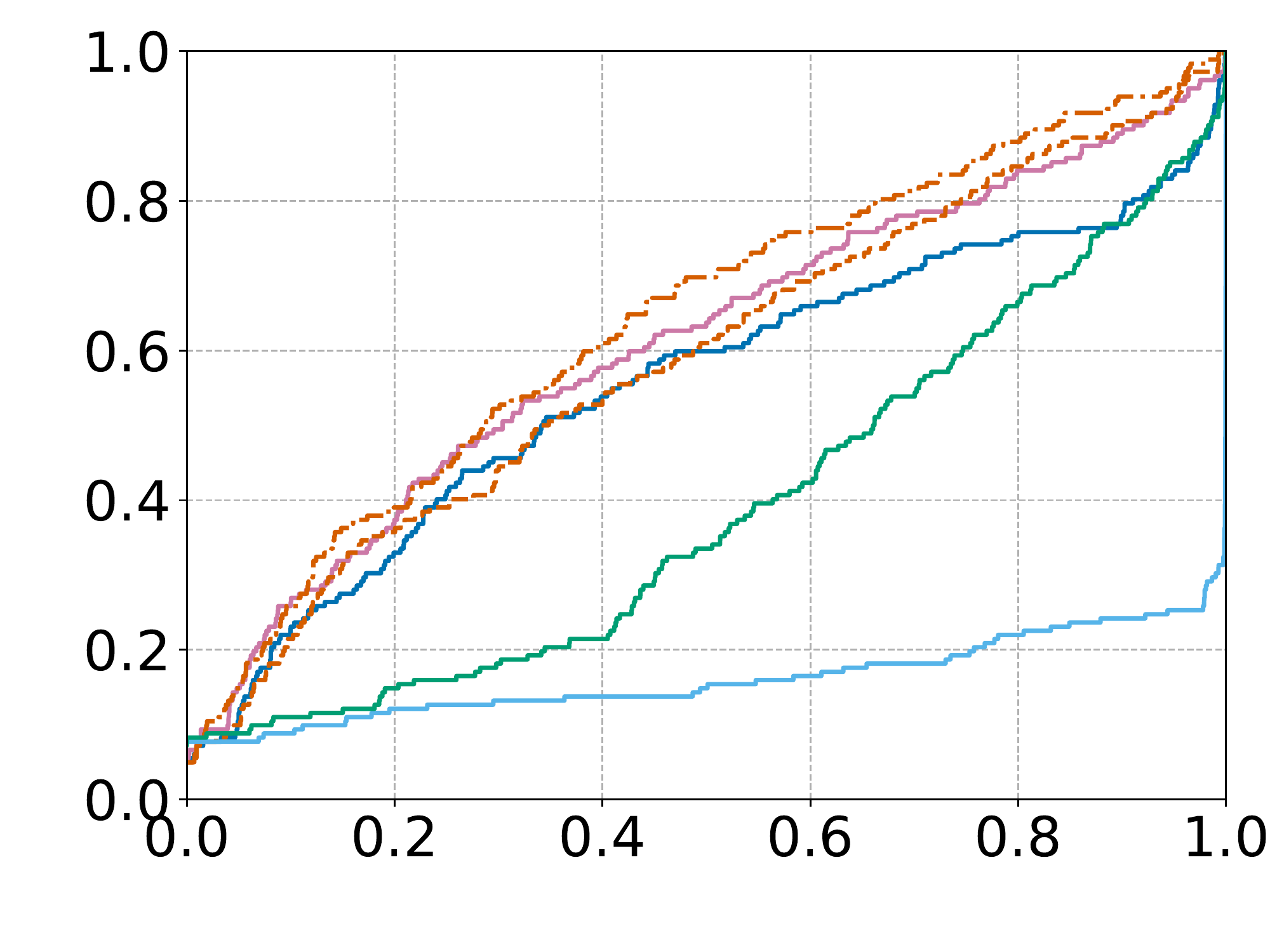}
&
\includegraphics[width=0.29\textwidth,valign=c,trim={0 0 0 0},clip]{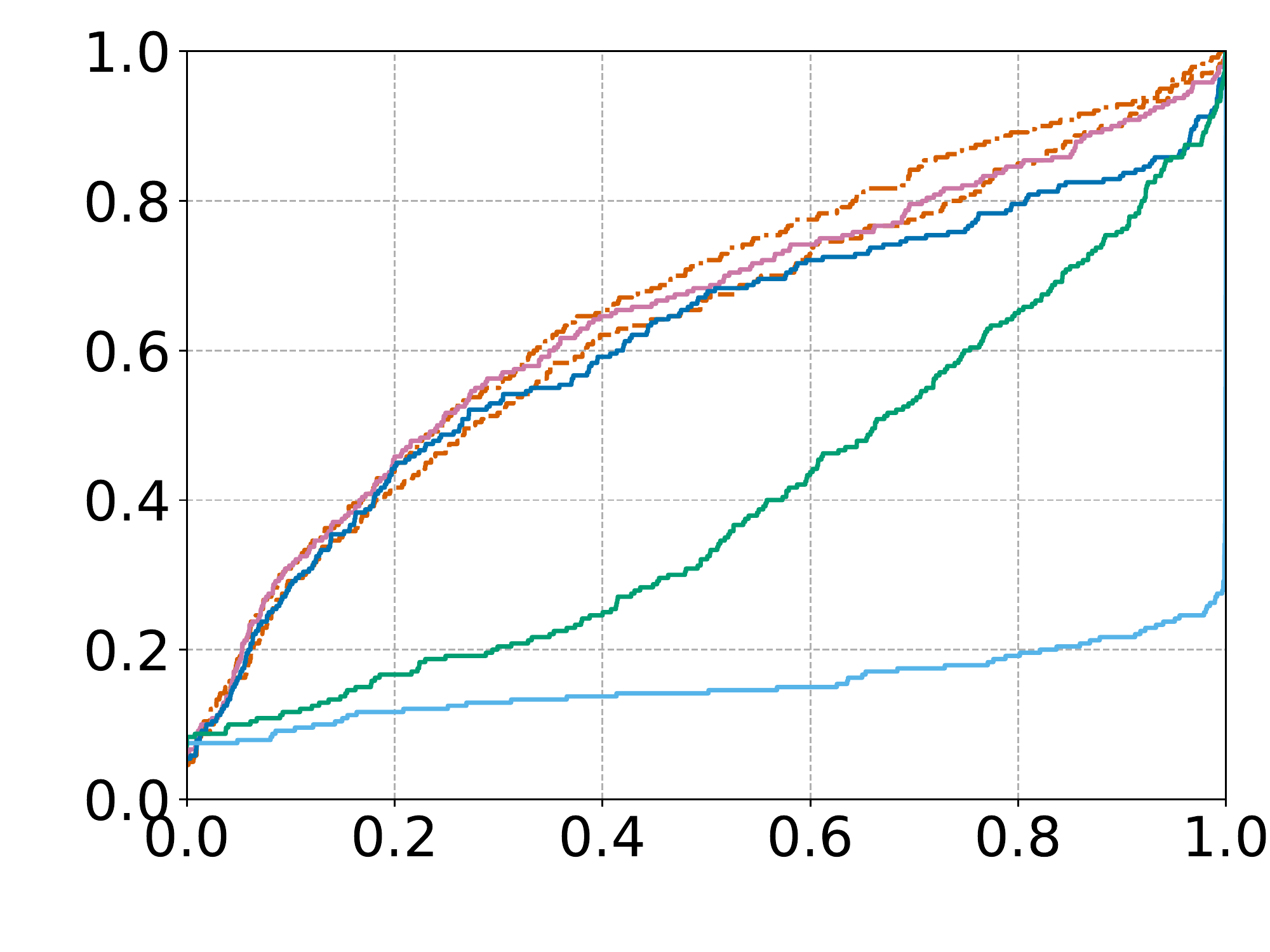}
\\\hline
\begin{tabular}{c}CYC \\ $p=0.5$\end{tabular}
& \includegraphics[width=0.29\textwidth,valign=c,trim={0 0 0 0},clip]{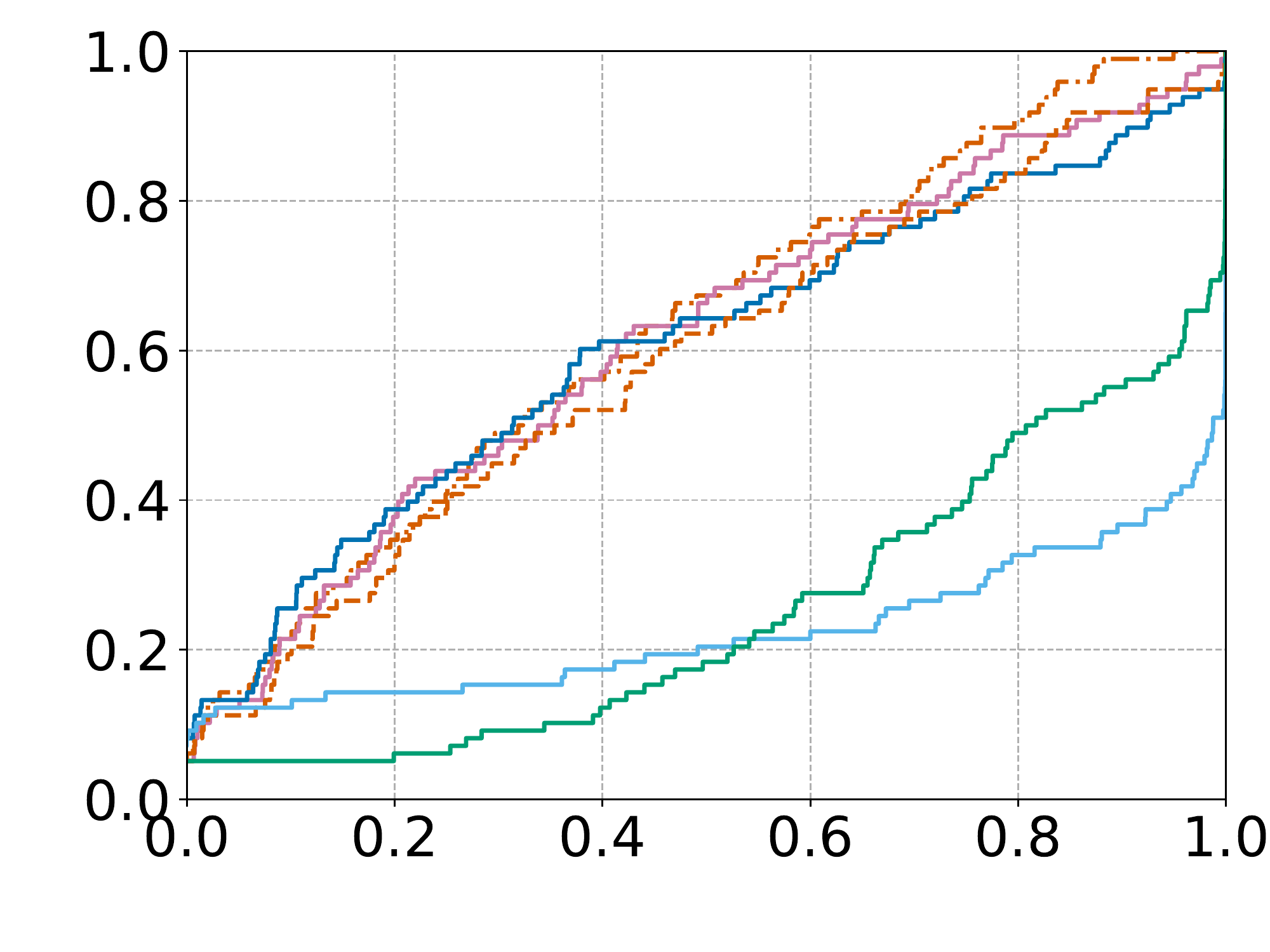}
&
\includegraphics[width=0.29\textwidth,valign=c,trim={0 0 0 0},clip]{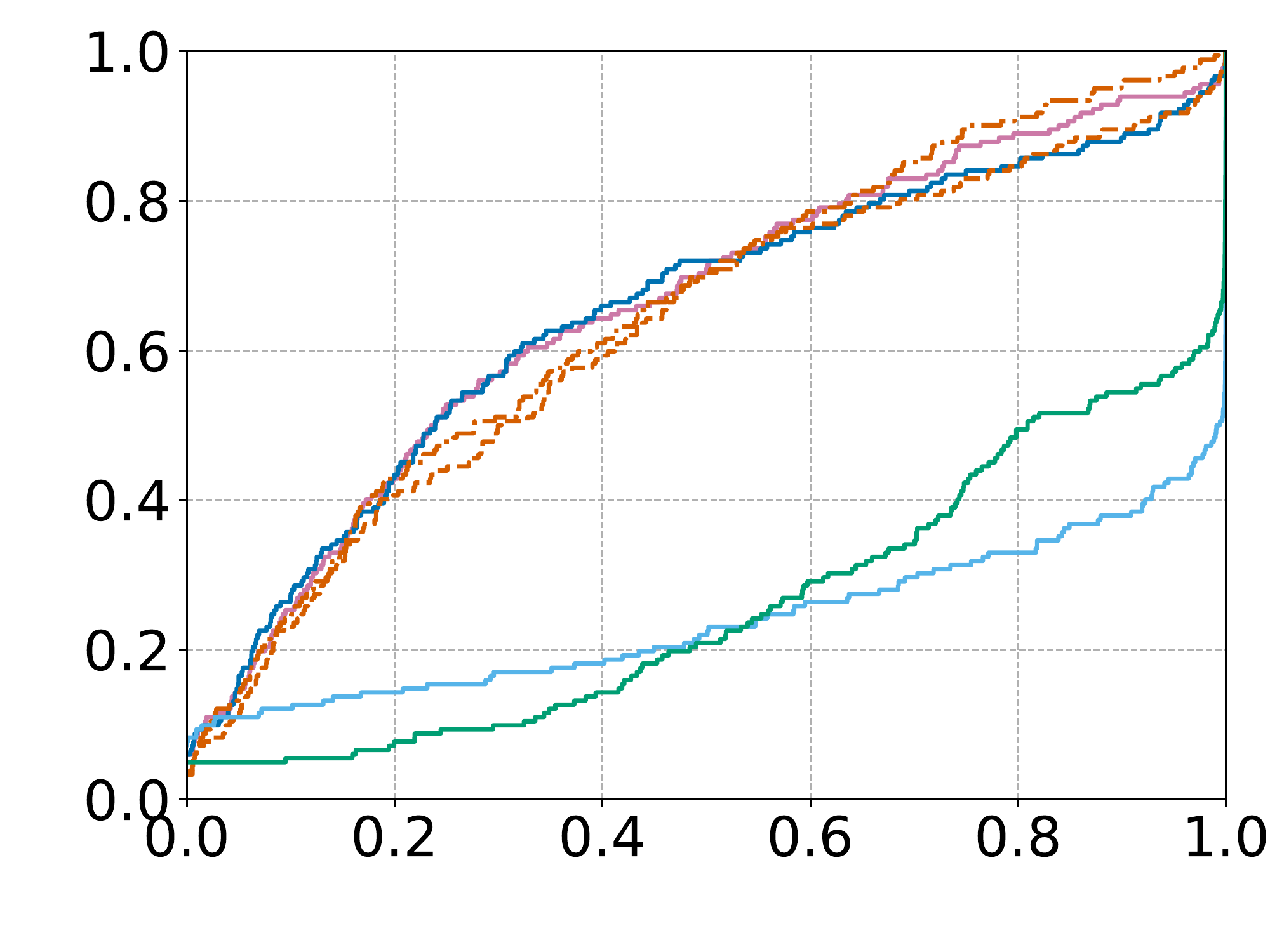}
&
\includegraphics[width=0.29\textwidth,valign=c,trim={0 0 0 0},clip]{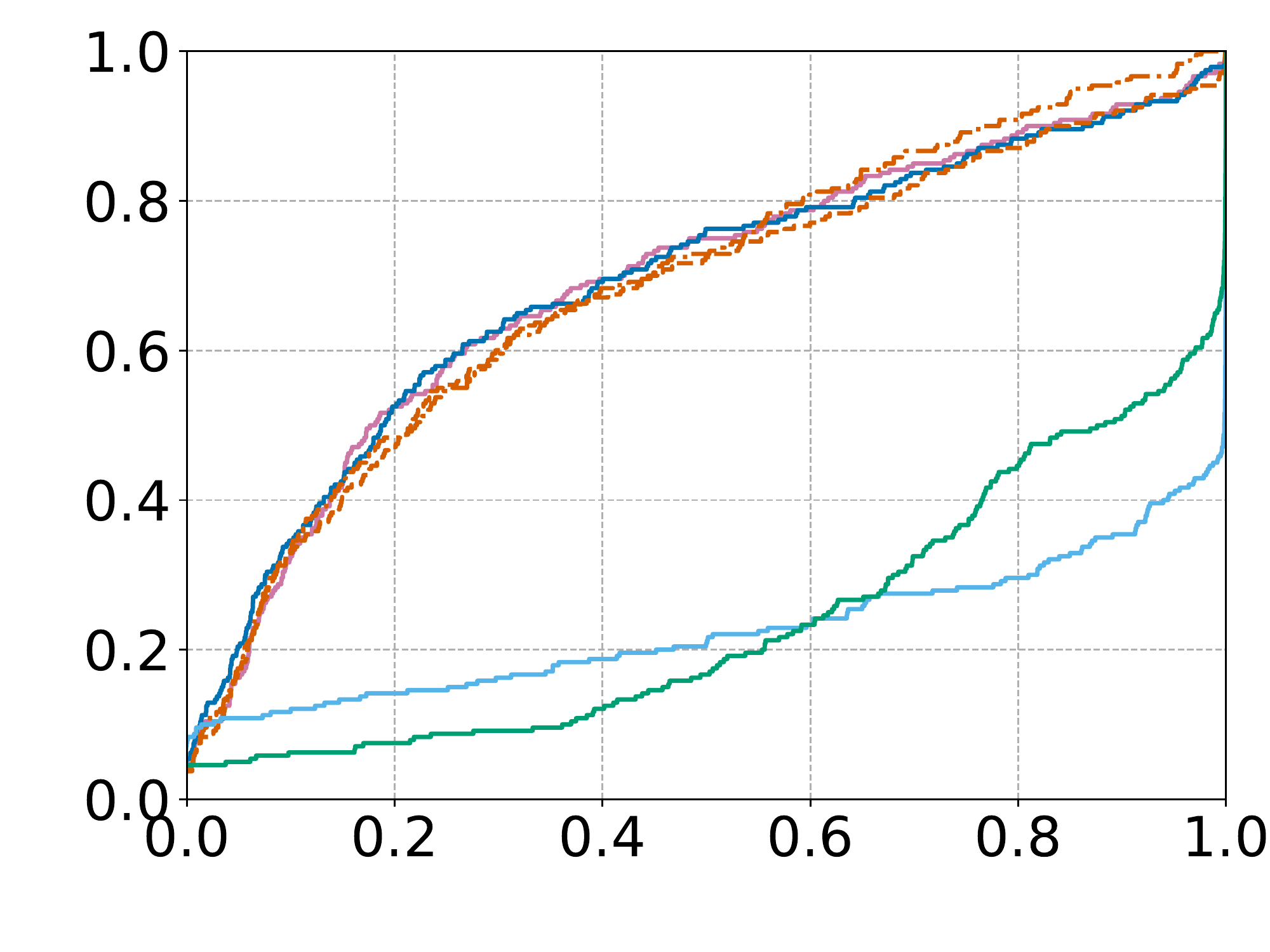}
\\\hline
\begin{tabular}{c}UAR \\ $p=1$\end{tabular}
& \includegraphics[width=0.29\textwidth,valign=c,trim={0 0 0 0},clip]{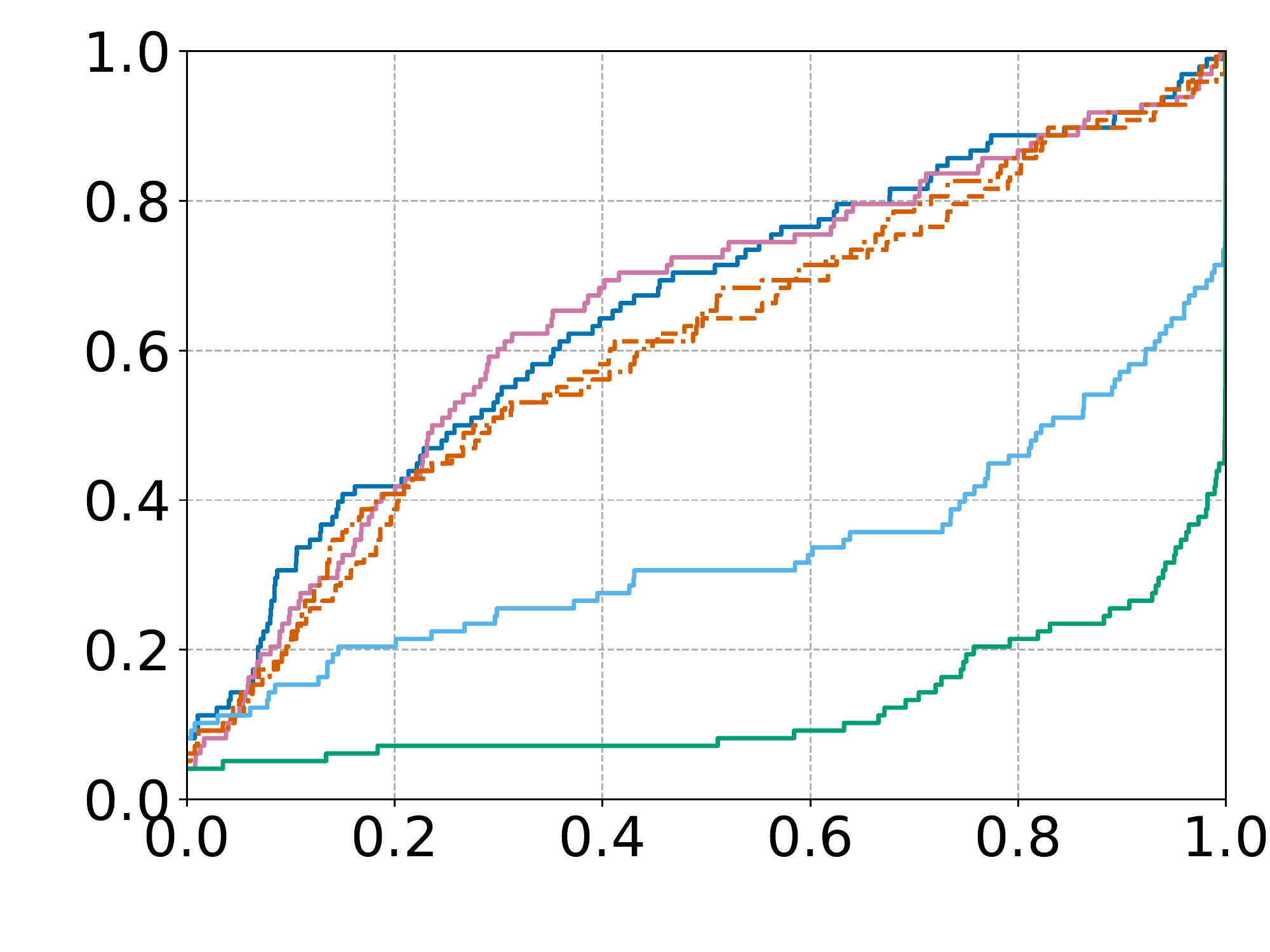}
&
\includegraphics[width=0.29\textwidth,valign=c,trim={0 0 0 0},clip]{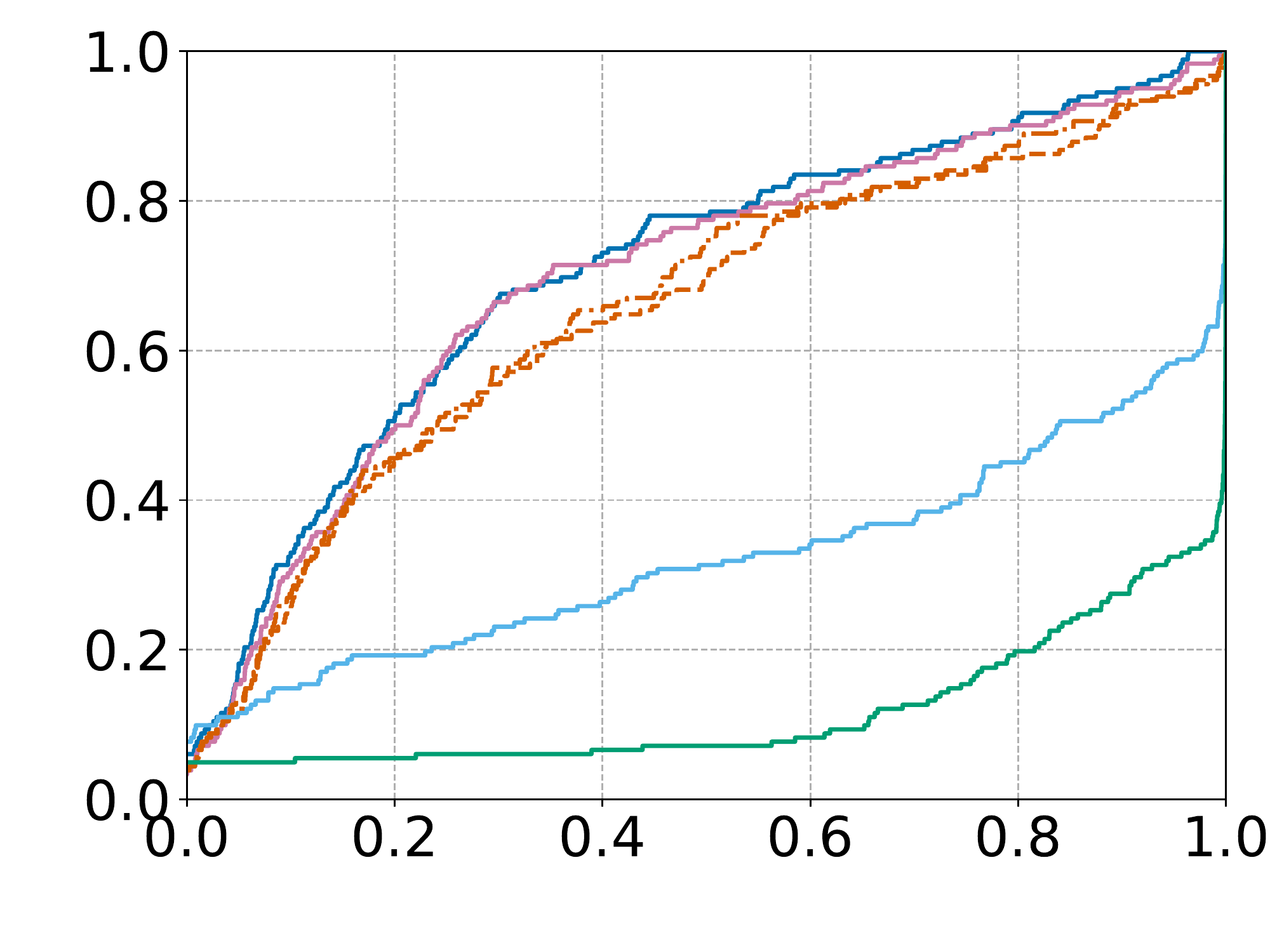}
&
\includegraphics[width=0.29\textwidth,valign=c,trim={0 0 0 0},clip]{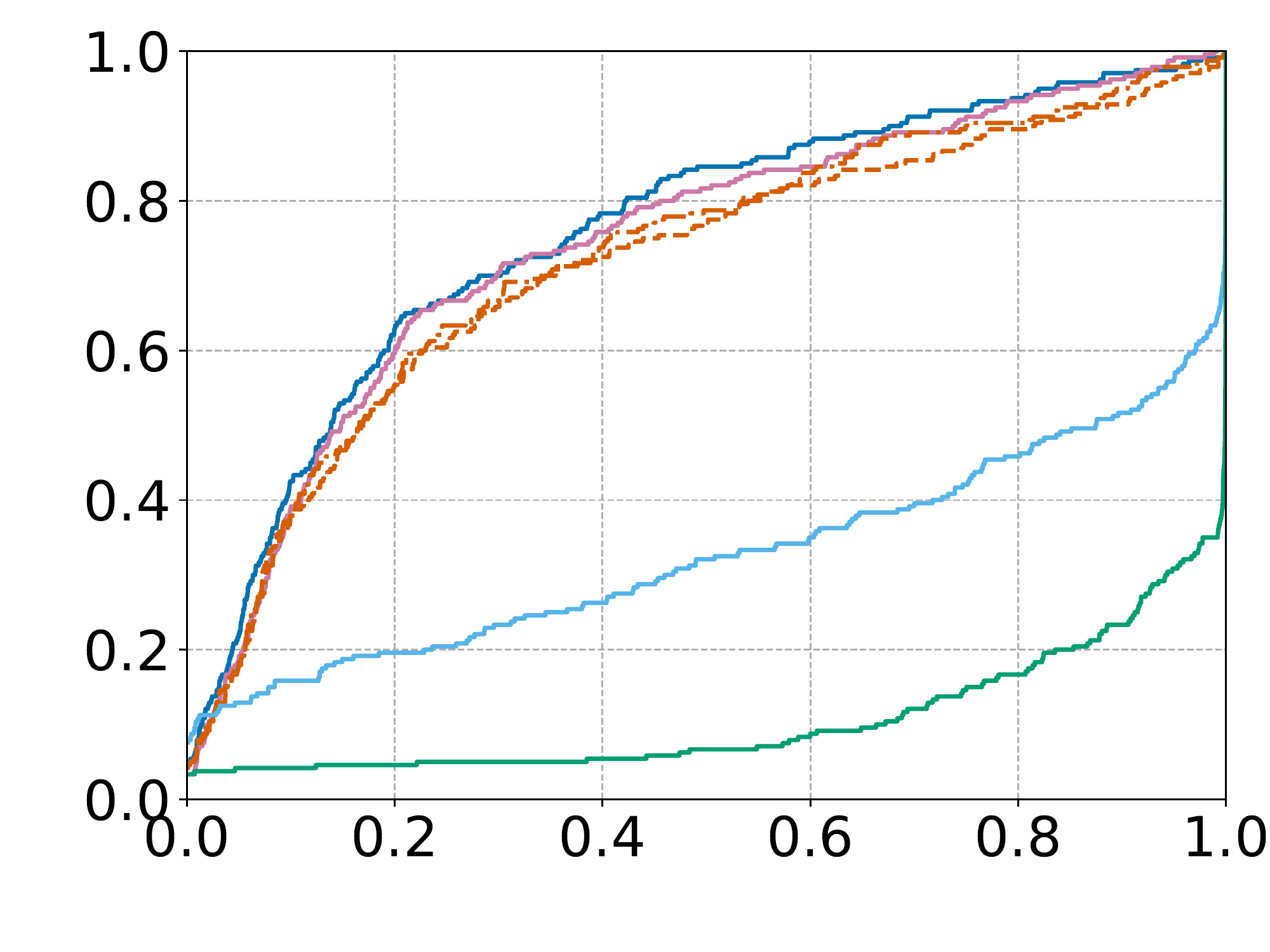}
\\\hline
\begin{tabular}{c}MAJ \\ $p=1$\end{tabular}
& \includegraphics[width=0.29\textwidth,valign=c,trim={0 0 0 0},clip]{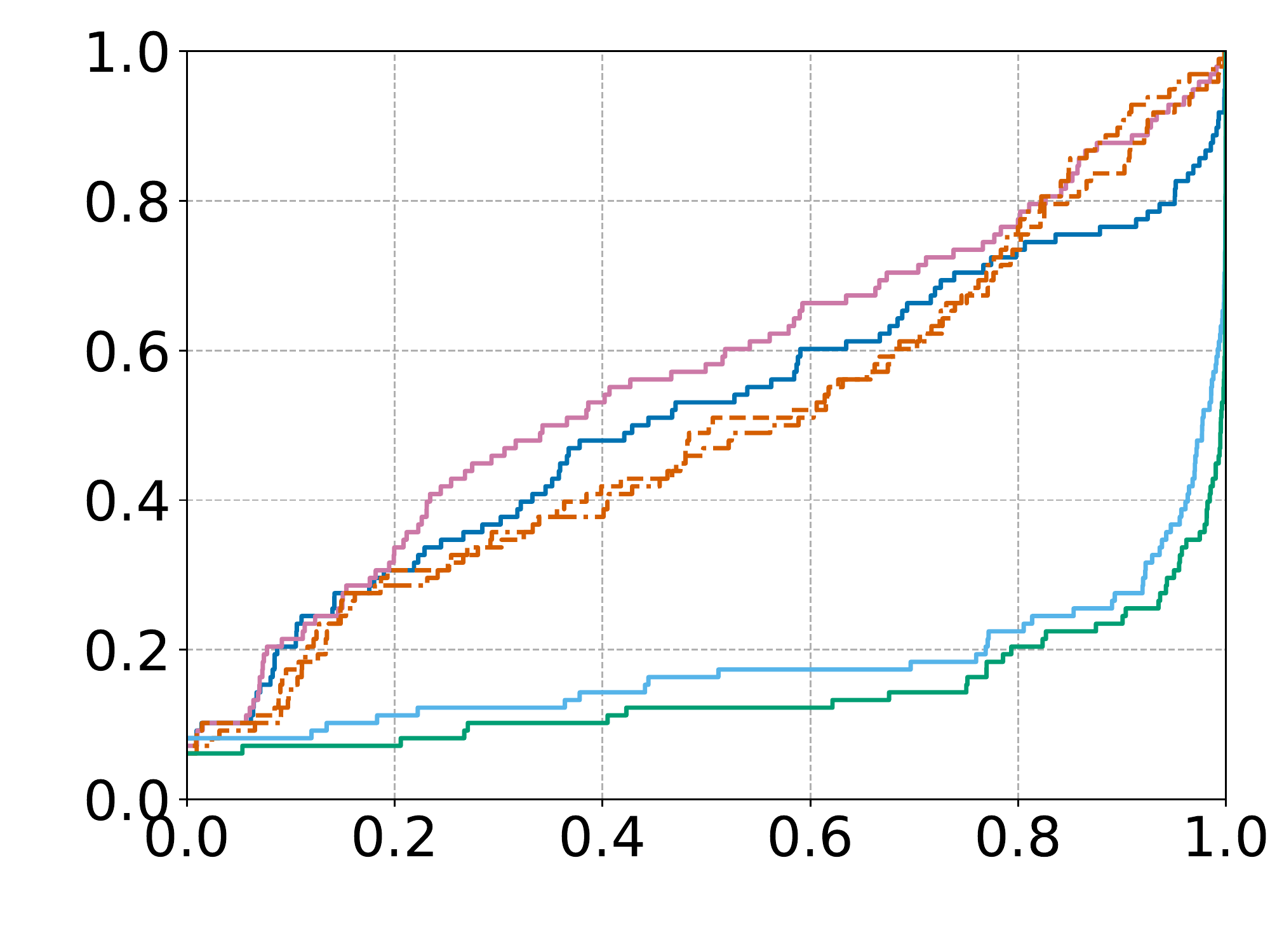}
&
\includegraphics[width=0.29\textwidth,valign=c,trim={0 0 0 0},clip]{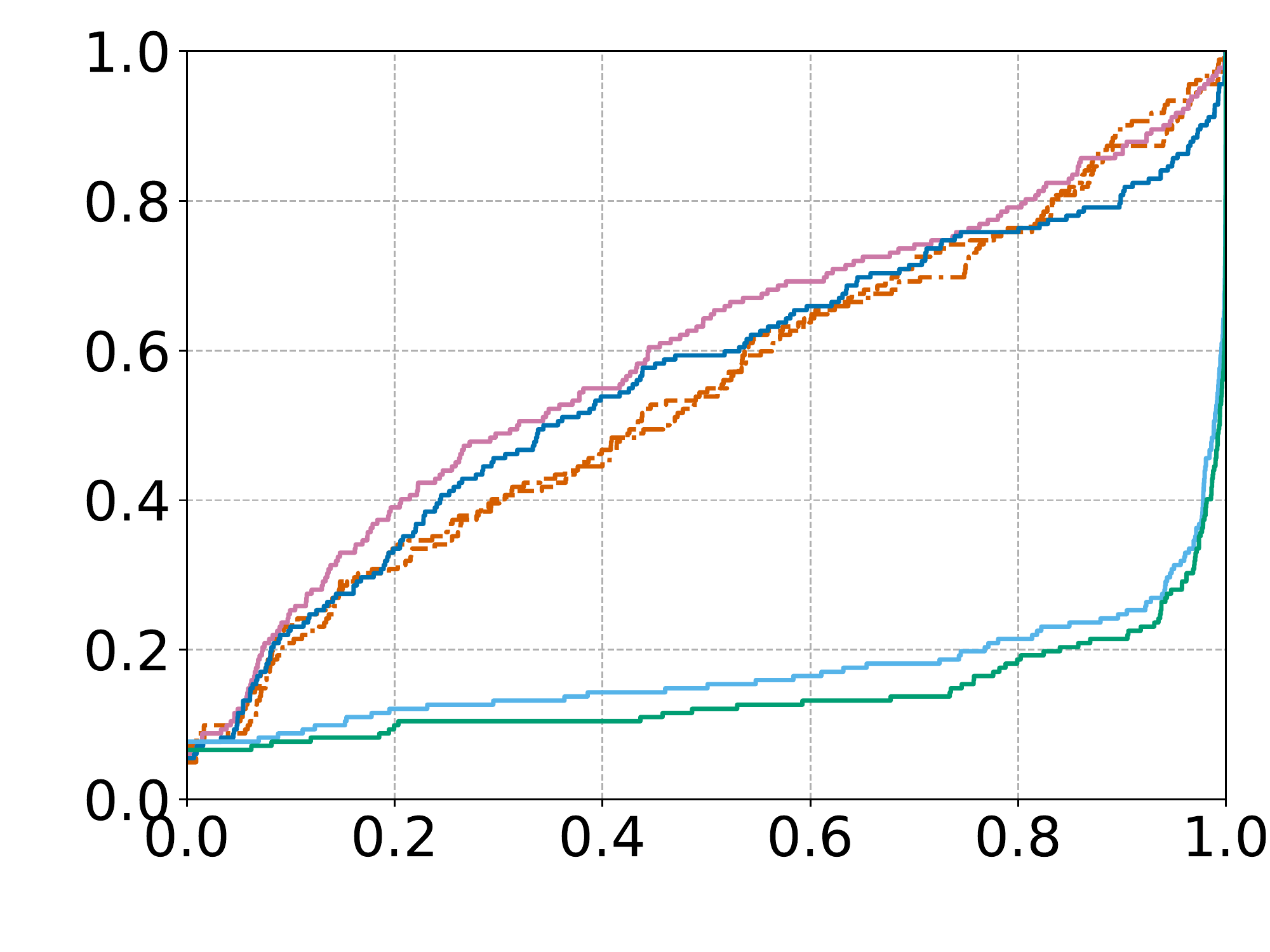}
&
\includegraphics[width=0.29\textwidth,valign=c,trim={0 0 0 0},clip]{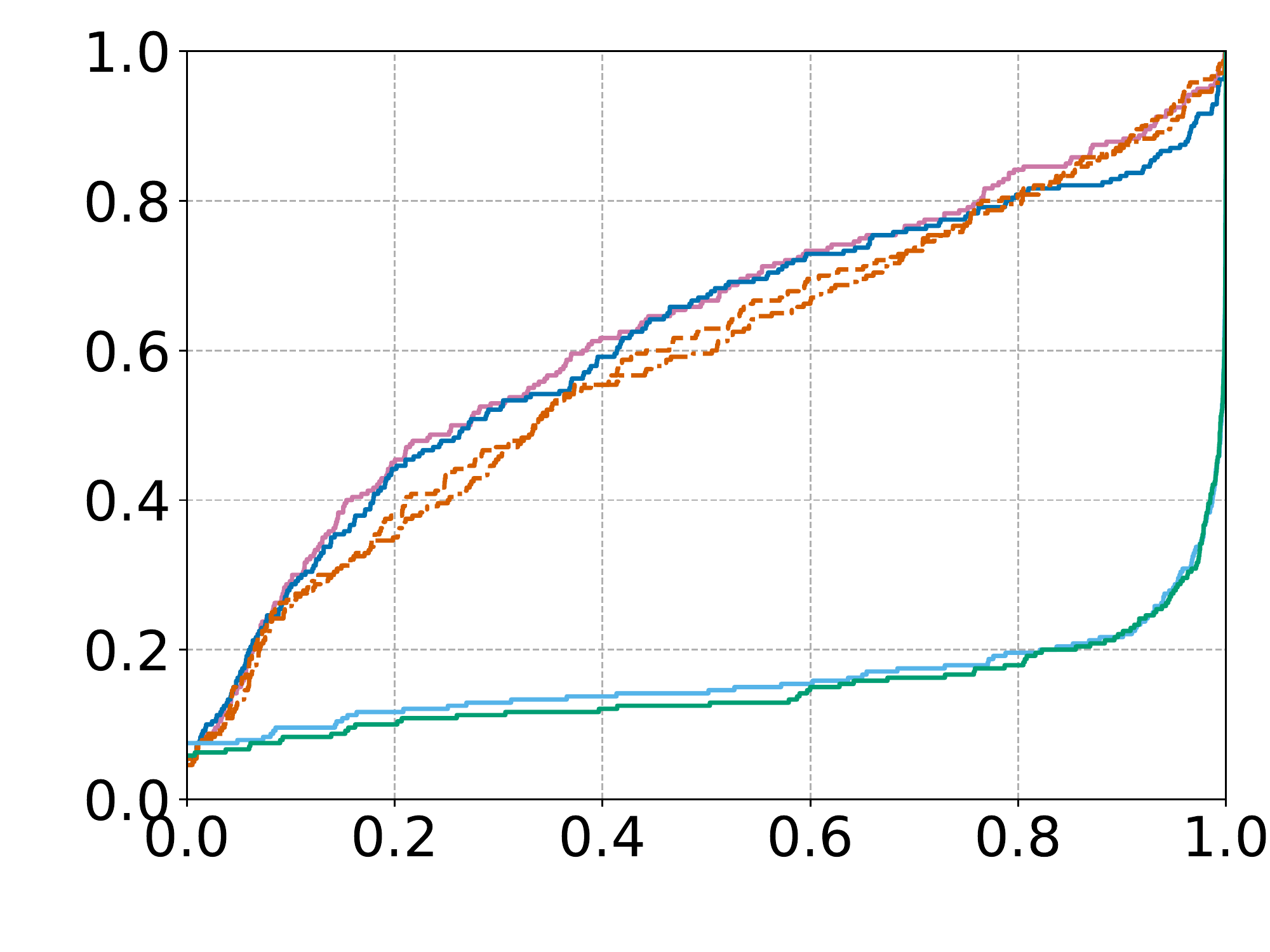}
\\\hline
\begin{tabular}{c}CYC \\ $p=1$\end{tabular}
& \includegraphics[width=0.29\textwidth,valign=c,trim={0 0 0 0},clip]{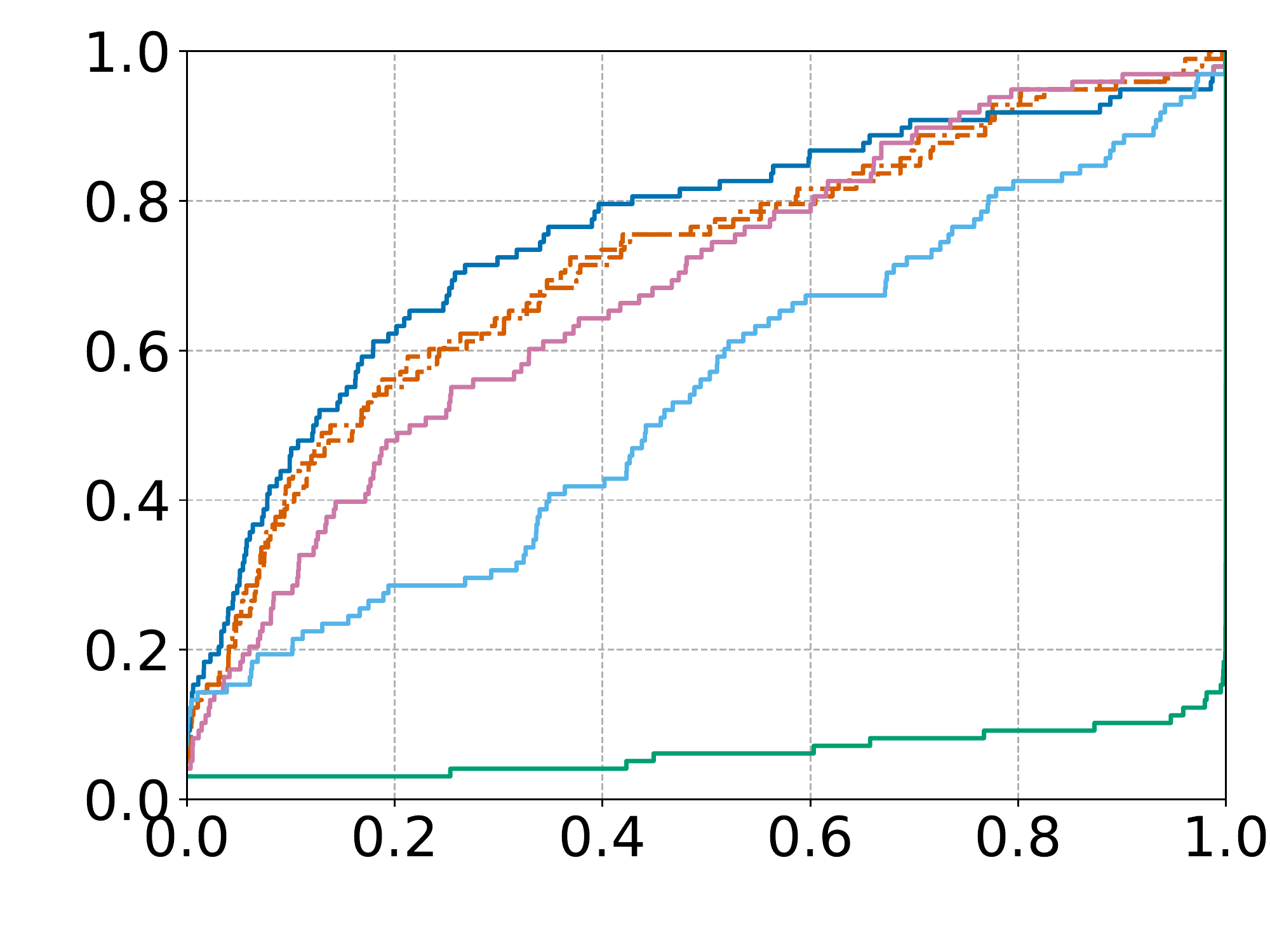}
&
\includegraphics[width=0.29\textwidth,valign=c,trim={0 0 0 0},clip]{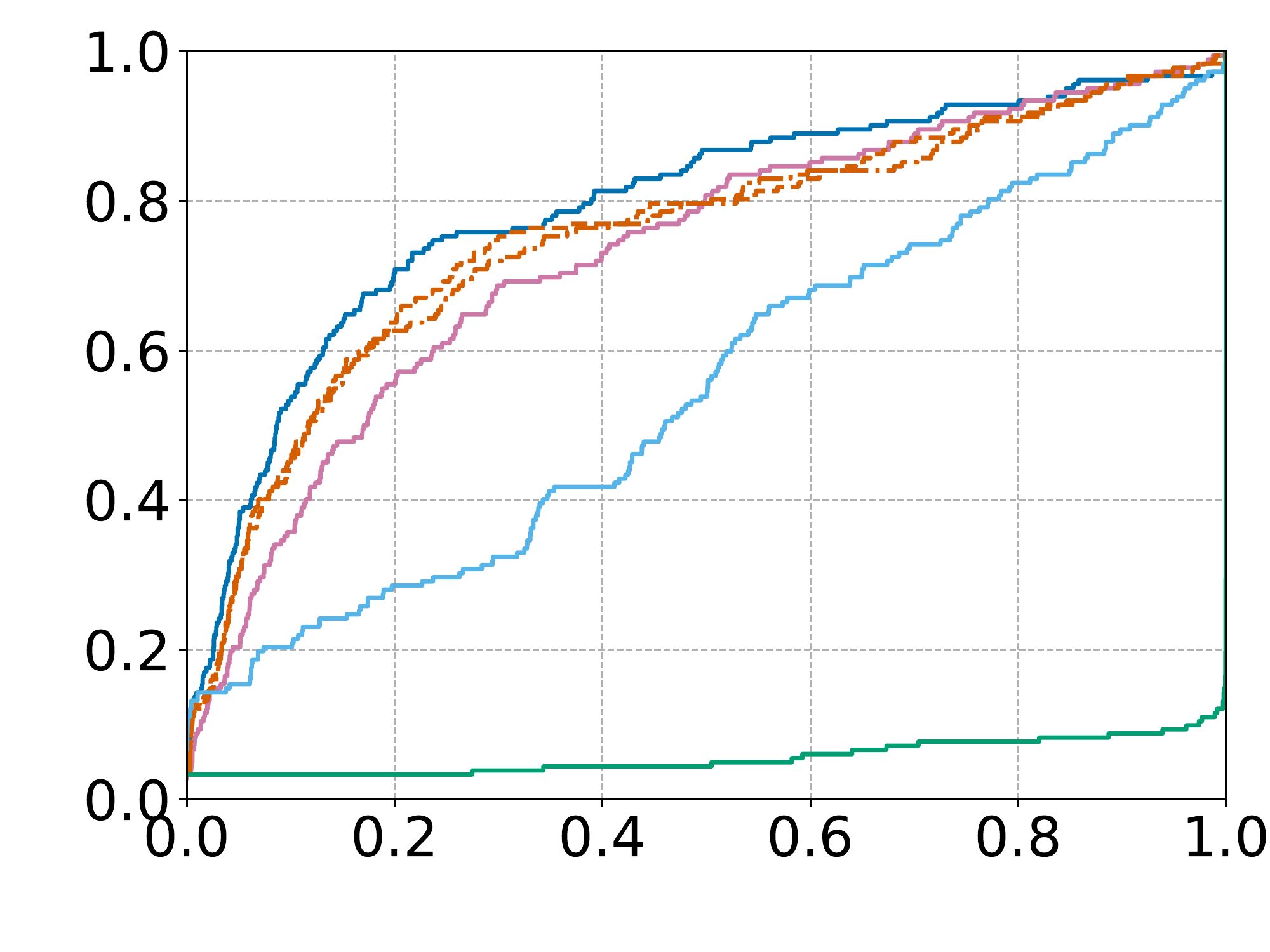}
&
\includegraphics[width=0.29\textwidth,valign=c,trim={0 0 0 0},clip]{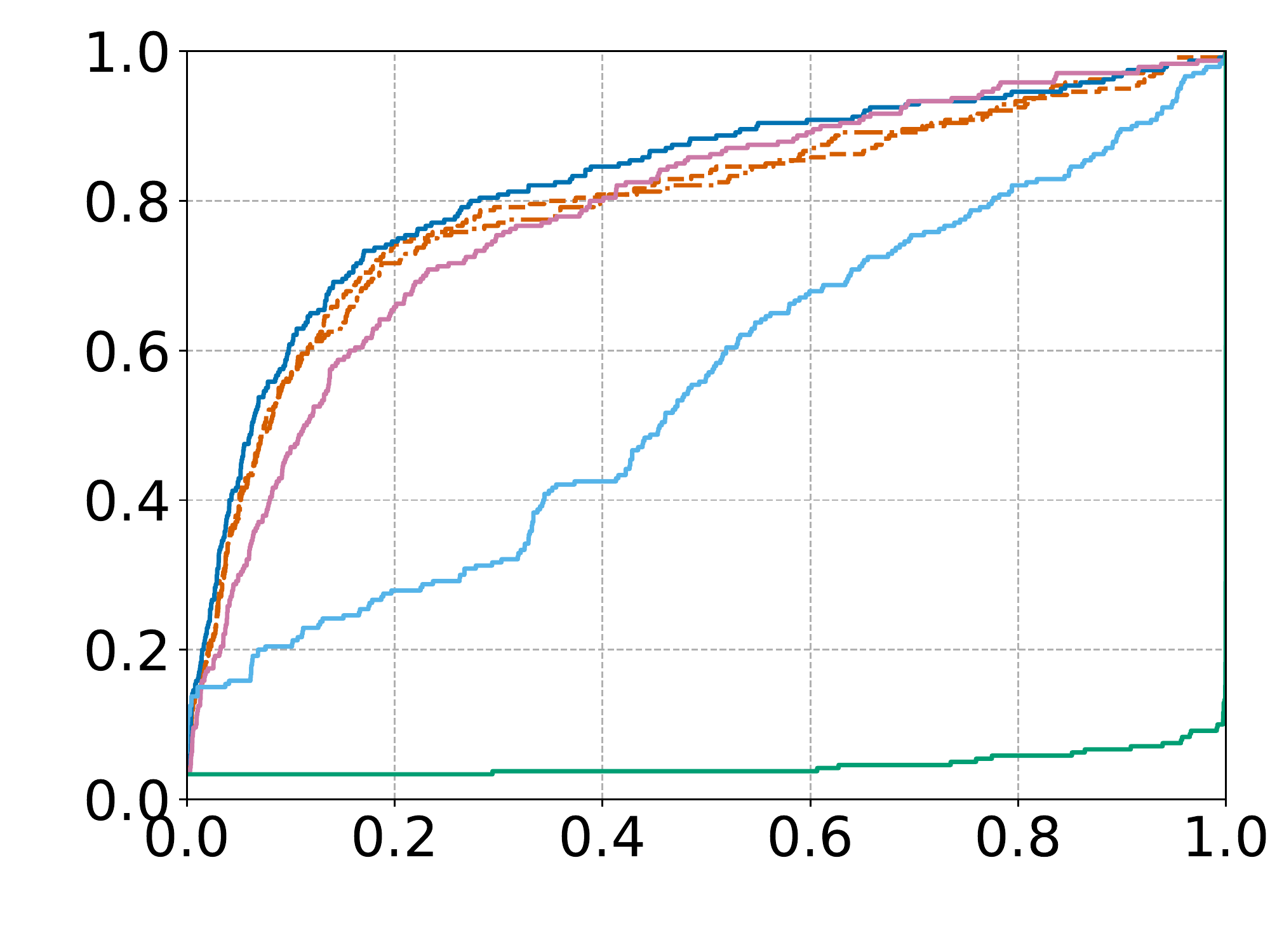}
\\\bottomrule
\multicolumn{4}{c}{
		\begin{subfigure}[t]{\textwidth}
    \includegraphics[width=\textwidth,trim={0 0 0 0},clip]{figs/cdfs/legends/legend_Lambda=8.pdf}
		\end{subfigure} }
\end{tabular}
\caption{Cumulative distribution functions (CDFs) for all evaluated algorithms, against different noise settings and warm start ratios in $\cbr{2.875,5.75,11.5}$. All CB algorithms use $\epsilon$-greedy with $\epsilon=0.00625$. In each of the above plots, the $x$ axis represents scores, while the $y$ axis represents the CDF values.}
\label{fig:cdfs-eps=0.00625-4}
\end{figure}

\begin{figure}[H]
\centering
\begin{tabular}{c | @{}c@{ }c@{ }c@{}} 
\toprule
& \multicolumn{3}{c}{ Ratio }
\\
Noise & 23.0 & 46.0 & 92.0
\\\midrule
\begin{tabular}{c}MAJ \\ $p=0.5$\end{tabular}
 & \includegraphics[width=0.29\textwidth,valign=c,trim={0 0 0 0},clip]{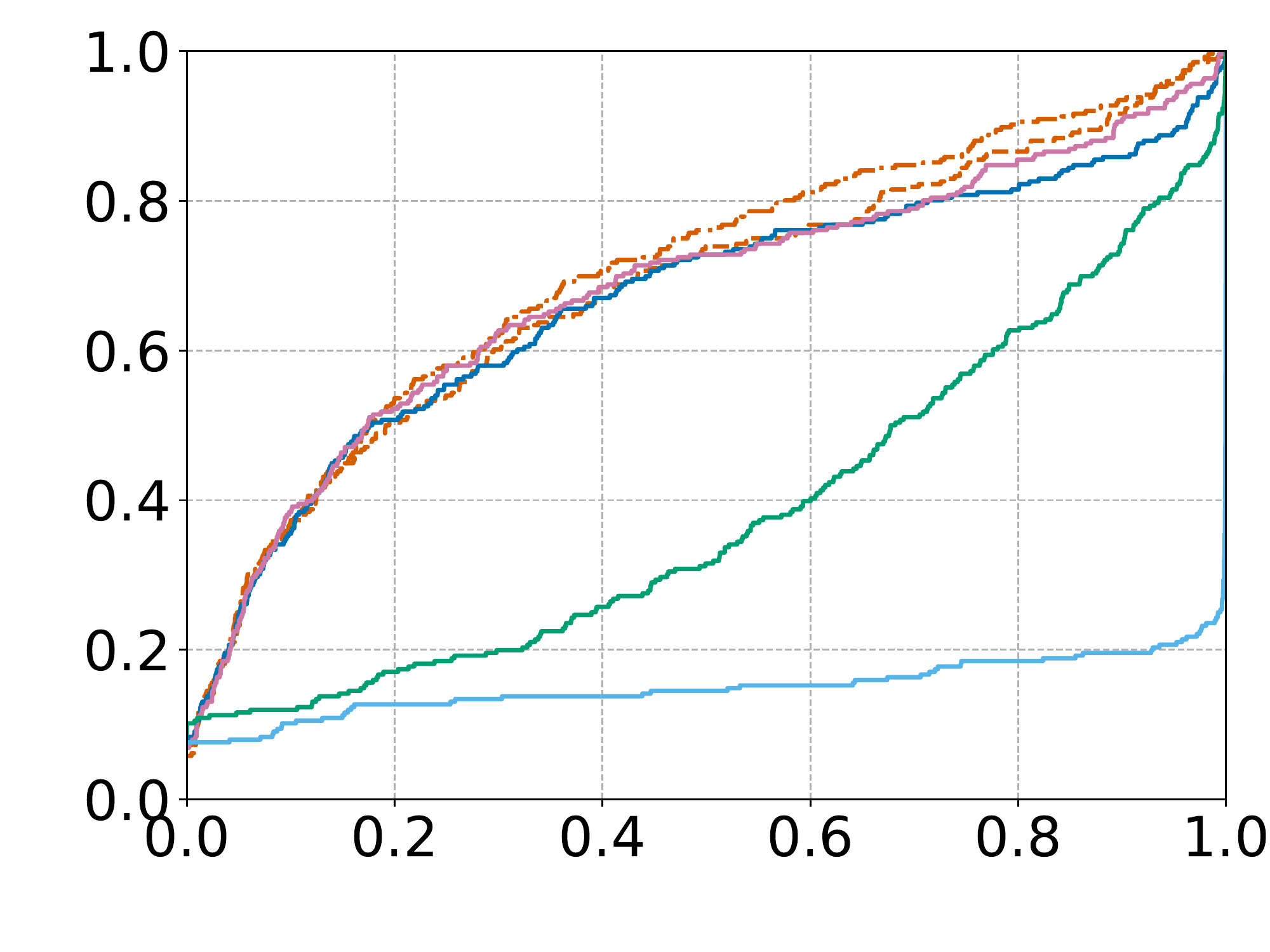}
&
\includegraphics[width=0.29\textwidth,valign=c,trim={0 0 0 0},clip]{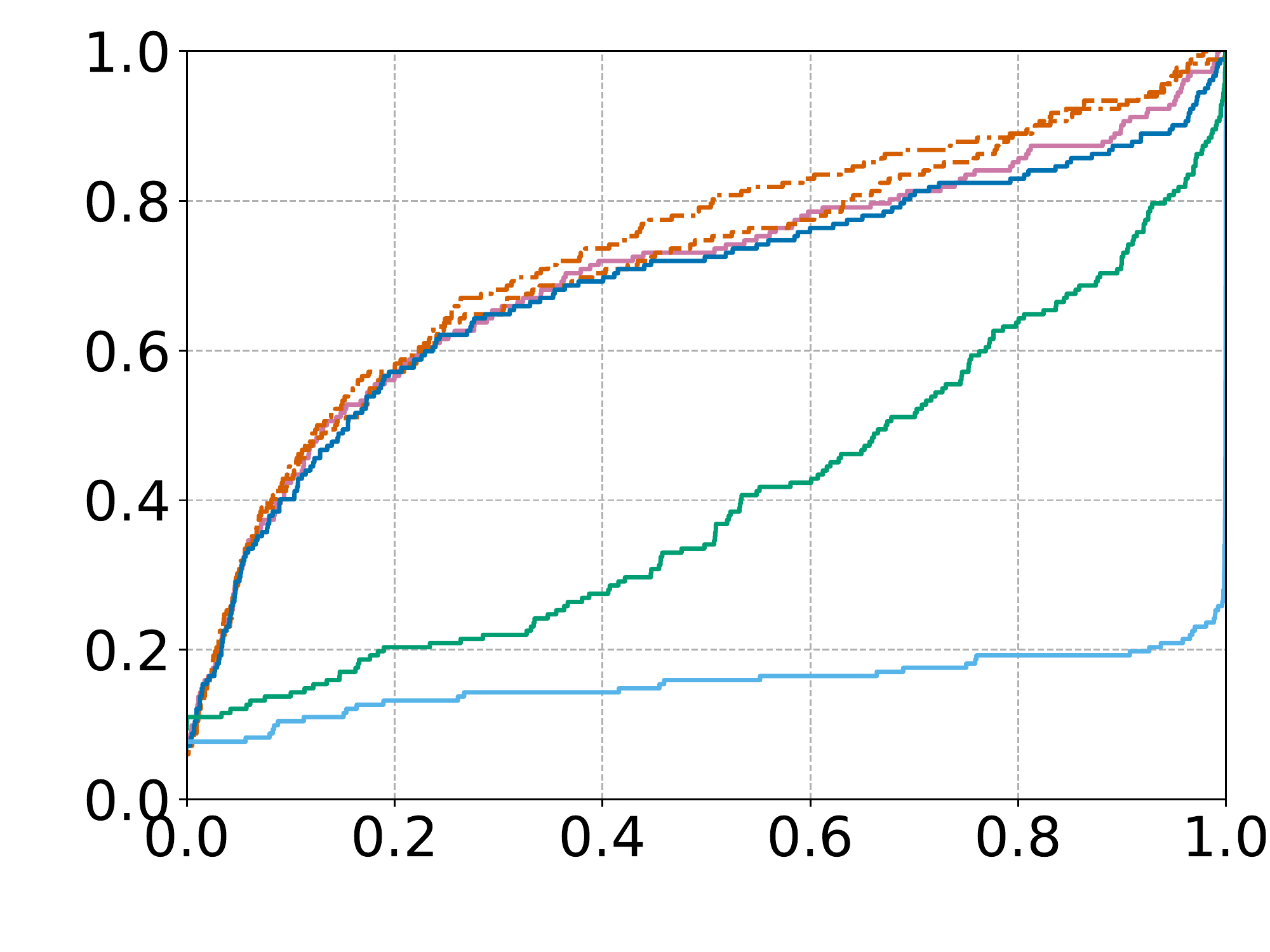}
&
\includegraphics[width=0.29\textwidth,valign=c,trim={0 0 0 0},clip]{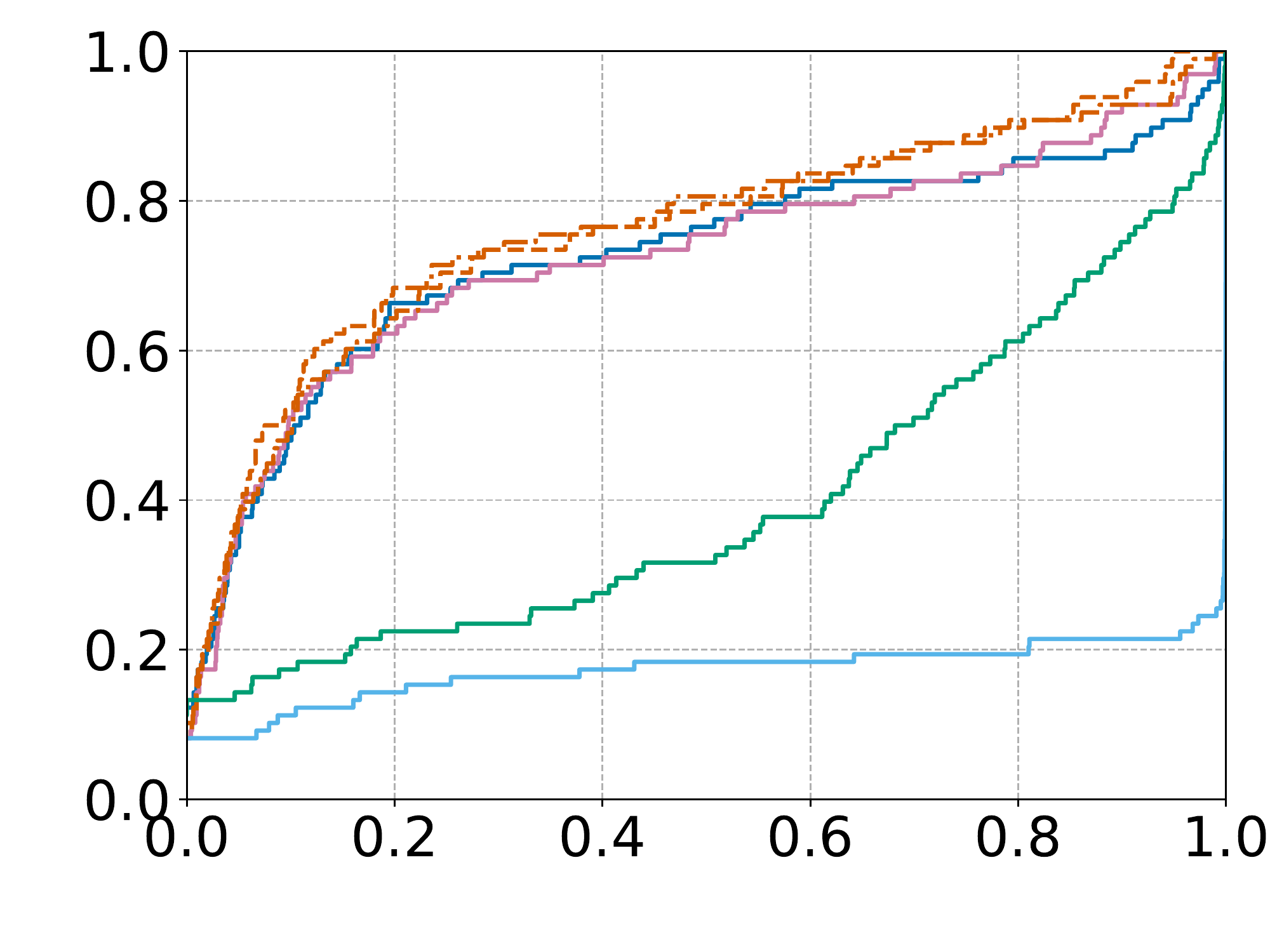}
\\\hline
\begin{tabular}{c}CYC \\ $p=0.5$\end{tabular}
& \includegraphics[width=0.29\textwidth,valign=c,trim={0 0 0 0},clip]{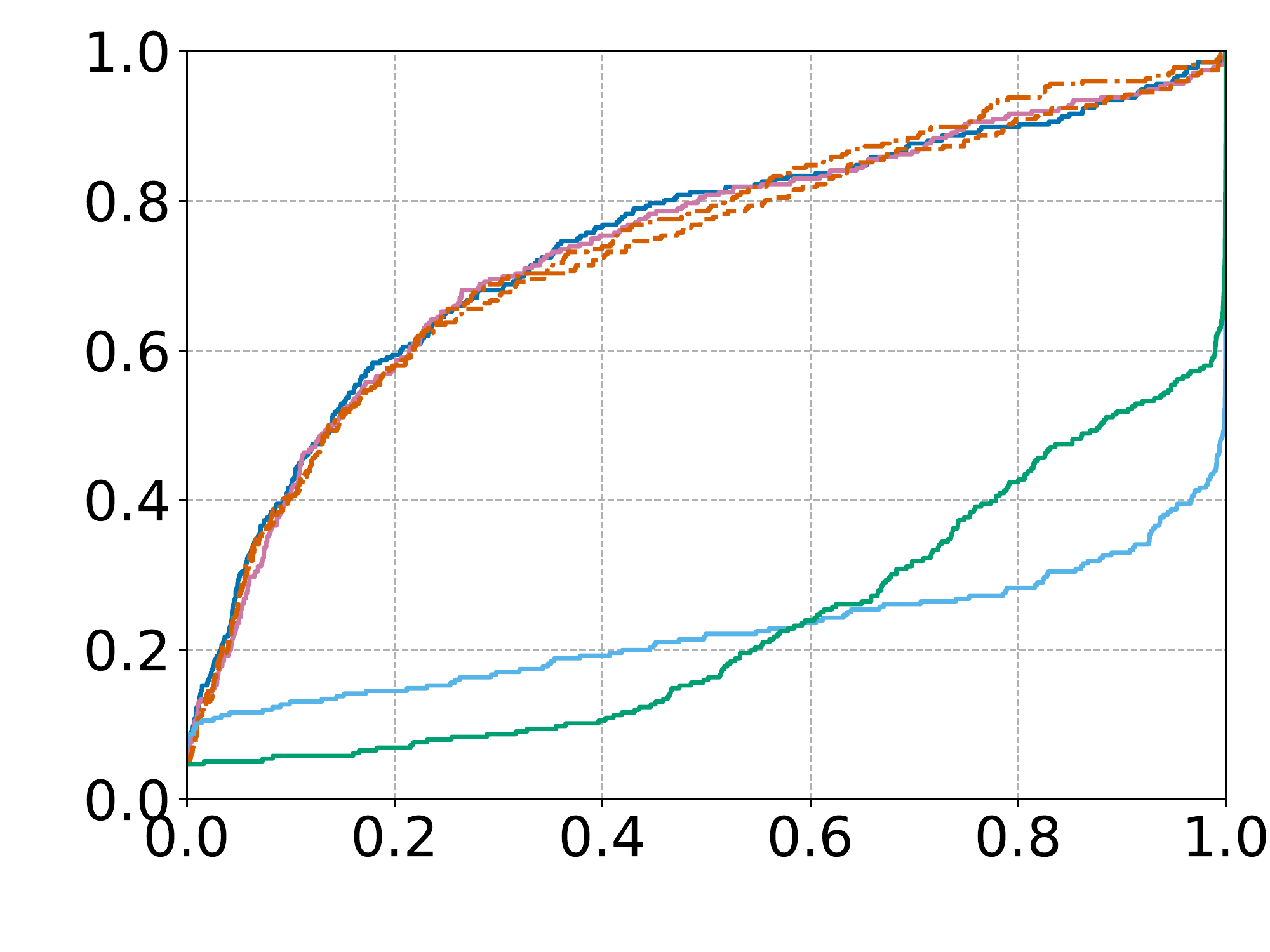}
&
\includegraphics[width=0.29\textwidth,valign=c,trim={0 0 0 0},clip]{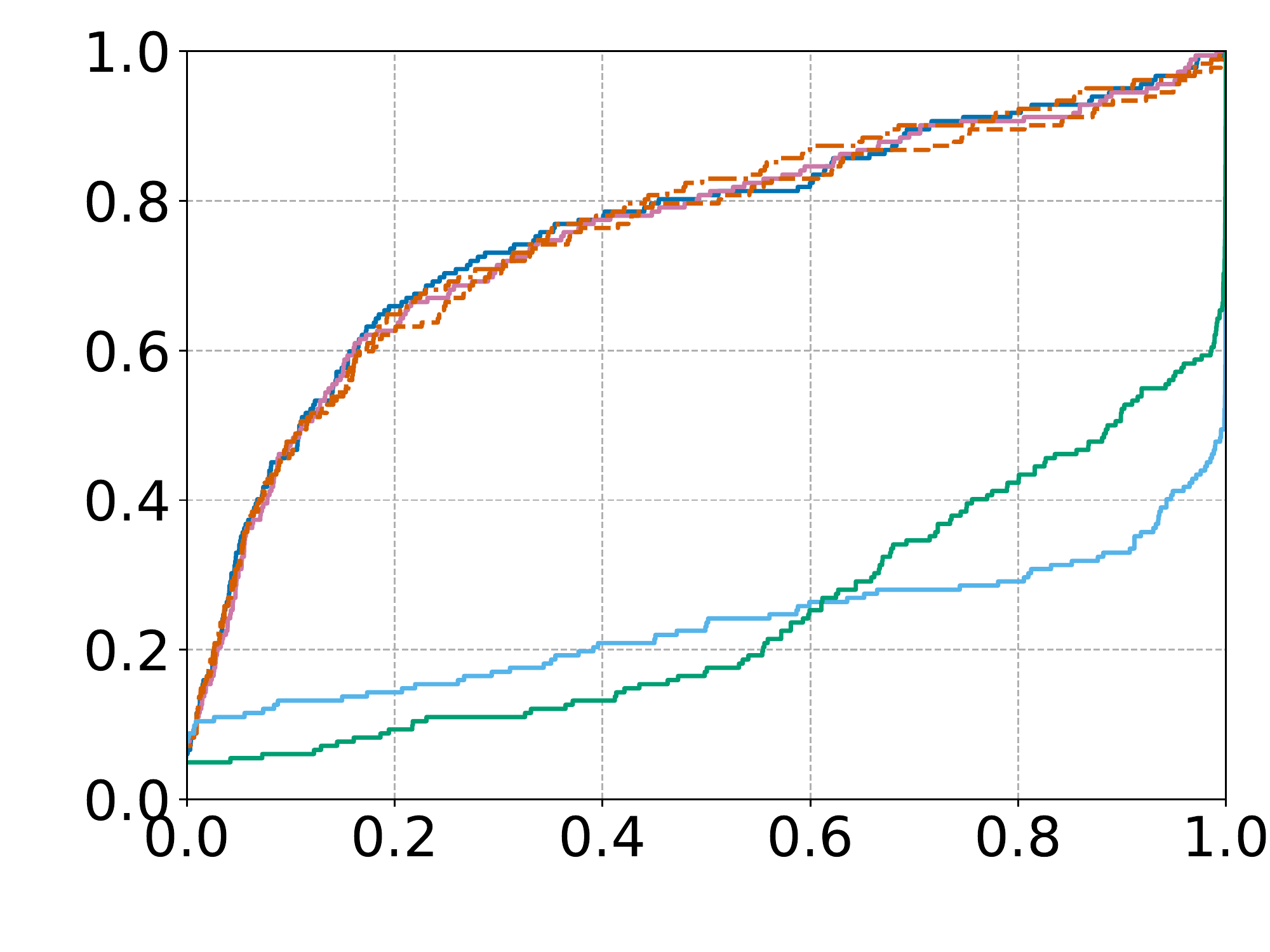}
&
\includegraphics[width=0.29\textwidth,valign=c,trim={0 0 0 0},clip]{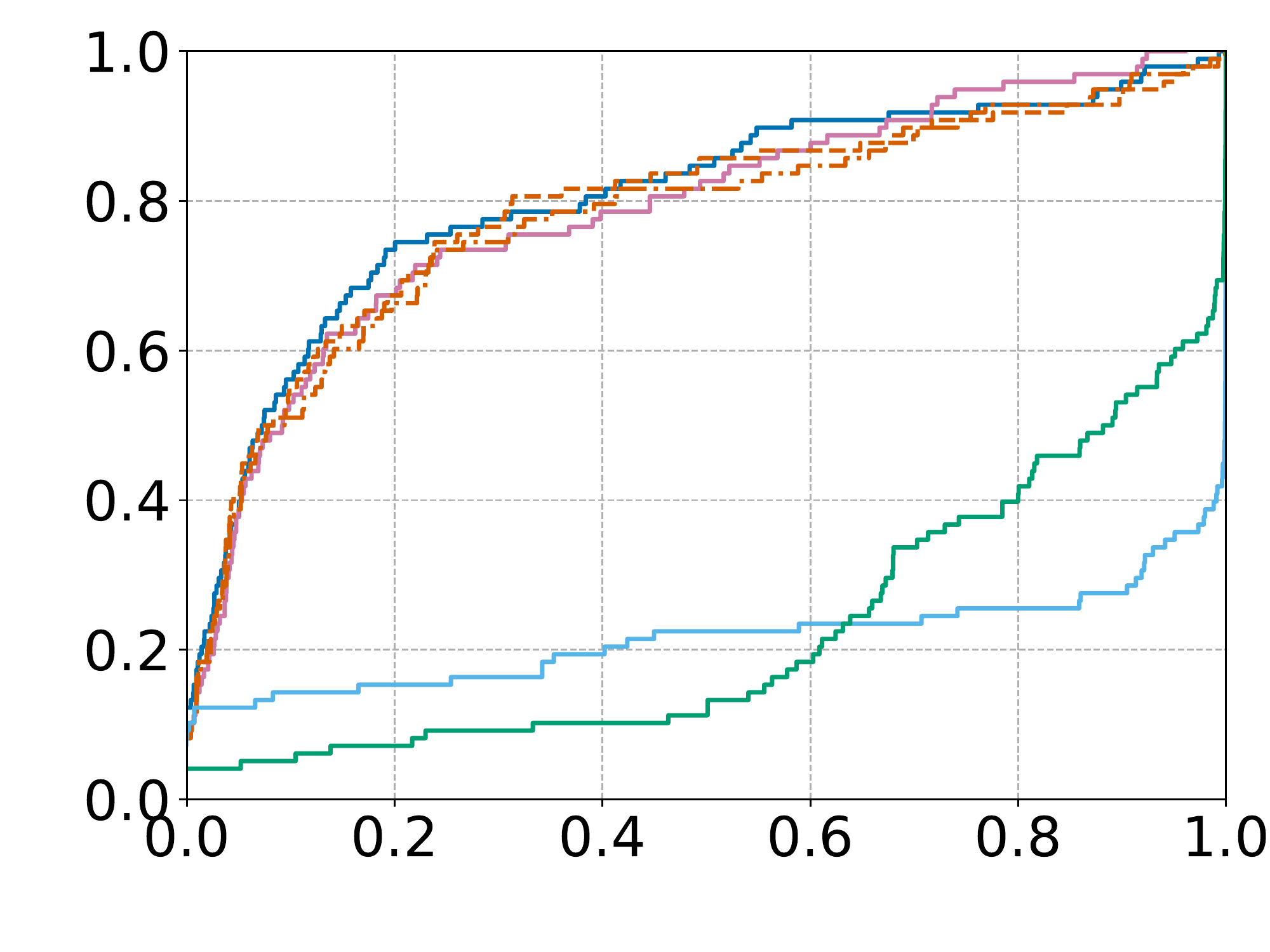}
\\\hline
\begin{tabular}{c}UAR \\ $p=1$\end{tabular}
& \includegraphics[width=0.29\textwidth,valign=c,trim={0 0 0 0},clip]{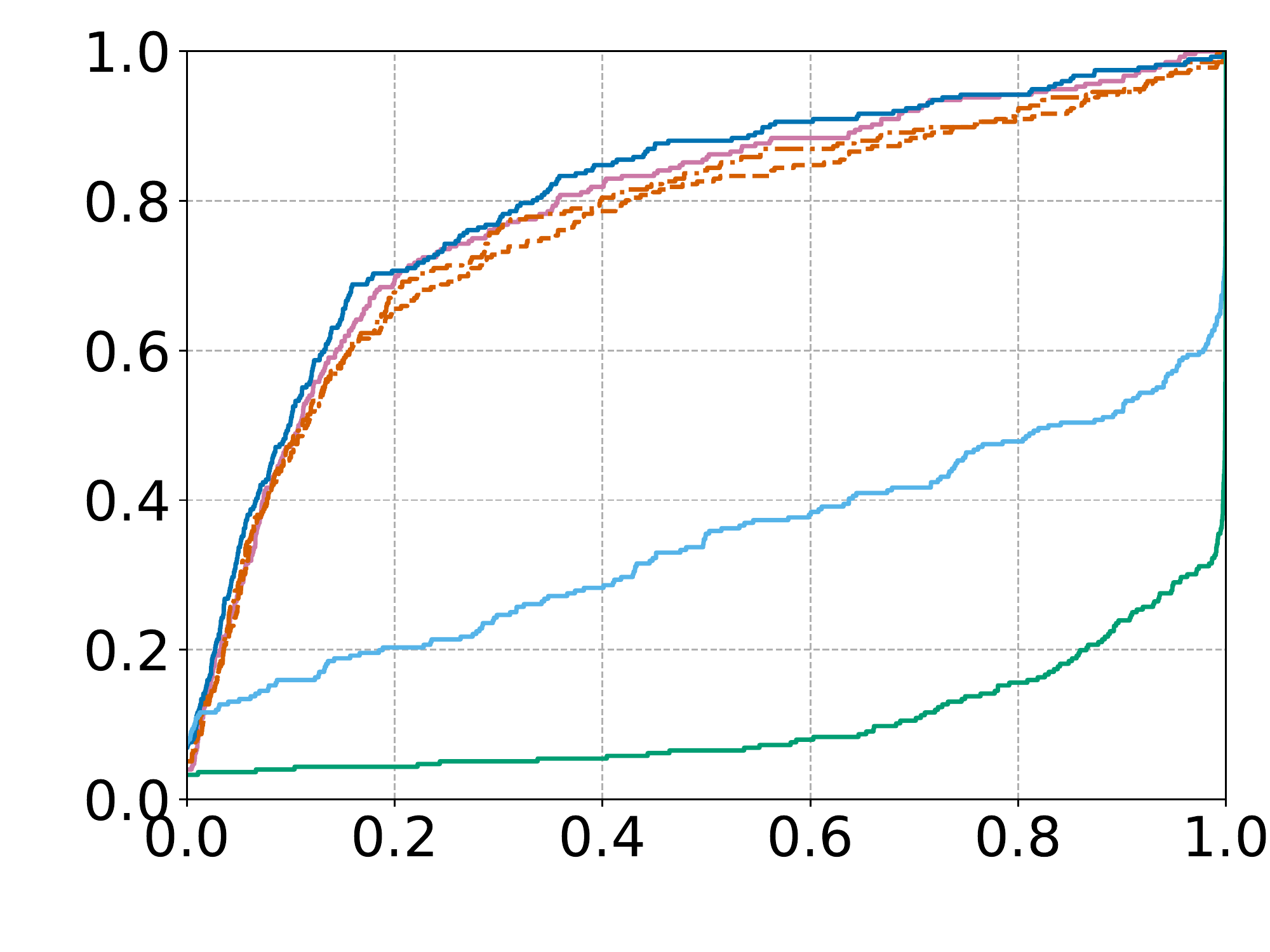}
&
\includegraphics[width=0.29\textwidth,valign=c,trim={0 0 0 0},clip]{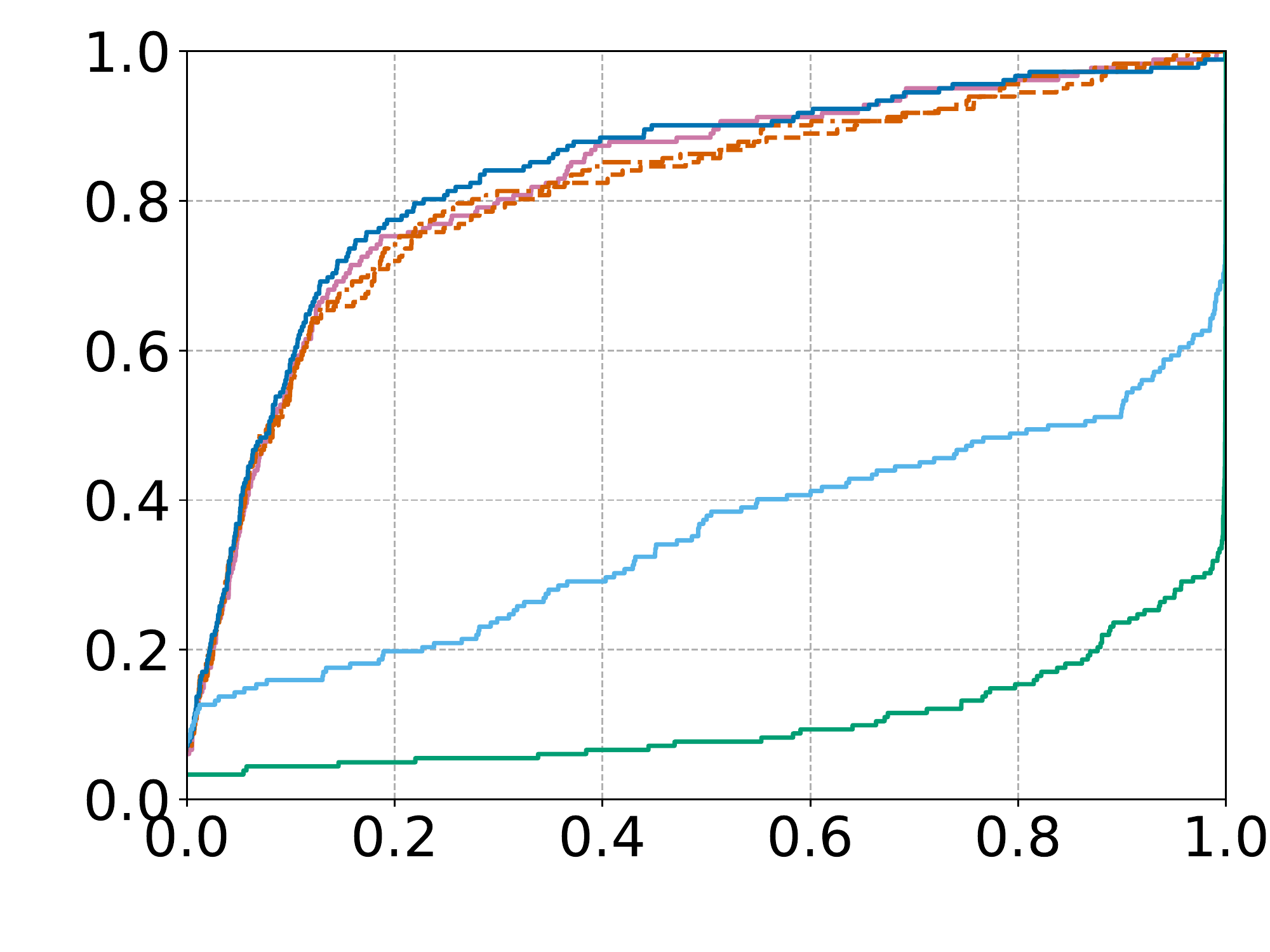}
&
\includegraphics[width=0.29\textwidth,valign=c,trim={0 0 0 0},clip]{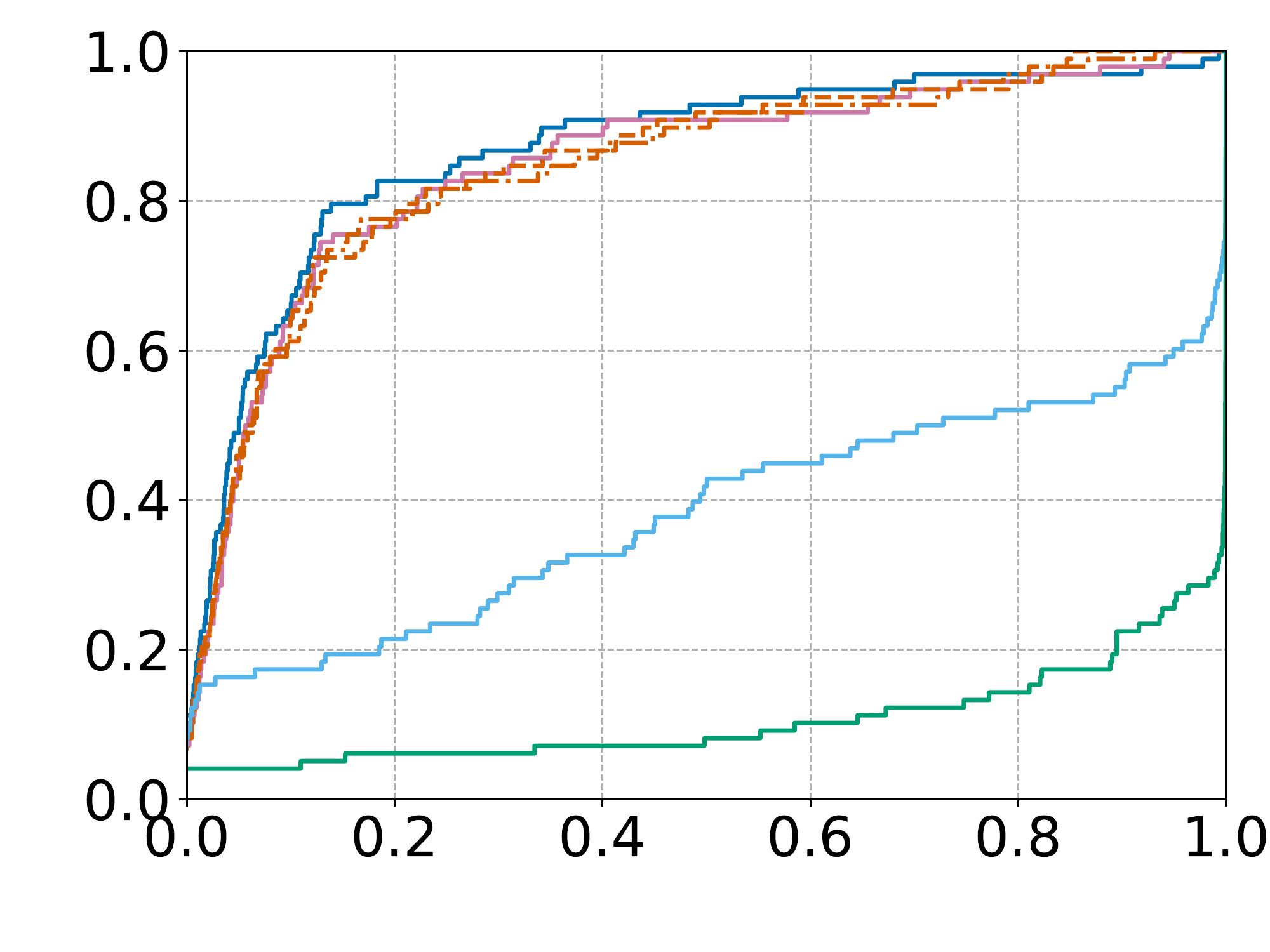}
\\\hline
\begin{tabular}{c}MAJ \\ $p=1$\end{tabular}
& \includegraphics[width=0.29\textwidth,valign=c,trim={0 0 0 0},clip]{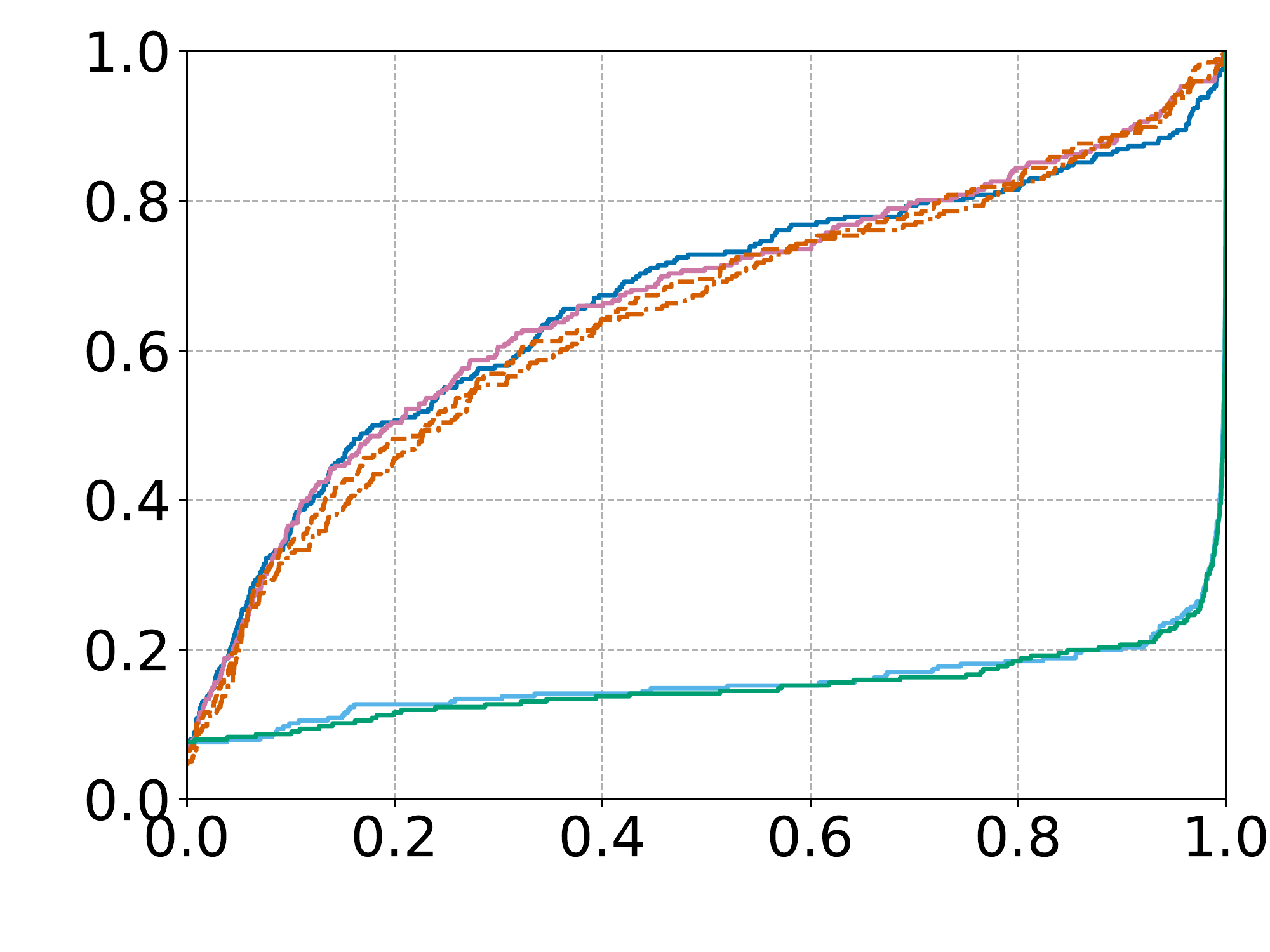}
&
\includegraphics[width=0.29\textwidth,valign=c,trim={0 0 0 0},clip]{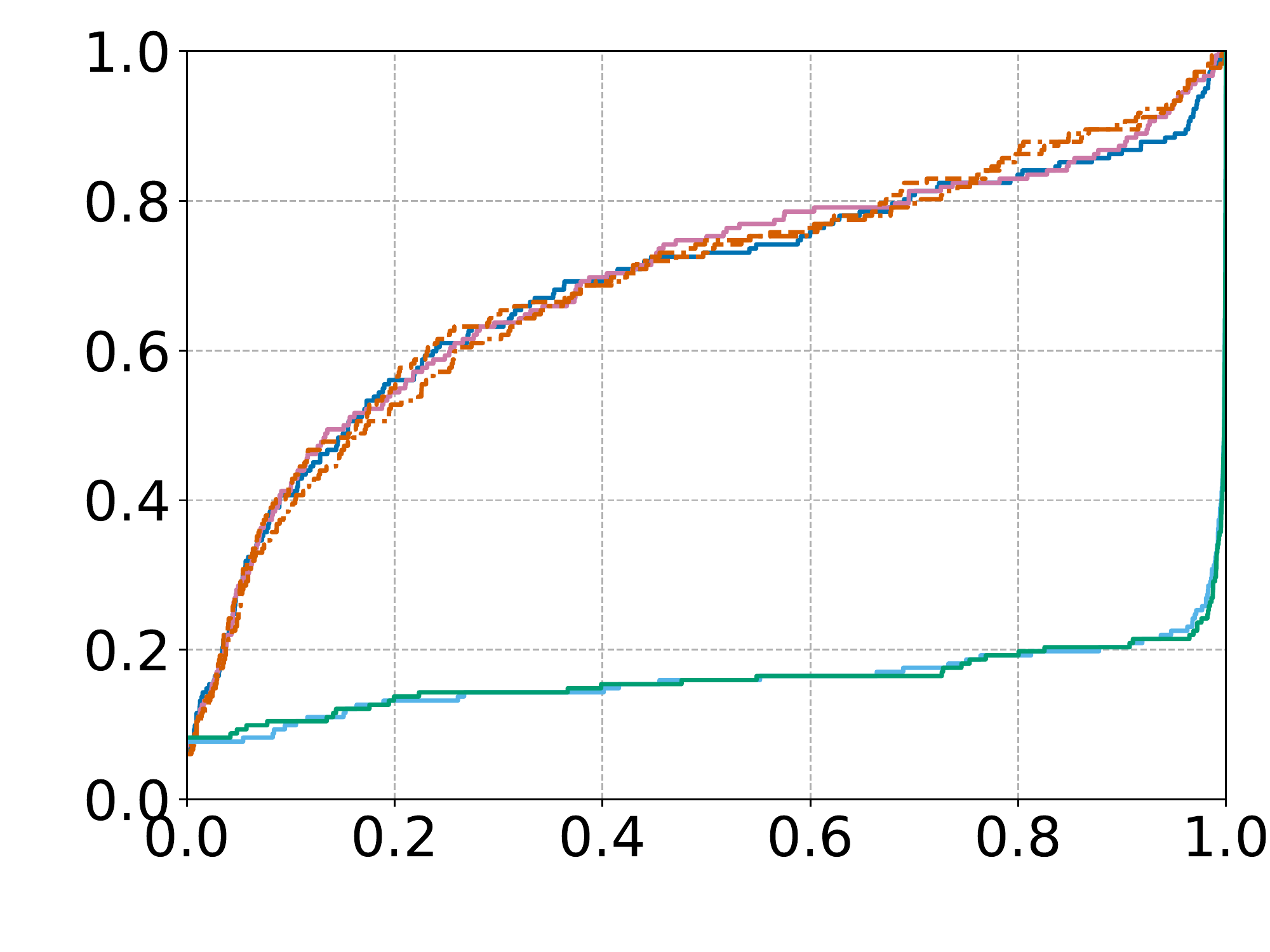}
&
\includegraphics[width=0.29\textwidth,valign=c,trim={0 0 0 0},clip]{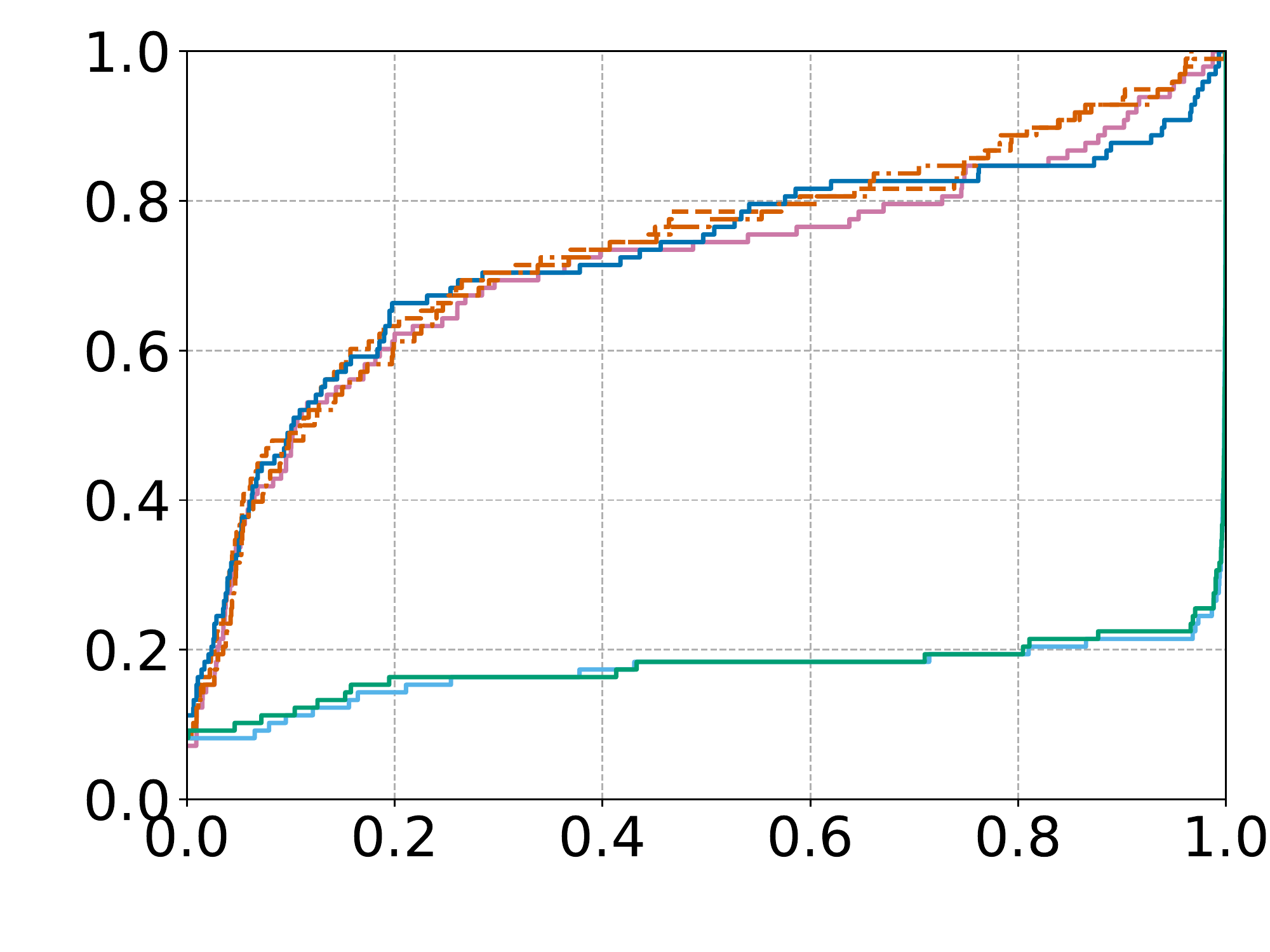}
\\\hline
\begin{tabular}{c}CYC \\ $p=1$\end{tabular}
& \includegraphics[width=0.29\textwidth,valign=c,trim={0 0 0 0},clip]{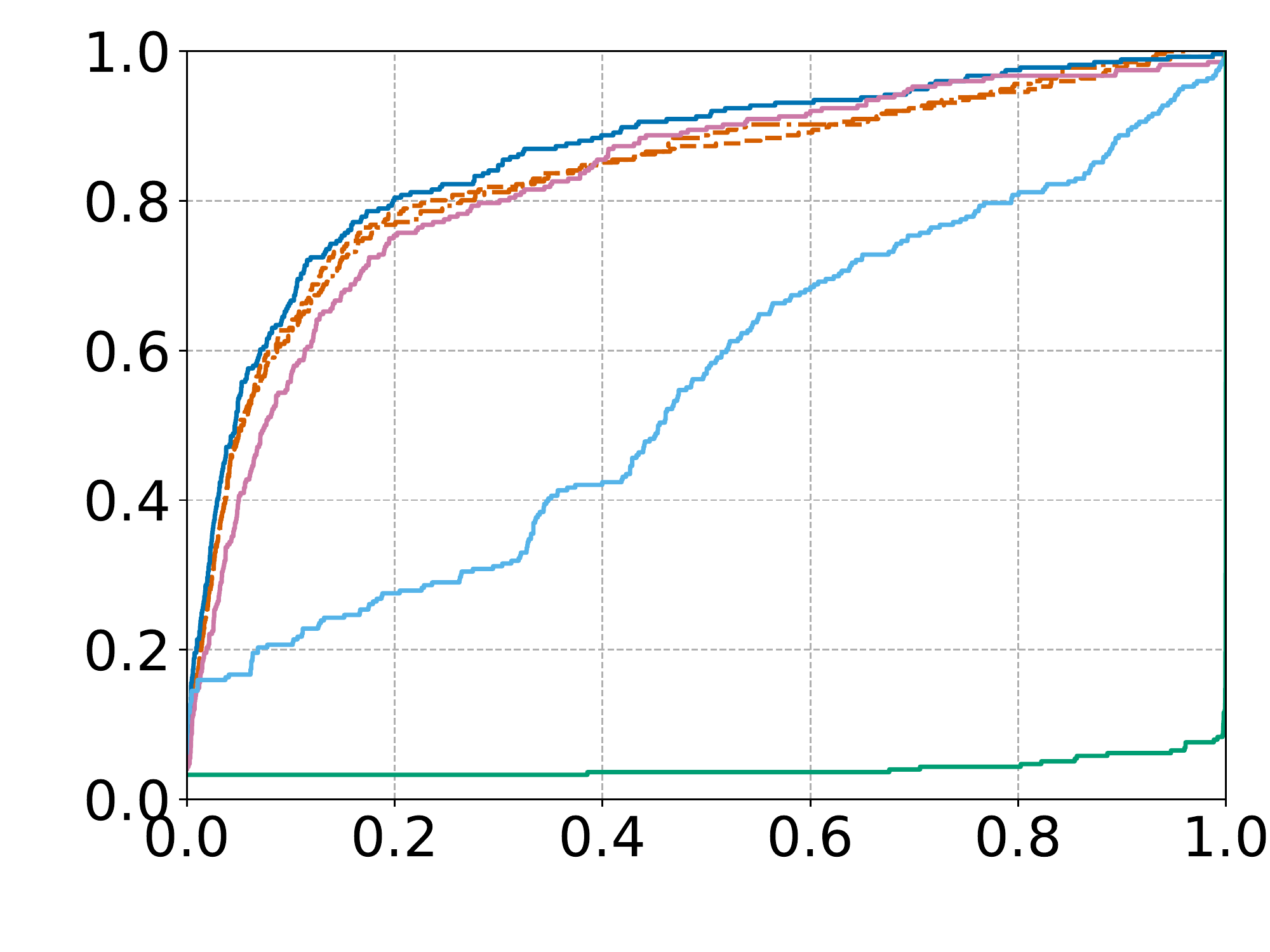}
&
\includegraphics[width=0.29\textwidth,valign=c,trim={0 0 0 0},clip]{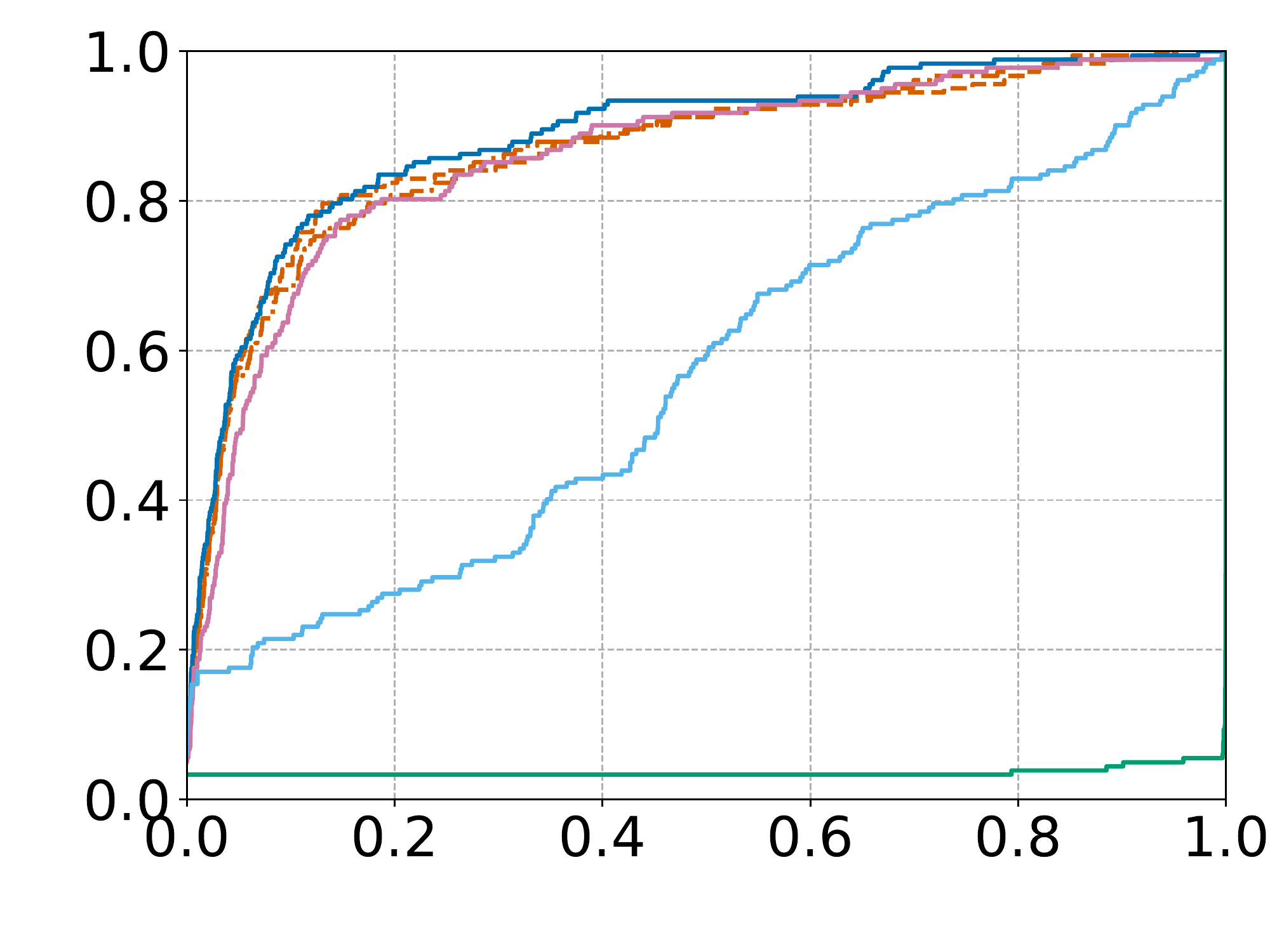}
&
\includegraphics[width=0.29\textwidth,valign=c,trim={0 0 0 0},clip]{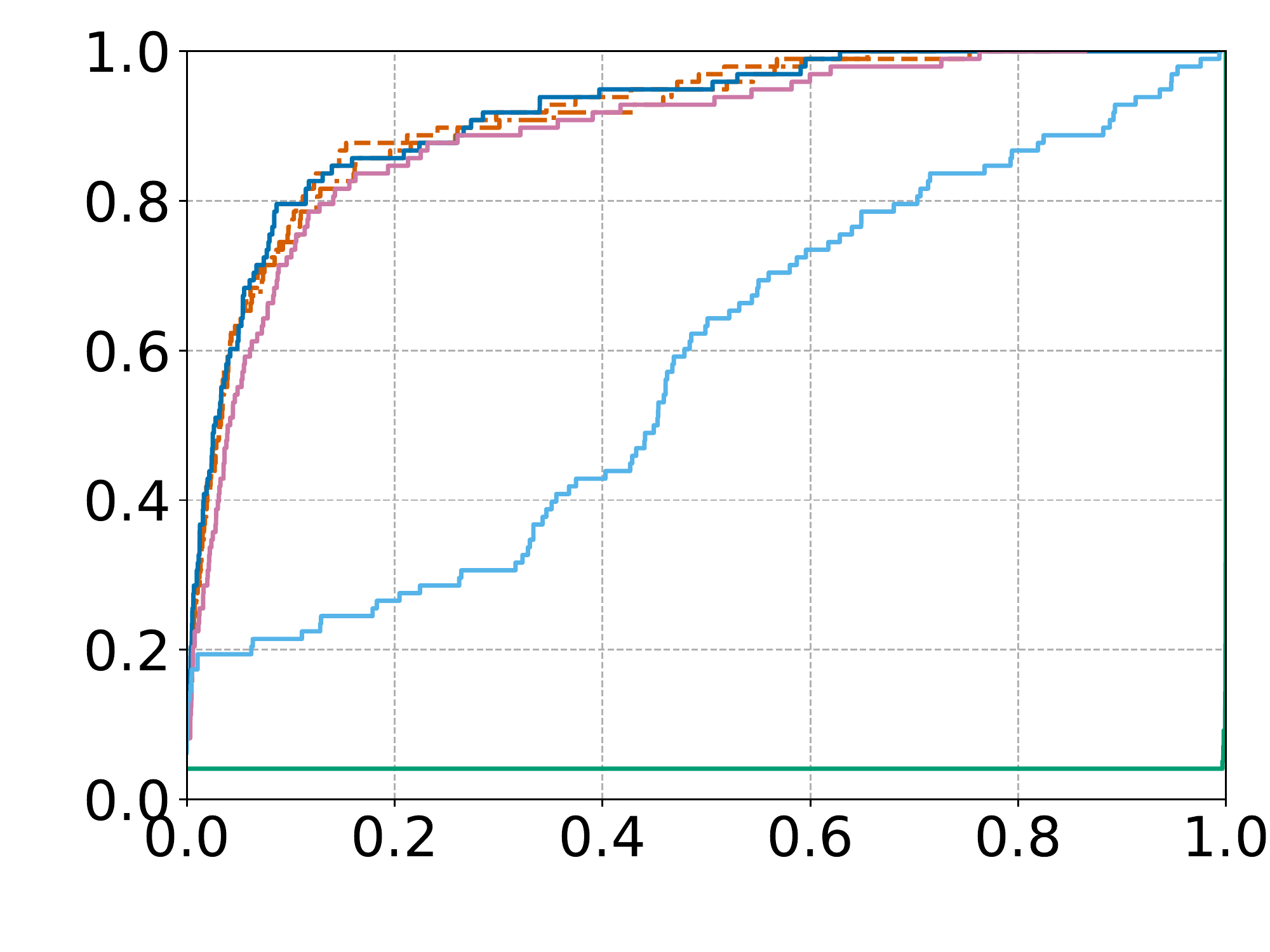}
\\\bottomrule
\multicolumn{4}{c}{
		\begin{subfigure}[t]{\textwidth}
    \includegraphics[width=\textwidth,trim={0 0 0 0},clip]{figs/cdfs/legends/legend_Lambda=8.pdf}
		\end{subfigure} }
\end{tabular}
\caption{Cumulative distribution functions (CDFs) for all evaluated algorithms, against different noise settings and warm start ratios in $\cbr{23.0,46.0,92.0}$. All CB algorithms use $\epsilon$-greedy with $\epsilon=0.00625$. In each of the above plots, the $x$ axis represents scores, while the $y$ axis represents the CDF values.}
\label{fig:cdfs-eps=0.00625-5}
\end{figure}

\begin{figure}[H]
\centering
\begin{tabular}{c | @{}c@{ }c@{ }c@{}} 
\toprule
& \multicolumn{3}{c}{ Ratio }
\\
Noise & 2.875 & 5.75 & 11.5
\\\midrule
Noiseless & \includegraphics[width=0.29\textwidth,valign=c,trim={0 0 0 0},clip]{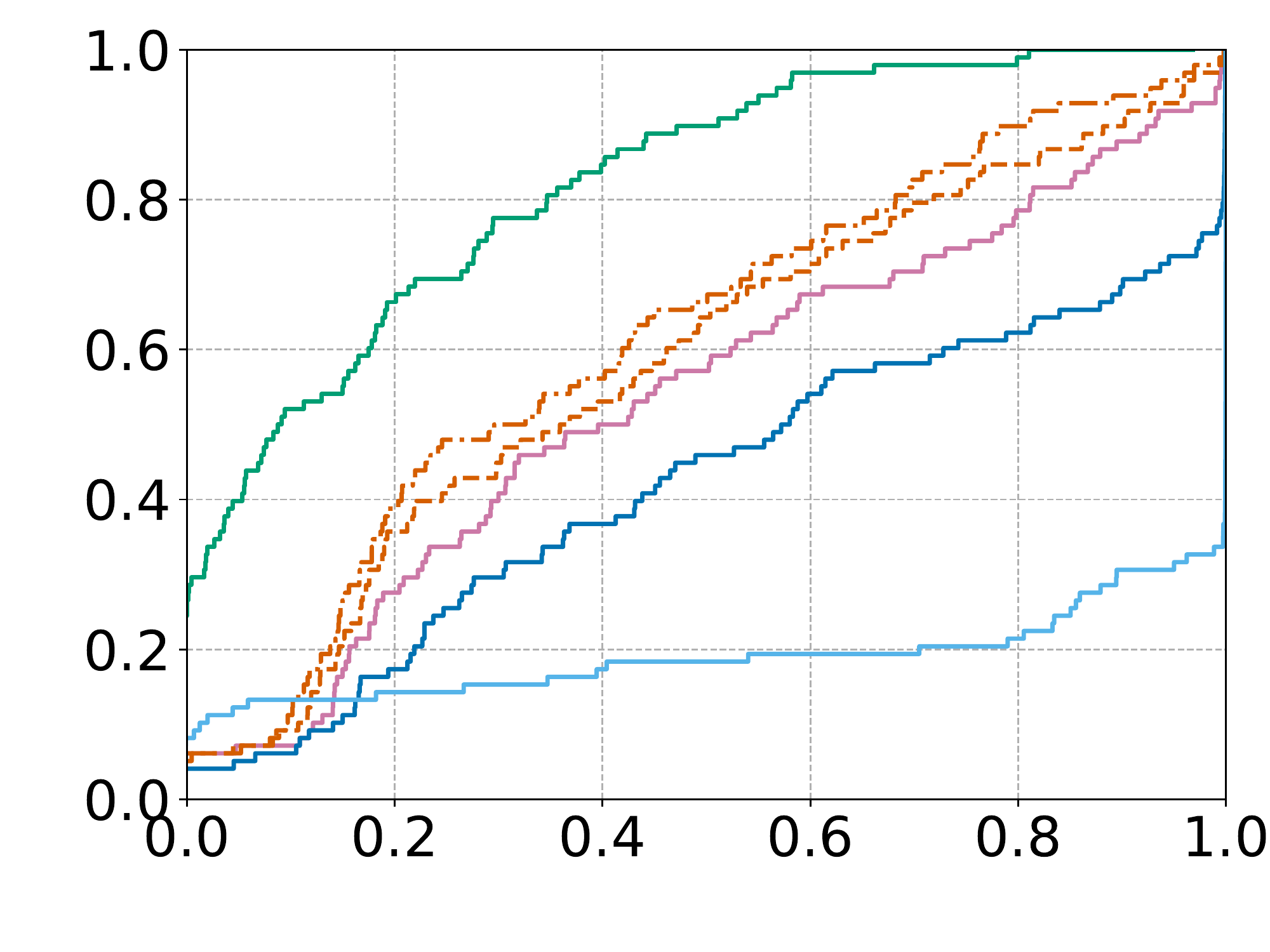}
&
\includegraphics[width=0.29\textwidth,valign=c,trim={0 0 0 0},clip]{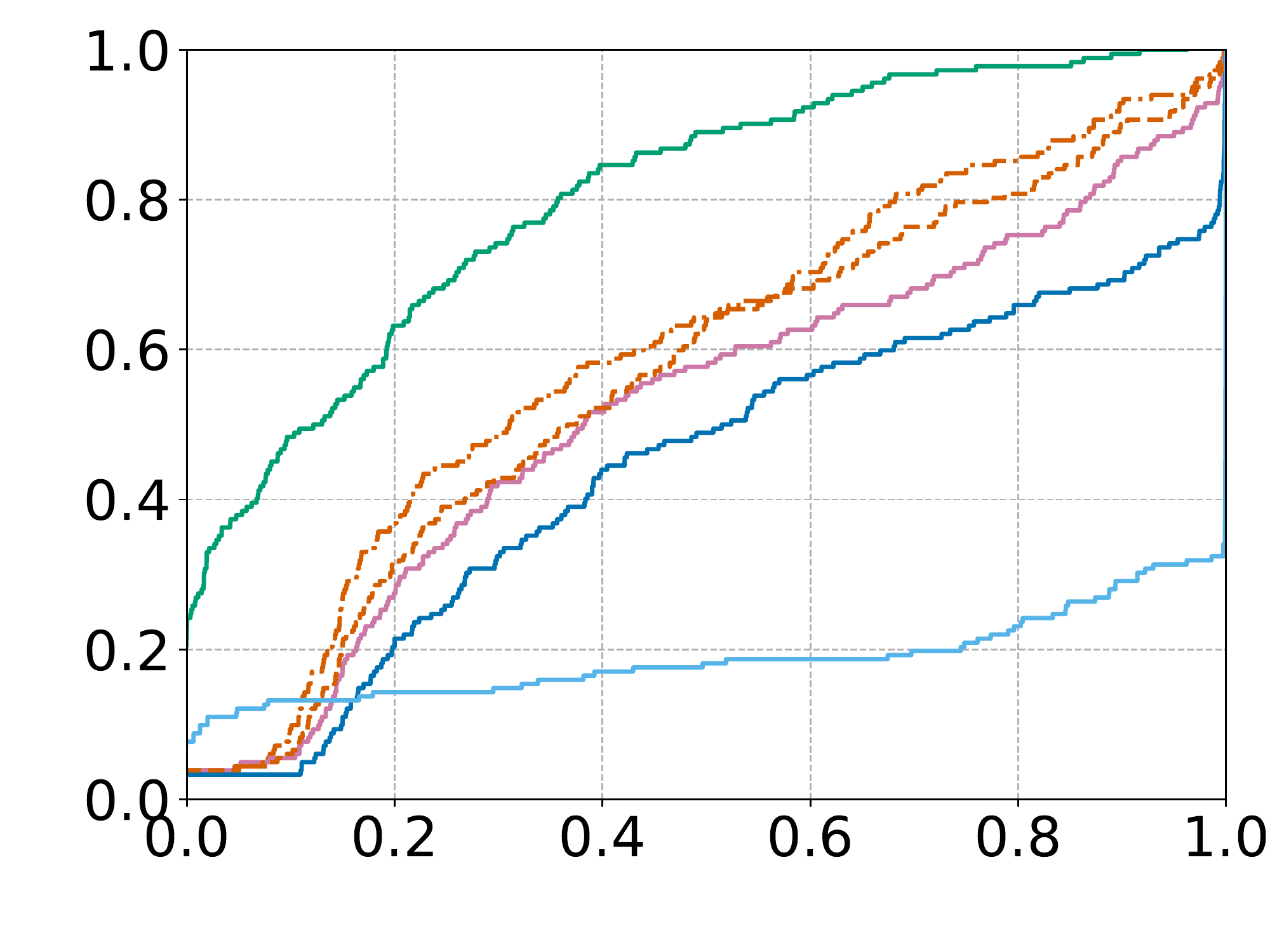}
&
\includegraphics[width=0.29\textwidth,valign=c,trim={0 0 0 0},clip]{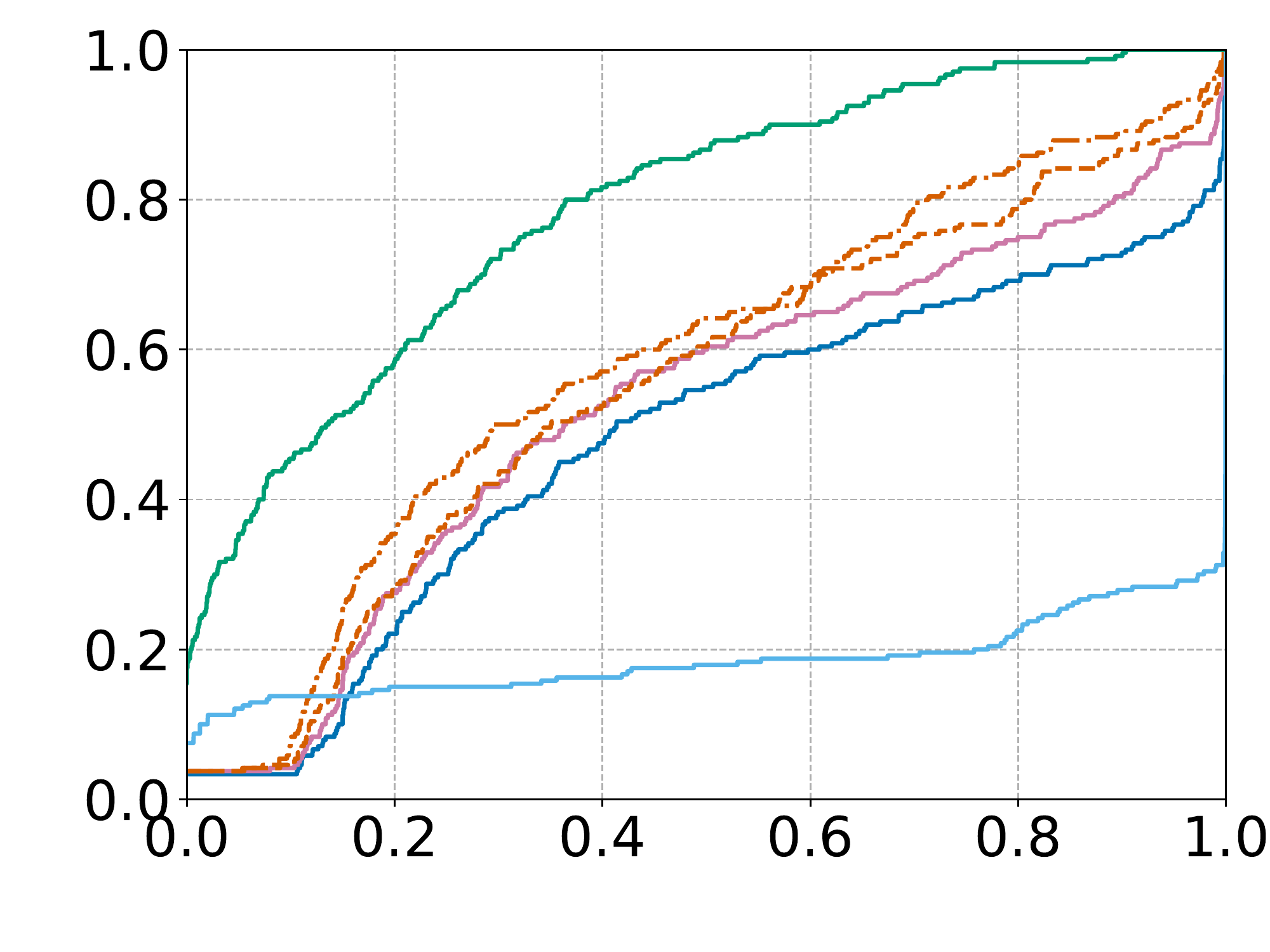}
\\\hline
\begin{tabular}{c}UAR \\ $p=0.25$\end{tabular}
& \includegraphics[width=0.29\textwidth,valign=c,trim={0 0 0 0},clip]{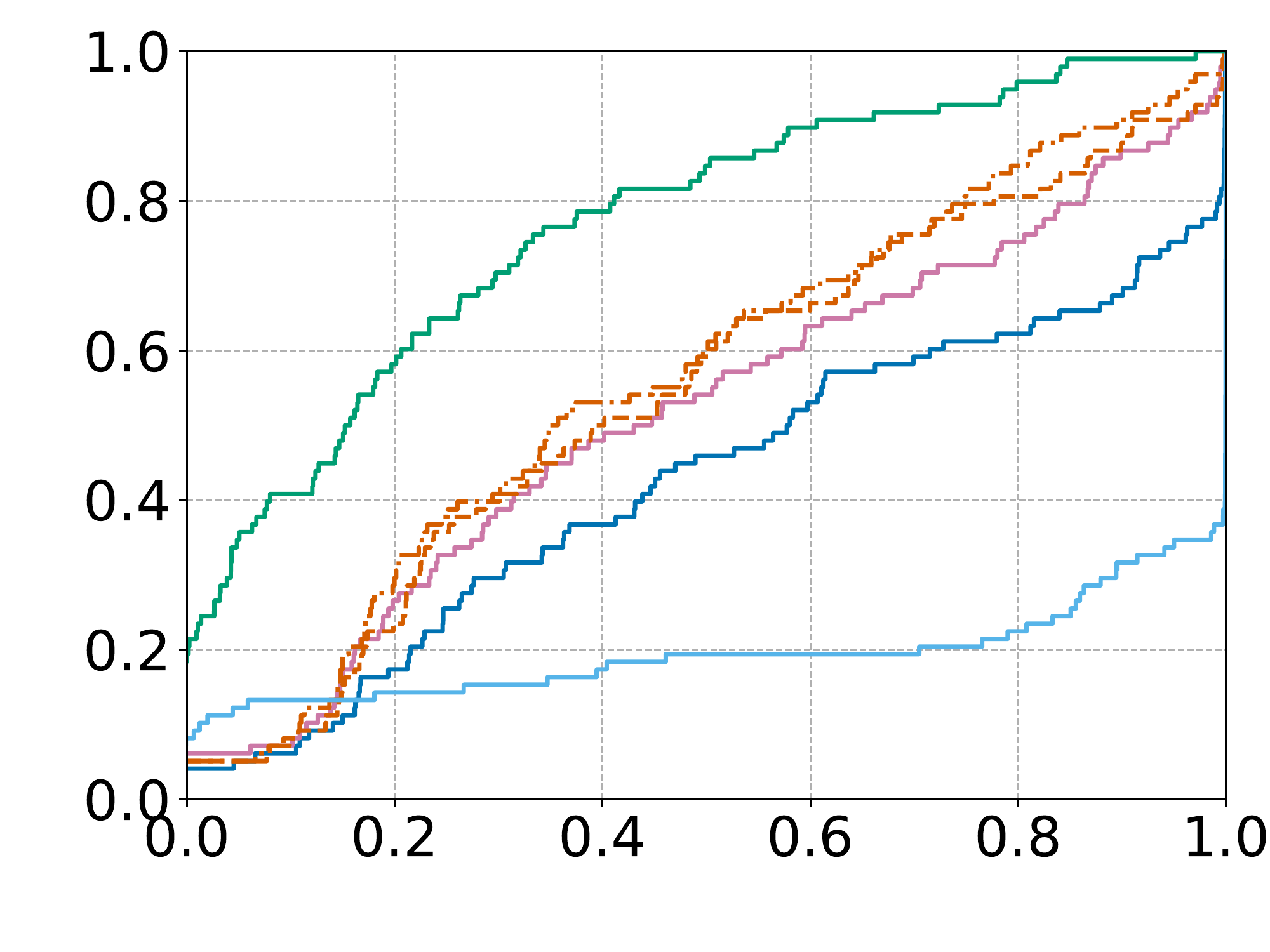}
&
\includegraphics[width=0.29\textwidth,valign=c,trim={0 0 0 0},clip]{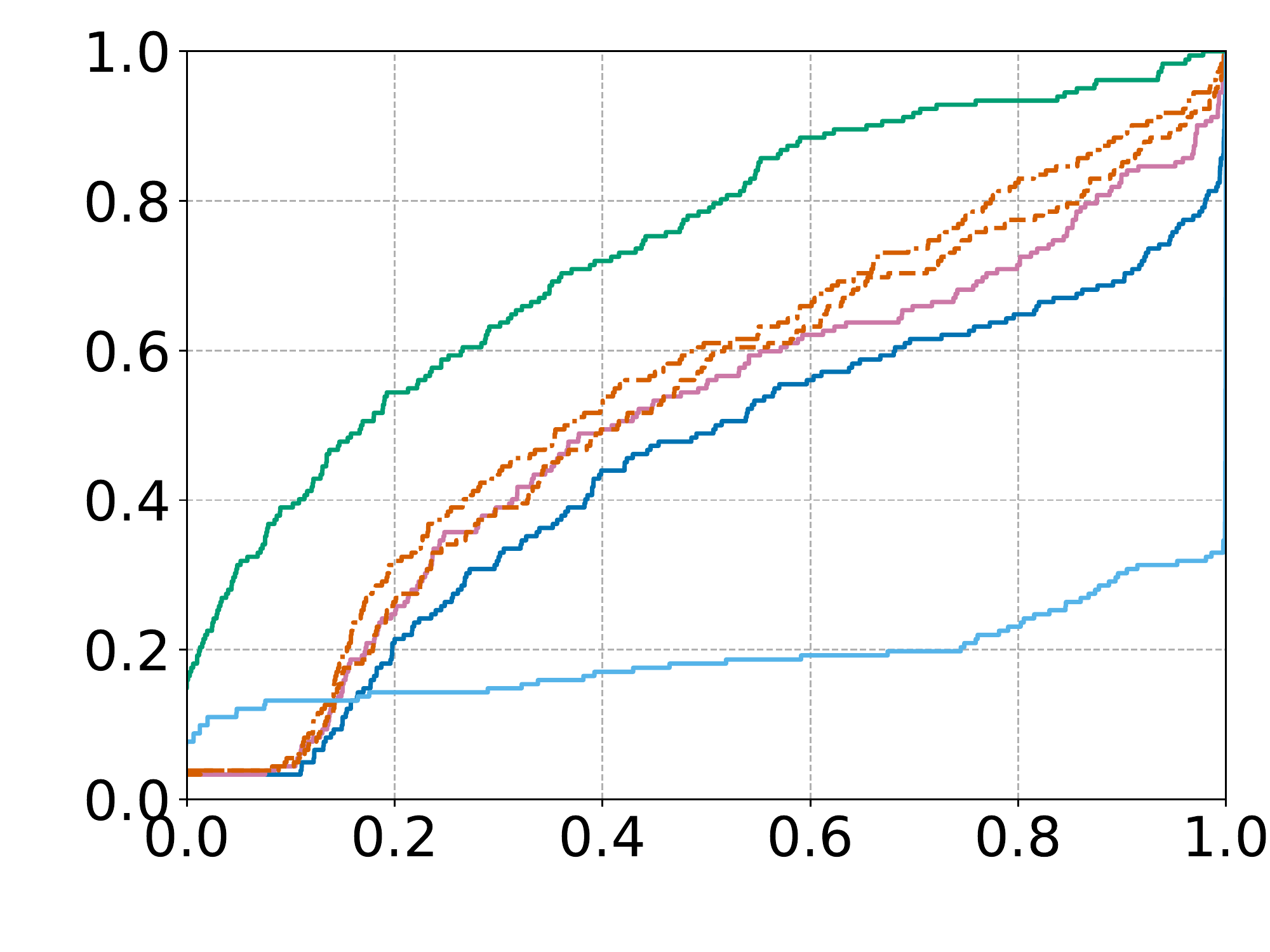}
&
\includegraphics[width=0.29\textwidth,valign=c,trim={0 0 0 0},clip]{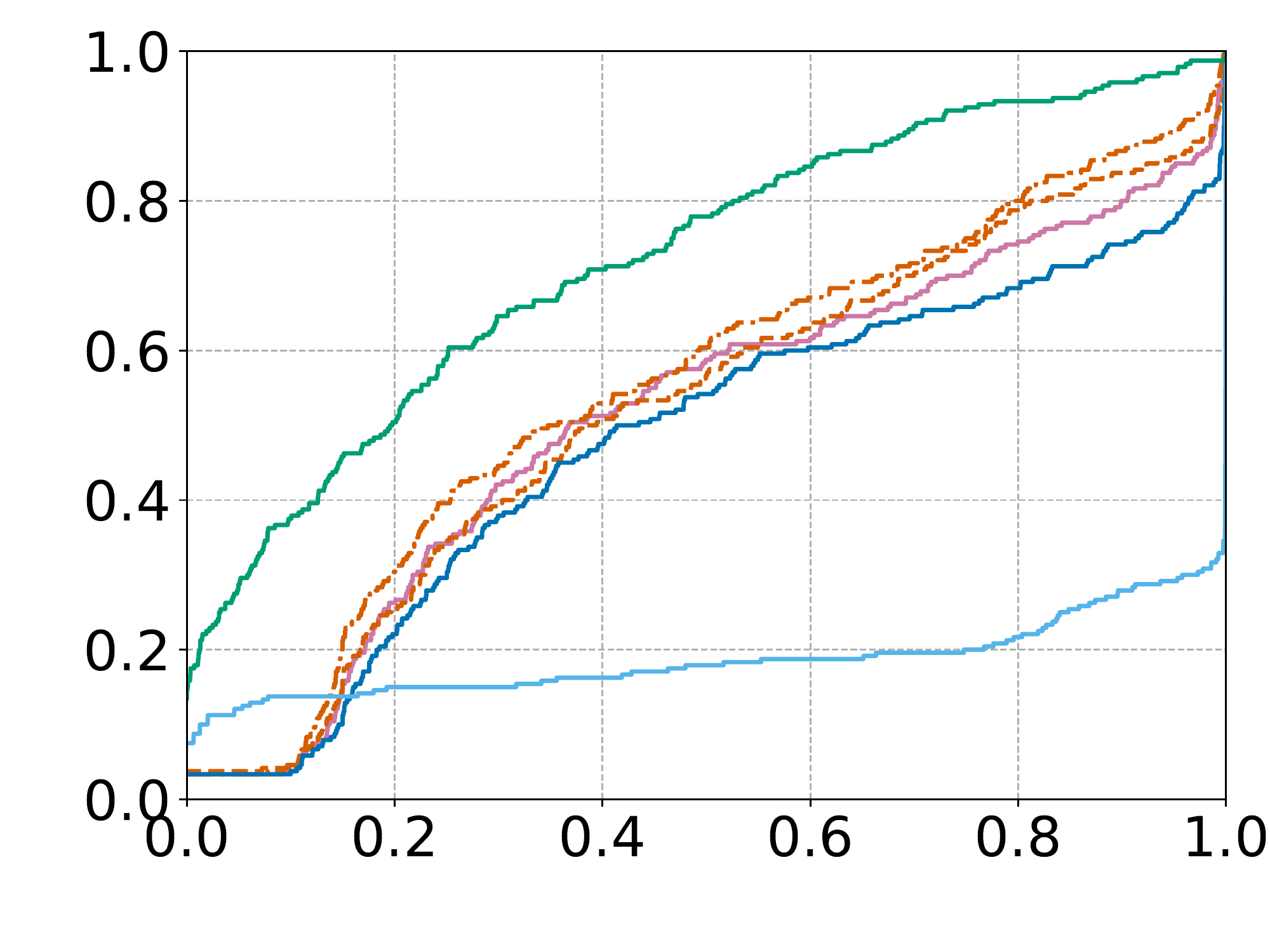}
\\\hline
\begin{tabular}{c}MAJ \\ $p=0.25$\end{tabular}
& \includegraphics[width=0.29\textwidth,valign=c,trim={0 0 0 0},clip]{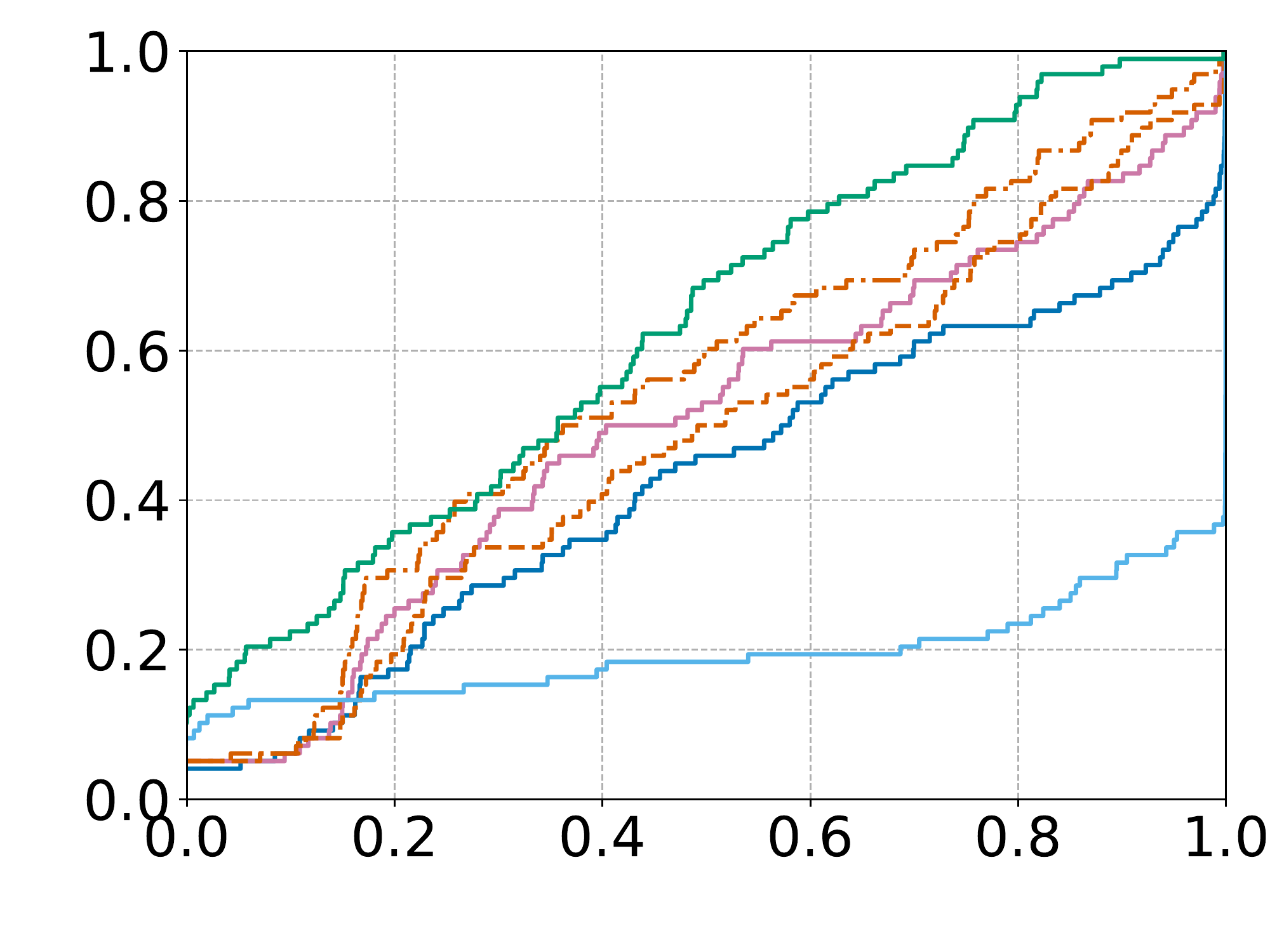}
&
\includegraphics[width=0.29\textwidth,valign=c,trim={0 0 0 0},clip]{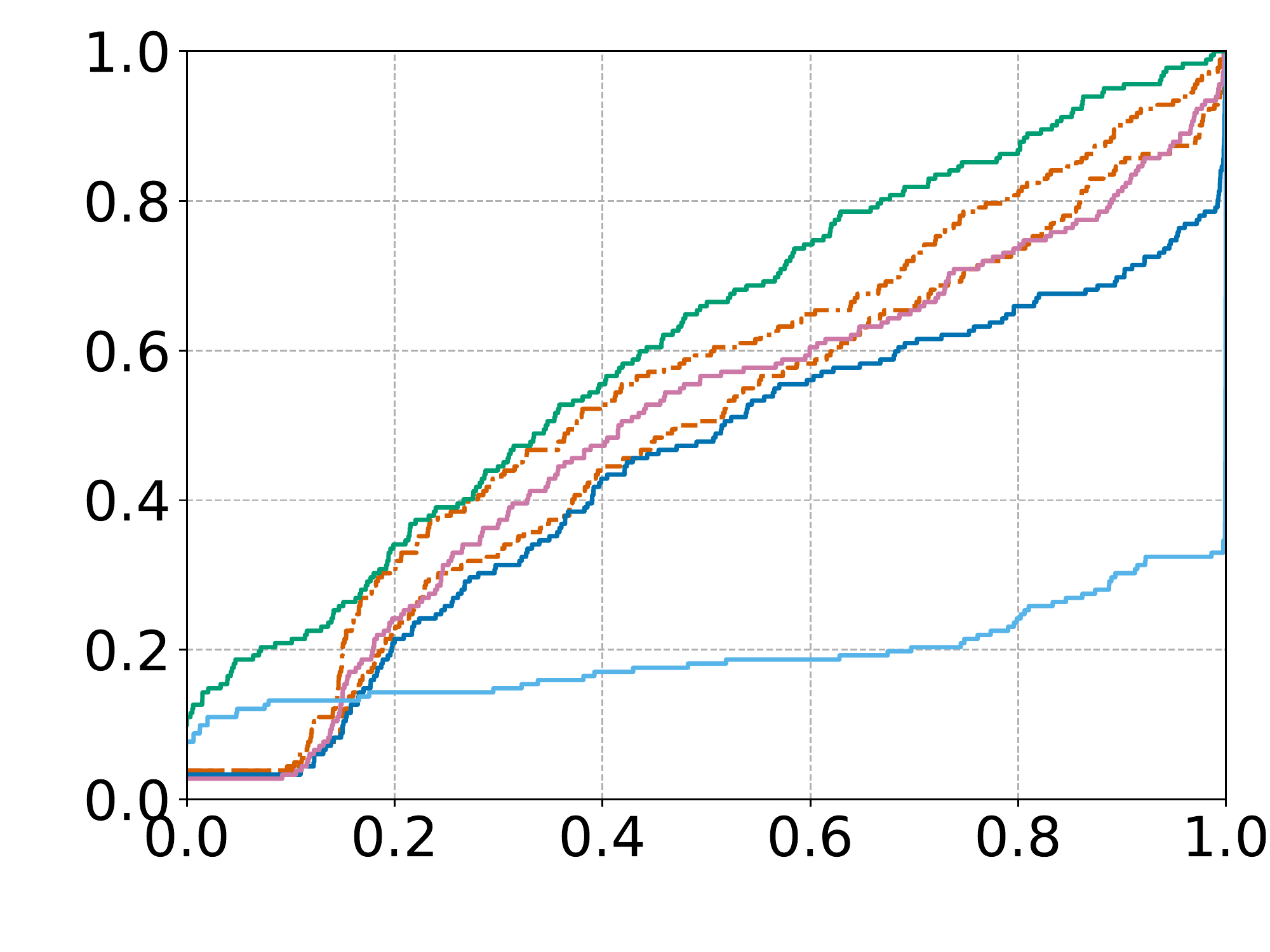}
&
\includegraphics[width=0.29\textwidth,valign=c,trim={0 0 0 0},clip]{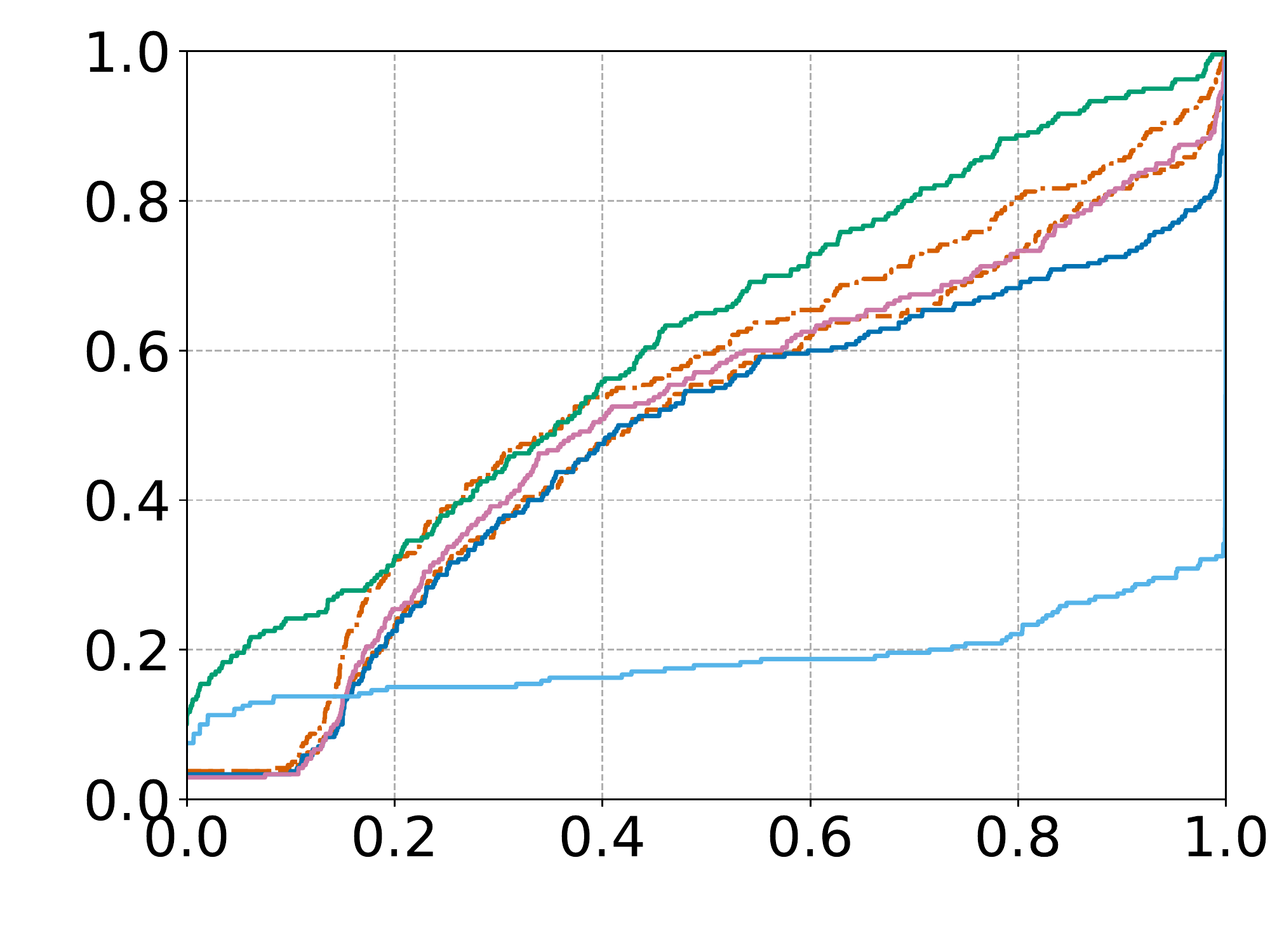}
\\\hline
\begin{tabular}{c}CYC \\ $p=0.25$\end{tabular}
& \includegraphics[width=0.29\textwidth,valign=c,trim={0 0 0 0},clip]{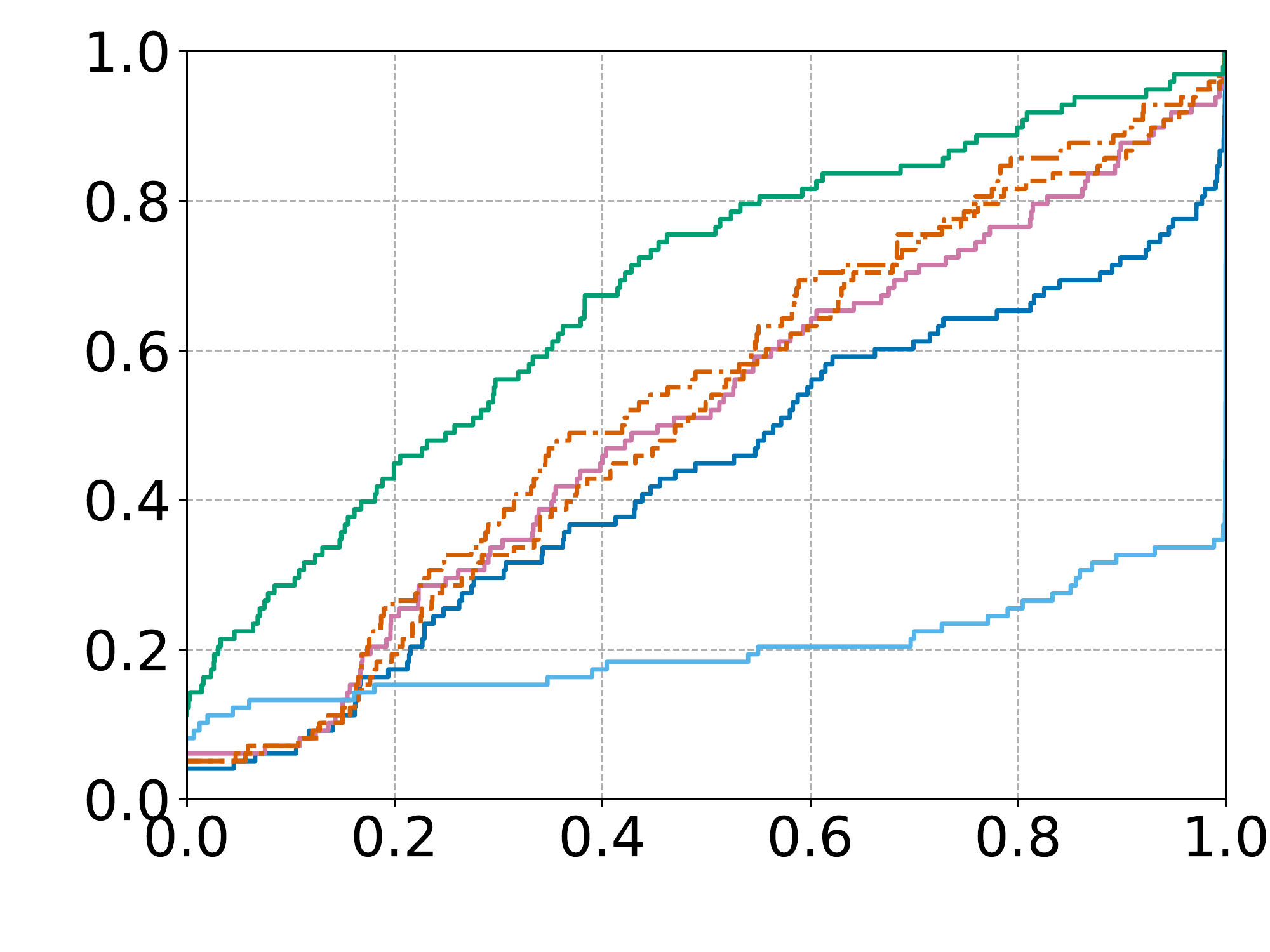}
&
\includegraphics[width=0.29\textwidth,valign=c,trim={0 0 0 0},clip]{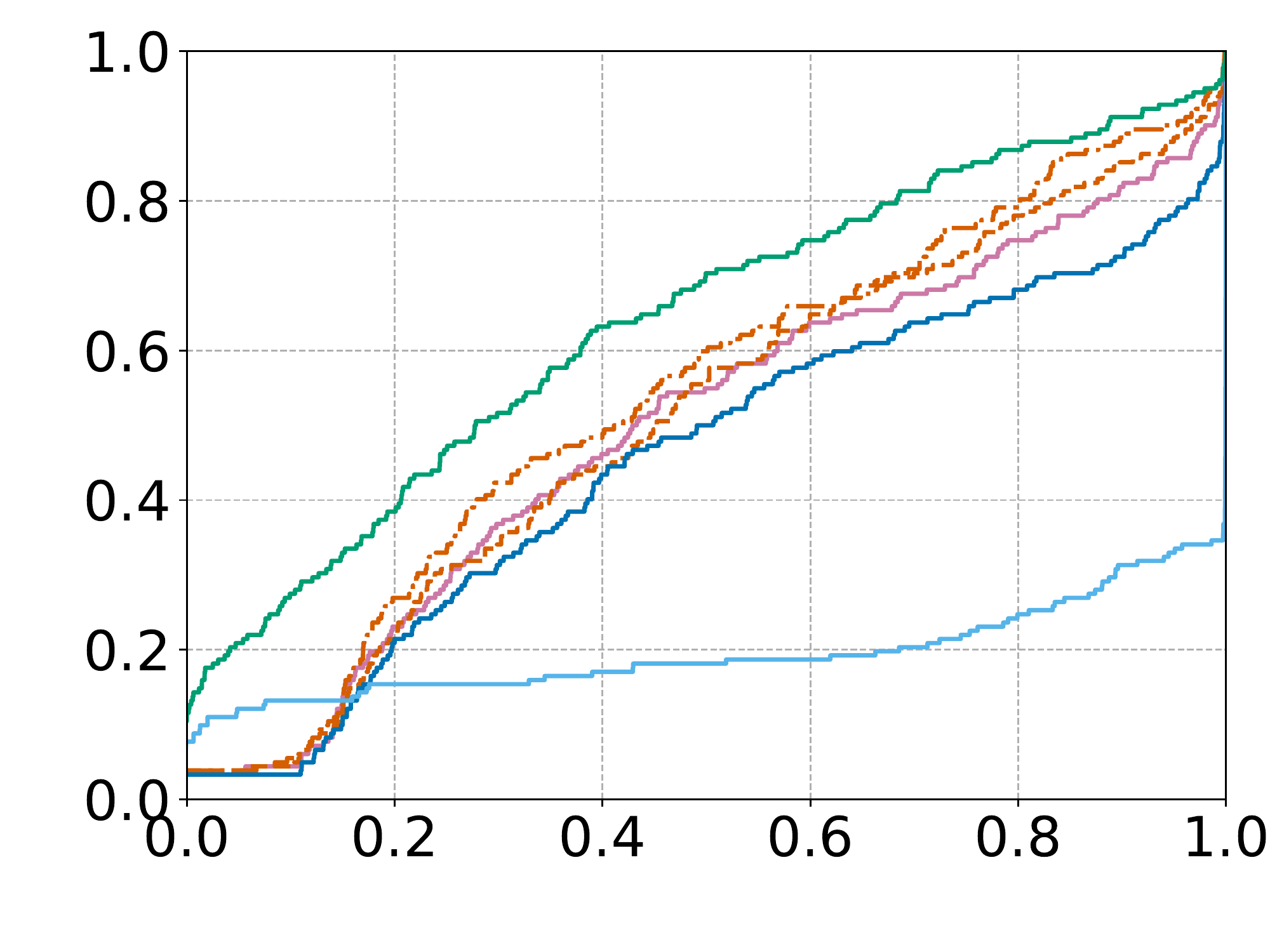}
&
\includegraphics[width=0.29\textwidth,valign=c,trim={0 0 0 0},clip]{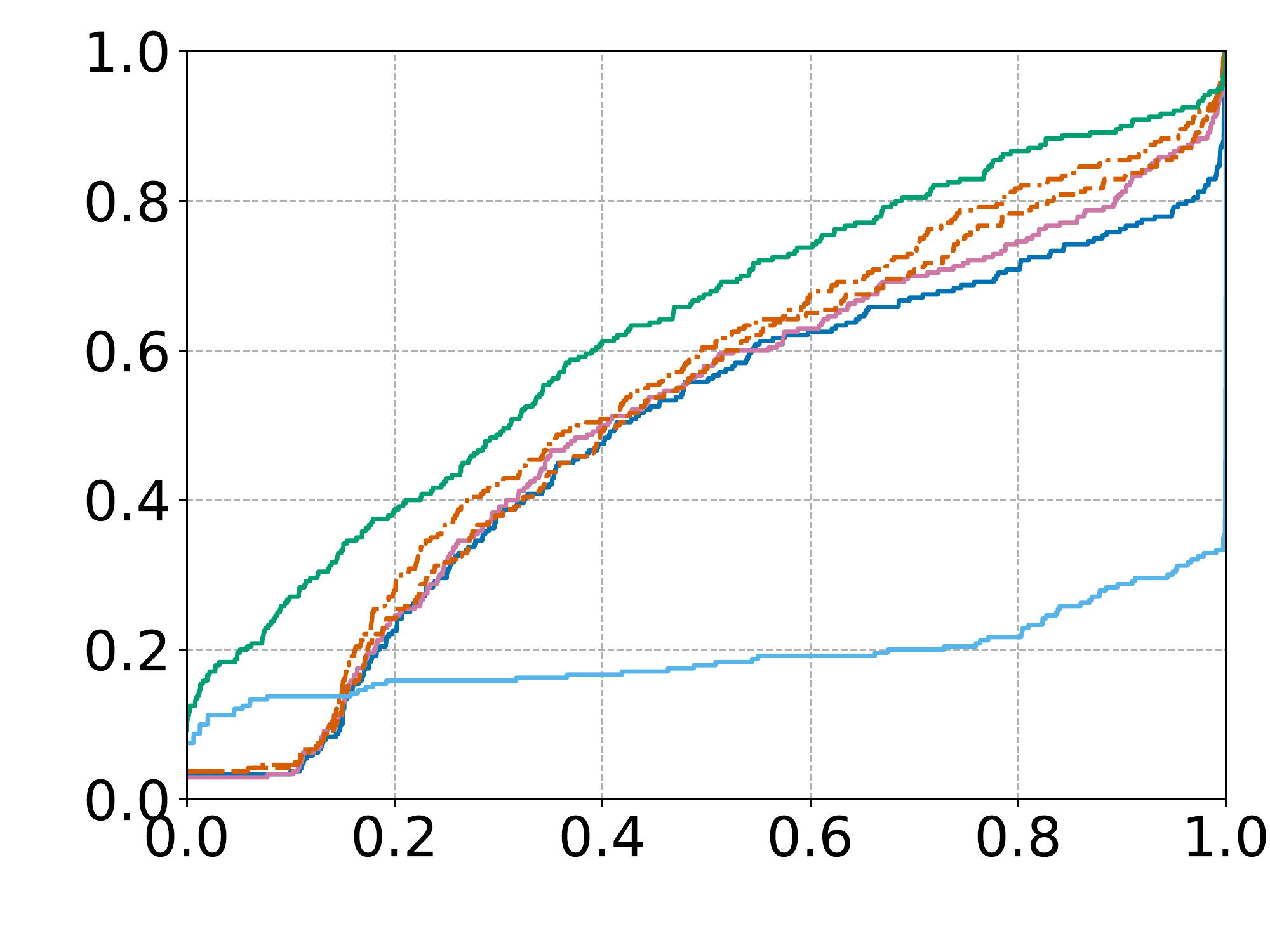}
\\\hline
\begin{tabular}{c}UAR \\ $p=0.5$\end{tabular}
& \includegraphics[width=0.29\textwidth,valign=c,trim={0 0 0 0},clip]{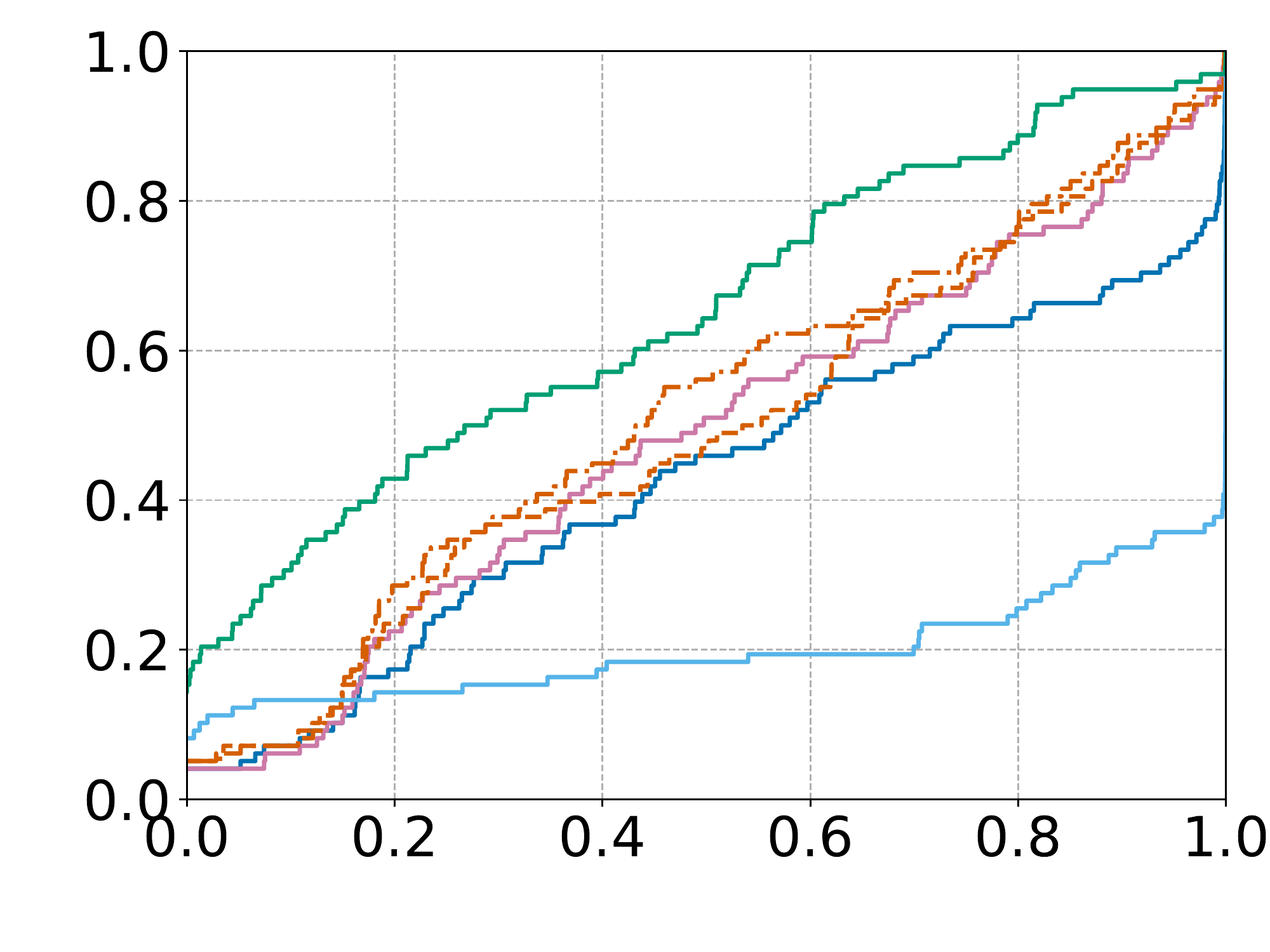}
&
\includegraphics[width=0.29\textwidth,valign=c,trim={0 0 0 0},clip]{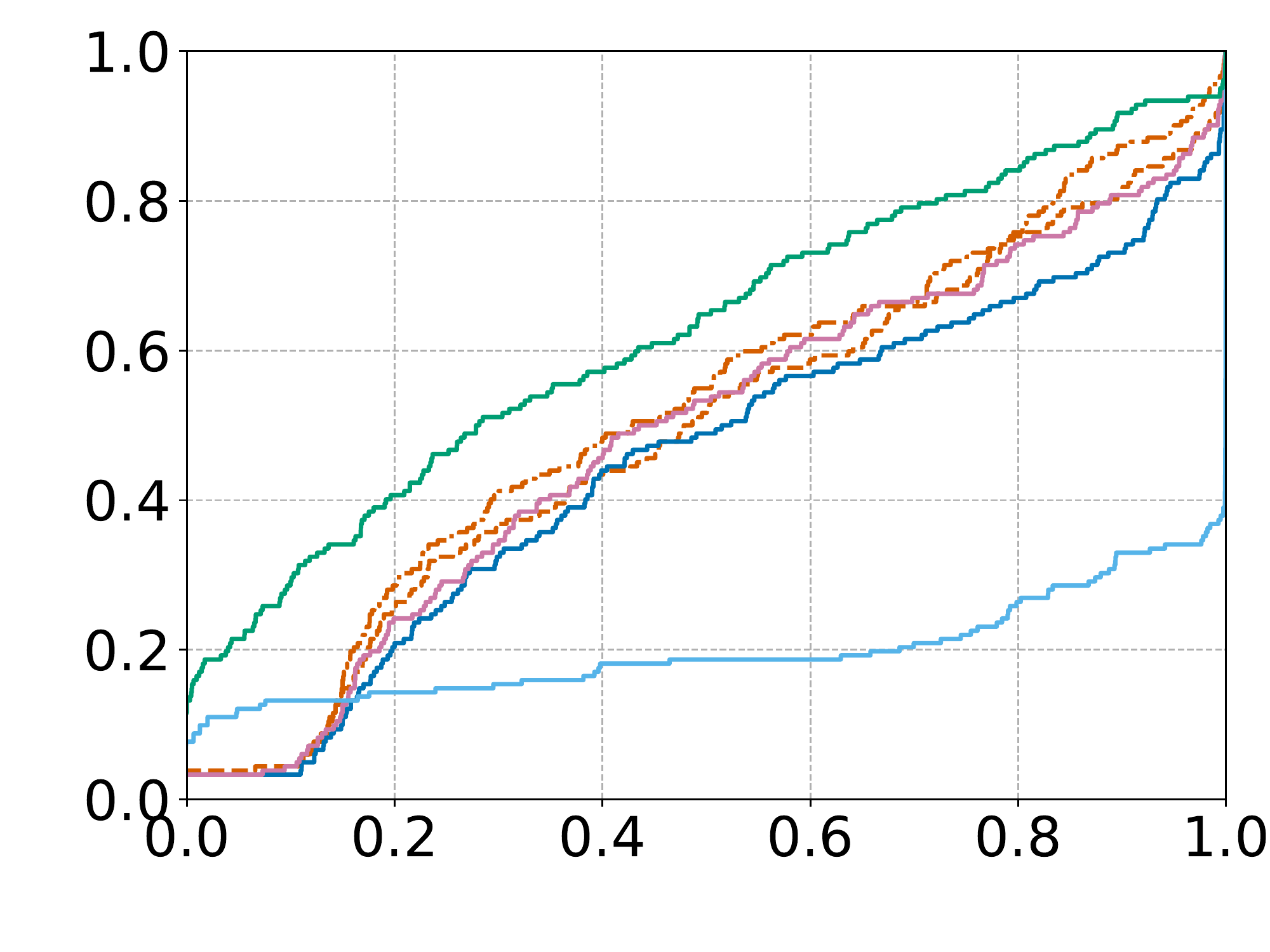}
&
\includegraphics[width=0.29\textwidth,valign=c,trim={0 0 0 0},clip]{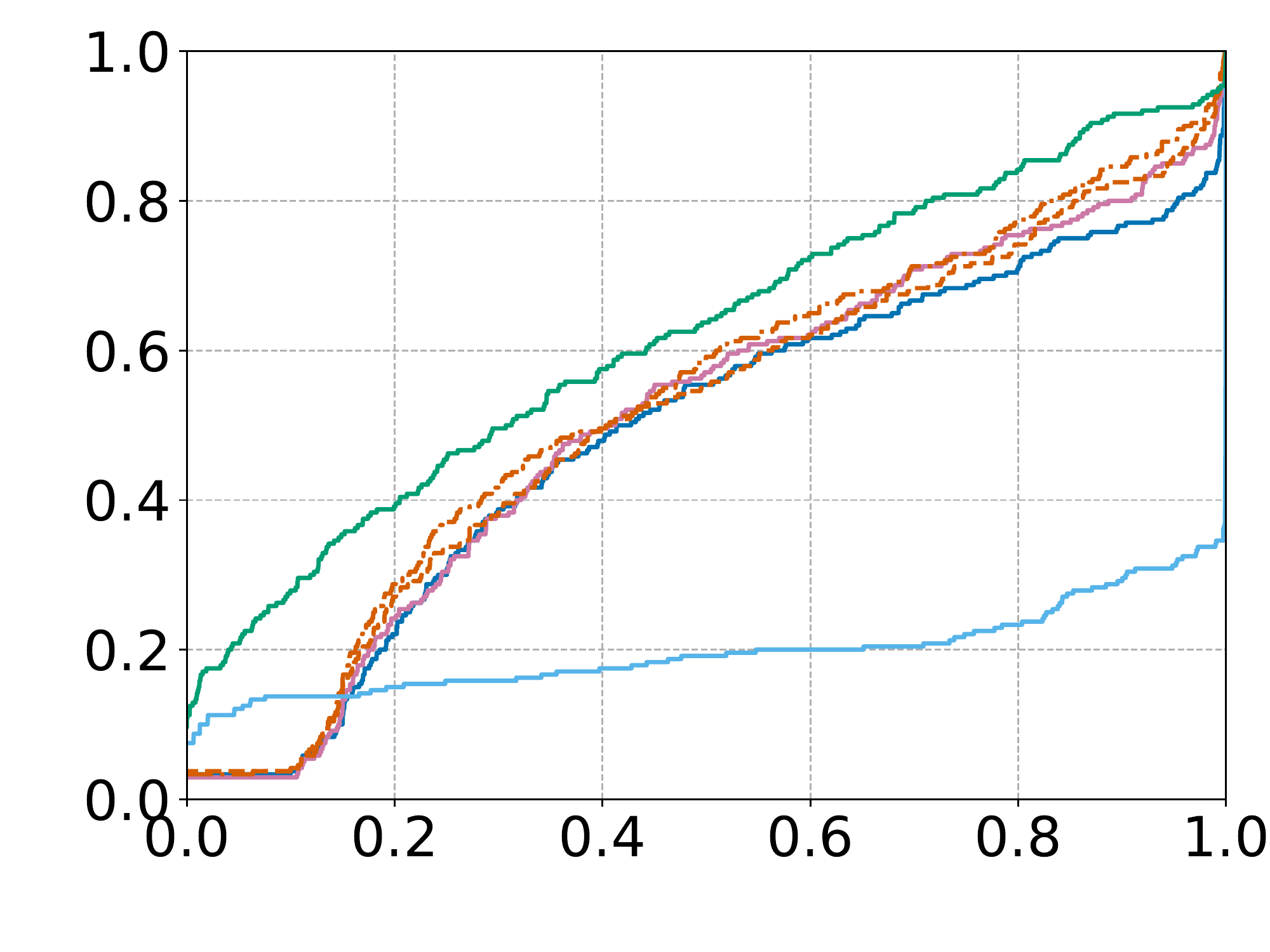}
\\\bottomrule
\multicolumn{4}{c}{
		\begin{subfigure}[t]{\textwidth}
    \includegraphics[width=\textwidth,trim={0 0 0 0},clip]{figs/cdfs/legends/legend_Lambda=8.pdf}
		\end{subfigure} }
\end{tabular}
\caption{Cumulative distribution functions (CDFs) for all evaluated algorithms, against different noise settings and warm start ratios in $\cbr{2.875,5.75,11.5}$. All CB algorithms use $\epsilon$-greedy with $\epsilon=0.1$. In each of the above plots, the $x$ axis represents scores, while the $y$ axis represents the CDF values.}
\label{fig:cdfs-eps=0.1-1}
\end{figure}

\begin{figure}[H]
\centering
\begin{tabular}{c | @{}c@{ }c@{ }c@{}} 
\toprule
& \multicolumn{3}{c}{ Ratio }
\\
Noise & 23.0 & 46.0 & 92.0
\\\midrule
Noiseless & \includegraphics[width=0.29\textwidth,valign=c,trim={0 0 0 0},clip]{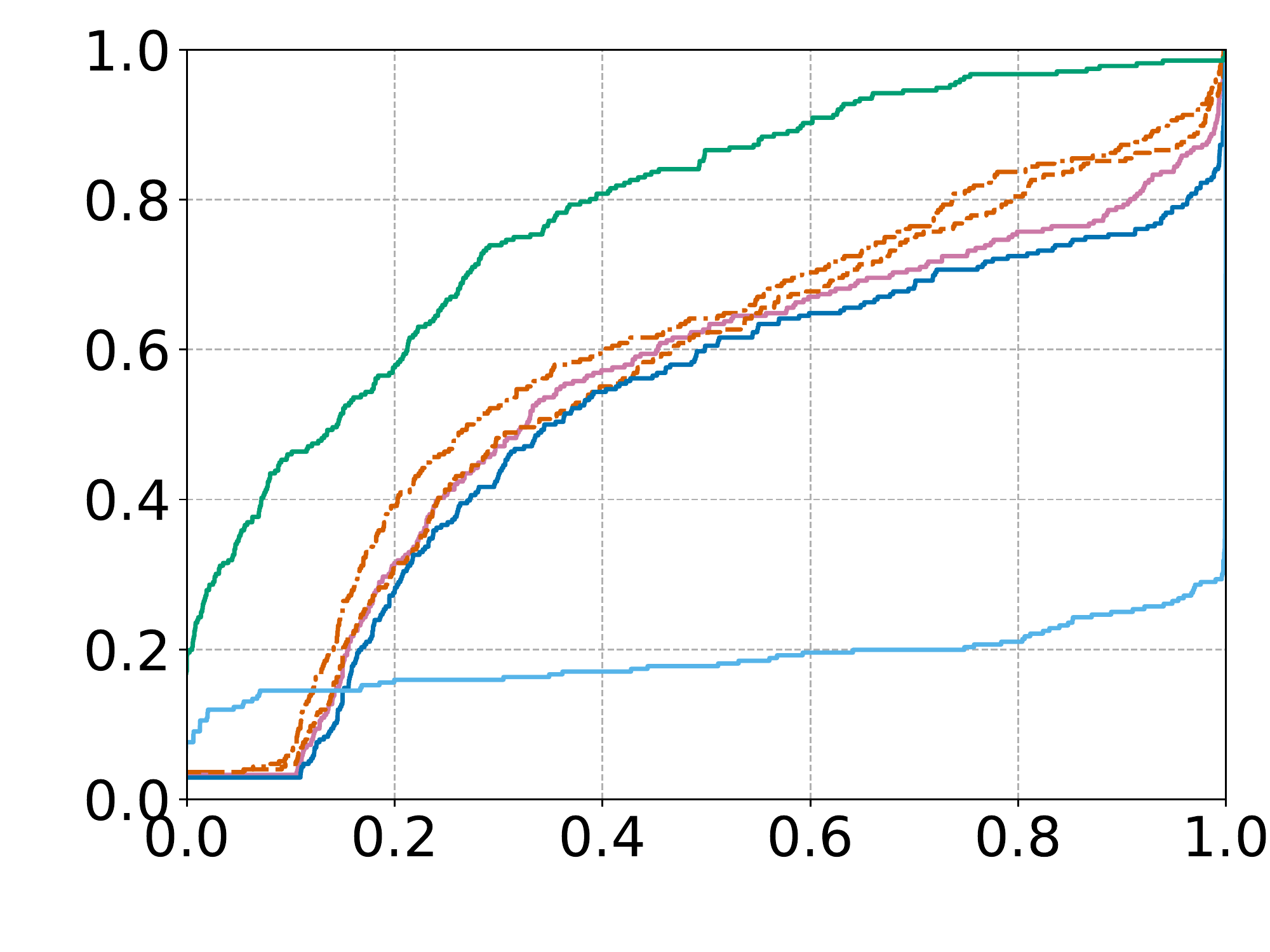}
&
\includegraphics[width=0.29\textwidth,valign=c,trim={0 0 0 0},clip]{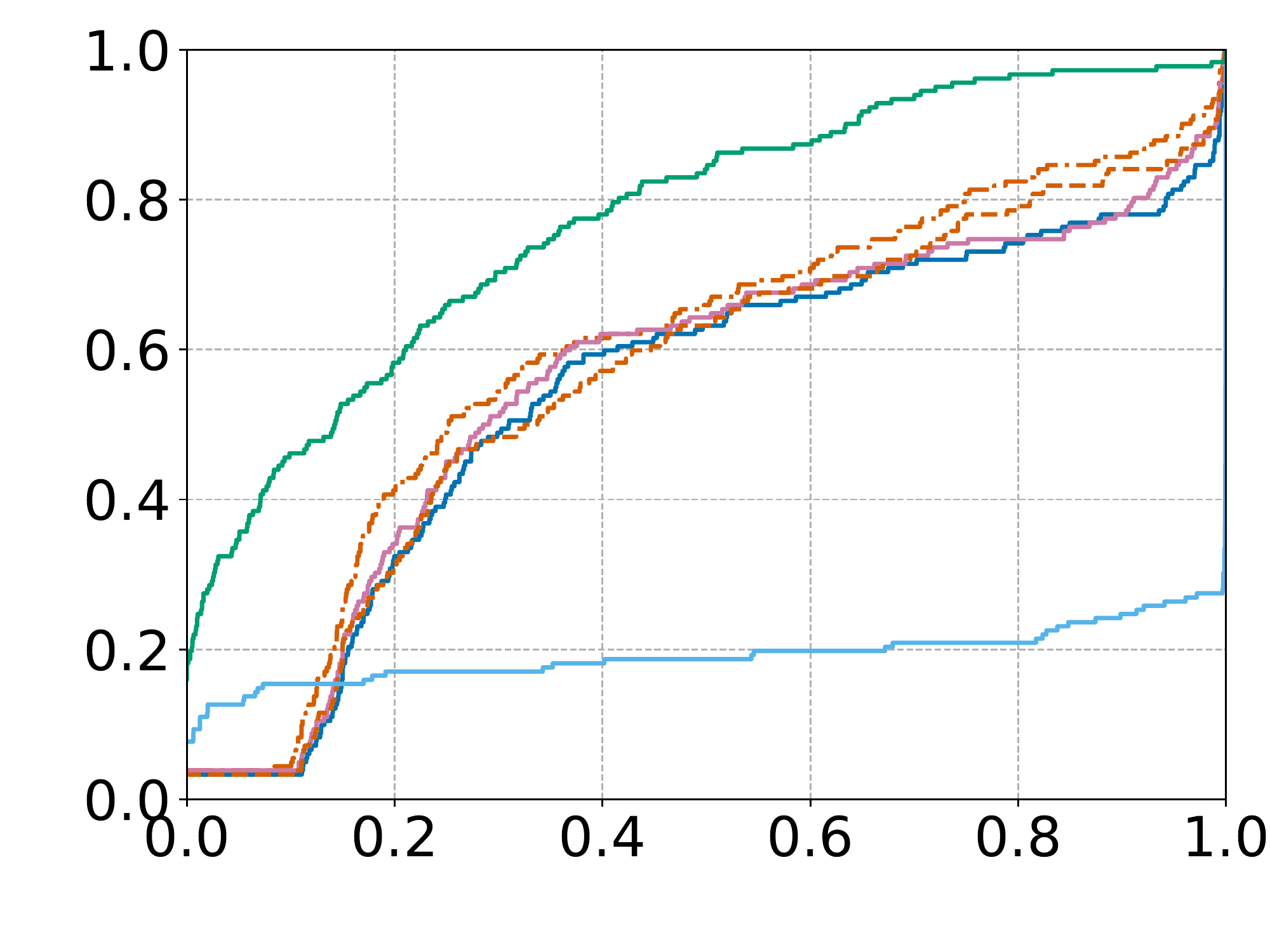}
&
\includegraphics[width=0.29\textwidth,valign=c,trim={0 0 0 0},clip]{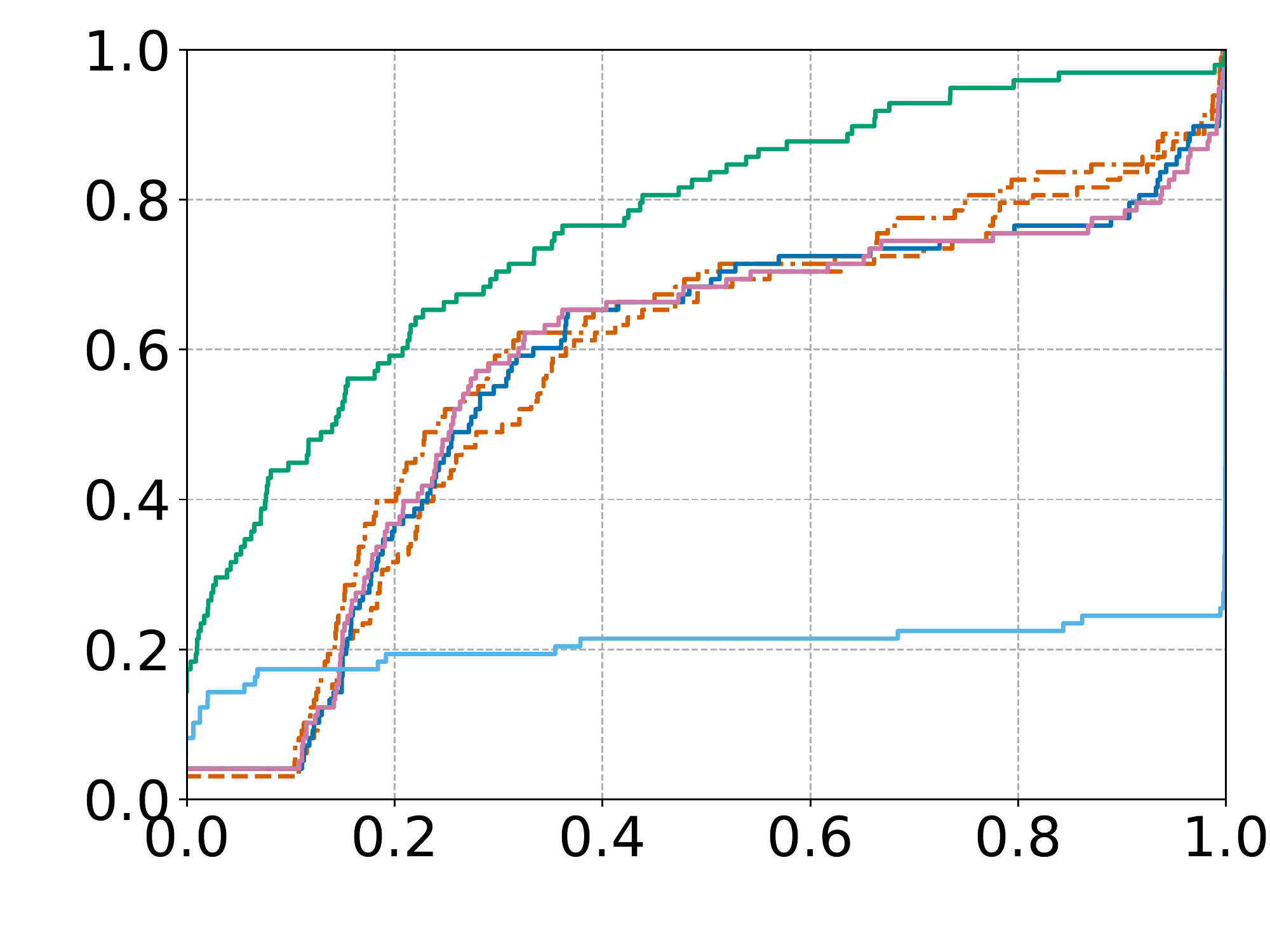}
\\\hline
\begin{tabular}{c}UAR \\ $p=0.25$\end{tabular}
& \includegraphics[width=0.29\textwidth,valign=c,trim={0 0 0 0},clip]{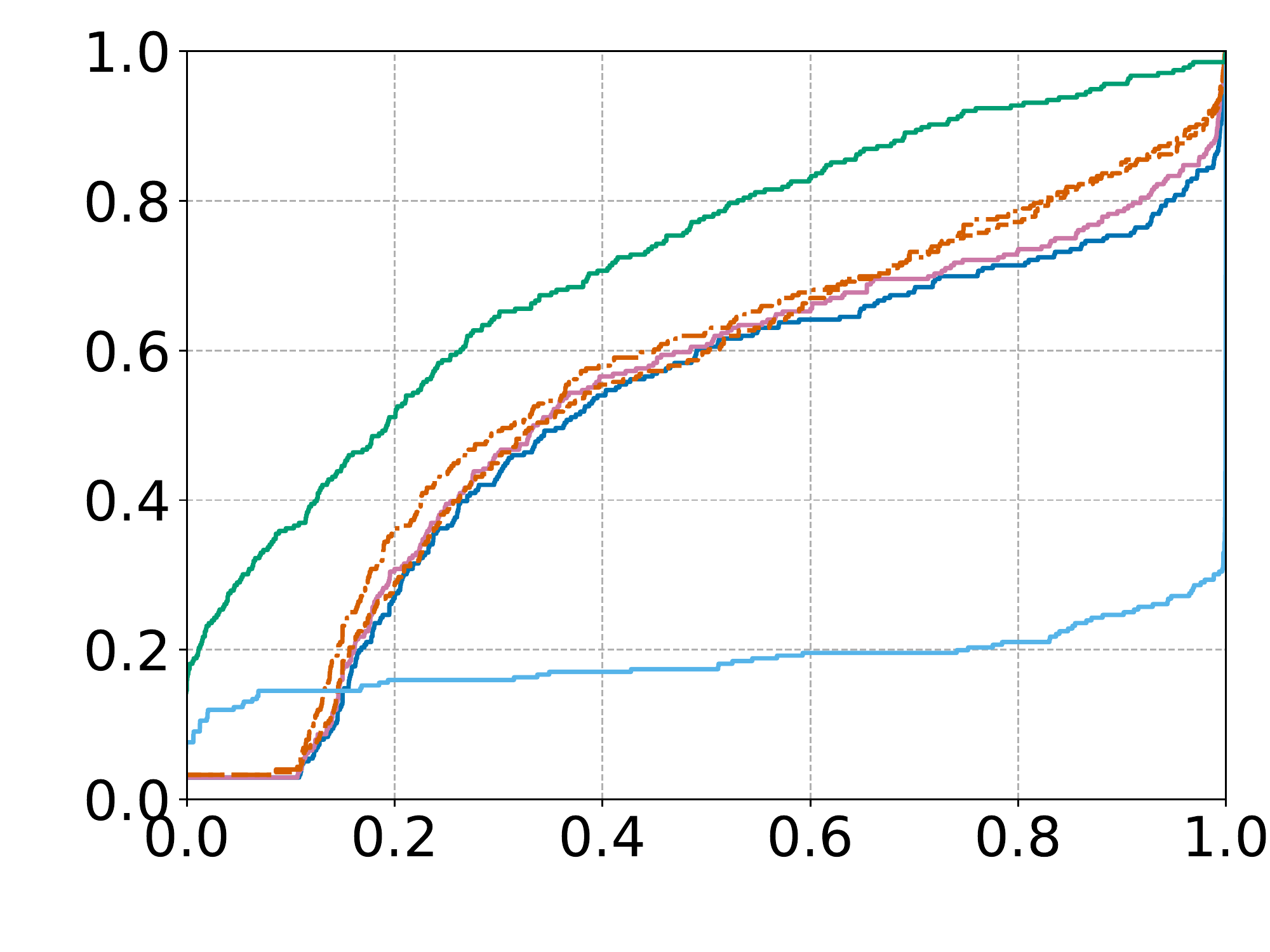}
&
\includegraphics[width=0.29\textwidth,valign=c,trim={0 0 0 0},clip]{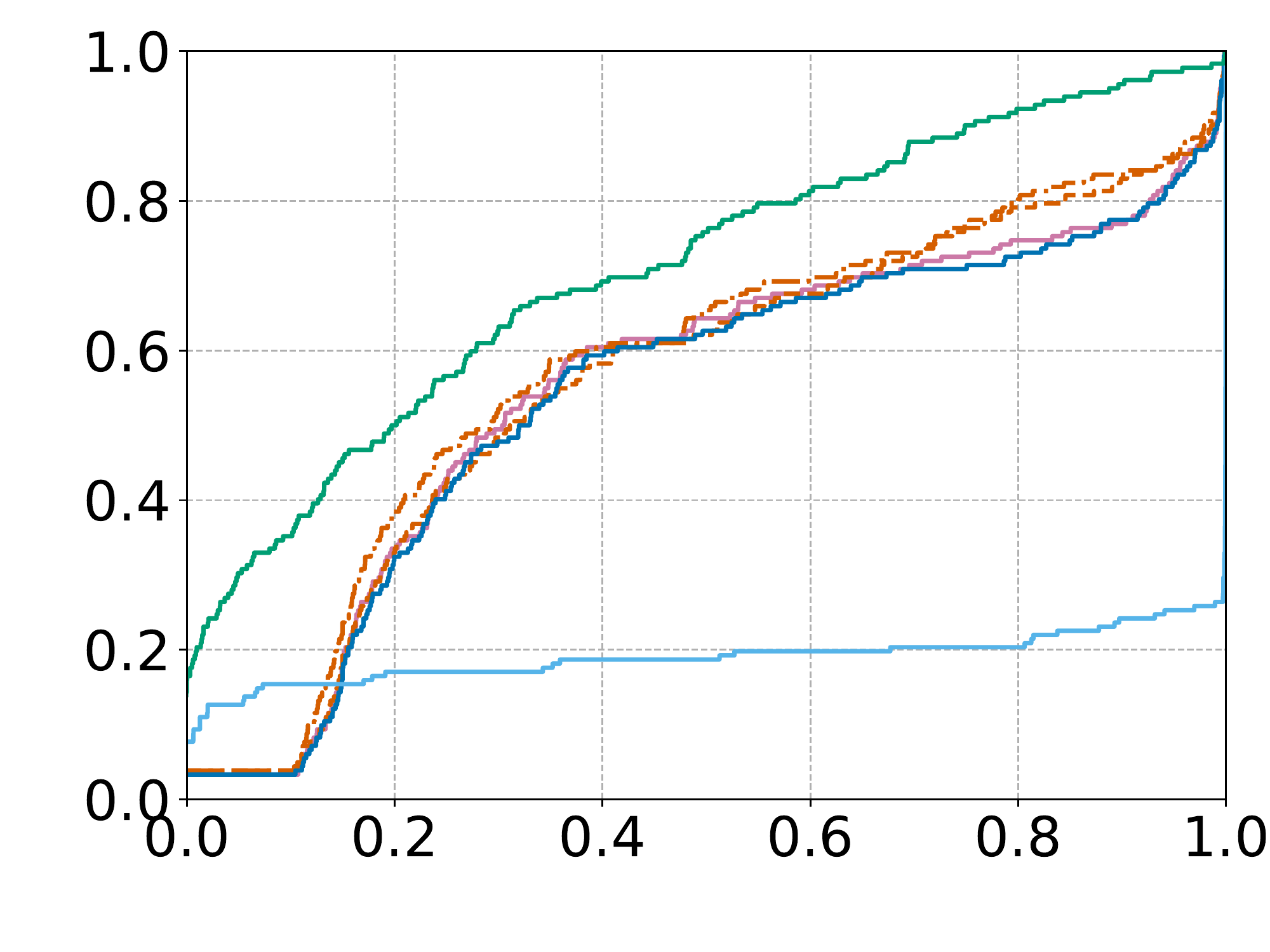}
&
\includegraphics[width=0.29\textwidth,valign=c,trim={0 0 0 0},clip]{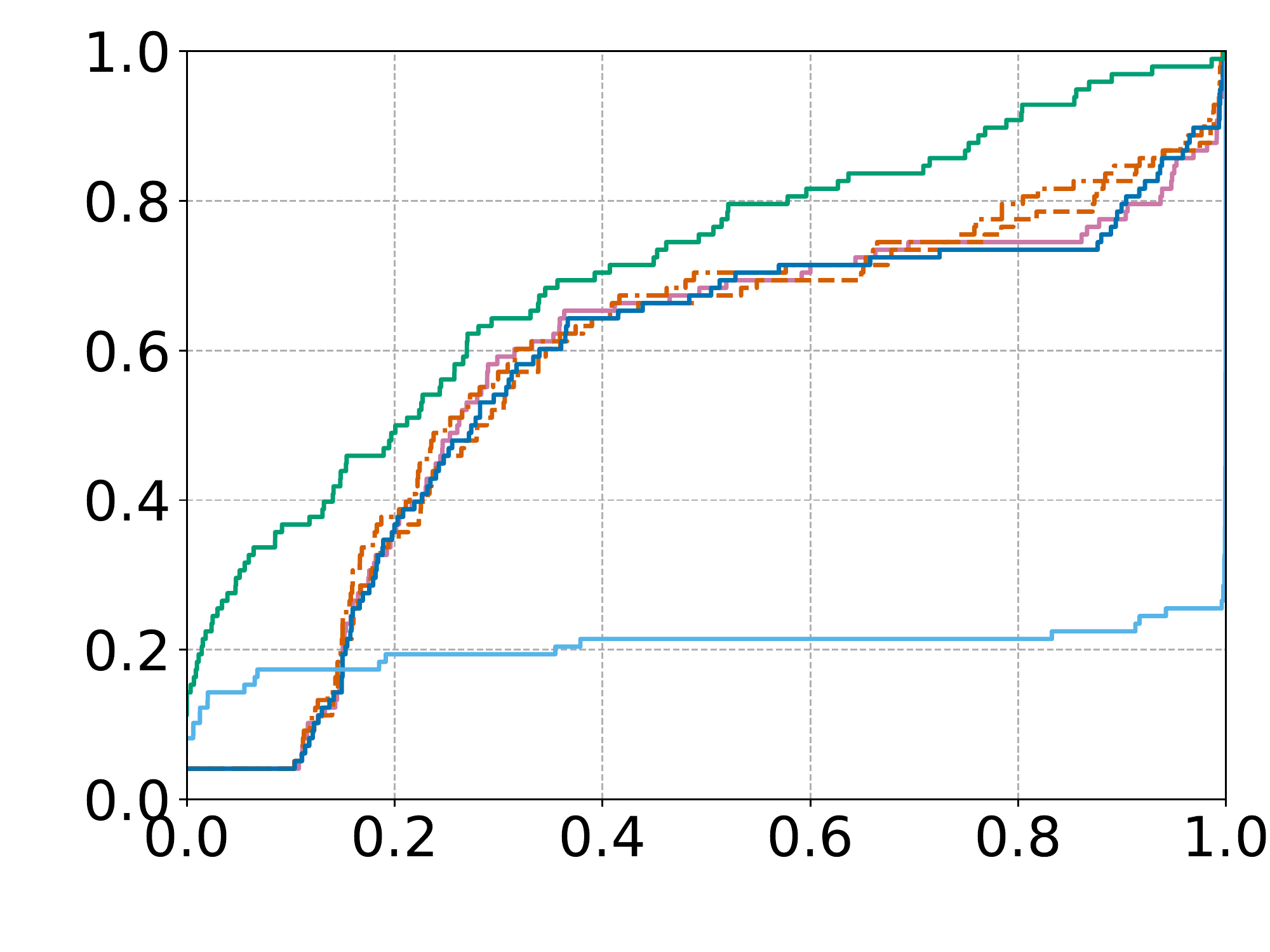}
\\\hline
\begin{tabular}{c}MAJ \\ $p=0.25$\end{tabular}
& \includegraphics[width=0.29\textwidth,valign=c,trim={0 0 0 0},clip]{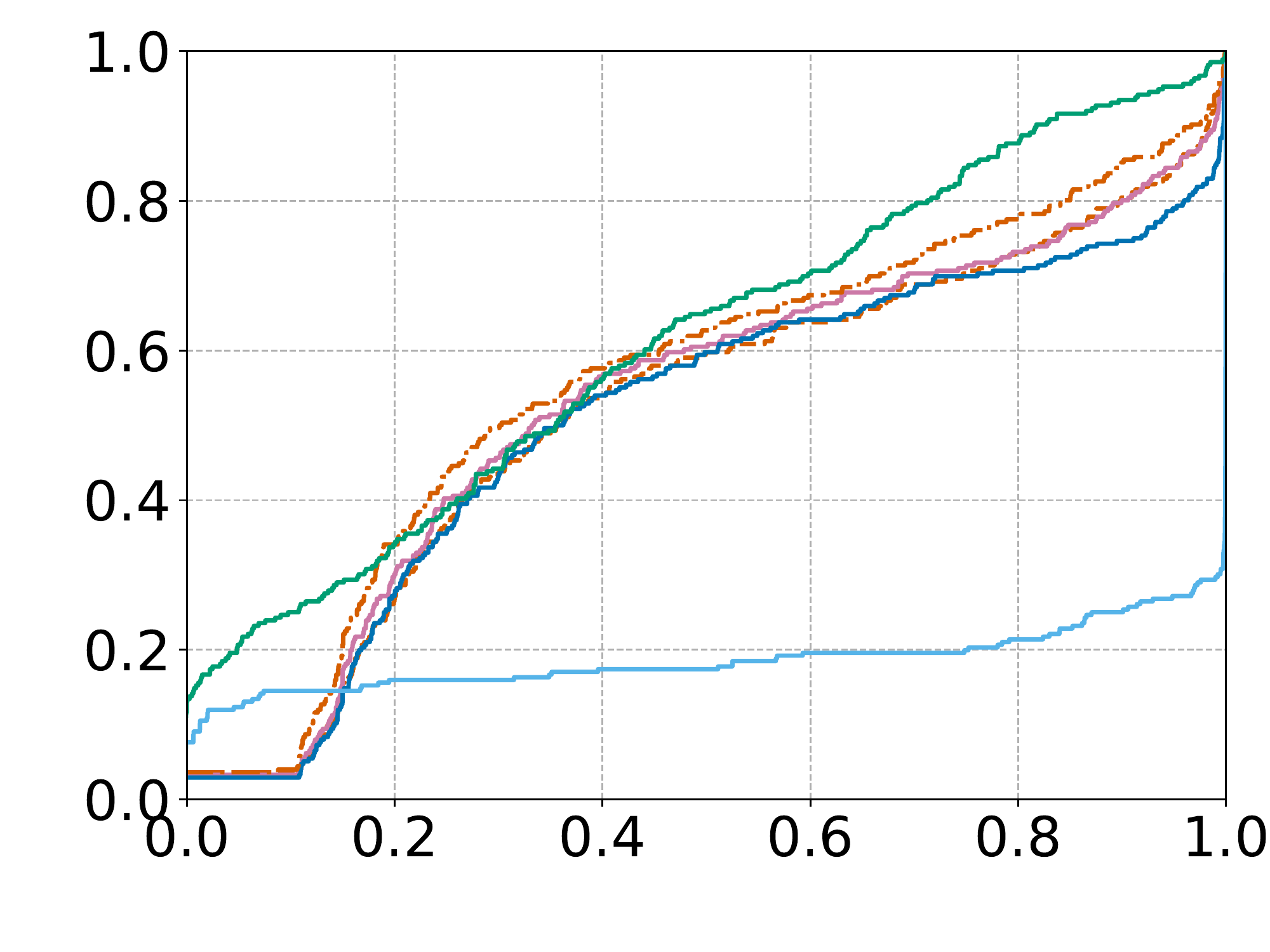}
&
\includegraphics[width=0.29\textwidth,valign=c,trim={0 0 0 0},clip]{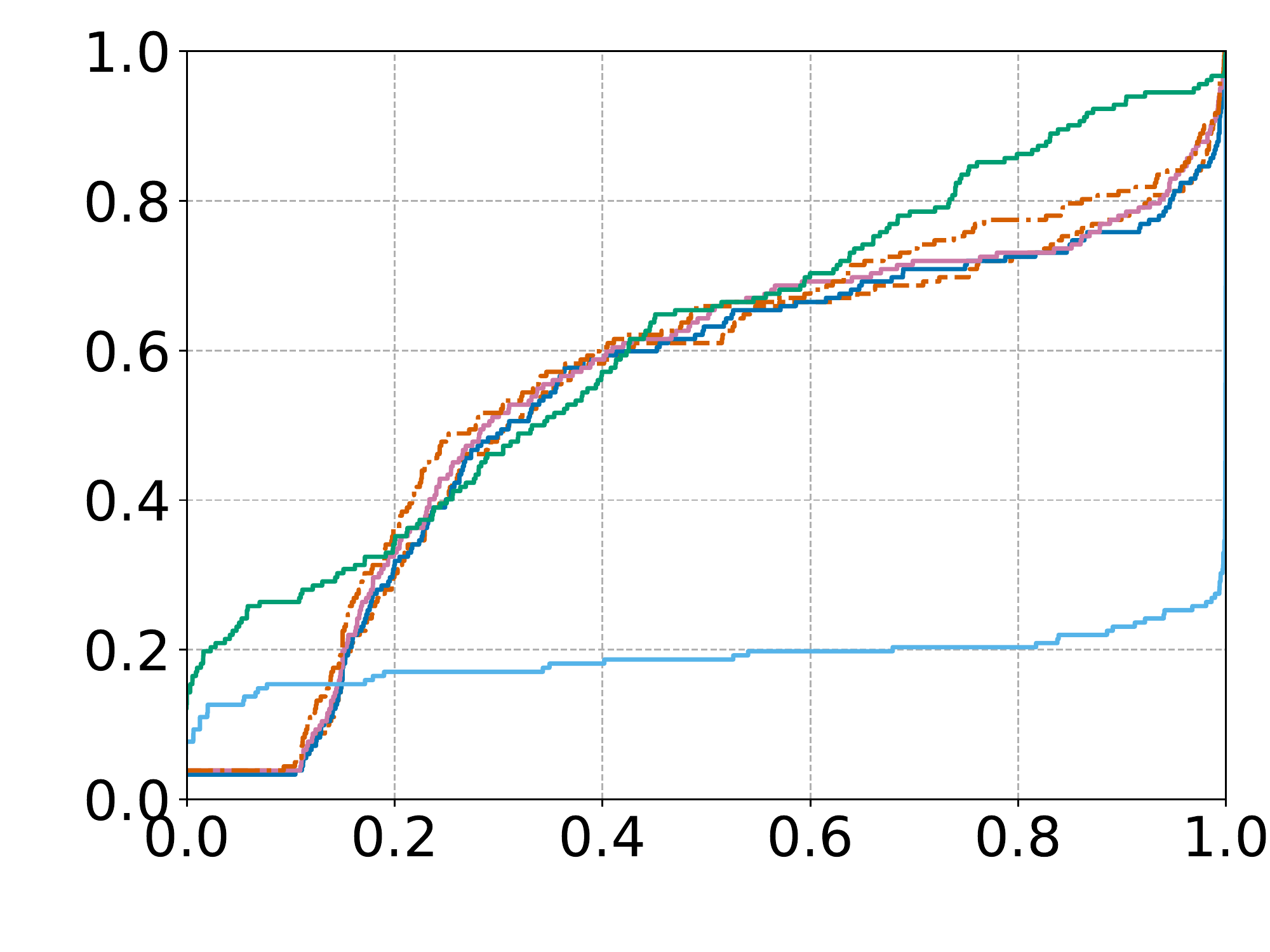}
&
\includegraphics[width=0.29\textwidth,valign=c,trim={0 0 0 0},clip]{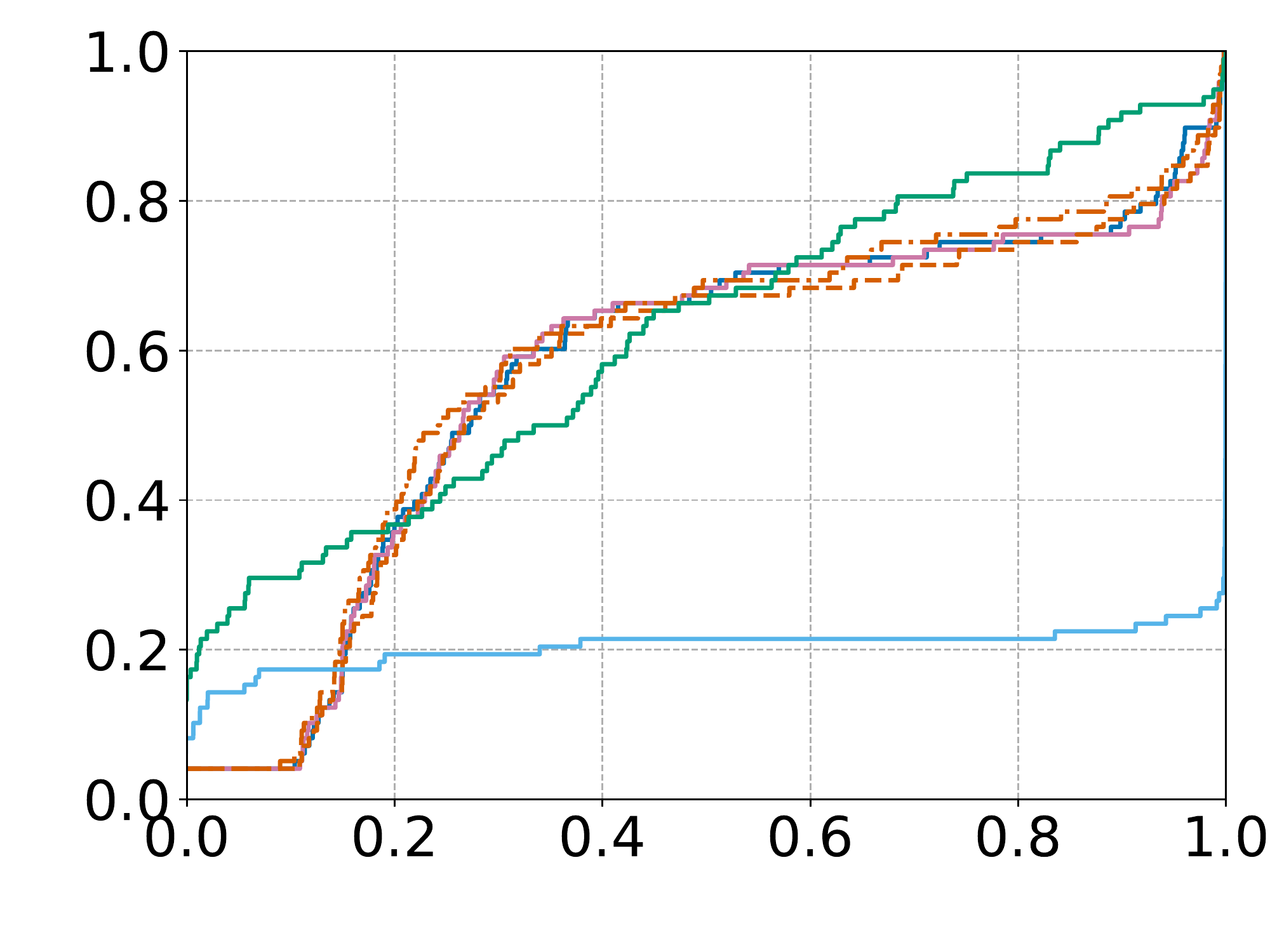}
\\\hline
\begin{tabular}{c}CYC \\ $p=0.25$\end{tabular}
& \includegraphics[width=0.29\textwidth,valign=c,trim={0 0 0 0},clip]{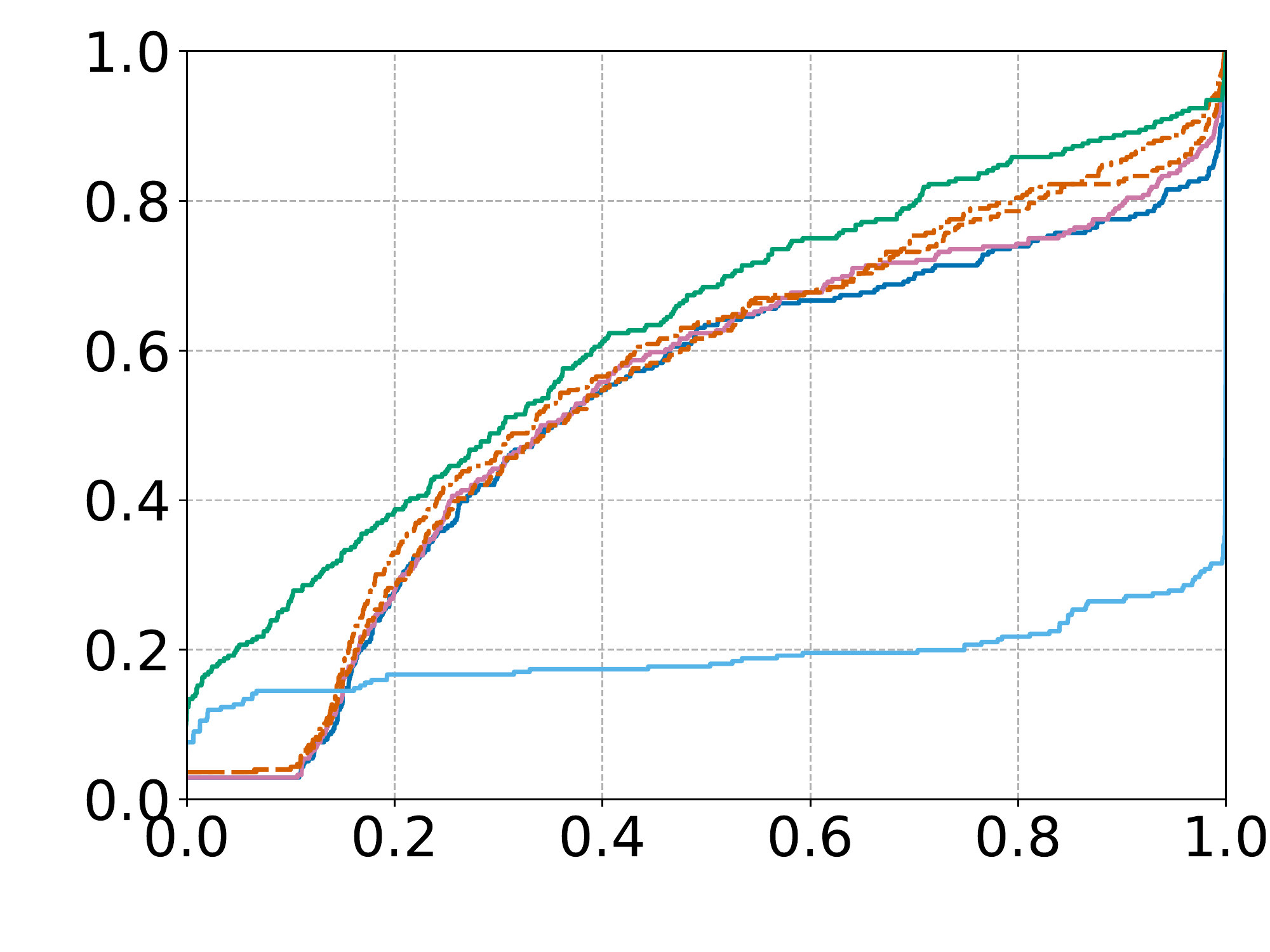}
&
\includegraphics[width=0.29\textwidth,valign=c,trim={0 0 0 0},clip]{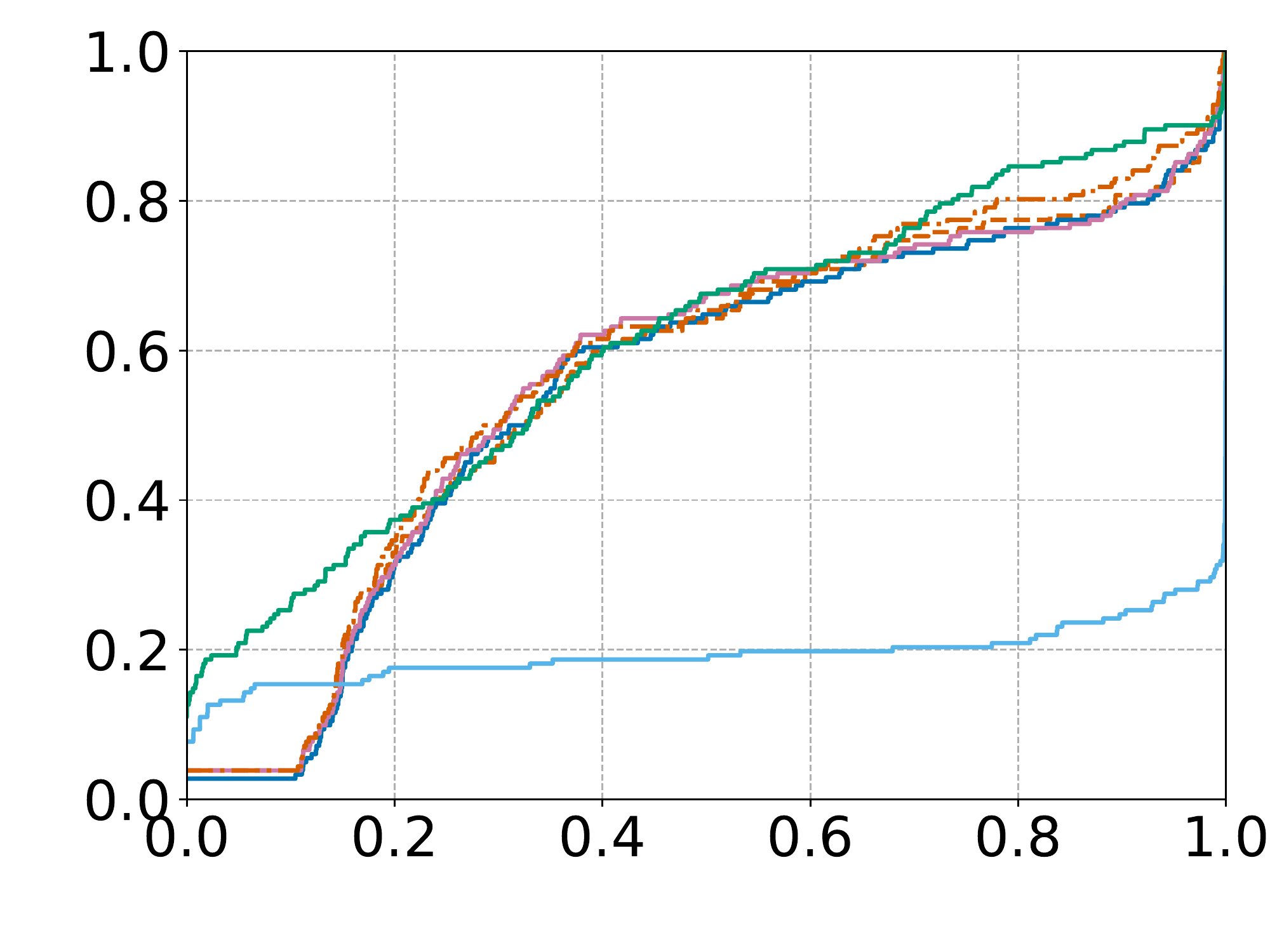}
&
\includegraphics[width=0.29\textwidth,valign=c,trim={0 0 0 0},clip]{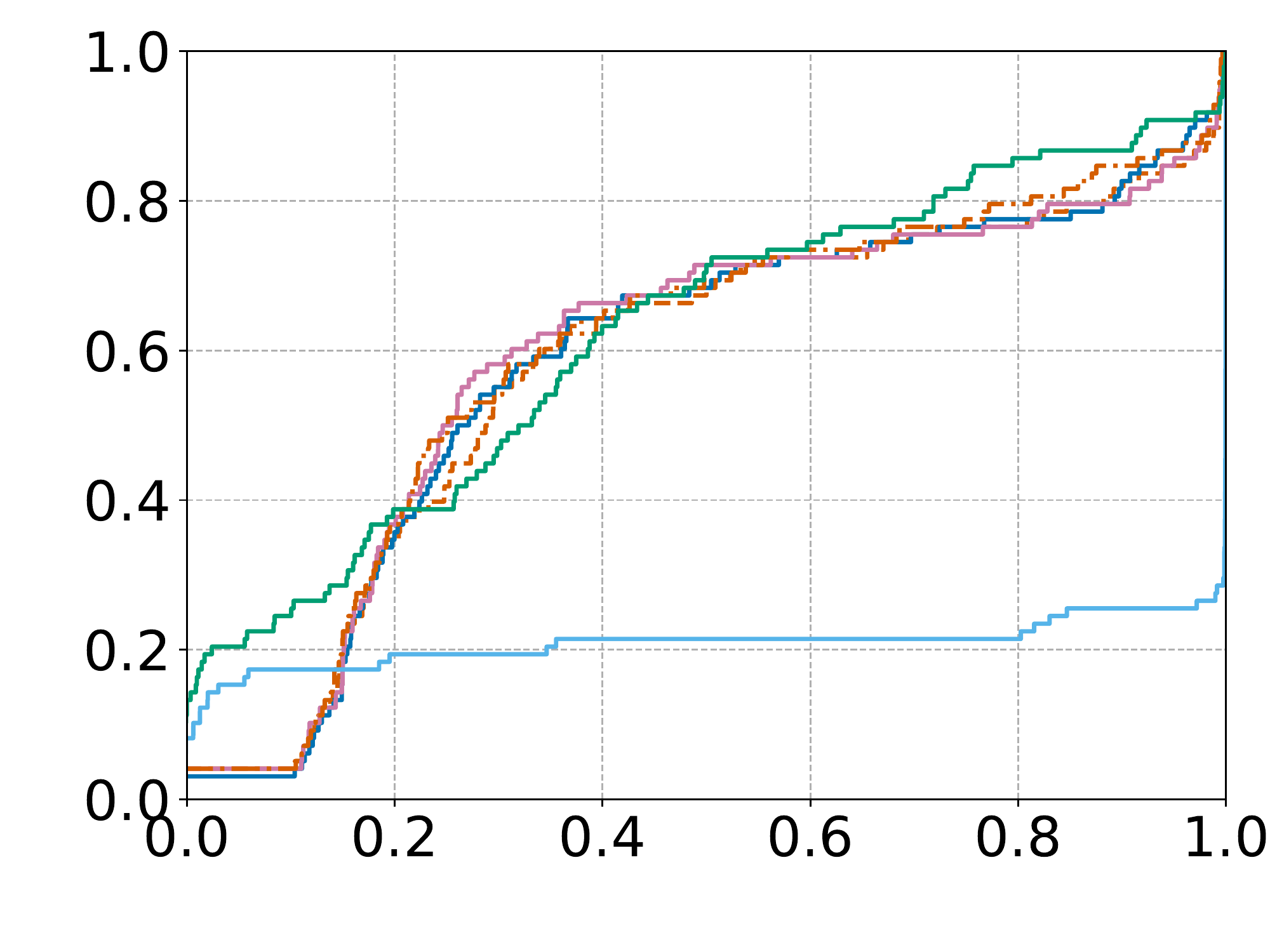}
\\\hline
\begin{tabular}{c}UAR \\ $p=0.5$\end{tabular}
& \includegraphics[width=0.29\textwidth,valign=c,trim={0 0 0 0},clip]{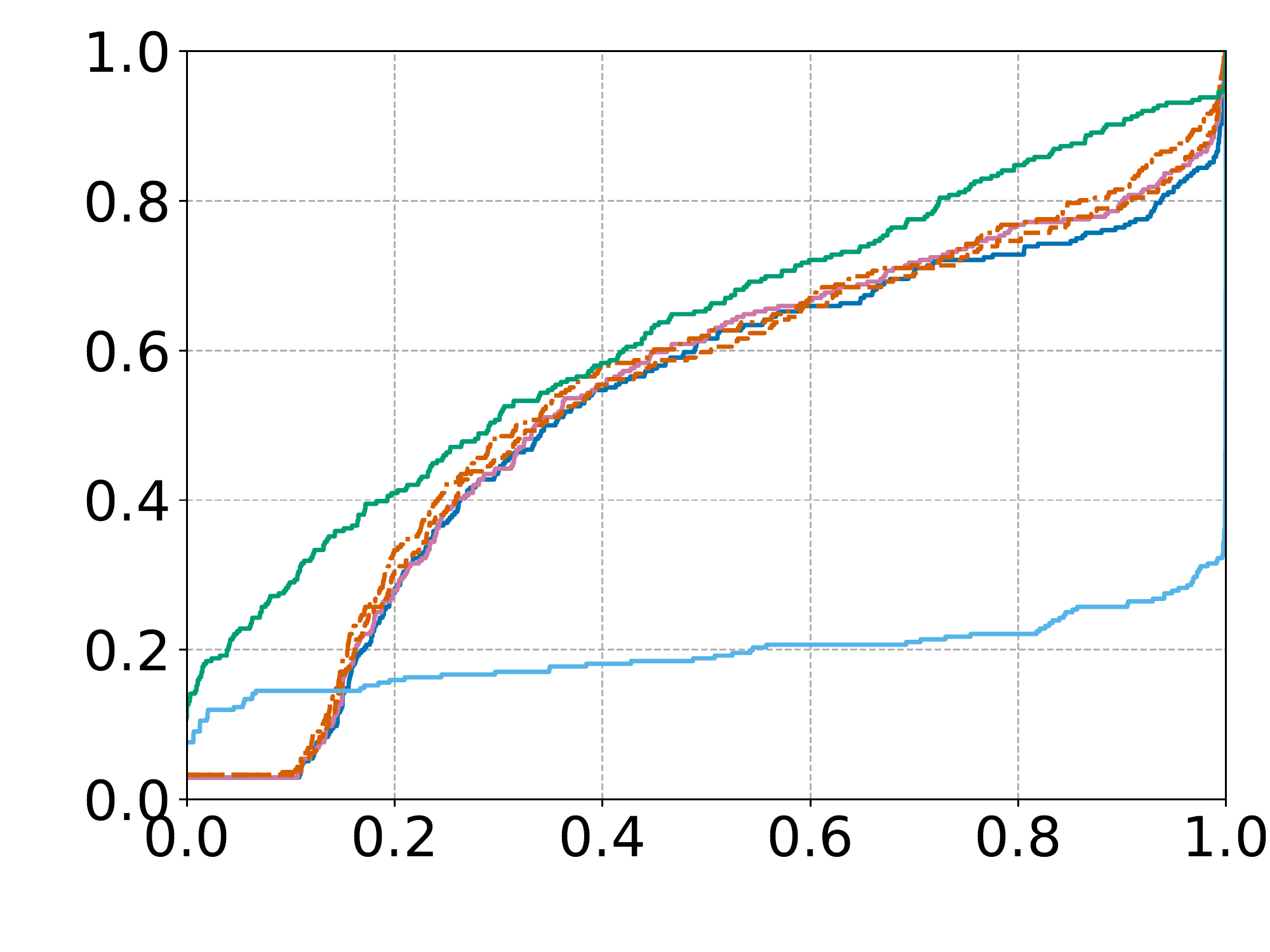}
&
\includegraphics[width=0.29\textwidth,valign=c,trim={0 0 0 0},clip]{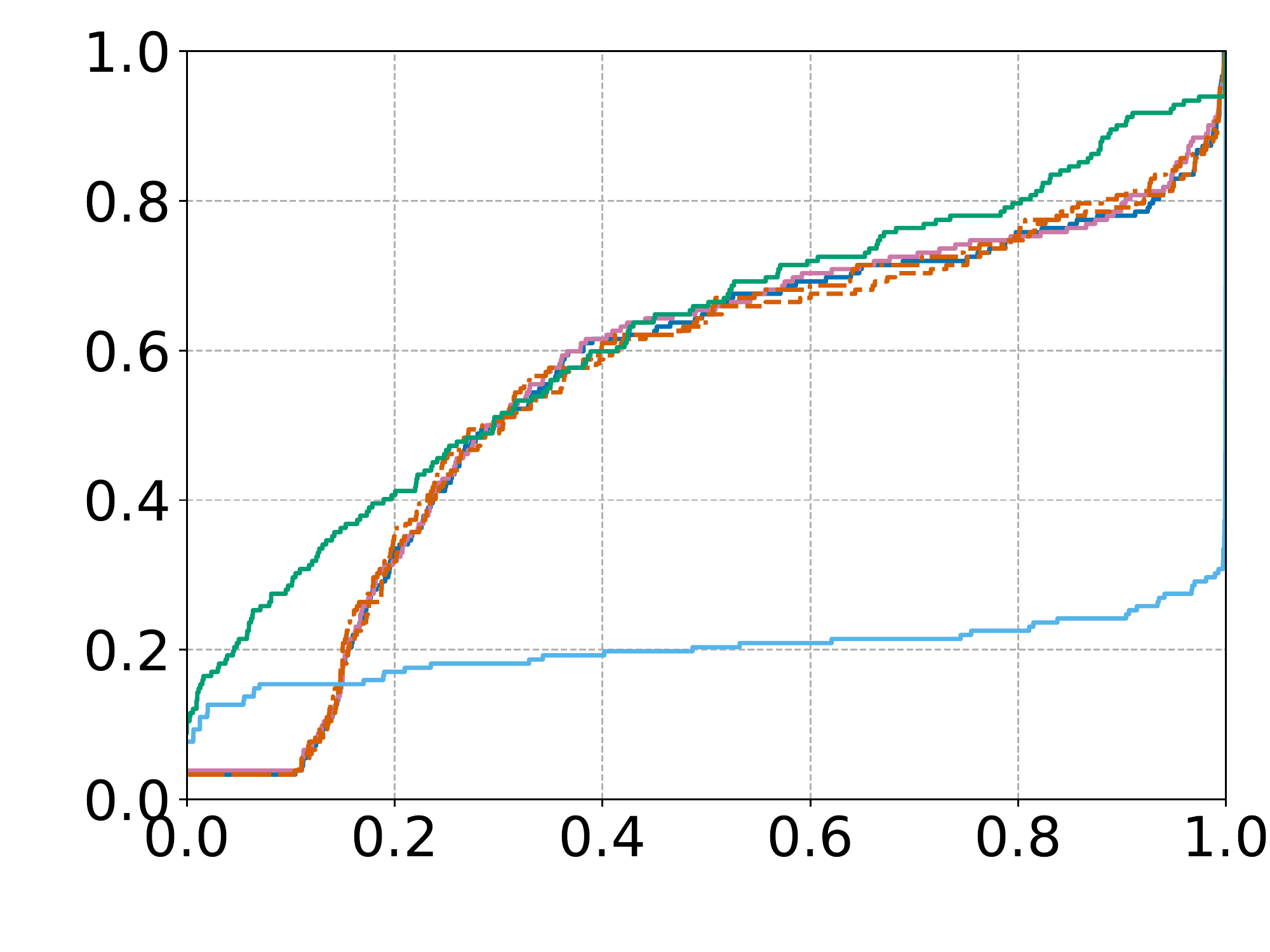}
&
\includegraphics[width=0.29\textwidth,valign=c,trim={0 0 0 0},clip]{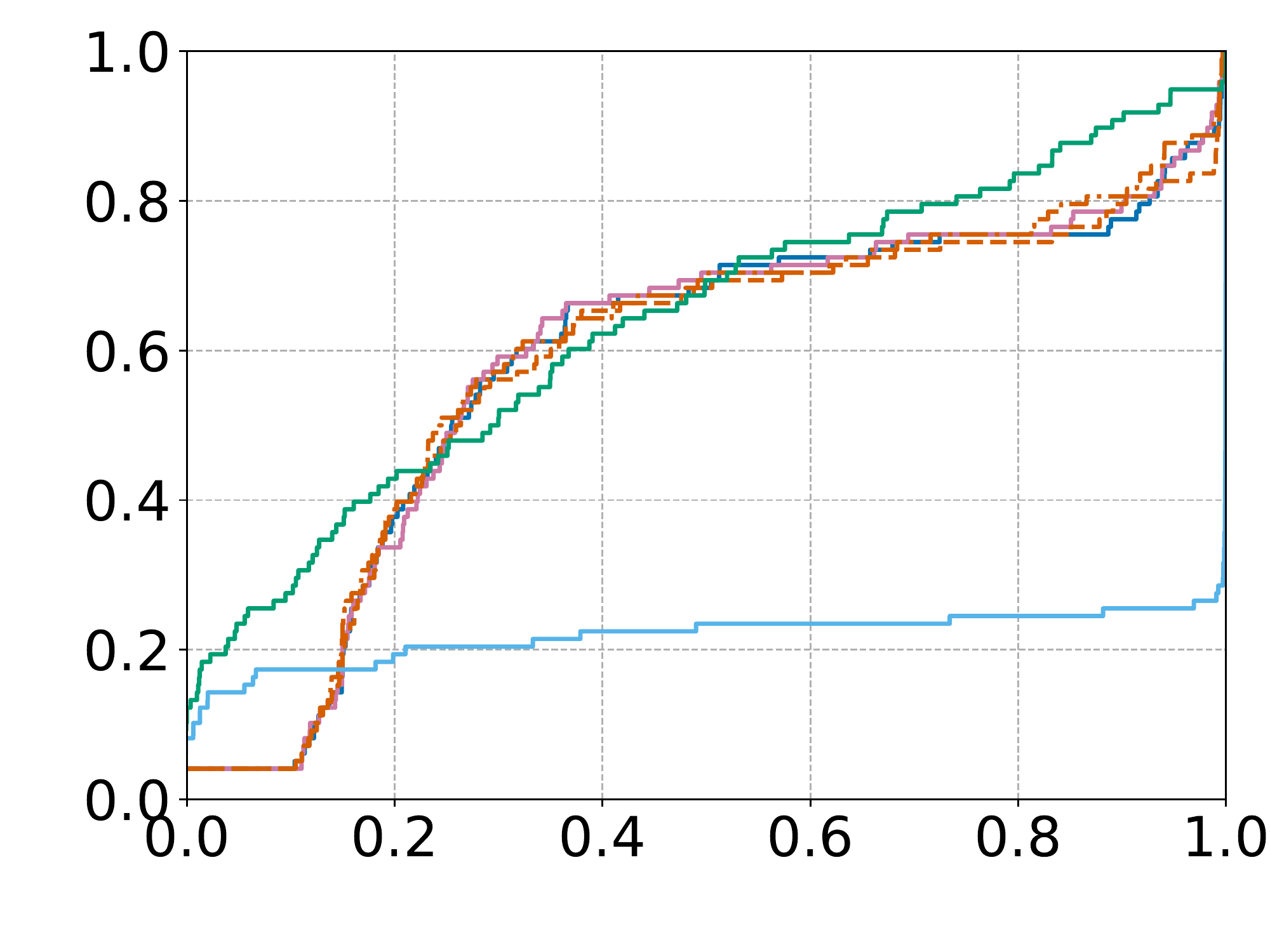}
\\\bottomrule
\multicolumn{4}{c}{
		\begin{subfigure}[t]{\textwidth}
    \includegraphics[width=\textwidth,trim={0 0 0 0},clip]{figs/cdfs/legends/legend_Lambda=8.pdf}
		\end{subfigure} }
\end{tabular}
\caption{Cumulative distribution functions (CDFs) for all evaluated algorithms, against different noise settings and warm start ratios in $\cbr{23.0,46.0,92.0}$. All CB algorithms use $\epsilon$-greedy with $\epsilon=0.1$. In each of the above plots, the $x$ axis represents scores, while the $y$ axis represents the CDF values.}
\label{fig:cdfs-eps=0.1-2}
\end{figure}

\begin{figure}[H]
\centering
\begin{tabular}{c |@{}c@{}} 
\toprule
& \multicolumn{1}{c}{ Ratio }
\\
Noise & 184.0
\\\midrule
Noiseless & \includegraphics[width=0.3\textwidth,valign=c,trim={0 0 0 0},clip]{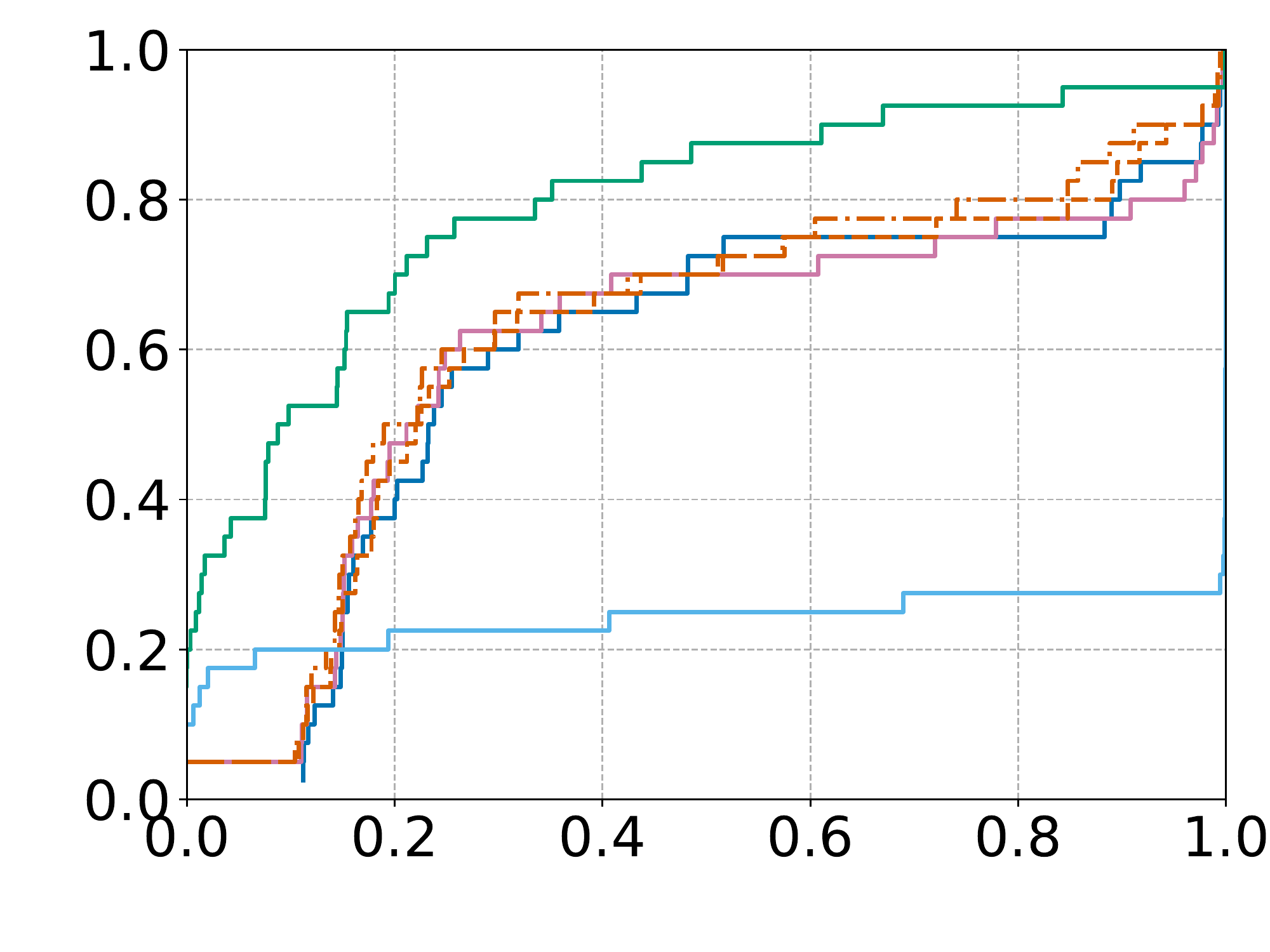}
\\\hline
\begin{tabular}{c}UAR \\ $p=0.25$\end{tabular}
& \includegraphics[width=0.3\textwidth,valign=c,trim={0 0 0 0},clip]{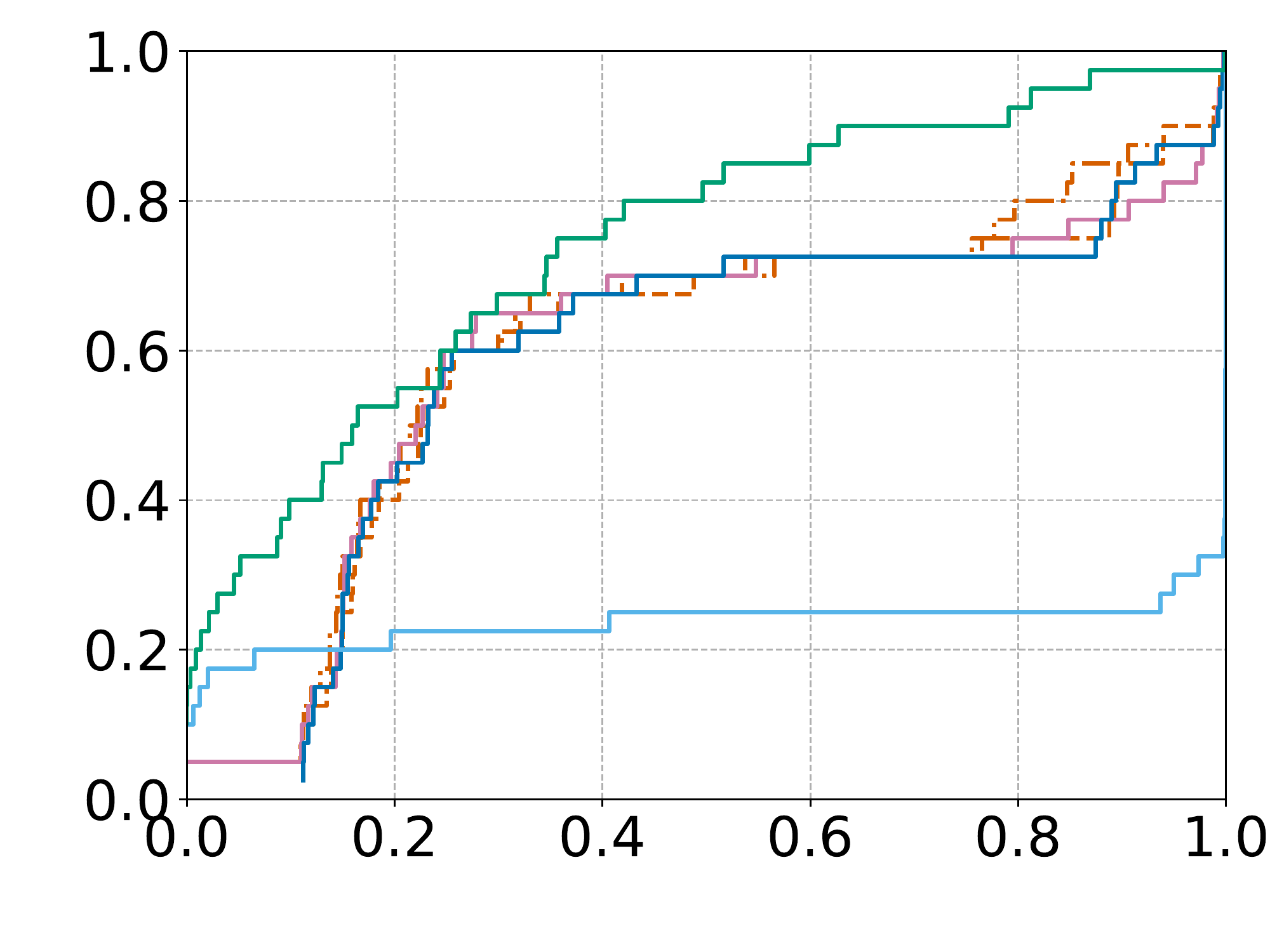}

\\\hline
\begin{tabular}{c}MAJ \\ $p=0.25$\end{tabular}
& \includegraphics[width=0.3\textwidth,valign=c,trim={0 0 0 0},clip]{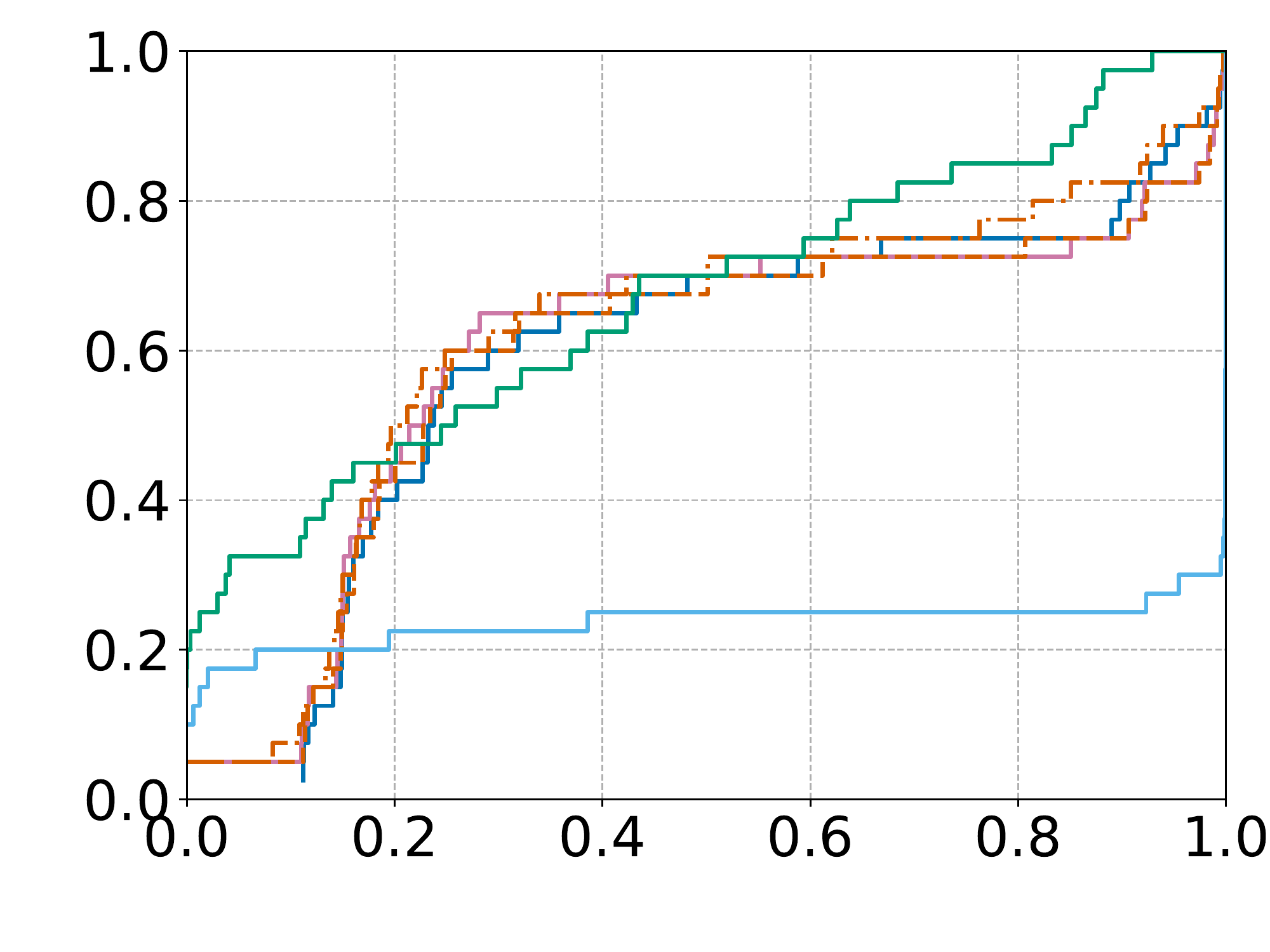}

\\\hline
\begin{tabular}{c}CYC \\ $p=0.25$\end{tabular}
& \includegraphics[width=0.3\textwidth,valign=c,trim={0 0 0 0},clip]{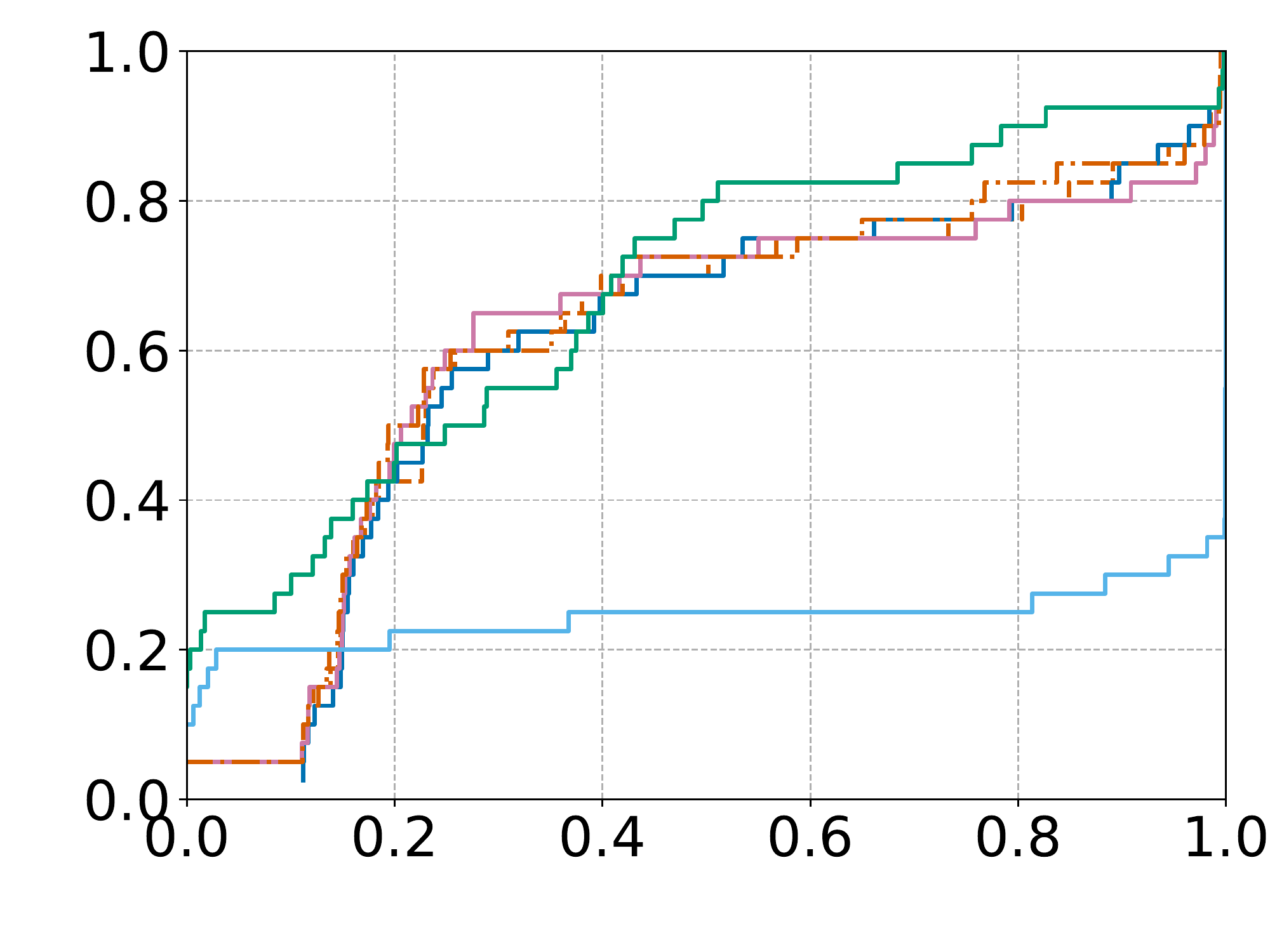}

\\\hline
\begin{tabular}{c}UAR \\ $p=0.5$\end{tabular}
& \includegraphics[width=0.3\textwidth,valign=c,trim={0 0 0 0},clip]{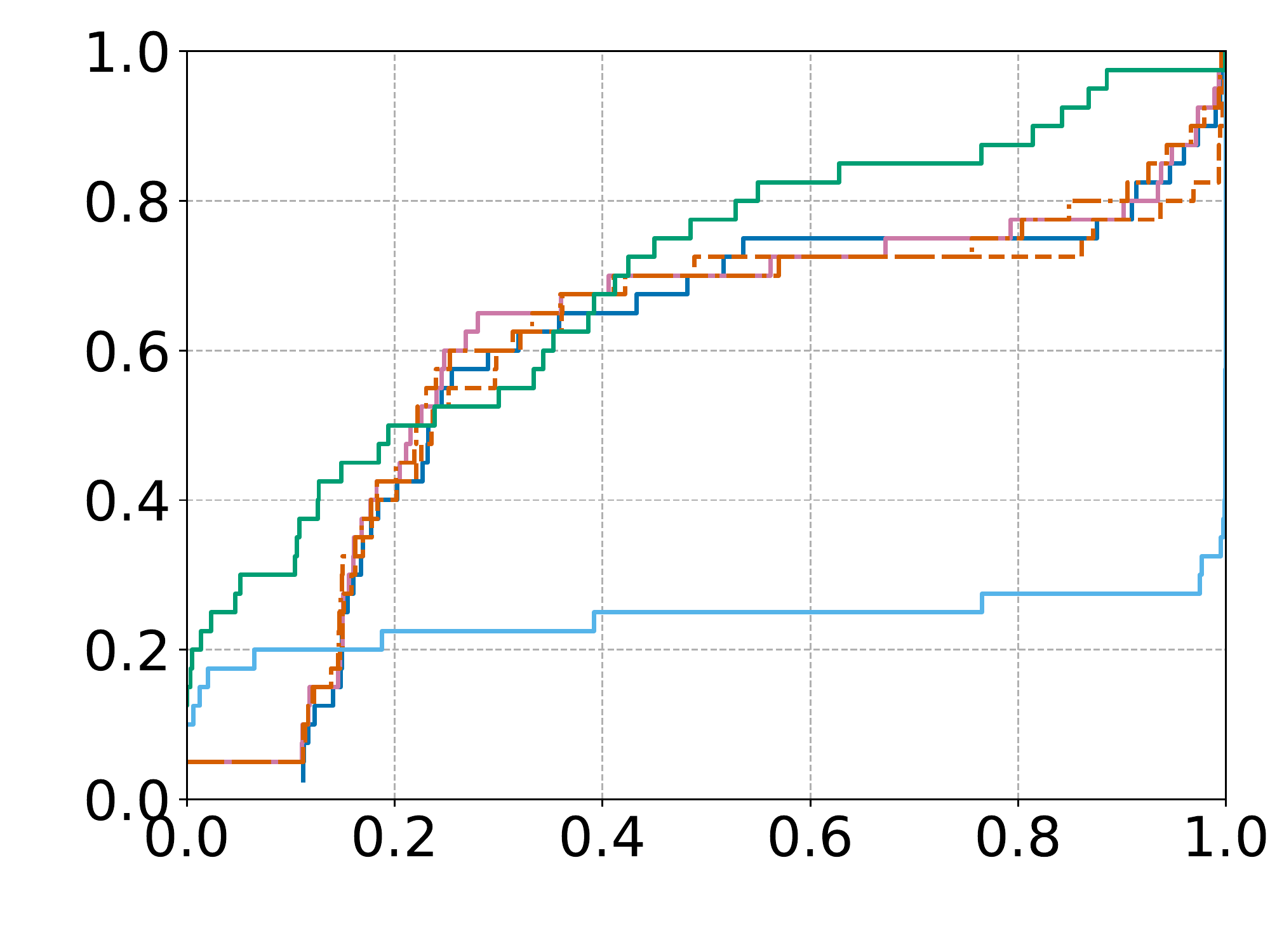}

\\\bottomrule
\multicolumn{2}{c}{
		\begin{subfigure}[t]{\textwidth}
    \includegraphics[width=\textwidth,trim={0 0 0 0},clip]{figs/cdfs/legends/legend_Lambda=8.pdf}
		\end{subfigure} }
\end{tabular}
\caption{Cumulative distribution functions (CDFs) for all evaluated algorithms, against different noise settings and warm start ratios in $\cbr{184.0}$. All CB algorithms use $\epsilon$-greedy with $\epsilon=0.1$. In each of the above plots, the $x$ axis represents scores, while the $y$ axis represents the CDF values.}
\label{fig:cdfs-eps=0.1-3}
\end{figure}

\begin{figure}[H]
\centering
\begin{tabular}{c | @{}c@{ }c@{ }c@{}} 
\toprule
& \multicolumn{3}{c}{ Ratio }
\\
Noise & 2.875 & 5.75 & 11.5
\\\midrule
\begin{tabular}{c}MAJ \\ $p=0.5$\end{tabular}
 & \includegraphics[width=0.29\textwidth,valign=c,trim={0 0 0 0},clip]{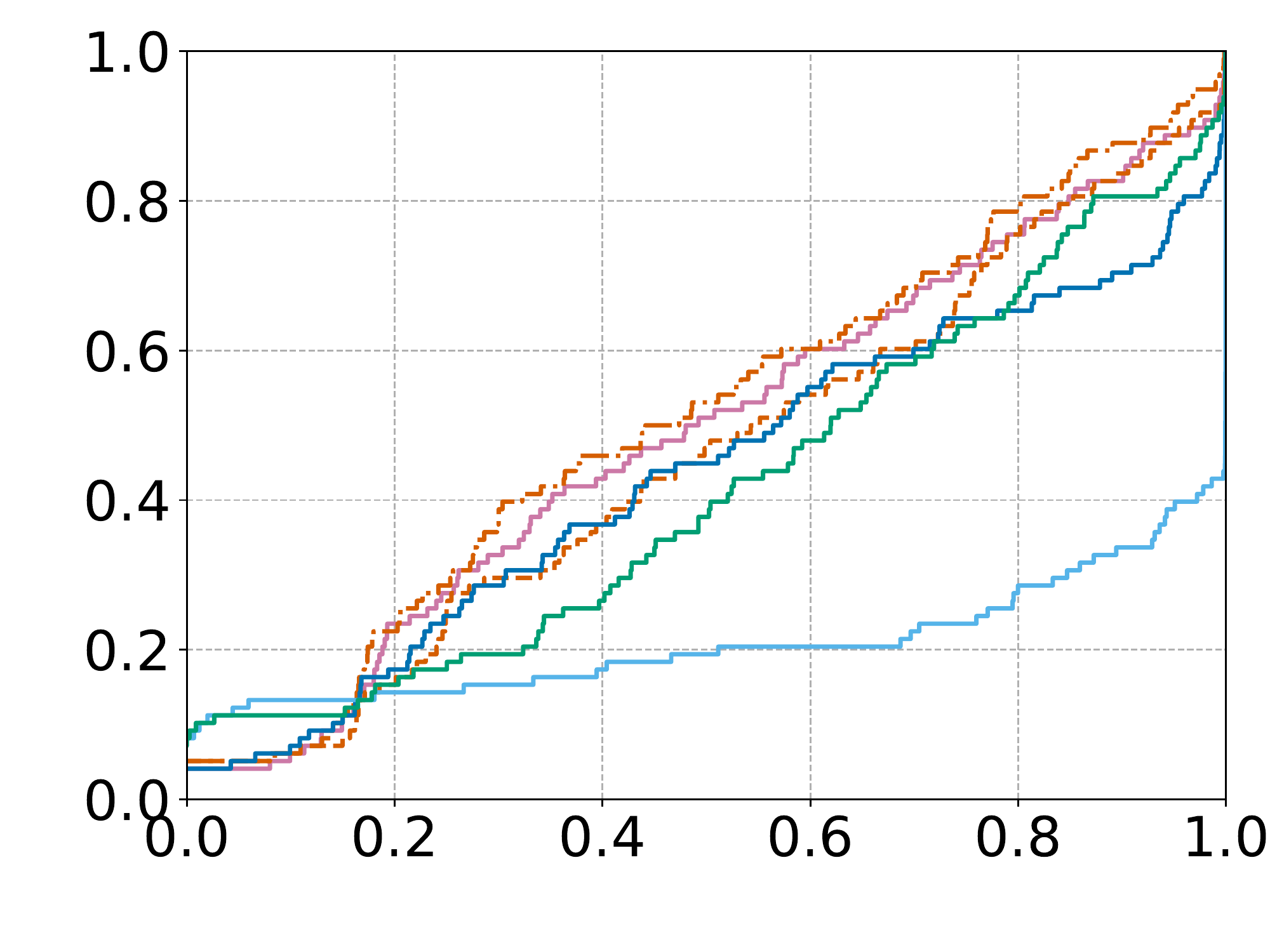}
&
\includegraphics[width=0.29\textwidth,valign=c,trim={0 0 0 0},clip]{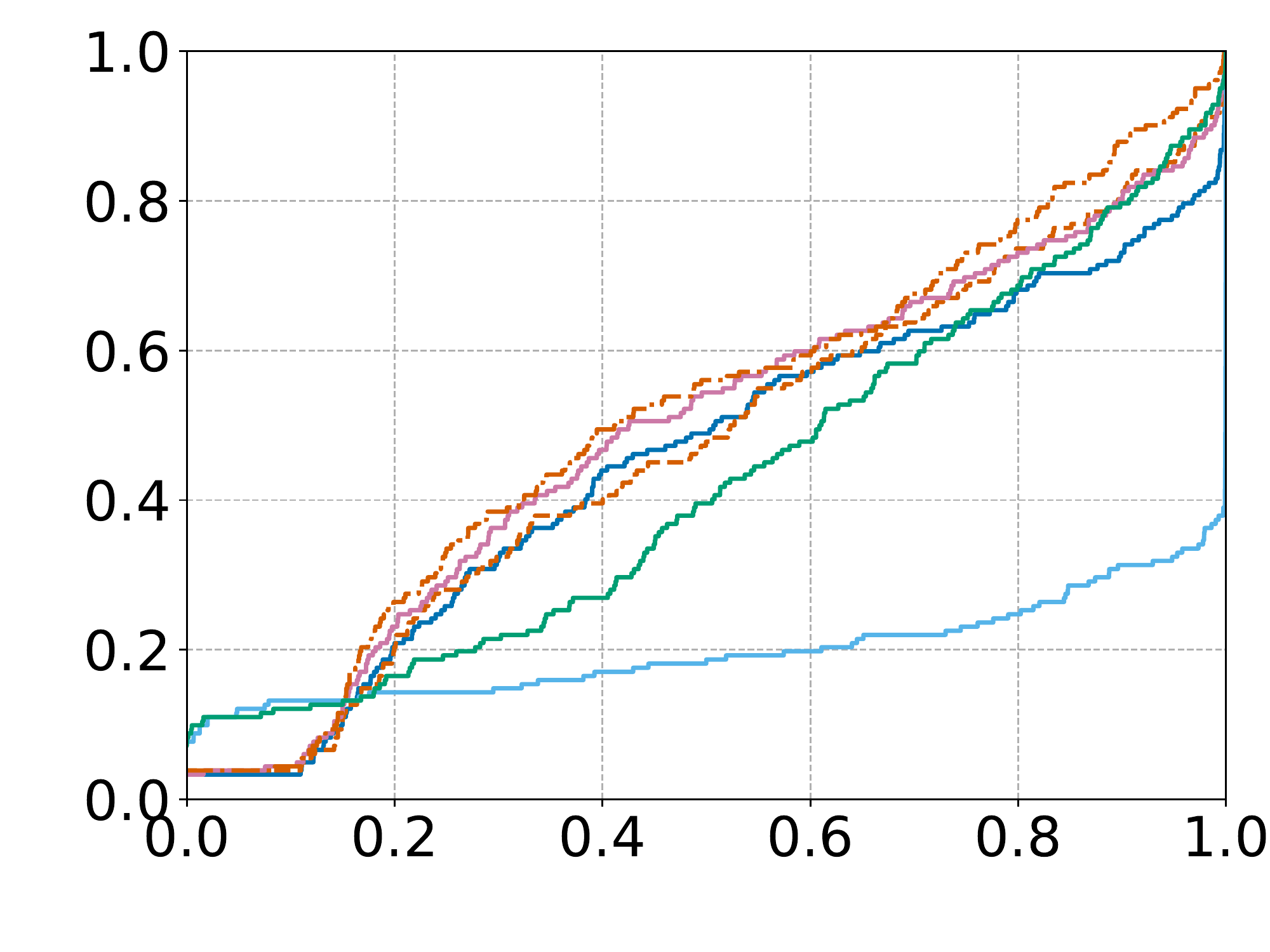}
&
\includegraphics[width=0.29\textwidth,valign=c,trim={0 0 0 0},clip]{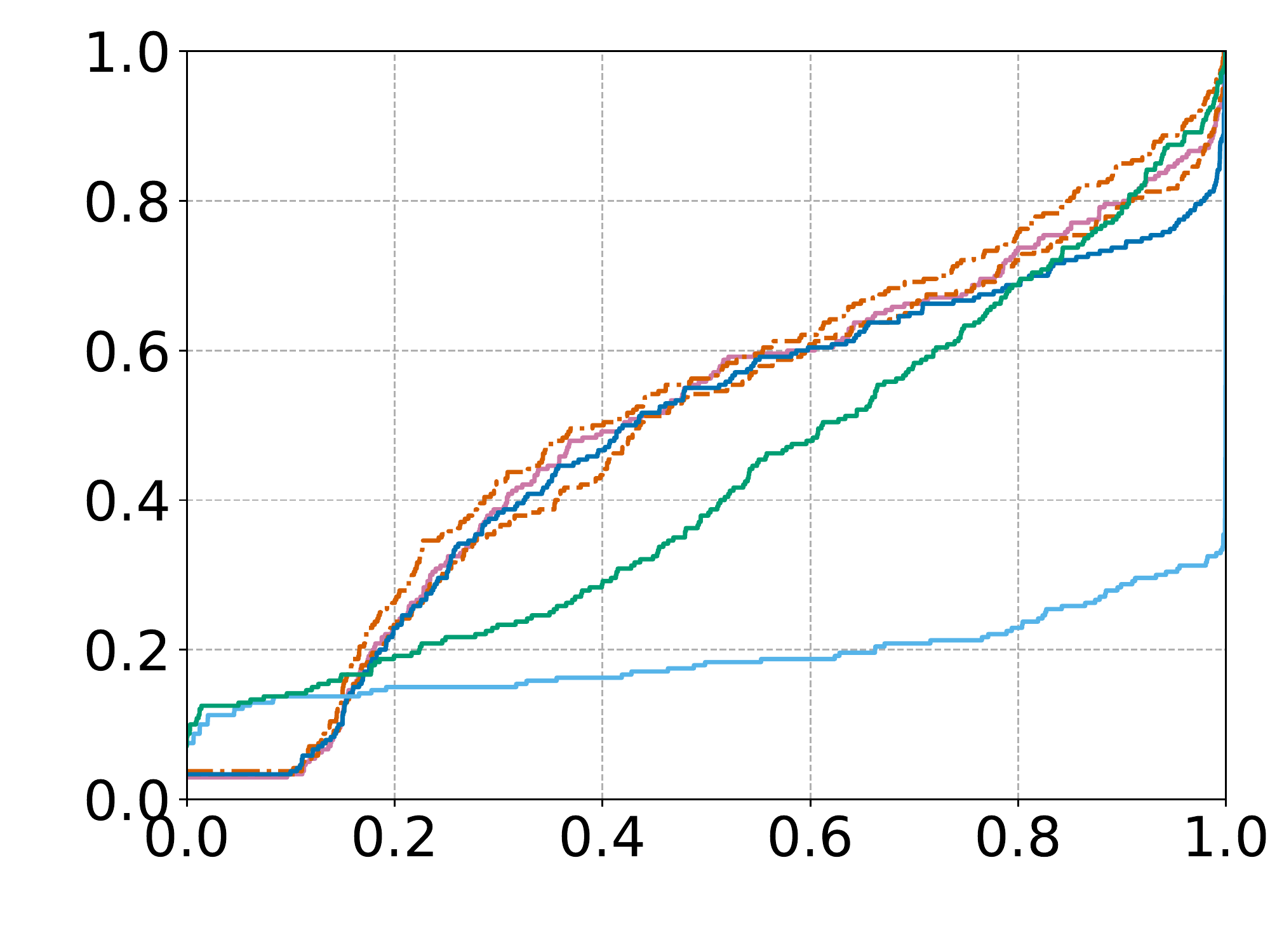}
\\\hline
\begin{tabular}{c}CYC \\ $p=0.5$\end{tabular}
& \includegraphics[width=0.29\textwidth,valign=c,trim={0 0 0 0},clip]{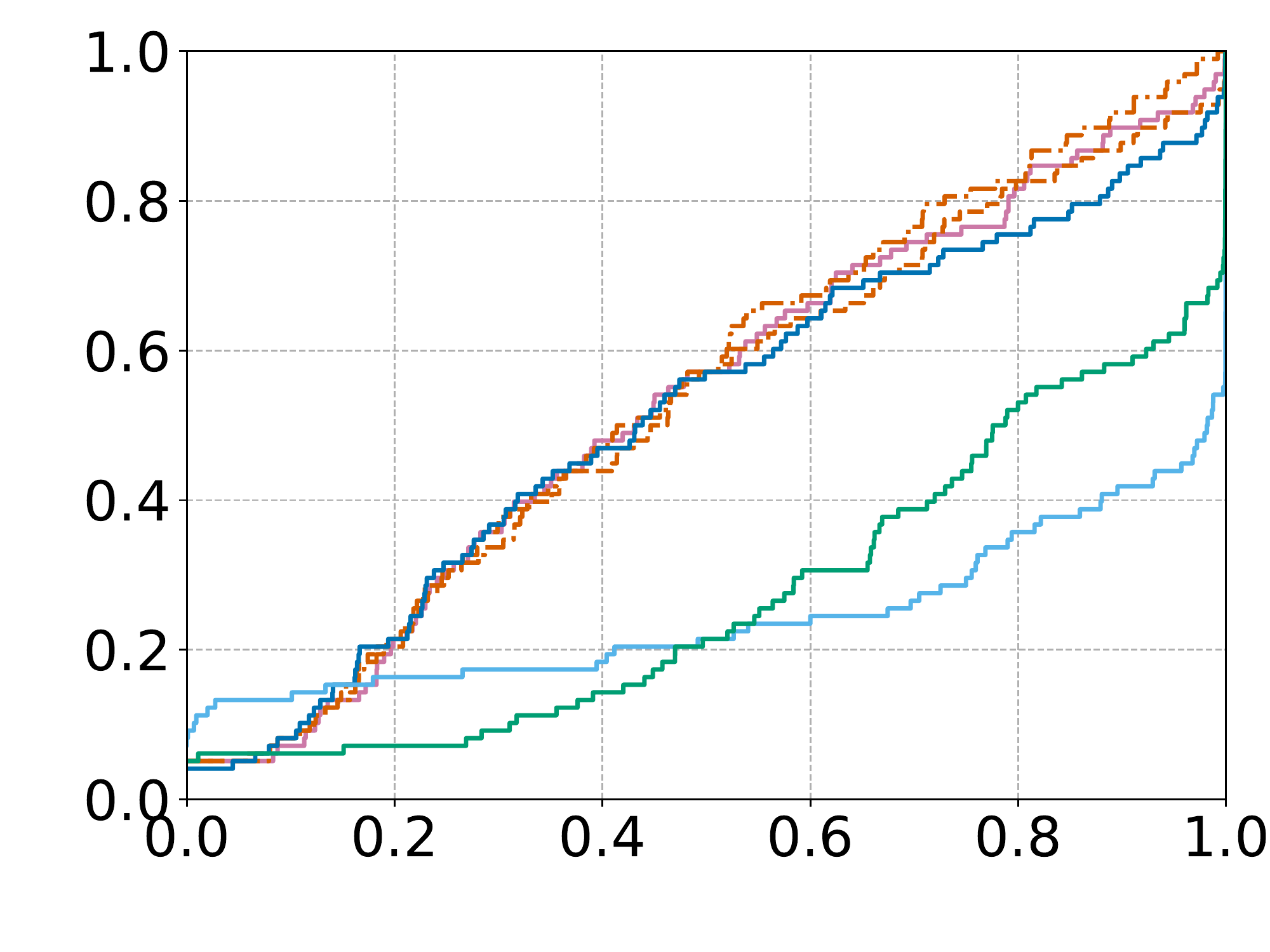}
&
\includegraphics[width=0.29\textwidth,valign=c,trim={0 0 0 0},clip]{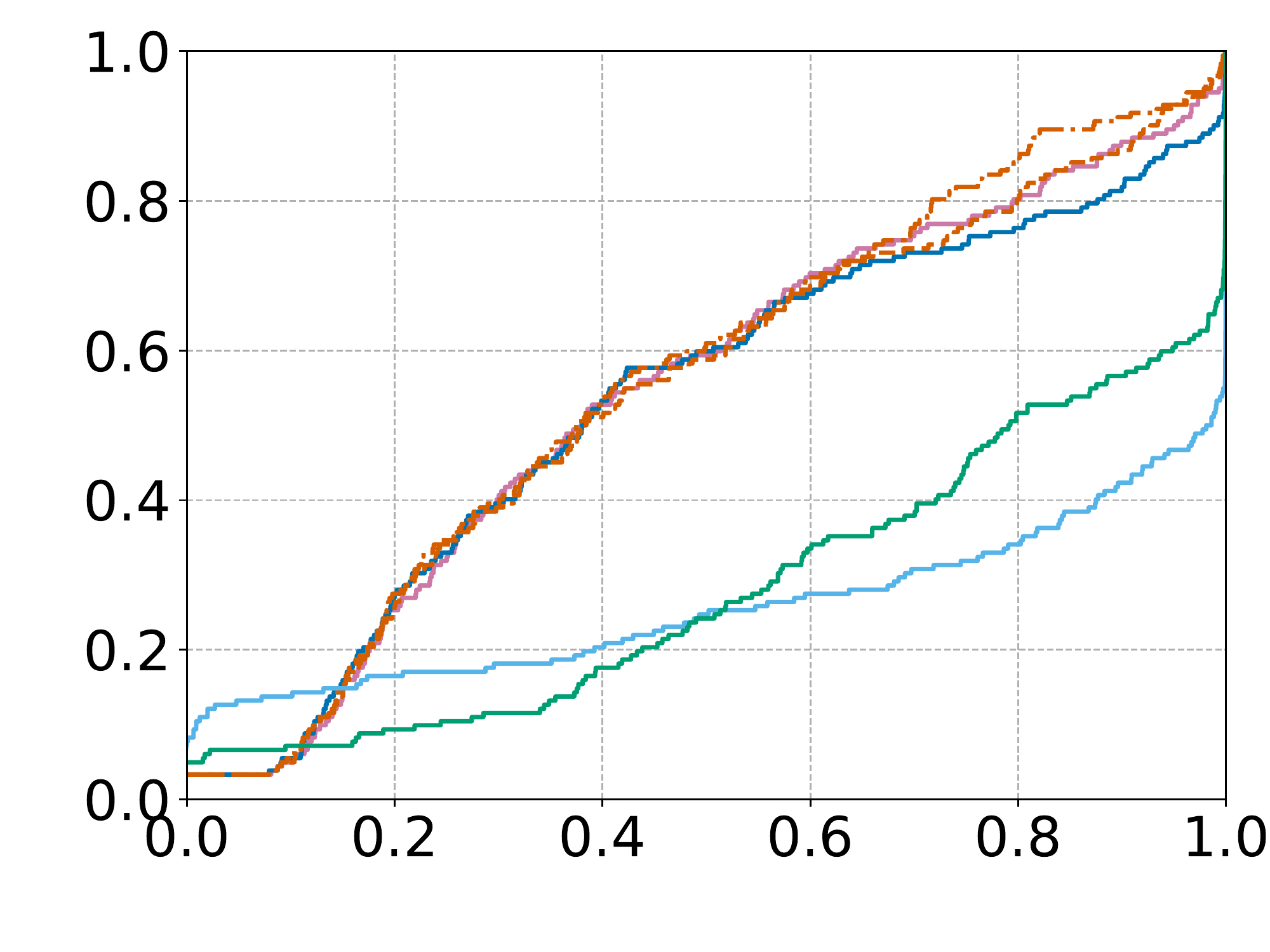}
&
\includegraphics[width=0.29\textwidth,valign=c,trim={0 0 0 0},clip]{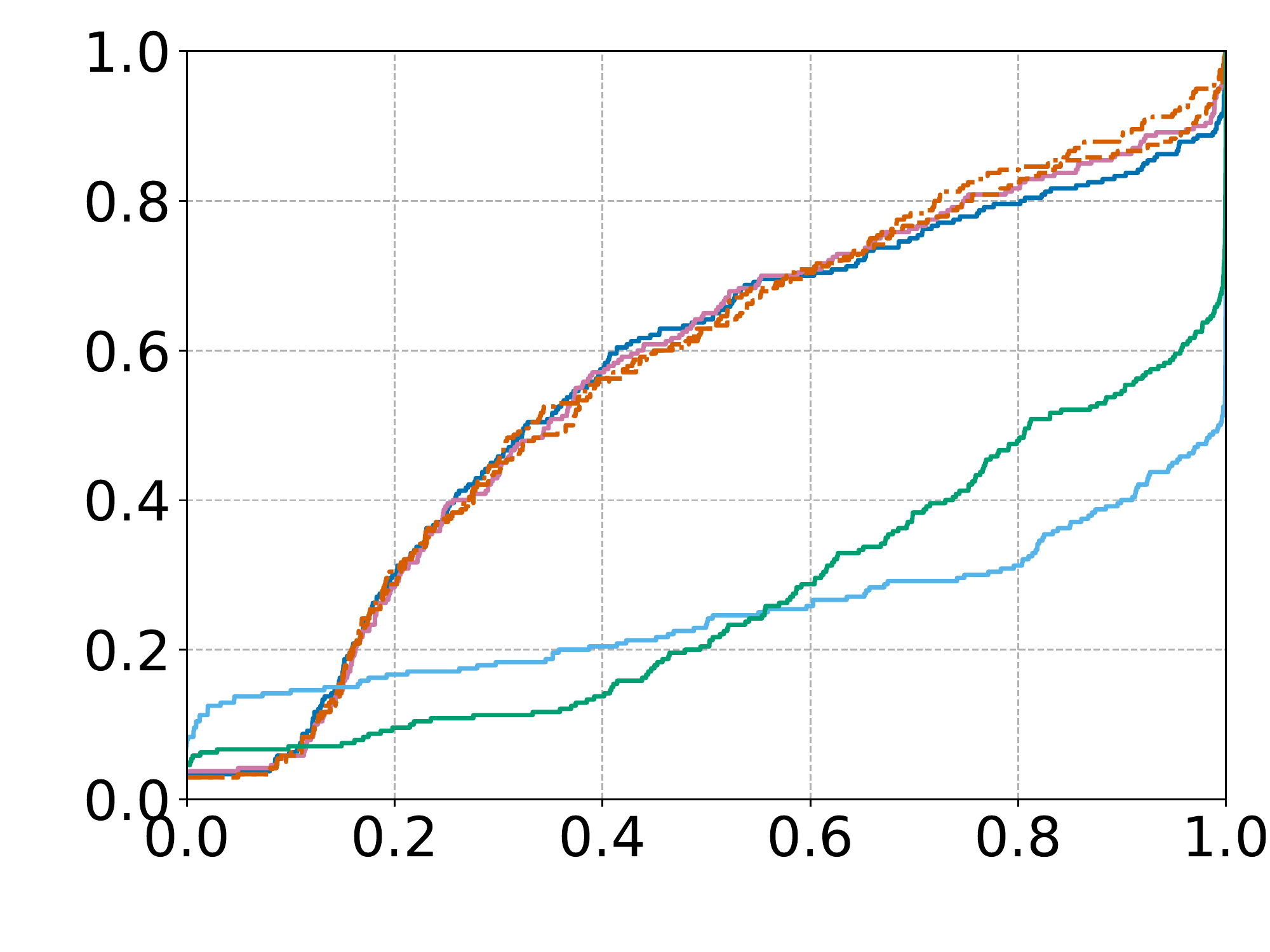}
\\\hline
\begin{tabular}{c}UAR \\ $p=1$\end{tabular}
& \includegraphics[width=0.29\textwidth,valign=c,trim={0 0 0 0},clip]{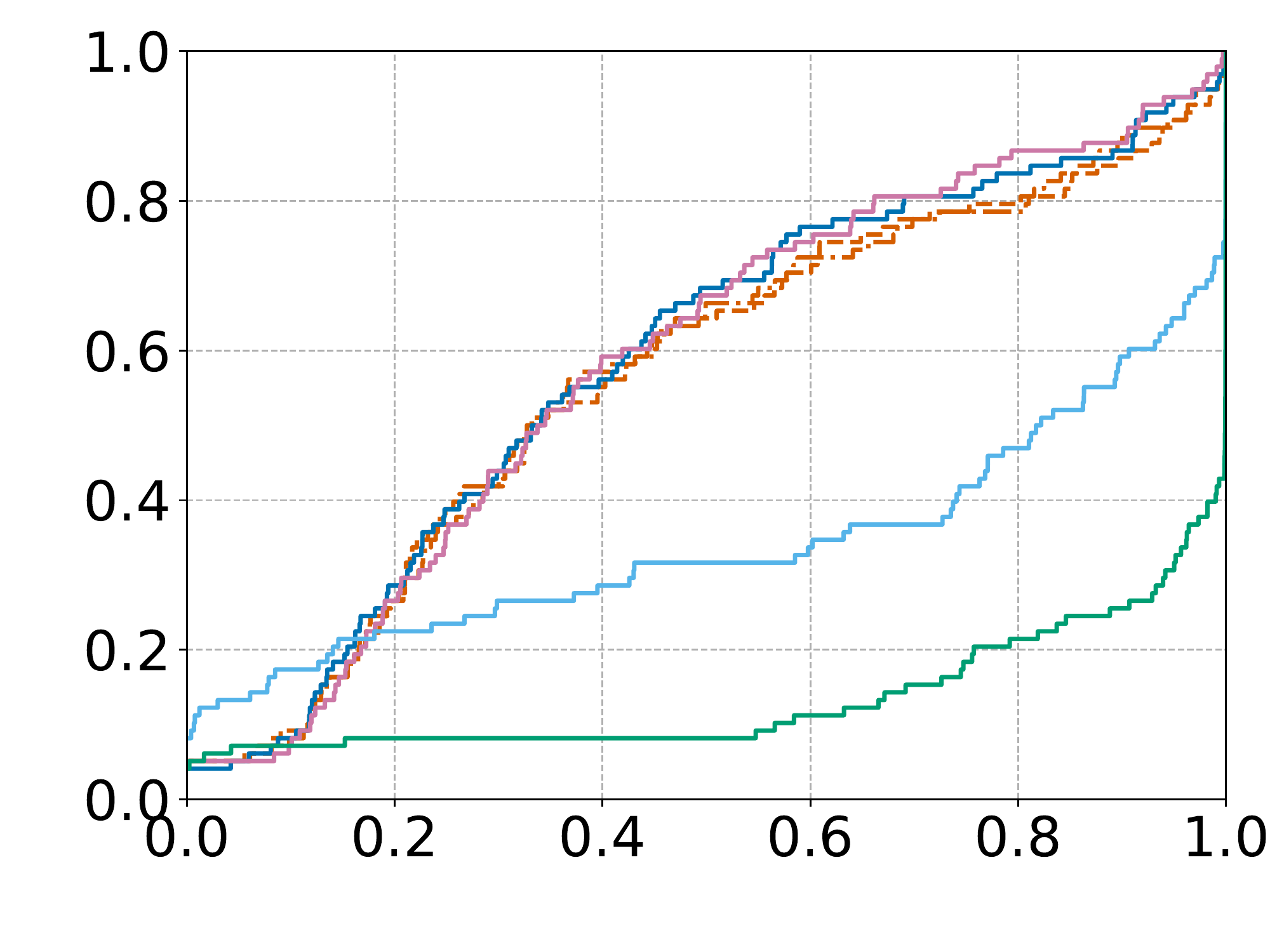}
&
\includegraphics[width=0.29\textwidth,valign=c,trim={0 0 0 0},clip]{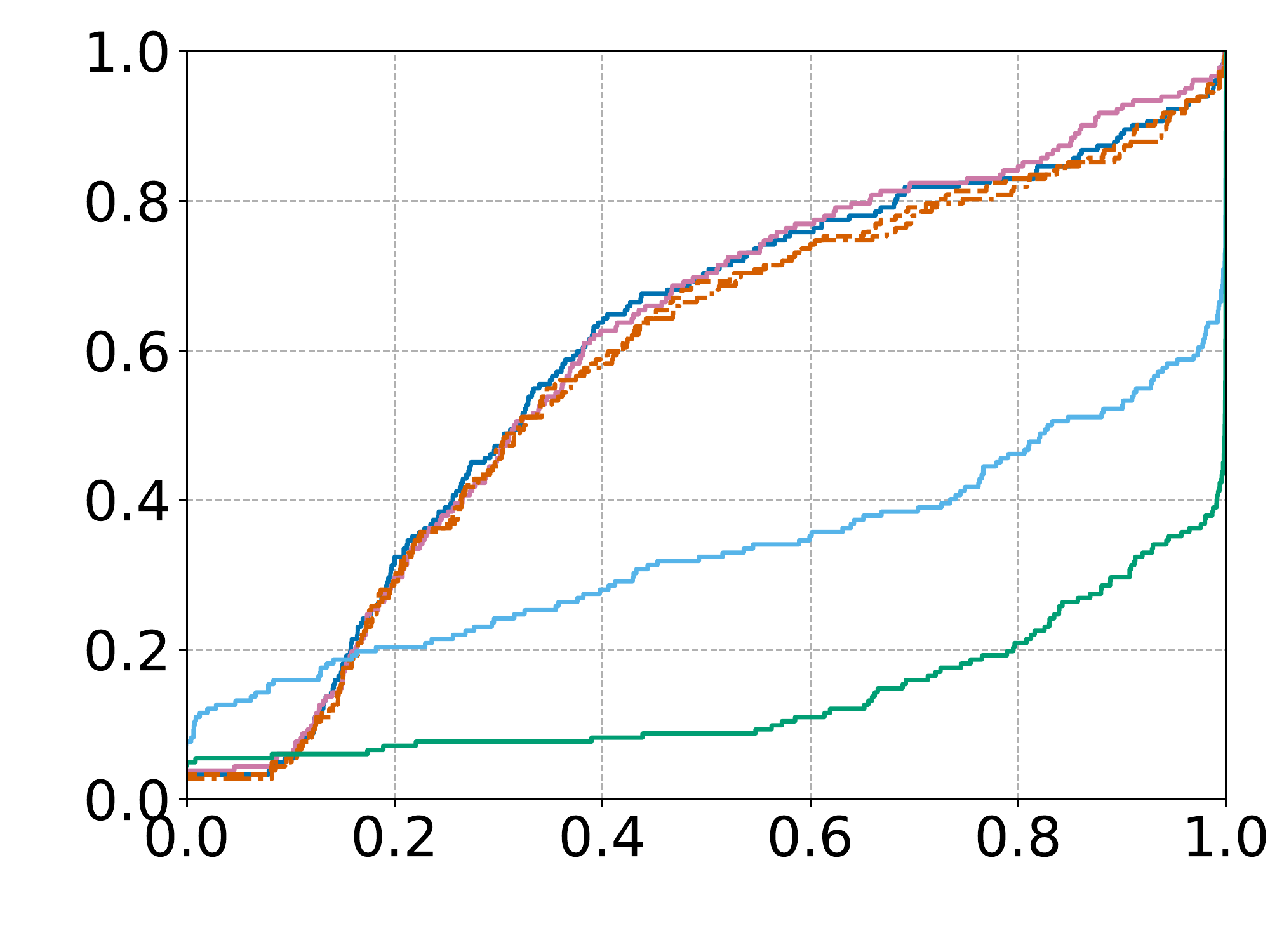}
&
\includegraphics[width=0.29\textwidth,valign=c,trim={0 0 0 0},clip]{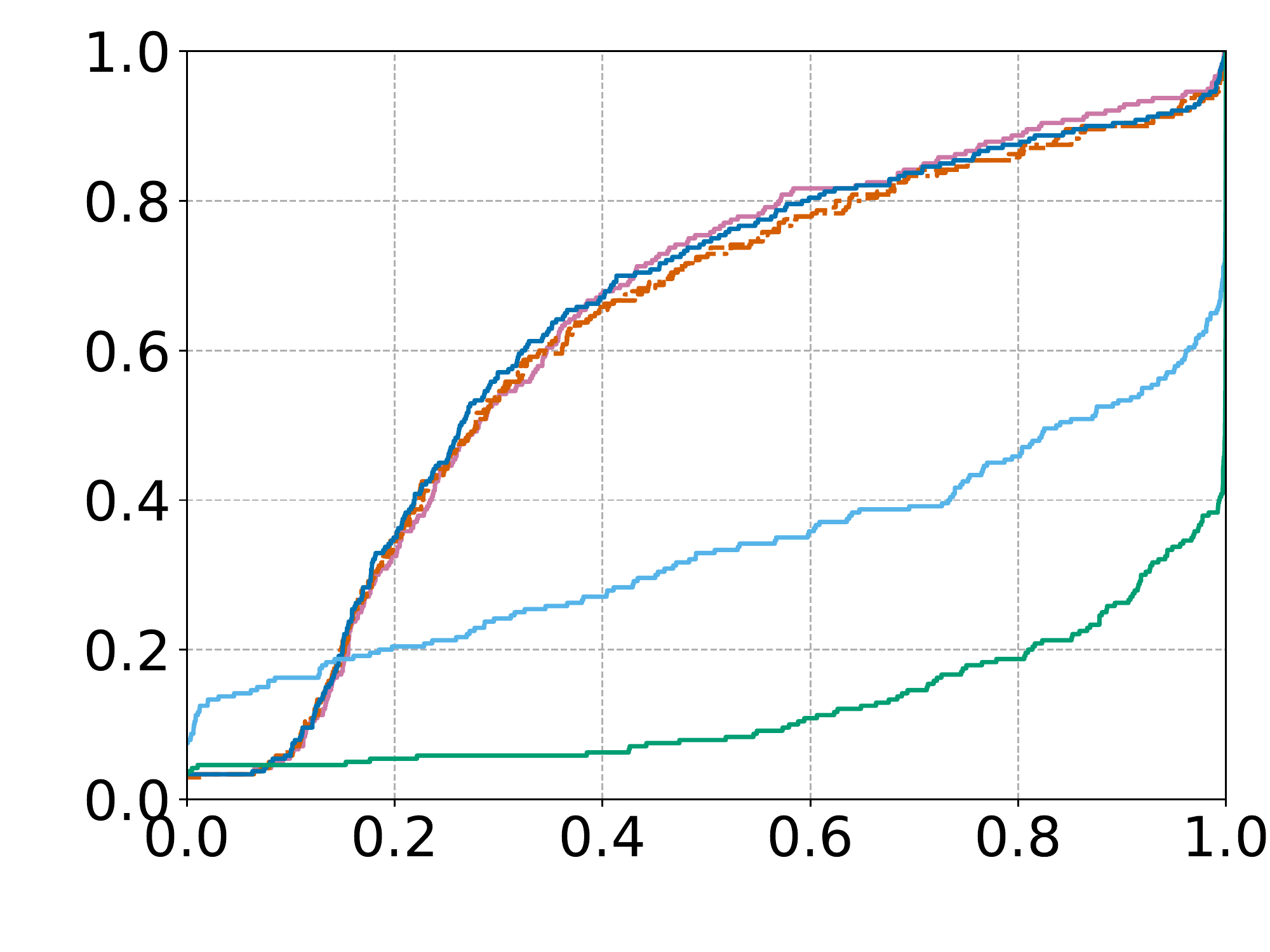}
\\\hline
\begin{tabular}{c}MAJ \\ $p=1$\end{tabular}
& \includegraphics[width=0.29\textwidth,valign=c,trim={0 0 0 0},clip]{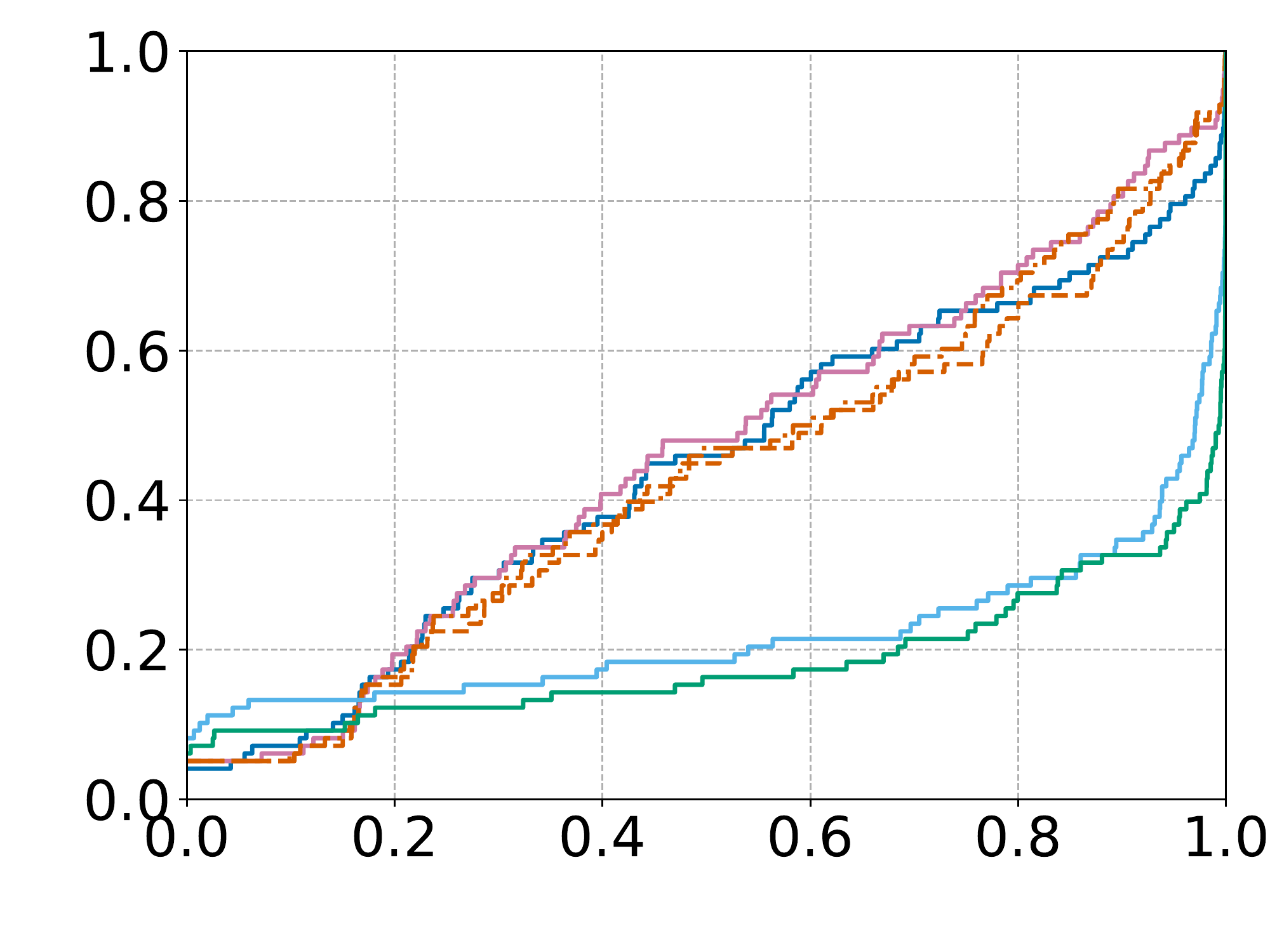}
&
\includegraphics[width=0.29\textwidth,valign=c,trim={0 0 0 0},clip]{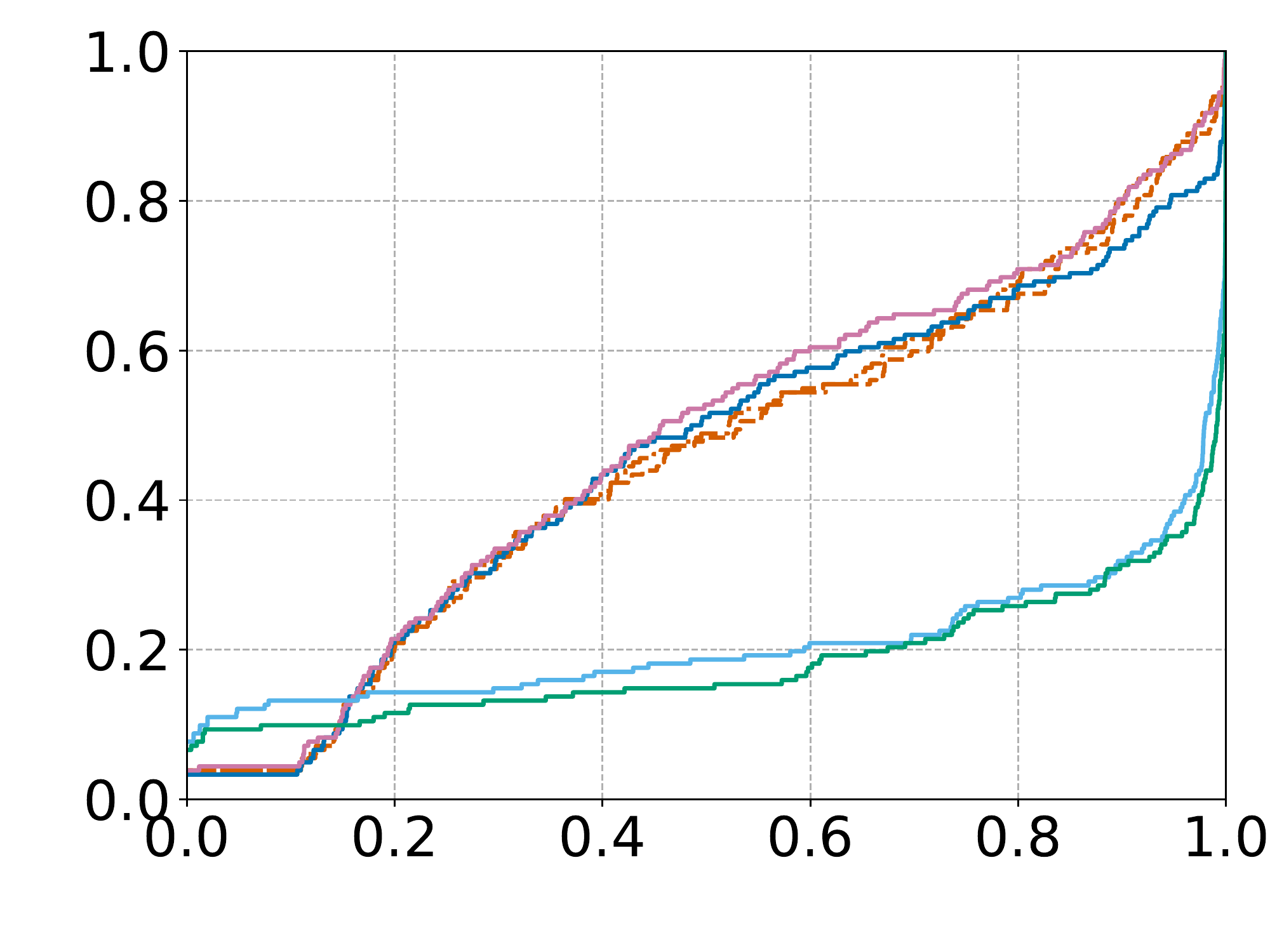}
&
\includegraphics[width=0.29\textwidth,valign=c,trim={0 0 0 0},clip]{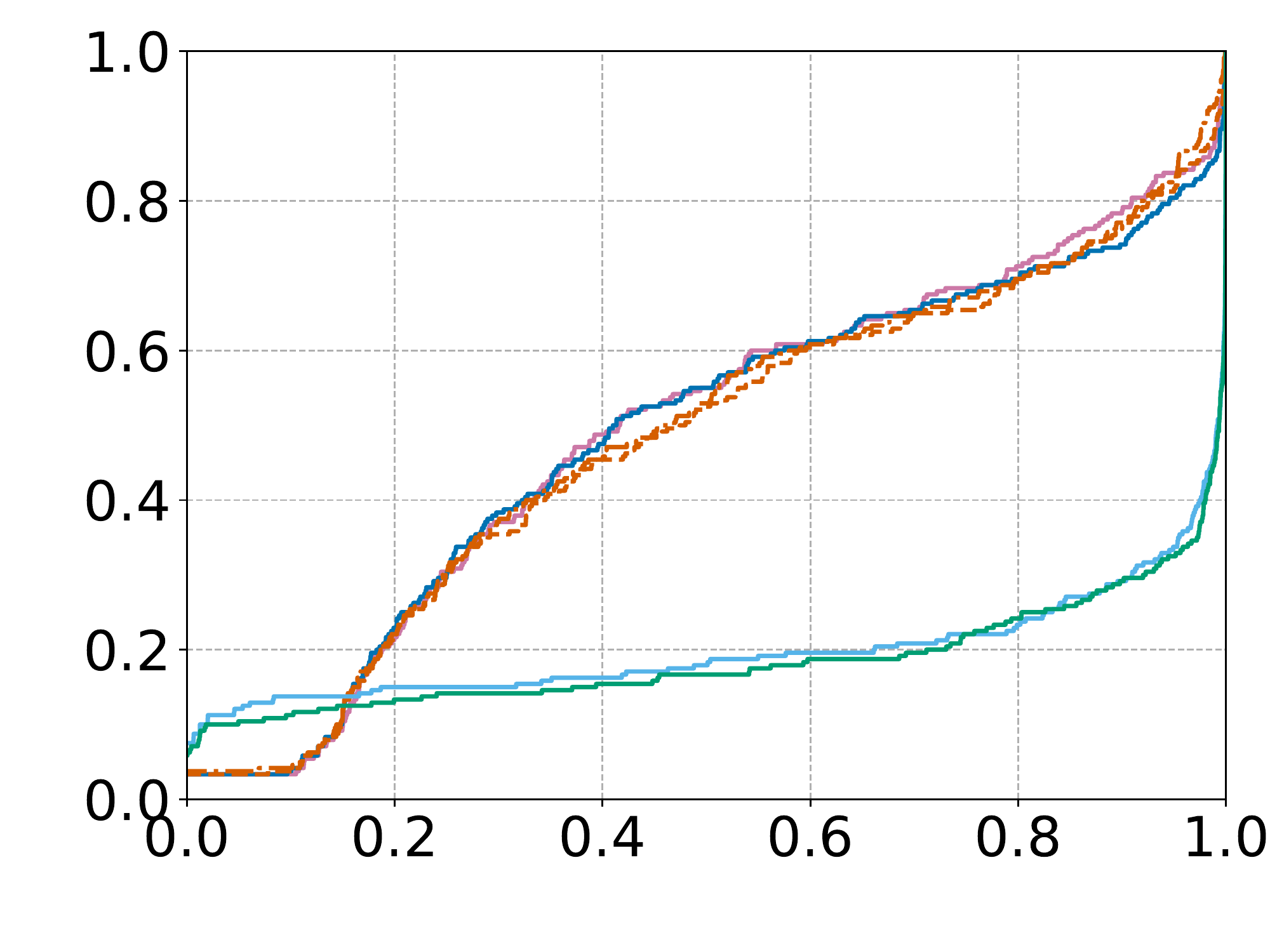}
\\\hline
\begin{tabular}{c}CYC \\ $p=1$\end{tabular}
& \includegraphics[width=0.29\textwidth,valign=c,trim={0 0 0 0},clip]{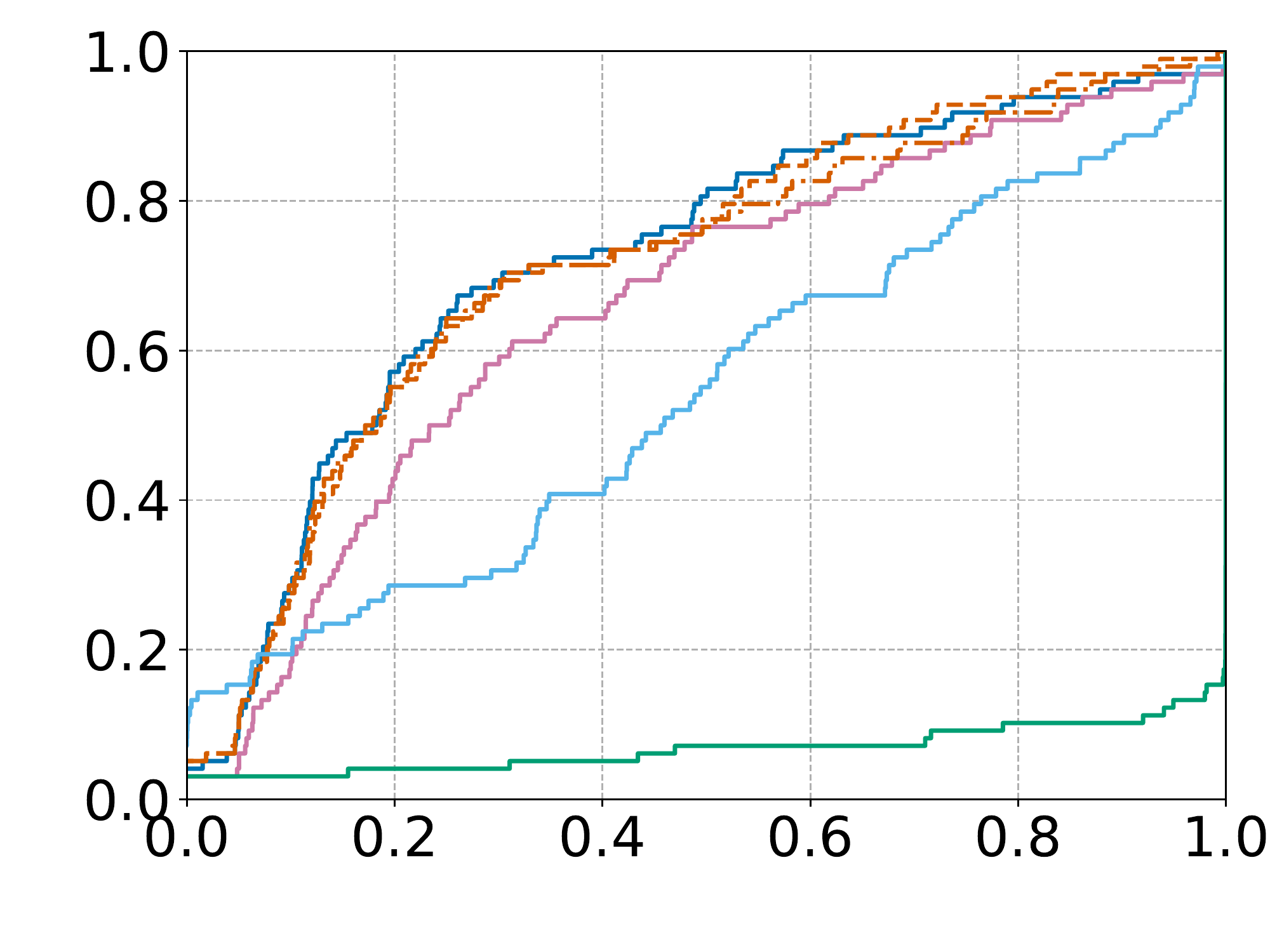}
&
\includegraphics[width=0.29\textwidth,valign=c,trim={0 0 0 0},clip]{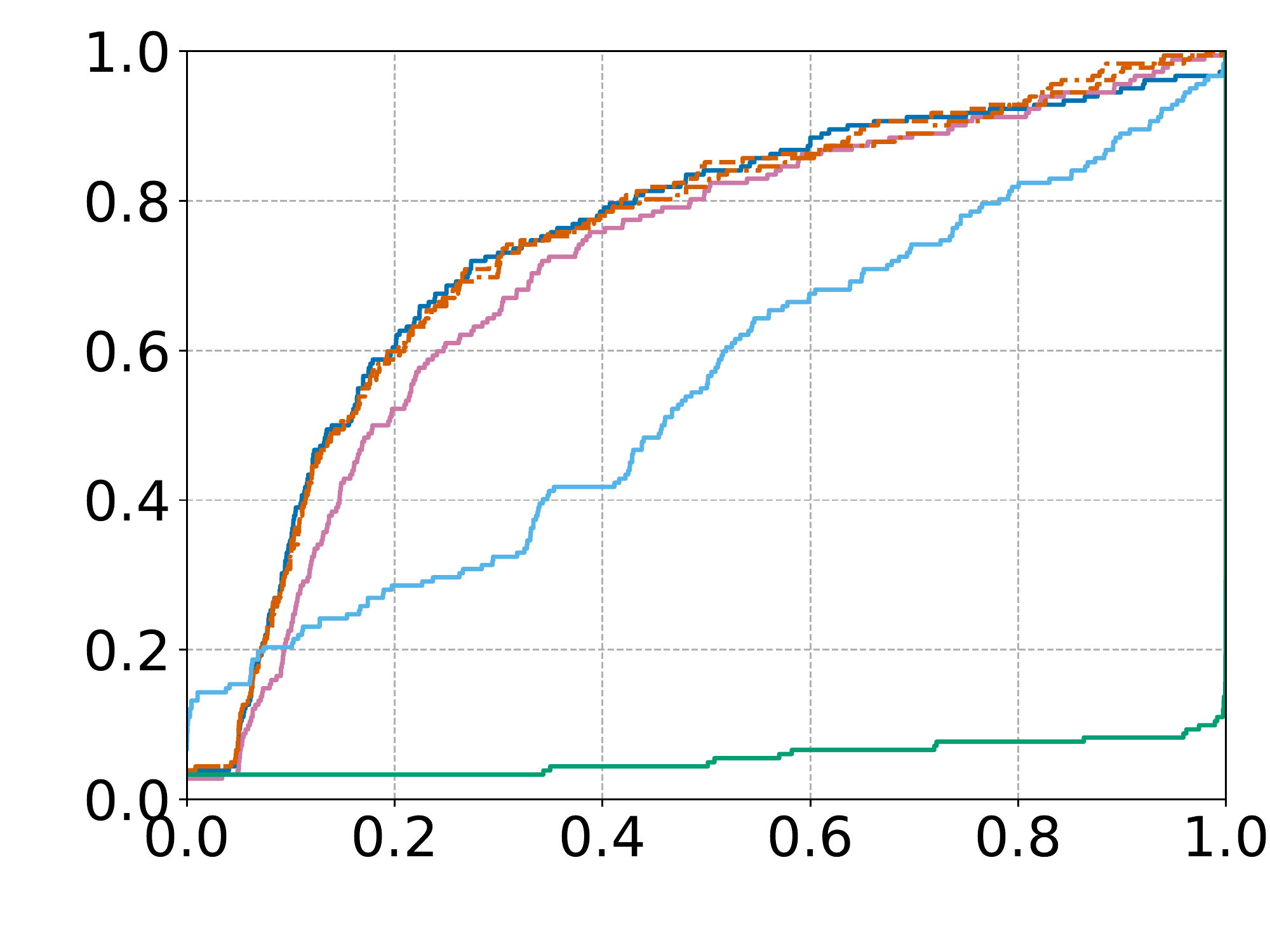}
&
\includegraphics[width=0.29\textwidth,valign=c,trim={0 0 0 0},clip]{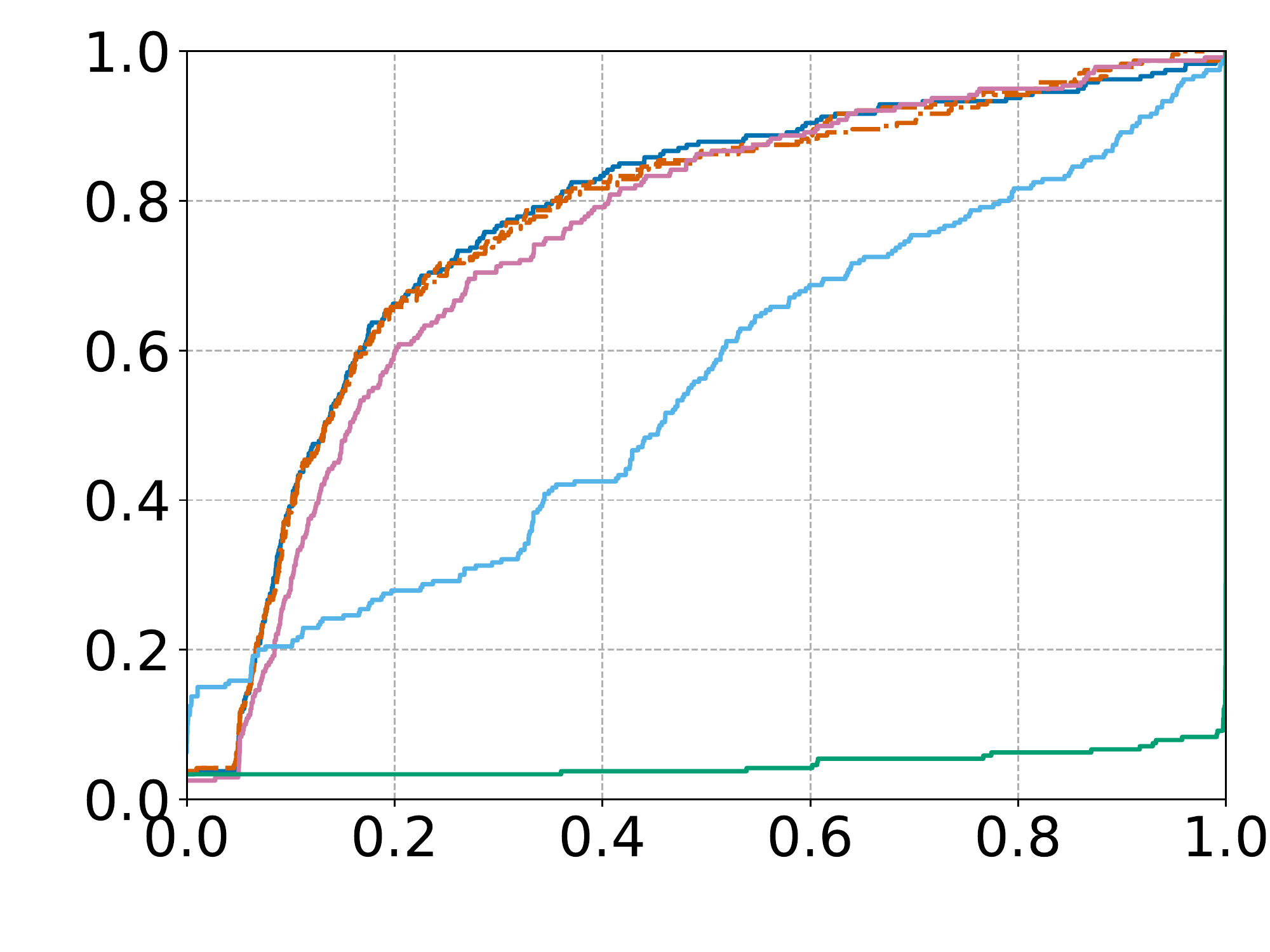}
\\\bottomrule
\multicolumn{4}{c}{
		\begin{subfigure}[t]{\textwidth}
    \includegraphics[width=\textwidth,trim={0 0 0 0},clip]{figs/cdfs/legends/legend_Lambda=8.pdf}
		\end{subfigure} }
\end{tabular}
\caption{Cumulative distribution functions (CDFs) for all evaluated algorithms, against different noise settings and warm start ratios in $\cbr{2.875,5.75,11.5}$. All CB algorithms use $\epsilon$-greedy with $\epsilon=0.1$. In each of the above plots, the $x$ axis represents scores, while the $y$ axis represents the CDF values.}
\label{fig:cdfs-eps=0.1-4}
\end{figure}

\begin{figure}[H]
\centering
\begin{tabular}{c | @{}c@{ }c@{ }c@{}} 
\toprule
& \multicolumn{3}{c}{ Ratio }
\\
Noise & 23.0 & 46.0 & 92.0
\\\midrule
\begin{tabular}{c}MAJ \\ $p=0.5$\end{tabular}
 & \includegraphics[width=0.29\textwidth,valign=c,trim={0 0 0 0},clip]{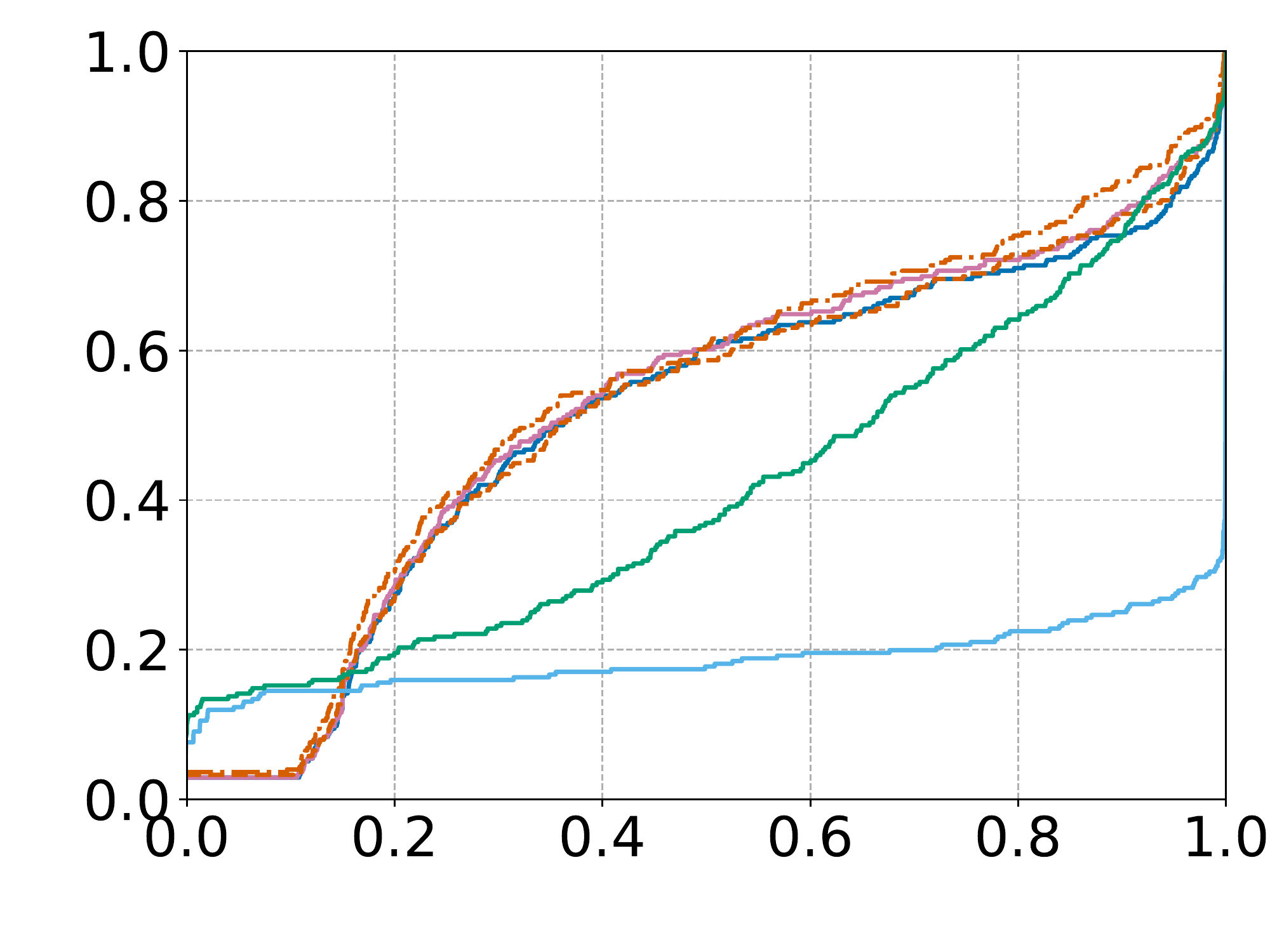}
&
\includegraphics[width=0.29\textwidth,valign=c,trim={0 0 0 0},clip]{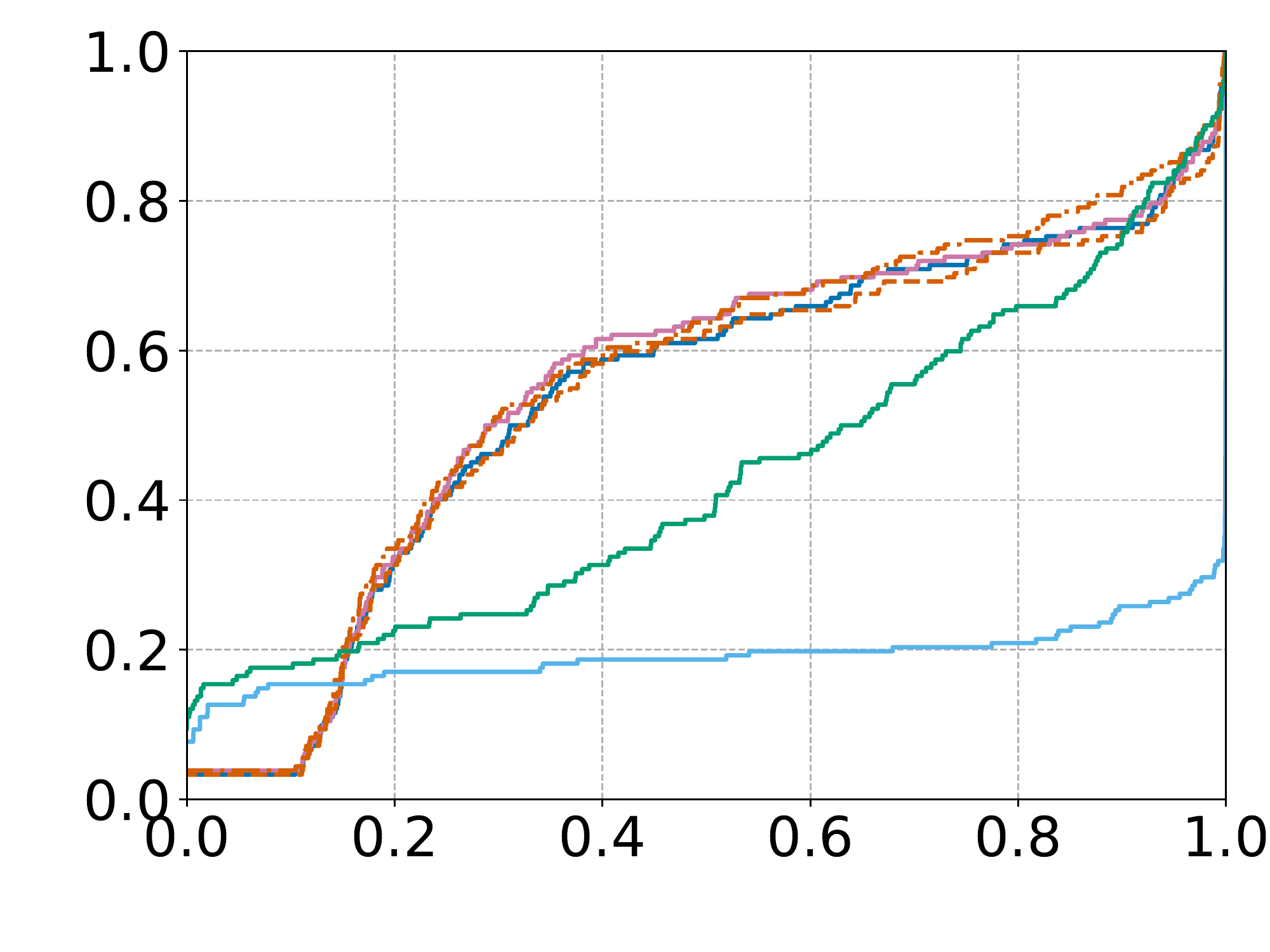}
&
\includegraphics[width=0.29\textwidth,valign=c,trim={0 0 0 0},clip]{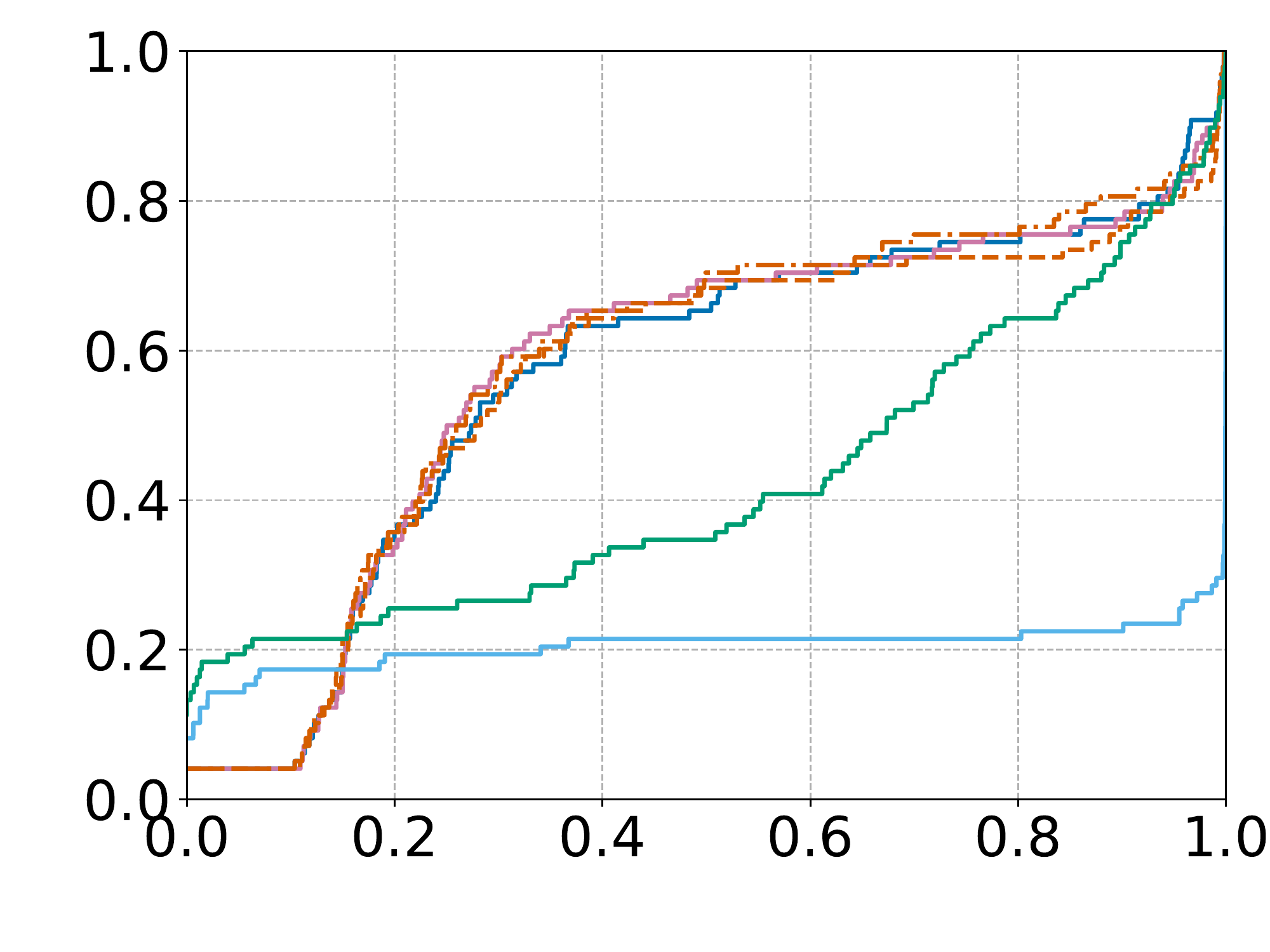}
\\\hline
\begin{tabular}{c}CYC \\ $p=0.5$\end{tabular}
& \includegraphics[width=0.29\textwidth,valign=c,trim={0 0 0 0},clip]{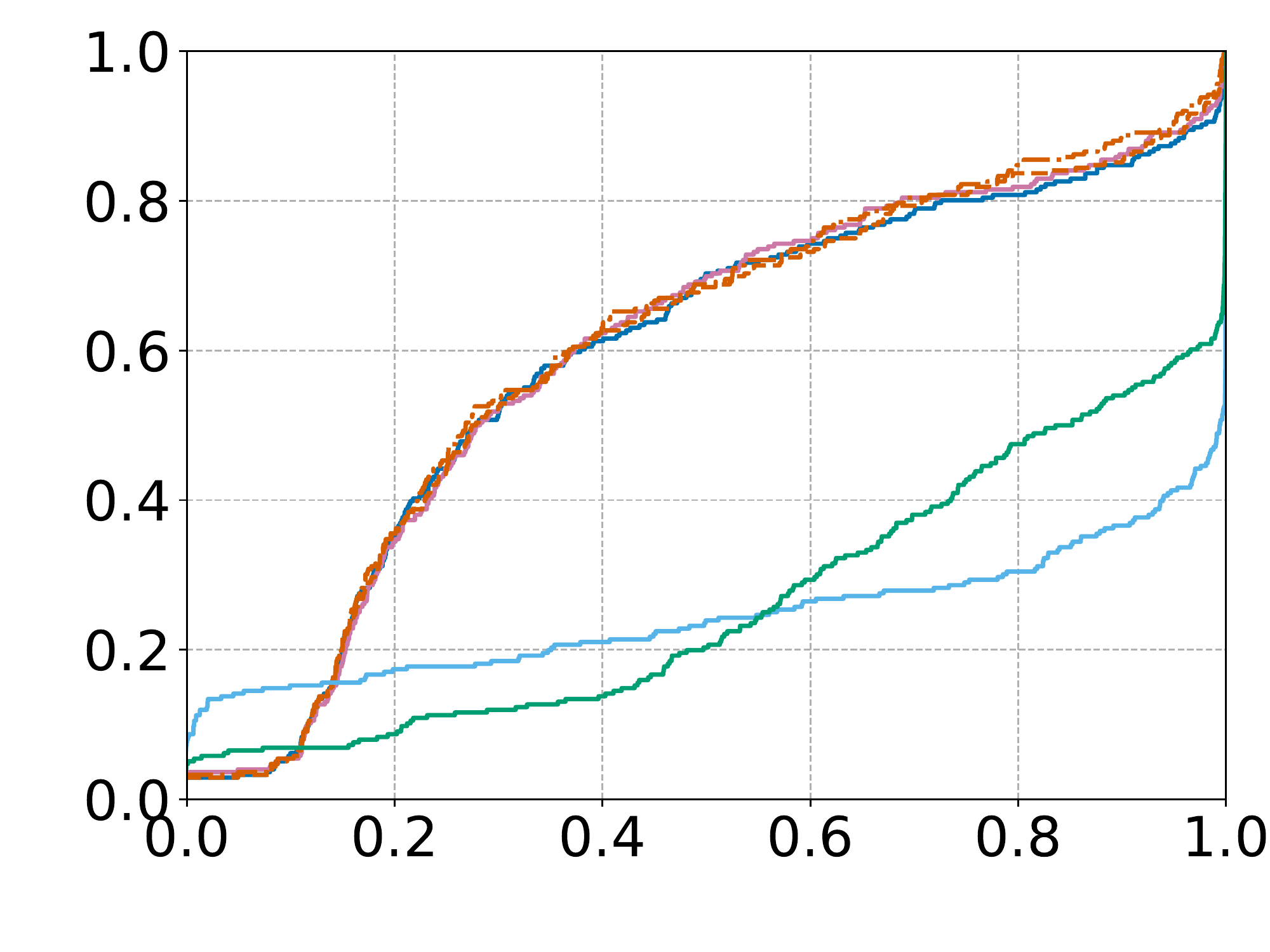}
&
\includegraphics[width=0.29\textwidth,valign=c,trim={0 0 0 0},clip]{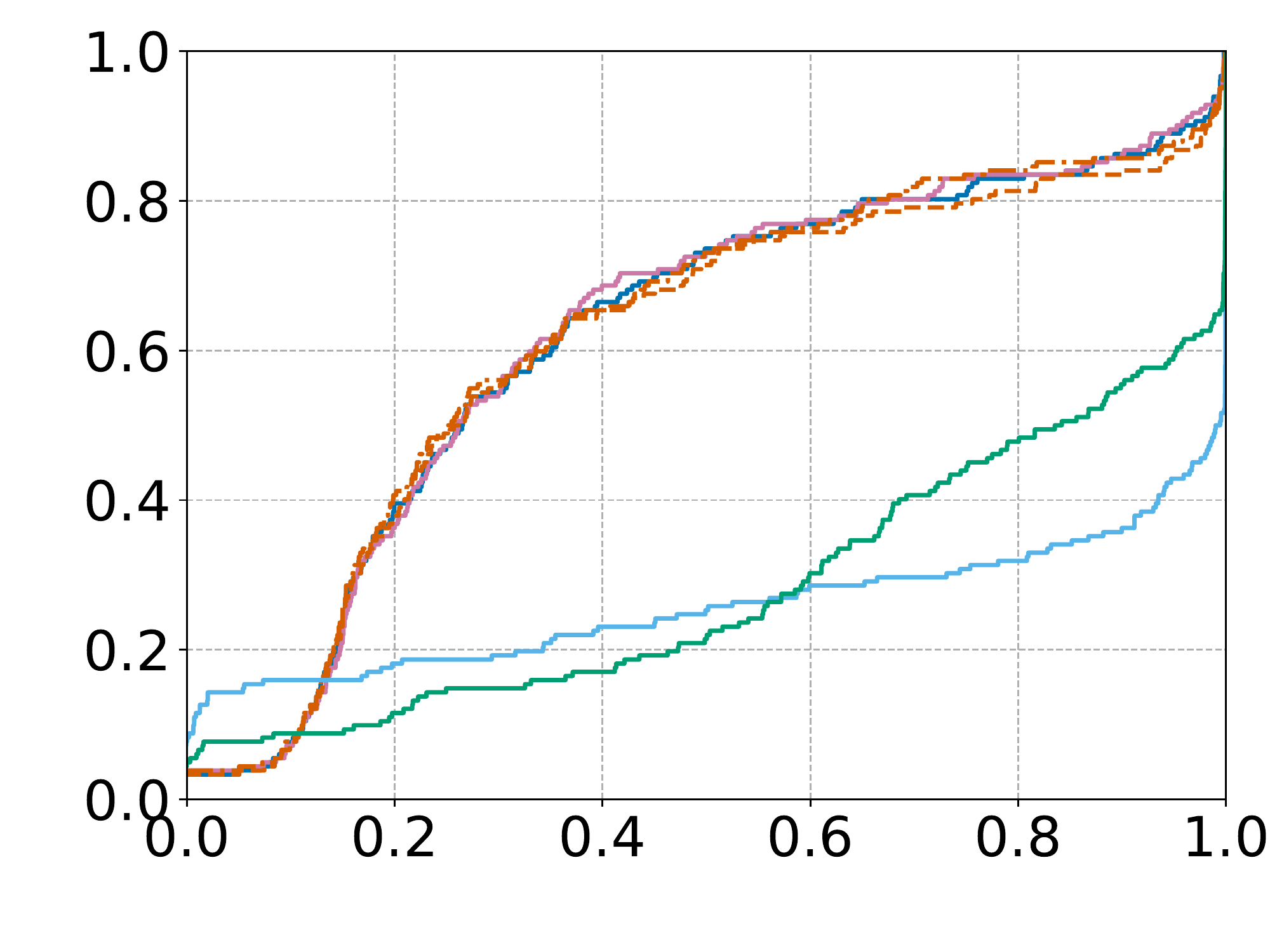}
&
\includegraphics[width=0.29\textwidth,valign=c,trim={0 0 0 0},clip]{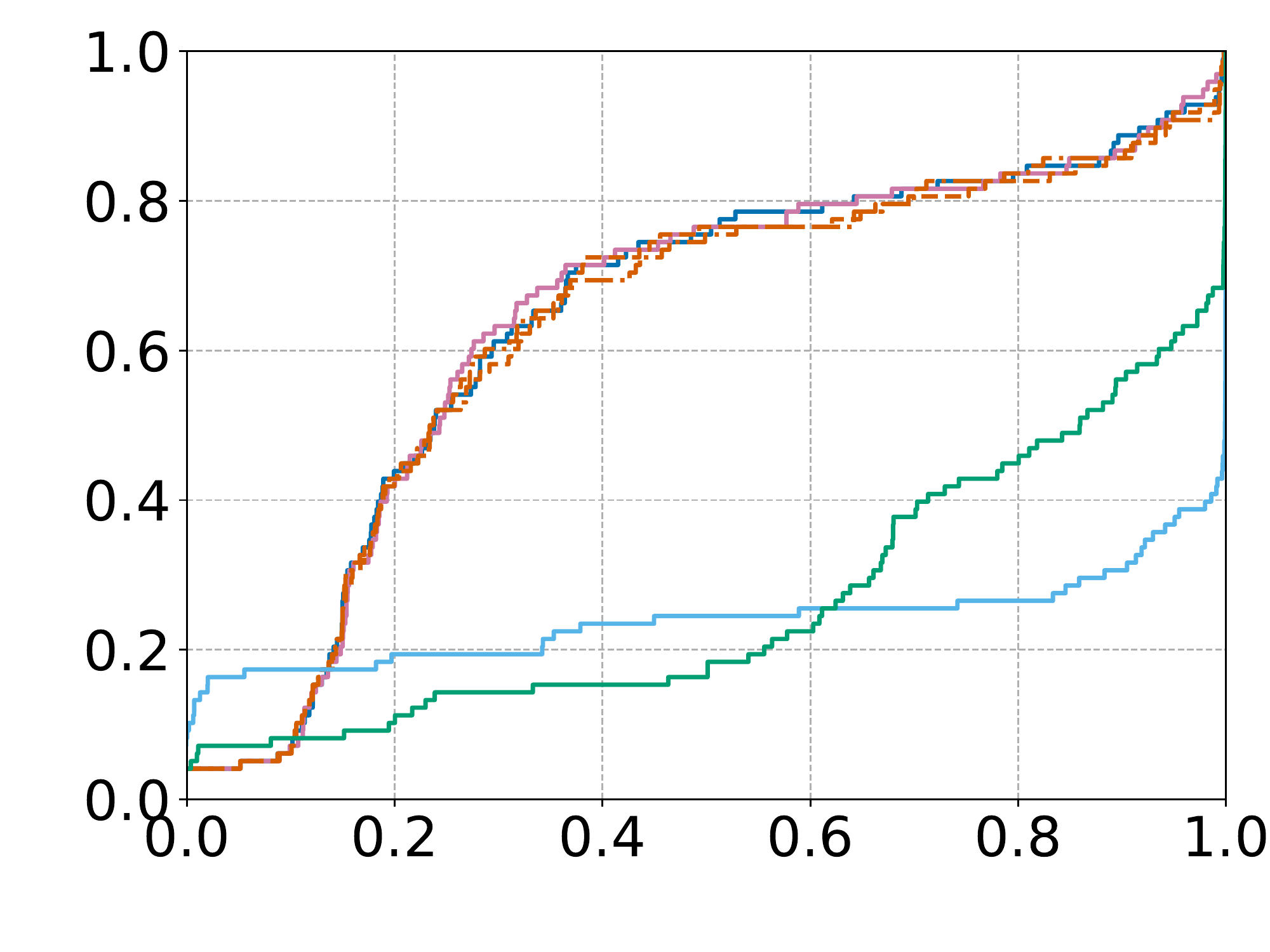}
\\\hline
\begin{tabular}{c}UAR \\ $p=1$\end{tabular}
& \includegraphics[width=0.29\textwidth,valign=c,trim={0 0 0 0},clip]{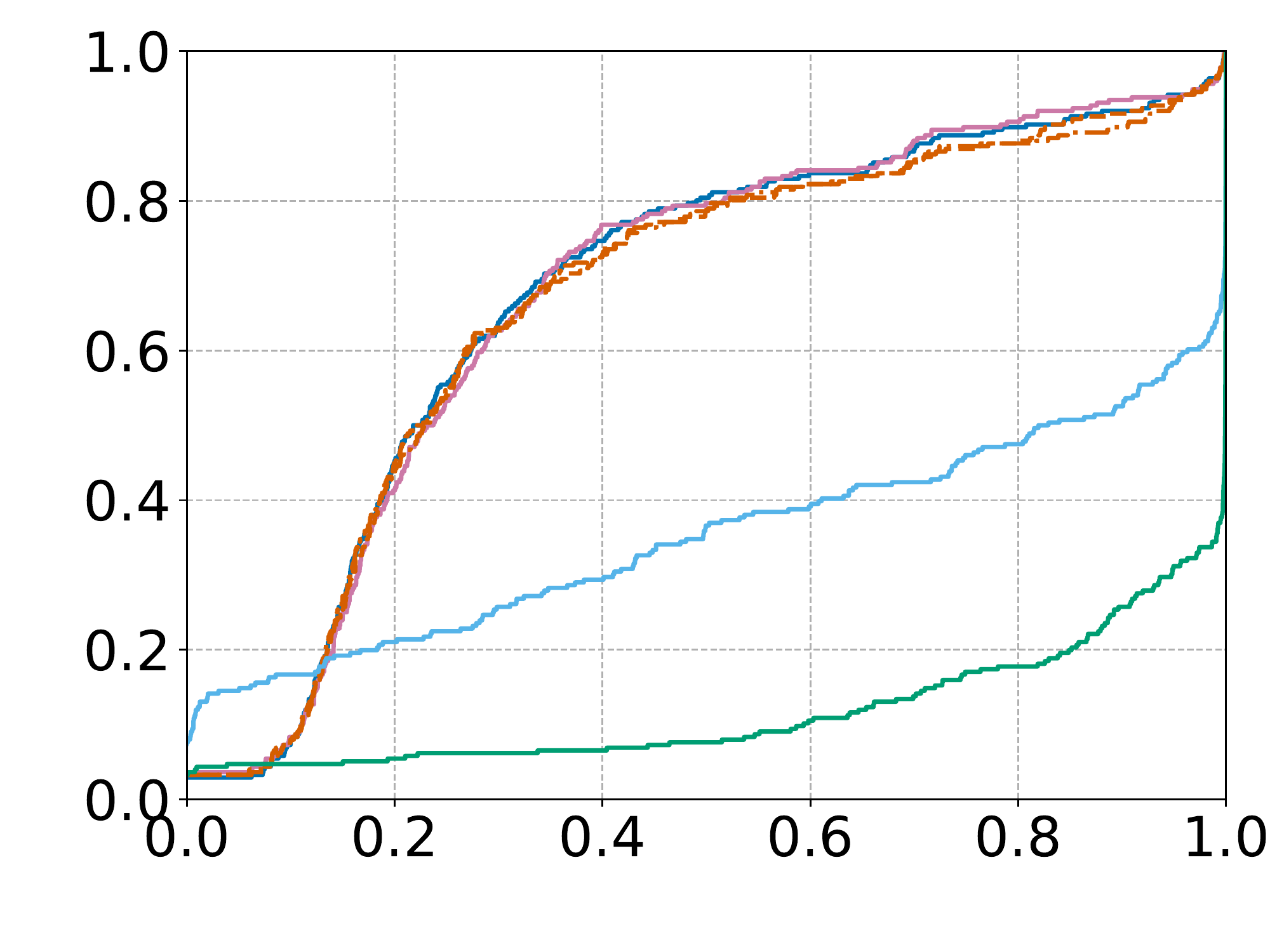}
&
\includegraphics[width=0.29\textwidth,valign=c,trim={0 0 0 0},clip]{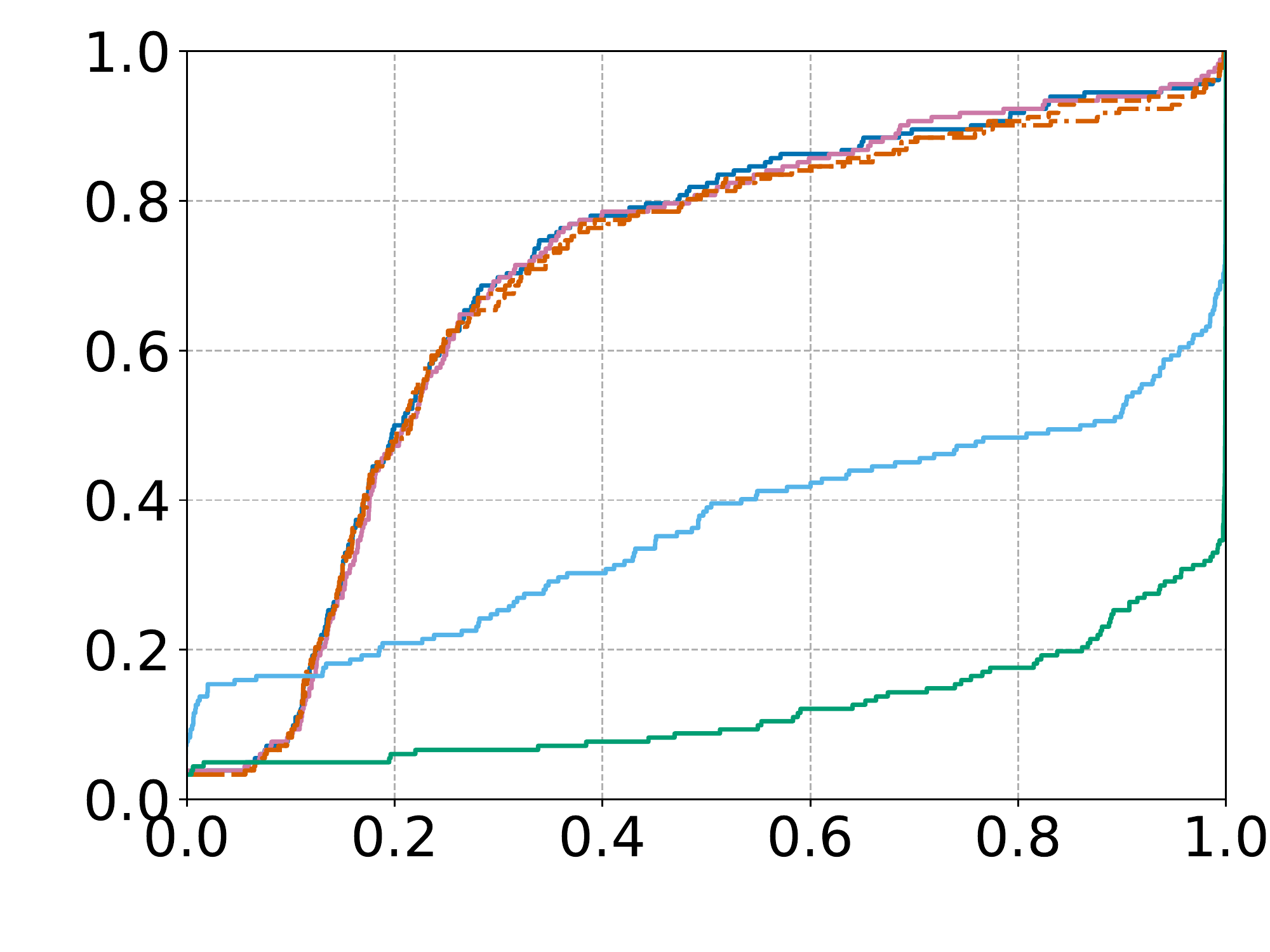}
&
\includegraphics[width=0.29\textwidth,valign=c,trim={0 0 0 0},clip]{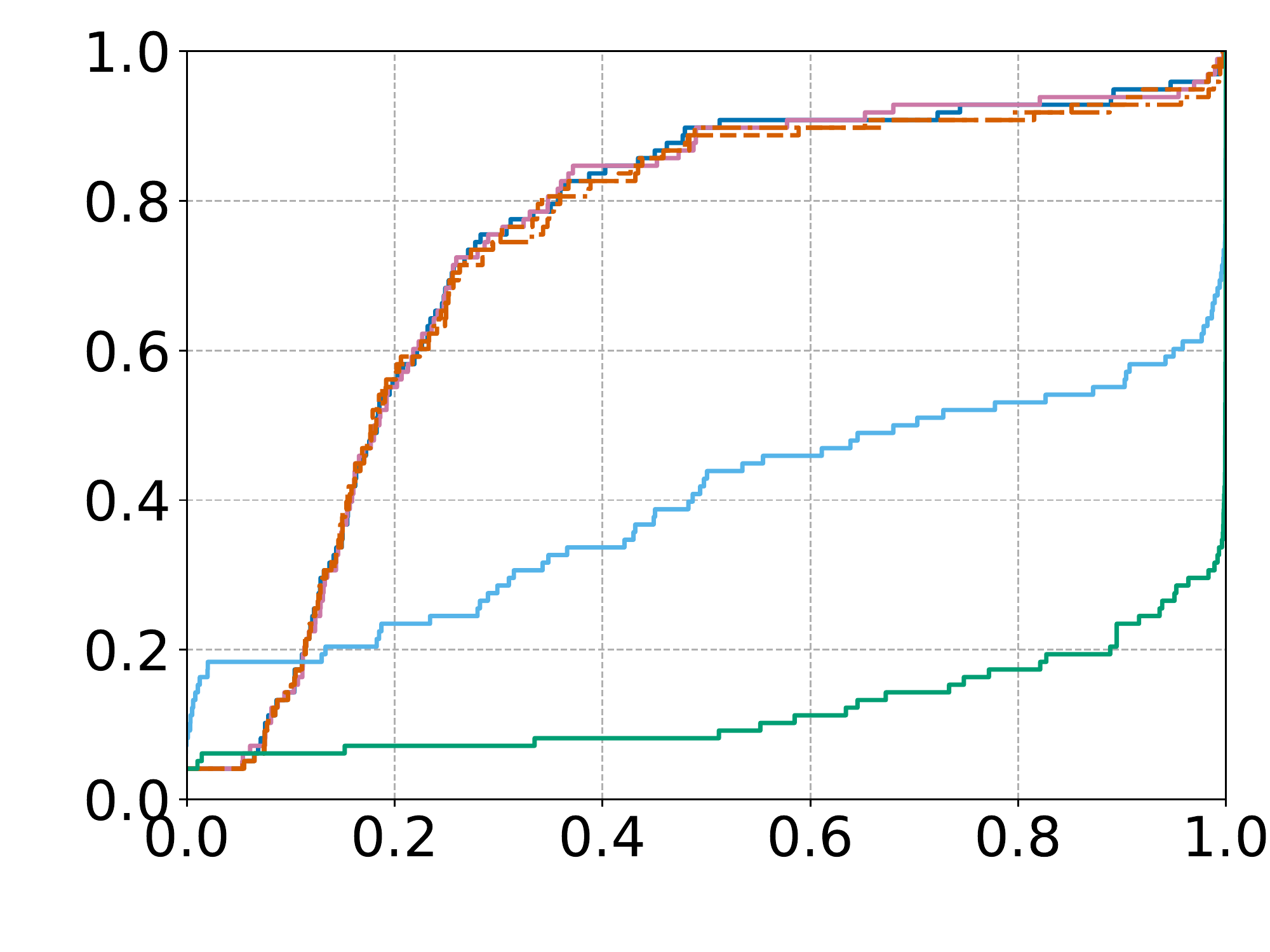}
\\\hline
\begin{tabular}{c}MAJ \\ $p=1$\end{tabular}
& \includegraphics[width=0.29\textwidth,valign=c,trim={0 0 0 0},clip]{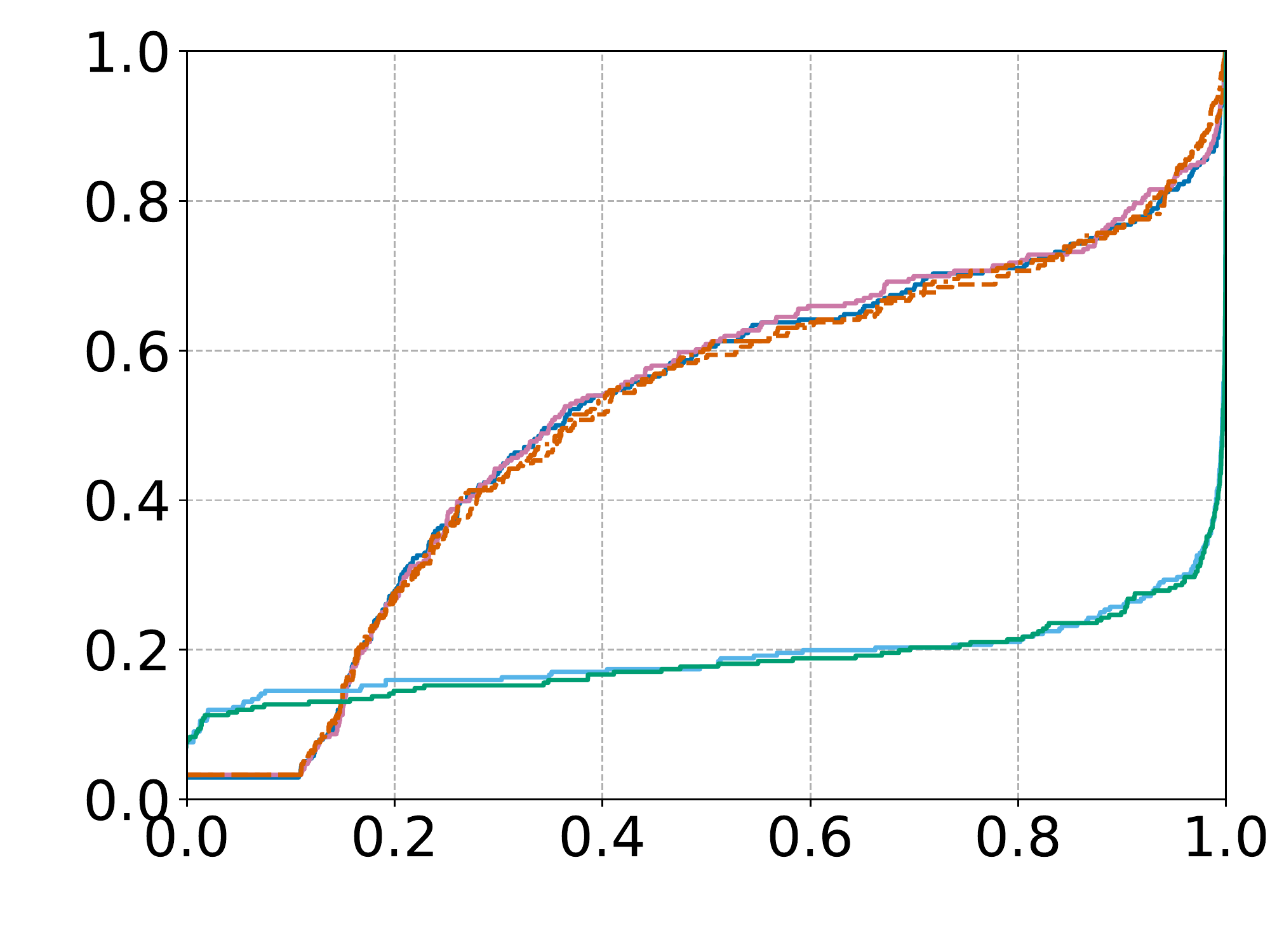}
&
\includegraphics[width=0.29\textwidth,valign=c,trim={0 0 0 0},clip]{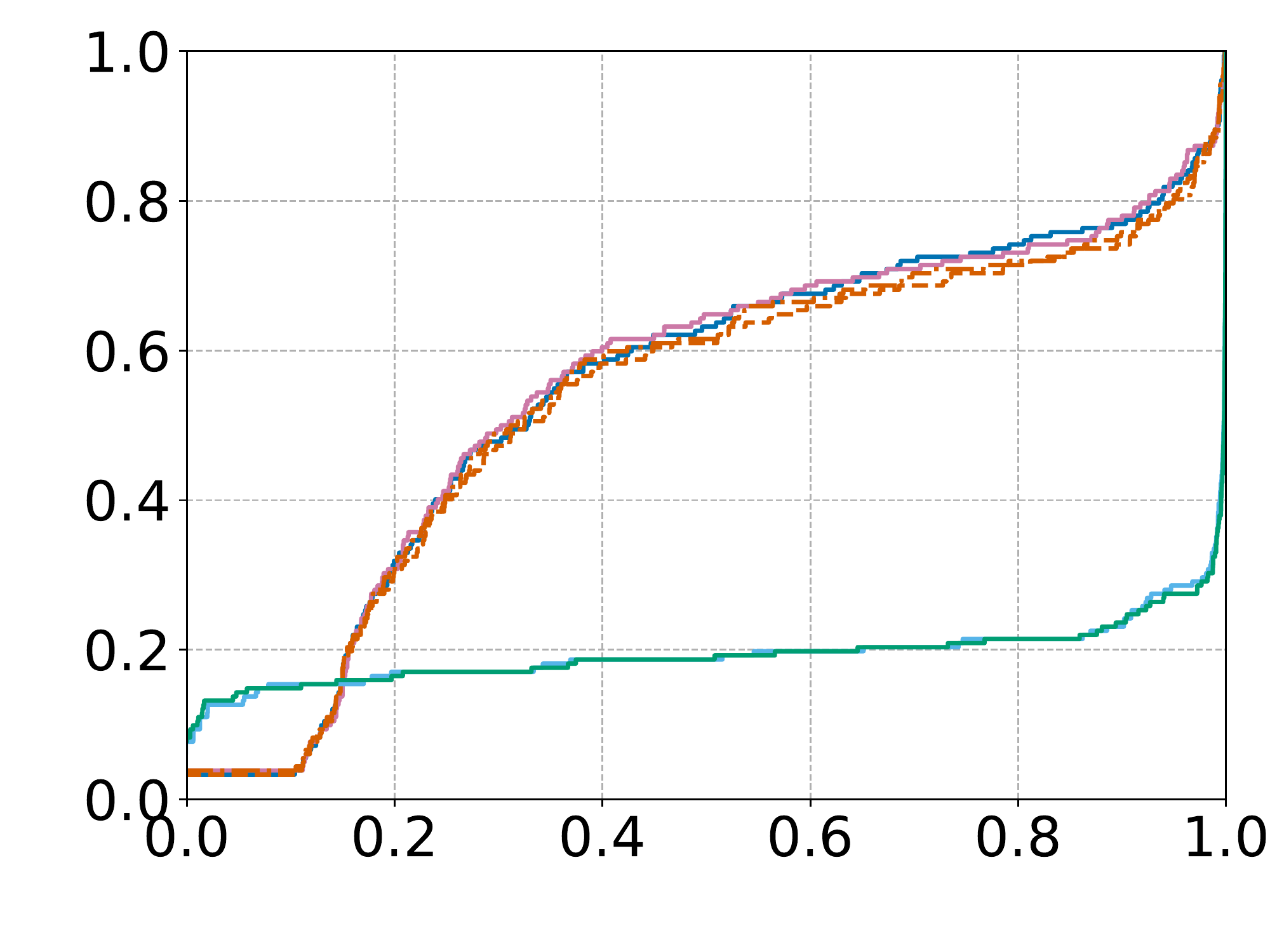}
&
\includegraphics[width=0.29\textwidth,valign=c,trim={0 0 0 0},clip]{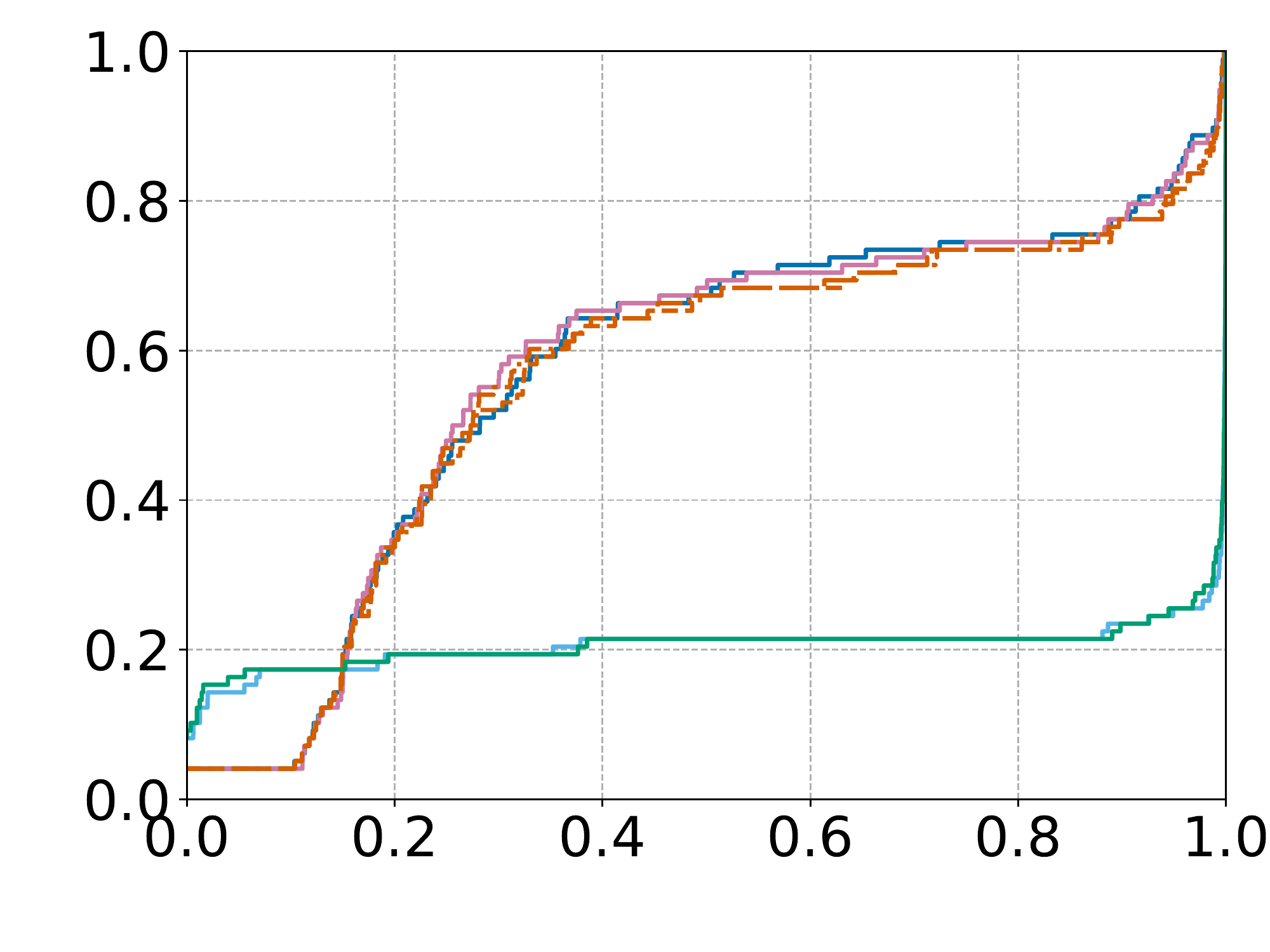}
\\\hline
\begin{tabular}{c}CYC \\ $p=1$\end{tabular}
& \includegraphics[width=0.29\textwidth,valign=c,trim={0 0 0 0},clip]{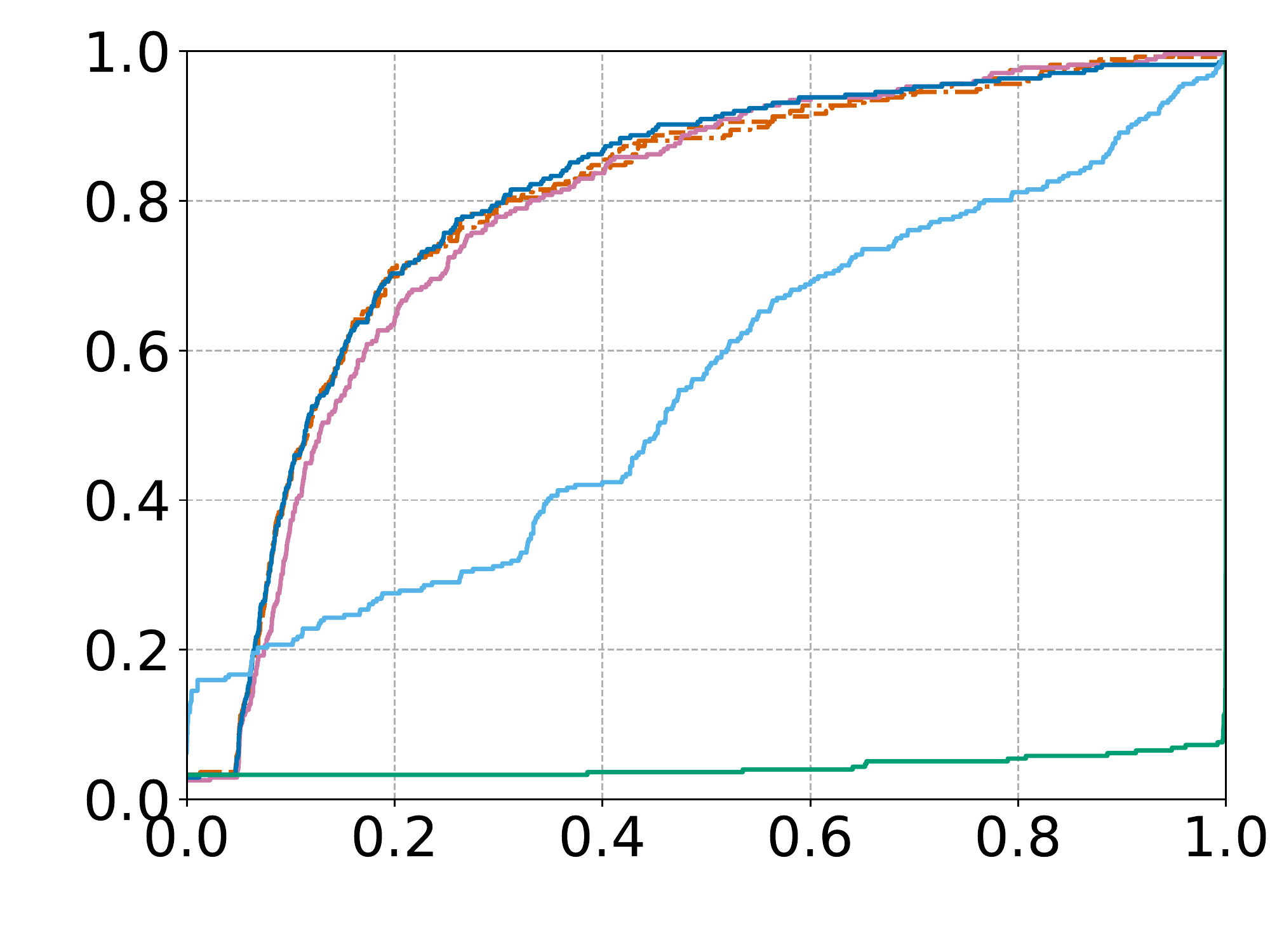}
&
\includegraphics[width=0.29\textwidth,valign=c,trim={0 0 0 0},clip]{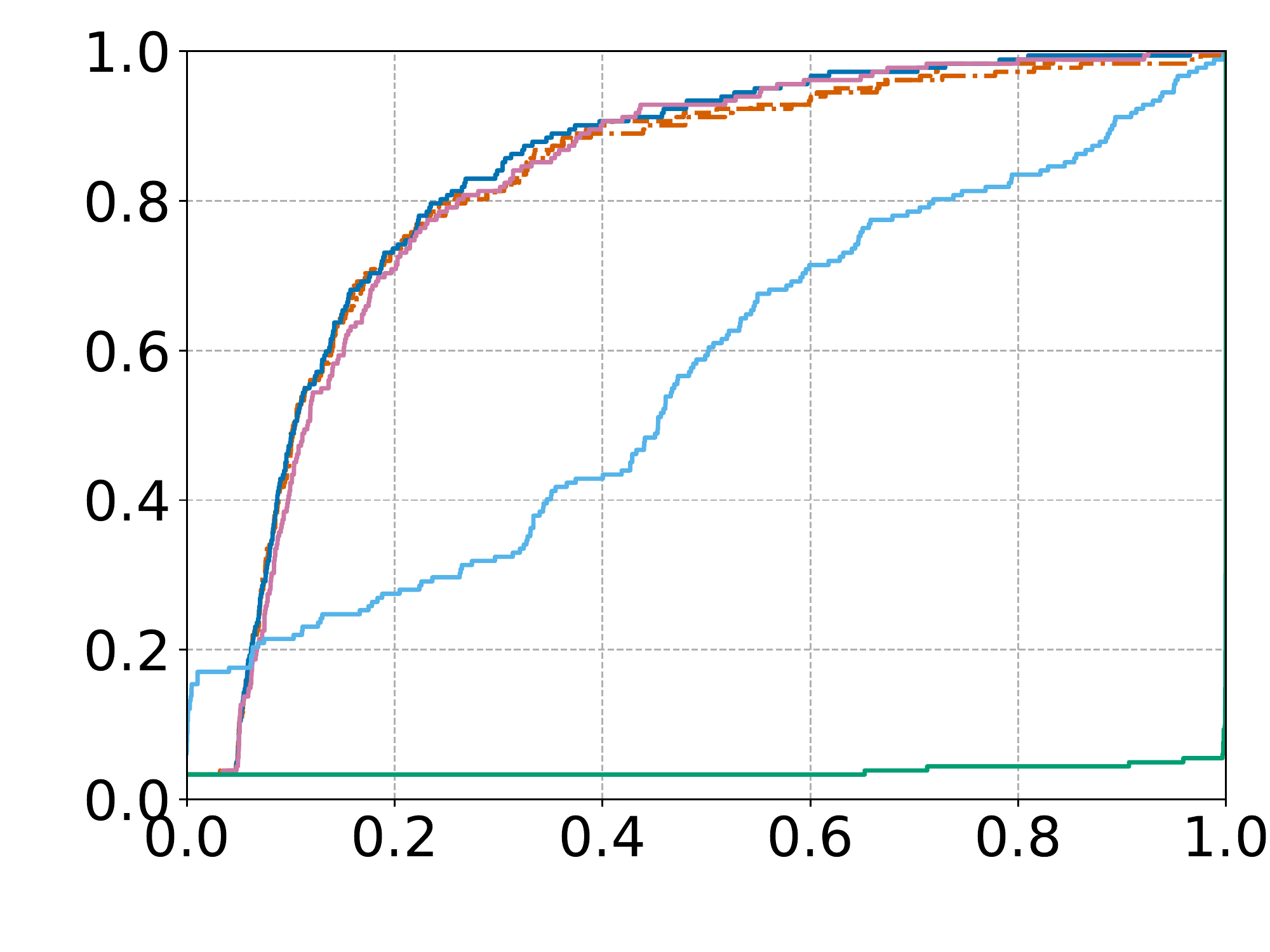}
&
\includegraphics[width=0.29\textwidth,valign=c,trim={0 0 0 0},clip]{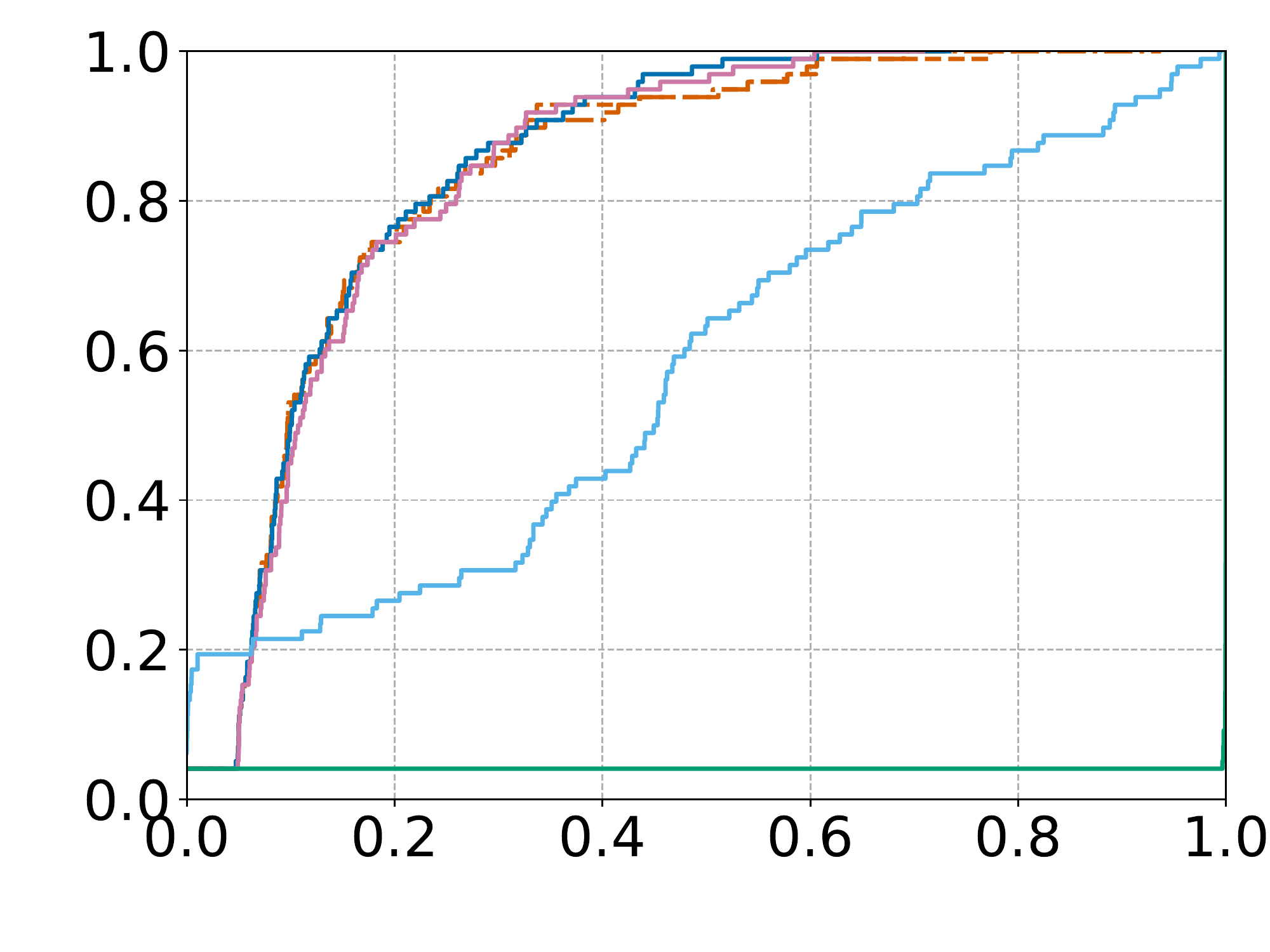}
\\\bottomrule
\multicolumn{4}{c}{
		\begin{subfigure}[t]{\textwidth}
    \includegraphics[width=\textwidth,trim={0 0 0 0},clip]{figs/cdfs/legends/legend_Lambda=8.pdf}
		\end{subfigure} }
\end{tabular}
\caption{Cumulative distribution functions (CDFs) for all evaluated algorithms, against different noise settings and warm start ratios in $\cbr{23.0,46.0,92.0}$. All CB algorithms use $\epsilon$-greedy with $\epsilon=0.1$. In each of the above plots, the $x$ axis represents scores, while the $y$ axis represents the CDF values.}
\label{fig:cdfs-eps=0.1-5}
\end{figure}

We give a collection of cumulative distribution functions (CDFs)
of our algorithms evaluated against different exploration parameters $\epsilon$, noise settings and warm start ratios below:

\begin{enumerate}
\item In Figures~\ref{fig:cdfs-eps=0.0125-1} to~\ref{fig:cdfs-eps=0.0125-6}, we present CDFs where all CB algorithms use $\epsilon$-greedy with parameter $\epsilon = 0.0125$.

\item In Figures~\ref{fig:cdfs-eps=0.00625-1} to~\ref{fig:cdfs-eps=0.00625-6}, we present CDFs where all CB algorithms use $\epsilon$-greedy with parameter $\epsilon = 0.00625$.

\item In Figures~\ref{fig:cdfs-eps=0.1-1} to~\ref{fig:cdfs-eps=0.1-6}, we present CDFs where all CB algorithms use $\epsilon$-greedy with parameter $\epsilon = 0.1$.
\end{enumerate}


The general trends in this more detailed comparison are similar to those observed in Section~\ref{sec:exp}. For less noisy and small warm-start ratios, \textsc{Sup-Only} is particularly difficult as a baseline since it performs no exploration. With extreme noise, \textsc{Bandit-Only} is the best as the supervised examples are misleading. \ourname competes well on both the extremes, while outperforming all methods in several regimes. Importantly, \ourname always beats the other methods that attempt to leverage both sources of data, and prevents the significant performance hit from relying on the wrong data source in either of the two extreme cases.

\newpage

\begin{figure}[H]
\centering
\begin{tabular}{c | @{ }c@{}} 
\toprule
& \multicolumn{1}{c}{ Ratio }
\\
Noise & 184.0
\\\midrule
\begin{tabular}{c}MAJ \\ $p=0.5$\end{tabular}
 & \includegraphics[width=0.29\textwidth,valign=c,trim={0 0 0 0},clip]{figs/cdfs/no_agg/corruption={st,ctws=3,cpws=0.5,cti=1,cpi=0.0},inter_ws_size_ratio={2.875},explore_method={expl,eps=0.0125},/cdf_notitle.pdf}

\\\hline
\begin{tabular}{c}CYC \\ $p=0.5$\end{tabular}
& \includegraphics[width=0.29\textwidth,valign=c,trim={0 0 0 0},clip]{figs/cdfs/no_agg/corruption={st,ctws=2,cpws=0.5,cti=1,cpi=0.0},inter_ws_size_ratio={2.875},explore_method={expl,eps=0.0125},/cdf_notitle.pdf}

\\\hline
\begin{tabular}{c}UAR \\ $p=1$\end{tabular}
& \includegraphics[width=0.29\textwidth,valign=c,trim={0 0 0 0},clip]{figs/cdfs/no_agg/corruption={st,ctws=1,cpws=1.0,cti=1,cpi=0.0},inter_ws_size_ratio={2.875},explore_method={expl,eps=0.0125},/cdf_notitle.pdf}

\\\hline
\begin{tabular}{c}MAJ \\ $p=1$\end{tabular}
& \includegraphics[width=0.29\textwidth,valign=c,trim={0 0 0 0},clip]{figs/cdfs/no_agg/corruption={st,ctws=3,cpws=1.0,cti=1,cpi=0.0},inter_ws_size_ratio={2.875},explore_method={expl,eps=0.0125},/cdf_notitle.pdf}

\\\hline
\begin{tabular}{c}CYC \\ $p=1$\end{tabular}
& \includegraphics[width=0.29\textwidth,valign=c,trim={0 0 0 0},clip]{figs/cdfs/no_agg/corruption={st,ctws=2,cpws=1.0,cti=1,cpi=0.0},inter_ws_size_ratio={2.875},explore_method={expl,eps=0.0125},/cdf_notitle.pdf}

\\\bottomrule
\multicolumn{2}{c}{
		\begin{subfigure}[t]{\textwidth}
    \includegraphics[width=\textwidth,trim={0 0 0 0},clip]{figs/cdfs/legends/legend_Lambda=8.pdf}
		\end{subfigure} }
\end{tabular}
\caption{Cumulative distribution functions (CDFs) for all evaluated algorithms, against different noise settings and warm start ratios in $\cbr{184.0}$. All CB algorithms use $\epsilon$-greedy with $\epsilon=0.0125$. In each of the above plots, the $x$ axis represents scores, while the $y$ axis represents the CDF values.}
\label{fig:cdfs-eps=0.0125-6}
\end{figure}

\begin{figure}[H]
\centering
\begin{tabular}{c | @{ }c@{}} 
\toprule
& \multicolumn{1}{c}{ Ratio }
\\
Noise & 184.0
\\\midrule
\begin{tabular}{c}MAJ \\ $p=0.5$\end{tabular}
 & \includegraphics[width=0.29\textwidth,valign=c,trim={0 0 0 0},clip]{figs/cdfs/no_agg/corruption={st,ctws=3,cpws=0.5,cti=1,cpi=0.0},inter_ws_size_ratio={2.875},explore_method={expl,eps=0.00625},/cdf_notitle.pdf}

\\\hline
\begin{tabular}{c}CYC \\ $p=0.5$\end{tabular}
& \includegraphics[width=0.29\textwidth,valign=c,trim={0 0 0 0},clip]{figs/cdfs/no_agg/corruption={st,ctws=2,cpws=0.5,cti=1,cpi=0.0},inter_ws_size_ratio={2.875},explore_method={expl,eps=0.00625},/cdf_notitle.pdf}

\\\hline
\begin{tabular}{c}UAR \\ $p=1$\end{tabular}
& \includegraphics[width=0.29\textwidth,valign=c,trim={0 0 0 0},clip]{figs/cdfs/no_agg/corruption={st,ctws=1,cpws=1.0,cti=1,cpi=0.0},inter_ws_size_ratio={2.875},explore_method={expl,eps=0.00625},/cdf_notitle.pdf}

\\\hline
\begin{tabular}{c}MAJ \\ $p=1$\end{tabular}
& \includegraphics[width=0.29\textwidth,valign=c,trim={0 0 0 0},clip]{figs/cdfs/no_agg/corruption={st,ctws=3,cpws=1.0,cti=1,cpi=0.0},inter_ws_size_ratio={2.875},explore_method={expl,eps=0.00625},/cdf_notitle.pdf}

\\\hline
\begin{tabular}{c}CYC \\ $p=1$\end{tabular}
& \includegraphics[width=0.29\textwidth,valign=c,trim={0 0 0 0},clip]{figs/cdfs/no_agg/corruption={st,ctws=2,cpws=1.0,cti=1,cpi=0.0},inter_ws_size_ratio={2.875},explore_method={expl,eps=0.00625},/cdf_notitle.pdf}

\\\bottomrule
\multicolumn{2}{c}{
		\begin{subfigure}[t]{\textwidth}
    \includegraphics[width=\textwidth,trim={0 0 0 0},clip]{figs/cdfs/legends/legend_Lambda=8.pdf}
		\end{subfigure} }
\end{tabular}
\caption{Cumulative distribution functions (CDFs) for all evaluated algorithms, against different noise settings and warm start ratios in $\cbr{184.0}$. All CB algorithms use $\epsilon$-greedy with $\epsilon=0.00625$. In each of the above plots, the $x$ axis represents scores, while the $y$ axis represents the CDF values.}
\label{fig:cdfs-eps=0.00625-6}
\end{figure}

\begin{figure}[H]
\centering
\begin{tabular}{c | @{ }c@{}} 
\toprule
& \multicolumn{1}{c}{ Ratio }
\\
Noise & 184.0
\\\midrule
\begin{tabular}{c}MAJ \\ $p=0.5$\end{tabular}
 & \includegraphics[width=0.29\textwidth,valign=c,trim={0 0 0 0},clip]{figs/cdfs/no_agg/corruption={st,ctws=3,cpws=0.5,cti=1,cpi=0.0},inter_ws_size_ratio={2.875},explore_method={expl,eps=0.1},/cdf_notitle.pdf}

\\\hline
\begin{tabular}{c}CYC \\ $p=0.5$\end{tabular}
& \includegraphics[width=0.29\textwidth,valign=c,trim={0 0 0 0},clip]{figs/cdfs/no_agg/corruption={st,ctws=2,cpws=0.5,cti=1,cpi=0.0},inter_ws_size_ratio={2.875},explore_method={expl,eps=0.1},/cdf_notitle.pdf}

\\\hline
\begin{tabular}{c}UAR \\ $p=1$\end{tabular}
& \includegraphics[width=0.29\textwidth,valign=c,trim={0 0 0 0},clip]{figs/cdfs/no_agg/corruption={st,ctws=1,cpws=1.0,cti=1,cpi=0.0},inter_ws_size_ratio={2.875},explore_method={expl,eps=0.1},/cdf_notitle.pdf}

\\\hline
\begin{tabular}{c}MAJ \\ $p=1$\end{tabular}
& \includegraphics[width=0.29\textwidth,valign=c,trim={0 0 0 0},clip]{figs/cdfs/no_agg/corruption={st,ctws=3,cpws=1.0,cti=1,cpi=0.0},inter_ws_size_ratio={2.875},explore_method={expl,eps=0.1},/cdf_notitle.pdf}

\\\hline
\begin{tabular}{c}CYC \\ $p=1$\end{tabular}
& \includegraphics[width=0.29\textwidth,valign=c,trim={0 0 0 0},clip]{figs/cdfs/no_agg/corruption={st,ctws=2,cpws=1.0,cti=1,cpi=0.0},inter_ws_size_ratio={2.875},explore_method={expl,eps=0.1},/cdf_notitle.pdf}

\\\bottomrule
\multicolumn{2}{c}{
		\begin{subfigure}[t]{\textwidth}
    \includegraphics[width=\textwidth,trim={0 0 0 0},clip]{figs/cdfs/legends/legend_Lambda=8.pdf}
		\end{subfigure} }
\end{tabular}
\caption{Cumulative distribution functions (CDFs) for all evaluated algorithms, against different noise settings and warm start ratios in $\cbr{184.0}$. All CB algorithms use $\epsilon$-greedy with $\epsilon=0.1$. In each of the above plots, the $x$ axis represents scores, while the $y$ axis represents the CDF values.}
\label{fig:cdfs-eps=0.1-6}
\end{figure}

\end{document}